\documentclass[11pt]{article}

\usepackage{PRIMEarxiv}
\usepackage{longtable}
\usepackage[utf8]{inputenc} %
\usepackage[T1]{fontenc}    %
\usepackage{url}            %
\usepackage{booktabs}       %
\usepackage{amsfonts}       %
\usepackage{nicefrac}       %
\usepackage{microtype}      %
\usepackage{lipsum}
\usepackage{bm}
\usepackage{fancyhdr}       %
\usepackage{graphicx}       %
\graphicspath{{media/}}     %
\usepackage{etoolbox}       %
\usepackage{makecell}  %
\usepackage{times}
\usepackage{soul}
\usepackage{url}
\usepackage[hidelinks]{hyperref}
\usepackage[table]{xcolor}
\usepackage{booktabs}
\usepackage[utf8]{inputenc}
\usepackage[small]{caption}
\usepackage{graphicx}
\usepackage{amsmath}    
\usepackage{amssymb}
\usepackage{amsthm}
\usepackage{wasysym}
\usepackage{booktabs}
\usepackage{algorithm}
\usepackage{algorithmic}
\usepackage[switch]{lineno}
\usepackage{xcolor}
\usepackage{listings}
\usepackage{tikz}
\usepackage{xparse}
\usepackage{enumitem}
\usepackage{fontawesome5}
\definecolor{mygreen}{RGB}{30, 128, 20}
\colorlet{shadecolor}{gray!20}
\usepackage{arydshln}
\usepackage{booktabs}
\usepackage{natbib}

\tikzset{chatstyle/.style={text width=2.8in,rounded corners=2pt}}

\definecolor{mygreen}{HTML}{88EABB}
\definecolor{OliveGreen}{HTML}{00693E}

\usepackage{wrapfig}
\usepackage{adjustbox}
\usepackage{xspace}
\usepackage{listings}
\usepackage{subcaption}
\usepackage[most]{tcolorbox}

\usepackage{pifont}

\usepackage{multicol}
\usepackage{multirow}
\usepackage{bbding}
\usepackage{circledsteps}

\usepackage{fancyhdr}
\usetikzlibrary{mindmap,shadows}

\definecolor{LightCyan}{RGB}{232,241,255}
\definecolor{LightRed}{RGB}{255,235,235}
\definecolor{LightPink}{RGB}{255,235,255}
\definecolor{LightGreen}{RGB}{218,255,234}
\definecolor{LightYellow}{RGB}{255,255,235}
\definecolor{LightGray}{RGB}{242,242,242}
\definecolor{Red}{RGB}{253, 239, 242}
\definecolor{Yellow}{RGB}{255, 255, 204}
\definecolor{Pink}{RGB}{255, 243, 254}
\definecolor{Gray}{RGB}{249, 249, 249}
\definecolor{Green}{RGB}{230, 255, 241}
\definecolor{Blue1}{RGB}{218, 232, 245}
\definecolor{Blue2}{RGB}{239, 248, 253}
\definecolor{Blue3}{RGB}{136, 190, 220}
\definecolor{Blue4}{RGB}{83, 157, 204}
\definecolor{Blue5}{RGB}{42, 122, 185}
\definecolor{Blue6}{RGB}{11, 85, 159}
\definecolor{GreenCheck}{RGB}{0, 102, 51}
\definecolor{LightBack}{RGB}{247,249,251}
\tcbset{
    colback=LightBack,colframe=black,boxsep=3pt, arc=3pt,boxrule=1pt,
    width=\linewidth,enhanced,frame hidden, overlay={\draw[line width=1pt] (frame.north west)--([xshift=\linewidth]frame.north west);\draw[line width=1pt] (frame.south west)--([xshift=\linewidth]frame.south west);}
}

\newcommand{\squishlist}{
\begin{list}{{{\small{$\bullet$}}}}
{\setlength{\itemsep}{1pt}      \setlength{\parsep}{5pt}
\setlength{\topsep}{-2pt}       \setlength{\partopsep}{0pt}
\setlength{\leftmargin}{2.5em} \setlength{\labelwidth}{1em}
\setlength{\labelsep}{1em} } }

\newcommand{\squishend}{  \end{list}  }
\definecolor{aigold}{RGB}{244,210, 1} 
\definecolor{aigreen}{RGB}{210,244,211} 

\sethlcolor{aigreen}
\definecolor{aired}{RGB}{255,180,181} 
\definecolor{lighterseafoam}{RGB}{194,218,184}

\newcolumntype{L}[1]{>{\raggedright\arraybackslash}p{#1}}
\newcolumntype{C}[1]{>{\centering\arraybackslash}p{#1}}
\newcolumntype{R}[1]{>{\raggedleft\arraybackslash}p{#1}}
\usepackage{datetime}
\usepackage{CJKutf8}
\title{LLMs4All: A Review of Large Language Models\\ Across Academic Disciplines}

\author{
{\bfseries Yanfang (Fanny) Ye\footnotemark[1]~~\footnotemark[2]~~\footnotemark[3]}\quad
{\bfseries Zheyuan Zhang\footnotemark[2]}\quad
{\bfseries Tianyi Ma\footnotemark[2]}\quad
{\bfseries Zehong Wang\footnotemark[2]}\quad 
{\bfseries Yiyang Li\footnotemark[2]}\quad \\
{\bfseries Shifu Hou\footnotemark[2]}\quad 
{\bfseries Weixiang Sun\footnotemark[2]}\quad 
{\bfseries Kaiwen Shi\footnotemark[2]}\quad
{\bfseries Yijun Ma\footnotemark[2]}\quad
{\bfseries Wei Song}\quad
{\bfseries Ahmed Abbasi}\quad\\
{\bfseries Ying Cheng}\quad
{\bfseries Jane Cleland-Huang}\quad
{\bfseries Steven Corcelli}\quad 
{\bfseries Robert Goulding}\quad \\
{\bfseries Ming Hu}\quad
{\bfseries Ting Hua}\quad 
{\bfseries John Lalor}\quad
{\bfseries Fang Liu}\quad
{\bfseries Tengfei Luo}\quad
{\bfseries Edward Maginn}\quad\\
{\bfseries Nuno Moniz}\quad
{\bfseries Jason Rohr}\quad
{\bfseries Brett Savoie}\quad
{\bfseries Daniel Slate}\quad
{\bfseries Matthew  Webber}\quad\\
{\bfseries Olaf Wiest}\quad
{\bfseries Johnny Zhang}\quad
{\bfseries Nitesh V Chawla\footnotemark[3]}\quad
\\
\\
{\bfseries University of Notre Dame}\quad
}

\begin{document}
\maketitle
\renewcommand{\thefootnote}{\fnsymbol{footnote}}
\footnotetext[1]{Lead Author.}
\footnotetext[2]{Major Contributions.}
\footnotetext[3]{Corresponding Authors: \href{mailto:yye7@nd.edu}{yye7@nd.edu}, \href{mailto:nchawla@nd.edu }{nchawla@nd.edu}.}
\footnotetext[4]{Latest Update: November 22, 2025.}

\begin{abstract}

Cutting-edge Artificial Intelligence (AI) techniques keep reshaping our view of the world. For example, Large Language Models (LLMs) based applications such as ChatGPT have shown the capability of generating human-like conversation on extensive topics. Due to the impressive performance on a variety of language-related tasks (e.g., open-domain question answering, translation, and document summarization), one can envision the far-reaching impacts that can be brought by the LLMs with broader real-world applications (e.g., customer service, education and accessibility, and scientific discovery). Inspired by their success, this paper offers an overview of state-of-the-art LLMs and their integration into a wide range of academic disciplines, including: (1) arts, letters, and law (e.g., history, philosophy, political science, arts and architecture, law), (2) economics and business (e.g., finance, economics, accounting, marketing), and (3) science and engineering (e.g., mathematics, physics and mechanical engineering, chemistry and chemical engineering, life sciences and bioengineering, earth sciences and civil engineering, computer science and electrical engineering). Integrating humanity and technology, in this paper, we explore how LLMs are shaping research and practice in these fields, while also discussing key limitations, open challenges, and future directions in the era of generative AI. The review of how LLMs are engaged across disciplines, along with key observations and insights, can help researchers and practitioners interested in exploiting LLMs to advance their works in diverse real-world applications.

\end{abstract}

\newpage
\tableofcontents
\newpage 
\begin{CJK*}{UTF8}{gbsn}
\section{Introduction}

Nowadays, cutting-edge technologies in Artificial Intelligence (AI) keep reshaping our view of the world. For example, as a foundation language model based on the Generative Pre-trained Transformer (GPT) architecture, ChatGPT \cite{achiam2023gpt} has shown its capability of generating human-like conversation on extensive topics, which makes it the fastest-growing application (i.e., with more than 100 million users within the first two months of its launch) \cite{mesko2023chatgpt}. Although limitations such as robustness and truthfulness it still remains, due to the impressive performance on a variety of language-related tasks (e.g., open-domain question answering, translation, and document summarization), ChatGPT could have a wide range of potential applications (e.g., customer service, personal assistants, and medical diagnosis). Besides the models like ChatGPT in Natural Language Procession (NLP), the pre-trained foundation models in Computer Vision (CV) such as Florence/Florence-2~\cite{yuan2021florence} and Qwen2.5-VL can achieve state-of-the-art performance on various vision-tasks (e.g., object detection, image segmentation, and video reasoning), which make them particularly useful for applications such as facial recognition, medical image analysis, and self-driving cars. This cross-domain convergence underscores the pivotal role of large language models (LLMs), which provide the representational and reasoning layer for embedding other modalities, positioning them as central components in the evolving ecosystem of AI-powered research and applications.

Motivated by recent advances, this paper surveys cutting-edge LLMs and their integration into a wide range of academic disciplines, including: (1) arts, letters, and law (history, philosophy, political science, arts and architecture, law), (2) economics and business (finance, economics, accounting, marketing), and (3) science and engineering (mathematics, physics and mechanical engineering, chemistry and chemical engineering, life sciences and bioengineering, earth sciences and civil engineering, computer science and electrical engineering). At the intersection of humanity and technology, we examine how LLMs may reshape research workflows and professional practice in each area, while also outlining major limitations, unresolved challenges, and promising directions in the era of generative AI. By synthesizing cross-disciplinary uses and distilling key takeaways, this review is intended to guide researchers and practitioners seeking to harness LLMs to advance their work in real-world applications. In the following, we outline the organization of the paper.

Building on recent breakthroughs, in \textbf{Chapter 2}, we ground the reader in what LLMs are and how to assess them. We begin with precise definitions and a concise history of LLMs. We then map the state of the art with an overview and focused profiles of major model families—GPT-series, OpenAI reasoning models, Claude 3, Gemini 2, Gork, Llama 3, Qwen 2, and DeepSeek—highlighting design choices and capabilities. We conclude with evaluation: the core task types, representative benchmarks, and commonly used methods, followed by a performance-at-a-glance synthesis. Together, these sections aim to provide a background, a comparative map of current models, and practical guidance for reading results and making methodologically sound choices.

For each academic discipline within the three clusters—arts, letters, and law; economics and business; and science and engineering—we begin by introducing the discipline through an overview of its major research tasks and traditional methodologies, with the highlight of its key contributions and significant impact. After identifying common research challenges that could be assisted by AI (particularly LLMs), we integrate disciplinary research with LLMs by providing a taxonomy that aligns with established disciplinary tasks while mapping them onto a unified computational input–output framework. This ensures both disciplinary relevance and algorithmic consistency for model development, benchmarking, and comparative analysis. Within each category, we review existing works on LLM-powered research and applications, examine current limitations, and explore future research directions. Finally, we conclude with representative benchmarks and critical discussions.

In \textbf{Chapter 3}, we survey how LLMs are transforming the humanities and law, moving from evidence to practice. In \textbf{history}, we cover narrative and interpretive uses (e.g., narrative generation and analysis), quantitative and scientific approaches (e.g., simulating historical psychological responses), and comparative and cross-disciplinary work, with benchmarks and a brief discussion. In \textbf{philosophy}, we review normative and interpretive applications (e.g., debate/dialogue generation), analytical and logical ones (e.g., symbol grounding diagnostics), and comparative and cross-disciplinary studies with benchmarks. In \textbf{political science}, we examine text analysis for policy insights, opinion simulation and forecasting, and the generation and framing of political messaging, integrating these with benchmark summaries and reflections. In \textbf{arts and architecture}, we outline model-assisted creation in visual, literary, and performing arts, as well as LLM-aided architectural design, creation, and analysis, followed by evaluations and takeaways. Finally, in \textbf{law}, we cover legal consultant question answering, contract and brief drafting, legal document understanding and case analysis, and judgment prediction, concluding with representative benchmarks and discussion.

In \textbf{Chapter 4}, we review how LLMs are being used across economics and business. In \textbf{finance}, we survey LLMs for trading and investment research, corporate finance, market analysis, financial intermediation and risk management, sustainable finance, and fintech, as well as how these systems are benchmarked. In \textbf{economics}, we cover behavioral and experimental studies, macroeconomic simulation and agent-based modeling, strategic and game-theoretic interactions, and economic reasoning/knowledge representation, with dedicated evaluations. In \textbf{accounting} section, we examine auditing, financial and managerial accounting, and taxation, alongside benchmarking. In \textbf{marketing} section, we review consumer insight and behavior analysis, content creation and campaign design, and market-intelligence and trend analysis, again with performance benchmarks. 

In \textbf{Chapter 5}, we chart how LLMs are used across science and engineering. We start with \textbf{mathematics}, including proof assistance, theoretical exploration and pattern recognition, math education, and targeted benchmarks. In \textbf{physics and mechanical engineering}, we cover documentation-centric tasks, design ideation and parametric drafting, simulation-support and modeling interfaces, multimodal lab and experiment interpretation, as well as interactive reasoning, followed by domain-specific evaluations and a discussion of opportunities and limits. In \textbf{chemistry and chemical engineering}, we examine molecular structure and reaction reasoning, property prediction, materials optimization, test/assay mapping, property-oriented molecular design, and reaction-data knowledge organization, followed by a comparison of benchmark suites. In \textbf{life sciences and bioengineering} section, we include genomic sequence analysis, clinical structured-data integration, biomedical reasoning and understanding, and hybrid outcome prediction, with attention to validation standards. In \textbf{earth sciences and civil engineering} section, we review geospatial and environmental data tasks, simulation and physical modeling, document workflows, monitoring and predictive maintenance, plus design/planning tasks, again with benchmarks. Finally, we close the chapter with \textbf{computer science and electrical engineering}: code generation and debugging, large-codebase analysis, hardware description language code generation, functional verification, and high-level synthesis, followed by purpose-built benchmarks and a concluding discussion of impacts and open challenges.

In \textbf{Chapter 6}, we conclude “Navigating the Present, Shaping the Future” by synthesizing what we learn across domains. We first outline emerging frontiers, then synthesize evidence across three arenas: (i) arts, letters, and law—shared opportunities, common limitations, and future paradigms from historical analysis to legal reasoning; (ii) economics and business—signals from finance, accounting, economics, and marketing translated into strategy with concrete paradigms; and (iii) science and engineering—models as instruments, with discipline probes yielding cross-cutting opportunities, constraints, and workflow-ready paradigms. We conclude with a path forward that integrates schema-aligned multimodality and grounded attribution; tool-augmented computation under formal constraints; rule-governed, reproducible agent simulation; temporal-causal adaptation; decision support with calibrated uncertainty and domain controls; human-in-the-loop oversight and transparent governance; and education-led capacity building with embedded safety—providing a practical, auditable, and scalable blueprint for cross-disciplinary adoption.

Altogether, in this paper, we chart the LLM landscape end-to-end—foundations and evaluation, then concrete uses across arts, letters, and law, economics and business, and science and engineering—showing what works today, where capabilities remain fragile, and how to measure progress. Readers can take away a common vocabulary and task taxonomy; guidance for selecting models and tools; recipes for building rigorous evaluations and benchmarks; and practical patterns for deployment that balance utility with safety, compliance, and human oversight. The content of this paper may not be exhaustive, and certain perspectives presented herein may be open to debate; furthermore, with the rapid advancement and continuous evolution of technology especially in the field of AI, the disciplines reviewed in this study are expected to witness ongoing developments. However, as an initial effort, this paper may help readers identify promising problem formulations, design defensible evaluations, estimate potential impact, and anticipate failure modes in their respective disciplines. We hope this synthesis equips researchers, practitioners, and policymakers to navigate the present responsibly—and to shape a future where LLMs deliver reliable, auditable, and genuinely useful capabilities—across a wide spectrum of academic disciplines.
\newpage
\section{Background}
\label{sec:background}

In this chapter, we orient the reader within the rapidly advancing field of LLMs by clarifying both what these models are and how their performance can be meaningfully assessed. To set the stage, we begin with precise definitions and a concise historical overview, tracing the key developments that have shaped the present landscape. Building on this foundation, the chapter then turns to the current state of the art (SOTA), presenting comparative portraits of major model families—including the GPT lineage, OpenAI’s reasoning-focused systems, Claude 3, Gemini 2, Gork, Llama 3, Qwen 2, and DeepSeek—highlighting the architectural choices and distinctive capabilities that define each. From this survey, we move naturally to the question of evaluation, examining how these models are tested in practice: the principal categories of tasks, the benchmark suites most widely referenced, and the methodological approaches commonly employed. We conclude with an integrated synthesis of comparative performance, offering readers both a high-level view and practical guidance. Together, these sections establish the necessary background, provide a structured map of the model landscape, and equip readers to interpret results and make sound methodological choices in the chapters that follow.

\subsection{What are Large Language Models (LLMs)? }

\subsubsection{Definition of LLMs }
\textbf{Large Language Models (LLMs)} are AI systems designed to comprehend and generate human-like language, through the training over massive corpora of text (and increasingly other modalities) to estimate the conditional distribution of the next token given prior context, typically using Transformer-based architectures.~\cite{vaswani2017attention,radford2018improving,radford2019language,brown2020language}. At their core, LLMs rely on deep learning techniques, especially self-attention mechanisms, allowing parallel processing of text~\cite{zhao2023survey}. The fundamental logic behind LLMs involves two main phases: pre-training and fine-tuning~\cite{zhao2023survey}. In the pre-training phase, LLMs are trained on extensive text corpora using self-supervised learning tasks, such as predicting masked tokens (words) or the next word in a sequence. This process allows them to implicitly learn grammar, syntax, semantics, and factual knowledge directly from raw text data, without explicit human labeling~\cite{brown2020language, devlin2019bert}. Subsequently, in the fine-tuning phase, these models are further trained on task-specific datasets or prompted with task examples, enabling them to perform diverse language understanding and generation tasks efficiently and effectively~\cite{li2021prefix,Sanh2021MultitaskPT,lin2020exploring}. From an application perspective, this pretrain–finetune paradigm allows non-experts, including professionals outside computer science, to benefit from powerful language understanding capabilities without requiring deep technical expertise or extensive labeled data. Users can adapt LLMs to their specific needs through minimal customization, making advanced AI tools more accessible across domains such as healthcare, law, and biology.

\noindent\textbf{How to Define ``Large''?} LLMs are characterized as ``large'' by their exceptionally high number of parameters—typically ranging from billions to hundreds of billions—paired with extensive training data. For instance, OpenAI’s GPT series contains hundreds of billions of parameters, allowing it to capture a wide range of linguistic patterns and knowledge. Beyond sheer size, the term ``large'' also signifies a critical scale at which emergent capabilities arise—abilities that are not present in smaller models and often cannot be predicted simply by extrapolating from smaller-scale performance \cite{wei2022emergent}. Driven by advances in computing power and guided by scaling laws, the size of language models has increased rapidly over recent years. Models once regarded as state-of-the-art have been quickly surpassed and are now considered relatively small by current standards. For example, GPT-2, released in 2019, contained 1.5 billion parameters \cite{radford2018improving}, whereas the smallest variants of contemporary LLMs typically begin at 7 billion parameters. This shift highlights the field’s rapid progression and the evolving definition of what constitutes a ``large'' model. 

\noindent\textbf{How to Categorize LLMs?} While there are many ways to categorize LLMs, the choice often depends on the intended application and user needs. Based on the current development of LLMs, we highlight two complementary perspectives, \textit{functionality-based} \cite{pahune2023several} and \textit{reasoning-based} \cite{patil2025advancing}, that offer practical guidance for both researchers and domain professionals aiming to exploit LLMs effectively. 

From the functionality-based perspective, LLMs can be broadly divided into general-purpose and domain-specific models. General-purpose LLMs, such as GPT-3 and GPT-4 \cite{brown2020language, achiam2023gpt}, are trained on large, diverse corpora and perform well across a wide array of tasks. In contrast, domain-specific LLMs are fine-tuned on specialized data to enhance performance in focused areas, as general-purpose models may often under-perform in specialized domains due to their lack of domain-specific knowledge \cite{kerner2024domain}. For example, BioBERT \cite{lee2020biobert} targets biomedical texts, while SciBERT \cite{beltagy2019scibert} is optimized for scientific literature. This distinction is especially important for professionals in domains like healthcare, legal services, and scientific research, who require models that understand domain-specific terminology and context, yet retain general linguistic competence. 

From the reasoning-based perspective, LLMs can be distinguished by their capacity for complex inference. Reasoning-capable models, such as GPT-4 with chain-of-thought prompting \cite{wei2022chain}, GPT o1 \cite{openaio1}, and Deepseek-R1 \cite{guo2025deepseek}, are designed for multi-step reasoning tasks like mathematical problem solving or logical deduction. These models are well suited for analytical or decision-support tasks where explainability and intermediate steps matter. In contrast, non-reasoning models—though less adept at complex inference—excel in tasks where surface-level language understanding is sufficient, such as summarization, classification, or named entity recognition \cite{zhao2023survey}. These models are often more efficient and robust, making them preferable in real-time applications or resource-constrained settings.

\subsubsection{History of LLMs}

\begin{figure}
    \centering
    \includegraphics[width=0.99\linewidth]{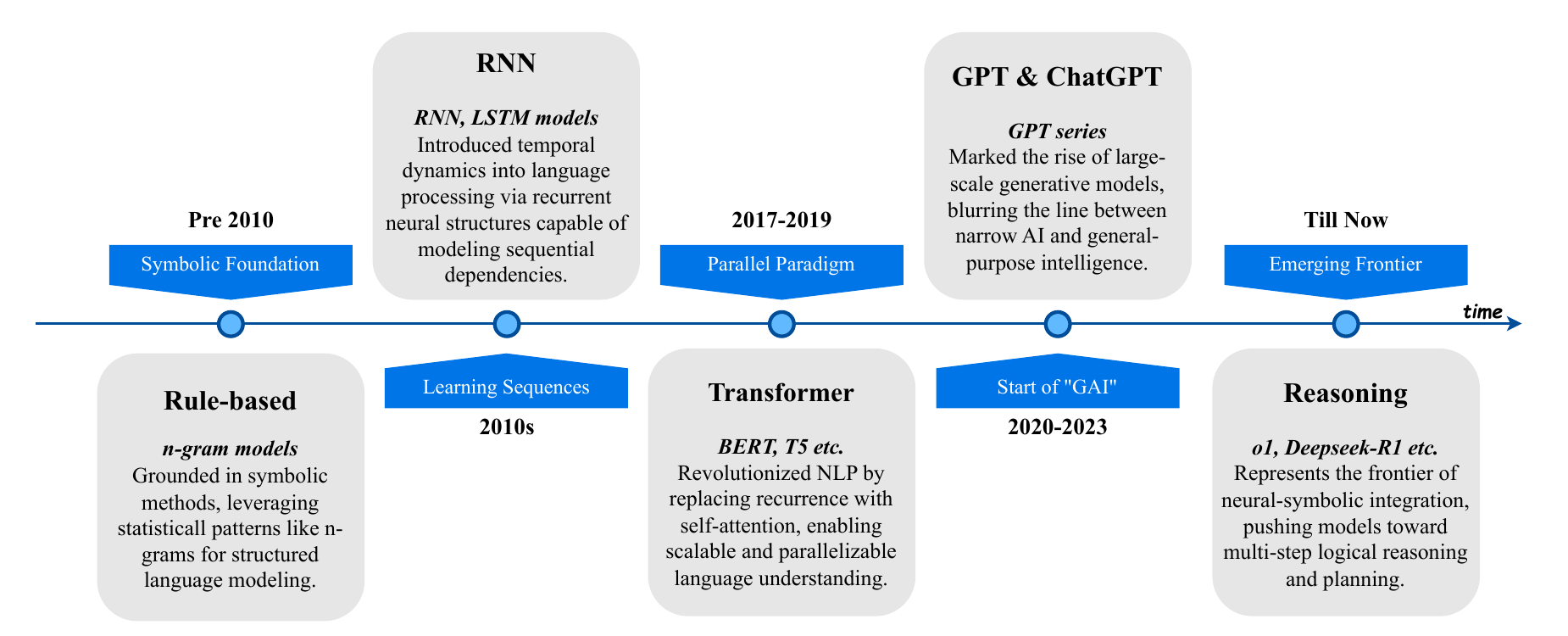}
    \caption{Key milestones in the development of LLMs.}
    \label{fig:timeline}
    \vspace{-10pt}
\end{figure}

Figure~\ref{fig:timeline} illustrates the key milestones in the development of LLMs, which will be introduced in detail below.

\noindent\textbf{Rule-based Systems: Symbolic Foundation. }The emergence of LLMs is the culmination of several decades of progress in natural language processing (NLP), moving from manual rules to statistical models to modern deep learning approaches~\cite{zhao2023survey,zhao2025surveylargelanguagemodels}. Early attempts at machine language understanding were rule-based and symbolic – researchers hand-crafted grammatical rules or expolited expert systems to parse and generate text~\cite{jelinek1998statistical,rosenfeld2000two,stolcke2002srilm}. While insightful, these systems were brittle and could not easily scale to the diversity of real-world language. By the 1990s, the field had shifted toward statistical methods: instead of manual rules, systems learned from data. For example, n-gram language models were used to predict text by learning probabilities of word sequences from large text corpora. Notably, IBM’s alignment models in the 1990s applied statistical methods to tasks like translation~\cite{brown1993mathematics,brown-etal-1988-statistical}, and by the 2000s researchers were building ever larger text datasets (``web-scale” corpor~\cite{kilgarriff2003introduction}) to train statistical language models~\cite{resnik2003web,banko2001scaling}. These data-driven models outperformed earlier symbolic approaches, as evidenced by the fact that by 2009, statistical language models had largely overtaken rule-based ones on many tasks. However, classical statistical models still had limitations — they typically considered only limited context (e.g. a fixed window of previous words) and could not capture long-range dependencies or deeper meanings effectively~\cite{zhao2023survey}.

\noindent\textbf{Recurrent Neural Networks: Learning Sequences.} A major paradigm shift came in the 2010s with the rise of neural network approaches to NLP~\cite{bengio2003neural,mikolov2010recurrent}. Inspired by successes in computer vision, researchers began applying deep learning to language. Recurrent neural networks (RNNs)~\cite{mikolov2010recurrent,kombrink2011recurrent}, especially ones with gating mechanisms like LSTM~\cite{hochreiter1997long,wang2016learningnaturallanguageinference}, were used to model sequences of text. RNN-based language models could maintain a hidden state that in principle captured information about prior words in a sequence, allowing them to handle longer contexts than n-gram models~\cite{tarwani2017survey,mienye2024recurrent}. By the mid-2010s, RNNs became state-of-the-art for tasks like translation – for instance, in 2016 Google Translate switched from a phrase-based system to a neural sequence-to-sequence model (an encoder-decoder network using LSTMs) for improved accuracy~\cite{googleFound,googleZero-Shot}. This neural takeover marked an improvement in handling context and generating more fluent outputs. Nonetheless, standard RNNs had drawbacks: they processed words sequentially and had difficulty with very long sequences or capturing long-term dependencies due to issues like vanishing gradients~\cite{zhao2023survey}.

Early progress was driven by advances in word embeddings. The introduction of Word2Vec (Mikolov et al., 2013) enabled the learning of dense vector representations of words, capturing rich semantic relationships and laying the groundwork for neural NLP. Building on this, the application of sequence-to-sequence (Seq2Seq) models with attention mechanisms (Bahdanau et al., 2015) enabled substantial improvements in tasks such as machine translation and text summarization.

\noindent\textbf{Transformers: a Parallel Paradigm Shift.} The next major breakthrough came with the introduction of the Transformer architecture in 2017, which profoundly transformed the development of language models~\cite{vaswani2017attention}. Transformer introduced the mechanism of self-attention to handle sequences, allowing the model to consider all words in a sentence in parallel rather than sequentially~\cite{vaswani2017attention}. Transformers could thus capture long-range relationships more effectively and be trained in parallel batches, dramatically improving scalability. This architecture fundamentally shifted the field, enabling the training of significantly larger models on vastly greater datasets than was previously feasible.

\begin{wrapfigure}{r}{0.4\textwidth}
  \centering
  \includegraphics[width=0.38\textwidth]{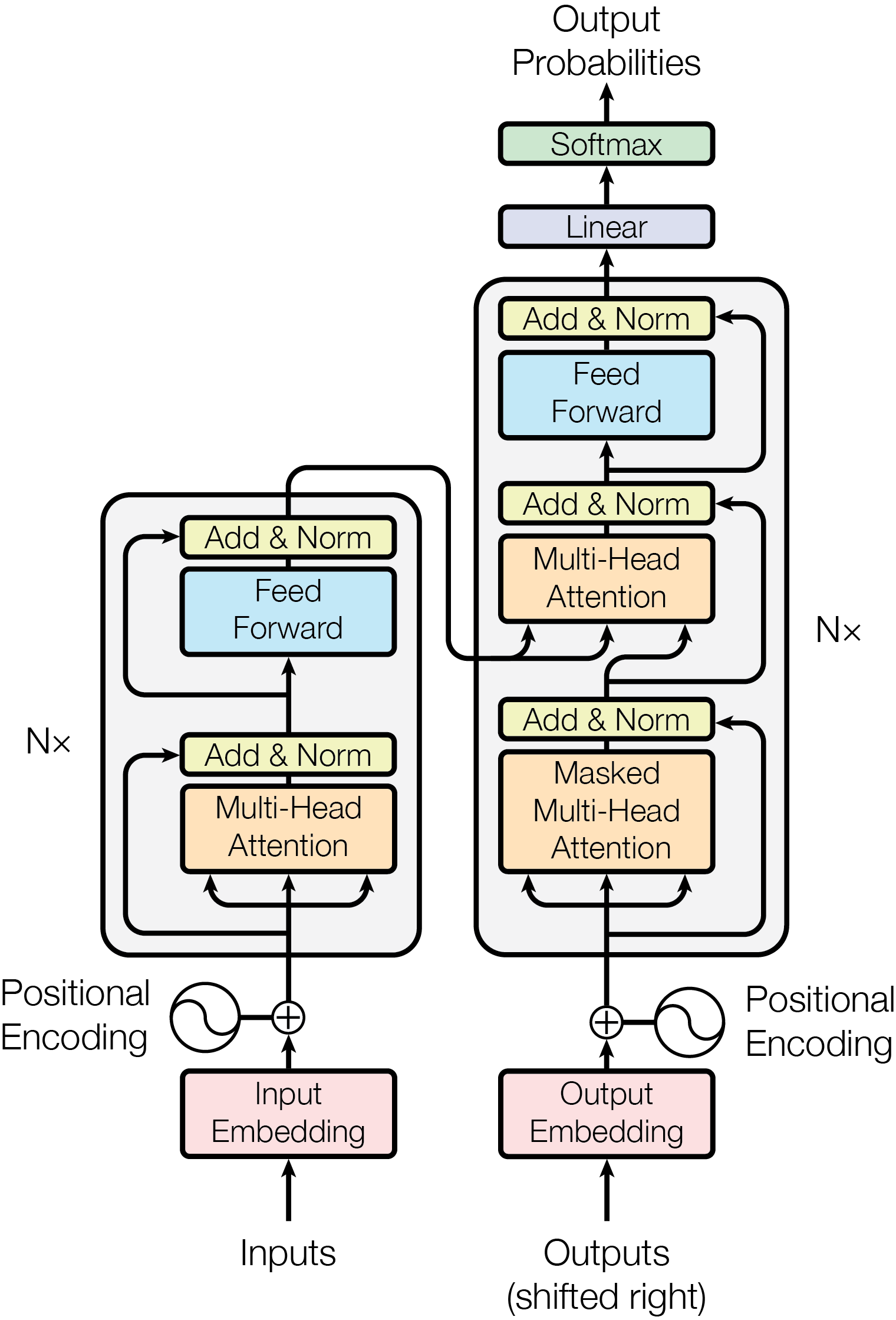}
  \caption{\centering The model architecture of Transformer (Figure source: the original paper~\cite{vaswani2017attention}).}
  \label{fig:transformer_arch}
\end{wrapfigure}

As shown in Figure~\ref{fig:transformer_arch}, Transformer model consists of two primary components: an encoder and a decoder, both built by stacking multiple identical layers. Each encoder layer contains two main sublayers: a multi-head self-attention mechanism and a feedforward neural network. The multi-head self-attention mechanism allows the model to attend to different positions of the input sequence across multiple subspaces, thereby capturing diverse semantic relationships. The self-attention mechanism computes attention weights based on the dot-product between linearly projected queries (Q) and keys (K), which are then used to aggregate values (V) through a weighted sum. By deploying multiple attention heads in parallel, the model learns rich, context-aware representations. 

The decoder shares a similar structure with the encoder but includes an additional encoder-decoder attention sublayer between the self-attention and feedforward layers. This component enables the decoder to focus on relevant parts of the input sequence during generation, which is critical for producing accurate and coherent outputs.

Within a year of its introduction, large Transformer-based models began to emerge. Notably, BERT (2018) employed the Transformer encoder to achieve unprecedented performance on natural language understanding tasks~\cite{devlin2019bert}. The success of BERT and subsequent models like T5~\cite{raffel2020exploring} popularized the pretraining-finetuning paradigm, where models are first pretrained on large corpora using objectives like masked language modeling and then fine-tuned for specific downstream tasks. Around the same time, OpenAI introduced the first Generative Pre-trained Transformer (GPT-1)\cite{radford2018improving}, which leveraged a Transformer decoder and an autoregressive training objective (predicting the next token) to specialize in text generation. GPT-1 is often regarded as the first modern LLM, despite its relatively modest scale of 117 million parameters by today’s standards. Unlike encoder-based models such as BERT, which are primarily designed for understanding tasks and lack autoregressive generation capability\cite{devlin2019bert,lan2019albert,liu2019roberta}, GPT-1 marked the beginning of a new paradigm: decoder-only architectures optimized for text generation.

\begin{table}[]
\renewcommand{\arraystretch}{1.3}  
    \centering
    \resizebox{\linewidth}{!}{
    \begin{tabular}{cccccc}
    \hline
        \textbf{Time Nodes} & \textbf{Model} & \textbf{Core Tech}  & \textbf{Contribution/Feature} & \textbf{Major Lab/Company} & \textbf{Release Time}\\
        \hline
        
        \multirow{3}{*}{\makecell{\textbf{Rule-based}\\Pre 2010}} & ELIZA~\cite{weizenbaum1966eliza} & Rule-based communication & First NLP communication system & MIT & 1966 \\
        & n-grams~\cite{brown1992class} & Markov language model & Based on statistical frequency modeling & IBM & 1992 \\
        & PCFG~\cite{johnson1998pcfg} & Context-independent probabilistic &  Syntactic analysis \& structural prediction & Brown Uni. & 1998 \\
        
        \hline

        \multirow{2}{*}{\makecell{\textbf{RNN}\\ 2010s}} & RNNLM~\cite{6424228} & Based on RNN & First RNN-based language model & Microsoft & 2010 \\
        & Seq2Seq~\cite{sutskever2014sequence} & Based on GRU / LSTM & Pioneered encoder-decoder for NLP & Google & 2014 \\

        \hline

        \multirow{3}{*}{\makecell{\textbf{Transformer}\\ 2017-2019}} & Transformer~\cite{vaswani2017attention} & Self-attention & Breaks sequential constraint, enables parallelism & Google & 2017 \\
        & BERT~\cite{devlin2019bert} & Masked LM + fine-tuning & Contextualized language representation & Google & 2018 \\
        & T5~\cite{raffel2020exploring} & Text-to-text transfer learning & Reformulates all NLP tasks as text generation & Google & 2019 \\

        \hline

        \multirow{4}{*}{\makecell{\textbf{GPT \& ChatGPT}\\ 2020-2023}} & GPT-1/2~\cite{radford2018improving,radford2019language} & Autoregressive transformer & Unified architecture for generation & OpenAI & 2018/2019 \\
        & GPT-3~\cite{brown2020language} & Large-scale autoregressive model & Kickstarted large model era & OpenAI & 2020 \\
        & ChatGPT~\cite{openai2022chatgpt} & Fine-tuned GPT-3 with RLHF & Interactive and aligned chatbot & OpenAI & 2022 \\
        & GPT-4~\cite{achiam2023gpt} & Multimodal, tool-use capabilities & Generalized across wide range of tasks & OpenAI & 2023 \\

        \hdashline
        \multirow{4}{*}{\makecell{\textbf{Other LLMs}\\ 2020-2023}} & Codex~\cite{chen2021evaluating} & Code-focused GPT fine-tuning & Natural language to code translation & OpenAI & 2021 \\
        & FLAN-T5~\cite{chung2024scaling} & Instruction-tuned T5 & Strong zero-shot generalization & Google & 2022 \\
        & QWen~\cite{qwen} & Tool-augmented transformer & Open-source model with strong instruction-following & Alibaba & 2023 \\
        & LLaMA~\cite{touvron2023llama} & Efficient transformer variants & High performance with fewer resources & Meta & 2023 \\

        \hline
        
        \multirow{4}{*}{\makecell{\textbf{Emerging Frontier}\\ Till Now}} & Mixtral~\cite{jiang2024mixtral} & Sparse Mixture-of-Experts & Efficient inference with high performance & MixtralAI & 2023 \\
        & DeepSeek-R1~\cite{guo2025deepseek} & Neural-symbolic reasoning & Multi-step logic  & DeepSeek & 2024 \\
        & o1~\cite{openaio1} & Experimental AGI prototype & Focus on generalized reasoning skills & OpenAI & 2024 \\
        \hline

    \end{tabular}}
    \vspace{8pt}
    \caption{Several important models in different time nodes.}
    \vspace{-20pt}
    \label{tab:my_label}
\end{table}

\noindent\textbf{GPT and ChatGPT: From Language Modeling to General Intelligence. }
The GPT series by OpenAI, based on the Transformer architecture, incorporated key design innovations that positioned it as a landmark in the advancement of LLMs. Unlike BERT’s masked token prediction, GPT models use causal (autoregressive) language modeling, predicting the next token given all previous tokens~\cite{radford2018improving,radford2019language,brown2020language}. This seemingly simple change allows GPT models to generate long, coherent passages of text. The architecture is \textbf{decoder-only}, stacking multiple Transformer blocks to model the distribution over sequences. The evolution from GPT-1~\cite{radford2018improving} through GPT-2~\cite{radford2019language} to GPT-3~\cite{brown2020language} (2018–2020) was marked by exponential growth in both model size and training data~\cite{zhao2023survey}. GPT-3~\cite{brown2020language} with 175 billion parameters, was trained on a massive and diverse dataset covering web text, books, Wikipedia, and more. It demonstrated few-shot and even zero-shot learning: the ability to perform unseen tasks when given just a prompt or a few examples, without additional training. This behavior, previously unseen, signaled an emergent form of general-purpose linguistic intelligence.

In fact, it was the release of ChatGPT in late 2022 that truly rendered LLMs accessible and practically useful at scale, introducing a conversational interface that brought LLM capabilities to millions of users and triggered a global wave of LLM applications. While based on GPT-3.5~\cite{brown2020language} and later GPT-4~\cite{achiam2023gpt}, ChatGPT introduced instruction tuning and reinforcement learning from human feedback (RLHF) to align the model with human values, dialogue etiquette, and safety constraints~\cite{achiam2023gpt}. Unlike vanilla GPT-3, which was often unpredictable, ChatGPT could follow instructions, engage in multi-turn conversations, and avoid unsafe outputs—making it viable for deployment in education, research, programming, and creative tasks.


The difference between GPT and vanilla Transformer is thus not just in architecture (both use the Transformer as a backbone), but in how GPT leverages generative pretraining, scaling laws~\cite{rae2021scaling,kaplan2020scaling,rae2021scaling}, few-shot prompting~\cite{kang2023large,song2023llm}, and instruction alignment~\cite{wang2023aligning} to transition from a language model to a general-purpose AI assistant. ChatGPT represents a milestone because it closes the loop between model, user, and feedback, making LLMs interactive, helpful, and widely usable.

\noindent\textbf{Reasoning: The Emerging Frontier. }
With the widespread adoption of LLMs (LLMs), researchers have observed that these models often struggle with multi-step problems and abstract logical tasks. To further enhance their capabilities and move closer to the goal of artificial general intelligence (AGI), substantial efforts have been made to improve their reasoning abilities. The introduction of chain-of-thought prompting~\cite{wei2023chainofthoughtpromptingelicitsreasoning} has emerged as a promising approach to address this limitation. By enabling models to decompose problems into intermediate reasoning steps, this method mirrors the way humans tackle complex tasks~\cite{wang2022towards,huang2022towards}. This paradigm shift has empowered LLMs to handle tasks requiring sequential logical reasoning, ranging from mathematical problem solving to complex decision-making processes~\cite{huang2022towards}.

The year 2023 saw continued innovation with the release of GPT-4, a multimodal model capable of processing both text and images, along with Claude (Anthropic), which emphasized alignment and safety through Constitutional AI, and LLaMA (Meta), which spurred a vibrant open-source LLM community. 
Most recently, GPT-4o (2024) introduced real-time multimodal capabilities—integrating text, vision, and audio—with improved conversational alignment and latency, signaling a new phase in the evolution of interactive and multimodal AI systems.

Complementing this approach, self-consistency techniques~\cite{wang2023selfconsistencyimproveschainthought} have been proposed to further refine reasoning performance~\cite{huang2022towards}. These methods sample multiple reasoning paths and select the most coherent outcome, significantly reducing error rates and enhancing the reliability of model outputs. Additionally, a variety of reinforcement learning-based strategies have been employed to improve reasoning capabilities. One notable example is the DeepSeek-R1~\cite{guo2025deepseek} model, which contains 671 billion parameters and is specifically optimized for tasks involving mathematics, programming, and logical reasoning. It leverages reinforcement learning with a rule-based reward mechanism (rule base reward + RL) to enhance its reasoning proficiency~\cite{guo2025deepseek}. As models continue to improve in planning and reasoning over complex problems, humanity appears to be gradually approaching AGI.

\subsection{State-of-the-art LLMs}

\begin{figure*}[t]
    \centering
    \includegraphics[width=\linewidth]{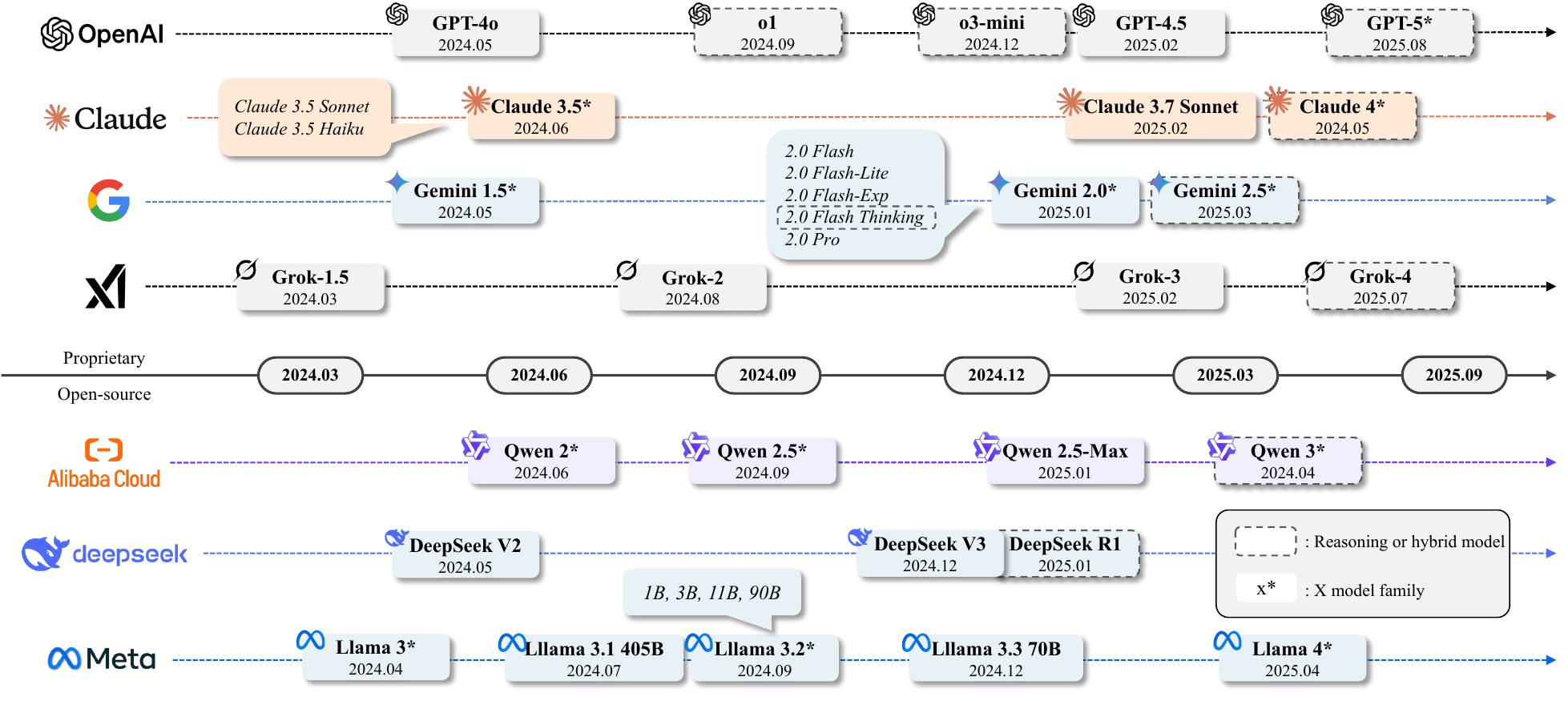}
    \caption{Chronological display of current SOTA LLMs.}
    \label{fig: sota llm}
\end{figure*}

\begin{table}[]
    \caption{Existing SOTA LLMs on benchmark datasets.
   }
    \label{tab:sota llm}
    \resizebox{\linewidth}{!}{
    \begin{tabular}{clcccccccccc}
    \toprule 
    \multirow{2}{*}{Family}   &     Model & {Context}  &  {Output}  & Input & \multicolumn{2}{c}{Cost (\$/1M Tok)}  & \multicolumn{2}{c}{Infer. Time} &  \multirow{2}{*}{Strengths}  & \multirow{2}{*}{Provider} \\
    & Name &  Window & Size & Modal & Input & Output & TPS  & TFT & \\
      \midrule\midrule
      \multirow{4}{*}{GPT 4~\cite{achiam2023gpt, openai2025gpt4.5}} 
      & GPT-4o &  128K & 16,384 & \multirow{7}{*}{T, I}  & 2.50 & 10.00 & 209.4 & 0.35 &  High TPS, General-Purpose Tasks  & \multirow{4}{*}{OpenAI} \\ 
      & GPT-4o mini & 128K  & 16,384&    & 0.15 & 0.60 & 74.3 & 0.35 &  Fine-tune / Distill, Focused Tasks & & \\ 
      & ChatGPT-4o & 128K & 16,384&    & 5.00 & 15.00 & 214.0 & 0.81 & Chat-based Applications & & \\ 
      & GPT-4.5 & 128K  & 16,384 &    & 75.00 & 150.00 & 11.5 & 1.47 & 
        Complex / Creative Tasks \\
      \midrule
 & GPT-5 &  400K & 128K & & 1.25 & 10.00 & 136.0 & 1.03 & Coding, Reasoning, and Agentic Tasks. & \\ 
      GPT 5~\cite{openai_gpt5_system_card_2025} & GPT-5-mini &  400K & 128K & & 0.25 & 2.00 & 73.0 & 0.98 &  Well-defined Tasks / Precise Prompts. & OpenAI \\
\bottomrule
    \end{tabular}
    \vspace{3mm}
    }
    \resizebox{\linewidth}{!}{ 
    \begin{tabular}{clccccccccccc}
    \midrule 
   Family & Model Name & Model Size & Context Window & Input Modal & Strengths & License & Provider \\
   \midrule\midrule
   \multirow{2}{*}{GPT-OOS~\cite{openai2025gptoss120bgptoss20bmodel}} & GPT-oos-20B & 21B & \multirow{2}{*}{131K} & \multirow{2}{*}{T} & \multirow{2}{*}{Reasoning and Agentic Tasks} & \multirow{2}{*}{Apache 2.0} & \multirow{2}{*}{OpenAI} \\
   & GPT-oss-120B & 117B & & & & & \\ 
   \midrule
   \multirow{4}{*}{Qwen 2~\cite{yang2024qwen2technicalreport, yang2024qwen25, qwq32b}}   & Qwen 2& 0.5B - 72B &  131,072 & T & Previous Flagship Models & & \multirow{4}{*}{Alibaba}\\ 
   & Qwen 2.5 &  0.5B - 72B   & 128K & T & General-Purpose Tasks & Apache & \\ 
   & Qwen 2.5-1M & 7B, 14B & 1M & T & Long Context Tasks & License &  \\
   & QwQ & 32B & 131,072 & T & Reasoning and QA Tasks&   & \\
   \midrule 
   \multirow{3}{*}{DeepSeek~\cite{guo2025deepseek, liu2024deepseek}}  & DeepSeek-V3 & 671B & \multirow{3}{*}{128K} & T & MoE, Math & \multirow{3}{*}{MIT License} & \multirow{3}{*}{DeepSeek}\\
   & DeepSeek-R1 & 671B &  & T &  Complex Reasoning Tasks & \\
   & DeepSeek-R1 Distill & 1.5B - 70B & & T & Efficient Reasoning Inference & \\ 
   \midrule 
   \multirow{4}{*}{Llama 3~\cite{grattafiori2024llama}} & Llama 3.1  & 8B, 70B, 405B & \multirow{4}{*}{128K} & T & Multilingual, Text Generation & \multirow{4}{*}{Custom License} & \multirow{4}{*}{Meta AI}\\
   & Llama 3.2 LW & 1B, 3B &  & T & Efficient Inference  &  & \\
   & Llama 3.2 MM & 11B, 90B & &  T, I & Multimodal, Image Tasks & & \\
   & Llama 3.3 & 70B & & T & -- & & \\
   \midrule
    \bottomrule
    \end{tabular}}
    Notes:  T, I, and A denote Text, Image, and Audio, respectively. 
    For time, we report the median time for 1K tokens prompt length. 
    TPS is the output \textbf{T}okens \textbf{P}er \textbf{S}econd, and TFT is the \textbf{T}ime to \textbf{F}irst \textbf{T}oken, which is also called latency. Gemini 2.0 Flash TK represents Gemini 2.0 Flash Thinking. All the cost information are collected from LLM provider official websites, and inference time (Infer. Time) are obtained from \href{https://artificialanalysis.ai/leaderboards/models}{Artificial Analysis}. The strengths of each LLM are summarized according to the corresponding model technical report,  \href{https://www.vellum.ai/llm-leaderboard}{Vellum LLM Leaderboard}, and benchmark reports.
    Dash means the information is not available.
\end{table}
\begin{table}[]
     \caption{Guidelines for LLMs selection based on tasks and corresponding constraints.}
    \label{tab: llm selection}
    \resizebox{\linewidth}{!}{
    \begin{tabular}{c|c|c|c|c|c|c|c|c}
    \toprule
       Task  & Modality &  Context Window & Latency & Privacy & Budget & Hardware & Perf. & LLMs \\
       \midrule\midrule
       \multirow{4}{*}{Conversational Task} & T, (I, A)  & $\ge$ 200K & Yes & No & Low & -- & Med & Gemini 2\\
       & T, (I) & $<$ 200K  & Yes & No & Med & -- & High &  ChatGPT-4o\\
       & T & $ <$ 200K & No & Yes & -- & Med & -- & Qwen 2.5 \\
        & T, (I) & $ <$ 200K &  No & Yes & -- & High & -- & Llama 3.2 MM \\
        \midrule
        \multirow{3}{*}{Reasoning Task} & T, (I, A) & $\ge$ 200K & Yes & No & -- & -- & High &  Gemini 2.5 Pro\\
        & T & $ <$ 200K & No & No & Med & -- & Med & OpenAI Reasoning \\ 
        & T & $ <$ 200K & Yes & Yes & -- & High & -- & DeepSeek-R1\\
        \midrule
        \multirow{3}{*}{Coding Task} &  T & $<$ 200K & Yes & No & Low & -- & Med & Claude 3.5 \\
        &  T & $<$ 200K & No & No & High & -- & High & Claude 3.7 \\
        & T & $\ge$ 200K & No & Yes & -- & Med & -- & Qwen 2.5-1M\\
        \midrule
        \multirow{3}{*}{Open-Domain QA}  
        & T, (I, A) &  $\ge$ 200K & Yes & No & Low & -- & Med & Gemini 2\\
        & T & $<$ 200K & Yes & No &  Med & -- & High & GPT-4o\\
        & T &  $<$ 200K & Yes & Yes & -- & Med & -- &QwQ 32B\\
        \midrule
        \multirow{3}{*}{Focused Tasks (Fine-tune)} & T & $<$ 200K & Yes & No &  Low & -- & -- & GPT-4o mini \\
        & T, (I) & $<$ 200K & No & Yes & -- & Med & -- &LLama 3.2 MM\\
        & T & $<$ 200K & Yes & Yes & -- & Low & -- & DeepSeek-R1 Distill\\
       \bottomrule
    \end{tabular}
   }
   Notes: T, I, and A denote Text, Image, and Audio, respectively. Brackets surround optional input modality. 
   In latency, YES indicates the task has a latency requirement. 
   For privacy, YES represents that the input data may be confidential. 
   For budget, hardware, and performance(perf.), we report in low, med (median), and high, based on the relative criteria. 
   For example, LOW BUDGET prefers costs less than \$1 per 1M tokens, HIGH HARDWARE indicates a requirement to run LLMs with model size greater than or equal to 90B. 
   The recommended LLMs are based on the performance, requirements, and strengths provided in the provider official websites, \href{https://artificialanalysis.ai/leaderboards/models}{Artificial Analysis}, and \href{https://www.vellum.ai/llm-leaderboard}{Vellum LLM Leaderboard}. 
   These provided LLMs in each scenario are \textbf{merely for referenes}, and are \textbf{not guaranteed to be the optimal solutions}. Users should \textbf{consider their application scenarios when choosing the LLMs}.
   
\end{table}

\subsubsection{Overview}
Nowadays, a variety of LLMs are available to address different downstream tasks.
These models can be broadly categorized as closed-source (accessed through APIs) or open-source (deployed locally).
Given varied requirements such as latency, performance, and input/output modalities, no single LLM can simultaneously satisfy all use cases.
In general, closed-source LLMs outperform open-source LLMs due to large model sizes, extensive training corpora, and training recipes.
They also tend to support larger context windows and higher token throughput, enabling more complex and longer interactions. 
However, they often require API usage, resulting in high time to first token (TFT), also called latency, and may incur substantial costs, especially for reasoning models, such as OpenAI o1-pro. 
Additionally, off-site inference can raise privacy concerns when handling sensitive data.
By contrast, open-source LLMs typically feature smaller model sizes, allowing cost-effective deployment on local hardware.
Open-source LLMs often yield faster response times (depending on the user's infrastructure) and stricter data confidentiality. 
Aside from the aforementioned basic factors, LLM capabilities should also be considered. 
We summarize the key features of SOTA LLMs in Table~\ref{tab:sota llm}.
In the subsequent sections, we undertake a systematic analysis of each SOTA LLM model family in detail. Note that with the rapid development of AI technology, LLMs are also evolving quickly. This section only covers the mainstream SOTA models up to the time of publication. As GPT plays a central and milestone role in the development of LLMs—by establishing the methodological paradigm, advancing large-scale modeling, and shaping the surrounding application ecosystem—we begin by introducing and discussing the GPT-series models in the following section (Section~\ref{sec: gpt series models}).

\subsubsection{GPT-series Models} \label{sec: gpt series models}
The Generative Pre-training Transformer (GPT) series represents a family of autoregressive language models based on Transformer architecture~\cite{vaswani2017attention}, introduced by OpenAI~\footnote{https://openai.com/}, 
These models employ a self-supervised learning paradigm that generates contextually coherent text through sequential token prediction.
Due to the excellent performance in generative language tasks, GPT models have introduced multiple breakthroughs in the NLP community~\cite{zhao2023survey, yenduri2024gpt}. 
In this section, we will discuss each GPT model and its contributions in detail.

\textbf{GPT-1~\cite{radford2018improving}.} Prior to the emergence of GPT, conventional NLP models are trained over large amounts of task-specific annotated datasets, leading to limited generalization capabilities across tasks beyond trained datasets.
To address this challenge, OpenAI developed GPT-1 in 2018, a decoder-only transformer architecture with 117 million parameters, and adopts a two-stage training recipe: (i) unsupervised pre-training on large text corpus; and (ii) supervised fine-tuning. 
One of the successes of GPT-1 is the excellent zero-shot performance across multiple downstream tasks, including sentiment analysis, question answering, etc. 
Besides the success in model performance, \textbf{the underlying principle to model natural language text, i.e., next token (word) prediction, also has a profound influence on the development of subsequent LLMs~\cite{yenduri2023generative}.}

\textbf{GPT-2~\cite{radford2019language}.} In late 2019, OpenAI released GPT-2 which employed similar architecture of GPT-1, with 10 times the size of GPT-1 (117M to 1.5B parameters). The model is trained over a newly collected webpage dataset, called WebText, which contains slightly over 8 million documents~\cite{radford2019language}. 
GPT-2 sought to perform multi-task via unsupervised learning, without explicit fine-tuning over labeled datasets. 
Motivated by existing studies for the probabilistic framework with task condition~\cite{kaiser2017one, finn2017model, mccann2018natural}, GPT-2 introduces a probabilistic framework for multi-task solving, formulated as
\begin{equation}
    p(\texttt{Output}| \texttt{Input}, \texttt{Task}), ~\label{eq: probability task condition}
\end{equation}
which generates output conditioned on the input and task information. 
Here, the task information can be regarded as the pioneer of the concept \textbf{In-Context Learning} (ICL) in the current LLMs community. 
Besides the proposed probabilistic model, the success of GPT-2 under the unsupervised multi-task learning settings is rooted in their training philosophy: \textbf{the global minimum of the unsupervised objective is also the global minimum of the supervised objective}~\cite{radford2019language}.

\textbf{GPT-3 and GPT-3.5~\cite{brown2020language} .} Although GPT-2 has provided significant insights into the LLM community, the overall performance of GPT-2 is lower than supervised SOTA models.
OpenAI extended GPT-2 to GPT-3 in 2020, which demonstrated incremental performance compared to GPT-2 and supervised fine-tuning models, by scaling up the model architectures to 175B parameters, i.e., the largest language model ever at that time. 
Although not explicitly stated, the performance gain of GPT-3 over GPT-2 validates the scaling law~\cite{kaplan2020scaling} that large models, in terms of model parameters, have stronger capabilities. 
Another contribution of GPT-3 is \textbf{the introduction of ICL, which instructs LLMs with a few demonstrations of the task at inference time}.
These demonstrations can be viewed as task conditioning in Equation~\ref{eq: probability task condition}.
To enhance the capability over complex tasks, such as code completion and math problems, OpenAI develops a stronger capability model than GPT-3 in complex problem solving, by training over code dataset, called GPT-3.5~\cite{chen2021evaluating}.
In addition to capability improvement, GPT-3.5 is also trained with a three-stage reinforcement learning algorithm from human feedback (RLHF) (first introduced in InstructGPT~\cite{ouyang2022training}), which helps enhance the ability to follow instructions and ease concerns regarding LLMs producing toxic responses or violation of local policies.
The research contribution of GPT-3.5 and RLHF can be summarized in three directions: training LLMs \textbf{(i) using human feedback, (ii) to assist human evaluation, and (iii) to do alignment research}~\cite{zhao2023survey, openai2023alignment}.

\textbf{ChatGPT.} 
OpenAI launched a conversation model called ChatGPT in November 2022, which achieves a pivotal milestone in the AI research community. 
ChatGPT is a sibling model to InstructGPT, while specially optimized on a human-generated conversation dataset~\cite{openai2022chatgpt}.
ChatGPT demonstrates superior ability in communications with humans, and the additional support of the plugin mechanisms enables ChatGPT to obtain external knowledge by interacting with plugins, such as a calculator and web search etc.
The plugin support can be viewed as the prototype of MCP~\cite{claude2025MCP}.
The success of ChatGPT marks the exploration of LLMs and has a significant impact on future research in the LLM domain.

\textbf{GPT-4*~\cite{achiam2023gpt, openai2025gpt4.5}.}
The aforementioned GPT series models only support text input.
To extend the model input from single text to multimodal signals (text and image), OpenAI published GPT-4 in March 2023. 
GPT-4 outperforms earlier GPT models, including ChatGPT, in various benchmark datasets.
Moreover, as reported in GPT-4 technical report~\cite{achiam2023gpt}, OpenAI spent six months on human alignment training in RLHF to alleviate the safety concerns of GPT-4. 
Besides the remarkable performance gain and alleviation of safety concerns, GPT-4 is trained over a new principle called \textbf{predictable scaling}, which refers to the \textbf{development of infrastructure that allows for reliable extrapolation of the performance across scales of compute and model sizes.} 
This approach is adopted by most later LLMs to minimize the need for extensive model-specific tuning, making the training process more efficient and systematic. Several GPT-4* models are released in late 2024 or early 2025, i.e., GPT-4o~\cite{openai2025gpt4o}, GPT-4o mini~\cite{openai2025gpt4oMini}, and GPT-4.5~\cite{openai2025gpt4.5}, which step further than GPT-4. 
GPT-4o and GPT-4o mini are multilingual, multimodal (text, image, audio, and video) models that generate any combination of text, audio, and image as output.
GPT-4.5 emphasizes improved writing capabilities, enhanced world knowledge, and a refined interaction experience. \\

\textbf{GPT-5*~\cite{openai_gpt5_system_card_2025}.}
    GPT-5 is introduced as a model family with multiple sizes, including GPT-5, GPT-5-mini, and GPT-5-nano, and two principal configurations: a standard model and a thinking variant optimized for extended, deliberate reasoning.
    The release highlights step-change improvements in core capabilities: stronger multi-step reasoning, more capable multimodality, and more reliable tool use and orchestration.
    On the training and alignment side, GPT-5 adopts an expanded RLHF pipeline and upgraded data curation, which together aim to improve instruction following, factuality, and controllability. 
    Safety systems are deepened with tighter gating and continuous monitoring, particularly for sensitive domains such as biosecurity, cybersecurity, and model autonomy. 
    Moreover, according to their report~\cite{openai_gpt5_system_card_2025}, GPT-5 has improved calibration and robustness, enabling more dependable model behavior under varied prompts and contexts.
    Operationally, GPT-5 delivers lower latency and better cost efficiency than prior GPT-4-class models, supported by serving-side and architectural optimizations. 
    Taken together, these changes position GPT-5 as a more capable, controllable, and practical foundation model for both general and high-stakes reasoning use cases, with a specialized thinking configuration for complex problem solving.

\subsubsection{OpenAI Reasoning Models} \label{sec: openai reasoning model intro}

OpenAI Reasoning Models, such as OpenAI o1 and o3-mini, represent a significant evolution in LLM design, specifically engineered to address complex, multi-step reasoning tasks \cite{openai_reasoning}. These models, collectively known as the "o" series, introduce innovations including reasoning tokens and advanced reinforcement learning techniques. Such features facilitate a structured "chain-of-thought" reasoning process, wherein the model generates internal reasoning steps prior to producing a final response. This approach proves particularly effective in domains demanding sophisticated problem-solving, including advanced programming, scientific inquiry, and strategic planning. Moreover, the models are available in configurations with varying computational demands, allowing users to balance speed and accuracy based on application-specific requirements. This emphasis on explicit, structured reasoning represents a departure from traditional LLM architectures, aiming to generate outputs that are not only accurate but also traceable through logical inference. OpenAI’s release of its reasoning models marks the onset of heightened competition among leading technology firms in the realm of reasoning-oriented artificial intelligence.

\subsubsection{Claude 3 Model Family}

Claude 3~\cite{claude3modelcard, claude3.7modelcard} is a family of LLMs developed by Anthropic (https://www.anthropic.com/), a company founded in 2021 by former OpenAI employees. 
Claude utilizes Transformer architecture and has gone through versions \textit{Claude 3 Opus}, \textit{Claude 3.5 Sonnet}, \textit{Claude 3.5 Haiku}, and \textit{Claude 3.7 Sonnet} from 2024 to 2025.
At the heart of Claude, the task is training LLMs to be helpful, honest and harmless~\cite{claude3modelcard}.
Claude achieves this by incorporating a \textbf{Constitution}, which contains predefined ethical and behavioral guidelines that shape the outputs of LLMs. 
Most of the principles in the Constitution are introduced in their earlier post~\cite{claudeconstitution}, with an additional principle based on feedback from the public input process that directs Claude to be empathetic and accessible to people with disabilities, thereby reducing model stereotype bias.

The Claude 3 family offers various models with capabilities to meet specific needs. 
Claude 3.5 Haiku, the fastest model in the Claude 3 family, is optimized for near-instant responses, which is suitable for real-time tasks to mimic human interactions, such as customer support, content moderation, etc.
Claude 3.5 Sonnet balances performance and speed, excelling in enterprise workloads like data processing and code generation at a low cost. 
Claude 3.7 Sonnet, the latest Claude model, is a hybrid reasoning model that provides the thinking process in the output. 
It includes a toggleable "extended thinking" model in which Claude produces a sequence of tokens as a "thinking process" to work through complex problems before delivering the final response. 
This mode, trained through reinforcement learning, allows for detailed step-by-step reasoning, which can be adjusted by the user to specify a token limit.
Experiments reported in Claude 3.7 Sonnet System Card~\cite{claude3.7modelcard} demonstrate that the "extend thinking" model is particularly valuable for challenging tasks, such as mathematical problems, complex analysis, and multi-step reasoning tasks.
Overall, the Claude 3 models are distinguished by their multi-modal capabilities, allowing them to process visual inputs alongside text. 
They demonstrate SOTA performance on vision-related benchmarks and quantitative reasoning tasks~\cite{claude2024introducing}. 
Additionally, Claude 3.5 Sonnet stands as the SOTA model in \textbf{coding benchmarks, and maintains strong performance for routine programming tasks in practical applications.}

\subsubsection{Gemini 2 Model Family}
Gemini 2.0 represents Google's latest advancement in multimodal LLMs, encompassing a comprehensive suite of models tailored to diverse computational needs \cite{google_gemini}. Central to this suite is Gemini 2.0 Flash, a high-performance model optimized for rapid response and efficient handling of general-purpose tasks. Accompanying this is Gemini 2.0 Flash-Lite, a more cost-effective variant designed to maintain substantial performance while reducing computational overhead. Additionally, Gemini 2.0 Pro demonstrates particular strengths in code generation, tool use, and the processing of complex prompts with its 2 million context window. Key innovations in the Gemini 2.0 series include native tool usage capabilities, image generation, and speech synthesis, all within an expanded multimodal framework \cite{google_gemini_dec2024}. \textbf{Enhanced context window size and improved integration of multiple input-output modalities} position Gemini 2.0 as a pivotal tool in the evolution toward agentic AI systems.

\subsubsection{Gork Model Family}
Grok 3, developed by xAI, integrates extensive pretraining with enhanced reasoning capabilities enabled by the Colossus supercluster, which offers an order-of-magnitude increase in computational power over previous state-of-the-art models \cite{xai_grok3}. Grok 3 exhibits marked improvements in areas such as logical reasoning, mathematics, coding, factual recall, and adherence to complex instructions. Its reasoning proficiency is reinforced through large-scale reinforcement learning, enabling the model to engage in extended problem-solving, correct internal inconsistencies, and explore alternative solution pathways. This iterative training process yields more accurate and dependable outputs. Benchmark evaluations reveal that Grok 3 achieves a leading Elo score of 1402 in the Chatbot Arena \cite{xai_grok3} (the score may change over time). In addition, xAI has released Grok 3 mini, a more computationally efficient variant that retains strong reasoning capabilities. However, at the time of writing, APIs for the Grok 3 series remain unavailable to the public.

Next, we provide reviews for open-source LLMs, i.e., GPT-OSS, Llama 3, Qwen, and DeepSeek Model Family, that the model parameters are available on public platforms, such as Huggingface (https://huggingface.co/) and Github (https://github.com/).
While open-source, the licensing terms of each model may vary significantly, necessitating careful consideration when deploying them in research or production environments.

\subsubsection{GPT-OSS}
\textbf{GPT} \textbf{O}pen-\textbf{S}ource \textbf{S}eries (GPT-OSS)~\cite{openai2025gptoss120bgptoss20bmodel} is a family of open-weight language models released by OpenAI under the Apache 2.0 license in August 2025, marking a notable shift in the company's stance toward open-source AI.
The models employ autoregressive MoE transformer architectures and come in two sizes: gpt-oss-120b and gpt-oss-20b.
Specifically, gpt-oss-120b consists of 36 layers, with 116.8B total parameters and 5.1B “active” parameters per token per forward pass, while gpt-oss-20b has 24 layers with 20.9B total and 3.6B active parameters.
Training combines RL with methods informed by OpenAI's most advanced internal system, e.g., o3 and related frontier models.
The models standardize on an extended o200k\_harmony tokenizer and a harmony chat schema that encodes role hierarchy and channels for CoT, tool calls, and final outputs, which support reliable multi-turn, agentic behavior and seamless interleaving of reasoning with function execution.
In evaluation, gpt-oss-120b  achieves near-parity with o4-mini on core reasoning benchmarks, while running efficiently on a single 80 GB GPU. 
The gpt-oss-20b model delivers performance comparable to o3-mini on common benchmarks and can run on consumer devices with just 16 GB of memory, making it ideal for on-device use cases, local inference, or rapid iteration without costly infrastructure. 
Both models also perform strongly on tool use, few-shot function calling, and CoT reasoning.
Together, these elements yield an open, deployable blueprint for long-context, tool-using, and compute-adaptive reasoning systems.

\subsubsection{Llama 3 Model Family}  
\textbf{L}arge \textbf{La}nguage Model \textbf{M}eta \textbf{A}I (Llama) is a family of LLMs introduced by Meta AI (https://www.meta.ai/), first released in February 2023. 
It serves as Meta's response to OpenAI's GPT models and is designed as a foundational model for various NLP tasks~\cite{tiernan2023chatgpt, nik2023meta}.
Llama 3 Model Family~\cite{grattafiori2024llama} has gone through versions Llama 3, to 3.3, with model parameters ranging from 1B to 405B.
All of these models are auto-regressive decoder-only models based on the Transformer with slight modifications for efficiency purposes. 
Specifically, Llama 3 leverages grouped query attention~\cite{ainslie2023gqa} with eight heads to improve inference speed and to reduce the size of key-value caches during decoding. 
The performance gain of Llama 3 compared with previous versions, i.e., Llama 1 \& 2, is primarily driven by improvement in data quality and diversity, as well as by increased training scale.
Unlike other LLM model families that employ RLHF for human preference alignment, Llama 3 adopts Direct Preference Optimization (DPO), which directly optimizes for the policy best satisfying the preferences with a simple classification object. 
Meta AI also explores Proximal Policy Optimization (PPO)~\cite{schulman2017proximal}, but found that DPO requires less computing for large-scale models and performs better. 
The success of Llama is rooted in the efficient training recipe and the huge high-quality training data.

\subsubsection{Qwen 2 Model Family}
Qwen 2 model family is developed by Alibaba Cloud (https://www.alibabacloud.com), also known as Tongyi Qianwen. 
These models are designed for a variety of downstream tasks, including NLP, multimodal understanding, and coding assistance. 
The first version of the Qwen 2 model family is Qwen 2~\cite{yang2024qwen2technicalreport} with model parameters ranging from 0.5B to 72B.
In September 2024, an extended version, called Qwen 2.5, was released with more model size options compared to Qwen 2, such as 3B, 14B, and 32B. 
In earlier 2025, Alibaba Cloude further released an advanced version, Qwen 2.5 Max, and a reasoning model called QWQ-32B.
The architecture of Qwen 2 model family is similar to Llama 3, which adopt Transformer-based decoder architecture with GQA~\cite{ainslie2023gqa} for efficient KV cache, SwiGLU activation~\cite{dauphin2017language} for non-linear activation, and RoPE~\cite{su2024roformer} for encoding position information~\cite{yang2024qwen25}.
All of the aforementioned Qwen 2 models are available at Huggingface. 
One success of the Qwen 2 model family is relevant to their innovative Mixture of Expert (Moe)~\cite{baldacchino2016variational}, which efficiently allocates computational resources, improving scalability and performance.
Moreover, Qwen 2 models support multilingualism across 29+ languages for global applications.

\subsubsection{DeepSeek Model Family}
\textbf{DeepSeek-V3.} DeepSeek-V3, released by DeepSeek in December 2024, employs a Mixture-of-Experts (MoE) architecture comprising 671 billion parameters, with 37 billion parameters activated per token \cite{liu2024deepseek}. This dynamic routing mechanism allows the model to selectively activate relevant subsets of parameters based on input characteristics, enhancing both computational efficiency and model performance. Trained on a vast multilingual dataset totaling 14.8 trillion tokens—primarily in English and Chinese—over a 55-day period, the training process utilized 2,048 NVIDIA H800 GPUs at an estimated cost of \$5.6 million. This is significantly more cost-efficient than comparable models such as GPT-4, whose training expenditures are estimated to range between \$50–100 million. Benchmark results indicate that DeepSeek-V3 surpasses models such as LLaMA 3.1 and Qwen 2.5, and achieves parity with leading models like GPT-4o and Claude 3.5 Sonnet.

\textbf{DeepSeek-R1.} DeepSeek-R1 \cite{guo2025deepseek}, introduced in January 2025 by DeepSeek, advances the reasoning capabilities of its predecessor through an enhanced MoE architecture and a multi-stage training regimen \cite{guo2025deepseek}. Like DeepSeek-V3, R1 consists of 671 billion parameters with 37 billion activated per token, optimizing the balance between scale and computational efficiency. A defining feature of DeepSeek-R1 is its emphasis on reinforcement learning (RL) to cultivate advanced reasoning behaviors. The model was initially subjected to supervised fine-tuning using a curated dataset of chain-of-thought exemplars—a phase referred to as the "cold start." This was followed by large-scale RL using the Group Relative Policy Optimization (GRPO) algorithm, which incentivizes autonomous development of reasoning strategies, including self-verification and error correction. This robust training strategy enables DeepSeek-R1 to attain reasoning performance on par with OpenAI's o1 model, while maintaining significantly lower training costs. Crucially, DeepSeek-R1 has been released under the permissive MIT License, granting the research community unrestricted access to its model weights and outputs, thereby fostering transparency and collaborative innovation in AI development.

\subsection{Evaluation on LLMs}
\label{sec:benchmarks}

Understanding the landscape of state-of-the-art (SOTA) LLMs requires not only a grasp of their architectural innovations, capabilities, and training paradigms, but also a clear evaluation of how these models perform across real-world tasks. While the previous section outlines the defining characteristics of leading LLMs—ranging from GPT-4.5 and Claude 3.7 Sonnet to open-source models like Llama 3 and DeepSeek-R1—this alone does not provide a full picture of their practical effectiveness. In this section, we shift focus to rigorous evaluation methods and benchmark results that quantify these models' strengths and trade-offs across diverse tasks such as reasoning, coding, multilingual understanding, and tool use. By linking architectural design with empirical performance, we aim to guide practitioners and researchers in making informed decisions when selecting LLMs for specific applications.

\subsubsection{Tasks}

The core function of LLMs is language modeling—predicting the next token based on the current input. This inherently requires both understanding and generating human language. Leveraging this foundational capability, LLMs can perform a wide variety of downstream tasks. We broadly categorize four key types of tasks that LLMs are capable of performing:

\noindent\textbf{Text Understanding}~\cite{devlin2019bert,liu2019roberta,yang2019xlnet}. Text understanding is a fundamental NLP task focused on identifying the intent, topic, or semantics of a given input, often within a long-context. It answers the question: "What is this text about?" This category includes several sub-tasks: (1) \textit{Sentiment detection} determines the writer's emotional tone—positive, negative, or neutral. For example, “I like this apple” expresses positivity, while “This is a total waste of time” conveys a negative sentiment. More nuanced categories, such as anger, joy, or frustration, can also be captured. (2) \textit{Information extraction} involves identifying specific entities or facts from text, such as names, dates, locations, or actions. From “Apple announced a new iPhone on March 15,” a model could extract “Apple” (organization), “iPhone” (product), and “March 15” (date). (3) \textit{Relationship understanding} tracks how entities are related across sentences. For instance, in “Sarah gave the book to Mike. He thanked her,” a model must infer that “he” refers to Mike and “her” refers to Sarah. (4) \textit{Summarization} reduces lengthy text to a concise version that preserves the main ideas. It may also simplify complex language or adjust the tone or style for different audiences.

\noindent\textbf{Text Generation}~\cite{keskar2019ctrl,brown2020language,raffel2020exploring}. Text generation refers to the ability of an LLM to produce coherent, fluent, and contextually appropriate text. This extends beyond stringing words together—it requires logic, relevance, and creativity. Key sub-tasks include: (1) \textit{Question answering}, which involves generating natural, complete answers based on a question and its context. This is essential in open-domain systems like digital assistants and educational tools. (2) \textit{Style transfer} rewrites text in a different tone or style while preserving its original meaning. For example, the formal sentence “I regret to inform you” might be rendered more casually as “Just a heads-up.” (3) \textit{Text completion} involves filling in or finishing partially written content. It powers autocomplete tools in emails, messaging apps, and writing assistants. (4) \textit{Machine translation} converts text from one language to another, not just literally, but with attention to grammar, idioms, and cultural nuance to preserve meaning and tone.

\noindent\textbf{Complex Reasoning}~\cite{kojima2022large,snell2024scaling,chen2021evaluating,ji2025test}. Complex reasoning involves deeper cognitive abilities such as logical inference, problem-solving, and structured thinking that go beyond simple pattern matching. Sub-tasks include: (1) \textit{Code generation}, where natural language instructions are translated into executable code. This allows users to generate scripts or programs using plain English descriptions. (2) \textit{Multi-step inference} requires synthesizing information across several logical steps. For example, answering “Which country hosted the Olympics after China in 2008?” requires knowing that China hosted in 2008 and the UK hosted in 2012. (3) \textit{Logical reasoning} tests the ability to apply deductive logic and identify valid conclusions or contradictions. A classic example: “If all cats are animals and some animals are black, can some cats be black?” (4) \textit{Commonsense reasoning} leverages everyday knowledge. For instance, given “He put the ice cream on the table in the sun,” the model should infer that the ice cream will melt.

\noindent\textbf{Knowledge Utilization}~\cite{lewis2020retrieval,gao2023retrieval,yang2023gpt4tools,xiong2024search}. Knowledge utilization refers to an LLM's ability to access, retrieve, and apply factual or procedural knowledge—either from internal memory or external sources—to solve tasks accurately. This includes: (1) \textit{Open-domain question answering}, where models retrieve and use up-to-date information to answer questions. For example, responding to “What are the current COVID-19 travel guidelines for Japan?” may require accessing recent data. (2) \textit{Tool-augmented reasoning} enhances LLM capabilities by integrating external tools such as calculators, databases, or code interpreters. For instance, to compute the square root of 41,324, a model may call a calculator tool. (3) \textit{Conversational search and retrieval} allows models to engage in interactive, multi-turn queries while dynamically retrieving and integrating relevant information. For example, answering “What are the side effects of this medication?” followed by “How does it compare to ibuprofen?” involves iterative search and context maintenance.

\subsubsection{Benchmarks}

\begin{table}[!t]
    \caption{Existing benchmarks on evaluating the capabilities of LLMs.}
    \label{tab:existing benchmarks}
    \resizebox{\linewidth}{!}{
        \begin{tabular}{cllll}
            \toprule
            \textbf{Tasks}                                   & \textbf{Subtask}                                            & \textbf{Relevant Benchmarks}                                        & \textbf{Domain}                & \textbf{Data Source}                  \\ \midrule
            \multirow{12}{*}{\textbf{Text   Understanding}}  & \multirow{3}{*}{\textbf{Sentiment Detection}}               & SST-2~\cite{socher-etal-2013-recursive}                             & Movie Reviews                  & Rotten Tomatoes                       \\
                                                             &                                                             & IMDB~\cite{maas-etal-2011-learning}                                 & Movie Reviews                  & IMDB                                  \\
                                                             &                                                             & Yelp Reviews~\cite{yelp2015}                                        & Business Reviews               & Yelp                                  \\ \cmidrule(l){2-5}
                                                             & \multirow{3}{*}{\textbf{Information   Extraction}}          & CoNLL-2003 (NER)~\cite{tjong-kim-sang-de-meulder-2003-introduction} & News Articles                  & Reuters Corpus                        \\
                                                             &                                                             & TACRED~\cite{zhang-etal-2017-position}                              & Various Domains                & News and Wikipedia                    \\
                                                             &                                                             & OpenIE~\cite{etzioni2008open}                                       & General Text                   & Web Sources                           \\
\bottomrule
        \end{tabular}
    }
\end{table}

As LLMs continue to advance, it becomes increasingly important to assess their capabilities across various domains, tasks, and reasoning skills. To evaluate how well LLMs perform in real-world scenarios, numerous benchmarks have been proposed across tasks such as mathematical reasoning, long-context understanding, and tool usage. We collect a selection of these benchmarks and categorize them by task type in Table~\ref{tab:existing benchmarks}. Below, we highlight several key benchmarks that are widely used to evaluate LLMs' abilities.

\begin{itemize}[leftmargin=10pt]
    \item \textbf{MMLU}~\cite{hendrycks2020measuring}: The Massive Multitask Language Understanding (MMLU) benchmark evaluates multitask accuracy across 57 diverse subjects, including humanities, social sciences, STEM fields, and professional domains like law and medicine. Each question is multiple-choice with four options, covering difficulty levels from elementary to professional. Questions are sourced from standardized test prep materials (e.g., GRE, USMLE) and university-level courses. The dataset comprises 15,908 questions split into training, validation, and test sets. MMLU assesses models in both zero-shot and few-shot settings, reflecting real-world conditions where no task-specific fine-tuning is applied. Human performance baselines are also provided, ranging from average crowdworkers to expert-level participants.

    \item \textbf{BIG-Bench}~\cite{srivastava2022beyond}: The Beyond the Imitation Game Benchmark (BIG-Bench) is a large-scale suite of 204 tasks designed to test LLMs on capabilities not captured by conventional benchmarks. Tasks span areas such as linguistics, mathematics, biology, social bias, and software engineering, and were contributed by researchers and institutions worldwide. Human experts also completed the tasks to establish reference baselines. BIG-Bench includes JSON tasks (with structured inputs/outputs) and programmatic tasks (which allow custom metrics and interaction). Evaluation metrics include accuracy, exact match, and calibration. A smaller curated subset, BIG-Bench Lite, contains 24 JSON tasks for lightweight and efficient evaluation.

    \item \textbf{HumanEval}~\cite{chen2021evaluating}: HumanEval is a benchmark for evaluating the functional correctness of code generation. It consists of 164 original Python programming problems, each with a function signature, descriptive docstring, and empty function body. A solution is deemed correct if it passes predefined unit tests, aligning with how developers assess code quality. The benchmark targets abilities such as comprehension, algorithmic reasoning, and basic mathematics. For safety, all code is executed in a secure sandbox to mitigate risks posed by untrusted or potentially harmful code.

    \item \textbf{TruthfulQA}~\cite{lin2021truthfulqa}: This benchmark is designed to assess whether LLMs generate truthful answers and avoid perpetuating misconceptions or factual inaccuracies. It includes 817 questions across 38 domains, such as health, finance, and law. The questions—typically concise, with a median length of 9 words—are crafted to exploit known weaknesses in LLMs, particularly their tendency to imitate common yet incorrect human text. The benchmark imposes rigorous truthfulness criteria, evaluating answers based on factual accuracy as supported by public sources like Wikipedia. Each question includes both true and false reference answers.

    \item \textbf{GSM8K}~\cite{cobbe2021training}: The Grade School Math 8K (GSM8K) dataset comprises 8.5K human-written arithmetic word problems suitable for gradeschool-level mathematics. Of these, 7.5K are training problems and 1K are test problems. Each problem typically requires 2 to 8 reasoning steps and involves basic arithmetic. The dataset emphasizes: (1) high quality, with a reported error rate below 2\%; (2) high diversity, avoiding repetitive templates and encouraging varied linguistic expression; (3) moderate difficulty, solvable using early algebra without advanced math concepts; and (4) natural language solutions, favoring everyday phrasing over formal math notation.
\end{itemize}

\subsubsection{Evaluation Methods}

Evaluating LLMs typically involves a combination of automatic and human-centered metrics, depending on the specific task at hand. We categorize existing evaluation methods into the following four classes:

\noindent\textbf{Basic Automatic Evaluation Metrics.}
Quantitative metrics are essential for assessing the performance of LLMs. These metrics vary based on the task type, such as classification, generation, or translation. For instance, in classification tasks, metrics like accuracy and F1-score are commonly used to compare predicted outputs with ground-truth labels. In contrast, for generative tasks such as question answering, metrics like BLEU (Bilingual Evaluation Understudy)~\cite{papineni2002bleu}, ROUGE (Recall-Oriented Understudy for Gisting Evaluation)~\cite{lin2004rouge}, and BERTScore~\cite{zhang2019bertscore} are employed to assess the semantic similarity between the generated and reference texts. Additionally, domain-specific evaluations exist. For example, in code generation, metrics like pass@k and functional correctness~\cite{chen2021evaluating} are utilized to measure the quality of the generated code.

\noindent\textbf{Advanced Automatic Evaluation Metrics Beyond Correctness. }
Beyond correctness, comprehensive evaluation of LLMs must account for broader behavioral attributes such as trustworthiness, toxicity, fairness, robustness, and reasoning quality~\cite{liang2022holistic,rauh2022characteristics,wang2023decodingtrust,cui2023fft,lightman2023let}. As LLMs are increasingly deployed in real-world applications—including education, healthcare, legal advising, and customer interaction—accuracy alone becomes an inadequate measure of performance. For instance, trustworthiness metrics assess whether models avoid hallucinations, misinformation, or unsupported claims~\cite{wang2023decodingtrust}, while toxicity evaluation captures the frequency of harmful, biased, or offensive content~\cite{cui2023fft}. Reasoning quality, on the other hand, examines whether a model can follow logically consistent, step-by-step problem-solving processes~\cite{lightman2023let}. Fairness metrics \cite{li2023survey} further ensure that model behavior remains equitable across diverse user groups, preventing systematic biases that could reinforce societal inequities. Additionally, robustness measures \cite{liu2023trustworthy} how stable model outputs are under adversarial or distributional shifts, and calibration evaluates whether a model's confidence levels meaningfully reflect its likelihood of being correct. In safety-critical or socially impactful scenarios, transparency and explainability become essential \cite{liao2023ai}, helping users understand, validate, or contest model outputs. These advanced metrics—often requiring dedicated datasets, probing techniques, or alignment with human values—offer a more holistic view of LLM capabilities and limitations. Moving beyond narrow accuracy-based benchmarks, they are essential for building LLMs that are not only powerful but also reliable, fair, and aligned with human expectations.

\noindent\textbf{Human Evaluation.}
Although automatic metrics offer scalability and efficiency, they often fail to capture nuanced qualities such as reasoning depth, factual accuracy, and practical utility~\cite{mishra2021cross}. Human evaluation addresses these gaps by involving human raters who assess LLM outputs for clarity, engagement, structure, and alignment with user intent~\cite{mishra2021cross}. This ensures higher reliability, safety, and real-world applicability. However, human evaluation is not without drawbacks: it is costly, time-consuming, and prone to variability due to subjective interpretations and potential rater bias. Moreover, it lacks scalability, making it infeasible for evaluating responses at a large scale.

\noindent\textbf{LLM-as-a-Judge.}
To overcome the limitations of human evaluation, the concept of using LLMs themselves as evaluators, LLM-as-a-judge~\cite{gu2024survey}, emerged. In this paradigm, human-aligned LLMs are employed to replace human raters~\cite{zheng2023judging,li2024generation}. A typical approach involves prompting the LLM to compare and rank two candidate answers. This method is more scalable, faster, and generally more cost-effective than relying on human reviewers, despite the computational costs of querying LLMs. Additionally, it allows for real-time feedback. However, this approach still faces several challenges, such as susceptibility to prompt sensitivity, potential biases in judgment, and the risk of hallucinations in comparative reasoning.

\subsubsection{Performance at a Glance}

In this section, we provide a performance-at-a-glance synthesis of selected SOTA LLM models. Table~\ref{tab:leaderboard-base} and Table~\ref{tab:leaderboard-reasoner}
report the performance of selected LLMs across a range of benchmarks. Table~\ref{tab:leaderboard-base} presents results on widely adopted basic benchmarks, while Table~\ref{tab:leaderboard-reasoner} highlights performance on more challenging reasoning tasks. The results are aggregated from both Chatbot Arena\footnote{\url{https://lmarena.ai/}} and the Vellum LLM Leaderboard\footnote{\url{https://www.vellum.ai/llm-leaderboard}}. We apply different colors to indicate the top-3 performances per benchmark. The selected evaluations cover diverse capabilities, including commonsense reasoning, code generation, tool use, multilingual understanding, and mathematical problem-solving, providing a holistic view of current LLM competence.

\begin{table}[!h]
    \centering
    \caption{The performance of popular LLMs on widely used benchmarks. We use different colors to indicate \colorbox{Blue4}{best}, \colorbox{Blue3}{second-best}, and \colorbox{Blue1}{third-best} performance. The first column is from \href{https://lmarena.ai/}{Chatbot Arena} and the remaining columns are from \href{https://www.vellum.ai/llm-leaderboard}{Vellum LLM leaderboard}.}
    \label{tab:leaderboard-base}
    \resizebox{0.9\linewidth}{!}{
        \begin{tabular}{lccccccc|c}
            \toprule
                                             & \textbf{Chatbot}      & \textbf{MMLU}            & \textbf{GPQA}            & \textbf{HumanEval}       & \textbf{Math}            & \textbf{BFCL}            & \textbf{MGSM}            & \textbf{Input Cost} \\
            \multirow{-2}{*}{\textbf{Model}} & \textbf{Arena}        & \textbf{(General)}       & \textbf{(Reasoning)}     & \textbf{(Coding)}        & \textbf{}                & \textbf{(Tool Use)}      & \textbf{(Multilingual)}  & \textbf{/ 1M Token} \\ \midrule
            Qwen 2.5                         & 1296                  & 70.20\%                  & 49\%                     & 88\%                     & 85\%                     & 61.31\%                  & -                        & -                  \\
            Grok 2                           & 1288                  & 87.50\%                  & 56\%                     & 88.40\%                  & 76.10\%                  & -                        & -                        & \$2.00              \\
            LLAMA 3.3                        & 1294                  & 86\%                     & 48\%                     & 88.40\%                  & 77\%                     & 77.50\%                  & \cellcolor{Blue3}91.10\% & -              \\
            DeepSeek V3                      & 1318                  & 88.50\%                  & 59.10\%                  & 82.60\%                  & 90.20\%                  & 57.23\%                  & 79.80\%                  & \$0.27              \\
            DeepSeek R1                      & \cellcolor{Blue1}1360 & \cellcolor{Blue3}90.80\% & \cellcolor{Blue3}71.50\% & -                        & \cellcolor{Blue3}97.30\% & -                        & -                        & \$0.55              \\
\bottomrule
        \end{tabular}
    }
\end{table}

\begin{table}[!h]
    \centering
    \caption{The performance of popular reasoning LLMs on challenging benchmarks. We use colors to indicate \colorbox{Blue4}{best}, \colorbox{Blue3}{second-best}, and \colorbox{Blue1}{third-best} performance. The first column is from \href{https://lmarena.ai/}{Chatbot Arena} and the remaining columns are from \href{https://www.vellum.ai/llm-leaderboard}{Vellum LLM leaderboard}.}
    \label{tab:leaderboard-reasoner}
    \resizebox{\linewidth}{!}{
        \begin{tabular}{lcccccccccc|c}
            \toprule
                                             & \textbf{Chatbot}      & \textbf{GPQA Diamond}    & \textbf{SWE-bench}       & \textbf{Tool Use}        & \textbf{Tool Use}        & \textbf{MMMLU}           & \textbf{MMMU}            & \textbf{IFEval}          & \textbf{MATH 500}        & \textbf{AIME 2024}       & \textbf{Input Cost} \\
            \multirow{-2}{*}{\textbf{Model}} & \textbf{Arena}        & \textbf{(Reasoning)}     & \textbf{(Agent coding)}  & \textbf{(Retail)}        & \textbf{(Airline)}       & \textbf{(Multilingual)}  & \textbf{(Visual)}        & \textbf{}                & \textbf{}                & \textbf{(Math)}          & \textbf{/ 1M Token} \\ \midrule
            DeepSeek R1                      & \cellcolor{Blue3}1360 & 71.50\%                  & \cellcolor{Blue1}49.20\% & -                        & -                        & -                        & -                        & 83.30\%                  & \cellcolor{Blue3}97.30\% & 79.80\%                  & \$0.55              \\
            Claude 3.5 Sonnet                & 1283                  & 65.00\%                  & 49.00\%                  & \cellcolor{Blue1}71.50\% & \cellcolor{Blue1}48.80\% & 82.10\%                  & 70.40\%                  & \cellcolor{Blue1}90.20\% & 78.00\%                  & 16.00\%                  & \$3.00              \\
            Claude 3.7 Sonnet                & 1296                  & 68.00\%                  & \cellcolor{Blue4}70.30\% & \cellcolor{Blue4}81.20\% & \cellcolor{Blue4}58.40\% & \cellcolor{Blue1}83.20\% & 71.80\%                  & \cellcolor{Blue3}90.80\% & 82.20\%                  & 23.30\%                  & \$3.00              \\
\bottomrule
        \end{tabular}
    }
\end{table}

From Table~\ref{tab:leaderboard-base}, models such as GPT-4.5, GPT-4o, and DeepSeek R1 exhibit strong and consistent performance across a broad array of tasks. Notably, GPT-4.5 sets the state of the art on HumanEval (92.40\%) and MATH (96.40\%), while also remaining competitive on MMLU. DeepSeek R1 places in the top three on MMLU (90.80\%), GPQA (71.50\%), and MATH (97.30\%), underscoring its strength in both reasoning and quantitative domains. Additionally, GPT-o1 and GPT-o3-mini deliver compelling results in mathematical reasoning (e.g., MATH: 97.90\%) and multilingual understanding (MGSM: up to 92.00\%). Table~\ref{tab:leaderboard-reasoner} further differentiates models based on their performance on advanced reasoning tasks. OpenAI o1 and Grok 3 Beta attain competitive scores on GPQA Diamond, MATH 500, and AIME 2024, reflecting solid reasoning abilities. Notably, Claude 3.7 Sonnet (reasoner) achieves leading results in tool use (Airline: 86.10\%), IF-Eval (93.20\%), and MATH 500 (96.20\%), highlighting its proficiency in multi-step reasoning and task completion. DeepSeek R1 continues to demonstrate top-tier performance, maintaining high scores on MATH 500 (97.30\%) and AIME 2024 (79.80\%).

From the foregoing analysis of LLM performance across a range of benchmarks, we draw the following observations:

\begin{itemize}[leftmargin=10pt]

\item \textbf{Insight 1: No single LLM dominates across all tasks.} GPT-4.5 ranks highest on Chatbot Arena (1398), suggesting strong general chatbot capabilities. However, it does not lead in benchmarks like GPQA (reasoning) or math. Conversely, DeepSeek R1 achieves the best scores on MMLU (90.8\%) and Math (97.3\%), but shows weaker results in HumanEval (coding) and multilingual tasks, indicating that top performance in one domain does not translate across all domains.

\item \textbf{Insight 2: Reasoning models outperform others in logical and structured tasks.} Claude 3.7 Sonnet (reasoner) achieves the highest GPQA Diamond score (84.8\%), the best Tool Use (Retail) result (81.2\%), and leads in MMMLU (86.1\%). These benchmarks emphasize reasoning and complex task execution, showcasing the value of models fine-tuned for reasoning.

\item \textbf{Insight 3: Specialized models often come with trade-offs.} Claude 3.5 Haiku performs well in general chatbot interaction but has one of the lowest GPQA scores (41.6\%) and math scores (69.4\%). Gemini 1.5 Pro and Gemini 2.0 Flash perform reasonably well in multilingual (MGSM) and tool-use (BFCL) tasks, but underperform in reasoning and coding, highlighting performance trade-offs between general-purpose and specialized capabilities.

\item \textbf{Insight 4:  Different models excel at different tasks, and should be selected accordingly.} For chatbot dialogue, GPT-4.5 and GPT-4o are top performers. Claude 3.7 Sonnet (reasoner) excels in reasoning, tool use, and multilingual benchmarks. Claude 3.5 Sonnet leads in coding with the best HumanEval score (93.7\%). For math-heavy benchmarks, DeepSeek R1 and GPT-03-mini both surpass 97\% on the Math benchmark. Thus, model selection should be guided by the specific task requirements.
    
\end{itemize}

\newpage
\section{LLMs for Arts, Letters, and Law}
\label{sec:discpline-1}

In this chapter, we explore how LLMs are reshaping the humanities and law, shifting emphasis from evidence to application. Specifically, we review five disciplines: history, philosophy, political science, arts and architecture, and law. In \textbf{history}, we address narrative and interpretive practices (e.g., story generation and analysis), quantitative and scientific methods (e.g., modeling historical psychological responses), as well as interdisciplinary and comparative approaches, supported by benchmarks and brief commentary. In \textbf{philosophy}, we consider normative and interpretive applications (e.g., generating debate or dialogue), analytical and logical domains (e.g., symbol grounding diagnostics), along with comparative and cross-disciplinary analyses, again connected to benchmark studies. In \textbf{political science}, we investigate text-based methods for extracting policy insights, simulating and forecasting opinions, and shaping and framing political messaging, while linking these to benchmark assessments and reflective discussion. In \textbf{arts and architecture}, we present model-enabled creation across the visual, literary, and performing arts, alongside LLM-supported architectural design, production, and analysis, followed by evaluations and key lessons. Finally, in \textbf{law}, we examine LLM use in consultant-style question answering, drafting contracts and briefs, parsing and analyzing legal documents and cases, and predicting judgments, concluding in representative benchmarks and discussion.

\subsection{History}

\subsubsection{Overview}

\begin{figure}[!t]
    \centering
    \includegraphics[width=\linewidth]{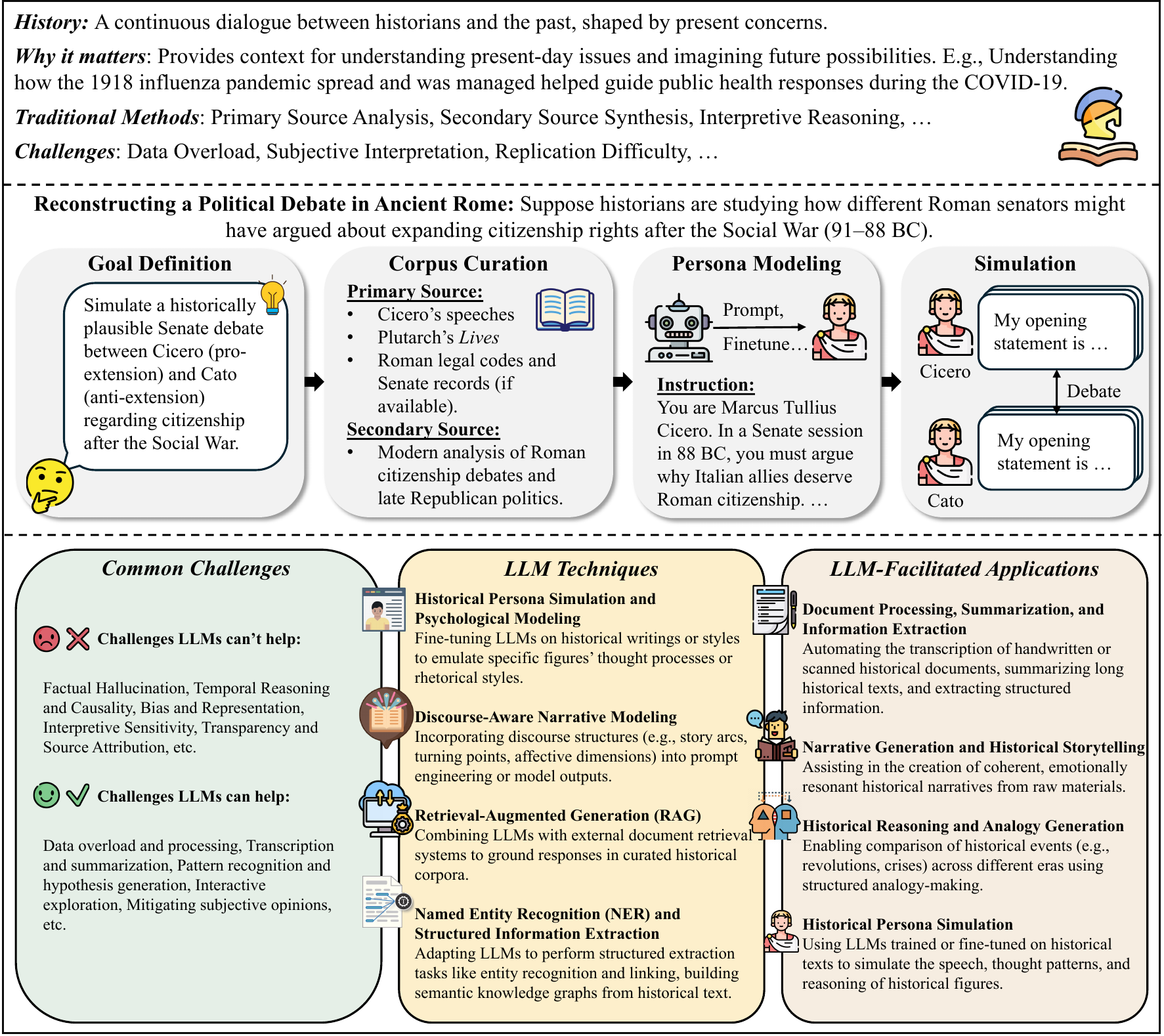}
    \caption{The history research in the era of LLMs.}
    \label{fig:history_framework}
\end{figure}

\noindent\textbf{Introduction.}
History is a continuous process of interaction between the historian and his facts, an unending dialogue between the present and the past \cite{what_is_history}. It helps us understand how the world has changed over time and provides valuable context for interpreting the present and imagining the future \cite{tosh2015pursuit,williams2014historian}. At its core, history is about reconstructing narratives, analyzing cause and effect, and interpreting the motivations and consequences of human actions across time \cite{jenkins2003rethinking,gaddis2002landscape}.

Traditionally, history research involves careful examination of primary sources—such as letters, legal documents, newspaper articles, and government records—as well as secondary analyses written by other historians \cite{howell2001reliable}. Scholars use these materials to craft explanations of past events, build timelines, uncover social patterns, and offer interpretations grounded in context \cite{marwick2001new,lloyd1993structures}. This process relies heavily on human reading, note-taking, and interpretive reasoning \cite{wineburg2010historical}.

However, traditional methods face growing limitations. First, the volume of historical data—digitized archives, scanned manuscripts, oral histories, and digital media—has grown far beyond what any individual or team can process manually \cite{jockers2013macroanalysis,moretti2013distant,milligan2019history}. Second, interpreting history often requires synthesizing multiple perspectives, which can be slow and subjective \cite{blevins2021paper}. Third, historical research is time- and labor-intensive, making it difficult to scale or replicate. These challenges have prompted interest in computational tools that can support, accelerate, or expand historical inquiry.

\noindent\textbf{The role of LLMs.} LLMs offer a powerful new toolkit for historical research. Trained on vast corpora of text, LLMs can read, summarize, translate, and generate human-like text at scale. They can identify patterns across large document collections, extract names and dates, simulate alternative narratives, or respond to questions using knowledge from multiple sources. These capabilities make LLMs especially suited for analyzing unstructured historical text, processing archival documents, and enabling interactive exploration of the past.

LLMs can assist historians in a variety of ways: by automating the transcription of handwritten documents, summarizing long articles or books, clustering related texts, or generating hypotheses about social or political dynamics. They can also support the creation of historical simulations or dialogue systems, enabling new forms of engagement with historical knowledge. Researchers have begun to explore LLM-based systems for historical thinking, comparative analysis, and even the modeling of historical psychology.

\noindent\textbf{Limitations of LLMs.}
Despite their promise, LLMs face important limitations when applied to historical research: \textit{Factual Hallucination:} LLMs may generate plausible-sounding but false or unverifiable historical claims, which undermines academic rigor and trust. \textit{Temporal Reasoning:} LLMs often struggle with chronology, causality, and contextual nuance—essential features of historical reasoning. \textit{Bias and Representation:} Because LLMs reflect the biases in their training data, they may reproduce skewed or incomplete views of history, overlooking marginalized voices or reinforcing dominant narratives. \textit{Interpretive Sensitivity:} History is not just about facts but about interpretation. LLMs may oversimplify or flatten complex debates by offering overly confident or decontextualized summaries. \textit{Transparency and Source Attribution:} LLMs generally do not cite specific sources, making it difficult for historians to verify information or trace interpretive lineage.

As such, LLMs should not be seen as replacements for human historians, but rather as tools that can extend their reach, suggest new questions, and assist in data exploration—especially when paired with domain expertise and critical oversight.

\noindent\textbf{Taxonomy.}
To better understand the potential of LLMs in history research, we organize the field into three broad categories, based on methodological approaches and research goals:

\begin{table}[!t]
    \centering
    \caption{Applications and insights of LLMs in historical research}
    \resizebox{\linewidth}{!}{
        \begin{tabular}{@{}p{3.5cm} p{4.5cm} p{6.5cm} p{6.5cm} p{2cm}@{}}
            \toprule
            \textbf{Historical Category} & \textbf{LLM Application Areas}                                                                                                            & \textbf{Use Case-Inspired Research Question} & \textbf{Key Insights and Contributions} & \textbf{References} \\
            \midrule

            Narrative and Interpretive History
                                            & Narrative Generation and Analysis
                                            & Can LLMs generate historically coherent and emotionally resonant narratives from primary source texts?
                                            & LLMs struggle with coherence and diversity; discourse-aware prompts improve storytelling; useful for story arcs and emotional analysis.
                                            & \cite{tian2024large, piper-bagga-2024-using}                                                                                                                                                                                                            \\\cmidrule{2-5}

                                            & Historical Research Assistance
                                            & How can LLMs simulate historical personas or provide conversational access to archives for education and research?
                                            & Used in chat-based exploration, simulated historical figures, and layered tools (e.g., KleioGPT); helpful for both scholars and students.
                                            & \cite{varnum2024large, zeng2024histolens, garcia2023if}                                                                                                                                                                                                 \\\cmidrule{2-5}

                                            & Historical Interpretation
                                            & Can LLMs improve consistency and reduce bias in interpretive historical analysis?
                                            & LLMs show more consistent judgments than humans; useful for reducing subjectivity in analysis.
                                            & \cite{celli2025language}                                                                                                                                                                                                                                \\\midrule

            Quantitative and Scientific History
                                            & Historical Thinking
                                            & How do LLMs enhance access to and reflection on historical content through transcription and summarization?
                                            & Help transcribe handwriting and summarize texts; make primary sources easier to explore and reflect on.
                                            & \cite{blevins2024}                                                                                                                                                                                                                                      \\\cmidrule{2-5}

                                            & Historical Data Processing
                                            & Can LLMs scale up entity extraction and temporal reasoning across vast historical corpora?
                                            & Enable name/date extraction, timeline generation, and semantic linking at scale.
                                            & \cite{hauser2024large, celli2024knowledge}                                                                                                                                                                                                              \\\cmidrule{2-5}

                                            & Simulating Historical Psychological Responses
                                            & Is it possible to model the psychological tendencies of past societies using LLMs trained on historical texts?
                                            & Can mimic cultural mindsets from historical texts, but results may reflect biases of elite sources.
                                            & \cite{varnum2024large}                                                                                                                                                                                                                                  \\\midrule

            Comparative and Cross-Disciplinary History
                                            & Historical Analogy Generation
                                            & How can LLMs retrieve or generate meaningful analogies between past and present events?
                                            & Generate cross-era comparisons; reduce hallucinations with structured frameworks and similarity checks.
                                            & \cite{li2024past}                                                                                                                                                                                                                                       \\\cmidrule{2-5}

                                            & Interdisciplinary Information Seeking
                                            & Can LLMs facilitate the discovery of relevant insights across disciplines for historical inquiry?
                                            & Tools like DiscipLink help explore diverse sources and connect ideas across fields.
                                            & \cite{zheng2024disciplink}                                                                                                                                                                                                                              \\
            \bottomrule
        \end{tabular}
    }
    \label{tab:history}
\end{table}

\begin{itemize}[leftmargin=10pt]
      \item \textbf{Narrative and Interpretive History:}
            This area emphasizes descriptive, subjective, and human-centered accounts of the past. It uses storytelling and meaning-making to explain events in context. LLMs can assist in narrative generation, reconstruct voices from historical texts, and interpret language use in personal accounts or memoirs.

      \item \textbf{Quantitative and Scientific History:}
            This approach applies statistical, computational, and formal methods to study historical data and trends. LLMs can process large datasets, simulate historical psychological responses, aid in historical reasoning tasks, and evaluate knowledge systems.

      \item \textbf{Comparative and Cross-Disciplinary History:}
            This domain integrates methods from sociology, economics, political science, and other disciplines to study similarities and differences across historical contexts. LLMs can Support comparative analysis, generate historical analogies, link concepts across eras or cultures.
\end{itemize}

Our taxonomy aligns with the framework outlined in reference \cite{green1999houses}, though it uses different terminology. It is conceptually consistent with how scholars commonly categorize historical research: narrative and interpretive history corresponds to traditional, humanistic approaches; quantitative and scientific history reflects data-driven, computational, and social-scientific methods; and comparative and cross-disciplinary history encompasses global, integrative, and hybrid perspectives.

\subsubsection{Narrative and Interpretive History}

Narrative and interpretive history refers to the study of the past through storytelling and critical interpretation, focusing on the meanings, motives, and lived experiences behind historical events. This method is often what people first imagine when they think of history: a crafted narrative that explains what happened, why it happened, and how it was experienced. Rather than just listing facts or dates, this approach aims to understand how individuals and societies made sense of their world. Historians working in this tradition analyze sources like letters, speeches, and cultural artifacts to reconstruct events with nuance, paying close attention to human intentions, emotions, and consequences. It is about weaving the past into coherent stories that connect with readers and shed light on the complexity of human life.

\noindent\textbf{Narrative Generation and Analysis.}
Recent work has critically examined the narrative generation and comprehension capacities of LLMs. \cite{tian2024large} introduce a comprehensive computational framework that evaluates LLM storytelling through three discourse-level aspects: story arcs (macro), turning points (meso), and affective dimensions (micro). Their analysis reveals substantial discrepancies between human- and machine-generated narratives: LLM outputs tend to be homogeneously positive, poorly paced, and lacking in suspense and diversity. Benchmarking on tasks such as story arc and turning point identification further confirms that models like GPT-4 and Claude underperform relative to humans in discourse-level narrative reasoning. However, the authors show that explicit integration of discourse structures into prompting strategies significantly improves storytelling outcomes, including greater emotional engagement and narrative diversity. Complementing this, the WNU 1.4 benchmark \cite{piper-bagga-2024-using} offers a cross-lingual, multi-task diagnostic tool to assess narrative understanding in LLMs. It reveals deficiencies in narrative coherence and character goal modeling, especially in low-resource languages. Together, these studies advocate for discourse-aware approaches in both evaluation and generation, marking a crucial step toward more human-like narrative competence in LLMs.

\noindent\textbf{Historical Research Assistance.}
Recent research also integrates LLMs into historical inquiry, proposing new methodologies that enhance both analysis and pedagogy. Varnum et al.~\cite{varnum2024large} advocate for using LLMs trained on historical texts to simulate responses of historical figures and generate contextualized insights, revealing both the interpretive potential and epistemological tensions inherent in such applications. Zeng~\cite{zeng2024histolens} introduces \textit{HistoLens}, an LLM-powered, multi-layered framework for historical text analysis, demonstrated via the Confucian-Legalist debates in the Western Han dynasty's \textit{Yantie Lun}. This framework combines thematic word frequency analysis, named entity recognition, knowledge graph construction, GIS-based spatial mapping, and machine teaching to construct interpretative pipelines that blend traditional scholarship with computational scalability. Complementing these innovations, Gonzalez Garcia and Weilbach~\cite{garcia2023if} present \textit{KleioGPT}, a retrieval-augmented conversational interface that enables historians to interact with curated corpora through natural language prompts. Their evaluation shows LLMs can assist with complex historiographic tasks, such as question answering and data extraction, while also underscoring challenges like hallucination and source attribution. Together, these works signal a paradigm shift in historical methodology, foregrounding LLMs as both analytical tools and collaborators in the humanities.

\noindent\textbf{Historical Interpretation.}
In addition, LLMs have extended their application to the domain of historical reasoning, focusing on interpretative consistency. Celli and Spathoulas \cite{celli2025language} empirically examine interpretative agreement between humans and LLMs over historical annotations using two cyclical theories—Structural Demographic Theory and the Big Cycle model. They find that LLMs, particularly large-scale models like GPT-4 and Claude 3.5, achieve significantly higher inter-annotator agreement than humans, suggesting a promising role for LLMs in minimizing subjective bias in historical analysis. Together, these studies reveal that LLMs not only possess the capacity to generate meaningful historical analogies but also exhibit interpretive consistency that may surpass human annotators, raising both opportunities and concerns for digital humanities and historical scholarship.

\subsubsection{Quantitative and Scientific History}

Quantitative and scientific history applies statistical, mathematical, and computational techniques to historical data in order to uncover patterns, test hypotheses, and make systematic comparisons. This method treats history a bit like a science experiment. Instead of focusing solely on individuals or events, it looks at big-picture trends—such as population changes, economic growth, or voting behavior—often using data like census records, trade statistics, or social surveys. It helps answer questions like “How did literacy rates change over a century?” or “What economic factors were linked to revolutions?” Thanks to digital tools and vast data archives, historians can now analyze much more information than before, revealing trends that would be impossible to spot just by reading documents one by one.

\noindent\textbf{Historical Thinking.}
The article \cite{blevins2024} by Cameron Blevins explores the use of LLMs like ChatGPT in historical research, particularly in transcribing and interpreting primary sources. LLMs can accurately transcribe handwritten historical documents, which were traditionally considered “machine-unreadable.” The article provides an example of an 1886 letter transcribed by ChatGPT with impressive accuracy. These tools can generate concise and accurate summaries of historical texts, as demonstrated with Benjamin Curtis's letter. However, errors can occur, especially with complex documents or non-English sources. A minor transcription mistake in the Curtis letter highlights how a lack of historical context can lead to missed nuances. Thus, LLMs can serve as research assistants, helping with tedious tasks like transcription while refining historians' own thoughts and interpretations.

\noindent\textbf{Historical Data Processing.}
Recent advancements in the processing of historical data through LLMs have demonstrated significant potential in enhancing the accuracy and efficiency of information retrieval and analysis. The work \cite{hauser2024large} introduces a framework for the intelligent extraction and visualization of Indonesian historical narratives using transformer-based LLMs. By integrating Named Entity Recognition (NER) and text classification models, the system efficiently identifies and links key entities and events within historical texts, offering an enriched semantic representation of historical knowledge. Complementarily, the comprehensive survey \cite{celli2024knowledge} categorizes and evaluates the capabilities of LLMs across diverse tasks, highlighting their utility in temporal reasoning, fact-checking, and structured data understanding, which are pivotal in historical research. Together, these studies underscore the transformative impact of LLMs on historical data processing, enabling more nuanced analysis, enhanced data curation, and scalable knowledge extraction from complex archival sources.

\noindent\textbf{Simulating Historical Psychological Responses.}
The integration of historical LLMs into behavioral science marks a novel methodological advance, enabling the simulation of psychological responses from temporally distant populations. Varnum et al. \cite{varnum2024large} argue that by training generative language models on corpora of historical texts—ranging from letters and fiction to scholarly works—it is possible to approximate the mental frameworks of historical societies. This approach allows researchers to transcend the temporal constraints of traditional psychological methods, which are inherently limited to living participants, and to infer the psychology of past cultures with greater fidelity than indirect proxies such as word frequency analyses. HLLMs could facilitate empirical investigations into cultural change, test the generalizability of psychological theories across time, and enrich historical scholarship with quantifiable insights. Despite promising early applications, such as MonadGPT and XunziALLM, the authors note significant challenges, including limited and elite-skewed training data, difficulties in benchmarking historical accuracy, and the risk of anachronistic biases. Nonetheless, with careful curation and interdisciplinary validation, HLLMs hold the potential to illuminate the cognitive landscapes of bygone eras and serve as valuable tools in both psychology and the historical sciences.

\subsubsection{Comparative and Cross-Disciplinary History}

Comparative and cross-disciplinary history involves analyzing historical phenomena across different cultures, regions, or time periods, often incorporating insights from other disciplines like sociology, anthropology, or political science. This method steps back to take a wide-angle view. Instead of looking at a single nation or event, comparative history puts multiple cases side by side to ask questions like, “Why did one country experience a revolution while another did not?” or “How did different empires manage trade or power?” Cross-disciplinary approaches further enrich this analysis by borrowing theories and methods from other fields, allowing historians to explore issues like class, race, environment, or technology with more depth. It's a powerful way to understand not just what happened, but why it happened in some places and not others.

\noindent\textbf{Historical Analogy Generation.} Li et al. \cite{li2024past} introduce the novel task of \textit{historical analogy acquisition}, which seeks to identify past events that are analogous to contemporary ones across multiple cognitive dimensions—topic, background, process, and result. They propose both retrieval-based and generative approaches, and notably develop a self-reflection framework to mitigate hallucinations and biases in free-form generation. Their automatic multi-dimensional similarity metric demonstrates high alignment with human evaluations, affirming the potential of LLMs in historical analogy tasks.

\noindent\textbf{Interdisciplinary Information Seeking.}
Interdisciplinary historical research necessitates traversing diverse disciplinary landscapes, often hindered by scattered knowledge, domain-specific terminologies, and cognitive load in assimilating unfamiliar perspectives. The \textit{DiscipLink} \cite{zheng2024disciplink} system exemplifies how LLMs can be harnessed to support such complex information-seeking endeavors through a human-AI co-exploration paradigm. Integrating mixed-initiative workflows, DiscipLink scaffolds the nonlinear process of interdisciplinary information seeking (IIS)—comprising orientation, opening, and consolidation—by eliciting exploratory questions (EQs) tailored to users' research goals, expanding search queries with domain-specific vocabulary, and synthesizing retrieved literature into contextualized themes. The system's design foregrounds researcher agency, allowing iterative engagement with LLM-generated suggestions while mitigating the risks of hallucination and bias. Empirical evaluations, including usability studies and case analyses, underscore its utility in helping scholars—especially those in fields like history—to navigate and integrate knowledge from adjacent disciplines such as sociology, education, and psychology. This work demonstrates the potential of LLMs not as autonomous agents but as collaborative partners in exploratory, interdisciplinary historical research.

\subsubsection{Benchmarks}

\begin{table*}[!t]
    \centering
    \caption{Overview of Benchmarks for Evaluating LLMs on Historical Data}
    \label{tab:history-benchmarks}
    \resizebox{\linewidth}{!}{
          \begin{tabular}{lp{3.5cm}p{4.2cm}p{3.8cm}p{3.5cm}}
                \toprule
                \textbf{Benchmark}                          & \textbf{Scope and Focus}                                     & \textbf{Data Composition}                                                                                                                             & \textbf{Evaluation Tasks}                                            & \textbf{Key Insights}                                                                                                                     \\ \midrule
                \textbf{TimeTravel} \cite{ghaboura2025time} & Multimodal evaluation of historical and cultural artifacts   & 10{,}250 expert-verified samples across 266 cultures and 10 regions; includes manuscripts, artworks, inscriptions, and archaeological findings        & Classification, interpretation, and historical reasoning             & Highlights LMMs' limitations in cultural/historical context understanding; sets new standards for AI in cultural heritage preservation    \\ \midrule
                \textbf{AC-EVAL} \cite{wei2024ac}           & Ancient Chinese language understanding by LLMs               & 3{,}245 multiple-choice questions on historical facts, geography, customs, classical poetry, and philosophy, categorized into three difficulty levels & General knowledge, short text understanding, long text comprehension & Chinese-trained models outperform English-trained ones; ancient Chinese remains a low-resource challenge; few-shot often introduces noise \\ \midrule
                \textbf{Hist-LLM} \cite{hauser2024large}    & Global historical knowledge evaluation using structured data & Subset of Seshat Global History Databank: 600 societies, 36,000 data points                                                                           & Multiple-choice on historical facts across global regions and eras   & Models show moderate accuracy; better on early periods and the Americas; weaker in Sub-Saharan Africa and Oceania                         \\
                \bottomrule
          \end{tabular}
    }
\end{table*}

\textbf{TimeTravel} \cite{ghaboura2025time} is a benchmark designed to evaluate large multimodal models (LMMs) on historical and cultural artifacts. It consists of 10,250 expert-verified samples spanning 266 cultures across 10 major historical regions. Unlike existing benchmarks that focus on modern objects and landmarks, TimeTravel prioritizes historical knowledge, contextual reasoning, and cultural preservation, providing structured datasets for AI-driven analysis of manuscripts, artworks, inscriptions, and archaeological findings. The benchmark enables the assessment of AI models in classification, interpretation, and historical comprehension, helping researchers identify strengths and limitations. TimeTravel sets a new standard for evaluating AI in cultural heritage preservation and historical discovery, with results demonstrating gaps in current AI capabilities while providing a foundation for future improvements.

\textbf{AC-EVAL} \cite{wei2024ac} is a benchmark designed to evaluate LLMs' understanding of ancient Chinese, addressing gaps in existing benchmarks that primarily focus on modern Chinese. It consists of 3,245 multiple-choice questions spanning historical facts, geography, social customs, philosophy, classical poetry, and prose, structured into three difficulty levels: general historical knowledge, short text understanding, and long text comprehension. The evaluation reveals that Chinese-trained LLMs outperform English-trained ones, highlighting ancient Chinese as a low-resource challenge for models like GPT-4. While chain-of-thought reasoning enhances performance in larger models, few-shot learning often introduces noise rather than improving accuracy. AC-EVAL provides a structured and comprehensive framework for assessing LLMs' proficiency in ancient Chinese, aiming to advance their application in language education and scholarly research.

\textbf{HiST-LLM} \cite{hauser2024large} is a benchmark for evaluating LLMs' historical knowledge using a subset of the Seshat Global History Databank. This dataset, covering 600 historical societies and 36,000 data points, enables the assessment of LLMs across various world regions and time periods. Seven models from OpenAI, Llama, and Gemini families were tested using multiple-choice questions on historical facts, showing accuracy ranging from 33.6\% to 46\%, outperforming random guessing but falling short of expert comprehension. The results indicate that models perform better on earlier historical periods and exhibit regional discrepancies, with stronger performance for the Americas and weaker results for Sub-Saharan Africa and Oceania. While LLMs demonstrate some expert-level historical knowledge, the benchmark reveals significant gaps and opportunities for improvement.

\subsubsection{Discussion}

\noindent\textbf{Opportunities and Impact.}
The integration of LLMs into historical research marks a profound shift in how the past can be explored, analyzed, and understood. As demonstrated across narrative and interpretive history~\cite{tian2024large, varnum2024large, celli2025language}, quantitative and scientific history~\cite{blevins2024, hauser2024large, varnum2024large}, and comparative and cross-disciplinary history~\cite{li2024past, zheng2024disciplink}, LLMs offer unprecedented capabilities for processing vast corpora of historical texts, assisting with complex interpretive tasks, and supporting interdisciplinary inquiries.

By automating labor-intensive processes such as transcription~\cite{blevins2024}, entity recognition~\cite{hauser2024large}, and analogical reasoning~\cite{li2024past}, LLMs can significantly augment the productivity of historians and open new avenues for exploring historical narratives and mentalities. They also enable novel methodologies, such as historical psychology simulation~\cite{varnum2024large} and interdisciplinary information retrieval~\cite{zheng2024disciplink}, bridging gaps between traditional humanities approaches and computational scalability. In doing so, LLMs have the potential to democratize access to historical research, enhance interpretive diversity, and foster richer engagements with the past.

\noindent\textbf{Challenges and Limitations.}
Despite their transformative potential, LLMs present serious limitations when applied to historical inquiry. Factual hallucination remains a critical risk~\cite{varnum2024large, garcia2023if}, undermining trust and academic rigor, especially when models generate plausible but unverified historical claims. Temporal reasoning deficiencies~\cite{tian2024large} hinder models' ability to understand chronology, causality, and contextual nuance—core competencies in historical scholarship.

Bias amplification is another major concern~\cite{celli2025language}. LLMs inherit the systemic biases present in their training data, which can lead to the overrepresentation of dominant historical narratives while marginalizing underrepresented perspectives. Moreover, the interpretive nature of history poses challenges: while LLMs can simulate interpretative consistency~\cite{celli2025language}, they may oversimplify complex historiographic debates or falsely present contested interpretations as settled facts.

Transparency and source attribution issues also persist~\cite{garcia2023if}. Without clear references to underlying sources, it becomes difficult for historians to verify claims, trace intellectual lineage, or engage in critical evaluation. Furthermore, the limited historical coverage and elite bias of available training corpora~\cite{varnum2024large} constrain the representativeness of LLM-assisted insights, particularly for studies focused on marginalized or non-Western histories.

\noindent\textbf{Research Directions.}
Building on current advances, several promising research directions emerge:

\begin{itemize}[leftmargin=10pt]
    \item \textbf{Historical Fine-Tuning and Corpus Curation.} Developing domain-specific LLMs fine-tuned on curated historical corpora—including diverse and marginalized sources—can improve factual reliability, bias mitigation, and interpretive depth~\cite{varnum2024large, zeng2024histolens}.
    
    \item \textbf{Temporal and Causal Reasoning Enhancement.} Advances in temporal modeling, causal inference, and discourse-level narrative comprehension~\cite{tian2024large} are critical for enabling LLMs to reason more accurately about historical sequences and cause-effect relationships.
    
    \item \textbf{Explainable and Source-Grounded Historical AI.} Building models that produce verifiable outputs, grounded in cited historical documents or structured datasets, can strengthen academic trust and facilitate critical engagement~\cite{garcia2023if}.
    
    \item \textbf{Collaborative Human-AI Historical Research.} Systems like \textit{DiscipLink}~\cite{zheng2024disciplink} suggest a promising future where LLMs act as exploratory partners rather than authoritative experts, supporting iterative, mixed-initiative workflows that preserve human scholarly agency.
    
    \item \textbf{Ethics and Epistemology of Digital History.} Further interdisciplinary studies are needed to critically examine how LLMs reshape historical knowledge production, interpretive authority, and educational practices, ensuring that technological augmentation remains aligned with historical rigor and ethical standards~\cite{celli2025language}.
\end{itemize}

\noindent\textbf{Conclusion.}
LLMs offer powerful new tools for historical research, enabling the scaling of traditional methods and the exploration of novel forms of inquiry. However, realizing their potential responsibly demands rigorous attention to their limitations: factual integrity, interpretive nuance, fairness, and transparency. Future advances must emphasize domain-specific fine-tuning, temporal reasoning, source attribution, and collaborative workflows that maintain critical human oversight. As the discipline of history adapts to the possibilities of the LLMs era, the most promising path lies in a balanced integration—where machine assistance amplifies, but does not replace, the historian’s craft.

\subsection{Philosophy} \subsubsection{Overview}

\noindent\textbf{Introduction.} Philosophy is the discipline that examines the most fundamental questions about existence, knowledge, value, reason, consciousness, and language. In contrast to the natural sciences, which rely on a narrow conception of empirical observation and experimentation, philosophy instead draws on general human experience and uses the tools of phenomenological reflection, logical modeling, narrative reasoning, conceptual analysis, and the intellectual resources offered by various wisdom and religious traditions to explore the foundations of human thought~\cite{sellars1962philosophy, russell2019problems}.

To put it more simply, philosophy is the pursuit of wisdom about ourselves and the universe. Rather than depending on empirical measurement, it draws on a broader conception of experience and perception and invites us to think deeply about the world and our place in it.

From this perspective, philosophical inquiry naturally leads to enduring questions: What is reality? Are skeptical worries philosophically justified? To what extent can philosophical thought be formalized? What is good human action? What is human nature? Do we possess free will? What constitutes a meaningful life? Although such questions may not yield definitive or formalizable answers, exploring them profoundly shapes how we think, act, and understand one another~\cite{kant1999critique, cohen2000aristotle}.

Because of this transformative capacity, the abstract value of philosophy lies in its ability to broaden and refine the scope of human reasoning, as well as in understanding the importance of culture and tradition in enabling genuine thought. As Aristotle remarked, “All men by nature desire to know”~\cite{cohen2000aristotle}. Similarly, Kant organized the field around four guiding questions—“What can I know? What ought I to do? What may I hope? What is man?”—to illuminate the major domains of inquiry~\cite{kant1999critique}. Likewise, Russell argued that philosophy’s value stems not from providing final solutions but from expanding intellectual possibility~\cite{russell2019problems}. Together, these views highlight philosophy’s essential role in cultivating clarity, curiosity, and critical reflection.

This abstract value becomes more concrete when we consider how philosophical ideas have both articulated currents of thought and interacted with historical developments. For instance, Aristotle's articulation of the virtues continues to be the basis for our understanding of what it is for human beings to flourish, and his philosophical insights about how genuine thought---both practical and theoretical---cannot be fully captured in logical or other formal models continues to resonate through the centuries.  Another example is John Locke’s social contract theory. Locke's ideas that individuals possess natural rights to “life, liberty, and property,” and that governments are legitimate only when founded on the consent of the governed~\cite{locke2013two}, were central to Enlightenment political thought and set the stage for major political transformations, not least the drafting of the \textit{Declaration of Independence}. 

One should note that it is always an open question whether such philosophical ideas are reflections of societal trends, or whether they are shaping such trends (or some admixture of the two). For instance, Elizabeth Anscombe famously argued that Oxford moral philosophy does not corrupt the youth, but only because it is already in line with the corrupt ideals of society at large (REF) -- if she is right, then it is really societal trends that are shaping much professional philosophy and not the other way round. At any rate, the truth of the matter is yet another item that is the subject of philosophical inquiry.
Taken together, these developments show that although philosophy involves theoretical reflection on questions of knowledge, morality, and human existence, philosophical developments not independent of culture, tradition, and societal trends. Through both its intellectual contributions and its complex relationship with political institutions, philosophy demonstrates an enduring capacity to deepen human understanding and to reflect on the structures of real-world societies.

\noindent\textbf{Core Domains of Philosophical Inquiry.}
Although philosophy spans a wide intellectual terrain, its central questions can be organized into four closely connected domains, each addressing a distinct dimension of human understanding. 

First, \textit{metaphysics and epistemology} investigate the nature of reality and the conditions of knowledge. Metaphysics examines what kinds of things exist (e.g. substances and accidents), while epistemology asks how we come to know what we know and what counts as justified belief~\cite{descartes2016meditations, kant1999critique}. These areas draw upon and systematize our common-sense experience and phenomenology, as well as what we know on the basis of various traditions of ethical and religious thought. 

\textit{Ethics and political philosophy} explore how we ought to act and how our societies ought to be organized. Ethics analyzes what is good, right, or virtuous in human behavior, whereas political philosophy evaluates the legitimacy of authority, the nature of justice, and the structure of political institutions~\cite{irwin2019nicomachean, rawls2017theory}. These inquiries help articulate the normative principles that guide law, governance, and interpersonal conduct.

Complementing these normative concerns, \textit{philosophy of mind and human existence} addresses questions about consciousness, selfhood, perception, and the existential conditions of human life. This includes debates over the mind--body relationship, free will, intentionality, and the meaning of life~\cite{heidegger1977basic, nagel2024like}. In contemporary contexts such as neuroscience and artificial intelligence, these issues take on renewed interdisciplinary significance.

Finally, \textit{logic, language, and philosophy of science} analyze the structure of reasoning, the nature of meaning, and the epistemic foundations of scientific inquiry. This domain includes symbolic logic, theories of reference, and discussions of realism, explanation, and the limits of scientific knowledge~\cite{frege1980foundations, kuhn1970structure}. By clarifying how arguments are formed and how scientific theories operate, it forms a bridge between abstract philosophical thought and empirical investigation.

These four domains not only delineate the contours of philosophical study but also anchor its connections with science, politics, art, and everyday human experience.

\begin{figure}[!t]
    \centering
    \includegraphics[width=\linewidth]{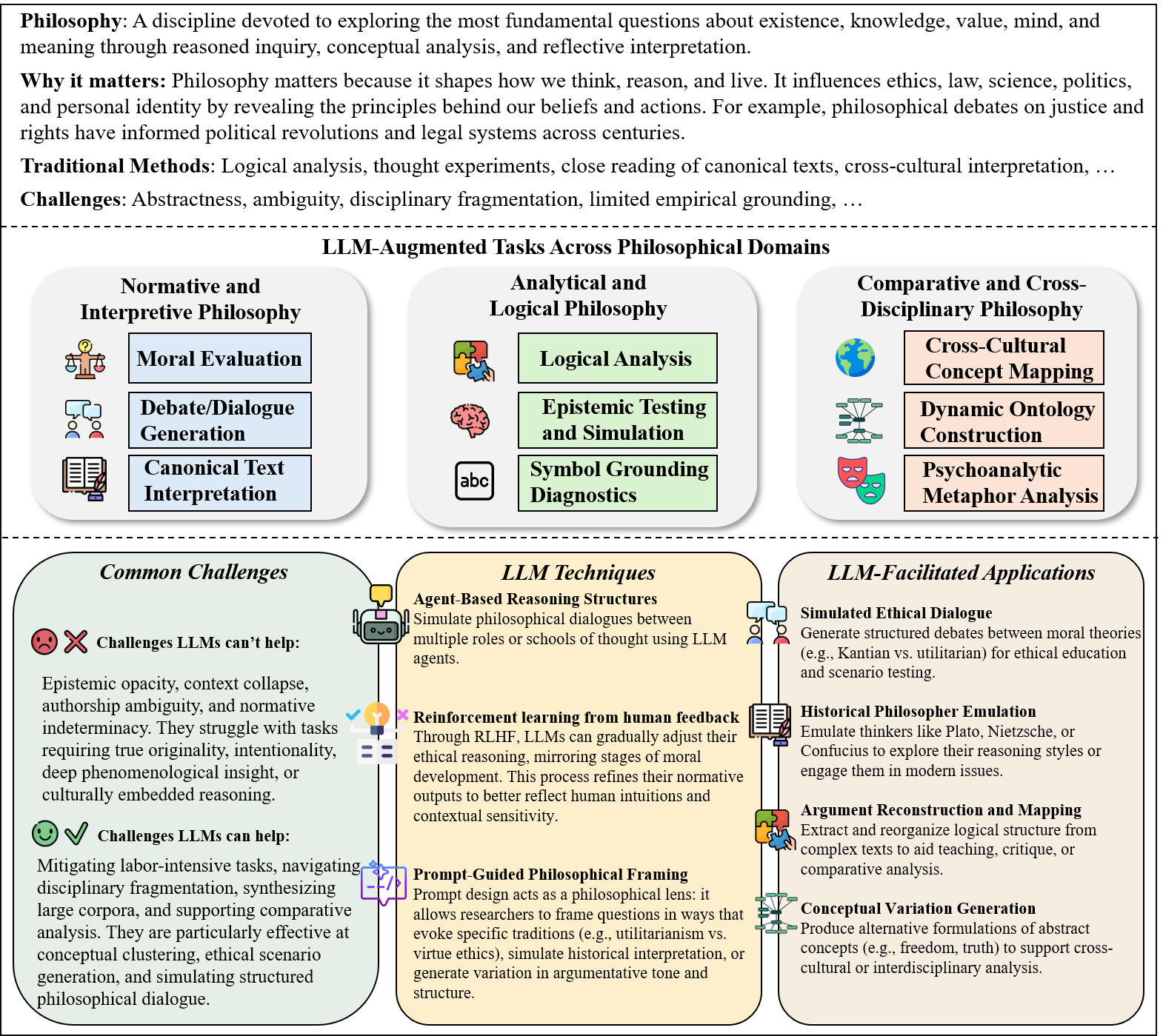}
    \caption{The philosophy research in the era of LLMs.}
    \label{fig:philosophy_framework}
\end{figure}

\noindent\textbf{The Role of LLMs.}
LLMs offer a transformative toolkit for philosophical research. Trained on vast corpora that include classical treatises, contemporary scholarship, and interdisciplinary works, LLMs can read, summarize, translate, and generate complex arguments at scale. By identifying patterns in philosophical discourse, drawing connections between seemingly distant ideas, and simulating dialectical exchanges, they enable new modes of inquiry that were previously impractical. As a result, LLMs are especially valuable for tasks such as literature synthesis, hypothesis generation, and preliminary conceptual analysis.

Beyond these high-level capabilities, LLMs can support philosophers in more concrete ways. They can automate the review of extensive literatures, condense dense arguments into digestible summaries, and cluster related concepts across traditions. Their generative capacities can also introduce novel perspectives that bridge established schools of thought. Moreover, LLMs facilitate interactive exploration of complex theoretical landscapes, helping researchers discover new insights and formulate innovative research questions. Early applications include simulating ethical debates, modeling epistemological frameworks, and mapping ontological structures across diverse intellectual traditions.

\noindent\textbf{Limitations of LLMs.}
Despite these promising contributions, LLMs also exhibit notable limitations when applied to philosophical research. \textit{Factual and logical hallucination} remains a central concern: LLMs may produce arguments that are coherent in form yet unfounded or erroneous in content, thereby undermining academic rigor and the reliability of conclusions~\cite{coeckelbergh2025llms}. In addition, \textit{abstract conceptualization} often poses a challenge, as deeply nuanced or highly theoretical concepts may be oversimplified or misinterpreted~\cite{overgaard2024clarification}. Because LLMs are trained on existing texts, they are also susceptible to \textit{bias and partial perspectives}, reproducing dominant paradigms while overlooking marginalized voices or alternative viewpoints~\cite{colombatto2024folk}. Finally, the issue of \textit{transparency and source attribution} complicates the integration of LLM-generated content into scholarly work, since outputs frequently lack explicit citations, making it difficult to verify claims or trace the lineage of ideas~\cite{heersmink2024phenomenology}.

\noindent\textbf{Taxonomy.}
To better understand the potential of LLMs in philosophical research, we organize the field into three broad categories based on methodological approaches and research goals:

\begin{itemize}[leftmargin=10pt] 

\item \textbf{Normative and Interpretive Philosophy:}
This area emphasizes the analysis of moral values, ethical dilemmas, and interpretive narratives within philosophical traditions. LLMs can assist in generating novel normative arguments, reconstructing historical ethical discourses, and offering fresh interpretations of canonical texts.

\item \textbf{Analytical and Logical Philosophy:}  
        This approach applies formal logic, precise argumentation, and rigorous conceptual analysis to explore philosophical problems. LLMs can support the systematic breakdown of complex arguments, facilitate comparative analyses of theoretical positions, and enhance clarity in logical reasoning.

\item \textbf{Comparative and Cross-Disciplinary Philosophy:}  
        This domain integrates insights from diverse fields—such as political science, sociology, and linguistics—to study philosophical questions from multiple angles. LLMs can aid in cross-cultural comparisons, generate analogies between disparate theories, and help synthesize interdisciplinary perspectives.
        
\end{itemize}

Our taxonomy aligns with frameworks found in contemporary philosophical research, reflecting the balance between traditional humanistic inquiry and data-driven, computational methods.

\subsubsection{Normative and Interpretive Philosophy}

Normative and interpretive philosophy examines the ethical, cultural, and historical dimensions of human life. It focuses on the values that guide action, the moral frameworks that shape deliberation, and the narratives through which individuals and societies understand right and wrong. Rather than prioritizing only formal rigor or abstract universals, this tradition emphasizes the richness of lived experience—foregrounding context, positionality, and meaning. Philosophers working in this domain therefore engage closely with literature, history, and socio-linguistic patterns, interpreting moral claims not simply as logical propositions but as situated expressions of ethical life.

Recent research demonstrates how LLMs are beginning to intersect with this tradition in complex ways. Dillion et al.~\cite{dillion2025ai} showed that laypeople sometimes judged GPT-4's ethical advice as slightly more moral, trustworthy, and thoughtful than that of a human ethicist, indicating a potential shift in public perceptions of moral expertise. Building on this, Kempt et al.~\cite{kempt2024towards} argued that ethical dialogue hinges not merely on formal consistency but on conversational and socio-linguistic appropriateness, offering a framework that captures the pragmatic character of real-world moral discourse. Extending these insights, Wang et al.~\cite{wang2025possibilities} drew on Deweyan models of moral growth to analyze how iterative learning processes in LLMs—particularly Reinforcement Learning from Human Feedback (RLHF)—parallel developmental trajectories of ethical learning, though without achieving full conceptual maturity. Colombatto et al.~\cite{colombatto2024folk} added yet another dimension by examining how people attribute consciousness and moral agency to LLMs based on their behavioral cues, raising important questions about anthropomorphism, folk psychology, and ethical projection.

Taken together, these studies suggest that LLMs are not merely tools for producing ethical content; they are reshaping expectations about what it means to reason morally in an era of artificial intelligence. At the same time, this emerging potential is tempered by ongoing concerns about contextual sensitivity, cultural bias, and the fundamental limits of machine understanding in normative domains.

\subsubsection{Analytical and Logical Philosophy}

Analytical philosophy prioritizes clarity, precision, and argumentative rigor. Its methods center on the careful dissection of philosophical problems, sustained attention to formal structure, and a commitment to logical coherence. Inquiry in this tradition typically proceeds through systematic critique, deductive reasoning, and conceptual analysis, drawing extensively on tools from mathematics, linguistics, and computer science.

LLMs have become increasingly relevant within this methodological framework. Mugleston et al.~\cite{mugleston2025epistemology} argued that LLMs challenge traditional epistemological categories by generating compressive but generative representations of knowledge that resist interpretation through standard propositional models. Heersmink et al.~\cite{heersmink2024phenomenology} further examined the epistemic status of LLM outputs, emphasizing the tension between the black-box opacity of these systems and the philosophical demand for transparency and traceability in epistemic justification. Coeckelbergh~\cite{coeckelbergh2025llms} addressed the epistemic risks posed by hallucination and misinformation, showing how LLM-generated errors can destabilize the informational foundations of democratic deliberation.

Bender et al.~\cite{bender2020climbing} offer a conceptual framework for evaluating whether systems like LLMs can meaningfully be said to “understand,” arguing that linguistic fluency alone is insufficient for genuine semantic competence. Their analysis clarifies the stakes of attributing understanding or quasi-consciousness to computational models.

Finally, Harnad~\cite{harnad2025language} revisited the symbol grounding problem, arguing that despite their linguistic sophistication, LLMs lack the sensorimotor grounding required for true semantic understanding. This critique highlights the inherent limits of machine “comprehension” from an analytical and cognitive perspective.

What makes analytical philosophy particularly significant in the context of LLMs is its emphasis on testability, validity, and formal consistency. With the rise of AI-generated reasoning, longstanding debates in the philosophy of language—such as the distinction between syntax and semantics, the nature of reference, and the force of classical challenges like the Chinese Room argument—have taken on renewed urgency. LLMs now serve both as subjects for examining these issues and as platforms for rethinking them. Moreover, LLMs increasingly assist philosophers by creating formal argument maps, identifying unstated premises, and evaluating deductive soundness. These capabilities have clear pedagogical value for teaching formal logic and informal reasoning. Yet their effectiveness depends heavily on prompt design and alignment mechanisms, which themselves call for philosophical scrutiny. Importantly, the proliferation of LLMs is prompting philosophers to revisit foundational questions: Can a non-conscious system produce genuine knowledge? What constitutes “understanding” in a computational model? Are LLM outputs assertions, simulations, or performative acts? These questions challenge not only epistemology and logic but also the methodological foundations of philosophy in the twenty-first century.

\subsubsection{Comparative and Cross-Disciplinary Philosophy}

Comparative and cross-disciplinary philosophy seeks to bridge traditions and methodologies across cultures and academic domains. It examines how different societies formulate responses to universal philosophical questions and brings together insights from political theory, sociology, cognitive science, psychoanalysis, and related fields. Rather than relying on a single analytic style or normative framework, this approach values synthesis, pluralism, and sustained dialogue across intellectual boundaries.

LLMs play a distinctive role in this area by enabling the integration of heterogeneous data sources and philosophical paradigms. Colombatto et al.\cite{colombatto2024folk} showed how people attribute consciousness to AI systems, revealing individual variability conditioned by experience and usage patterns—an important precursor to broader cross-cultural comparisons.

Heimann et al.\cite{heimann2025extimate} provided an additional interdisciplinary perspective by applying a Lacanian psychoanalytic framework to metaphorical disruptions in LLM outputs, drawing parallels between psychotic structures and the disjointed metaphor use seen in machine-generated language. Harnad et al.\cite{harnad2025language} further enriched this conversation by situating LLMs within the broader arc of human symbolic evolution, arguing that computational models both reflect and distort the underlying architecture of meaning.

Through these diverse engagements, LLMs function as tools for comparative ontology, cultural critique, and interdisciplinary synthesis. They illuminate differences in philosophical worldviews while generating new frameworks for exploring global patterns of thought.

\subsubsection{Benchmarks}

\begin{table}[htbp]
\centering
\caption{Curated Benchmarks and Resources for Philosophy and Humanities}
\begin{tabular}{p{2cm} p{2.5cm} p{3cm} p{2.5cm} p{4cm}}
\toprule
\textbf{Benchmark} & \textbf{Scope and Focus} & \textbf{Data Composition} & \textbf{Evaluation Tasks} & \textbf{Key Insights} \\
\midrule

\textbf{Project Gutenberg}
~\cite{projectgutenberg}&
Classical philosophical text access in the public domain &
75k of digitized philosophical and humanities books from major thinkers across eras &
Long-form reasoning, document understanding, language modeling &
Serves as foundational source for philosophy QA, argument parsing, and training philosophically informed LLMs \\
\addlinespace

\textbf{PhilPapers} 
~\cite{philpapers}&
Bibliographic indexing and philosophical research discovery &
Index of over 2.5M philosophy papers with abstracts, metadata, and categorization across subfields &
Document retrieval, topic modeling, citation graph analysis &
Widely used in research discovery and argument mining; useful for concept clustering and academic stance tracking \\
\bottomrule
\end{tabular}
\label{tab:philosophy-benchmarks}
\end{table}

To evaluate the ability of LLMs to understand abstract concepts, perform long-form reasoning, and support scholarly exploration, researchers often draw on curated corpora and metadata resources from philosophy and the humanities. Table~\ref{tab:philosophy-benchmarks} highlights two widely used sources—Project Gutenberg and PhilPapers—that, while not benchmarks in the strict sense, serve as foundational materials for constructing philosophy-oriented evaluation tasks.

\textbf{Project Gutenberg} is a large public-domain digital library that hosts thousands of classical philosophical and literary works, including texts by Plato, Aristotle, Descartes, Hume, and other major figures. Although originally created for public access rather than machine learning, its extensive collection of historical prose has become an important corpus for pretraining and evaluating LLMs on long-form reasoning, stylistic analysis, and document understanding. While Gutenberg does not provide explicit benchmark tasks, its texts are frequently used to build datasets for philosophical question answering, thematic classification, and argument-recognition studies.

\textbf{PhilPapers} is a comprehensive bibliographic database indexing more than 2.5 million philosophy-related academic works. It provides metadata, abstracts, keywords, and a detailed taxonomy of philosophical subfields. Although the full texts are usually unavailable for model training due to copyright restrictions, PhilPapers metadata plays a crucial role in tasks such as topic modeling, concept clustering, academic stance detection, and the construction of retrieval-based evaluation sets. Its structured categorization system makes it a valuable resource for designing benchmarks that test an LLM’s ability to reason within well-defined philosophical domains.

Together, these resources provide broad coverage of philosophy as both a historical tradition and a contemporary research discipline. They serve as essential building blocks for constructing benchmarks that assess how well LLMs can handle complex, ambiguous, and abstract content—traits central to human philosophical inquiry.

\subsubsection{Discussion}

\noindent\textbf{Opportunities and Impact.}
The adoption of LLMs in philosophical research introduces powerful new methods for engaging long-standing questions of ethics, meaning, and knowledge. As demonstrated in normative contexts—such as laypeople rating GPT-4’s ethical advice as highly trustworthy~\cite{dillion2025ai}—and in epistemic analysis~\cite{heersmink2024phenomenology}, LLMs assist in interpreting dense arguments, simulating ethical debates, and synthesizing diverse intellectual traditions. They support the examination of canonical texts, the generation of novel conceptual variations, and the mapping of ontological structures across philosophical schools.

LLMs also reduce barriers of scale and access by automating tasks such as argument classification, stylistic translation, and comparative summarization. They allow philosophers to simulate dialogues between traditions (e.g., utilitarianism vs.\ deontology), emulate historical voices, and reframe problems from alternative perspectives. Through these capabilities, LLMs have the potential to broaden access to philosophical education and complement traditional scholarly methods.

\noindent\textbf{Challenges and Limitations.}
Despite these advantages, the use of LLMs in philosophy is constrained by serious risks. Hallucination and fabrication threaten intellectual rigor, particularly when models generate arguments without grounding in verifiable sources~\cite{coeckelbergh2025llms}. LLMs may oversimplify abstract ideas or misrepresent nuanced debates. Biases in training data can lead models to reproduce Western-centric canons while marginalizing non-Western or alternative philosophical traditions~\cite{colombatto2024folk}.

A further challenge is the lack of transparency and epistemic accountability. Since LLMs frequently omit citations and cannot trace their claims to specific sources, their contributions are difficult to integrate into rigorous scholarly debate~\cite{heersmink2024phenomenology}. Finally, the absence of intentionality, phenomenological consciousness, or cultural embodiment limits the depth of questions LLMs can meaningfully address: they simulate reasoning but lack lived experience, self-awareness, or contextual grounding.

\noindent\textbf{Research Directions.} Several promising avenues for future work include:

\begin{itemize}[leftmargin=10pt]

\item \textbf{Philosophical Fine-Tuning and Corpus Design.}  
Carefully curated and diverse corpora can mitigate bias and expand philosophical coverage~\cite{paullada2021data}.

\item \textbf{Logical Structure and Argument Mining.}  
Developing tools for extracting and visualizing formal arguments can enhance transparency and pedagogical value.  
See Lawrence et al.’s survey of argument mining~\cite{lawrence2019argument} for foundational methods applicable to philosophical texts.

\item \textbf{Ontology Mapping and Comparative Frameworks.}  
LLMs may assist in comparing metaphysical and ethical schemas across cultural traditions. Cross-cultural studies of consciousness attribution~\cite{salazar2023cross} show how cultural context shapes interpretations of AI minds.

\item \textbf{Interactive Dialogue Systems for Teaching.}  
LLMs deployed as Socratic partners or role-played historical thinkers can deepen philosophical engagement~\cite{dillion2025ai}.

\item \textbf{Epistemology and AI Ethics.}  
Further interdisciplinary work is needed to assess the epistemic status of LLM outputs, their limits of understanding, and their role in reshaping human inquiry~\cite{heersmink2024phenomenology}.

\end{itemize}

\noindent\textbf{Conclusion.}
LLMs offer compelling tools for augmenting philosophical research and education. From dialogical simulations to conceptual experimentation, they enhance scalability and diversify access to philosophical content. Yet realizing this promise responsibly requires deeper attention to epistemic integrity, bias, and philosophical coherence. Future developments must focus on transparency, inclusivity, and alignment with human critical faculties. Rather than replacing human reflection, LLMs should be regarded as tools for extending and enriching the practice of philosophy in contemporary contexts.

\subsection{Political Science} 
\subsubsection{Overview}

\noindent\textbf{Introduction.} Political science is the systematic study of power, governance, institutions, behavior, and how authority is shared in societies. It looks at how decisions are made, how policies are created and carried out, and how political actors compete, work together, and defend their right to rule in local and global systems \cite{hunter1962governs, eulau1954political, gerth1946politics}.

In simpler terms, political science explores how people organize themselves to make group decisions and solve conflicts. It studies elections, governments, protests, ideologies, and laws—how they work, why they break down, and what values they support.

For example, it asks why democracies grow or fall, how authoritarian governments keep control, and how public opinion affects policy. Political science helps not only scholars, but also journalists, activists, diplomats, and everyday citizens.


\begin{figure}[!t]
    \centering
    \includegraphics[width=\linewidth]{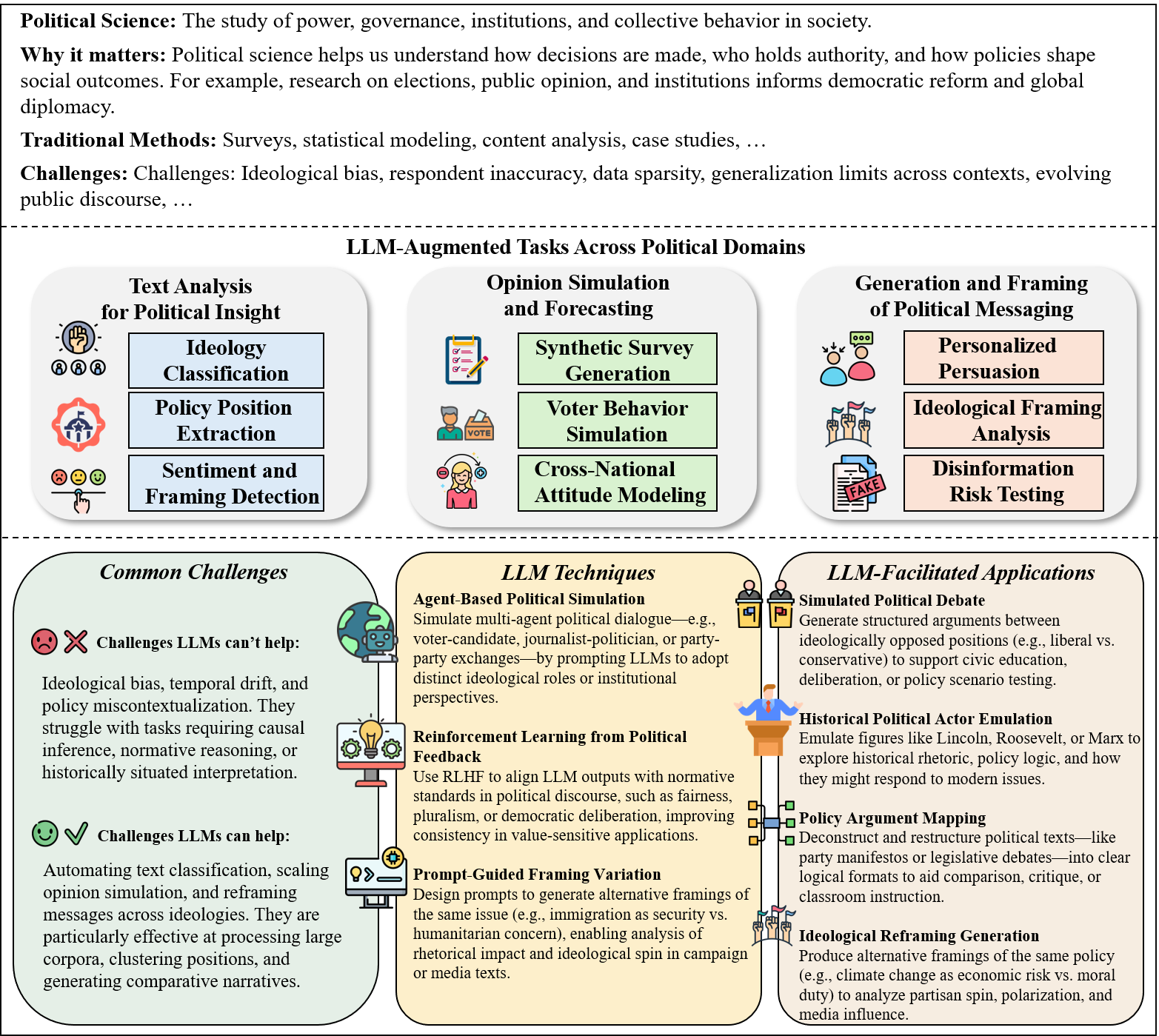}
    \caption{The political research in the era of LLMs.}
    \label{fig:philosophy_framework}
\end{figure}

\noindent\textbf{Traditional Political Research.}  
Political scientists tried diverse methods to investigate questions of power, governance, and behavior. 
These approaches include:
\textit{Textual and Historical Analysis.} The close reading of constitutions, political speeches, and canonical works to uncover ideological foundations and institutional change~\cite{skinner2002visions}.
\textit{Comparative Case Studies.} Cross-national or historical comparisons to identify causal patterns in political systems and transitions~\cite{lijphart1971comparative}.
\textit{Survey Research and Public Opinion Analysis.} Empirical measurement of citizens’ beliefs, voting behavior, and attitude structures through questionnaires and polls~\cite{converse2006nature}.
\textit{Formal Modeling and Game Theory.} Use of abstract models and strategic reasoning to represent decision-making among rational actors~\cite{downs1957economic}.
\textit{Statistical and Computational Approaches.} Application of regression, network analysis, and text-as-data methods to detect patterns in political discourse and social dynamics~\cite{grimmer2022text}.
Together, these methods span philosophical, empirical, and quantitative traditions. However, they also face challenges in scalability, integration, and the interpretation of complex social meaning.

\noindent\textbf{The role of LLMs.}
LLMs can change how political scientists study texts, model behavior, and create new ideas. They can handle large amounts of unstructured political data, such as speeches, manifestos, news articles, and social media posts. This gives researchers new ways to study ideology, public opinion, and how political language works \cite{aoki2024large}.

LLMs can also help classify sentiment and ideological positions \cite{grimmer2022text}. They can create realistic answers for survey simulations and build policy or campaign messages that fit specific themes. In archival work, they can help people explore political history by summarizing, translating, or rewriting historical texts. These abilities widen the range and speed of political analysis and reduce the need for manual coding, while allowing fast testing of ideas.

\noindent\textbf{Limitations of LLMs.}
Despite their promise, LLMs face significant limitations when applied to political research: \textit{Hallucination:} LLMs often generate plausible but factually incorrect content, posing risks for misinformation and faulty inference \cite{zhang2025siren}.
\textit{Temporal and Institutional Fragility:} They may struggle to track historical timelines, legislative sequences, or institutional nuances across contexts.
\textit{Bias Propagation:} Because they inherit biases from training data, LLMs may reinforce stereotypes or marginalize dissenting perspectives, raising concerns about fairness and representation \cite{bang2024measuring}.
\textit{Shallow Interpretation:} LLMs often lack the ability to perform deep causal or normative reasoning, limiting their utility in theory-building or critical analysis.
\textit{Opacity and Attribution:} Their outputs typically lack transparent source citations, complicating verification and undermining reproducibility.

For these reasons, LLMs are best viewed not as replacements for interpretive or empirical political methods, but as augmentative tools that can enhance—but not substitute—core disciplinary judgment.

\noindent\textbf{Taxonomy.}  
To organize the expanding applications of LLMs in political science, we propose a task-based taxonomy that reflects how these models are reshaping core research practices. This framework consists of three domains:

\begin{itemize}[leftmargin=10pt]
  \item \textbf{Text Analysis for Political Insight:}  
  LLMs assist in analyzing political texts such as speeches, party platforms, policy documents, and social media discourse. They are widely used for tasks like sentiment analysis, policy position classification, ideological scaling, and automated topic modeling, enabling scalable and consistent textual interpretation at unprecedented scale.

  \item \textbf{Opinion Simulation and Forecasting:}  
  LLMs can simulate public opinion, generate synthetic survey respondents (“silicon samples”), and forecast electoral outcomes through multi-step reasoning. These capabilities are particularly valuable for behavioral modeling, comparative analysis, and data augmentation in contexts where real-world survey data is limited.

  \item \textbf{Generation and Framing of Political Messaging:}  
  LLMs are increasingly used to craft persuasive political language, adapt messages to different audiences, and analyze framing strategies. These applications include generating campaign slogans and policy narratives, auditing ideological bias in generated outputs, and studying the persuasive dynamics of AI-authored political content.
\end{itemize}

\subsubsection{Text Analysis for Political Insight}

A main area where LLMs show strong potential is in the automated study of political text. Political scientists often use speeches, party platforms, social media posts, and legislative records to learn about ideology, sentiment, important issues, and how political leaders communicate. LLMs provide scalable tools that can support or replace older methods of content analysis, often with similar or even better performance.

Ornstein et al.~\cite{ornstein2025train} show that few-shot prompting with GPT-3.5 and GPT-4 can handle complex classification and topic modeling on political documents without needing large labeled datasets. Le Mens and Gallego~\cite{le2025positioning} introduce an “ask-and-average” method that lets LLMs place political texts on an ideological scale, with results that match expert surveys and roll-call–based measures. In the same way, O’Hagan and Schein~\cite{o2023measurement} find that LLM-based estimates of ideological positions are stable and easy to interpret, and they can perform as well as traditional text scaling methods while relying on fewer assumptions.

Beyond ideological scaling, LLMs also excel in classification tasks central to empirical political research. Liu and Shi~\cite{liu2024poliprompt} introduce PoliPrompt, a cost-effective framework for political text classification that achieves high accuracy in stance detection and sentiment labeling by combining prompt engineering with consensus-based inference. Törnberg~\cite{tornberg2023chatgpt} further finds that ChatGPT-4 surpasses both expert coders and crowd workers in classifying political tweets, even in zero-shot settings.

Together, these studies show that LLMs can interpret political language at large scale and can produce detailed annotations and hidden features that broaden the tools available in computational political science. As LLMs grow more advanced, their use in political text analysis is likely to speed up empirical research and also bring up important questions about transparency, replicability, and clear concepts.

\subsubsection{Opinion Simulation and Forecasting}

LLMs are increasingly used to simulate public attitudes, generate synthetic survey data, and forecast electoral outcomes—tasks that lie at the core of political behavior research. These models can act as proxies for survey respondents, offering cost-effective tools for hypothesis testing in contexts where traditional data collection is limited or infeasible.

Qu and Wang~\cite{qu2024performance} demonstrate that ChatGPT can simulate public opinion across diverse issue domains with surprising fidelity, particularly on moral and political questions. However, they also note that the model’s representational capacity varies across regions and demographics, raising concerns about generalizability. Yu et al.~\cite{yu2024large} propose a multi-step simulation framework for modeling political decision-making using LLMs, incorporating voter profiles and ideological cues to emulate realistic behavioral patterns at scale.

Role-conditioning has emerged as a key strategy to enhance realism. Karanjai et al.~\cite{karanjai2025synthesizing} show that embedding demographic and personality attributes into prompts enables LLMs to generate diverse and policy-relevant opinion distributions—what they call “synthetic public spheres.” Meanwhile, Lee et al.~\cite{lee2024can} empirically test whether LLMs can estimate population-level opinion distributions on global warming, finding promising accuracy when covariates are incorporated, but persistent bias in underrepresented subgroups.

At the predictive frontier, Bradshaw et al.~\cite{bradshaw2024llm} introduce a novel distribution-based method for forecasting U.S. electoral outcomes using token probability distributions from LLMs. Their work highlights how generative uncertainty can be turned into a probabilistic prediction tool, offering new metrics for electoral inference.

Collectively, these studies suggest that LLMs are more than textual tools—they are evolving into simulators of political behavior, capable of reflecting, and at times amplifying, the complexities of democratic publics. Yet, these advances also necessitate careful scrutiny of bias, alignment, and methodological transparency.

\subsubsection{Generation and Framing of Political Messaging}
LLMs are increasingly being used not only to analyze but to generate persuasive political content. This includes tasks such as crafting campaign messages, personalizing voter appeals, reframing controversial issues, and, in darker contexts, synthesizing disinformation. These capabilities place LLMs at the center of contemporary political communication, raising both practical opportunities and urgent ethical questions.

Hackenburg et al.~\cite{hackenburg2024evidence} show that the persuasive effectiveness of political messages generated by LLMs increases logarithmically with model size. Using 24 different models and 720 U.S.-focused prompts, their study finds that while larger models are more persuasive, returns diminish rapidly beyond a certain scale. This implies that performance can often be optimized through prompt engineering rather than model expansion alone. In related work, Aldahoul et al.~\cite{aldahoul2025large} evaluate the ideological positioning of LLMs and show that many models exhibit politically extreme and inconsistent behaviors across prompts—yet still persuade users effectively, even when merely conveying neutral information.

One emerging trend is the use of LLMs for personalized persuasion at scale. Matz et al.~\cite{matz2024potential} demonstrate that ChatGPT, when tailored to users' psychological profiles, can generate messages that are significantly more effective than generic ones. These findings point toward a future where AI-driven campaign strategies are hyper-individualized, adapting tone, framing, and issue salience to individual voters.

On the other hand, the same generative capacities pose risks. Williams et al.~\cite{williams2025large} assess how easily LLMs can be prompted to generate false but persuasive election-related narratives. Their DisElect dataset and experimental findings show that several popular LLMs consistently produce high-quality disinformation that is difficult for humans to distinguish from truth. These results underscore the importance of guardrails, transparency, and disinformation detection frameworks.

Finally, Šola et al.~\cite{vsola2025human} combine AI-driven eye-tracking and LLM-based message generation to redesign political campaign materials. Their study highlights that human-centered design principles—when merged with generative models—can enhance attention and emotional engagement in political communication.

In sum, LLMs are not passive describers of political reality; they are active agents in shaping it. Whether optimizing legitimate political messaging or amplifying manipulative content, their generative potential reconfigures the dynamics of persuasion, participation, and propaganda in the digital public sphere.

\subsubsection{Benchmarks}

\begin{table}[htbp]
\centering
\caption{Curated Benchmarks and Datasets for Political NLP}
\begin{tabular}{p{2cm} p{2.5cm} p{3cm} p{2.5cm} p{4cm}}
\toprule
\textbf{Benchmark} & \textbf{Scope and Focus} & \textbf{Data Composition} & \textbf{Evaluation Tasks} & \textbf{Key Insights} \\
\midrule

\textbf{POLITICS Dataset}
~\cite{kulkarni2018multi} &
Ideology and stance detection from media coverage &
3.6M political news articles from diverse media sources (left, center, right) &
Stance detection, ideology prediction, text classification &
Highlights media bias and ideological polarization; useful for evaluating fairness and framing in political text models \\
\addlinespace

\textbf{U.S. Congressional Records} 
~\cite{congressionalrecord} &
Modeling formal political discourse and legislative language &
Standardized collection of U.S. congressional transcripts and debates &
Summarization, intent classification, dialogue analysis &
Ideal for studying legal-parliamentary styles; enables models to learn structured argumentation in policymaking \\
\addlinespace

\textbf{American Presidency Project}
~\cite{woolley1999american} &
Presidential speech analysis and political rhetoric tracking &
Public speeches, press conferences, and presidential communications from 1789 onward &
Rhetorical style analysis, time-based policy comparison &
Enables longitudinal study of U.S. executive discourse; supports leadership style profiling across administrations \\
\addlinespace

\textbf{Media Bias \& Fact-check Dataset} 
~\cite{mediabiasfactcheck} &
Detecting factual reliability and political bias in media &
Fact-check ratings and bias scores from MediaBiasFactCheck and AllSides &
Text labeling (bias/fact), misinformation detection &
Supports training of bias-aware LLMs; widely used in political misinformation detection and source validation \\
\addlinespace

\textbf{TruthfulQA}
~\cite{lin2022truthfulqa} &
Evaluating truthful generation in politically sensitive contexts &
Manually curated QA dataset focusing on false beliefs and politically sensitive facts &
Question answering, hallucination detection &
Used to stress-test LLMs on truthfulness and bias in politically charged content; benchmarks robustness to disinformation \\
\bottomrule
\end{tabular}
\label{tab:political-benchmarks}
\end{table}
To evaluate the capabilities of LLMs in political analysis and discourse understanding, several specialized benchmarks have been introduced across media, legislative, and rhetorical domains. Table~\ref{tab:political-benchmarks} summarizes five representative datasets that capture different layers of political language processing—from ideology classification and rhetorical structure to factual consistency and bias detection.

\textbf{POLITICS Dataset}~\cite{kulkarni2018multi} is a large-scale corpus of 3.6 million news articles collected from left-, center-, and right-leaning media outlets. It supports tasks such as political stance classification, ideological alignment detection, and media bias recognition. By modeling political signals embedded in content and hyperlinks, the dataset helps identify framing strategies and polarization in news discourse. It has been widely used for training and evaluating fairness-aware and ideology-sensitive language models.

\textbf{U.S. Congressional Records}~\cite{congressionalrecord} provide a structured textual record of congressional debates and proceedings, enabling fine-grained modeling of formal political speech. Tasks supported by this dataset include legislative summarization, policy intent classification, and speaker-style modeling. These records are particularly valuable for studying deliberative argumentation, procedural discourse, and political framing across party lines.

\textbf{The American Presidency Project}~\cite{woolley1999american} compiles over two centuries of presidential communications, including speeches, executive orders, and press statements. This resource supports rhetorical analysis and temporal comparison of executive discourse. It enables longitudinal studies of leadership language, agenda-setting patterns, and institutional tone shifts across administrations.

\textbf{The Media Bias and Fact-check Dataset}~\cite{mediabiasfactcheck} combines factuality ratings and political bias labels from platforms like MediaBiasFactCheck and AllSides. It supports tasks such as source bias prediction, factual reliability assessment, and misinformation flagging. As LLMs are increasingly used in content moderation and source attribution, this dataset is essential for stress-testing their alignment with factual standards and political neutrality.

\textbf{TruthfulQA}~\cite{lin2022truthfulqa} is designed to test whether language models can resist producing factually incorrect or misleading responses to questions based on false premises or human misconceptions. With 817 curated QA items across 38 categories—including politics, law, and health—it evaluates truthfulness and informativeness under adversarial conditions. The benchmark reveals that larger models may generate more fluent yet less truthful answers, underscoring the need for truth-aligned training in politically sensitive domains.

Collectively, these political NLP benchmarks offer a multi-faceted framework for evaluating LLMs' ability to understand, generate, and fact-check politically charged content. They are essential for ensuring the safety, fairness, and integrity of AI systems deployed in democratic discourse and governance contexts.

\subsubsection{Discussion}

\noindent\textbf{Opportunities and Impact.} LLMs are transforming the scope and method of political science by automating textual analysis, simulating voter behavior, and generating persuasive political content. As demonstrated across core tasks like text classification~\cite{ornstein2025train, le2025positioning, o2023measurement}, synthetic opinion generation~\cite{qu2024performance, karanjai2025synthesizing}, and campaign message design~\cite{hackenburg2024evidence, matz2024potential}, LLMs enable political scientists to explore ideological dynamics, electoral predictions, and persuasive framing at unprecedented scale and speed. These models open new avenues for behavioral experimentation, hypothesis testing, and the synthesis of fragmented discourse.

LLMs also contribute to political pedagogy and public engagement by simulating survey responses, summarizing legislative debates, and facilitating multilingual access to policy documents. In low-resource environments or emerging democracies, they can support faster information processing and broader access to civic discourse. Moreover, LLMs enhance reproducibility and consistency in textual coding, long a bottleneck in political communication and media studies.

\noindent\textbf{Challenges and Limitations.} Despite these gains, LLM-based political analysis faces serious risks. Hallucination and misinformation generation remain persistent challenges~\cite{zhang2025siren}, particularly when models are prompted with politically charged or adversarial inputs. Issues of fairness, bias amplification, and ideological skew~\cite{bang2024measuring} raise concerns about LLMs reinforcing dominant narratives while marginalizing dissent.

Interpretive shallowness further limits LLM utility in theory-building: while models can simulate preference distributions or generate policy arguments, they rarely reflect causal depth, institutional nuance, or normative coherence. Their black-box nature and lack of source attribution complicate scholarly transparency and verification. Finally, the political misuse of LLMs—for generating disinformation~\cite{williams2025large} or manipulating electoral behavior~\cite{aldahoul2025large} —raises urgent ethical questions that require governance and auditing frameworks.

\noindent\textbf{Research Directions.} Building on recent developments, we can outline several promising directions for future research:

\begin{itemize}[leftmargin=10pt]
\item \textbf{Fine-Tuned Political Models.} Domain-specific fine-tuning on legislative records, public opinion corpora, and multilingual political texts can enhance accuracy and reduce ideological bias.
\item \textbf{Causal and Temporal Modeling.} Combining LLMs with structured models or reasoning frameworks may improve their ability to infer causality, detect agenda dynamics, and simulate policy feedback.
\item \textbf{Bias Detection and Auditing.} Systematic tools to audit, mitigate, and document political biases in LLM outputs are essential for responsible deployment.
\item \textbf{Interactive Political Simulation.} LLMs can be embedded in deliberative platforms to support role-playing, voter education, or negotiation training in civic and educational settings.
\item \textbf{Disinformation Defense.} Techniques such as adversarial prompting, fact-checking augmentation, and truth-conditioned training can be used to defend against political misuse.
\end{itemize}

\noindent\textbf{Conclusion.} LLMs offer transformative potential for political science, enabling scalable behavioral modeling, discourse analysis, and personalized messaging. Yet, their integration into democratic contexts must be critically governed to preserve transparency, fairness, and civic trust. Future efforts should combine computational innovation with domain-sensitive safeguards, ensuring that political AI supports deliberation and accountability rather than distortion and manipulation.

\subsection{Arts and Architecture}

\subsubsection{Overview}

\begin{figure}[!t]
    \centering
    \includegraphics[width=\linewidth]{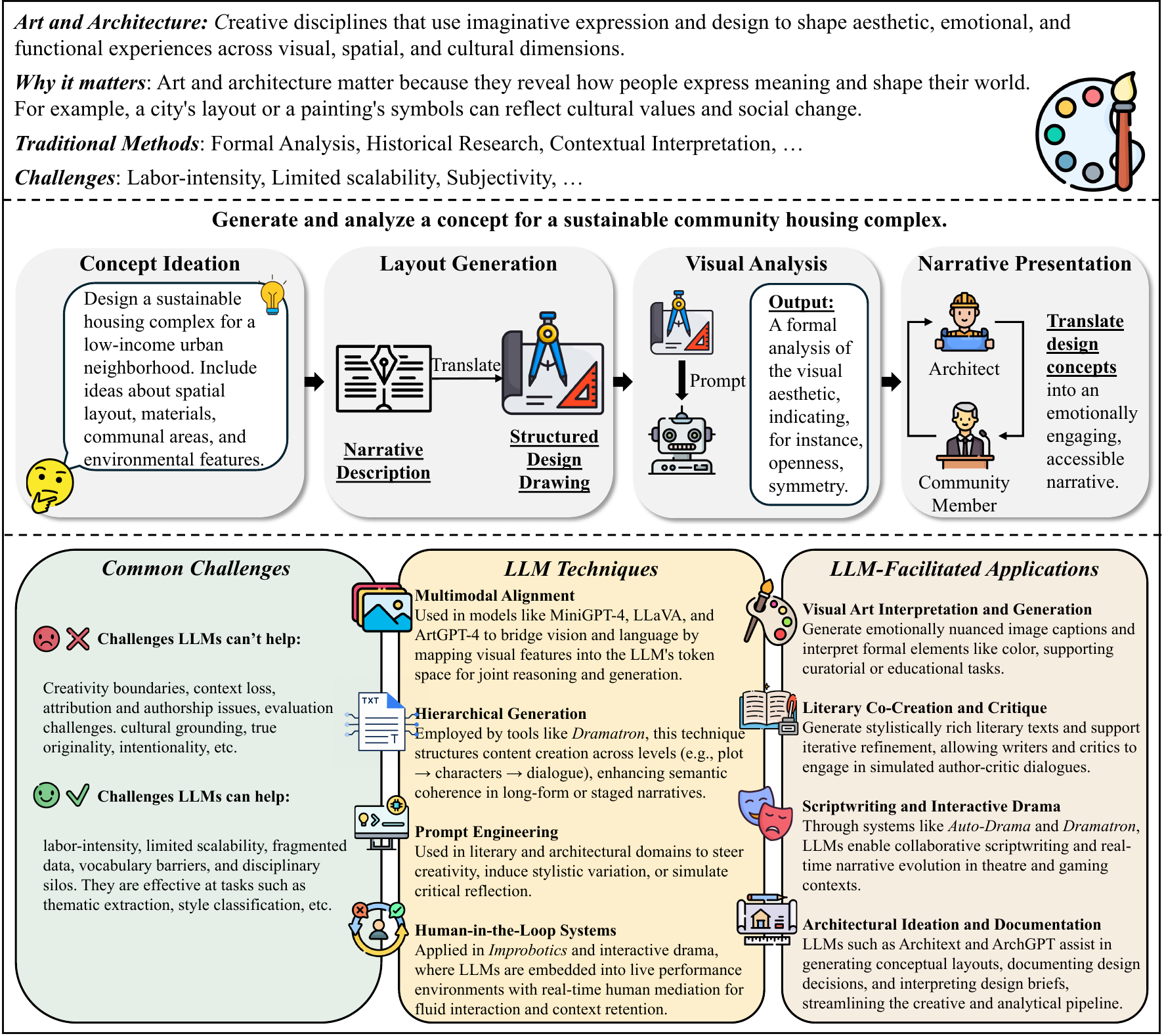}
    \caption{The art and architecture research in the era of LLMs.}
    \label{fig:art_framework}
\end{figure}

\noindent\textbf{Introduction.}
Art is the conscious use of imagination and creative skill to produce works that express aesthetic, emotional, or conceptual ideas, often through mediums such as visual images, sound, performance, or language \cite{davies2005definitions}. Architecture is the art of designing and constructing buildings and physical structures, combining functionality, aesthetics, and environmental consideration to shape human environments \cite{eisenman1970notes,bravoco1985requirement}.

Research in art and architecture traditionally combines theoretical inquiry, historical investigation, and practice-based exploration \cite{baxandall1985patterns}. In the arts, scholars engage with artworks through methods such as formal analysis (examining visual elements like color, line, and composition) \cite{panofsky1970meaning}, iconography (interpreting symbols and motifs), and contextual analysis (situating works within their social, political, or historical setting) \cite{denzin2003performance}. Literary and performance studies may also involve close reading, narrative analysis, and ethnographic observation, as well as archival research to uncover original scripts, notes, or recordings \cite{taylor2003archive}.

In architecture, research spans multiple domains—from studying the evolution of design movements and analyzing urban development to conducting material testing or building performance simulations. Scholars may investigate the sociocultural impact of built environments, model spatial behavior, or explore the use of sustainable technologies \cite{frampton2001studies}. Methodologies often blend technical modeling (e.g., CAD, BIM), environmental simulation, historical documentation, and conceptual critique \cite{hillier1989social,groat2013architectural}. In both art and architecture, qualitative approaches (interviews, fieldwork, visual diaries) and quantitative methods (e.g., surveys, pattern recognition in design databases) are increasingly used in combination \cite{drucker2014graphesis}.

Despite this methodological richness, traditional research in these fields faces several limitations. It is often labor-intensive, time-consuming, and limited by the scale of what a researcher can manually process or observe. Analyzing thousands of artworks, texts, or architectural plans for patterns or cross-cultural comparisons is rarely feasible without computational tools. Furthermore, insights are often locked behind domain-specific expertise or specialized vocabularies, making it challenging to connect findings across artistic disciplines or integrate them with computational models. These challenges highlight the need for scalable, cross-modal, and linguistically aware systems—such as LLMs—to support the next generation of art and architectural research.

\begin{table}[!t]
    \centering
    \caption{LLM Applications and Insights in Art and Architecture}
    \resizebox{\linewidth}{!}{
        \begin{tabular}{@{}p{3.5cm} p{4.5cm} p{6.5cm} p{6.5cm} p{2cm}@{}}
            \toprule
            \textbf{Type of Art} & \textbf{Subtasks}                                                                                                                    & \textbf{Use Case-Inspired Research Question} & \textbf{Insights and Contributions} & \textbf{Citations} \\
            \midrule

            \textbf{Visual Art}
                                 & Creation (Prompting, Image Generation)
                                 & How can LLM-vision models generate visually compelling and stylistically diverse art through prompt-based interaction?
                                 & Multimodal models like MiniGPT-4 and LLaVA support creative generation; ArtGPT-4 enhances aesthetics using image adapters.
                                 & \cite{zhu2023minigpt, liu2023visual, yuan2023artgpt}                                                                                                                                                                                           \\\cmidrule{2-5}

                                 & Analysis (Symbolism, Style Classification)
                                 & Can LLMs classify artistic styles and interpret visual symbolism without dedicated visual encoders?
                                 & GalleryGPT reduces hallucinations via structured prompts; CognArtive uses SVGs to enable symbolic reasoning without visual models.
                                 & \cite{bin2024gallerygpt, khadangi2025cognartive}                                                                                                                                                                                               \\

            \midrule
            \textbf{Literary Art}
                                 & Creation (Storytelling, Stylistic Writing)
                                 & How can LLMs be guided to produce stylistically rich and emotionally resonant literary texts?
                                 & Prompt tuning and temperature settings support diverse styles; enables multi-voice storytelling and creative expression.
                                 & \cite{shanahan2023evaluating, gomez-rodriguez-williams-2023-confederacy}                                                                                                                                                                       \\\cmidrule{2-5}

                                 & Analysis (Critique, Interpretation)
                                 & Can LLMs simulate literary critics or authors to support textual analysis and interpretation?
                                 & Interactive and self-reflective prompting enables critical reading and style-aware interpretation.
                                 & \cite{wang2024evaluating, yang2024analyzing}                                                                                                                                                                                                   \\

            \midrule
            \textbf{Performing Art}
                                 & Creation (Scriptwriting, Interactive Drama)
                                 & How can LLMs co-write scripts and structure dramatic arcs in interactive performances?
                                 & Tools like Dramatron build layered plots; Auto-Drama uses classical structure (e.g., Aristotelian elements) to shape narrative flow.
                                 & \cite{mirowski2023co, wu2024role}                                                                                                                                                                                                              \\\cmidrule{2-5}

                                 & Performance (Live Interaction, Improvisation)
                                 & Can LLMs participate in live improvisational performance alongside human actors?
                                 & Improbotics blends live dialogue with LLM input for co-creative scenes; handles multi-party flow and timing.
                                 & \cite{Branch2024DesigningAE}                                                                                                                                                                                                                   \\

            \midrule
            \textbf{Architecture}
                                 & Design and Creation (Concept Ideation, Layout Generation, Decision Records)
                                 & How can LLMs support early-stage design, spatial layout, and architectural decision-making processes?
                                 & Tools like Architext generate floorplans; GPT-based pipelines assist with ideation and accurate design documentation.
                                 & \cite{Galanos2023ArchitextLG, Ma2024ExploringTC, Dhar2024CanLG}                                                                                                                                                                                \\\cmidrule{2-5}

                                 & Analysis (Heritage Assessment, Design Comparison)
                                 & Can LLMs assist in heritage assessment and design analysis by integrating semantic knowledge with user intent?
                                 & ArchGPT retrieves context-aware insights; supports restoration, regulation checks, and expert collaboration.
                                 & \cite{Zhang2024ArchGPTHL}                                                                                                                                                                                                                      \\

            \bottomrule
        \end{tabular}
    }
    \label{tab:llm_art_architecture}
\end{table}

\noindent\textbf{The Role of LLMs.}
LLMs offer new possibilities for supporting, augmenting, and democratizing research in the arts and architecture. These models can analyze and generate text, describe images, interpret creative language, and assist in ideation. Their ability to understand and produce language-rich content makes them particularly useful in domains that rely on creative description, interpretation, or storytelling.

In the arts, LLMs are being used to co-create visual and literary works, simulate historical or fictional voices, summarize critical interpretations, and generate poetic or narrative content. In architecture, LLMs assist in conceptual design, help interpret building codes, summarize design briefs, and even simulate conversations with historical architects. Multimodal extensions of LLMs (e.g., combining text with image or spatial data) further enhance their role in creative and analytical workflows.

\noindent\textbf{Limitations of LLMs.}
Despite their growing adoption, LLMs introduce challenges when applied to creative disciplines: \textit{Creativity Boundaries:} LLMs can generate stylistic content but lack true originality, intentionality, or cultural grounding. Their outputs may mimic but not innovate meaningfully without human curation. \textit{Context Loss:} Art and architecture are deeply tied to historical, social, and cultural contexts. LLMs often fail to retain or interpret this embedded context, leading to surface-level or anachronistic outputs. \textit{Attribution and Authorship:} The use of LLMs in co-creation raises ethical questions about authorship, originality, and credit—especially in arts communities where individual expression is core. \textit{Evaluation Challenges:} Unlike scientific domains, there are no objective benchmarks for creativity or artistic value, making it hard to evaluate LLM-generated contributions meaningfully.

Thus, LLMs should be viewed as collaborators or assistants in the creative process—not as substitutes for human imagination, expertise, or cultural insight.

\noindent\textbf{Taxonomy.}
Based on the current literature, we outline a taxonomy of how LLMs are being applied across different artistic and design domains:

\begin{itemize}[leftmargin=10pt]

      \item\textbf{Visual Arts.}
            Visual arts include creative practices such as painting, photography, illustration, and video. LLMs assist artists by generating image prompts, describing visual scenes, and helping with conceptual development. They also support analysis by interpreting symbolism, classifying art styles, and summarizing critical commentary.

      \item\textbf{Literary Arts.}
            Literary arts encompass creative writing forms such as poetry, drama, fiction, and essays. LLMs can generate literary content, mimic specific authorial styles, and assist in drafting narratives. They also enable textual analysis, such as thematic extraction, stylistic comparison, and literary critique.

      \item\textbf{Performing Arts.}
            Performing arts include music, dance, theater, and opera—forms that rely on embodied performance. LLMs can generate scripts, lyrics, or librettos, and simulate performative dialogue for interactive settings. They also help tag and analyze archival materials, enabling exploration of movement, expression, and performance history.

      \item\textbf{Architecture.}
            Architecture focuses on designing functional and aesthetic spaces, integrating art, engineering, and environmental factors. LLMs support architects by generating design narratives, proposing spatial ideas, and interpreting project briefs. They also assist in summarizing regulations, comparing architectural styles, and evaluating design decisions.

\end{itemize}

\subsubsection{Visual Art}
Visual art is formally defined as any art form that is primarily visual in nature, encompassing disciplines such as painting, drawing, photography, sculpture, printmaking, and video art. It emphasizes the use of imagery, color, composition, and form to convey meaning, emotion, or aesthetic value. In simpler terms, visual art is what we usually think of when we visit an art gallery or museum—works that are meant to be seen. These can include a painting on canvas, a photograph on a wall, or even a digital video installation. Visual art can be realistic or abstract, decorative or political, and it often reflects how artists interpret the world around them.

\noindent\textbf{Visual Art Creation.}
Recent advancements in LLMs have significantly enhanced the capabilities of AI in visual art creation. Notably, models such as MiniGPT-4 \cite{zhu2023minigpt}, LLaVA \cite{liu2023visual}, and ArtGPT-4 \cite{yuan2023artgpt} demonstrate varied architectural innovations to improve artistic image understanding and generation. MiniGPT-4 employs a frozen vision encoder and LLM backbone, utilizing a Q-former for feature mapping, which facilitates efficient training but limits alignment flexibility. LLaVA overcomes this by enabling partial fine-tuning of the LLM, achieving better multimodal integration at the cost of computational efficiency. ArtGPT-4 innovates further by introducing trainable image adapters within the LLM architecture, allowing for parameter-efficient fine-tuning that preserves the interpretive depth of artistic content. These adapters, implemented with bottleneck structures and placed after attention layers, help the model capture subtle artistic features and emotional undertones with human-like sensitivity. Moreover, ArtGPT-4's superior performance on benchmarks such as ArtEmis and ArtMM demonstrates its proficiency in conveying nuanced emotional and aesthetic qualities of visual art.

\noindent\textbf{Visual Art Analysis.}
Also, the advancements in LLMs have significantly enhanced the capacity for automated visual art analysis by integrating visual perception with linguistic reasoning. GalleryGPT \cite{bin2024gallerygpt}, a specialized LMM fine-tuned on a high-quality dataset called \textit{PaintingForm}, exemplifies this trend by addressing the limitations of traditional image-text models which often exhibit ``LLM-biased visual hallucination''—an overreliance on memorized textual knowledge rather than genuine visual understanding. By focusing on formal analysis—encompassing visual features like composition, color, and light—GalleryGPT leverages structured supervision to interpret and describe artworks based purely on their visual elements. Meanwhile, CognArtive \cite{khadangi2025cognartive} explore an alternative strategy by translating image content into SVG-based textual formats, enabling off-the-shelf language models to perform visual reasoning, classification, and even generative tasks without dedicated visual encoders \cite{cai2023leveraging}. These methods highlight a paradigm shift from recognition-centric models to visually-grounded analytical systems, empowering LMMs to articulate nuanced, expert-level interpretations of art.

\subsubsection{Literary Art}
Literary art refers to creative works expressed through written or spoken language, including genres such as poetry, fiction, drama, and essays. It is characterized by the imaginative use of words to convey ideas, emotions, and stories, often with attention to aesthetics, structure, and cultural significance. Put simply, literary art is the art of storytelling, whether through novels, plays, or poems. It helps people share experiences, express feelings, and explore ideas using language. We encounter literary art not only in classic literature but also in everyday forms like spoken word poetry or online short stories.

\noindent\textbf{Literary Art Creation.}
The interdisciplinary methods presented in recent studies highlight how LLMs can significantly enhance literary art creation by offering a dynamic interplay between creative generation and critical interpretation. The work \cite{shanahan2023evaluating} demonstrates that techniques such as creative dialogue, temperature modulation, and multi-voice prompting can elicit complex, stylistically nuanced texts from LLMs, which in turn become fertile material for literary critical analysis. These methods invite iterative engagement, mirroring traditional author-critic interactions and foregrounding the process of refinement and self-critique. Literary critical frameworks—concerned not with appraisal but with interpretive depth—allow researchers to investigate elements like diction, narrative coherence, and stylistic innovation in AI-generated outputs, thereby moving beyond reductive measures of quality or creativity. Similarly, the evaluation protocols outlined in the comparative study of LLMs on creative writing tasks emphasize human-centered assessment across dimensions of craft, originality, and stylistic fidelity, reaffirming the necessity of qualitative, rubric-based evaluation to capture the performative and affective aspects of literary creation \cite{gomez-rodriguez-williams-2023-confederacy}.

\noindent\textbf{Literary Art Analysis.}
Recent advancements in LLMs have catalyzed innovative methodologies for facilitating literary art analysis. \cite{shanahan2023evaluating} demonstrate that techniques such as interactive prompting, temperature modulation, and multi-voice generation not only enable the generation of literarily rich texts but also support critical examination through a literary lens. Their qualitative evaluation \cite{wang2024evaluating} reveals that creative dialogue between a user and an LLM, characterized by iterative feedback and stylistic refinement, mirrors traditional pedagogical practices in creative writing. Modulating the temperature parameter allows for varying degrees of stylistic experimentation, producing texts ranging from conventional to lexically and syntactically avant-garde, thus expanding the interpretive canvas for literary critics. Moreover, the self-reflective, multi-voice generation method \cite{yang2024analyzing}—where the model simulates both authorial and critical roles—exemplifies a novel approach to self-critique and textual development, further blurring the lines between analysis and creation. These methods not only challenge conventional notions of authorship and creativity but also provide new tools for exploring intertextuality, narrative voice, and stylistic nuance within computational frameworks.

\subsubsection{Performing Art}
Performing art is defined as artistic expression conveyed through live performance, typically involving the human body, voice, or presence, and includes disciplines such as music, dance, theatre, and opera. These art forms unfold over time and are often intended for an audience, emphasizing temporality, embodiment, and interaction. In everyday terms, performing art includes concerts, plays, dance shows, or operas—any art form you watch and experience as it happens. It can be scripted or improvised, classical or experimental, and often brings together multiple art forms like movement, music, and storytelling.

\noindent\textbf{Performing Art Creation.}
LLMs significantly influenced performing art creation, particularly in scriptwriting and dramaturgy. One prominent example is \textit{Dramatron} \cite{mirowski2023co}, a hierarchical story generation tool that leverages LLMs to co-author theatre scripts and screenplays with human writers. By structuring content generation across multiple layers---from log lines and character descriptions to plot outlines and dialogues---Dramatron addresses the challenge of long-term semantic coherence, which is often a limitation in traditional flat LLM generations. \textit{Auto-Drama} \cite{wu2024role} significantly advances the field of interactive drama by enabling dynamic, immersive storytelling experiences. The authors introduce a comprehensive framework for \textit{LLM-based interactive drama}, grounded in six redefined Aristotelian dramatic elements: plot, character, thought, diction, spectacle, and interaction. This model facilitates rich, player-driven narrative construction where users can engage directly with characters and environments. Key methodological innovations, such as the \textit{Narrative Chain}, offer granular control over narrative development by segmenting the storyline into interconnected sub-narratives, allowing seamless plot progression while preserving player autonomy.

\noindent\textbf{Interaction Performance.}
The methods \cite{Branch2024DesigningAE} employed in the \textit{Improbotics} study demonstrate significant advancements in facilitating interactive performance between AI and human actors, particularly within the dynamic context of improvisational theatre. By integrating LLMs into live performances through a human-in-the-loop system, the researchers enabled real-time, multi-party dialogue that simulates authentic co-creative interaction. The system employed speech recognition, manual metadata input, and curated line selection, allowing the AI (presented as a ``Cyborg'') to contribute meaningfully to spontaneous scene development. This architecture addresses core challenges of multi-party conversational AI, such as speaker identification and context retention, by relying on both automated and human-mediated input. Iterative design with actor feedback ensured that the AI's role evolved from a comedic object to a viable ensemble member, encouraging collaboration rather than mere novelty. The curated stream of LLM-generated responses enabled performers to incorporate AI contributions fluidly, supporting narrative coherence and audience engagement. Furthermore, game formats such as ``Speed Dating'' and ``Wedding Speech'' were strategically developed to evaluate the AI's responsiveness to social cues and narrative arcs. While limitations in speech recognition and delayed response selection occasionally hindered fluidity, the hybrid human-AI interaction model fostered audience curiosity and performer creativity. Overall, this approach reframes AI from a static content generator to a dynamic performance partner, illustrating the potential of LLMs in embodied, context-sensitive artistic collaboration.

\subsubsection{Architecture}
Architecture is formally defined as the art and science of designing and constructing buildings and physical spaces that fulfill functional, structural, and aesthetic requirements. It involves the integration of form, environment, culture, and technology to shape the built environment. More simply, architecture is what surrounds us every day—from houses and schools to skyscrapers and public plazas. It is not just about making buildings stand up, but about how they make us feel, how they function, and how they reflect the identity of communities. Good architecture blends beauty, usefulness, and meaning.

\noindent\textbf{Architecture Design and Creation.} Architectural design is facilitated by methods like Architext \cite{Galanos2023ArchitextLG}, which leverage LLMs to generate valid and diverse floorplans from natural language prompts. By fine-tuning PLMs on synthetic semantic representations of layouts, Architext enables intuitive and scalable design workflows, reducing dependency on expert knowledge or specialized software. Its models consistently achieve high validity and correctness, demonstrating that language-based generation can serve as an effective tool for structured design tasks, transforming conceptual architectural descriptions directly into usable layout configurations. The work \cite{Ma2024ExploringTC} facilitate architecture design and creation by enabling the generation of large, diverse sets of conceptual solutions through LLMs. By manipulating generative parameters and employing various prompt engineering strategies—including critique prompting and few-shot learning—the research demonstrates how LLMs like GPT-4 can support early-stage design ideation. These approaches provide designers with rapid access to broad solution spaces, enhancing creativity and overcoming fixation during concept development stages. In addition, \cite{Dhar2024CanLG} facilitate architectural design by enabling the generation of Architecture Decision Records (ADRs) from contextual inputs. Through zero-shot, few-shot, and fine-tuning approaches, LLMs can assist architects in documenting architectural decisions efficiently. Notably, fine-tuned smaller models like Flan-T5 demonstrate performance comparable to large models, offering scalable, privacy-conscious solutions. These methods enhance knowledge management and streamline decision-making, though human oversight remains essential for quality assurance.

\noindent\textbf{Architecture Analysis.} 
ArchGPT \cite{Zhang2024ArchGPTHL} facilitates architecture analysis by leveraging LLMs to parse user intents, retrieve domain-specific knowledge, and coordinate external tools. Through task-specific modules—such as image classification, semantic retrieval, and visual rendering—ArchGPT streamlines complex renovation tasks. Its structured workflow enhances interdisciplinary communication, automates heritage assessment, and ensures compliance with architectural guidelines. This methodology bridges expert knowledge with AI reasoning, offering efficient, adaptive solutions for heritage conservation and restoration.

\subsubsection{Benchmarks}

\begin{table*}[!t]
    \centering
    \caption{Overview of Benchmarks for Evaluating LLMs and Generative Models in Art and Architecture Domains}
    \label{tab:art-architecture-benchmarks}
    \resizebox{\linewidth}{!}{
        \begin{tabular}{lp{3.8cm}p{4.1cm}p{4.2cm}p{3.8cm}}
            \toprule
            \textbf{Benchmark} & \textbf{Scope and Focus} & \textbf{Data Composition} & \textbf{Evaluation Tasks} & \textbf{Key Insights} \\ \midrule

            \textbf{ArtBench-10} \cite{liao2022artbench} & Artwork generation and style classification & 60{,}000 curated images spanning 10 artistic styles with balanced classes and cleaned labels & Generative modeling using GANs, VAEs, diffusion; FID, IS, KID, precision/recall scores & StyleGAN2+ADA achieves best results; highlights diversity and fidelity gaps in GAN variants \\ \midrule

            \textbf{AKM} \cite{Dhar2024CanLG} & Architectural design decisions from text prompts & Context-rich design scenarios and model inputs for GPT/T5-family models & Zero-shot, few-shot, and fine-tuned generation of architectural decisions & GPT-4 performs best in zero-shot; GPT-3.5 effective with few-shot; Flan-T5 benefits from fine-tuning \\ \midrule

            \textbf{ADD (DRAFT)} \cite{dhar2025draft} & Domain-specific architectural decision generation & 4{,}911 real-world Architectural Decision Records (ADRs) with labeled rationale and context & Few-shot vs. RAG vs. fine-tuning using DRAFT for architectural reasoning & DRAFT outperforms baselines in accuracy and efficiency; avoids reliance on large proprietary LLMs \\ \midrule

            \textbf{AGI \& Arch} \cite{ploennigs2024generative} & Generative AI's knowledge of architectural history & 101M+ Midjourney prompts + qualitative and factual analysis of style descriptions & Historical style recognition, hallucination detection, generative image-text alignment & ChatGPT shows inconsistencies in confidence vs. accuracy; Midjourney trends analyzed via reverse prompts \\ \midrule

            \textbf{WenMind} \cite{cao2024wenmind} & Chinese Classical Literature and Language Arts (CCLLA) & 42 fine-grained tasks across Ancient Prose, Poetry, and Literary Culture; tested on 31 LLMs & Question answering, translation, rewriting, interpretation across genres & ERNIE-4.0 best performer with 64.3 score; major gap in LLM proficiency for classical Chinese content \\ \midrule

            \textbf{AIST++} \cite{li2021ai} & Music-conditioned 3D dance motion generation & 5.2 hours of 3D motion data across 10 dance genres + musical accompaniment & Motion synthesis with FACT transformer; evaluation via FID, diversity, beat alignment & FACT model generates realistic, synchronized, long-sequence dances better than prior baselines \\

            \bottomrule
        \end{tabular}
    }
\end{table*}

\textbf{ArtBench-10} \cite{liao2022artbench} is introduced as the first standardized, class-balanced, and high-quality benchmark dataset for artwork generation. It consists of 60,000 images spanning 10 distinct artistic styles and addresses common issues in previous datasets, such as class imbalance, noisy labels, and near-duplicates. The authors provide extensive benchmarking experiments using various generative models, including GANs, VAEs, and diffusion-based methods, and analyze their performance using metrics like FID, IS, precision, recall, and KID. The results highlight StyleGAN2 + ADA as the leading approach, while Projected GAN struggles with diversity. ArtBench-10 establishes a rigorous framework for evaluating generative models in the context of artwork synthesis

\textbf{AKM} \cite{Dhar2024CanLG} is a benchmark the effectiveness of LLMs in generating architectural design decisions based on a given context. The study evaluates GPT and T5-based models using zero-shot, few-shot, and fine-tuning approaches, measuring their performance against human-level decision-making. Results indicate that GPT-4 performs best in zero-shot settings, producing relevant and accurate decisions, while cost-effective models like GPT-3.5 achieve comparable results with few-shot learning. Additionally, smaller models such as Flan-T5 perform competitively after fine-tuning, suggesting that LLMs can assist in generating architectural decisions but require further research to reach human-level performance and standardization.

\textbf{ADD} \cite{dhar2025draft} The paper evaluates the effectiveness of DRAFT—Domain-specific Retrieval Augmented Few-shot Tuning—for generating Architectural Design Decisions (ADDs). The benchmark includes comparisons against few-shot learning, retrieval-augmented generation (RAG), and fine-tuning across a dataset of 4,911 Architectural Decision Records (ADRs). DRAFT consistently outperforms other approaches in accuracy and relevance while maintaining efficiency. Automated metrics and human evaluations indicate that DRAFT generates high-quality ADDs without requiring proprietary or resource-intensive LLMs, making it a viable solution for organizations facing privacy and infrastructure constraints.

\textbf{AGI \& Arch} \cite{ploennigs2024generative} benchmarks the architectural knowledge of generative AI models—specifically ChatGPT for text generation and Midjourney for image generation—by systematically analyzing their ability to recognize and accurately describe historical architectural styles. Through quantitative assessments, the authors evaluate ChatGPT's reliability in identifying styles, naming architects, and avoiding hallucinations, finding inconsistencies in its confidence versus factual accuracy. Meanwhile, Midjourney's capability to generate images based on architectural style prompts is assessed by reversing the generation process to see if the AI can correctly describe its own outputs. The benchmark includes a large-scale analysis of over 101 million Midjourney queries to determine popular architectural styles and trends. Ultimately, the study exposes the strengths and limitations of generative AI models in preserving and understanding architectural history, offering a structured methodology for evaluating their knowledge fidelity.

\textbf{WenMind} \cite{cao2024wenmind} is a newly proposed benchmark designed to evaluate LLMs in the domain of Chinese Classical Literature and Language Arts (CCLLA). It encompasses three sub-domains—Ancient Prose, Ancient Poetry, and Ancient Literary Culture—spanning 42 fine-grained tasks across multiple question formats and evaluation scenarios. Through rigorous testing of 31 representative LLMs, the study finds that even the best-performing model, ERNIE-4.0, scores only 64.3, highlighting a significant gap in LLM proficiency in CCLLA. The benchmark provides insights into the strengths and weaknesses of various models and underscores the importance of pre-training data in achieving better results. WenMind sets a new standardized baseline for CCLLA research, offering a valuable resource for future advancements in this domain.

\textbf{AIST++} \cite{li2021ai} is the largest multi-modal dataset of 3D dance motion and music, along with the Full-Attention Cross-modal Transformer (FACT) network for generating 3D dance motion conditioned on music. AIST++ serves as a benchmark for music-conditioned dance generation, providing 5.2 hours of 3D motion across 10 dance genres. The FACT model outperforms prior state-of-the-art methods by employing full-attention transformers with future-N supervision, generating realistic and diverse long-sequence motions. The benchmark evaluations include motion quality (FID scores), diversity, and motion-music correlation (Beat Alignment Score), showing that the FACT model produces more natural and synchronized dance movements compared to existing approaches.

\subsubsection{Discussion}

\noindent\textbf{Opportunities and Impact.}
LLMs are catalyzing a profound evolution in how creative disciplines like art and architecture are practiced, analyzed, and reimagined. As demonstrated across visual arts~\cite{yuan2023artgpt, bin2024gallerygpt}, literary arts~\cite{shanahan2023evaluating, gomez-rodriguez-williams-2023-confederacy}, performing arts~\cite{mirowski2023co, Branch2024DesigningAE}, and architecture~\cite{Galanos2023ArchitextLG, Zhang2024ArchGPTHL}, LLMs offer unprecedented capabilities for augmenting creative processes, automating interpretive analysis, and expanding access to specialized knowledge.

By facilitating narrative generation, multimodal interpretation, and conceptual ideation, LLMs can assist artists, writers, performers, and architects in overcoming traditional bottlenecks related to scale, documentation, and cross-disciplinary synthesis. They democratize access to creative exploration, allowing a broader range of individuals—including those without specialized training—to engage with and contribute to artistic and architectural production. Furthermore, LLMs open up new methodological possibilities: structured formal analysis of artworks~\cite{bin2024gallerygpt}, hybrid human-AI dramaturgy~\cite{mirowski2023co}, iterative literary critique~\cite{shanahan2023evaluating}, and language-driven spatial design~\cite{Galanos2023ArchitextLG} exemplify how machine collaboration can enhance and diversify creative inquiry.

\noindent\textbf{Challenges and Limitations.}
Nonetheless, the application of LLMs in artistic and architectural domains raises significant challenges. First, creativity remains bounded: while LLMs can generate stylistic approximations, they lack genuine intentionality, emotional grounding, and cultural specificity, often producing outputs that are derivative or decontextualized~\cite{shanahan2023evaluating, yuan2023artgpt}. Second, contextual sensitivity is a critical limitation. Art and architecture are deeply embedded in social, historical, and material contexts that LLMs may fail to adequately capture or respect, leading to superficial interpretations or anachronistic outputs~\cite{bin2024gallerygpt}.

Authorship and originality also present complex ethical questions. In disciplines where individual vision and innovation are central, the blending of human and machine contributions challenges traditional notions of creative ownership, citation, and value attribution~\cite{mirowski2023co, Branch2024DesigningAE}. Moreover, evaluation remains difficult: unlike objective domains, assessing artistic quality or creative utility is inherently subjective, requiring nuanced, context-sensitive frameworks that go beyond automated metrics~\cite{gomez-rodriguez-williams-2023-confederacy}.

Finally, multimodal integration—especially in architecture and performing arts—remains in its early stages. Current LLM systems still struggle to fully align spatial, temporal, and embodied aspects of creativity within coherent generative frameworks~\cite{Zhang2024ArchGPTHL, Branch2024DesigningAE}.

\noindent\textbf{Research Directions.}
Several promising research directions emerge to address these challenges:

\begin{itemize}[leftmargin=10pt]
    \item \textbf{Contextual Grounding and Cultural Sensitivity.} Developing models fine-tuned on curated, diverse artistic and architectural corpora can enhance cultural depth and historical accuracy, mitigating risks of decontextualization~\cite{yuan2023artgpt, bin2024gallerygpt}.
    
    \item \textbf{Multimodal and Multisensory Integration.} Advancing multimodal architectures that tightly integrate language, visual, spatial, and auditory modalities will be essential for fully capturing the richness of artistic and architectural expression~\cite{liu2023visual, khadangi2025cognartive}.
    
    \item \textbf{Co-Creative and Iterative Systems.} Building interactive frameworks that emphasize iterative collaboration between human creators and LLMs—as seen in dramaturgy~\cite{mirowski2023co} and creative writing~\cite{shanahan2023evaluating}—can preserve human agency while enhancing machine assistance.
    
    \item \textbf{Ethical Authorship and Attribution Frameworks.} Establishing clear guidelines for authorship attribution, ethical usage, and creative credit in LLM-augmented works will be critical to ensuring fair recognition and responsible innovation~\cite{mirowski2023co}.
    
    \item \textbf{Qualitative Evaluation Metrics.} Developing domain-specific, human-centered evaluation rubrics—focused on creativity, authenticity, and emotional resonance—will be necessary to meaningfully assess LLM contributions in artistic fields~\cite{gomez-rodriguez-williams-2023-confederacy}.
\end{itemize}

\noindent\textbf{Conclusion.}
LLMs are expanding the boundaries of creative exploration, offering powerful tools for augmenting artistic production, critical analysis, and interdisciplinary synthesis. Yet their application demands careful attention to contextual sensitivity, creative agency, and ethical responsibility. As art, literature, performance, and architecture increasingly intersect with AI, the future lies not in replacing human creativity, but in forging synergistic partnerships—where machine intelligence extends human imagination, enriches critical engagement, and fosters new forms of cultural expression.

\subsection{Law}

\subsubsection{Overview}

\noindent\textbf{Introduction.} 
From regulating everyday activities such as signing a lease, starting a business, or entering into marriage to safeguarding free speech, resolving disputes, and promoting fairness, law governs the basic institutional and interpersonal relations of society. Law is a system of rules and procedures, implemented and enforced by governments and institutions, that helps order and coordinate how people and organizations interact. These rules are found in many forms: laws passed by legislatures, decisions made by courts, regulations from agencies, and long-standing customs that have legal force~\cite{law2015dictionary,sutera1993history,herrnstein1970law}.

In short, law acts as society's rulebook. It lays out our rights and responsibilities, provides tools to resolve conflicts, and guides behavior in everything from everyday transactions to high-stakes corporate or constitutional decisions. Understanding and applying the law, however, is not always straightforward. Legal documents are often long, complex, and written in specialized language. Figuring out what a regulation means, how a court case applies, or what terms go into a contract often requires time, training, and expertise.

Legal work is deeply text-based. Lawyers and legal professionals spend much of their time reading and writing: interpreting laws, analyzing past court decisions (precedents), drafting agreements, responding to client questions, and preparing arguments. This work traditionally relies on years of professional training, manual research, and reasoning skills developed through legal education and practice~\cite{kennedy1982legal,hutchinson2012defining,levi2013introduction}. Tools like Westlaw and LexisNexis have helped by making legal materials easier to search, but they still depend on users knowing exactly what to look for and how to navigate technical legal databases.
More broadly, legal research refers to the process of discovering, understanding, and analyzing legal information to answer specific questions or develop arguments. It covers a wide range of activities: finding relevant statutes and regulations, identifying controlling precedents from court rulings, comparing legal principles across jurisdictions, and understanding how laws apply in different factual situations. To solve these research problems, legal professionals rely on doctrinal analysis \cite{hutchinson2012defining,bhaghamma2023comparative,md2019legal}, case law reasoning \cite{levi2013introduction,atkinson2005legal,bench2003model}, analogical reasoning \cite{sunstein1993analogical,posner2005reasoning}, and manual document review—skills refined through rigorous training and experience~\cite{kennedy1982legal}.

Despite these advances, legal research still faces unique challenges. A single word in a statute can carry enormous weight, but its meaning often depends on how courts have interpreted it over time. Arguments must be built on precedent, logic, and interpretation, while also accounting for evolving social, political, and institutional contexts. Such nuances, the blend of rigorous logic, textual analysis, and human judgment, are the most important issues in legal research. At the same time, the legal world is facing new challenges. The amount of legal text—court rulings, statutes, regulations, filings, contracts, and more—is growing faster than ever. Legal professionals must now sift through massive amounts of information, stay up to date on changing rules, and make sense of complex relationships across documents. Doing this manually is time-consuming and expensive, and it creates barriers for individuals or smaller organizations that cannot afford high-end legal support.

\begin{figure}[!t]
    \centering
    \includegraphics[width=\linewidth]{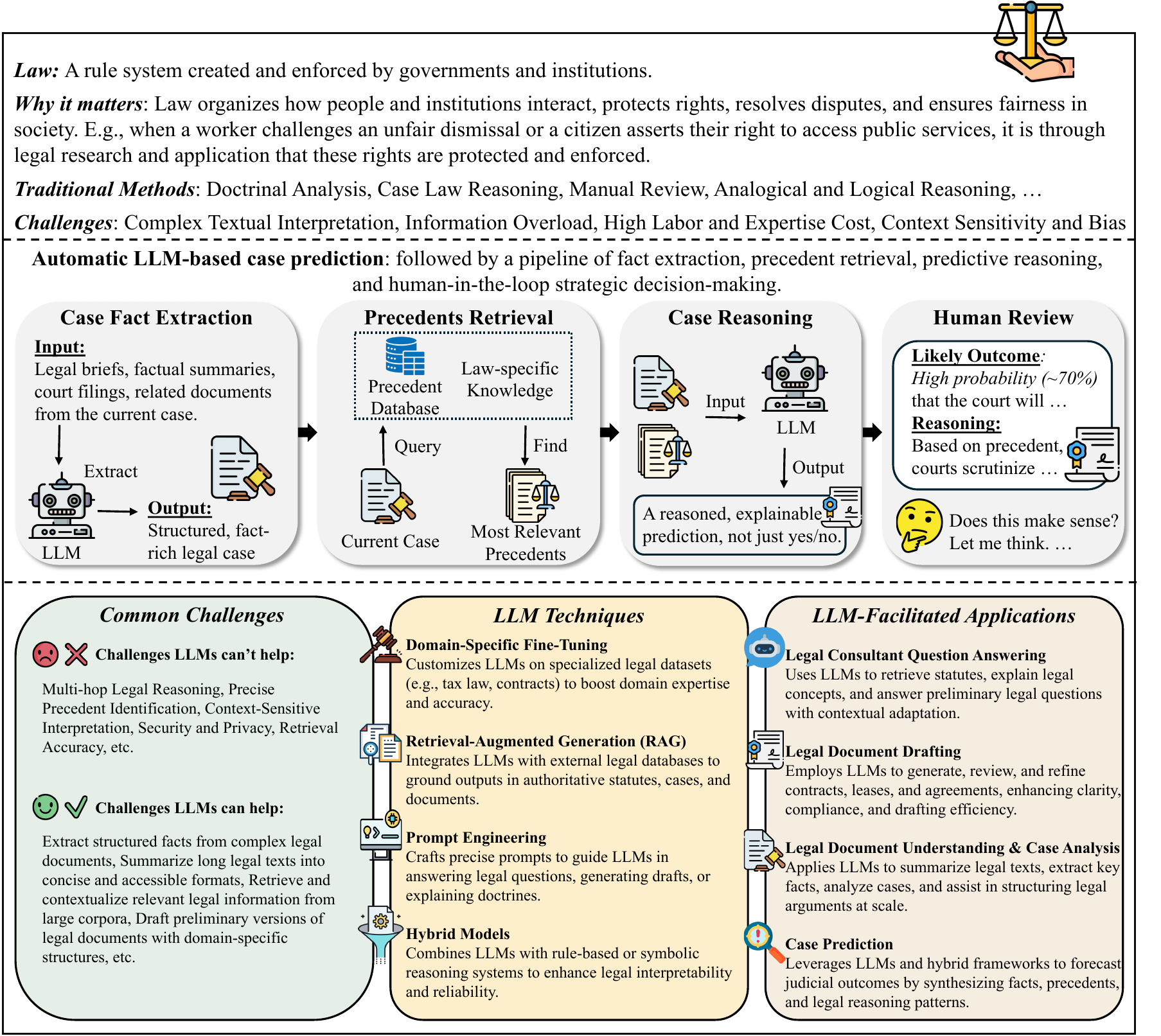}
    \caption{
    \textbf{Overview of LLM Applications in Legal Research.}
    This figure illustrates the core challenges in legal work, key tasks enabled by LLMs, and the dominant techniques used to deploy them. It highlights how LLMs address traditional limitations—such as bias and information overload—by supporting tasks like legal Q\&A, document drafting, case analysis, and prediction. On the methods side, it showcases powerful techniques including Retrieval-Augmented Generation (RAG), Domain-Specific Fine-Tuning, Prompt Engineering, and Hybrid Modeling.
    }
    \label{fig:law-framework}
\end{figure}

\textbf{The role of LLMs.} LLMs are advanced AI systems trained to understand and generate human-like
language.
Their strengths in natural language understanding, summarization, retrieval, and question answering position them well to support legal tasks.
Recent advances show that LLMs can extract structured facts from legal documents, identify key issues, generate coherent legal drafts, and assist users in navigating legal systems~\cite{yao2024survey,li2025system}.
Unlike traditional rule-based and keyword search tools, LLMs can engage with legal text at the level of meaning, making them more adaptable and user-friendly.

\begin{figure}[!t]
    \centering
    \includegraphics[width=\linewidth]{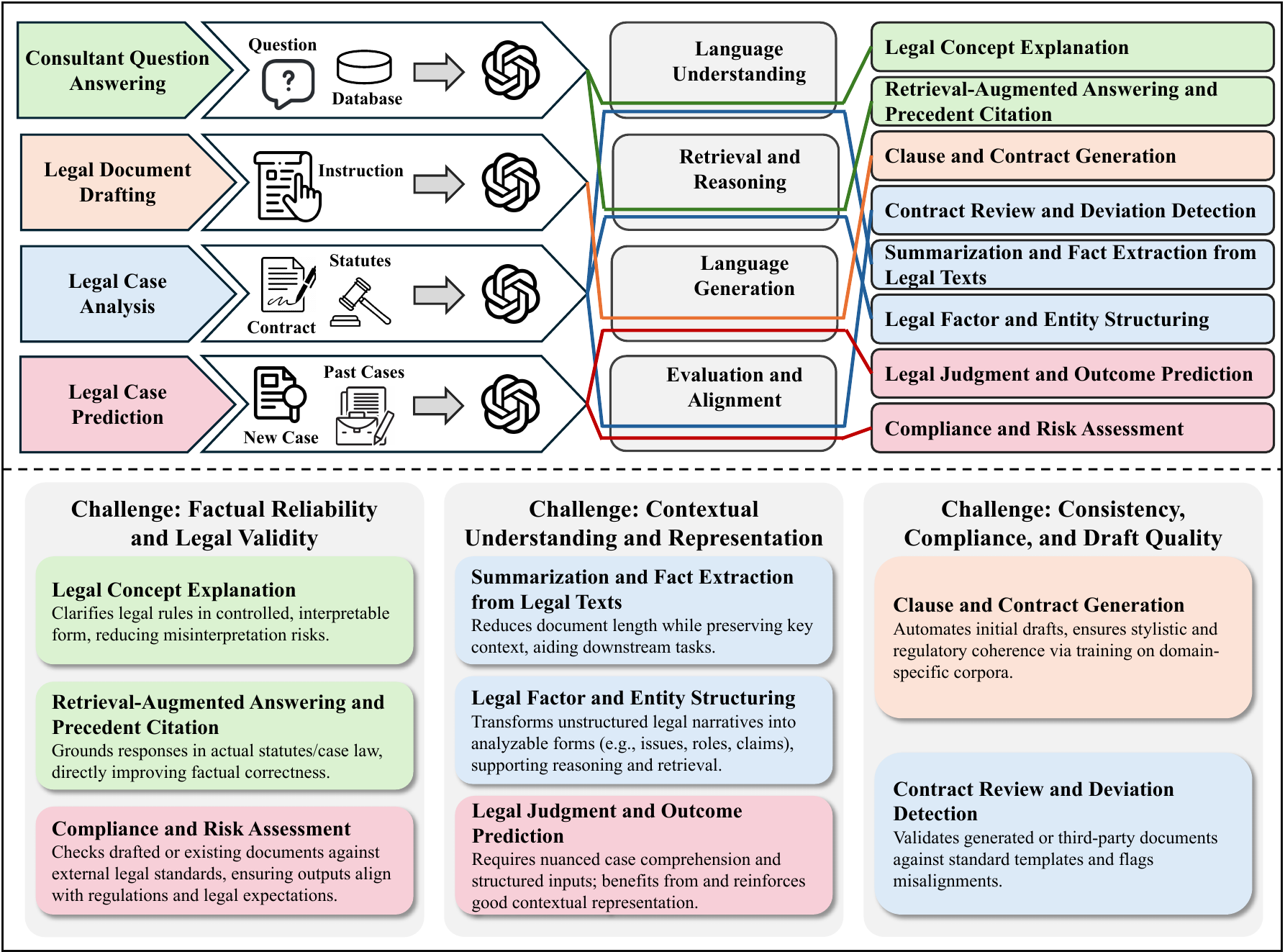}
    \caption{
    \textbf{LLM-Driven Workflow in Legal Tasks.} This figure illustrates the end-to-end paradigm of applying LLMs in various legal processes. For each task—Consultant QA, Document Drafting, Case Analysis, and Case Prediction—the figure outlines typical inputs (e.g., legal queries, cases, statutes), how the LLM processes these inputs, and the corresponding outputs such as query answers, contract drafts, key legal facts, and judicial predictions.
    }
    \label{fig:law-paradigm}
\end{figure}

\textbf{Limitations of LLMs.} Despite their promise, LLMs also face important limitations in legal applications. \textit{Security and Privacy:} Legal documents often contain sensitive or privileged information. Using cloud-hosted LLMs risks inadvertent disclosure of confidential data, raising ethical and regulatory concerns, especially in jurisdictions with strict data protection laws. \textit{Multi-hop Reasoning:} Many legal questions require synthesizing information across multiple sources, e.g., linking a statute to case interpretations and factual contexts. While LLMs are fluent in language generation, they often struggle with the structured, multi-step reasoning needed in legal contexts~\cite{li2025system}. \textit{Retrieval Accuracy:} Legal outcomes frequently hinge on identifying the most relevant precedents. Yet LLMs may retrieve plausible but non-binding cases, miss key controlling authorities, fail to distinguish nuances in legal reasoning, or, in the extreme, produce fictitious, non-existent legal sources via hallucinations.

\noindent\textbf{Taxonomy.} To better understand how LLMs contribute to legal research and practice, we propose a taxonomy that captures a set of core categories of legal tasks where language plays a central role:

\begin{itemize}[leftmargin=10pt]
      \item \textbf{Legal Consultant Question Answering.}
            Many legal queries—such as “What are the elements of negligence?” or “Does the General Data Protection Regulation (GDPR) apply to this situation?”—involve retrieving statutory definitions, summarizing doctrines, or explaining precedent. LLMs can function as legal assistants, offering plain-language explanations, surfacing relevant laws, and contextualizing rules. This enables broader access to legal knowledge and supports both laypersons and professionals during early-stage legal reasoning.

      \item \textbf{Legal Document Drafting.}
            Drafting contracts, policies, and filings involves significant repetition and domain knowledge. LLMs can generate initial templates, propose clauses, and adapt documents to specific scenarios or jurisdictions. This accelerates document production, reduces drafting overhead, and promotes standardization, especially useful for small firms or high-volume legal operations.

      \item \textbf{Legal Document Understanding and Case Analysis.}
            Interpreting statutes, summarizing opinions, or identifying relevant facts is core to legal analysis. LLMs can extract key information, highlight legal entities or issues, and support case comparison. This improves comprehension, reduces time spent on manual review, and helps structure arguments and decisions based on large textual corpora.

      \item \textbf{Case Prediction.}
            Predicting legal outcomes—based on case facts, prior rulings, and jurisdictional context—is valuable for risk assessment and litigation strategy. While final outcomes are shaped by human judgment and evolving law, LLMs can surface patterns, suggest likely outcomes, and support probabilistic reasoning based on precedent, helping users plan and prioritize cases.
\end{itemize}

\begin{table}[!t]
    \centering
    \caption{Applications and insights of LLMs in legal research and practice}
    \resizebox{\linewidth}{!}{
        \begin{tabular}{@{}p{3.8cm} p{4.5cm} p{6.5cm} p{6.5cm} p{2cm}@{}}
            \toprule
            \textbf{Legal Domain} & \textbf{LLM Application Areas} & \textbf{Use Case-Inspired Research Question} & \textbf{Key Insights and Contributions} & \textbf{References} \\
            \midrule

            Legal Consultant Question Answering
                & Interactive Legal Q\&A Systems
                & Can LLMs answer basic legal questions with accurate references to statutes and case law?
                & Emergent legal reasoning observed; retrieval-augmented prompting reduces hallucination; GPT-4 can approximate legal explanations with improved accuracy.
                & \cite{nay2023llm, savelka2023explaining, choi2023llmforempirical} \\\cmidrule{2-5}

                & Domain-Specific Legal Models
                & How can LLMs be fine-tuned to better address legal reasoning tasks?
                & Models like LawLLM improve U.S. law reasoning through fine-tuning and task adaptation (retrieval, precedent matching, judgment prediction).
                & \cite{shu2024lawllm} \\\cmidrule{2-5}

                & Legal Factor Extraction
                & Can LLMs extract and define core legal factors from court opinions?
                & Supports building expert systems; improves structure and consistency of legal analysis.
                & \cite{gray2024using} \\
            \midrule

            Legal Document Drafting
                & Clause Generation
                & Can LLMs autonomously generate domain-compliant legal clauses?
                & LLMs generate grammatically and legally sound clauses; useful in reducing drafting effort.
                & \cite{lam2023contractdrafting, carneiro2025automated} \\\cmidrule{2-5}
            
                & Draft Comparison via NLI
                & How can LLMs verify consistency between generated and template contracts?
                & NLI tasks help identify deviations and inconsistencies, enabling automated review.
                & \cite{narendra2024comparison} \\\cmidrule{2-5}
            
                & Legal Validity of Prompt-Based Contracts
                & What legal risks arise when contracts are generated using prompts?
                & Raises issues with doctrines like the parol evidence rule; prompt provenance matters.
                & \cite{wang2024promptcontracts} \\
            \midrule

            Legal Document Understanding and Case Analysis
                & Document Summarization \& Entity Extraction
                & How well can LLMs extract facts and citations from unstructured legal texts?
                & Enhanced summarization, fact extraction, and legal entity recognition using retrieval-based and fine-tuned models.
                & \cite{chalkidis2020legal, savelka2023explaining, shu2024lawllm} \\\cmidrule{2-5}

                & Large-Scale Legal Analysis
                & Can LLMs support empirical research over large legal corpora?
                & Enables scalable judgment pattern extraction; useful for comparative legal studies.
                & \cite{choi2023llmforempirical} \\\cmidrule{2-5}

                & E-Discovery and Compliance
                & Can LLMs assist in regulatory compliance and legal review at scale?
                & RAG-based systems improve due diligence and compliance decisions; multi-agent LLMs aid in document relevance prediction.
                & \cite{ lahiri2024litigation, chang2024mainrag} \\
            \midrule

            Legal Case Prediction
                & Judgment Forecasting
                & Can LLMs accurately predict legal case outcomes?
                & LLMs outperform traditional models; retrieval-augmented LLMs improve consistency and generalization.
                & \cite{wu2023precedent, cao2024pilot} \\\cmidrule{2-5}
            
                & Hybrid Legal Reasoning
                & How can LLMs be integrated with expert systems to improve prediction accuracy?
                & Hybrid systems improve interpretability and performance by aligning LLM outputs with legal logic.
                & \cite{mumford2022hybrid} \\
            \bottomrule

        \end{tabular}
    }
    \label{tab:legal_llm}
\end{table}

\subsubsection{Legal Consultant Question Answering}

Legal consultant question answering focuses on enabling LLMs to respond to foundational legal inquiries \cite{chalkidis2020legal}. To answer the typical questions previously mentioned ``What are the elements of negligence?'' or ``Does the GDPR apply to this case?'', it requires rapid access to statutes, case law, and established legal doctrines \cite{ashley2017artificial,surden2012machine}. Positioned as an initial layer of legal assistance, these systems provide immediate, scalable, and context-aware guidance, thereby expanding access to legal knowledge and reducing routine workloads for legal professionals \cite{surden2012machine}.
Historically, legal question-answering relied on rule-based systems and keyword search over curated legal databases \cite{ashley2017artificial}. While effective in narrow contexts, these traditional approaches struggle with limited coverage, sensitivity to linguistic variation, and cross-jurisdictional adaptability. Recent advances in LLMs, however, offer promising solutions to these longstanding challenges.

A growing body of research demonstrates the potential of LLMs in specialized legal tasks. Nay et al. \cite{nay2023llm} showed that LLMs can be adapted to function as tax attorneys by supplying domain-relevant legal texts and a small number of examples. Their findings indicate that such adaptation markedly improves performance on complex tax law queries, illustrating the emergent legal reasoning capabilities of LLMs. 
Similarly, Savelka et al. \cite{savelka2023explaining} evaluated GPT-4's ability to explain legal concepts. They found that integrating retrieval mechanisms not only enhanced the clarity of generated explanations but also decreased factual errors, improving the reliability of outputs for legal practitioners.
Building on this direction, Shu et al. \cite{shu2024lawllm} proposed LawLLM, a model explicitly tailored to the U.S. legal domain. LawLLM excels in tasks such as similar-case retrieval, precedent recommendation, and judgment prediction, benefiting from domain-specific fine-tuning and retrieval augmentation. Their work demonstrates how targeted adaptations can support legal decision-making in real-world settings.
Other researchers have explored new forms of legal knowledge extraction. Gray et al. \cite{gray2024using} developed a method for automatically extracting legal factors and definitions from judicial opinions using LLMs, facilitating more efficient legal analysis and improving the construction of expert systems. 
Meanwhile, Choi et al. \cite{choi2023llmforempirical} examined the role of LLMs in empirical legal research, assessing best practices and identifying both benefits and limitations when applying LLMs to large-scale legal document analysis.

Taken together, these studies highlight the rapidly expanding capabilities of LLMs in supporting legal inquiry. They emphasize that while LLMs can substantially democratize access to legal information, effective deployment requires careful contextual adaptation, continuous updating of legal sources, and robust safety mechanisms.

\subsubsection{Legal Document Drafting}

Legal document drafting concerns the generation, refinement, and compliance checking of standardized legal texts—such as contracts, leases, and other agreements—by embedding relevant legal principles into clear and coherent documents \cite{ashley2017artificial,thomas2020construction}. The task demands precise formulation of legal clauses and strict adherence to regulatory requirements to minimize ambiguity, reduce disputes, and ensure consistency across documents \cite{thomas2020construction}. Beyond producing legally sound outputs, an equally important goal is to streamline drafting workflows and decrease the manual burden on legal practitioners \cite{ashley2017artificial}.

Traditionally, legal document drafting has been carried out either manually by trained professionals or through rule-based systems that rely on fixed templates and keyword searches within curated legal repositories \cite{ashley2017artificial,thomas2020construction}. Although effective in constrained settings, these methods are labor-intensive, rigid, and susceptible to human error, often leading to inconsistencies and limited adaptability to evolving legal norms \cite{thomas2020construction}. Recent progress in LLMs offers a compelling alternative: LLM-based systems can interpret unstructured legal language, retrieve relevant information semantically, and generate coherent, legally grounded drafts with substantially less human intervention \cite{chalkidis2020legal,savelka2023explaining}. When enhanced through domain-specific fine-tuning or retrieval-augmented generation, these models further improve in accuracy, consistency, and compliance, thereby increasing both the efficiency and reliability of the drafting process \cite{shu2024lawllm}.

A series of recent studies demonstrates how LLMs are reshaping legal drafting practices. Lam et al. \cite{lam2023contractdrafting} investigate LLM-based assistance for contract clause drafting and find that such models can support clearer, more coherent, and more efficient clause formulation and refinement. Expanding on domain-specific applications, Carneiro-Diaz \cite{carneiro2025automated} shows that LLMs can automatically analyze rental contract clauses, highlighting their capacity to both generate and interpret specialized contractual language.
In addition, Narendra et al. \cite{narendra2024comparison} introduce an LLM-driven document comparison framework that employs Natural Language Inference (NLI) to identify inconsistencies between drafted contracts and reference templates. This method provides an automated pathway for quality assurance, assisting in template verification and contract negotiation. 
Another line of work examines the legal implications of LLM-based drafting. Wang et al. \cite{wang2024promptcontracts} analyze how prompt-driven drafting interacts with traditional legal doctrines and discuss the associated benefits and risks related to authorship, provenance, and enforceability in workflows that rely on LLM-generated prompts.

Collectively, these developments underscore the transformative potential of LLMs in legal document drafting. By automating precise clause generation, enabling systematic quality checks through NLI, clarifying the legal ramifications of prompt usage, and supporting domain-specific drafting needs, LLM-based approaches can help enhance the accuracy, consistency, and compliance of legal documents while substantially streamlining drafting workflows.

\subsubsection{Legal Document Understanding and Case Analysis}

The ability to interpret legal texts (e.g.,  statutes, judicial opinions, and court filings) is central to legal practice and research, particularly for judges, clerks, and scholars \cite{choi2023llmforempirical}. Traditionally, this work has depended on meticulous human review, annotation, and summarization, often supported by rule-based tools or keyword-driven retrieval systems designed to extract relevant facts and citations \cite{ashley2017artificial}. While effective in limited settings, these conventional approaches are slow, inconsistent, and increasingly inadequate for managing the growing volume and complexity of legal documents \cite{ashley2017artificial}.

Recent advances in LLMs have presented powerful capabilities. Modern LLMs can efficiently summarize lengthy legal documents, extract critical facts and citations, identify legal entities, and highlight key issues with notable speed and consistency \cite{chalkidis2020legal,savelka2023explaining}. Leveraging deep semantic understanding and contextual reasoning, these models process unstructured texts to produce coherent, legally grounded analyses, thereby supporting faster and more informed judicial and research workflows \cite{shu2024lawllm}. When enhanced with retrieval augmentation or fine-tuned on domain-specific corpora, LLMs provide even more accurate, context-aware insights for legal document understanding and case analysis \cite{gray2024using,choi2023llmforempirical}.

A growing body of research highlights the expanding role of LLMs in addressing the challenges of legal document understanding and case analysis. Shu et al. \cite{shu2024lawllm} introduced LawLLM, a multi-task system tailored to the U.S. legal domain, which excels not only in judgment prediction but also in retrieving relevant statutes and cases, enabling comprehensive document interpretation. Similarly, Nay et al. \cite{nay2023llm} demonstrated that with access to pertinent legal texts and minimal examples, LLMs can reason through complex tax regulations and case details, showcasing their capacity for specialized legal analysis.
LLMs have also shown promise in scaling empirical legal research. Choi et al. \cite{choi2023llmforempirical} illustrated how LLMs can analyze large legal corpora to extract patterns and summarize judicial opinions. Savelka et al. \cite{savelka2023explaining} evaluated GPT-4’s ability to generate legislative explanations and found that retrieval-augmented methods substantially improve clarity and factual accuracy. Gray et al. \cite{gray2024using} further proposed techniques for automatically extracting key legal factors and definitions from court opinions, producing structured outputs that support expert systems and legal reasoning.
Beyond these foundational contributions, recent studies have extended LLM applications into practical legal workflows, particularly in e-discovery and large-scale document review. Lahiri et al. \cite{lahiri2024litigation} presented DISCOvery Graph, a hybrid framework combining graph neural networks with LLM reasoning to strengthen relevance prediction in litigation document analysis. Chang et al. \cite{chang2024mainrag} introduced MAIN-RAG, a multi-agent filtering approach for retrieval-augmented generation in high-stakes domains such as regulatory compliance, where accuracy and interpretability are critical.
Additional work has explored LLMs in digital forensics and investigative contexts. Wickramasekara et al. \cite{wickramasekara2024forensics} reviewed the potential and limitations of LLMs in forensic investigations, emphasizing challenges of auditability and legal compliance. Yin et al. \cite{yin2025forensics} examined how LLMs may reshape forensic workflows while cautioning against risks such as hallucination and systemic bias. Complementing these analyses, Pai et al. \cite{pai2023exploration} evaluated several open-source LLMs for e-discovery tasks, offering comparative insights into their strengths in summarization, key evidence detection, and domain adaptability.

Taken together, these studies demonstrate the rapidly expanding capabilities of LLM-based approaches in legal document understanding and case analysis. By automating tasks such as summarization, relevance assessment, and legal fact extraction, LLMs promise more scalable, efficient, and consistent legal workflows, while simultaneously underscoring the need for careful domain-specific tuning, robust safety mechanisms, and responsible deployment.

\subsubsection{Legal Judgment Prediction}

Legal judgment prediction represents another central task at the intersection of AI and legal informatics, aiming to forecast judicial outcomes from structured records or unstructured case descriptions. This capability is highly valuable in legal practice, informing litigation strategy, risk evaluation, and policy development. Earlier approaches typically relied on statistical or machine learning models trained on hand-engineered features and structured representations of legal cases~\cite{ashley2017artificial}. While effective in narrow domains, such systems often struggle to generalize across jurisdictions or capture the nuanced reasoning embedded in narrative legal texts.
The advent of LLMs has significantly advanced this field. Unlike traditional models, LLMs can interpret and generate complex legal language, making legal reasoning amenable to formulation as a classification, ranking, or generation task. Their strengths in semantic understanding, contextual reasoning, and few-shot learning position them as powerful tools for judicial prediction, particularly when enhanced through retrieval components or specialized reasoning modules.

A growing line of research highlights the impact of LLMs on improving accuracy, robustness, and interpretability in legal judgment prediction. Cao et al.~\cite{cao2024pilot} proposed the PILOT framework, which incorporates precedent retrieval and temporal modeling to reflect the dynamic evolution of legal doctrines. Tested on the ECHR2023 dataset, PILOT substantially outperformed traditional baselines, underscoring the role of time-sensitive and precedent-aware reasoning.
Wu et al.~\cite{wu2023precedent} introduced the PLJP framework, which aligns LLM-generated outputs with key legal factors extracted from relevant precedents. This approach enhances predictive accuracy while grounding model decisions in structured legal reasoning, thereby improving interpretability.
Shui et al.~\cite{shui2023comprehensive} conducted an extensive empirical evaluation of multiple LLMs on judgment prediction tasks. Their results show that prompting methods—such as incorporating in-context examples or framing predictions as multiple-choice questions—significantly influence model performance. These findings offer practical guidance for optimizing LLM-based prediction without requiring fine-tuning.

To advance interpretability further, Deng et al.~\cite{deng2024adapt} proposed the ADAPT framework, which decomposes legal cases into fact segments, candidate charges, and potential outcomes. This structured reasoning pipeline supports more transparent inference, particularly in complex multi-label settings like criminal judgment prediction.
Additionally, Nigam et al.~\cite{nigam2024rethinking} examined real-world deployment challenges using the Indian Supreme Court corpus. Their study highlights the necessity of retrieval-augmented generation (RAG) and jurisdiction-specific customization, especially in legal systems with unique precedent structures or linguistic characteristics. They demonstrate that models such as GPT-3.5 Turbo can achieve or surpass traditional benchmarks when properly adapted to domain-specific requirements.
Collectively, these contributions demonstrate the transformative potential of LLM-based methods in legal judgment prediction, offering improvements in predictive performance, interpretability, and adaptability across diverse legal contexts.

\subsubsection{Benchmarks}

The evaluation of LLMs in the legal domain has become an increasingly active research area, propelled by the demand for AI systems that can understand, retrieve, and reason over complex legal texts. Compared to general NLP tasks, legal applications require precise reasoning under rigid formal logic, interpretive nuance, and jurisdiction-specific language. Thus, benchmark datasets are essential not only for comparing model performance but also for formalizing what constitutes legal competence in machine systems.

Before the rise of LLMs, legal NLP primarily relied on task-specific benchmarks such as CUAD, CaseHOLD, EUR-Lex, and COLIEE. As summarized in Table~\ref{tab:pre-llm-legal-benchmarks}, these benchmarks typically focused on narrow sub-tasks like contract clause extraction, case conclusion classification, or statute retrieval. Models in this era were mainly based on support vector machines (SVMs), rule-based systems, or fine-tuned BERT variants. Evaluation was conducted using task-specific metrics like accuracy, precision, or recall, often lacking generalization testing or complex inference. 

\begin{table}[htbp]
\centering
\caption{Pre-LLM Era Benchmarks in Legal NLP}
\begin{tabular}{p{2cm} p{2.5cm} p{2.5cm} p{2.5cm} p{4.56cm}}
\toprule
\textbf{Benchmark} & \textbf{Scope and Focus} & \textbf{Data Composition} & \textbf{Evaluation Tasks} & \textbf{Key Insights} \\
\midrule

\textbf{CUAD}
~\cite{hendrycks2021cuad} &
Contract clause extraction and risk detection in legal documents &
13,000+ expert-annotated examples across 41 clause types, sourced from real commercial contracts &
Clause identification, named entity recognition (NER), binary classification &
Focused on practical contract review tasks; emphasized precision in extraction under legal ambiguity; widely used in contract AI \\
\addlinespace

\textbf{CaseHOLD}
~\cite{zheng2021does} &
Case law judgment understanding via conclusion prediction &
53,000+ U.S. appellate court case summaries with multiple-choice legal holdings &
Multiple-choice question answering; outcome selection from candidates &
Tests nuanced legal entailment and fact-to-holding inference; served as early benchmark for transformer-based legal models \\
\addlinespace

\textbf{EUR-Lex
}~\cite{loza2010efficient} &
Multilabel legal topic classification for EU directives and regulations &
55,000+ European legal documents tagged with 3,956 EuroVoc labels &
Multilabel text classification &
One of the earliest and most cited legal NLP datasets; highly imbalanced and hierarchical label space inspired development of label-aware classifiers \\
\addlinespace

\textbf{SCOTUS}
~\cite{spaeth1994supreme} &
Supreme Court decision classification and ideological alignment analysis &
U.S. Supreme Court opinions, annotated with justice ideology, vote splits, and case topics &
Binary and multiclass classification, ideological trend analysis &
Used in political science and legal prediction; supported early quantitative legal studies using machine learning \\
\addlinespace

\textbf{COLIEE}
~\cite{rabelo2022overview} &
Legal information retrieval and entailment challenge (competition format) &
Multiple years of formal tasks including Japanese Bar Exam questions and Canadian legal cases/statutes &
IR, entailment classification, statute retrieval, legal QA &
Serves as international benchmark challenge; evaluated both retrieval and inference under strict logic constraints \\
\bottomrule
\end{tabular}
\label{tab:pre-llm-legal-benchmarks}
\end{table}

With the emergence of foundation models such as GPT series, and their multilingual counterparts, the research and practitioner communities began designing new benchmarks that match the scale and depth of LLM capabilities. These new benchmarks reflect a shift from task-isolated pipelines toward comprehensive, multi-task and multi-language evaluations emphasizing zero-shot, few-shot, and instruction-following reasoning. Table~\ref{tab:legal-benchmarks} summarizes six representative legal benchmarks developed in the LLM era: LegalBench, LawBench, LexGLUE, LAiW, Swiss-Judgment-Prediction, and LJP-IV. These benchmarks collectively span English and Chinese legal systems, multiple legal traditions (common law, civil law), and a variety of task formats (classification, retrieval, generation, and reasoning). Each is designed to capture a particular aspect of legal intelligence, including factual recall, statutory alignment, interpretive inference, and multilingual adaptability.

\begin{table}[htbp]
\centering
\caption{Overview of Benchmarks for Evaluating LLMs in the Legal Domain}
\begin{tabular}{p{2cm} p{2.5cm} p{3cm} p{2.5cm} p{4.5cm}}
\toprule
\textbf{Benchmark} & \textbf{Scope and Focus} & \textbf{Data Composition} & \textbf{Evaluation Tasks} & \textbf{Key Insights} \\
\midrule
\textbf{LegalBench}
~\cite{guha2023legalbench} &
Legal reasoning across 162 subtasks from diverse real-world legal contexts &
Expert-authored prompts covering six reasoning types (e.g., rule recall, rule application, interpretation) &
Zero-shot and few-shot reasoning, structured answer generation &
Provides a fine-grained taxonomy of legal reasoning skills; reveals that even top-tier LLMs struggle with multi-step statutory logic and implicit legal inference, making it ideal for stress-testing general-purpose LLMs in high-reasoning settings \\
\addlinespace

\textbf{LawBench}
~\cite{fei2024lawbench} &
Legal knowledge probing for Chinese LLMs across memory, understanding, and application tiers &
20 distinct tasks, including classification, extraction, and generation; built from 90K+ QA pairs sourced from Chinese laws and judicial documents &
Legal QA, clause classification, summarization, legal sentence rewriting &
Designed to test hierarchical legal comprehension in Chinese LLMs; emphasizes generalization limitations across task types and uncovers alignment gaps in open-source vs. proprietary models \\
\addlinespace

\textbf{LexGLUE}
~\cite{chalkidis2022lexglue} &
English legal language understanding and document classification across European legal systems &
Seven curated datasets (e.g., ECtHR, EUR-Lex, SCOTUS) labeled with case outcomes, topics, and rulings &
Multiclass and multilabel classification, case judgment prediction &
Pioneered standardization of English-language legal NLP tasks; facilitates benchmarking across both general-purpose LMs and legal-domain-pretrained models; bridges legal and general NLP evaluation efforts \\
\addlinespace

\textbf{LAiW
}~\cite{laiw2023github} &
Comprehensive evaluation suite for Chinese legal LLMs across major task families &
Tasks include judgment prediction, statute extraction, article matching, and legal QA based on Chinese court records and codes &
Classification, matching, QA, retrieval, generation &
Captures real-world judicial task structures in China’s legal system; reveals domain misalignment in Chinese LLMs and encourages community efforts in localized, high-quality legal datasets \\
\addlinespace

\textbf{Swiss-Judgment-Prediction}
~\cite{niklaus2021sjp} &
Multilingual legal case outcome prediction in Swiss federal courts &
Over 85K real-world legal cases in German, French, and Italian, annotated with ruling outcomes and metadata &
Judgment prediction, multilingual classification, temporal generalization &
Highlights challenges in cross-lingual legal learning and domain temporal drift; supports development of multilingual legal LLMs, especially for underrepresented legal languages \\
\addlinespace

\textbf{LJP-IV}
~\cite{zhang2024beyondguilt} &
Fine-grained Chinese criminal verdict prediction with inclusion of “innocent” cases &
Extended version of Chinese LJP dataset with new trichotomous labels (guilty, partially guilty, innocent) &
Few-shot, multi-label classification, reasoning over multiple charges &
Fills gap in prior LJP datasets that lacked nuanced outcome granularity; pushes models to reason over exoneration cases and subtle charge distinctions in realistic criminal scenarios \\
\bottomrule
\end{tabular}
\label{tab:legal-benchmarks}
\end{table}

\textbf{LegalBench.} LegalBench is one of the most fine-grained and widely used benchmarks for legal reasoning, introduced by the Hazy Research group at Stanford. It contains 162 subtasks across six categories of legal reasoning (rule recall, rule application, contextual interpretation, rhetorical analysis, etc.), with prompts written by legal experts. LegalBench emphasizes few-shot and zero-shot settings, testing whether models can generalize to complex legal logic without task-specific fine-tuning. Evaluation is primarily based on accuracy and exact match. Leaderboard results are presented in Table~\ref{tab:legalbench-leaderboard}.

\textbf{LawBench.} LawBench evaluates Chinese legal models through 20 distinct tasks structured across three tiers: memory, comprehension, and application. It features over 90,000 QA pairs built from legal statutes and court rulings. It is designed to examine how well Chinese LLMs can perform legal retrieval, clause classification, and open-domain legal analysis. It serves as a counterpart to LegalBench in the Chinese legal environment.

\textbf{LexGLUE.} LexGLUE unifies seven English-language legal datasets (including ECtHR, EUR-Lex, and SCOTUS) into a benchmark suite for legal classification and judgment prediction. It supports both general-purpose and domain-pretrained models, providing a standardized platform for evaluating legal NLP systems. LexGLUE helped establish best practices for legal model development and evaluation in European and U.S. contexts.

\textbf{LAiW.} The LAiW benchmark targets typical Chinese legal tasks such as statute retrieval, judgment prediction, and article matching. Based on real-world Chinese judicial data, it simulates courtroom decision flows and legal reasoning chains. LAiW exposes domain adaptation gaps in general Chinese LLMs and highlights the need for legal domain alignment in foundation models.

\textbf{Swiss-Judgment-Prediction.} This benchmark addresses multilingual judgment prediction in Switzerland’s trilingual legal system (German, French, Italian). It consists of over 85,000 annotated cases and emphasizes robustness to temporal drift and cross-lingual inference. It enables the evaluation of legal AI systems in multi-jurisdictional settings where language and law evolve concurrently.

\textbf{LJP-IV.} LJP-IV introduces a third label—“innocent”—to the widely used Chinese legal judgment prediction datasets, enabling trichotomous classification (guilty, partially guilty, innocent). It emphasizes fine-grained reasoning and helps investigate fairness and model bias in criminal judgment prediction.

\textbf{Evaluation tasks and metrics.} Across both pre-LLM and post-LLM benchmarks, common legal NLP tasks include legal question answering, statutory article retrieval, case outcome prediction, legal summarization, and precedent matching. Evaluation metrics vary by task: classification tasks use accuracy, F1, and exact match; generation tasks use BLEU, ROUGE, or GPTScore; retrieval tasks use recall@K or mean average precision (MAP). Notably, post-LLM benchmarks increasingly incorporate zero/few-shot reasoning settings and human evaluation of legal correctness and explainability.

\textbf{LegalBench leaderboard.} Among all benchmarks, LegalBench is the most prominent for leaderboard-based performance comparison. Table~\ref{tab:legalbench-leaderboard} reports the results of leading models such as Gemini 2.5 Pro Exp, GPT-4.1, and Grok 3. While Gemini Pro leads in accuracy, cost-effective models like Gemini Flash Preview deliver comparable results with lower latency and API cost. This suggests that highly competitive legal reasoning models can now be deployed efficiently at scale.

\begin{table}[htbp]
\centering
\caption{LegalBench Benchmark Leaderboard (Top 10 Models)}
\begin{tabular}{clccc}
\hline
\textbf{Rank} & \textbf{Model} & \textbf{Accuracy} & \textbf{Cost (In / Out)} & \textbf{Latency (s)} \\
\hline
1 & Gemini 2.5 Pro Exp & 83.6\% & \$1.25 / \$10.00 & 3.51 \\
2 & Gemini 2.5 Flash & 82.8\% & \$0.15 / \$0.60 & 0.43 \\
3 & o3 & 82.5\% & \$10.00 / \$40.00 & 5.14 \\
4 & Grok 3 Mini (High Reasoning) & 82.0\% & \$0.60 / \$4.00 & 4.92 \\
5 & Grok 3 Beta & 82.0\% & \$3.00 / \$15.00 & 0.44 \\
6 & GPT-4.1 & 81.9\% & \$2.00 / \$8.00 & 0.42 \\
7 & Gemini 2.5 Flash (Thinking) & 81.8\% & \$0.15 / \$3.50 & 2.66 \\
8 & o1 Preview & 81.7\% & \$15.00 / \$60.00 & 10.33 \\
9 & Grok 3 Mini (Low Reasoning) & 81.6\% & \$0.60 / \$4.00 & 3.38 \\
10 & DeepSeek V3 & 80.1\% & \$0.90 / \$0.90 & 4.13 \\
\hline
\end{tabular}
\label{tab:legalbench-leaderboard}
\end{table}

\textbf{Summary.} Compared to the pre-LLM era, where legal benchmarks targeted narrow tasks with limited scope and manually tuned models, the post-LLM era features broad, multi-faceted, and more cognitively demanding benchmarks. These new benchmarks reflect the increasing ambition to model real legal reasoning rather than isolated subtasks. Still, current benchmarks remain limited in areas such as procedural simulation, adversarial robustness, and cross-jurisdiction alignment. Bridging this gap will require new benchmarks that reflect practical legal workflows, account for legal change over time, and support real-time AI-human collaboration in high-stakes legal environments.

\subsubsection{Discussion}

\noindent\textbf{Opportunities and Impact.}
LLMs are reshaping the landscape of legal practice by enhancing efficiency, accessibility, and interpretability across core tasks such as legal question answering, document drafting, document understanding and case analysis, and judgment prediction. These models enable the automation of traditionally manual and expert-dependent processes, thereby democratizing access to legal knowledge and reducing costs associated with legal services. For instance, in legal consultant question answering, LLMs offer context-aware and scalable assistance, improving upon the limitations of rule-based systems~\cite{nay2023llm,savelka2023explaining}. In legal document drafting, LLMs support the generation of compliant, precise clauses through semantic understanding and template alignment~\cite{lam2023contractdrafting, carneiro2025automated}.. Legal document understanding and case analysis are similarly enhanced, with models like LawLLM and GPT-4 enabling structured extraction of legal factors and empirical insights from complex legal texts~\cite{shu2024lawllm, gray2024using}. Finally, in legal judgment prediction, models are beginning to capture the nuances of precedent, legal reasoning, and jurisdictional variation through frameworks such as PILOT and PLJP~\cite{cao2024pilot, wu2023precedent}, offering significant utility in litigation planning and risk assessment.

\noindent\textbf{Challenges and Limitations.}
Despite these promising developments, significant challenges remain. One key limitation is factual reliability: LLMs may hallucinate statutes or precedents if not coupled with robust retrieval mechanisms~\cite{savelka2023explaining}. Moreover, the interpretability of model outputs, especially in high-stakes legal decisions, can hinder trust and accountability, particularly when models act as opaque black boxes~\cite{nigam2024rethinking}. Domain specificity also poses a critical barrier; general-purpose models often fail to capture jurisdictional nuances or evolving regulatory standards, requiring continual fine-tuning and curated legal datasets~\cite{carneiro2025automated}. In drafting tasks, questions of legal liability arise when AI-generated text is ambiguous or misaligned with established legal doctrines, as highlighted by concerns surrounding prompt-based generation and doctrines.
Furthermore, in judgment prediction, risks of reinforcing historical biases, legal inequities, or overfitting to precedent without understanding intent must be carefully managed~\cite{deng2024adapt, nigam2024rethinking}.

\noindent\textbf{Research Directions.}
To address these challenges and realize the full potential of LLMs in legal domains, several research directions merit emphasis:

\begin{itemize}[leftmargin=10pt]
    \item \textbf{Retrieval-Augmented and Fact-Verified Generation.} Integrating retrieval-augmented generation (RAG) systems that anchor responses in verifiable legal texts can reduce hallucination and improve accuracy~\cite{gray2024using}.
    
    \item \textbf{Jurisdiction-Aware and Temporal Modeling.} Developing models sensitive to jurisdictional differences and evolving case law, such as time-aware frameworks like PILOT,can enhance the contextual reliability of legal predictions~\cite{cao2024pilot}.
    
    \item \textbf{Human-in-the-Loop Oversight and Interpretability.} Embedding human review processes and generating structured rationales for outputs, particularly in document drafting and judgment prediction, will improve transparency and accountability~\cite{wu2023precedent, deng2024adapt}.
    
    \item \textbf{Ethical and Regulatory Frameworks.} Establishing governance standards for the deployment of LLMs in legal contexts, including audit trails, liability attribution, and responsible AI usage, will be essential to mitigate misuse and legal uncertainty~\cite{ wickramasekara2024forensics}.
    
    \item \textbf{Open-Source and Domain-Specific Legal Datasets.} Continued development and release of domain-specific corpora, especially in underrepresented legal systems, will support equitable research and application~\cite{ choi2023llmforempirical}.
\end{itemize}

\noindent\textbf{Conclusion.}
LLMs are emerging as transformative tools in legal informatics, streamlining workflows, enhancing access to justice, and supporting complex legal reasoning. However, their successful deployment demands careful alignment with legal standards, domain expertise, and interpretability requirements. Rather than replacing legal professionals, LLMs function best as assistive technologies—amplifying legal insight, accelerating analysis, and fostering more equitable, data-informed legal systems.

\newpage
\section{LLMs for Economics and Business}
\label{sec:discpline-2}

In this chapter, we examine how LLMs are applied in economics and business domains. In particular, we review four disciplines: finance, economics, accounting, and marketing. In \textbf{finance}, we cover trading and investment research, corporate finance, market analytics, financial intermediation and risk management, sustainable finance, and fintech, and explain how these tools are evaluated. In \textbf{economics}, we consider behavioral and experimental studies, macroeconomic and agent-based simulation, strategic and game-theoretic interactions, and systems for economic reasoning and knowledge representation, with targeted assessments. In \textbf{accounting} section, we examine auditing, financial and managerial accounting, and taxation, together with benchmarking. 
In \textbf{marketing} section, we review consumer insight and behavior analysis, content creation and campaign design, and market-intelligence and trend analysis, with associated performance benchmarks.

\subsection{Finance}

\subsubsection{Overview}

\noindent\textbf{Introduction.}
Finance is the study of how individuals, institutions, and governments acquire, spend and manage money and other financial resources over time under conditions of uncertainty~\cite{gitman2015principles, brealey2014principles}. It encompasses the mechanisms of asset valuation, the behavior of financial markets, and the strategic decision-making of economic agents under risk. In simpler terms, finance is about managing money—how people and organizations save, invest, borrow, and plan for the future while dealing with risks. Whether it's a family budgeting for a home, a company raising capital, or a trader investing in the stock market, finance provides the principles and tools to make informed decisions.

Finance research covers a broad range of tasks, including asset pricing, portfolio optimization, risk management, financial forecasting, and corporate decision analysis. Traditionally, these tasks are approached using mathematical modeling, econometric techniques, statistical inference, and more recently, machine learning algorithms~\cite{karatzas1998methods, ruppert2011statistics, dixon2020machine}. For example, regression models and time-series analysis are widely used for forecasting~\cite{abu1996introduction, kim2003financial, sezer2020financial}, while stochastic calculus and optimization methods are central to asset pricing and portfolio theory~\cite{korn2001option, lamberton2011introduction}.

Traditional quantitative methods have been instrumental in advancing finance as a rigorous and impactful discipline. Groundbreaking models like the Capital Asset Pricing Model (CAPM)~\cite{sharpe1964capital}, Black-Scholes option pricing~\cite{black1973pricing}, and the Fama-French factor models~\cite{fama1993common, fama2015five}have not only deepened our understanding of market dynamics but also shaped financial regulation, investment strategies, and risk management practices. The integration of economics, statistics, and computation in finance has driven scientific innovation and enabled practical tools used daily by investors, policymakers, and financial institutions.

Despite recent advancements, finance research continues to face persistent and multifaceted challenges. One major hurdle is the explosion of unstructured data, including news articles, earnings call transcripts, regulatory filings, and social media posts~\cite{goldstein2021big}. These sources are rich in information but difficult to analyze systematically due to their variability in format, tone, and relevance. For example, assessing the market impact of a CEO’s offhand comment during an earnings call requires deep contextual understanding beyond traditional models.

Another key challenge is the manual workload involved in processing and analyzing financial data. Tasks like data labeling, report summarization, or sentiment annotation often rely on significant human effort, making them time-consuming, resource-intensive, and prone to error. This bottleneck can slow down decision-making and hinder scalability in fast-paced financial environments.

Finance also operates within a highly dynamic environment, where market conditions, policy landscapes, and investor behaviors change rapidly. Static models trained on historical data frequently fail to adapt to new patterns, such as those observed during financial crises or global events like the COVID-19 pandemic. This volatility complicates the development of robust, long-lasting analytical models.

Finally, interpretability remains a critical concern. Many high-performing models, especially those based on deep learning, function as “black boxes,” offering little transparency into how predictions are made. This lack of clarity can undermine trust in model outputs, particularly in high-stakes decisions such as credit approval, fraud detection, or investment recommendations, where explainability is essential for compliance and stakeholder confidence.

In this context, LLMs represent a significant technological shift. Their ability to process and reason over vast amounts of both structured and unstructured text data positions them as valuable tools for supporting and enhancing financial research~\cite{nie2024survey}.

\begin{figure}[!t]
    \centering
    \includegraphics[width=0.99\linewidth]{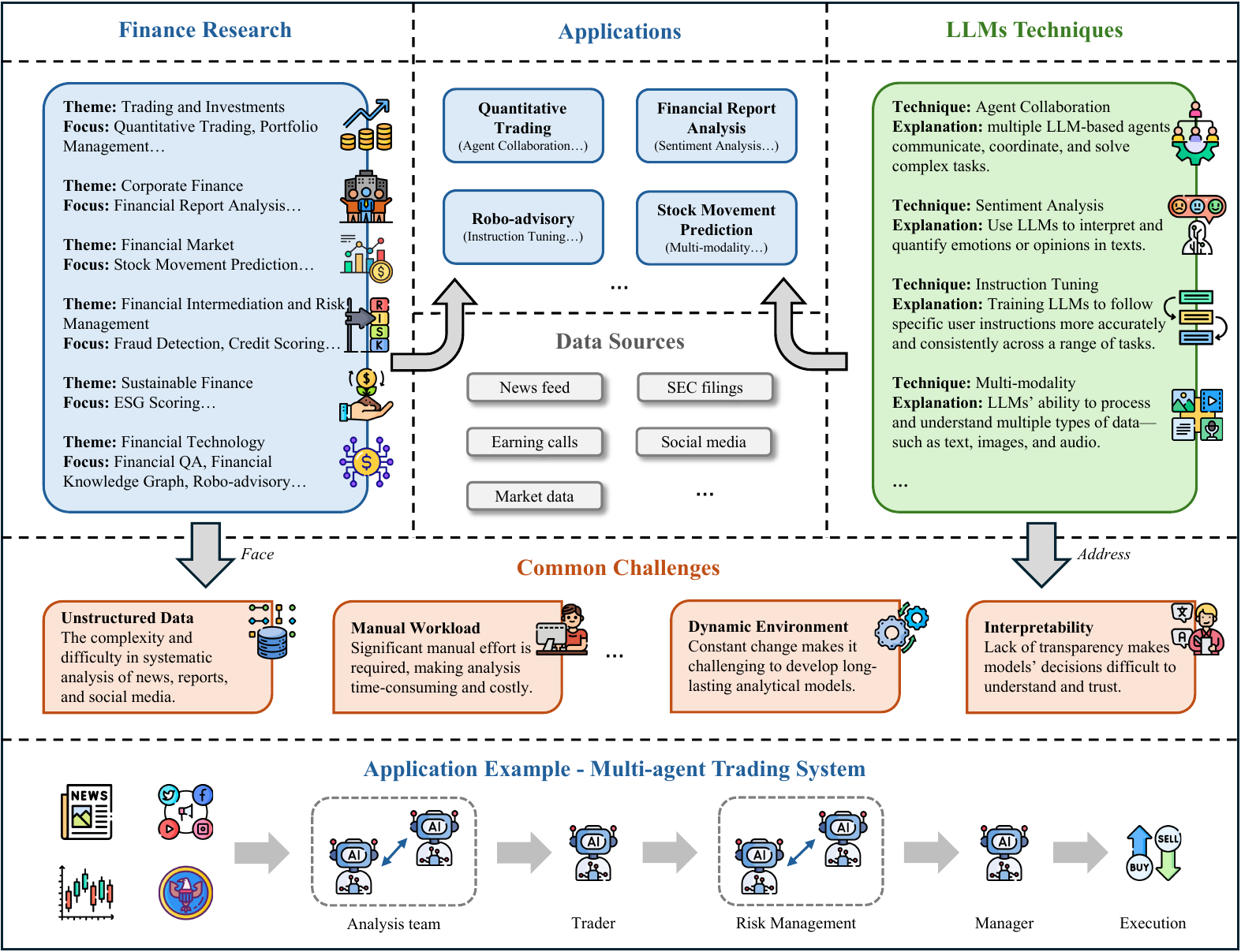}
    \caption{Overview of LLMs' Applications in Finance Research.}
    \label{fig:finance}
    \vspace{-10pt}
\end{figure}

\noindent\textbf{The Role of LLMs.}
However, integrating LLMs into financial research must be approached with caution. Certain problem areas remain beyond their current capabilities. Tasks that require precise numerical computation, high-frequency decision-making, or real-time financial modeling demand low-latency inference, quantitative rigor, and often, regulatory robustness—areas where traditional models still hold a clear advantage. That said, there are specific categories of research problems where LLMs are especially well-positioned to contribute. These include extracting structured information from unstructured financial documents~\cite{li2025extracting}, interpreting and generating textual financial reports~\cite{kim2024financial}, and answering complex domain-specific questions~\cite{islam2023financebench}. The strength of LLMs lies in their ability to process and synthesize large volumes of text, adapt to domain-specific jargon through fine-tuning or prompt engineering, and facilitate interactive exploration of financial knowledge~\cite{yang2024financial}. As such, LLMs should be seen not as replacements for traditional tools, but as powerful complements—particularly in domains where language and knowledge representation play a central role.

\begin{table}[!t]
\centering
\caption{Applications and insights of LLMs in finance research}
\resizebox{\linewidth}{!}{
\begin{tabular}{@{}p{4cm} p{5cm} p{7cm} p{7cm} p{2.5cm} @{}}
\toprule
\textbf{Field} & \textbf{Subfield} & \textbf{Key Insights and Contributions} & \textbf{Examples} & \textbf{Citations} \\
\midrule
\multirow{2}{*}{\parbox[t]{2.8cm}{\textbf{Trading and \\Investments}}}
    & \textbf{Quantitative Trading} 
    & LLM-based trading agents demonstrate enhanced interpretability, adaptability, and profitability in simulating and executing financial strategies across diverse market conditions 
    & FinCon~\cite{yu2024fincon}: Multi-agent LLM system improves trading via verbal risk reinforcement. \newline
    FinMem~\cite{Yu2023FinMemAP}: Layered memory and character design enhance LLM-based trading decisions.
    & \cite{wang2024ExploringLC, fatouros2024can, zhang2024ai, zhang2024multimodal, ding2024large, Gao2024SimulatingFM, kirtac2024sentiment, wang2024quantagent, Yu2023FinMemAP, Li2023TradingGPTMS, yu2024fincon, xiao2024tradingagents} \\
\cmidrule(lr){2-5}
    & \textbf{Portfolio Management} 
    & LLMs enhance financial decision-making by enabling adaptive, explainable, and multi-modal portfolio strategies through agent collaboration, sentiment reasoning, and dynamic alpha mining 
    & Ko \& Lee~\cite{ko2024can}: ChatGPT enhances portfolio diversification and asset selection across classes. \newline
    Kou et al.~\cite{kou2024automate}: LLM-driven agents mine multimodal alphas, dynamically adapt trading strategies.
    & \cite{luo2025llm, abe2024leveraging, unnikrishnan2024financial, gu2024adaptive, kou2024automate, ko2024can, Wu2024PortfolioPB, Wang2023AlphaGPTHI} \\
\midrule
\textbf{Corporate Finance} 
    & \textbf{Financial Report Analysis} 
    & LLMs enhance financial report analysis by enabling accurate, explainable, and scalable extraction and generation through multimodal processing, domain-specific fine-tuning, and tool-augmented reasoning 
    & XBRL-Agent~\cite{han2024xbrl}: LLM agent analyzes XBRL reports using retriever and calculator tools.
    & \cite{le2024auto, han2024xbrl, gomes2024automating} \\
\midrule
\textbf{Financial Markets} 
    & \textbf{Stock Movement Prediction} 
    & LLMs can effectively predict and explain stock movements by extracting sentiment, factors, and insights from financial text through self-reflection, instruction tuning, and domain-specific prompting 
    & Koa et al.~\cite{koa2024learning}: Self-reflective LLM explains stock predictions via reinforcement learning framework. \newline
    Ni et al.~\cite{ni2024harnessing}: QLoRA-fine-tuned LLM predicts stocks using rich earnings data.
    & \cite{lopez2023can, ni2024harnessing, Bhat2024StockPT, wang2024llmfactor, koa2024learning, Zhang2023UnveilingTP} \\
\midrule
\multirow{2}{*}{\parbox[t]{2.8cm}{\textbf{Financial \\ Intermediation \\ and Risk \\ Management}}}
    & \textbf{Fraud Detection} 
    & LLMs significantly enhance financial fraud detection by enabling accurate, scalable, and robust identification of anomalies and manipulations through prompt engineering, hybrid modeling, and adversarial benchmarking 
    & Fraud-R1~\cite{yang2025fraud}: Benchmark tests LLM defenses against multi-round fraud and phishing. \newline
    RiskLabs~\cite{cao2024risklabs}: LLM fuses multi-source data to forecast volatility and financial risk.
    & \cite{yang2025fraud, korkanti2024enhancing, bakumenko2024advancing, boskou4897041exploring, cao2024risklabs} \\
\cmidrule(lr){2-5}
    & \textbf{Credit Scoring} 
    & LLMs enhance credit risk assessment by improving prediction accuracy, generalization, and explainability through hybrid modeling, text integration, and domain-specific fine-tuning
    & CALM~\cite{feng2023empowering}: LLM scores credit risk across tasks with fairness checks. \newline
    LGP~\cite{Teixeira2023EnhancingCR}: Prompted LLMs use Bayesian logic to generate insightful risk reports.
    & \cite{sanz2024credit, drinkall2025forecasting, feng2023empowering, Teixeira2023EnhancingCR} \\
\midrule
\textbf{Sustainable Finance} 
    & \textbf{ESG Scoring} 
    & Leverage LLMs for classification, rule learning, data extraction, greenwashing detection, readability assessment, and multi-lingual understanding, thereby enhancing transparency and decision-making in sustainable finance.
    & ESGReveal~\cite{zou2025esgreveal}: Leverage LLMs and RAG to systematically extract and analyze ESG data from corporate reports.  & \cite{hyojeong2024esg, zhao2024revolutionizing, calamai2025corporate, shimamura2025evaluating, lin2024gpt4esg, birti2025optimizing, tian2024esg, zou2025esgreveal, gu2024survey, gupta2024systematic} \\
\midrule
\multirow{3}{*}{\parbox[t]{2.8cm}{\textbf{Finance \\ Technology}}} 
    & \textbf{Financial Question Answering} 
    & LLMs enhance financial QA in accurate, context-aware reasoning over complex, multi-source financial data.
    &  TAT-QA~\cite{zhu2021tat}: a Financial QA benchmark that contains over 16,000 questions built from real-world financial reports that combines tabular and textual data.  & \cite{al2025exploring,  al2025exploring, kahl2024llms, zacher2024can,  zhu2021tat, lai2024sec, tao2024finqa, shah2024multi} \\
\cmidrule(lr){2-5}
    & \textbf{Knowledge Graph Construction} 
    & LLMs enable automated KG construction from financial data, retrieval from KG and support multi-document financial QA. 
    & FinKG~\cite{kertkeidkachorn2023finkg}: A curated core financial knowledge graph built from authoritative sources like corporate reports and stock data, structured to enable systematic analysis and applications in financial forecasting, risk assessment, and decision-making through semantically rich relationships between entities. &  \cite{senechal2024llm, unknown2024Revolutionizing, kutumbe2024transforming, han2025rag, shah2024multi, kertkeidkachorn2023finkg} \\
\cmidrule(lr){2-5}
    & \textbf{Robo-advisory} 
    & LLMs enhance robo-advisors for novice investors, but still lag behind humans in performance and trust.
    & Jung et al~\cite{jung2018robo}: A earlier work that propose the concept "Robo-Advisory" that leverage AI to provide automatic financial advisory servies for a broader range of investors. & \cite{feng2024can, jung2018robo, Gill2024financial, tzanetos2024robo, singh2024empowering, zhao2024revolutionizing, tzanetos2024robo} \\
\bottomrule
\end{tabular}
}
\end{table}

\noindent\textbf{Taxonomy.}
To understand the potential application of LLMs in finance research, we propose a taxonomy that reflects the diversity of tasks across the field:

\begin{itemize}[leftmargin=10pt]
    \item \textbf{Trading and Investments.} Trading pursues short-term gains while investment emphasizes long-term value through diversification and analysis. Traditional methods struggle with large-scale, complex data, whereas LLMs offer new capabilities for processing unstructured information, enhancing forecasting, and supporting strategies in quantitative trading and portfolio management.
    
    \item \textbf{Corporate Finance.} Corporate finance manages funding, capital structure, and investment to drive growth. Conventional approaches like financial modeling and discounted cash flow analysis are labor-intensive and limited under fast-changing conditions. LLMs streamline tasks such as financial report analysis, improving efficiency and accuracy in strategic decision-making.
    
    \item \textbf{Financial Markets.} Financial markets allocate resources and manage risk through the trading of assets. While econometric models and machine learning aid analysis, they face challenges with today’s data scale and complexity. LLMs advance this field by processing unstructured information and enabling applications such as stock movement prediction.
    
    \item \textbf{Financial Intermediation and Risk Management.} Banks and insurers channel capital while managing risks, but traditional statistical models and manual processes lag in dynamic environments. LLMs improve performance by analyzing diverse datasets, with emerging applications in fraud detection and credit scoring.
    
    \item \textbf{Sustainable Finance.} Sustainable finance incorporates ESG factors into investment decisions. Standard scoring systems often overlook rich unstructured data from reports and media. LLMs can extract and synthesize such information, offering more context-aware and adaptive ESG insights.
    
    \item \textbf{Financial Technology.} FinTech reshapes financial services through innovations like digital banking, blockchain, and robo-advisory. Traditional solutions emphasize automation but lack flexibility. LLMs expand FinTech by powering financial question answering, knowledge graph construction, and conversational advisory, enhancing personalization and accessibility.
\end{itemize}

Each of these areas presents distinct opportunities for LLMs to enhance or extend traditional approaches. For instance, In Financial Report Analysis, LLMs can interpret complex narratives in earnings reports, extract key metrics or risk factors, and even flag inconsistencies or anomalies that may be missed by rule-based systems~\cite{han2024xbrl}. In ESG Scoring, LLMs can analyze qualitative disclosures across environmental, social, and governance dimensions, enabling more comprehensive and up-to-date assessments that incorporate nuanced language cues and public sentiment~\cite{wu2024susgen}.

Across these use cases, LLMs offer new forms of insight, not by replacing models that price assets or manage risk, but by bridging the gap between language and data, enabling better contextual awareness, more transparent decision-making, and broader access to financial knowledge.

\subsubsection{Trading and Investment}

Trading and investment constitute the bedrock of capital markets, encompassing the strategic allocation of financial resources to generate returns and manage risk. From a financial perspective, trading often focuses on short-term opportunities, driven by price volatility, liquidity dynamics, and market microstructure~\cite{harris2002trading}, while investment emphasizes long-term value creation through fundamental analysis, asset diversification, and portfolio optimization \cite{bodie2014investments}. These activities, though distinct in horizon and methodology, share a common goal: the maximization of expected utility under uncertainty.

Historically, trading and investment decisions have relied heavily on traditional methodologies, such as fundamental analysis—assessing the intrinsic value of assets based on economic indicators, company financials, and industry conditions—and technical analysis, which examines historical price patterns and market trends~\cite{williams1938theory, graham1951security}. However, these conventional methods face notable limitations, including the challenge of systematically processing vast amounts of information, susceptibility to subjective bias, and difficulty in accurately interpreting complex, unstructured data.

Recent advancements in LLMs have provided promising tools to overcome these limitations. By capturing and analyzing nuanced sentiments, attitudes, traits, and patterns hidden within structured and unstructured massive textual data, these advanced models can offer improved insights, enhanced forecasting capabilities, and more informed decision-making processes for trading and investment professionals. Two notable applications in this evolving landscape include Quantitative Trading and Portfolio Management, which are further discussed below:

\textbf{Quantitative Trading. }
Quantitative trading involves the use of mathematical models and algorithms to identify and exploit trading opportunities~\cite{chan2021quantitative}. LLMs can be used to incorporate textual data (e.g., news sentiment, social media, analyst reports) into predictive models, enhancing signal generation for algorithmic trading strategies.

Recent research in LLM-driven financial trading agents has advanced across several interconnected fronts, notably in memory architecture, agent individuality, multimodal intelligence, and simulation environments. A foundational trend is the emergence of LLM agents equipped with layered memory and character design, as seen in TradingGPT~\cite{Li2023TradingGPTMS}, FinMem~\cite{Yu2023FinMemAP}, and FinCon~\cite{yu2024fincon}. These systems introduce cognitive structures mimicking human memory stratification (short, medium, and long-term) while embedding distinctive agent personalities. This not only improves interpretability and decision diversity but also enhances adaptability in volatile markets.

Building on the concept of self-improvement, QuantAgent~\cite{wang2024quantagent} proposes a dual-loop learning system that evolves its domain-specific financial knowledge through iterative simulation and real-world feedback. Similarly, FS-ReasoningAgent~\cite{wang2024ExploringLC} innovatively segments reasoning into factual and subjective channels to optimize cryptocurrency trades, revealing nuanced insights into LLM reasoning preferences under different market conditions.

A parallel direction emphasizes realistic multi-agent financial ecosystems. TradingAgents~\cite{xiao2024tradingagents} introduces a hierarchical LLM-agent framework that mirrors the structure of real-world trading firms, organizing agents into specialized roles (e.g., fundamental analysts, sentiment analysts, traders, risk managers). Through structured inter-agent communication and decision protocols, it fosters collaborative, debate-driven decision-making, demonstrating superior cumulative returns and Sharpe ratios in simulation. Complementarily, FinCon~\cite{yu2024fincon} proposes a synthesized manager-analyst hierarchy with dual-level risk control—one handling within-episode volatility via CVaR alerts, and another guiding strategic belief updates across episodes via conceptual verbal reinforcement—achieving strong performance in both single-stock and portfolio trading tasks.

Meanwhile, simulation-based evaluation of trader behavior remains an active area. ASFM~\cite{Gao2024SimulatingFM} and StockAgent~\cite{zhang2024ai} construct rich multi-agent environments with LLMs as trader proxies, allowing researchers to analyze policy impacts and behavioral biases in realistic, yet controlled, financial ecosystems. These studies highlight the use of LLMs not only as traders but as tools to deepen our understanding of market dynamics and agent interactions.

In parallel, LLMs’ capacity for sentiment-driven trading has also been substantiated~\cite{kirtac2024sentiment}, where models like GPT-3 significantly outperform traditional sentiment dictionaries in predicting returns, achieving strong Sharpe ratios and cumulative profits. MarketSenseAI~\cite{fatouros2024can} and FinAgent~\cite{zhang2024multimodal} further extend this line with multimodal inputs and tool-augmented architectures, synthesizing text, charts, and fundamentals to simulate generalist financial agents with impressive profitability and reasoning transparency.

Collectively, these works mark a shift toward intelligent, explainable, and increasingly autonomous AI trading systems grounded in human-aligned cognition, hierarchical coordination, and complex environmental simulations.

\textbf{Portfolio Management. }
Portfolio management involves strategically selecting and overseeing a set of financial assets, such as stocks, bonds, commodities, currencies, and cryptocurrencies, to meet specified investment objectives~\cite{reilly2002investment}. LLMs can analyze large volumes of qualitative and quantitative data, interpret market news and sentiments, forecast trends, and assist in strategic decision-making. 

The application of LLMs in portfolio and trading strategy research has evolved rapidly, branching into several thematic directions. A foundational area centers on strategy generation and alpha mining. Alpha-GPT~\cite{Wang2023AlphaGPTHI} introduces a human-AI interactive paradigm for alpha discovery, empowering quants to translate intuitive trading ideas into formulaic signals via natural language interfaces. Similarly, Kou et al.~\cite{kou2024automate} extends this concept through a multi-agent and multimodal framework that dynamically evaluates market conditions and adapts trading strategies accordingly.

A second important line of work investigates portfolio construction and management via LLMs. Studies like Ko and Lee~\cite{ko2024can} and Abe et al.~\cite{abe2024leveraging} explore how ChatGPT and persona-based ensembles enhance asset selection and diversification. These systems often outperform random or traditional strategies, particularly during specific market conditions such as rising inflation or high volatility. Gu et al.~\cite{gu2024adaptive} takes this further by introducing a margin trading model that adaptively reallocates between long and short positions using an LLM-RL hybrid system, combining real-time reasoning and transparency for better risk management.

Parallel to these efforts, researchers have focused on news-driven sentiment analysis and reinforcement learning (RL). Wu~\cite{Wu2024PortfolioPB} and Unnikrishnan~\cite{unnikrishnan2024financial} demonstrate how LLMs can extract actionable sentiment from financial news, significantly improving the performance of RL agents when managing portfolios or single-stock trading. These works confirm that LLM-enhanced strategies outperform baseline RL models and even historical benchmarks under various market scenarios.

Lastly, in the domain of crypto portfolio management, Luo et al.~\cite{luo2025llm} proposes an LLM-powered multi-agent system that integrates multimodal data and collaborative reasoning across expert agents. This approach captures the complex nature of digital assets, showing strong performance in both asset selection and portfolio return across leading cryptocurrencies.

Together, these studies illustrate a shift from traditional, rule-based approaches to more adaptive, interpretable, and human-aligned systems powered by LLMs. They mark a significant leap in integrating advanced language technologies into financial decision-making and portfolio optimization.

\subsubsection{Corporate Finance}

Corporate finance is formally defined as the area of finance that focuses on how corporations handle their sources of funding, capital structuring, and investment decisions~\cite{brealey2014principles, tirole2010theory}. In simpler terms, corporate finance is about how companies manage their money, deciding where to get funds, how to invest them effectively, and how to ensure the company grows profitably and sustainably over time.

Traditionally, corporate finance relies heavily on methods such as financial modeling, discounted cash flow analysis~\cite{modigliani1958cost}, capital budgeting techniques~\cite{dean1951capital}, and sensitivity analysis. These conventional approaches, though proven effective, often face critical limitations—they tend to be labor-intensive, heavily reliant on manual computations, susceptible to human error~\cite{steiger2010validity}, and sometimes inflexible when dealing with rapidly changing market conditions or complex financial scenarios.

In recent years, LLMs have emerged as powerful tools capable of addressing these limitations. LLMs have the potential to automate and enhance many corporate finance tasks, such as investment analysis, merge and acquisition (M\&A) forecasting and insolvency forecasting, enabling more accurate, insightful, and efficient financial management~\cite{kropmans2024application}. Especially when complementing traditional financial models \cite{yang2023getting,xing2025ai}. Among these applications, one particularly promising area is financial report analysis, which plays a critical role in informing corporate financial strategies. Specifically, we discuss the recent progress in financial report analysis in the following section.

\textbf{Financial Report Analysis. } Financial report analysis is crucial for understanding corporate performance and guiding investment decisions. Traditionally involving labor-intensive methods like ratio and trend analysis, it has increasingly benefited from automation via LLMs \cite{mousavi2025lexicons}, which excel at extracting, summarizing, and interpreting complex financial information.

Recent studies highlight various applications of LLMs in financial analysis. Han et al.~\cite{han2024xbrl} demonstrated improvements in eXtensible Business Reporting Language (XBRL) analysis by integrating retrieval-augmented generation and specialized calculation tools, while Le~\cite{le2024auto} illustrated the effectiveness of financially fine-tuned LLMs for earnings report generation. Ziegler~\cite{gomes2024automating} underscored the superior accuracy of multimodal approaches for ESG data extraction. Moving forward, key research opportunities include advancing multimodal integration, refining domain-specific fine-tuning strategies, enhancing interpretability for robust decision-making, and developing comprehensive platforms that seamlessly integrate LLM-based analyses into financial workflows.

\subsubsection{Financial Market Analysis}

A financial market can be formally defined as a structured marketplace where individuals, institutions, and governments engage in the buying and selling of financial instruments such as stocks, bonds, commodities, derivatives, and currencies~\cite{pilbeam2018finance}. It serves as a pivotal platform that facilitates efficient allocation of resources, liquidity management, and risk distribution, underpinning economic growth and financial stability on a global scale~\cite{greenwood1997financial, bond2012real}. In simpler terms, financial markets function similarly to any marketplace, like a grocery store or an auction, but instead of groceries or antiques, they deal with assets like shares of companies, government bonds, or commodities such as gold and oil. Buyers and sellers come together to trade these financial assets, each seeking profit, stability, or a way to mitigate their risks.

Historically, traditional methods employed within financial markets have revolved primarily around statistical and econometric models~\cite{lai2008statistical}, fundamental analysis~\cite{wafi2015fundamental}, and machine learning algorithms~\cite{chen2003application}. While these approaches have served as the foundation for investment decisions for decades, they possess inherent limitations, particularly concerning their predictive accuracy, adaptability, and capacity to handle vast, complex, and unstructured datasets prevalent in today's fast-moving digital financial landscape~\cite{hasan2020current, shen2018big}.

In recent years, advances in LLMs present promising solutions to address many limitations encountered by traditional methodologies. By leveraging their remarkable abilities to process extensive amounts of textual and numerical data, understand contextual nuances, and discern patterns, LLMs offer powerful tools to interpret financial data more accurately, effectively enhancing decision-making processes in the financial industry~\cite{deng2023llms}. One notable application of these models, Stock Movement Prediction, are attracting considerable interest, and will be discussed in subsequent section.

\textbf{Stock Movement Prediction. }
Stock movement prediction involves forecasting the direction or magnitude of stock price changes, which is a pivotal task for investors and financial analysts~\cite{bustos2020stock}. With recent advancements, LLMs have shown substantial promise in this domain by leveraging vast amounts of unstructured textual data, such as financial news, earnings reports, and investor sentiment, which traditional numerical models might overlook or fail to adequately analyze.

Recent research has seen a surge of interest in leveraging LLMs for stock movement prediction, evolving from simple sentiment analysis to sophisticated, explainable, and data-integrated forecasting frameworks. A foundational line of work demonstrates that LLMs can extract economically meaningful sentiment signals from financial headlines. For example, Lopez-Lira and Tang~\cite{lopez2023can} show that ChatGPT can predict short-term stock returns from news headlines without explicit financial fine-tuning, outperforming traditional methods and revealing that model size correlates with economic efficacy. Extending this, Zhang et al.~\cite{Zhang2023UnveilingTP} explore LLMs in the context of Chinese financial texts, comparing different LLM architectures for sentiment factor extraction and establishing a standardized pipeline for backtesting trading strategies derived from Chinese news sentiment. Similarly, Bhat and Jain~\cite{Bhat2024StockPT} focus on distilled LLMs to perform emotion analysis on headlines, showing that emotion-labeled news can predict stock trends as reliably as traditional financial indicators.

Moving beyond raw sentiment, several works emphasize explainability and structured prediction. Koa et al.~\cite{koa2024learning} introduce the Summarize-Explain-Predict (SEP) framework, where a self-reflective LLM iteratively trains itself to generate human-readable justifications for stock predictions, eliminating the need for expert-annotated data and offering robust performance in both classification and portfolio tasks. Wang et al.~\cite{wang2024llmfactor} propose LLMFactor, which extracts interpretable economic “factors” from news via prompt engineering (Sequential Knowledge-Guided Prompting), significantly improving explainability and aligning predictions with market-relevant insights. Meanwhile, Ni et al.~\cite{ni2024harnessing} tailor LLMs for earnings report analysis using a QLoRA-enhanced model that integrates both firm-specific financials and external macro factors, outperforming GPT-4 in predictive accuracy.

Together, these studies reflect a broader transition in the field: from sentiment extraction as a proxy for return signals to multimodal, interpretable, and fine-tuned approaches that push the boundaries of AI-enabled financial forecasting. They underscore the growing ability of LLMs to not only predict stock movement but also to articulate why and how such predictions are made, thereby enhancing transparency and trust in AI-driven investment decisions.

\subsubsection{Financial Intermediation and Risk Management}

Financial Intermediation and Risk Management refers to the systematic process through which financial institutions, such as banks, insurance companies, and investment firms, facilitate the efficient allocation of capital by channeling funds from savers to borrowers, while simultaneously managing the inherent financial risks associated with such transactions~\cite{allen1997theory}. In simpler terms, financial intermediation involves institutions acting as middlemen who pool money from individuals and businesses with surplus funds, and then lend or invest those funds in activities that need financing. Risk management involves strategies that these institutions use to identify, assess, and control risks, ensuring the stability and health of the financial system.

Traditionally, financial intermediation and risk management have relied heavily on statistical methods, manual reviews, and rule-based systems~\cite{mcneil2015quantitative, pritchard2014risk}. Institutions typically used historical data analysis, regulatory frameworks, and expert judgment to make lending and investment decisions, as well as to evaluate and mitigate risks~\cite{muhlbauer2004pipeline}. However, these traditional methods have notable limitations, including their reliance on historical data, which may not effectively capture rapidly changing market dynamics or unexpected events. Additionally, manual processes can be time-consuming, error-prone, and inefficient when handling large volumes of data, thus limiting the ability of financial institutions to respond quickly and effectively to new financial risks.

Advances in LLMs have the potential to significantly enhance financial intermediation and risk management practices. LLMs can process vast amounts of structured and unstructured data efficiently, while offer novel capabilities in analyzing textual data, interpreting complex patterns, and providing actionable insights, thus enabling institutions to manage risk more proactively and dynamically. Two important applications where LLMs show promise are Fraud Detection and Credit Scoring, which will be introduced in the following sections.

\textbf{Fraud Detection. }
Fraud detection is essential for safeguarding financial institutions from illicit activities that result in significant financial losses and damage stakeholder trust~\cite{west2016intelligent}. LLMs offer powerful capabilities for enhancing fraud detection, as they can efficiently process vast volumes of textual and numerical data, detect subtle anomalies, and recognize complex patterns indicative of fraudulent behavior. Leveraging LLMs enables financial institutions to improve accuracy, reduce detection time, and swiftly adapt to emerging threats.

Recent advancements illustrate the potential of LLMs across diverse financial fraud scenarios. Notably, Bakumenko et al.~\cite{bakumenko2024advancing} present a compelling methodology that leverages sentence-transformer embeddings to encode non-semantic categorical features in financial journal entries. This approach effectively mitigates feature sparsity and heterogeneity, yielding substantial performance gains across multiple classification models. Their results suggest that LLM-based embeddings capture latent structures traditional encodings overlook, marking a pivotal step in audit-grade anomaly detection.

Complementing this, Korkanti~\cite{korkanti2024enhancing} integrates customized LLMs with state-of-the-art predictive analytics and anomaly detection techniques, producing a robust framework that significantly boosts both precision and recall. The model demonstrates heightened sensitivity to subtle, high-risk indicators within real-time transactional and communication datasets, addressing key gaps in adaptability and responsiveness prevalent in existing systems.

Meanwhile, studies like Boskou et al.~\cite{boskou4897041exploring} and Cao et al.~\cite{cao2024risklabs} affirm the broader versatility of LLMs. The former employs prompt-engineered interactions with ChatGPT-4 to identify deception in corporate disclosures, yielding moderately successful classification scores without fine-tuning. The latter, through a multi-modal fusion of earnings calls, time-series data, and news content, shows promise for general financial risk prediction, though its direct fraud-detection utility remains underexplored.

On the benchmarking front, Yang et al.~\cite{yang2025fraud} introduce the Fraud-R1 dataset—an extensive, multilingual, multi-round benchmark designed to rigorously assess LLM robustness against realistic phishing and fraud scenarios. Their results expose persistent weaknesses in role-play and cross-lingual settings, emphasizing the need for multilingual, adaptive, and context-aware models.

In sum, LLMs are reshaping the fraud detection landscape by enabling deeper contextual comprehension and cross-modal reasoning. Future efforts should focus on real-time multilingual capabilities, explainability, and adversarial robustness to ensure deployment-ready solutions in dynamic financial environments.

\textbf{Credit Scoring. }
Credit scoring is a critical tool in financial risk management used by financial institutions to assess the likelihood that individuals or entities will fulfill their credit obligations. Traditionally, it relies on statistical methods such as logistic regression and decision trees~\cite{dastile2020statistical}. However, recent advancements of LLMs offer the potential to enhance predictive accuracy and provide deeper contextual understanding through textual data analysis. LLMs can complement traditional numeric risk indicators, improving both predictive performance and interpretability.

The integration of LLMs into credit risk assessment has opened new avenues for enhancing both predictive performance and interpretability across diverse financial settings. One line of research explores LLMs as generalist credit scoring tools, exemplified by Feng et al.'s CALM model~\cite{feng2023empowering}, which leverages instruction tuning across nine datasets to create a versatile and benchmarked LLM for credit and risk assessment. This approach reveals LLMs' promise in democratizing access to sophisticated scoring tools while also highlighting concerns around fairness and potential bias.

Another core stream investigates LLMs for hybrid modeling and interpretability enhancement. Teixeira et al.~\cite{Teixeira2023EnhancingCR} introduce Labeled Guide Prompting (LGP), which augments GPT-4 outputs with Bayesian network reasoning and labeled examples to generate credit reports preferred by human analysts, blending human-like judgment with machine consistency.

Several studies address LLMs in specialized or constrained financial environments. Sanz-Guerrero and Arroyo~\cite{sanz2024credit} apply a fine-tuned BERT model to peer-to-peer (P2P) lending platforms, where traditional financial variables are sparse. By extracting risk signals from loan narratives, they show that LLM-derived scores can improve prediction and reshape how credit models weigh traditional features, especially across different loan purposes. Similarly, Drinkall et al.~\cite{drinkall2025forecasting} conduct a critical evaluation of generative LLMs in corporate credit rating prediction, finding that despite strong textual encoding, models like GPT underperform compared to multimodal baselines like XGBoost when numerical reasoning is essential.

Collectively, these works illustrate a maturing field where LLMs are not merely augmenting legacy models, but increasingly redefining how creditworthiness is evaluated—particularly in terms of generalization, explainability, and ethical oversight. They point to a future in which credit assessment becomes more adaptive, inclusive, and transparent through the responsible deployment of AI.

\subsubsection{Sustainable Finance}
The integration of Environmental, Social, and Governance (ESG) factors into investment decisions has gained significant traction in recent years. 
Traditional ESG scoring methodologies often rely on structured data and predefined metrics. 
However, the increasing availability of unstructured textual data related to ESG, such as news articles and company reports, offers a rich source of information that LLMs are well-suited to process. 
Recent research~\cite{zhao2024revolutionizing} indicates that LLMs can potentially contribute to ESG scoring by learning implicit evaluation criteria directly from text. 
Moreover, several studies~\cite{calamai2025corporate, shimamura2025evaluating, lin2024gpt4esg} have explored the use of LLMs to extract structured information from sustainability reports, which can then be employed for ESG-relevant classification tasks~\cite{birti2025optimizing, tian2024esg}.
Beyond information extraction and classification, the concept of using LLMs as evaluators in ESG scoring is also gaining attention~\cite{zou2025esgreveal}.
A recent survey~\cite{gu2024survey} mentions the promising applications of the "LLM-as-a-judge" paradigm in various financial domains, including ESG scoring. 
This suggests a future where LLMs could synthesize information from diverse textual sources and make an overall judgment on a company's ESG performance, potentially mimicking the role of human ESG analysts. 
This could lead to more dynamic and context-aware scoring mechanisms that go beyond traditional data aggregation methods. 
However, it is important to note that research has also highlighted the potential brittleness of LLMs when processing long contexts, such as large sustainability reports, which could impact their reliability in ESG scoring tasks~\cite{gupta2024systematic}.

\subsubsection{Financial Technology}
Financial Technology (Fintech) refers to the integration of technology into the offerings of financial services companies to improve their use and delivery to consumers~\cite{leong2018fintech}. 
FinTech has implications not merely for the financial industry, and includes virtually all forms of e-commerce and spreads to other industry fields, especially in blockchain technology, such as insurance, banking services, trading on capital markets, and risk management~\cite{stojakovic2020fintech}.

\textbf{Financial QA.}
The capability to quickly and accurately access financial information is crucial in the fast-paced world of finance. LLMs are emerging as powerful tools to build sophisticated financial question answering systems. 
Studies in this topic aim to explore the capabilities of LLMs to understand and respond to a wide range of financial queries. 
For instance, FinCausal 2025 shared task~\cite{al2025exploring} proposes a task specifically focused on the ability of LLMs to detect causal relationships within financial texts through question answering. 
The study evaluates both LLMs and discriminative approaches, with the findings indicating the strong potential of generative LLMs, particularly in scenarios where only a few examples are available for learning. 
This suggests that LLMs can effectively reason about financial causality even with limited task-specific training data.

Recognizing the need for high-quality data to train and evaluate these systems, researchers have curated specialized datasets. 
SocialFinanceQA~\cite{kahl2024llms} introduces a benchmark dataset, comprising a vast collection of financial questions and answers extracted from Reddit's finance-focused communities. 
The rationale behind this is that these real-world discussions reflect the actual language and types of questions that individuals have about finance, providing a valuable resource for fine-tuning and aligning LLMs to better serve user needs in this domain. 
A recent survey FinLLMs~\cite{lee2025large} also highlights financial QA as a primary benchmark task, and lists several benchmark datasets for financial QA task evaluation.

The practical application of LLMs in financial question answering is already evident in consumer-facing platforms. 
The Amplework blog~\cite{singh2024empowering} highlights NerdWallet as an example of a platform that utilizes AI to provide users with personalized answers to their financial questions related to investing, personal finance, and debt management in real-time. 
This demonstrates the ability of LLMs to make complex financial information more accessible and understandable to a wider audience. 
Furthermore, another article~\cite{zacher2024can} mentions the testing of LLMs on CPA exam preparation materials.
LLMs, such as ChatGPT-4, demonstrate strong capabilities in business analysis and reporting automation, which inherently involves the ability for complex financial QA.

\textbf{Financial Knowledge Graph.}
Financial data is characterized by intricate relationships between various entities, including companies, financial instruments, market indicators, and economic events. 
Knowledge graphs offer a powerful way to represent these connections in a structured format, enabling more sophisticated analysis and information retrieval. 
LLMs are proving instrumental in automating the construction of these financial knowledge graphs from the vast amounts of unstructured textual data available~\cite{shah2024multi}. 
For example, LLM Knowledge Graph Builder~\cite{senechal2024llm}  allows users to transform unstructured data, including financial documents, into dynamic knowledge graphs without requiring specialized coding skills. 
It leverages LLMs to automatically identify and extract key entities and the relationships between them, subsequently converting this information into a graph structure that can be stored and queried efficiently.

This blog ~\cite{unknown2024Revolutionizing} further elaborates on the synergy between LLMs and knowledge graphs. 
LLMs excel at tasks like Named Entity Recognition (NER) and relationship extraction from unstructured text, which are fundamental steps in building knowledge graphs. 
By understanding the nuances of language and the context in which financial entities are mentioned, LLMs can accurately identify these entities and the connections between them, automatically populating and enriching financial knowledge graphs.

A practical application of this technology is detailed in the article~\cite{kutumbe2024transforming}, which provides a step-by-step guide on using the Neo4j LLM Knowledge Graph Builder to transform financial statements into knowledge graphs. 
This process involves uploading financial data, extracting key financial entities and their relationships (such as the relationship between revenue, expenses, and net income), and then querying the resulting knowledge graph using natural language via GraphRAG~\cite{han2025rag}. 
This allows financial analysts to gain deeper insights from financial statements more intuitively and efficiently.

While the examples above focus on specific tools and platforms, the underlying principles are broadly applicable. 
Furthermore, research has shown that knowledge graphs constructed by LLMs can enhance the performance of financial question answering systems. 
KG-RAG~\cite{shah2024multi} uses knowledge graph triples (constructed using a fine-tuned small language model) as context for multi-document financial question answering, achieving better results than traditional RAG methods.

\textbf{Robo-advisory.}
Robo-advisors have become a popular way for individuals to manage their investments through automated platforms that offer financial planning and investment management services~\cite{feng2024can}. 
Integrating LLMs into these platforms has the potential to significantly enhance their capabilities, making them more personalized, interactive, and effective. 

Akira.ai~\cite{Gill2024financial} discusses in detail how AI agents powered by LLMs are transforming financial robo-advisory. 
These agents can engage in more natural and human-like conversations with clients, understand their financial goals and risk tolerance expressed in natural language, and provide highly personalized investment recommendations. 
Moreover, LLMs can analyze vast amounts of real-time market data to provide timely insights and support portfolio adjustments, and they can even assist with tasks like tax optimization by identifying tax-efficient investment strategies.

Several leading robo-advisor platforms are already leveraging AI models, including LLMs, to enhance their services. 
This blog~\cite{singh2024empowering} highlights platforms like Betterment and Wealthfront, which use AI-driven algorithms to offer personalized investment portfolios tailored to users' financial situations and goals.
These systems continuously monitor market conditions and rebalance portfolios as needed. LLMs also contribute to real-time market analysis, sentiment detection, and the automation of savings and investment strategies on these platforms. 
Additionally, robo-advisors like Ellevest use AI-powered tools to offer personalized financial education alongside investment management, catering to users' specific financial literacy levels.

A fwe studies also indicate the broader potential of LLMs in various tasks relevant to robo-advisory. 
A recent survey~\cite{zhao2024revolutionizing} notes the increasing use of LLMs in finance for automating financial report generation, forecasting market trends, analyzing investor sentiment, and offering personalized financial advice. These capabilities can be directly integrated into robo-advisory platforms to provide more comprehensive and sophisticated services. 
While this article~\cite{tzanetos2024robo} emphasizes that human advisors still hold an edge in terms of personalization, it acknowledges the increasing sophistication of robo-advisors powered by AI and the use of techniques like retrieval-augmented generation (RAG) to improve their ability to provide relevant and accurate advice.

\subsubsection{Benchmarks}

\input{tables/finance_prellm_vs_llm}

Recent work, such as the FinLLMs survey~\cite{lee2025large}, has thoroughly summarized the existing financial benchmarks and datasets for specific tasks. To provide a more intuitive understanding of the performance improvements brought by LLM-based methods over traditional approaches across various financial tasks, we further compile a comprehensive comparison, as shown in Table~\ref{tab:prellm_vs_llm}. However, with the rapid emergence of LLMs and autonomous agents in financial applications, existing benchmarks are increasingly inadequate for evaluating model performance. On one hand, the unified and powerful language capabilities of LLMs, combined with their extensive knowledge base, enable them to tackle a wide range of financial tasks, necessitating a more comprehensive assessment of their overall competencies in finance. On the other hand, as LLMs continue to advance, there is growing interest in deploying them in practical financial scenarios rather than isolated, idealized experimental settings. Therefore, this section addresses these two critical aspects: first, we introduce recent multi-perspective, multi-task benchmarks designed to evaluate LLMs’ comprehensive abilities in finance; second, we discuss how to construct datasets that better reflect real-world conditions using publicly available data sources.

\textbf{Multi-Task Benchmarks}

\input{tables/benchmark_finance}

\textbf{FinBen.} FinBen~\cite{xie2024finben} comprises 36 datasets covering 24 tasks across seven critical financial dimensions: information extraction, textual analysis, question answering, text generation, risk management, forecasting, and decision-making. Its key innovations include the introduction of stock trading evaluation, agent-based and retrieval-augmented generation (RAG) assessments, and the development of novel datasets specifically designed for summarization, regulatory QA, and trading. Evaluations of prominent LLMs reveal that while current models excel at tasks such as information extraction and sentiment analysis, they still struggle with complex reasoning and domain-specific challenges, particularly in forecasting and decision-making. These findings underscore the potential and current limitations of LLMs in financial applications.

\textbf{R-Judge.} R-Judge~\cite{yuan2024r} is a comprehensive benchmark designed to assess the risk awareness of LLMs in agent-based environments, particularly focusing on their ability to detect and judge safety risks arising from multi-turn interactions. The dataset consists of 569 carefully curated agent interaction records, spanning 27 real-world scenarios across five application categories—programming, IoT, software, web, and finance—and covering ten distinct risk types, including financial loss, privacy leakage, and property damage. Within the financial domain, R-Judge introduces scenarios where LLM agents must make decisions that could lead to monetary losses or regulatory violations, thereby providing a practical testbed for evaluating LLMs' judgment under real-world financial constraints. Experimental results reveal substantial performance gaps: most models perform at or below random baselines, struggling especially with the financial and high-risk categories. These findings highlight the significant challenges in equipping LLMs with robust safety reasoning and underscore the importance of domain-specific fine-tuning and richer contextual understanding for deploying LLMs in sensitive financial environments.

\textbf{FinEval.} FinEval~\cite{zhang2023fineval} is a comprehensive benchmark designed to evaluate LLMs in the Chinese financial domain across both foundational knowledge and real-world application scenarios. The benchmark consists of 8,351 questions spanning four core areas: Financial Academic Knowledge, Financial Industry Knowledge, Financial Security Knowledge, and Financial Agent. It supports multiple evaluation settings—zero-shot, few-shot, and chain-of-thought prompting—and includes objective, subjective, and open-ended formats. Experimental results show that while models perform competitively in financial security and basic knowledge tasks, their performance drops significantly on agent tasks requiring dynamic planning and reasoning. These results underscore the current limitations of LLMs in simulating expert-level financial behavior and highlight the need for further advancement in domain adaptation and task complexity handling.

\input{tables/benchmark_finance_results}

\textbf{CFinBench.} CFinBench~\cite{nie2024cfinbench} is the most comprehensive Chinese financial benchmark to date. It consists of 99,100 questions spanning 43 second-level categories across four core dimensions aligned with real-world financial career progression: Financial Subject, Financial Qualification, Financial Practice, and Financial Law. These categories test LLMs' mastery of theoretical foundations (e.g., economics, auditing), professional certification knowledge (e.g., CPA, securities), applied job skills (e.g., tax consulting, asset appraisal), and legal compliance (e.g., banking and commercial law). The benchmark includes three question types (single-choice, multiple-choice, and judgment), enabling diverse and realistic assessment formats. Evaluations show that the best accuracy remains just over 60\%, indicating substantial challenges and room for improvement in financial domain adaptation. 

\textbf{UCFE.} UCFE~\cite{yang2024ucfe} presents a user-centric financial expertise benchmark designed to evaluate LLMs in real-world financial scenarios through dynamic, multi-turn user interactions. It introduces 17 task types, encompassing both zero-shot and few-shot formats across domains such as stock prediction, risk assessment, financial consulting, and regulatory compliance. Grounded in a large-scale user study with 804 participants, including analysts, financial professionals, regulators, and the general public. The benchmark tests LLMs not only for factual accuracy but also for their adaptability, response depth, and alignment with user satisfaction. Results across 11 models demonstrate that domain-specific LLMs like Tongyi-Finance-14B and CFGPT2-7B outperform general-purpose counterparts in delivering context-aware, actionable financial insights. These findings highlight the critical importance of user alignment and task interactivity for advancing LLMs in finance.

\textbf{Hirano.} Hirano~\cite{hirano2024construction} constructs the first large-scale benchmark specifically designed to evaluate LLMs in the Japanese financial domain. It comprises five core tasks reflecting both professional knowledge and real-world financial practices in Japan: sentiment analysis of securities reports (chabsa), fundamental knowledge in securities analysis (cma\_basics), auditing questions from the Certified Public Accountant exam (cpa\_audit), multiple-choice questions from the national financial planner certification (fp2), and practice exams for securities broker representatives (security\_sales\_1). These tasks cover a range of difficulties and formats—binary, multiple-choice, and judgment-based—enabling a nuanced assessment of LLM capabilities. Benchmarking results across a wide array of models show that even top-performing models face challenges in complex, domain-specific tasks. The analysis further reveals that training data quality and domain relevance substantially impact performance, emphasizing the need for tailored model development.

\textbf{Discussion.} Based on the conclusions drawn from the aforementioned benchmarks, we summarize the following insights regarding the current capabilities and limitations of LLMs in the financial domain:

\begin{itemize}[leftmargin=10pt]
    \item \textbf{Challenges in complex financial tasks:} Current LLMs still struggle with tasks that require deep domain knowledge, logical reasoning, and multi-step decision-making.
    
    \item \textbf{Effectiveness of domain-specific fine-tuning:} Fine-tuning LLMs on domain-specific corpora continues to yield notable performance gains, demonstrating its importance in enhancing model specialization.
    
    \item \textbf{Benchmark coverage vs. real-world applicability:} While these benchmarks effectively assess LLMs’ comprehensive capabilities in finance, they are primarily diagnostic and not tailored to specific application scenarios. Practical use cases often require the design of dedicated, task-specific benchmarks.
    
    \item \textbf{Need for broader evaluation dimensions:} Additional attention should be given to other meaningful evaluation perspectives, such as user alignment (e.g., UCFE) and risk awareness (e.g., R-Judge), which are crucial for safe and effective real-world deployment.
\end{itemize}

To provide an intuitive overview of the performance of mainstream LLMs on these benchmarks, we also present a selection of evaluation results from the aforementioned benchmarks, as shown in Table~\ref{tab:benchmark_results}. Given the substantial data gaps and the continuous updates across different LLM providers, we recommend that interested readers conduct these benchmark evaluations independently to support their own assessments.

\textbf{Financial Dataset Construction}

Although LLM-based financial agents~\cite{xiao2024tradingagents, Yu2023FinMemAP, yu2024fincon} are attracting increasing interest, standardized benchmarks~\cite{li2024investorbench} and datasets remain scarce. To support researchers in developing their own evaluation environments and datasets for trading and investment tasks, we present a curated list of raw data sources that can serve as the foundation for constructing custom benchmarks (Table~\ref{tab:financial_data_sources}). These include the following six categories of data:

\input{tables/benchmark_finance_raw_data}

\begin{itemize}[leftmargin=10pt]
    \item \textbf{General Financial Data:} Provides access to real-time and historical stock prices, fundamental financial indicators, and corporate financial statements. Such data are critical for simulating trading environments, developing investment strategies, conducting market forecasts, and evaluating algorithmic trading agents.

    \item \textbf{Cryptocurrency Data:} Offers market prices, trading volumes, and metadata for cryptocurrencies. These datasets are particularly useful for research on crypto trading strategies, market microstructure analysis, and portfolio optimization involving digital assets.

    \item \textbf{Regulatory Filings:} Includes official company disclosures, such as quarterly and annual reports (10-Q, 10-K), and other significant events (8-K filings). Regulatory filings are essential for fundamental analysis, event-driven trading, and financial sentiment extraction.

    \item \textbf{Analyst Reports:} Consists of investment opinions, earnings forecasts, and qualitative assessments from financial analysts. These resources are valuable for sentiment analysis, opinion aggregation, and modeling the impact of market expectations on asset prices.

    \item \textbf{News Data:} Covers financial news, press releases, and market commentary from a variety of media outlets. News data is critical for developing event-driven trading strategies, market volatility prediction, and detecting sentiment shifts in real-time.

    \item \textbf{Social Media Data:} Comprises user-generated content from platforms such as Twitter (X) and Reddit. Social media data enables the study of retail investor sentiment, information diffusion, and the dynamics of attention-driven market movements.
\end{itemize}

These raw data sources provide flexibility for researchers to create customized datasets, simulate trading scenarios, and design evaluation benchmarks tailored to specific financial tasks and market conditions. For instance, INVESTORBENCH~\cite{li2024investorbench} constructs a comprehensive benchmark environment for stock, cryptocurrency, and ETF trading tasks by aggregating multi-source data such as OHLCV market data from Yahoo Finance and CoinMarketCap, regulatory filings from the SEC EDGAR database, and news data from public datasets and Refinitiv. These data are further enriched with sentiment annotations generated by GPT-3.5 and structured into time-series segments for warm-up and testing phases, enabling rigorous simulation of real-world trading scenarios. TradingAgents~\cite{xiao2024tradingagents} formulates its evaluation environment by collecting diverse financial signals—including historical stock prices, corporate fundamentals, and real-time news—from APIs and public financial repositories. FinMem~\cite{Yu2023FinMemAP} integrates high-frequency signals like stock price series, social media sentiment, and financial news into short-term memory, while storing earnings reports and macroeconomic indicators in long-term layers with decay-based retention. FinCon~\cite{yu2024fincon} constructs a multi-modal financial environment by combining text-based corporate disclosures, tabular time series, and audio transcripts of earnings calls. 

\subsubsection{Discussion}

\noindent\textbf{Opportunities and Impact.}
LLMs are ushering in a transformative era in financial research and applications, offering a unique combination of scalability, adaptability, and interpretability across diverse financial domains. As demonstrated in trading and investment~\cite{Li2023TradingGPTMS, wang2024quantagent, fatouros2024can}, corporate finance~\cite{han2024xbrl, gomes2024automating}, financial markets~\cite{lopez2023can, wang2024llmfactor}, financial intermediation and risk management~\cite{bakumenko2024advancing, feng2023empowering}, sustainable finance~\cite{zhao2024revolutionizing, zou2025esgreveal}, and fintech innovation~\cite{shah2024multi, Gill2024financial}, LLMs provide new tools to process vast unstructured datasets, generate strategic insights, and support decision-making processes with unprecedented efficiency. Additionally, their ability to integrate multimodal inputs, perform natural language reasoning, and adapt across financial subdomains positions LLMs not just as supplementary tools, but as pivotal contributors to the future of finance. 

\noindent\textbf{Challenges and Limitations.}
Despite their promise, significant challenges remain. First, numerical reasoning remains a major limitation for LLMs~\cite{drinkall2025forecasting}, where tasks involving high-frequency trading, real-time credit scoring, and precision-focused financial modeling still favor traditional econometric or hybrid approaches. Second, the interpretability and explainability of LLM decisions, especially in black-box architectures, must be improved to meet the stringent compliance standards in regulated industries such as banking and insurance~\cite{yang2025fraud, feng2023empowering}.

Another critical concern is robustness to distribution shifts. Financial markets are notoriously volatile and non-stationary; LLMs fine-tuned on historical data often struggle to generalize to unseen macroeconomic conditions or systemic shocks~\cite{ni2024harnessing, yang2025fraud}. Similarly, context length limitations pose challenges for applications like ESG scoring, where processing long, detailed sustainability reports remains difficult~\cite{gupta2024systematic}.

Moreover, the fairness and bias inherent in LLM training~\cite{feng2023empowering} raises ethical concerns, particularly when models are deployed in high-stakes areas like credit scoring or fraud detection. Addressing these issues requires interdisciplinary collaboration between AI developers, financial domain experts, and regulators.

\noindent\textbf{Research Directions.}
Several promising research directions emerge from current trends:

\begin{itemize}[leftmargin=10pt]
    \item \textbf{Domain-Specific Fine-Tuning and Instruction Tuning.} Tailoring LLMs to financial subdomains through specialized datasets, instruction tuning, or hybrid neuro-symbolic architectures can significantly improve domain alignment and trustworthiness~\cite{feng2023empowering, han2024xbrl}.
    
    \item \textbf{Multimodal and Multi-Agent Systems.} Integration of textual, numerical, and visual financial data streams in multi-agent architectures, as seen in TradingAgents~\cite{xiao2024tradingagents} and FinCon~\cite{yu2024fincon}, offers pathways toward more human-aligned, context-aware financial agents.
    
    \item \textbf{Explainable AI and Trustworthy Reasoning.} Developing frameworks like SEP~\cite{koa2024learning} and LLMFactor~\cite{wang2024llmfactor} that combine predictive performance with human-readable explanations will be critical for adoption in regulated industries.
    
    \item \textbf{Real-Time and Adaptive Learning.} Future models must be capable of online learning and rapid adaptation to changing market conditions, with architectures that support dynamic knowledge updates and feedback-driven improvement~\cite{wang2024quantagent}.
    
    \item \textbf{Ethics, Fairness, and Regulatory Compliance.} Research must prioritize fairness audits, bias mitigation strategies, and interpretability mechanisms to ensure that LLM-based systems meet ethical and legal standards in finance~\cite{feng2023empowering}.
\end{itemize}

\noindent\textbf{Conclusion.}
LLMs offer a compelling expansion of the financial analytics toolkit, promising new capabilities in understanding, forecasting, and decision-making under uncertainty. However, realizing their full potential requires careful attention to their limitations, rigorous fine-tuning for financial contexts, and an ongoing commitment to ethical, robust, and interpretable AI development. As finance continues to evolve, the synergistic integration of LLMs with traditional quantitative methods, domain knowledge, and human oversight will likely define the next frontier in financial innovation.

\subsection{Economics}

\subsubsection{Overview}

\noindent\textbf{Introduction.}
Economics, in its formal rigor, is the study of how agents allocate scarce resources to maximize utility or profits under constraints, guided by models rooted in mathematical logic and empirical testing~\cite{Solow2008TheEO}. More intuitively, it is the science of everyday choices—why people buy certain things, how firms decide what to produce, and how governments respond to inflation. Whether analyzing a shopper choosing between apples and oranges, or central banks managing interest rates, economics provides frameworks to understand behavior and institutional interactions.

Economics spans a wide range of subfields, each with its own research focus and methodological approach. Microeconomics studies individual decision-making and the functioning of markets~\cite{pindyck2018microeconomics}, while macroeconomics looks at the economy as a whole, analyzing phenomena such as economic growth, inflation, and unemployment~\cite{barro1997macroeconomics}. Specialized areas like labor economics~\cite{borjas2010labor}, industrial organization~\cite{tirole1988theory}, public finance~\cite{rosen1992public}, and behavioral economics~\cite{barberis2003survey} delve into specific aspects of economic activity—from how institutions shape labor markets to how cognitive biases influence individual choices. Across these domains, economists employ tools such as optimization theory, econometrics, and general equilibrium modeling~\cite{judd1998numerical, hayashi2011econometrics}. Foundational frameworks like the IS-LM model~\cite{hicks1937mr}, the Solow growth model~\cite{solow1956contribution}, and Nash equilibrium in game theory~\cite{nash1950equilibrium} have not only advanced theoretical understanding but also played a pivotal role in shaping public policy and economic thought more broadly.

Traditional economic methods have made profound contributions to both scientific understanding and real-world decision-making. Empirical techniques like randomized controlled trials have redefined development economics—evident in Banerjee, Duflo, and Kremer’s field experiments~\cite{banerjee2011poor} that earned the 2019 Nobel Prize and influenced global aid policy. Structural models, such as DSGE frameworks~\cite{smets2007shocks}, are now central to monetary policy in institutions like the European Central Bank and the U.S. Federal Reserve. Game-theoretic tools underpin modern auction design~\cite{milgrom2004putting}, informing spectrum auctions worldwide and contributing to the 2020 Nobel Prize in Economics. Together, these traditional methods exemplify how rigorous economic modeling and empirical analysis have not only advanced scientific theory but also shaped impactful policies across global institutions.

\begin{figure}[!t]
    \centering
    \includegraphics[width=0.99\linewidth]{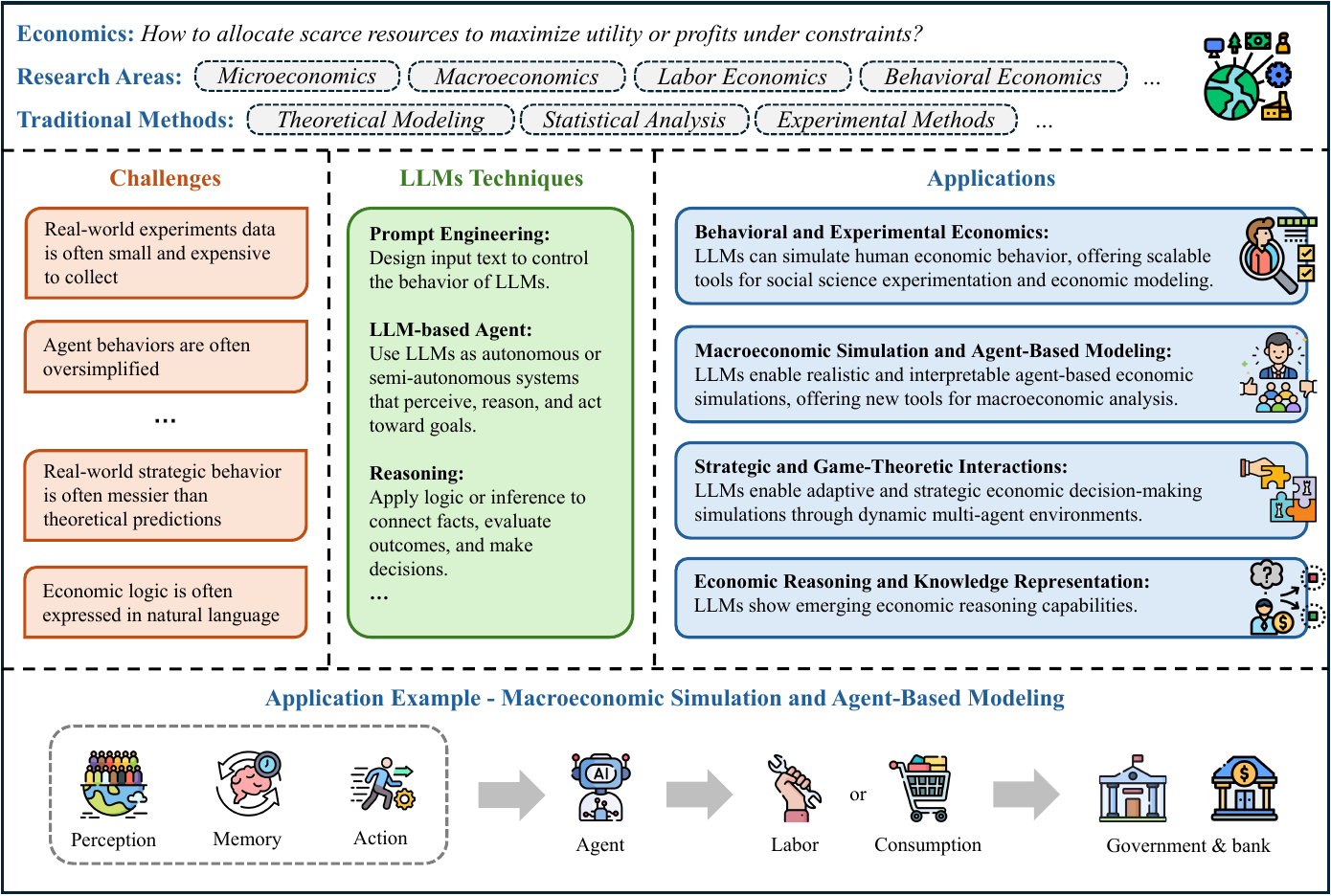}
    \caption{Overview of LLMs' Applications in Economics Research.}
    \label{fig:economics}
    \vspace{-10pt}
\end{figure}

Despite the progress in economic research, significant challenges persist that limit the realism and applicability of traditional models. Real-world economic environments are inherently dynamic, characterized by constantly evolving institutions, shocks, and information flows. Economic agents are diverse in their preferences, constraints, information, and decision-making processes—yet conventional models often rely on simplifying assumptions such as representative agents, rational expectations, or static equilibria. These abstractions can obscure important heterogeneities and interactions that drive actual outcomes.

Furthermore, collecting high-quality experimental or observational data is frequently expensive, time-consuming, and limited in scale, which constrains empirical validation. Agent behaviors are often oversimplified, failing to capture bounded rationality, adaptive learning, or context-dependent strategies. Real-world strategic interactions—such as those seen in financial markets, policy negotiations, or consumer choices—tend to be far more nuanced and unpredictable than what classical theory anticipates.

In addition, economic reasoning is frequently articulated in natural language, making it challenging to formally encode into models or simulations. Problems like modeling evolving preferences, simulating large-scale economies with heterogeneous agents, or generating credible counterfactual scenarios remain analytically intractable or computationally intensive. These limitations highlight the need for more flexible, data-rich, and behaviorally informed approaches to economic analysis.

\noindent\textbf{The Role of LLMs.}
LLMs, as text-based agents trained on vast human knowledge and capable of simulating reasoning and dialogue, offer a new toolset. Yet their capacity is domain-dependent: they may struggle with tasks demanding precise quantitative forecasting, consistent economic logic across contexts, or nuanced understanding of causality. Nonetheless, in specific domains, LLMs demonstrate potential. They can model human-like decisions in behavioral experiments~\cite{Filippas2023LargeLM}, simulate agents in macroeconomic environments~\cite{Taghikhah2023ACF}, and engage in strategic reasoning akin to game-theoretic thinking~\cite{Shapira2024GLEEAU, Guo2024EconNLIEL}. Their fluency in natural language allows them to interpret and generate policy scenarios, simulate negotiations, or even mimic human fairness preferences~\cite{Mei2024ATT}. These affordances make them well-suited to complement existing tools in behavioral and experimental economics, agent-based modeling, and knowledge representation.

\input{tables/economics}

\noindent\textbf{Taxonomy.}
To systematically explore these intersections, the following taxonomy organizes research at the nexus of economics and LLMs into four task domains:

\begin{itemize}[leftmargin=10pt]
    \item \textbf{Behavioral and Experimental Economics.} This field studies how real people make decisions, often deviating from the rational “homo economicus” model. Experiments with games like the dictator, ultimatum, and trust games reveal biases such as fairness concerns and the endowment effect. LLMs complement these methods by simulating diverse decision behaviors and allowing rapid pre-testing of economic experiments.
    
    \item \textbf{Macroeconomic Simulation and Agent-Based Modeling.} ABMs simulate how individual agents interact to shape aggregate outcomes like inflation or unemployment. Unlike equilibrium-based models, they capture dynamic, bottom-up processes but often lack realistic human behavior. LLMs enrich ABMs by powering adaptive, communicative agents, bringing greater realism and flexibility to macroeconomic simulations.
    
    \item \textbf{Strategic and Game-Theoretic Interactions.} Game theory examines how outcomes depend on the choices of multiple agents, requiring competition, cooperation, and anticipation. Traditional approaches rely on simplified assumptions, limiting realism. LLMs enable agents with recursive reasoning and natural language interaction, offering richer simulations of strategic scenarios.
    
    \item \textbf{Economic Reasoning and Knowledge Representation.} Economic reasoning analyzes trade-offs under scarcity, while knowledge representation encodes concepts for computational use. Rule-based methods struggle with complexity and scalability. LLMs simulate reasoning in natural language and generalize across contexts, though they remain sensitive to prompt design and prone to oversimplification.
\end{itemize}

Across economic research, LLMs do not replace traditional models or empirical methods, but extend the field’s analytical reach—by linking language with behavior, enabling simulation of human-like agents, and enhancing strategic reasoning. They bridge the gap between narrative and formal analysis, offering new tools for interpreting, modeling, and experimenting with economic decision-making.

\subsubsection{Behavioral and Experimental Economics}

Behavioral and Experimental Economics is a subfield of economics that studies how people actually make decisions, often deviating from the idealized “rational agent” model~\cite{Kahneman2002FoundationsOB}. Unlike classical economics, which assumes individuals are perfectly rational, consistent, and self-interested (homo economicus), this field acknowledges that humans are prone to biases, emotions, and heuristics. Experimental economists design controlled lab or field experiments to observe behaviors such as fairness, cooperation, time inconsistency, or risk aversion. The vividness of this discipline comes from its attention to how real people respond to incentives and information—like how someone might overvalue an item they own (endowment effect) or hesitate to switch default options even if better alternatives exist (status quo bias)~\cite{Kahneman1991AnomaliesTE}. These insights are foundational to areas like policy design, marketing, and behavioral finance.

Traditionally, behavioral and experimental economics relies on human subjects in lab settings, often using games like the dictator game~\cite{Engel2010DictatorGA}, ultimatum game~\cite{thaler1988anomalies}, or trust game~\cite{Johnson2011TrustGA} to infer preferences and behaviors. But these methods are time-consuming, costly, and constrained by participant availability and ethical considerations. Moreover, running variations of an experiment—for example, tweaking the framing of a question or demographic profile of the subject—requires significant effort. Another challenge is that findings often suffer from limited external validity due to small sample sizes or artificial settings. Here is where LLMs open up a new frontier. Because LLMs can be prompted to act as decision-makers in structured economic tasks, researchers can simulate thousands of hypothetical agents with varying preferences or constraints, rapidly exploring behavioral responses across scenarios~\cite{horton2023large, Ross2024LLMEM}. This opens the door to low-cost, scalable pre-testing of economic experiments, probing theoretical boundaries, or even generating novel hypotheses for subsequent testing with human subjects.

Recent research illustrates both the promise and nuance of using LLMs in behavioral and experimental economics. Horton positions LLMs as synthetic stand-ins for experimental participants~\cite{horton2023large}. He shows that GPT-3 can replicate results from classic behavioral experiments, like status quo bias in budget allocation or reactions to fairness in pricing scenarios, and that by simply altering the agent's "social preferences" via prompts, one can elicit predictable shifts in decision-making. This mirrors how economists model preference heterogeneity, suggesting that LLMs might serve not just as simulation tools but as exploratory partners in theory development.

Another group of studies tests whether LLMs exhibit economic rationality in a formal sense. Chen et al.~\cite{Chen2023TheEO} apply revealed preference theory to GPT by placing it in structured budget allocation tasks across domains like risk, time, and social preferences. They find GPT's choices to be remarkably consistent with utility maximization, often more so than actual human subjects. Yet the models show sensitivity to framing, indicating that context still matters—a pattern familiar from human experiments.

Complementing these findings, Ross et al.~\cite{Ross2024LLMEM} take a utility-theoretic approach to map LLM biases. They assess the degree to which LLMs manifest behavioral regularities like loss aversion, risk aversion, and time discounting. While LLMs like GPT-4 show patterns that align with both rational agent models and human-like heuristics, the consistency and magnitude of these biases vary across models and prompting strategies \cite{gao2025take,liu2025evaluating}. The implication is that LLMs, while powerful, are not monolithic agents. They behave in context-sensitive ways and can be shaped or nudged via design choices.

In sum, LLMs are emerging not just as tools for automating economic reasoning but as experimental agents in their own right. Their utility lies not in replacing human subjects but in extending the experimental toolkit of economists, providing new ways to generate hypotheses, test theories, and explore behavioral nuance at scale.

\subsubsection{Macroeconomic Simulation and Agent-Based Modeling}

Macroeconomic simulation and agent-based modeling (ABM) are computational tools used to study how individual economic decisions aggregate into macro-level phenomena~\cite{Axtell2025AgentBasedMI}. Unlike traditional models that assume representative agents and equilibrium, ABM simulates economies from the bottom up, with heterogeneous agents, like households and firms, interacting in dynamic environments. Picture a digital society where agents choose whether to work, consume, or save based on personal traits and environmental cues. As these decisions ripple through the system, they give rise to emergent outcomes like inflation, unemployment, or economic cycles.

Traditional ABMs often rely on rule-based or statistical models that are rigid, require expert calibration, and struggle to represent real human behavior~\cite{Macal2005TutorialOA}. Behavioral and experimental economics aim to capture more realistic decision-making, but integrating these insights into scalable simulations is difficult. Agent heterogeneity and adaptive behavior remain major gaps. LLMs offer a powerful solution. With their capacity for reasoning, memory, and language understanding, LLMs can simulate more human-like agents that respond to complex environments, communicate, and evolve over time, reducing the need for hand-coded rules and improving realism.

Recent work illustrates this shift. EconAgent~\cite{Li2023EconAgentLL} uses LLM-powered agents to simulate macroeconomic activities like labor supply and consumption. These agents perceive their environment and reflect on past experiences, producing macro patterns that align with real-world phenomena such as the Phillips Curve. Woo et al.~\cite{woo2024llm} develop a reinforcement learning-enhanced ABM where LLMs generate indices like perceived value and information spread, enabling agents to make context-sensitive purchasing decisions and model social influence. Hao and Xie~\cite{Hao2025AMF} push heterogeneity further by using different LLMs to represent socioeconomically distinct agents. Their Multi-LLM-Agent-Based (MLAB) framework captures both cognitive and contextual diversity, allowing nuanced policy simulations, such as responses to taxation across income groups.

Together, these studies signal a new generation of economic modeling. By embedding LLMs into ABMs, researchers can build richer, more adaptive simulations that bring us closer to modeling the true complexity of economic behavior.

\subsubsection{Strategic and Game-Theoretic Interactions}

Strategic and game-theoretic interactions in economics involve decision-making where each agent’s payoff depends on the choices of others. Game theory provides formal tools to model such settings, capturing the logic of anticipation, competition, and cooperation~\cite{Samuelson2016GameTI}. Classic examples include auctions, bargaining, and coordination games. These environments often require recursive reasoning—thinking about what others think you will do—and can be challenging to model or simulate with traditional economic tools.

Standard approaches use mathematical models with strict assumptions about rationality or controlled human experiments~\cite{ichiishi2014game}, both of which struggle with complexity, communication, and scalability. LLMs, by contrast, offer a novel way to simulate agents in strategic settings. Their ability to process natural language and adapt strategies on the fly makes them uniquely suited for interactive, multi-agent environments.

Recent work shows both promise and limitations. Guo et al.\cite{Guo2024EconomicsAF} introduced EconArena, where LLMs compete in games like beauty contests and auctions. Results show that models like GPT-4 and Claude2 display bounded rationality and strategic adaptation, especially when given game history. However, none consistently reach Nash equilibrium, and rule-following varies by model. This suggests LLMs can reason about others’ strategies, but not perfectly. Similarly, Mei et al.~\cite{Mei2024ATT} find through a behavioral Turing test that LLMs like ChatGPT-4 can exhibit behaviors remarkably close to human distributions in classic economic games, although their responses tend toward greater altruism and cooperation compared to typical human actions. Importantly, these models adapt their strategies based on context and past interactions, mirroring human learning and responsiveness to framing. Shapira et al.~\cite{Shapira2024GLEEAU}, using their GLEE framework, further emphasize the role of communication style in sequential strategic interactions such as bargaining and negotiation, highlighting that LLMs can realistically mimic human behavior, particularly where language nuances matter.

Together, these works highlight the potential of LLMs as tools for modeling strategic behavior, especially in dynamic or linguistically complex settings. Despite limitations regarding perfect rationality, their adaptability, expressiveness, and scale make them valuable complements to traditional economic methods.

\subsubsection{Economic Reasoning and Knowledge Representation}

Economic reasoning involves the structured analysis of how individuals, firms, and institutions make decisions under conditions of scarcity~\cite{Parkes2015EconomicRA}. Knowledge representation in economics, on the other hand, refers to how economic facts, theories, and relationships are formally encoded so that they can be understood and utilized by computational systems. These two areas are interlinked: reasoning cannot happen without structured knowledge, and knowledge must be reasoned with to be useful.

Traditionally, economic reasoning and knowledge representation have relied on rule-based systems or statistical simulations. While these methods are methodologically rigorous, they often falter when faced with the ambiguity of natural language or the complexity of real-world causal chains. Encoding economic knowledge has typically involved labor-intensive manual curation, resulting in systems that are brittle and difficult to scale. LLMs, by contrast, offer a fundamentally different approach. They are capable of detecting patterns in economic text, simulating chains of reasoning, and generalizing across diverse contexts through zero-shot~\cite{Kojima2022LargeLM} and few-shot (in-context learning)~\cite{Dong2022ASO} paradigms, as well as with Chain-of-Thought prompting~\cite{Wei2022ChainOT}. Nonetheless, their effectiveness is highly sensitive to prompt design, and they remain prone to hallucinations or conceptual misapplications, particularly in more nuanced or domain-specific scenarios.

Recent work has begun to systematically explore these capabilities. For instance, the EconQA dataset~\cite{Patten2023EvaluatingDS} evaluates LLMs on multiple-choice questions sourced from economics textbooks, probing their ability to answer definitional, factual, and conceptual questions. Chain-of-thought (CoT) prompting—encouraging the model to reason step by step—improved performance modestly, especially in tasks requiring multi-step logic. However, results suggest that not all prompt formats yield consistent gains and that some models can reason effectively without explicit CoT cues.

Complementing this, EconNLI~\cite{Guo2024EconNLIEL} evaluates models on their ability to infer causal relationships between economic events. This dataset requires not just linguistic understanding but familiarity with economic theories like the quantity theory of money. Models like GPT-4 performed better than random but still made frequent reasoning errors, particularly when the causal link wasn't surface-level or intuitive. This highlights the gap between surface-level fluency and deep economic understanding. EconLogicQA~\cite{Quan2024EconLogicQAAQ} pushes LLMs further by asking them to sequence multiple interrelated economic events logically. This task mirrors real-world economic planning or policy analysis, where understanding the progression from cause to effect is crucial. Here, even advanced models like GPT-4 achieve only moderate accuracy, underscoring the complexity of sequential reasoning in economics. Unlike static inference, this task tests whether a model can integrate multiple facts into a coherent narrative, a skill essential for economic decision-making.

Together, these efforts illustrate both the promise and limitations of LLMs in economics. They can parse and process economic knowledge at scale and can support reasoning when guided by well-designed prompts. However, challenges remain in grounding their reasoning in robust theory and avoiding confident but incorrect inferences. As research progresses, combining LLMs with formal economic models or hybrid systems that inject domain knowledge may offer a fruitful path forward.

\subsubsection{Benchmarks}

\textbf{GLEE}~\cite{Shapira2024GLEEAU} provides a unified framework and benchmark for studying language-based interactions in economic environments. It focuses on two-player sequential games, such as bargaining, negotiation, and persuasion, where communication occurs through natural language. GLEE standardizes the evaluation of LLM-based agents across multiple economic contexts by defining consistent parameterizations, degrees of freedom, and economic metrics such as efficiency and fairness. The benchmark includes extensive datasets from LLM vs. LLM and human vs. LLM interactions and enables controlled experimentation to study how language affects strategic behavior and outcomes in economic settings.

\textbf{EconLogicQA}~\cite{Quan2024EconLogicQAAQ} is a benchmark designed to assess the sequential reasoning abilities of LLMs within the domains of economics, business, and supply chain management. Unlike traditional benchmarks that evaluate models on isolated events, EconLogicQA requires models to logically sequence multiple interconnected events extracted from real-world business news articles. The benchmark presents multi-event multiple-choice questions, challenging LLMs to understand temporal and causal relationships in economic scenarios. A rigorous human review process ensures the quality and difficulty of the dataset, which serves as a tool for probing models' reasoning depth in complex economic contexts.

\textbf{EconNLI}~\cite{Guo2024EconNLIEL} introduces a natural language inference task specifically targeting economic event reasoning. EconNLI evaluates whether an LLM can correctly determine causal relationships between pairs of economic events or generate plausible consequent events based on a given premise. Unlike traditional NLI tasks based on semantic entailment, EconNLI requires understanding of economic theories and principles to infer causal links. Extensive experiments reveal that current LLMs struggle significantly with economic reasoning tasks, highlighting substantial gaps between surface-level language proficiency and domain-specific reasoning capabilities.

\subsubsection{Discussion}

\noindent\textbf{Opportunities and Impact.}
The integration of LLMs into economics research offers a profound expansion of the traditional methodological toolkit. Across behavioral and experimental economics~\cite{horton2023large, Ross2024LLMEM}, macroeconomic simulation~\cite{Li2023EconAgentLL, woo2024llm}, strategic interaction modeling~\cite{Guo2024EconomicsAF, Mei2024ATT}, and economic reasoning~\cite{Patten2023EvaluatingDS, Guo2024EconNLIEL}, LLMs provide new capacities: simulating human-like decision-making, enabling large-scale agent-based models with greater realism, supporting dynamic and adaptive strategic interactions, and offering scalable, language-driven representations of economic knowledge.

These tools open new frontiers for both theoretical exploration and empirical experimentation. Researchers can rapidly prototype behavioral experiments, simulate macroeconomic scenarios with heterogeneous agents, model strategic negotiations dynamically, and test causal reasoning across economic contexts at a scale and speed previously unattainable. For example, on the EconNLI dataset, even the best-performing traditional encoder-only model, fine-tuned BERT, achieved an F1 score of below \textbf{0.8}. In contrast, relatively small-scale LLMs like LLAMA2-7B, after supervised fine-tuning, easily surpassed this benchmark with F1 score around \textbf{0.87}, illustrating the superior reasoning capabilities of LLMs~\cite{Guo2024EconNLIEL}. Additionally, LLMs allow economics to bridge formal models and narrative analysis, capturing nuances that traditional mathematical abstractions may miss. This flexibility is critical as economic phenomena increasingly span complex, dynamic, and information-rich environments.

\noindent\textbf{Challenges and Limitations.}
Despite these promising advances, LLM-driven economics research faces notable limitations. First, economic consistency and coherence remain major concerns. While LLMs can mimic economic reasoning patterns, they are prone to hallucinations, oversimplifications, or logical inconsistencies, particularly in tasks requiring multi-step causal reasoning or deep theoretical grounding~\cite{Guo2024EconNLIEL, Quan2024EconLogicQAAQ}.

Second, prompt sensitivity and model variability introduce fragility into experimental results. Different prompt designs or minor variations in phrasing can lead to divergent outputs, complicating reproducibility and interpretation~\cite{Ross2024LLMEM, Patten2023EvaluatingDS}.

Third, rationality and strategic depth are bounded. While LLMs can engage in first-order strategic reasoning (anticipating others' actions), they often struggle with deeper levels of recursive reasoning necessary for achieving equilibrium behaviors in complex games~\cite{Guo2024EconomicsAF}.

Additionally, the context alignment problem persists. LLMs are typically trained on broad internet data and may import biases or unrealistic assumptions when simulating economic agents, particularly when real-world institutional, cultural, or contextual factors are crucial.

Finally, ethical considerations emerge. Using LLMs as synthetic economic agents raises questions about bias propagation, external validity of simulated results, and the responsible interpretation of LLM-based experimental findings.

\noindent\textbf{Research Directions.}
Several key research directions emerge to address these challenges:

\begin{itemize}[leftmargin=10pt]
    \item \textbf{Domain Adaptation and Fine-Tuning.} Fine-tuning LLMs on economics-specific corpora or augmenting them with structured economic knowledge bases could improve coherence, realism, and domain fidelity~\cite{Guo2024EconNLIEL, Patten2023EvaluatingDS}.
    
    \item \textbf{Structured Prompting and Experimental Design.} Developing standardized, transparent prompting methodologies—analogous to experimental protocols—will be essential for ensuring replicability and robustness in LLM-driven economic experiments~\cite{Ross2024LLMEM}.
    
    \item \textbf{Hybrid Modeling Approaches.} Combining LLMs with traditional economic models (e.g., embedding LLMs within agent-based frameworks governed by formal economic constraints) can leverage the strengths of both systems while mitigating weaknesses~\cite{Li2023EconAgentLL, woo2024llm}.
    
    \item \textbf{Advanced Reasoning and Chain-of-Thought Methods.} Enhancing multi-step, causal, and counterfactual reasoning through methods like Chain-of-Thought prompting or tool-augmented reasoning frameworks offers pathways to deeper, theory-consistent outputs~\cite{Wei2022ChainOT, Quan2024EconLogicQAAQ}.
    
    \item \textbf{Ethical Evaluation and External Validation.} Establishing benchmarks, guidelines, and ethical frameworks for using LLMs as economic agents—alongside systematic validation against real human data—will be crucial for credible scientific practice. Another important area relates to the economics of open versus closed LLMs, financial and geographical accessiblity barriers, and new forms of digital divides \cite{oketch-etal-2025-bridging}. 
\end{itemize}

\noindent\textbf{Conclusion.}
LLMs represent a transformative tool for economics research, enabling novel forms of behavioral simulation, strategic modeling, macroeconomic exploration, and reasoning automation. Yet they are not substitutes for traditional theory or empirical grounding. Instead, when carefully validated and thoughtfully deployed, they can serve as powerful complements—extending economists’ ability to model, simulate, and interpret complex economic behavior in ways that were previously infeasible. Future progress will depend not only on technical advancements but also on establishing robust methodological foundations that blend linguistic fluency with economic rigor.

\subsection{Accounting}

\subsubsection{Overview}

\noindent\textbf{Introduction.}
Accounting is formally defined as the systematic process of recording, classifying, summarizing, and interpreting financial information to support decision-making and ensure transparency and accountability~\cite{american1966statement, bushman2003transparency}. It plays a central role in shaping how economic activity is communicated and regulated~\cite{hopwood2013trying, bushman2001financial}. At its essence, accounting serves as the language of business: it tells the story of where money comes from, where it goes, and what it means. Similar to a medical professional analyzing vital signs, decision-makers rely on accounting to assess an organization’s financial health. It elucidates whether a company is profitable, solvent, or experiencing financial distress, thereby guiding strategic choices and fostering trust among stakeholders.

Accounting research spans several major areas that mirror professional practice. For example, financial reporting examines how firms communicate with investors and regulators, often evaluating the quality and usefulness of disclosures~\cite{weygandt2018financial}; Auditing focuses on the verification of financial information, exploring how auditors assess risk, detect fraud, or maintain independence~\cite{imhoff2003accounting}; Managerial accounting centers on internal decision-making, providing tools for budgeting, cost control, and performance evaluation~\cite{garrison2021managerial}; Taxation research investigates how firms respond to tax policies, design strategies, or manage compliance~\cite{hoogendoorn1996accounting, graham2012research}. Each of these directions reflects accounting’s broader role in supporting economic coordination, regulatory oversight, and informed decision-making in markets and organizations.

Traditional research methods, e.g., econometrics, archival analysis, case studies, and content analysis, have significantly advanced both academic understanding and practice. Regression models, for instance, have helped uncover patterns in earnings management, shaping reforms in reporting standards and corporate governance~\cite{coakley2000artificial, callen1996neural}. These methods have yielded tangible benefits. Studies linking financial reporting quality to lower cost of capital have informed investor behavior and policy~\cite{nwaobia2013financial}. Improvements in audit procedures have reduced fraud and increased public trust in financial markets~\cite{imhoff2003accounting, rezaee2004restoring}. One study~\cite{allen2023tax} estimates that the widespread preference among VC-backed startups for C-corporations over tax-advantaged LLCs has led to \$43.9 billion in foregone tax savings—equivalent to 4.9\% of total invested equity—highlighting how accounting-related decisions can carry substantial financial consequences. Though often resource-intensive and limited in scope, these methods form the empirical foundation of accounting knowledge. Their rigor has helped accounting evolve into a robust, evidence-based profession.

Modern accounting encounters a range of novel challenges, particularly in managing the vast and growing volumes of both structured and unstructured data~\cite{vasarhelyi2015big}. Financial reports, audit notes, and regulatory disclosures are increasingly complex and voluminous, making them difficult to analyze using traditional tools~\cite{richins2017big, cockcroft2018big, warren2015big}. Manual review and rule-based text analysis are not only labor-intensive but also brittle in the face of evolving language and formats. Furthermore, accounting professionals must navigate ever-changing legislation, which is difficult to track and interpret consistently across jurisdictions. Complex and evolving reporting standards like IFRS and GAAP introduce additional ambiguity, often leading to inconsistent application and interpretation. The pressure for faster, more accurate reporting adds urgency to these issues, especially as stakeholders demand real-time insights and greater transparency. Collectively, these challenges demand more adaptive, intelligent systems capable of understanding and processing the rich, nuanced content found in modern financial environments.

\begin{figure}[!t]
    \centering
    \includegraphics[width=0.99\linewidth]{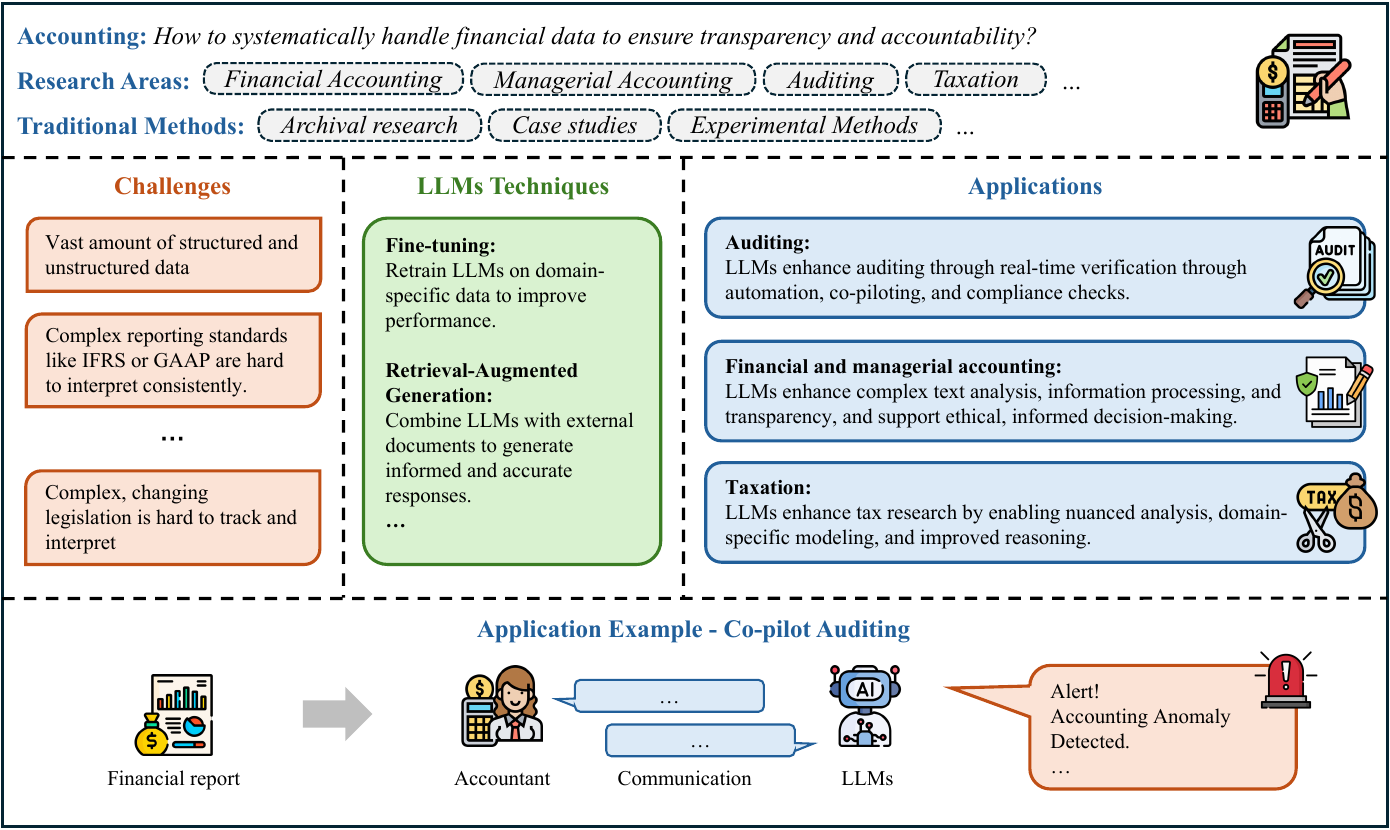}
    \caption{Overview of LLMs' Applications in Accounting Research.}
    \label{fig:accounting}
    \vspace{-10pt}
\end{figure}

\noindent\textbf{The Role of LLMs.}
LLMs provide an alternative. Their primary strength lies in interpreting and generating human-like text, rendering them effective for tasks such as summarizing disclosures, elucidating accounting standards, or extracting insights from audit reports. Their adaptability makes them valuable in both research and professional settings~\cite{alshurafat2023usefulness, zhao2024unleashing, dong2024scoping, street2023let}. Nevertheless, LLMs also exhibit certain limitations. They can generate convincing yet erroneous statements, lack domain-specific reasoning, and pose risks when applied to sensitive data~\cite{huang2025survey, wang2023survey, yao2024survey}. They cannot serve as replacements for expert judgment, particularly in tasks involving legal interpretation or ethical evaluation. Nevertheless, when employed judiciously, LLMs can augment human capabilities. They offer speed, scalability, and linguistic flexibility that can enhance productivity in tasks such as document review, preliminary analysis, or instructional support. Their value resides in complementing—rather than supplanting—traditional methods and expertise.

\noindent\textbf{Taxonomy.}
To understand the potential application of LLMs in accounting research and practice, we propose a taxonomy that reflects the diversity of tasks across the field:

\begin{itemize}[leftmargin=10pt]
    \item \textbf{Auditing.} Traditionally reliant on manual, sample-based checks, auditing struggles with rising data volumes and fraud complexity. LLMs can automate text analysis, flag anomalies, and expand audit coverage, enabling smarter AI-assisted audits while underscoring the need for transparency and safeguards.
    
    \item \textbf{Financial and Managerial Accounting.} Both functions are central to decision-making but increasingly burdened by complex disclosures and fragmented systems. LLMs help extract insights, streamline reporting, and convert unstructured data into actionable analysis, strengthening transparency, accuracy, and strategic value.
    
    \item \textbf{Taxation.} Taxation involves intricate laws and resource-constrained enforcement, with traditional systems often missing nuances in legal texts. LLMs can interpret tax codes, analyze unstructured filings, and support compliance and enforcement, offering new efficiency while raising questions of trust and adaptability.
\end{itemize}

Across accounting tasks, LLMs do not replace traditional methods of measurement or verification, but instead unlock new forms of understanding—by interpreting complex language, streamlining routine processes, and expanding access to financial reasoning. They bridge the gap between narrative and numbers, enhancing clarity, efficiency, and insight in the accounting profession.

\input{tables/accounting}

\subsubsection{Auditing}

Auditing is a cornerstone of financial accountability, designed to ensure that financial statements are accurate, complete, and compliant with regulatory standards~\cite{imhoff2003accounting}. At its core, auditing involves the systematic examination of a company’s financial records and transactions to assess their fairness and reliability~\cite{boynton2005modern, antipova2023auditing}. In more practical terms, an audit can be visualized as a multi-layered process where auditors sift through mountains of structured and unstructured financial data, like journal entries, invoices, contracts, and disclosures, looking for inconsistencies, errors, or signs of fraud. This work demands high attention to detail, extensive knowledge of accounting standards, and a careful balance of skepticism and judgment~\cite{kumar2015auditing}. Yet, the traditional tools at their disposal often rely on sampling and manual procedures, limiting coverage and efficiency~\cite{coderre2009internal, dittenhofer2001internal}.

Conventional audit methods have long faced limitations. Auditors typically work with samples rather than full datasets due to time and resource constraints, making audits prone to oversight~\cite{christensen2015behind, gepp2018big}. They also contend with increasingly complex data types, regulatory changes, and mounting pressure for faster, higher-quality audits. Manual processing of financial texts and repetitive audit tasks remain time-consuming, while fraud detection often hinges on patterns too subtle for traditional rules-based systems.

LLMs, like ChatGPT, offer transformative potential for addressing these gaps. By understanding and generating human-like text, LLMs can assist auditors in a wide range of tasks, including analyzing financial narratives, extracting key indicators from disclosures, automating documentation, and even performing real-time risk assessments. For instance, Gu et al.~\cite{gu2024artificial} introduced the concept of "co-piloted auditing," where auditors work collaboratively with LLMs to analyze journal entries, perform ratio analysis, and identify anomalies using natural language prompts. The result is a more dynamic and scalable auditing process that goes beyond simple automation to offer contextual insights and proactive risk detection. Moreover, in continuous auditing settings, LLMs enable full population testing and real-time data cross-verification. Li et al.~\cite{li2024enhancing} showcased a real-world implementation in a government payroll audit where ChatGPT parsed regulatory text and matched it against accounting records with 96\% accuracy, reducing verification time by 83\%.

Nevertheless, LLMs are not without drawbacks. Their use introduces concerns around model hallucinations, data privacy, audit independence, and ethical accountability~\cite{fotoh2023exploring}. Ensuring model outputs are transparent and grounded in verifiable sources is essential to maintaining audit integrity.

Recent research reflects a surge in interest in applying AI and LLMs to auditing. Fedyk et al.~\cite{fedyk2022artificial} provide empirical evidence that audit firms investing in AI experience improved audit quality, reduced audit fees, and a shift in labor dynamics—particularly the displacement of junior audit staff. Gu et al.~\cite{gu2024artificial} advance this by conceptualizing AI not merely as a tool, but as a collaborative partner in the audit process. Their "co-pilot" model positions ChatGPT as a flexible assistant capable of reasoning through complex audit tasks using chain-of-thought prompting, showing how fine-tuned LLMs can contribute meaningfully to judgment-intensive activities.

Meanwhile, Wang et al.~\cite{wangauditbench} propose AuditBench, a benchmark specifically designed to evaluate LLM performance in financial statement auditing, including tasks such as error identification, standards citation, and corrective action. The study highlights both the promise and current limitations of LLMs in reliably executing full audit cycles. Berger et al.~\cite{berger2023towards} explore regulatory compliance verification through LLMs, focusing on whether models can assess conformity between financial text and legal mandates. The study by Föhr et al.~\cite{fohr2023deep} introduces a comprehensive framework for integrating deep learning and LLMs into risk-based auditing, emphasizing the need for organizational readiness and technical infrastructure. Meanwhile, Fotoh and Mugwira~\cite{fotoh2023exploring} explore ethical implications, emphasizing the importance of safeguarding independence, privacy, and professional judgment in an AI-augmented audit environment.

Other contributions expand the practical landscape. Emett et al.~\cite{emett2024leveraging} document ChatGPT's integration into the internal audit workflows of a multinational firm, reporting efficiency gains of 50–80\% in audit preparation, fieldwork, and reporting. Similarly, Eulerich and Wood~\cite{eulerich2023demonstration} demonstrate a wide array of LLM applications across the entire audit lifecycle, from planning and risk assessment to documentation and follow-up, highlighting its value for both professional and academic audiences.

These studies collectively illustrate a shifting paradigm in auditing—from static, manual procedures to dynamic, AI-assisted processes. While promising, the literature also cautions against over-reliance, stressing the need for robust controls, explainability, and updated ethical standards.

\subsubsection{Financial and Managerial Accounting}

Financial and managerial accounting form the backbone of modern business decision-making, albeit with different audiences and goals~\cite{warren2020financial}. Financial accounting focuses on standardized reporting for external stakeholders—investors, regulators, and creditors—by presenting a historical snapshot of a firm’s financial health. In contrast, managerial accounting supports internal decision-making through more granular, often real-time data on costs, performance metrics, and forecasts~\cite{williams2018financial}. 

Despite their centrality, both fields have long faced challenges. Financial accounting has become increasingly burdened by voluminous and complex disclosures that hinder rather than help decision-making~\cite{krahel2015consequences, saha2019disclosure}. Managerial accounting struggles with fragmented systems and the difficulty of converting raw figures into actionable insights~\cite{heidmann2008exploring, granlund2002moderate}. These pain points are exacerbated by the rapid growth of unstructured data, requiring more than traditional tools to manage effectively~\cite{bhimani2014digitisation}.

LLMs, like ChatGPT, offer a new frontier for addressing these challenges. Their capacity to interpret, generate, and summarize complex text allows them to streamline both reporting and analysis. For financial accounting, LLMs can parse regulatory filings, reduce information overload, and improve transparency. A recent study by Kim et al.~\cite{kim2023bloated} demonstrates that LLM-generated summaries of financial disclosures not only condense content but often sharpen its informativeness, better aligning with market reactions. This ability to cut through “bloated” disclosures can enhance price efficiency and reduce information asymmetry.

In managerial contexts, LLMs bring value through integration with Robotic Process Automation (RPA), enabling them to not only interpret data but act on it. As shown in Li and Vasarhelyi’s~\cite{li2024applying} framework, such integrations automate tasks like transaction coding and report generation, overcoming limitations of standalone APIs or user interfaces. Beerbaum~\cite{beerbaum2023generative} further emphasizes the scalability of this approach, proposing RPA-LLM systems as ethically conscious and operationally robust tools for routine accounting tasks. The growing body of research also reflects this shift. One emerging theme centers on financial forecasting and performance estimation. For example, Comlekci et al.~\cite{comlekci2023can} used ChatGPT to project financial outcomes for publicly traded firms based on past data and sector developments. While accuracy varied by metric, the study highlighted LLMs’ potential in generating early-stage projections, especially when enhanced with external textual context.

Another important theme is the role of LLMs in auditing and assurance, especially in sustainability reporting. With the rise of ESG disclosures and regulatory initiatives like the EU’s CSRD, verifying qualitative statements has become more critical. Studies by Föhr et al.~\cite{fohr2023assuring} and Ni et al.~\cite{ni2023paradigm}introduce LLM-powered tools for benchmarking disclosures against frameworks like the TCFD, showing how domain-specific prompt engineering and retrieval mechanisms can produce traceable, high-quality evaluations.

At the same time, critical voices have emerged. De Villiers et al.~\cite{de2024will} caution that generative AI may inadvertently replicate biased or "greenwashed" content if left unchecked. Their analysis calls for greater transparency and accountability in how LLMs are trained and deployed in reporting contexts, particularly in areas like sustainability, where narrative control is often strategic.

Another important line of work explores behavioral patterns embedded in financial texts. Harris~\cite{harris2024managers} leverages LLMs to analyze how perceived competitive pressure influences managerial earnings manipulation, uncovering subtle textual signals in 10-K filings that correlate with discretionary accounting choices. This points to the growing role of LLMs in behavioral accounting research, where language not only describes but shapes economic decisions.

In sum, the integration of LLMs into financial and managerial accounting offers the potential to enhance efficiency, accuracy, and transparency, while also raising new questions about accountability and interpretation. As these models evolve, their role is not just as passive tools but as collaborators in reshaping how financial information is produced, analyzed, and understood.

\subsubsection{Taxation}

Taxation refers to the process through which governments collect financial contributions from individuals and organizations to fund public services, redistribute wealth, and support economic policy~\cite{besley2013taxation}. Taxes come in many forms, e.g., income tax, corporate tax, sales tax, and property tax, and their design involves balancing efficiency, equity, and administrative feasibility~\cite{salanie2011economics}. A fair and effective tax system is essential to a functioning society, yet its administration is often opaque and heavily reliant on expert interpretation.

Traditionally, the field of taxation has faced several persistent challenges. First, the tax code is extraordinarily complex, constantly evolving, and filled with exceptions, loopholes, and jurisdiction-specific nuances. For researchers, practitioners, and taxpayers alike, this complexity makes compliance difficult and error-prone~\cite{laffer2011economic, Brady2024}. Second, the enforcement of tax laws, such as through audits, is constrained by resource limitations and data opacity~\cite{slemrod2019tax}. Third, tax research and policy analysis have long depended on structured data, which fails to capture the richness and nuance embedded in textual disclosures, court decisions, and regulatory interpretations~\cite{hanlon2010review}.

This is where LLMs, such as GPT-4 and domain-specific models like TaxBERT~\cite{hechtner2025design}, offer a compelling leap forward. LLMs can process vast amounts of unstructured textual data, understand nuanced language, and generate contextually rich responses—capabilities well-suited to the linguistically dense and interpretive domain of taxation. For example, Choi and Kim~\cite{choi2024firm} leverage GPT-4 to develop a novel firm-level measure of tax audits by extracting insights from boilerplate narrative disclosures in 10-K filings. Their study reveals that tax audits significantly deter tax avoidance, with lasting effects post-audit, while also introducing corporate risks such as reduced investment and increased volatility. These insights, previously out of reach due to data limitations, underscore the transformative potential of LLMs in empirical tax research.

On the practitioner side, generative AI is gaining traction among tax professionals. According to a 2023 Thomson Reuters Institute survey~\cite{ThomsonReuters2023}, nearly three-quarters of tax professionals recognize the utility of ChatGPT and similar models for tasks like tax research, return preparation, and client advisory services. These models can help taxpayers understand complex tax codes, identify deductible expenses, and evaluate compliance risks without incurring the costs of professional services. However, concerns persist about model accuracy and transparency. As demonstrated in various evaluations, generic models such as ChatGPT-3.5 may "hallucinate" or produce legally incorrect advice when not properly grounded~\cite{huang2025survey, wang2023survey}, emphasizing the need for specialized or fine-tuned models~\cite{alarie2023rise, zhang2023four}.

Recent academic contributions can be categorized into three major streams. First, empirical measurement using LLMs is exemplified by Choi and Kim ~\cite{choi2024firm}, who construct a validated, firm-level proxy for tax audits using GPT-4, unlocking new research avenues into audit impacts and tax compliance behavior. Second, domain-specific model development is represented by Hechtner et al.~\cite{hechtner2025design}, who introduce TaxBERT, a BERT-based model fine-tuned on tax-specific corpora. Their work demonstrates the superior performance of specialized models over generic LLMs in parsing nuanced tax disclosures. Third, model benchmarking and evaluation are advanced by Choi et al.~\cite{choi2025taxation}, who present PLAT, a benchmark of complex tax penalty cases requiring reasoning beyond statutory interpretation. Their results show that while vanilla LLMs struggle with these tasks, multi-agent LLM architectures with retrieval and self-reflection mechanisms can substantially improve performance.

In sum, while the application of LLMs in taxation is still emerging, early evidence from both research and practice illustrates substantial promise. Future work should further explore how LLMs can enhance taxpayer support, assist regulators in enforcement, and drive more nuanced academic insights—all while addressing critical issues of reliability, interpretability, and domain adaptation.

\subsubsection{Benchmarks}

\textbf{PLAT}~\cite{choi2025taxation} is a benchmark designed to evaluate the ability of LLMss to reason about complex taxation issues, specifically focusing on the legitimacy of additional tax penalties. Unlike earlier datasets that emphasize deductive reasoning from explicit tax statutes, PLAT challenges models to make nuanced judgments based on ambiguous legal standards and factual contexts drawn from Korean court precedents. The dataset requires comprehensive legal understanding and conflict resolution between competing principles, such as taxpayer responsibility versus the protection of legitimate expectations. Experiments show that while baseline LLMs struggle with these tasks, retrieval augmentation, self-reasoning, and multi-agent role-playing significantly enhance performance.

\textbf{SARA}~\cite{holzenberger2020dataset} is an earlier dataset focused on computational statutory reasoning in U.S. tax law. It consists of simplified statutes from the U.S. Internal Revenue Code, along with natural language questions formulated as textual entailment and numerical prediction tasks. Each case requires the application of tax laws to specific fact patterns, emphasizing deductive reasoning based solely on statutory rules. SARA highlights the challenges of grounding language understanding in prescriptive legal norms and motivates the development of models capable of logical inference from structured legal texts.

\subsubsection{Discussion}

\noindent\textbf{Opportunities and Impact.}
The emergence of LLMs presents a transformative opportunity across accounting research and practice. As demonstrated in auditing~\cite{gu2024artificial, wangauditbench}, financial and managerial accounting~\cite{kim2023bloated, li2024applying}, and taxation~\cite{choi2024firm, hechtner2025design}, LLMs offer unprecedented capabilities for processing unstructured financial data, automating repetitive tasks, interpreting regulatory language, and assisting in judgment-intensive activities.

In auditing, LLMs enhance traditional practices by expanding coverage, reducing manual errors, and enabling real-time anomaly detection. In financial and managerial accounting, they streamline reporting, improve transparency, and support strategic analysis by synthesizing complex disclosures. In taxation, LLMs facilitate empirical research, regulatory compliance, and taxpayer support through efficient navigation of complex legal language. Demonstrating this potential at scale, Deloitte’s Zora AI aims to transform financial operations by \textbf{saving up to 25\% in costs} and \textbf{boosting productivity by up to 40\%}, freeing thousands of hours annually for finance teams~\cite{Deloitte2025ZoraAI}. This real-world example shows that LLMs are already powerful collaborators driving efficiency and better decision-making across accounting disciplines.

\noindent\textbf{Challenges and Limitations.}
Despite their potential, significant challenges must be addressed before LLMs can be widely and responsibly deployed in accounting contexts.

First, accuracy and hallucination risks remain a persistent concern. LLMs can generate plausible yet incorrect outputs, posing serious risks in high-stakes settings such as audits, regulatory compliance, or tax advisory~\cite{huang2025survey, fotoh2023exploring}. Second, lack of domain-specific reasoning limits their ability to consistently interpret complex accounting standards or apply nuanced legal doctrines, particularly when precise statutory interpretation is required~\cite{hechtner2025design, choi2025taxation}.

Third, ethical and regulatory challenges arise from the integration of LLMs into sensitive accounting workflows. Issues such as audit independence, data privacy, confidentiality, and accountability for AI-assisted outputs must be carefully managed~\cite{fotoh2023exploring, de2024will}.

Moreover, prompt sensitivity and model opacity complicate the interpretability and reproducibility of LLM-based results~\cite{wangauditbench}, requiring the development of robust validation protocols and transparent deployment strategies.

\noindent\textbf{Research Directions.}
To realize the full potential of LLMs in accounting while mitigating risks, several critical research directions emerge:

\begin{itemize}[leftmargin=10pt]
    \item \textbf{Domain-Specific Fine-Tuning and Customization.} Fine-tuning LLMs on accounting-specific corpora—such as financial statements, audit reports, tax codes, and regulatory filings—can substantially improve relevance, accuracy, and trustworthiness~\cite{hechtner2025design, li2024applying}.
    
    \item \textbf{Hybrid Human-AI Systems.} Emphasizing "co-piloted" models where human expertise remains central will ensure that LLMs complement rather than replace professional judgment, particularly in high-risk domains like auditing and taxation~\cite{gu2024artificial}.
    
    \item \textbf{Explainable and Verifiable Outputs.} Research into techniques such as chain-of-thought prompting, retrieval-augmented generation (RAG), and citation-grounded generation will be vital for enhancing auditability, interpretability, and regulatory compliance~\cite{berger2023towards, fohr2023deep}.
    
    \item \textbf{Robust Evaluation Benchmarks.} Developing comprehensive, accounting-specific benchmarks (e.g., AuditBench, PLAT) will facilitate objective assessment of LLM capabilities, biases, and limitations in real-world financial tasks~\cite{wangauditbench, choi2025taxation}.
    
    \item \textbf{Ethical Governance Frameworks.} Accounting research must proactively address ethical, privacy, and regulatory concerns by formulating best practices, disclosure standards, and accountability mechanisms for AI deployment~\cite{fotoh2023exploring, de2024will}.
\end{itemize}

\noindent\textbf{Conclusion.}
LLMs are poised to become powerful allies in accounting research and practice, enhancing the ability to process, interpret, and reason over complex financial information. However, realizing this potential responsibly demands rigorous domain adaptation, human-centered oversight, transparent evaluation, and a strong ethical foundation. Rather than replacing traditional accounting methods and judgment, LLMs should be viewed as enablers—tools that expand the scope, depth, and inclusivity of financial reasoning in an increasingly complex and dynamic economic environment.

\subsection{Marketing}

\subsubsection{Overview}

\noindent\textbf{Introduction.}
Marketing, in its formal sense, refers to the set of institutions, processes, and activities for creating, communicating, delivering, and exchanging offerings that have value for customers, clients, partners, and society at large~\cite{kotler2015marketing}. At its core, marketing is a discipline rooted in understanding and influencing consumer behavior to fulfill organizational objectives. In simpler terms, marketing is about understanding what people need or want and finding effective ways to present and deliver products or services that meet those needs. It encompasses everything from identifying target audiences and researching their preferences, to designing products, setting prices, promoting offers, and ensuring smooth delivery.

Within the field of marketing, researchers address a range of tasks aimed at generating insights to inform managerial decision-making. These tasks typically fall under categories such as market segmentation, consumer behavior analysis, product positioning, pricing strategy, brand management, advertising effectiveness, and customer satisfaction measurement. Traditional research methods for approaching these tasks include qualitative techniques like interviews and focus groups~\cite{calder1977focus, mariampolski2001qualitative}, as well as quantitative approaches such as surveys, experiments, conjoint analysis, and econometric modeling~\cite{franses2001quantitative, dzwigol2020innovation}. These techniques have provided robust frameworks for testing hypotheses, identifying causal relationships, and quantifying consumer preferences~\cite{malhotra2020marketing}.

The contribution of these traditional methods has been substantial. They have grounded the field of marketing in scientific rigor, enabling systematic inquiry into consumer psychology and market dynamics. Surveys, for instance, are a cornerstone of quantitative research, with online surveys being utilized regularly by 85\% of market research professionals, followed by mobile surveys at 47\% and proprietary panels at 32\%~\cite{backlinko2025marketresearch}. Through this, researchers and practitioners alike have developed theories and models that remain foundational to both academia and industry. These approaches have also proven their value in real-world applications, informing strategies that drive competitive advantage, optimize customer experiences, and maximize return on investment. The field has benefitted immensely from its interdisciplinary nature, drawing from psychology, economics, sociology, and statistics, thereby enriching both theoretical developments and practical implementations~\cite{fligstein2007sociology, mazzocchi2008statistics, glaeser2004psychology}.

Despite its strengths, marketing research faces a host of persistent challenges. One major difficulty lies in the inherent complexity and unpredictability of human behavior, which resists reduction into fixed, universal models. Consumers often act inconsistently, influenced by subtle emotional, social, and contextual cues that traditional research tools struggle to capture. Moreover, issues such as data sparsity, high dimensionality, and response biases, ranging from social desirability effects to non-response tendencies, undermine the reliability of collected data~\cite{insight7_limitations_2024}. In today’s fast-moving digital landscape, the volume and velocity of information further complicate matters~\cite{pascucci2023digital}. Researchers must grapple with monitoring vast amounts of unstructured data from social media, product reviews, forums, and industry publications, all of which demand new analytic capabilities. Manual tasks like survey coding and qualitative response analysis remain time-consuming and resource-heavy, while the growing demand for real-time, personalized content creates pressure to scale insights rapidly without sacrificing nuance. Together, these factors present a significant challenge for marketing professionals seeking to generate timely, accurate, and actionable insights.

\begin{figure}[!t]
    \centering
    \includegraphics[width=0.99\linewidth]{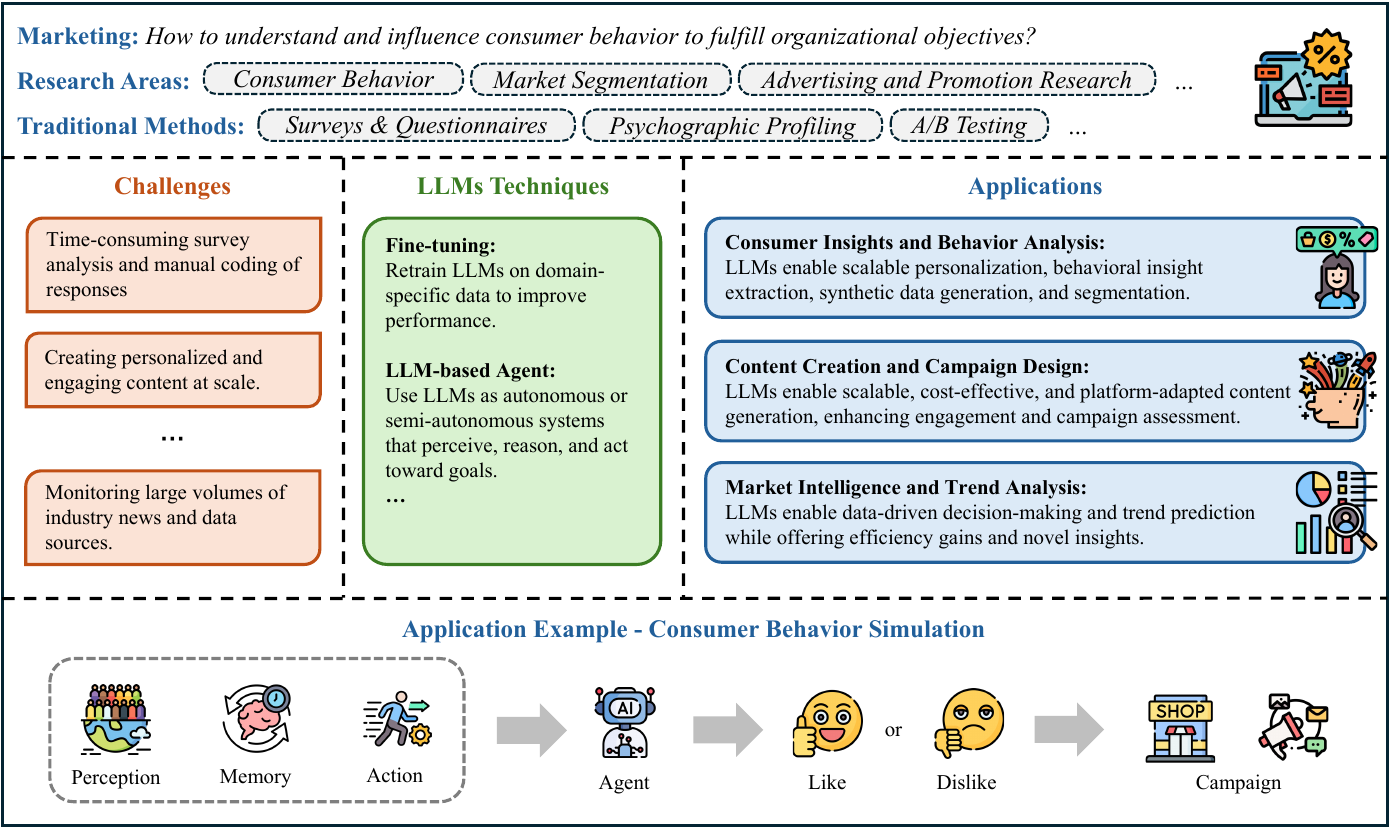}
    \caption{Overview of LLMs' Applications in Marketing Research.}
    \label{fig:marketing}
    \vspace{-10pt}
\end{figure}

\noindent\textbf{The Role of LLMs.}
Some of these challenges are beyond the reach of current LLMs. For instance, LLMs cannot autonomously conduct randomized controlled trials or derive causal inference with high reliability without carefully structured datasets and external validation. Nor can they fully replicate the depth of human empathy or ethical judgment necessary for designing sensitive interventions in consumer psychology~\cite{albrecht2022despite}. However, LLMs are particularly well-suited for tasks where the core requirement is understanding, generating, or summarizing large volumes of unstructured text. They excel at automating tasks such as sentiment analysis, topic modeling, content generation, customer service automation, and text-based survey coding. Their ability to synthesize information across sources, generate coherent narratives, and respond adaptively to prompts positions them as powerful tools for both qualitative and quantitative marketing research~\cite{Amini2024AnOO, Verma2021ArtificialII, Jain2023ThePA, Aghaei2025HarnessingTP}. Specifically, LLMs have shown utility in early-stage ideation, exploratory analysis, automated reporting, and even in the creation of synthetic respondents for preliminary hypothesis testing. The reason for their effectiveness lies in their training on vast corpora of language data, enabling them to generalize linguistic patterns, identify latent themes, and approximate conversational nuance with increasing sophistication.

\input{tables/marketing}

\noindent\textbf{Taxonomy.}
To understand the potential application of LLMs in marketing research, we propose a taxonomy that reflects the diversity of tasks across the field:

\begin{itemize}[leftmargin=10pt]
    \item \textbf{Consumer Insights and Behavior Analysis.} Consumer Insights and Behavior Analysis focuses on understanding the motivations behind consumer thoughts, feelings, and actions to inform effective marketing strategies. While traditional methods like surveys and interviews offer value, they often struggle with the scale and nuance of modern, unstructured data sources. LLMs are transforming this field by enabling scalable, nuanced analysis of language-rich data, offering deeper, real-time insights into consumer behavior.
    \item \textbf{Content Creation and Campaign Design.} Content Creation and Campaign Design are key pillars of marketing, combining creative storytelling with strategic planning to engage audiences and achieve business goals. Traditionally reliant on manual effort and intuition, this process has faced challenges in scalability, personalization, and real-time feedback. Today, LLMs are transforming how content is ideated, produced, and optimized, enhancing creativity, streamlining workflows, and enabling more dynamic, data-driven campaigns.
    \item \textbf{Market Intelligence and Trend Analysis.} Market Intelligence and Trend Analysis are essential for guiding strategic marketing decisions, helping businesses monitor competitors, anticipate consumer shifts, and navigate evolving market conditions. Traditionally grounded in surveys and expert insights, these methods often lag behind today’s fast-paced digital environment. LLMs are revolutionizing this space by enabling real-time analysis of vast, unstructured data sources, offering marketers faster, deeper, and more forward-looking insights to stay competitive and adaptive in a rapidly changing landscape.
\end{itemize}

Across marketing research tasks, LLMs do not replace established theories or empirical methods, but instead expand the field’s interpretive and creative capabilities by translating unstructured language into insight, automating content generation, and enhancing strategic awareness. They bridge the gap between human communication and data-driven analysis, enabling more personalized engagement, faster iteration, and richer understanding of consumer behavior and market dynamics.

\subsubsection{Consumer Insights and Behavior Analysis}

Consumer insights and behavior analysis lie at the heart of marketing strategy, aiming to understand why people think, feel, and act as they do in the marketplace~\cite{Solomon1993ConsumerBB}. These insights help businesses tailor products, messaging, and experiences to meet real consumer needs~\cite{Hawkins1997ConsumerBB, Peter1990ConsumerBA}. Traditionally, marketers have relied on surveys, interviews, and focus groups to collect this data. While useful, these tools often fall short in capturing the scale, complexity, and subtlety of modern consumer behavior, especially in a digital landscape teeming with unstructured data like reviews, chats, and social posts.

LLMs offer a powerful alternative. With their ability to understand and generate human-like text, LLMs can analyze vast amounts of consumer content, identifying sentiment, extracting themes, and revealing hidden patterns~\cite{Paul2023ChatGPTA}. Unlike manual methods, LLMs can scale effortlessly and respond in real time, making them well-suited for decoding consumer reviews, personalizing communication, and even generating synthetic consumer data. For instance, marketers can use LLMs to simulate customer personas or segment audiences based on open-text responses, bridging gaps left by rigid demographic approaches~\cite{Wang2024LargeLM, Sarstedt2024UsingLL, LiConsumerSW,chen2025personatwin}.

Recent work highlights the breadth of LLM applications in this space. Praveen et al.~\cite{Praveen2024CraftingCL} show that fine-tuned models outperform traditional tools in extracting emotion and sentiment from consumer reviews, especially in service industries. Li et al.~\cite{LiConsumerSW} demonstrate how LLMs improve consumer segmentation by embedding and clustering open-ended survey data, while also enabling high-accuracy persona simulations. Sarstedt et al.~\cite{Sarstedt2024UsingLL} explore the use of “silicon samples”—synthetic respondents generated by LLMs—and suggest they are especially valuable in pretesting or pilot stages, though not always interchangeable with human data.

Wang et al.~\cite{Wang2024LargeLM} address this concern by proposing a data augmentation approach that blends real and synthetic data to reduce bias in conjoint analysis. Their method enhances accuracy without the high costs of traditional data collection. Similarly, Goli and Singh~\cite{Goli2024FrontiersCL} find that LLMs can struggle to fully replicate human preferences but show promise in uncovering heterogeneity when guided by structured prompting techniques like “chain-of-thought conjoint”.

Broader applications of LLMs also include content generation, hyperpersonalization, and dynamic customer interaction, as shown in research by Paul et al.~\cite{Paul2023ChatGPTA} and Kumar et al.~\cite{Naik2024UnlockingBE}, further emphasizing their role as both analytical engines and interactive tools in modern marketing.

In sum, LLMs are reshaping how consumer insights are gathered and interpreted. While they are not without limitations, their ability to process complex language at scale marks a step-change in understanding and anticipating consumer behavior.

\subsubsection{Content Creation and Campaign Design}

In the realm of marketing, content creation and campaign design are central to how brands communicate their message and engage audiences. Content creation involves crafting valuable, relevant, and compelling digital materials, such as social media posts, blog entries, videos, or infographics, that resonate with the target audience~\cite{Balte2015ContentM}. Campaign design, on the other hand, is the strategic process of planning and organizing this content across various channels to achieve specific marketing objectives, such as brand awareness, lead generation, or customer loyalty~\cite{Dwivedi2015SocialMM, Johnson2005OnTS}. One can think of a brand as a storyteller in a crowded marketplace. The content is the voice, tone, and story, while the campaign design is the stage, spotlight, and choreography that guide the narrative and make sure it reaches the right eyes and ears at the right moment.

Traditionally, content creation and campaign design have relied heavily on human creativity and intuition. Marketers would brainstorm ideas, conduct market research, and produce content manually—a process that is often time-consuming and resource-intensive. Challenges in this traditional approach include difficulties in consistently generating high-quality content, the substantial time and effort required for research and planning, and the complexities involved in tailoring content to diverse audience segments across various platforms. Additionally, measuring the effectiveness of campaigns and obtaining real-time feedback posed significant hurdles, making it challenging to optimize strategies promptly~\cite{flyingvgroup_content_marketing}.

The advent of LLMs has introduced transformative solutions to these challenges. LLMs are advanced AI systems capable of understanding and generating human-like text, enabling marketers to automate and enhance various aspects of content creation and campaign design~\cite{Wahid2023EditorialWB, ivanov2023ai_marketing}. For instance, LLMs can generate drafts, headlines, social media captions, and even long-form articles within minutes, significantly boosting productivity and allowing marketers to focus more on strategy and creativity. Furthermore, LLMs can analyze vast amounts of data to personalize content for specific audience segments, ensuring that the messaging resonates more effectively with diverse groups. This personalization extends to creating engaging social media captions that enhance user interaction and brand presence~\cite{sharma2024llms}.

A growing body of research explores this shift. Reisenbichler et al.~\cite{Reisenbichler2022FrontiersSC} proposed a semiautomated content generation framework combining natural language generation (NLG) with human editing to produce SEO-optimized website content. Their findings show that machine-generated content, once refined by humans, can outperform human-written content in search rankings, while also reducing production costs by over 90\%. Similarly, Ivanov~\cite{ivanov2023ai_marketing} investigated audience perceptions of AI versus human-generated marketing content and found minimal differences, suggesting that consumers may not distinguish between the two when quality is high. In more experimental territory, Kasuga and Yonetani~\cite{Kasuga2024CXSimulatorAU} introduced a CXSimulator that uses LLM embeddings to simulate user behavior and assess the likely outcomes of different marketing campaigns, offering a way to pre-test ideas without real-world deployment.

Meanwhile, Aldous et al.~\cite{Aldous2024UsingCI} demonstrated how ChatGPT-generated content could outperform human-created posts in terms of emotional appeal and engagement, particularly on platforms like Facebook, highlighting the model’s ability to adapt content to platform-specific norms. This adaptability and personalization have been emphasized as crucial in recent editorial work on generative AI in content marketing~\cite{GolabAndrzejak2023, Huh2023ChatGPTAA}, where the focus is shifting from merely producing content to orchestrating entire advertising campaigns where AI helps strategize, generate, test, and iterate in rapid cycles.

Altogether, these developments point to a broader evolution: content creation and campaign design are becoming increasingly algorithmic, yet no less human. Rather than replacing creativity, LLMs are reconfiguring it—helping marketers shift from being sole creators to becoming skilled orchestrators of human-AI collaboration. As the boundaries blur, what emerges is a hybrid model of marketing practice, powered not only by technology but also by new workflows, new expectations, and new possibilities.

\subsubsection{Market Intelligence and Trend Analysis}

Market Intelligence (MI) and Trend Analysis play a vital role in modern marketing, serving as the compass by which organizations navigate competitive landscapes and shifting consumer demands. At its core, Market Intelligence involves collecting and analyzing data about market conditions, consumer behavior, competitors, and emerging trends to support strategic decision-making~\cite{Jenster2009MarketIB, Wee2001TheUO}. Trend Analysis, in tandem, looks beyond immediate patterns to identify longer-term movements in customer preferences, technology, and broader economic forces. One can think of MI and Trend Analysis as the business equivalent of weather forecasting: companies read the “climate” of the market to prepare for storms, seize sunshine, and stay ahead of seasonal changes.

Traditionally, businesses have relied on surveys, focus groups, expert panels, and basic statistical tools to conduct MI and Trend Analysis. These methods, though foundational, face significant limitations in speed, scalability, and responsiveness. For instance, survey-based data collection can be expensive and time-consuming, often reflecting stale consumer sentiments by the time analysis is complete. Moreover, as Soykoth et al.~\cite{Soykoth2024ResearchTI} highlight, the exponential growth of digital data has made it increasingly difficult to extract timely insights using manual or fragmented approaches. The inability to process unstructured data like social media posts, product reviews, or real-time feedback further constrains traditional methods.

This is where LLMs, such as ChatGPT, present a transformative opportunity. LLMs can process and generate human-like language, making them powerful tools for interpreting vast volumes of textual data, uncovering hidden patterns, and even simulating consumer responses. Unlike rigid models that require structured inputs, LLMs can parse unstructured content such as tweets, reviews, news, forum posts and distill meaningful insights with minimal setup. Mutoffar et al.~\cite{mutoffarRoleChatGPTInnovative2024} found that ChatGPT can enhance the ability of firms to predict market trends by offering nuanced interpretations of consumer sentiment and behavioral shifts.

Recent research and case studies further illuminate this evolution. For example, Saputra et al.~\cite{Saputra2023TheIO} demonstrated that using ChatGPT to generate Instagram marketing content improved campaign effectiveness by driving higher user engagement under the AIDA (Attention, Interest, Desire, Action) framework. These cases suggest not only improved efficiency but also enhanced creativity and adaptability in digital marketing practices. A compelling narrative of progress also emerges from the literature around MarkBot~\cite{Kushwaha2021MarkBotA}, a chatbot framework powered by language models that leverages user-generated content to minimize the lead time for deployment in marketing environments. Similarly, Yeykelis et al.~\cite{Yeykelis2024UsingLL} demonstrated that LLMs could replicate human responses in marketing experiments with impressive fidelity, opening the door to AI-powered replication and prediction in media effects research.

These developments also resonate with the data-augmentation framework proposed by Wang et al.~\cite{Wang2024LargeLM}, who show that LLMs, when properly integrated with real-world data, can dramatically reduce the cost and time associated with conjoint analysis and preference modeling, albeit with careful consideration of biases and calibration.

Taken together, the literature suggests that we are witnessing a pivotal shift: from retrospective, human-labor-intensive MI and Trend Analysis to a forward-looking, AI-augmented ecosystem. While traditional methods still provide foundational rigor, LLMs offer a flexible and scalable supplement—or in some contexts, a powerful alternative. This hybrid approach may well define the next frontier in market intelligence.

\subsubsection{Benchmarks}

Although dedicated datasets and benchmarks specifically targeting LLM applications in marketing are still scarce, existing large-scale public datasets provide valuable resources for exploratory research and model evaluation. Google BigQuery, in particular, offers several comprehensive datasets related to user behavior, advertising, and public trends. These datasets serve as promising foundations for developing and benchmarking marketing-oriented LLM systems. Below, we list several notable examples (Table~\ref{tab:marketing_datasets}).

\input{tables/benchmark_marketing.tex}

\textbf{GA4 E-commerce Sample} provides anonymized Google Analytics 4 event export data from the Google Merchandise Store. It captures detailed user behavior events such as pageviews, add-to-cart actions, and purchases, enabling analysis of customer journeys, conversion funnels, and marketing campaign effectiveness.

\textbf{TheLook E-commerce} is a synthetic e-commerce dataset designed for business intelligence and analytics training. It includes users, orders, products, inventory, and events data, suitable for customer segmentation, sales forecasting, and conversion analysis.

\textbf{Google Trends} contains search interest data over time for various keywords and regions. It enables marketers to analyze public interest dynamics, track emerging topics, and correlate search patterns with marketing efforts.

\textbf{GDELT} collects worldwide news event metadata, including actors, locations, event types, and sentiment. Marketers can use it to monitor brand exposure, public sentiment, and global events affecting market behavior.

\textbf{Google Ads Transparency Report} provides data on political and issue-based advertising on Google's platforms, including ad content, spending, and targeting. It is valuable for studying advertising strategies, campaign transparency, and audience targeting trends.

These datasets significantly expand the opportunities for applying LLMs in marketing research. They offer rich, diverse, and real-world sources of structured and unstructured information that LLMs can effectively process to generate actionable insights. For example, LLMs can be used to perform large-scale customer journey analysis by synthesizing event-level data from GA4 exports, or to extract latent market trends by analyzing temporal patterns in Google Trends data. TheLook E-commerce dataset enables training and evaluation of LLM-based models for tasks such as automated customer segmentation, personalized recommendation generation, and sales forecasting. Similarly, the GDELT database allows LLMs to monitor and summarize brand mentions, sentiment shifts, and emerging crises across global news streams, supporting real-time market intelligence applications. The Ads Transparency Report can serve as a foundation for studying advertising strategies, targeting behaviors, and political messaging dynamics through LLM-driven content analysis and clustering. Together, these datasets provide a practical and scalable substrate for developing, fine-tuning, and benchmarking LLM systems in marketing contexts, advancing both academic research and industry practice.

\subsubsection{Discussion}

\noindent\textbf{Opportunities and Impact.} The application of LLMs in marketing research presents transformative opportunities across multiple domains. As seen in consumer insights~\cite{Wang2024LargeLM, Sarstedt2024UsingLL, LiConsumerSW}, content creation~\cite{ivanov2023ai_marketing, Reisenbichler2022FrontiersSC, Aldous2024UsingCI}, and market intelligence~\cite{mutoffarRoleChatGPTInnovative2024, Wang2024LargeLM}, LLMs greatly enhance scalability, speed, and analytical depth. In consumer insights, LLMs uncover latent sentiments and behavioral patterns from vast unstructured data. In content creation, they accelerate ideation and enable hyper-personalization. In market intelligence, they provide real-time synthesis of competitive landscapes and emerging trends.

A 2024 AMA survey reports that \textbf{90\% of marketers now use generative AI}, with 71\% using it weekly and nearly 20\% daily. \textbf{Over 85\% report productivity gains}, and half cite improvements in both the quality and quantity of creative output~\cite{cashion2024generative}. These developments show that LLMs are not just automation tools but strategic partners redefining creativity and decision-making in marketing.

\noindent\textbf{Challenges and Limitations.}
Despite their considerable promise, deploying LLMs in marketing research raises important challenges. First, bias and hallucination risks persist. LLMs trained on broad internet corpora can perpetuate stereotypes or generate plausible yet inaccurate outputs, which, if unrecognized, could misguide marketing strategies~\cite{Paul2023ChatGPTA, Sarstedt2024UsingLL}. Second, lack of causal inference and experimental rigor limits LLMs' ability to replace traditional empirical methods like randomized controlled trials~\cite{albrecht2022despite}. Their outputs, while rich, are often correlational rather than causal, necessitating careful interpretation and external validation. Third, prompt sensitivity and reproducibility concerns complicate the use of LLMs for systematic research. Slight variations in prompting or task framing can yield inconsistent results~\cite{Sarstedt2024UsingLL, Wang2024LargeLM}, undermining reliability. Finally, ethical considerations arise regarding transparency, authorship, and authenticity, especially as AI-generated content becomes increasingly indistinguishable from human-created work~\cite{GolabAndrzejak2023}.

\noindent\textbf{Research Directions.}
Addressing these challenges points to several important research directions:

\begin{itemize}[leftmargin=10pt]
    \item \textbf{Hybrid Approaches Combining LLMs with Human Oversight.} Structuring workflows where human judgment complements AI output will help mitigate risks of error, bias, and unethical deployment.
    
    \item \textbf{Domain-Specific Fine-Tuning and Contextualization.} Fine-tuning LLMs on marketing-specific corpora and continuously updating them with domain-relevant data can improve accuracy, nuance, and relevance in marketing applications~\cite{mutoffarRoleChatGPTInnovative2024}.
    
    \item \textbf{Benchmarking and Standardized Evaluation.} Creating rigorous benchmarks for LLM performance in marketing tasks, such as synthetic persona realism, content personalization efficacy, and market trend prediction accuracy, will be vital for advancing scientific rigor.
    
    \item \textbf{Explainability and Interpretability.} Incorporating methods such as chain-of-thought prompting, citation grounding, and attribution tracing will enhance transparency and user trust in LLM-assisted marketing insights~\cite{Wang2024LargeLM}.
    
    \item \textbf{Ethical Guidelines and Best Practices.} Marketing researchers must proactively develop frameworks for ethical AI use, covering disclosure norms for AI-generated content, fairness in segmentation, and protection against manipulative targeting.
\end{itemize}

\noindent\textbf{Conclusion.}
LLMs are redefining the landscape of marketing research by unlocking new capacities for insight generation, content creation, and strategic analysis. However, their integration must be thoughtful, combining the creative power of human marketers with the linguistic and analytical capabilities of AI. Moving forward, marketing will likely evolve into a hybrid discipline, where the agility, scale, and personalization offered by LLMs complement traditional empirical rigor, creativity, and ethical responsibility to shape a more dynamic, insightful, and consumer-centered future.

\newpage
\section{LLMs for Science and Engineering}
\label{sec:discpline-3}

In this chapter, we chart how LLMs are utilized across science and engineering. The chapter spans mathematics; physics and mechanical engineering; chemistry and chemical engineering; life sciences and bioengineering; earth sciences and civil engineering; and computer science and electrical engineering. We open with \textbf{mathematics} including proof support, theoretical exploration and pattern discovery, math education, and targeted benchmarks. In \textbf{physics and mechanical engineering}, we cover documentation-centric tasks, design ideation and parametric drafting, simulation-aware and modeling interfaces, multimodal lab and experiment interpretation, and interactive reasoning, followed by domain-specific evaluations and a look at opportunities and limits. In \textbf{chemistry and chemical engineering}, we examine molecular structure and reaction reasoning, property prediction, materials optimization, test/assay mapping, property-oriented molecular design, and reaction-data knowledge organization, then compare benchmark suites. In \textbf{life sciences and bioengineering} section, we include genomic sequence analysis, clinical structured-data integration, biomedical reasoning and understanding, and hybrid outcome prediction, with emphasis on validation standards. In \textbf{earth sciences and civil engineering} section, we review geospatial and environmental data tasks, simulation and physical modeling, document workflows, monitoring and predictive maintenance, plus design/planning tasks, again with benchmarks. We close with \textbf{computer science and electrical engineering}: code generation and debugging, large-codebase analysis, hardware description language code generation, functional verification, and high-level synthesis, capped by purpose-built benchmarks and a final discussion of impacts and open challenges.

\subsection{Mathematics and Statistics}

\subsubsection{Overview}
\textbf{Introduction.} 
Mathematics is the study of abstract structures and relationships conceived through axiomatic systems and logical deduction, focusing on quantities, shapes, patterns, and transformations~\cite{eves1997foundations, courant1996mathematics, horsten2007philosophy}.
It provides a framework for formulating hypotheses precisely, exploring them using consistent rule-based reasoning, and deriving theorems that remain valid whenever the initial assumptions hold.
In other words, mathematics is a way to understand and describe the world by looking at patterns, shapes, and numbers.
Contemporary mathematical research covers a broad landscape, from both the abstract realms of pure mathematics and the practical studies of applied mathematics~\cite{national2013mathematical, treves2016topological, lax2016scattering}. 
Specifically, pure mathematical research centers on formulating and investing original mathematical concepts, aiming at advance mathematical knowledge without direct practical applications~\cite{hersh1998mathematics}, while applied mathematics aims at developing mathematical techniques for applications in science and other domains, or leveraging techniques from other fields to contribute to mathematics~\cite{holmes2009introduction, logan2013applied}.
Statistics is the science of collecting, analyzing, interpreting, and communicating data to understand uncertainty and make informed decisions. It provides the mathematical foundation for drawing conclusions from imperfect or incomplete information, enabling us to identify patterns, quantify variability, evaluate evidence, and predict future outcomes. Statistics plays a central role  in a wide spectrum of domains -- from scientific research and public policy to business, health, and technology, in transforming raw data into meaningful insights that guide actions and advance knowledge.

Theoretical mathematical and statistical research focuses on developing abstract theories and rigorous proofs to advance fundamental understanding and it faces both traditional and emerging challenges, particularly as the boundaries between disciplines continue to blur.  Modern mathematical problems increasingly arise from complex real-world systems in fields such as  biology, physics, economics, and engineering. This shift requires mathematicians and statisticians not only to master advanced theory but also to engage with domain-specific knowledge, navigate diverse research cultures, and collaborate effectively across disciplines. The need to translate abstract ideas into models that address practical questions while preserving mathematical rigor adds an additional layer of complexity. 

On the other hand,  traditional mathematical inquiry itself presents substantial barriers to entry and significant challenges for collaboration 
First, 
Solving a mathematical research problem requires sustained effort and intense concentration, commonly extending over months or even years~\cite{etingofPRIMES}.
For instance, engaging with prior mathematical literature often requires deep familiarity with underlying theories, as the arguments can be highly technical and conceptually demanding. 
In addition, traditional mathematical research can also present \textbf{considerable barriers to entry and promote challenges in collaboration}~\cite{epoch2024}. 
Acquiring the foundational background knowledge required to contribute meaningfully to mathematical fields often typically demands years of dedicated study. 
Even for senior researchers, effective collaboration can be hindered by the need for a shared understanding across different, often highly specialized, subdomains of mathematics. 
Moreover, \textbf{formalizing arguments into rigorous mathematical formats},  especially those suitable for verification by machines, is often a \textbf{laborious and long process}. 
Furthermore, the abstraction and complexity of many mathematical objects also pose significant limitations. 
Working with \textbf{concepts involving infinite structures, high-dimensional space, complex sets of equations} often stretches the \textbf{limits of human intuition and comprehension}. 
Similarly, understanding abstract patterns within extremely large datasets or highly complex mathematical systems can be prohibitively difficult to perform manually~\cite{adviser2021Mathematicians}.

As for mathematics education, it typically involves direct instruction where teachers explicitly explain how to solve specific types of problems, with several standard methods~\cite{thurston2005mathematical}. 
It emphasizes procedural methods, formula memorization, and repetitive practice to master mathematical concepts. 
Skills and concepts are usually taught in a logical sequence, with a focus on building foundational arithmetic skills before progressing to more complex topics like algebra and geometry. 
Standardized testing is frequently used to assess understanding and retention.
Despite the success of the aforementioned mathematical education, several studies have argued for its limitations and challenges.
For example, a study~\cite{kajander2014mathematical} mentions that traditional methods \textbf{overemphasize memorization and rote learning, failing to promote a deep conceptual understanding of mathematical principles}. 
Students may be able to perform calculations without understanding how the procedures work.
In addition, Peter, E. E.(2012)~\cite{peter2012critical} claims that focusing on procedural fluency can \textbf{ negate} the development of \textbf{critical thinking and problem-solving skills needed to apply mathematical knowledge in novel situations and real-world contexts}.

\begin{figure}[!t]
    \centering
    \includegraphics[width=\linewidth]{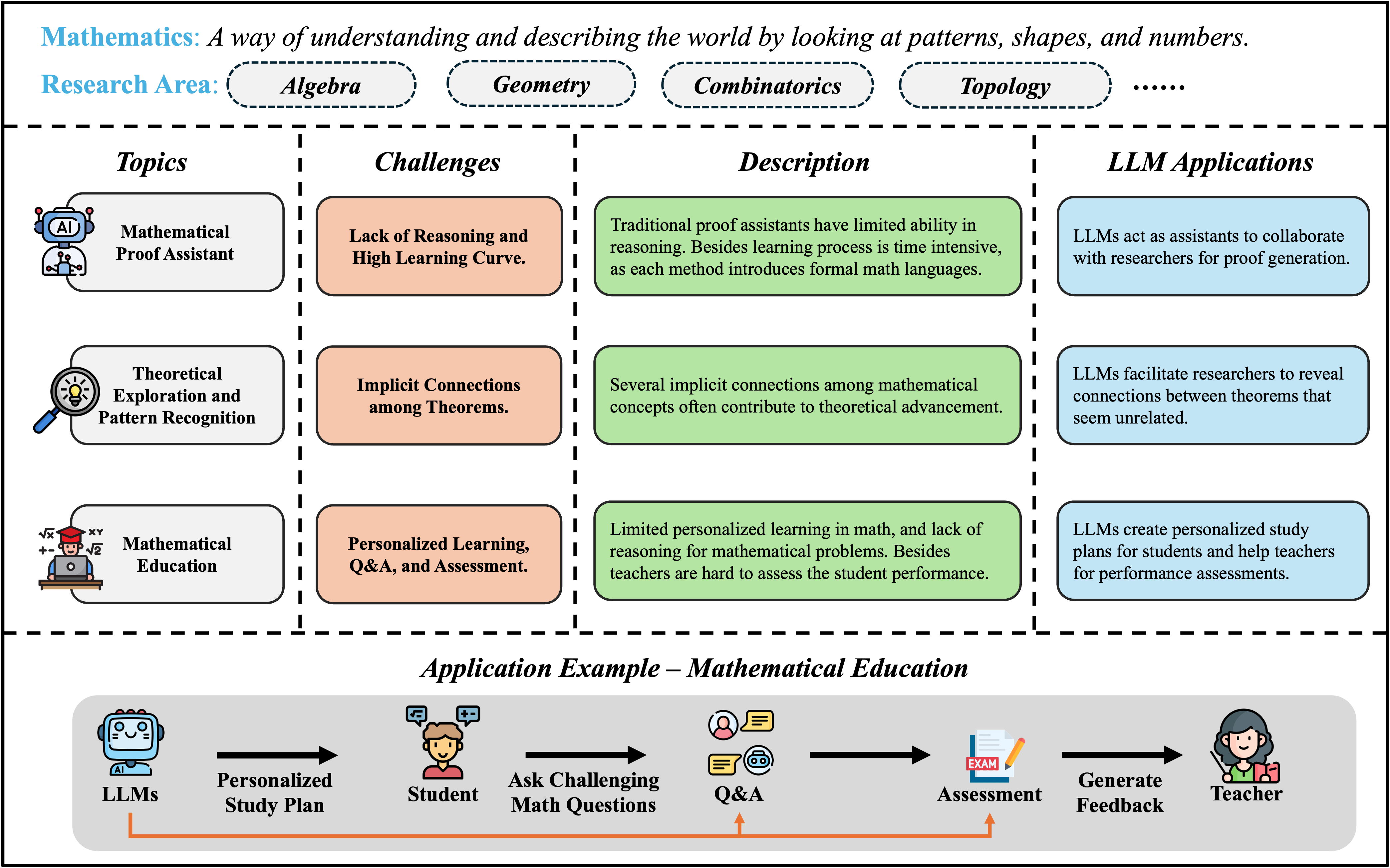}
    \caption{Overview of topics in Mathematics Discipline with LLMs solution.}
    \label{fig:math}
\end{figure}

\textbf{Role of LLMs.}
LLMs offer a promising avenue to address several limitations inherent in traditional mathematical research. 
One of the most significant potential contributions lies in \textbf{assisting with proof development and verification}~\cite{lama2024benchmarking}. 
LLMs can assist mathematicians in trivial tasks, such as identifying relevant lemmas and previously established theorems that might apply to the current proofs and formalizing mathematical arguments into rigorous proofs, allowing researchers to focus on more creative and conceptual tasks~\cite{song2025leancopilotlargelanguage}. 
Moreover, LLMs are also capable of \textbf{generating novel conjectures and hypotheses}~\cite{romera2024mathematical}.
By analyzing vast quantities of mathematical data and literature, LLMs can automatically identify patterns and propose new conjectures or relationships that might not be immediately apparent to human researchers.
Another significant benefit of LLMs is their potential for \textbf{lowering barriers to entry and enhancing accessibility within mathematics}. 
Besides, LLMs can act as consultants to explain core mathematical terms and fundamental intuitions, making specialized knowledge more accessible to researchers who are new to fields. 
For mathematics education, LLMs have the potential to assist teachers by providing personalized learning experiences, enhanced conceptual understanding, development of problem-solving skills, and automated assessment and feedback to students.

\textbf{Limitations of LLMs.}
Despite the potential of LLMs to assist in mathematical domains, several drawbacks and limitations must be carefully considered.
Firstly, several studies~\cite{mirzadeh2024gsm, cheng2025empowering} have discussed the limitations in \textbf{genuine logical inference} exhibited by LLMs.
Consequently, the performance of LLMs on mathematical tasks can significantly decline as the complexity of the questions increases, particularly when dealing with unseen or novel problems~\cite{agrawalexploring}. 
LLMs may also face difficulties with \textbf{abstract concepts and long-term reasoning}~\cite{li2025one, davoodi2024llms, ferrag2025reasoning}, and may raise the \textbf{risks of errors and hallucinations}. 

\textbf{Taxonomy.}
To summarize the main applications of LLMs in mathematical research, we introduce the following taxonomy: (i) Mathematical Proof Assistant; (ii) Theoretical Exploration and Pattern Recognition; and (iii) Mathematic Education.
\input{tables/mathematics}
Our taxonomy aligns with this survey~\cite{liang2024mathematics}, while we adopt a different viewpoint, emphasizing LLMs for mathematical research.
In the following sections, we will discuss each group of applications in detail. 
We list the key studies in Table~\ref{tab: mathematics}.

\subsubsection{Mathematical Proof Assistant}
The motivations for developing machine-assisted proof generation tools stem from the fact that proofs for complex mathematical problems may be extremely lengthy, involving numerous logical steps or extensive case-by-case analysis (a.k.a. proof by exhaustion)~\cite{liang2024mathematics, de1994mathematical}. 
Computer algorithms, embedded with rigorous deductive rules, could assist researchers in verifying the correctness of a proof or by automatically handling the proof-by-exhaustion steps.
For a more detailed discussion of earlier automated reasoning and machine-assisted proof approaches, we refer the reader to studies~\cite{harrison2014history, maric2015survey}.

Traditional proof assistants, i.e., MetaMath~\cite{megill2019metamath}, Isabelle/HOL~\cite{nipkow2002isabelle}, Rocq (Coq)~\cite{coq1996coq}, and Lean~\cite{moura2021lean}, aim to create mathematical proofs or review the proofs through structured logical frameworks.
However, these tools cannot generate the reasoning steps and only verify the steps provided by users.
Moreover, studying to utilize these systems may be time-consuming, as each method introduces a formal language to express mathematical statements.
With the advances of LLMs, several studies~\cite{yu2023metamath, polu2020generative} develop automated theorem-proving tools to facilitate researchers to generate reasoning steps or convert mathematical statements into formal languages.
GPT~\cite {polu2020generative}, PACT~\cite{han2021proof}, and Thor~\cite{jiang2022thor} introduce language model-based automated theorem proving tools interfacing with Metamath, Lean, and Isabelle/HOL, respectively.
Several advanced methods are further proposed to train LLMs through annotated datasets for proof assistant tasks, such as HyperTree Proof Search (HTPS)~\cite{lample2022hypertree}, Baldur~\cite{first2023proof}, COPRA~\cite{Amitayush2024language}, LeanDojo~\cite{yang2023leandojo}, AlphaProof~\cite{deepmind2024alphaproof}, and Lyra~\cite{zheng2024lyraorchestratingdualcorrection}.
To further demonstrate the LLMs' effectiveness in mathematical proof assistants, compared with conventional methods, we list the performance of several LLM-based methods and tree search methods in Table~\ref{tab: math evidence}.
According to the Table, we find that 
(i) Conventional approaches, i.e., tree search methods generally exhibit lower accuracies compared to the highest performing LLM-based methods, with hypertree proof search achieving 41.0\% and curriculum learning methods ranging between 29.2\% and 36.6\%.	
(ii) LLM-based methods show variability in performance based on the sample budget, evidenced by DeepSeekMath's performance improvement from 27.5\% with 128 samples to 52.0\% under a cumulative sampling strategy.		
(iii) Increasing the sample budget for methods such as DeepSeek-Prover (from 64 to 65,536 samples) leads to consistent accuracy gains, reaching up to 50.0\% at the highest sample count.

Beyond quantitative results, the breakthrough by Google DeepMind, i.e., AlphaEvolve~\cite{alphaevolve2025}, exemplifies its significance as a transformative moment in the intersection of artificial intelligence and mathematics.
Specifically, AlphaEvolve employs evolutionary programming and self-play, leveraging LLMs to autonomously generate and refine algorithms.
Unlike FunSearch, which searches for single functions, AlphaEvolve is capable of optimizing entire codebases, including the interactions between different functions.
However, AlphaEvolve still relies on human expertise in identifying interesting problems, defining clear evaluation metrics, and incorporating promising solutions into the iterative cycle. 
Several advancements of AlphaEvolve in the mathematical domain can be summarized as follows:
\begin{itemize}[leftmargin=*]
    \item Improving the $4
    \times 4$ matrix multiplication algorithm, reducing computation steps from 49 to 48—a record untouched since the Strassen algorithm in 1969~\cite{huss1996implementation, li2011strassen}.
    \item Advancing the \textit{hexagon packing problem} by finding better solutions for arranging 11 and 12 hexagons inside a larger hexagon, surpassing human achievements after 16 years of stagnation.
    \item Making progress on the \textit{kissing number problem}, a mathematical challenge unsolved for over 300 years~\cite{musin2008kissing}.
\end{itemize}

\input{tables/math_evidence}

\subsubsection{Theoretical Exploration and Pattern Recognition}
Several major theoretical advances often revolve around revealing deep connections among mathematical concepts that seem unrelated~\cite{liang2024mathematics}. 
A few studies leverage LLMs to collaborate with researchers for theoretical exploration.
For example, a recent study~\cite{johansson2023exploring} employs LLMs for conjecture generation within the Isabelle/HOL, and G-LED~\cite{gao2024generative} explores the potential of combining transformer models and diffusion models to forecast complex dynamical systems.
Their ability to process and understand external knowledge allows them to identify non-obvious connections and generate novel mathematical statements that can then be explored and potentially proven by mathematicians.
A prominent example in this direction is FunSearch~\cite{romera2024mathematical}, an evolutionary procedure that pairs a pre-trained LLM with a systematic evaluator to discover new constructions and heuristics for established open problems.

\subsubsection{Mathematical Education}
LLMs present a dual-edged potential for mathematics education, offering personalized learning experiences by tailoring problems and explanations to individual student needs~\cite{wang2024large,xu2024large, kumar2023math}.
LLMs can also enhance conceptual understanding by generating diverse explanations and providing step-by-step solutions. 
A recent study~\cite{kumar2023math} investigates the impact of LLMs, such as GPT-4, in learning mathematics. 
In particular, educators face concerns that students might use these tools to bypass genuine learning, but there is also hope that LLMs might serve as scalable and personalized tutors. 
The study provides empirical evidence on whether and how LLM-generated explanations affect student learning in math.
In Table~\ref{tab: math education}, we present the performance of students in mathematical problem solving, as reported in the study~\cite{kumar2023math}.	
The column Strategy refers to the method used to generate explanations for the questions, i.e.,
"Answer Only" indicates that no explanation is provided to the students, 
"Stock LLM" means the explanation is generated by an official LLM, and 
"Customized LLM" denotes that the explanation is tailored to the student.
Moreover, "See Answer First" means that students receive both the question and the correct answer before attempting to solve it, while "Try It First" denotes that students initially attempt to solve the problem on their own before receiving the correct answer for practice. 
According to Table~\ref{tab: math education}, we find that:
(i) Participants who received LLM-generated explanations performed better on subsequent test questions than those who saw only correct answers.
(ii) The greatest learning gains occurred when participants attempted problems, i.e., "Try It First", on their own before consulting the LLM explanations.
Besides facilitating students' mathematical learning, they can assist teachers in creating engaging problems and automating assessment, potentially freeing educators for more direct student interaction~\cite{hadzhikoleva2024automated}. 
However, LLMs are limited by their lack of true mathematical reasoning and conceptual understanding, often relying on pattern matching rather than genuine logic.
This can lead to reliability issues, inaccurate results, and the generation of incorrect information. 
Additionally, current LLMs struggle with visual and spatial reasoning, which are crucial in mathematics. 
Therefore, while LLMs can serve as valuable tools to augment mathematics education, they are not a replacement for human instruction and require careful oversight.
    
\input{tables/math_education}

\input{tables/math_bench.tex}

\subsubsection{Benchmarks}
Recent advances in LLMs have necessitated the development of sophisticated benchmarks to access mathematical reasoning capabilities.
In Table~\ref{tab: math bench}, we list most existing mathematics datasets, which can be categorized into four groups: competition-level, education-focused, math word problem, and mathematical reasoning. 
In the following paragraphs, we will introduce key benchmarks in detail.

\textbf{MATH.} 
The Mathematics Aptitude Test of Heuristics (MATH) dataset remains a foundational benchmark in examining the mathematic abilities of LLMs. 
MATH comprises 12,500 problems from prestigious competitions, including AMC 10, AMC 12, and AIME.
Each problem has a full step-by-step solution, which can be used for each model to generate answer derivations and explanations.
Moreover, MATH contains the following features:
(i) Five difficulty levels mirroring human problem-solving capabilities
(ii) Detailed LaTeX-formatted solutions with boxed answers
(iii) Coverage of seven mathematical domains, including combinatorics and number theory.
A smaller version of MATH, i.e., \textbf{MATH 500}, consists of 500 diverse problems from MATH, which serves as a standardized evaluation set for benchmarking and comparing the mathematical reasoning abilities of LLMs in a more manageable and reproducible way.

\textbf{AIME.} 
\textbf{A}merican \textbf{I}nvitational \textbf{M}athematics \textbf{E}xamination (AIME) is a collection of problems for a prestigious high school mathematics competition in the US and Canada.
Furthermore, each annual AIME examination comprises 15 questions, with each solution represented as a three-digit integer ranging from 000 to 999. The complexity of the problems escalates as the examination advances.
The sample in AIME includes the year, problem number, the full problem statement, and the correct answer.

\textbf{GSM8K.} 
Grade School Math 8K (GSM8K) is a dataset of 8,500 high-quality, linguistically diverse grade school math word problems, designed for evaluating and training models on mathematical problem-solving tasks. 
These problems commonly require 2 to 8 steps to solve, and solutions are provided in natural languages.

The performance of existing LLMs on the aforementioned datasets is provided in Table~\ref{tab: math benchmark}.
According to the table, we find that: 
(i) DeepSeek R1 scores 97.30 in MATH, which appears to outperform several models.
(ii) In MATH 500 benchmark, GPT-o1 is particularly strong here with \(97.90\%\), closely followed by DeepSeek R1 at approximately \(97.30\%\)
(iii) For AIME benchmarks, Grok 3 achieves the best performance in both years, i.e., 93.30\%.
(iv) Most models that display high accuracy on GSM8K. For instance, Qwen 2.5 and DeepSeek R1 are in the mid-to-hgih 90 percent range, with DeepSeek R1 at \(96.13\%\) and Qwen 2.5 at \(91.50\%\).
GPT-o1 and GPT-o3-mini also showcase robust performance with scores around \(95.58\%\) and \(95.83\%\) respectively.
The performance of LLMs in GSM8K showcases that for elementary math and reasoning, most models achieve near ceiling performance.

Overall, models, DeepSeek R1 and certain GPT variants (especially GPT-o1 and GPT-o3-mini) demonstrate \textbf{consistent performance} across various benchmarks, making them \textbf{attractive for applications} that require a broad spectrum of math problem-solving capabilities.
Although some models, such as GPT-o1, have a higher cost per token, their performance on certain benchmarks, i.e., MATH 500, might justify the expense for research or high-stakes applications. 
In contrast, DeepSeek R1 offers a very competitive performance at a much lower token cost.
The performance variation across different benchmarks, i.e., MATH vs. AIME vs. GSM8K, suggests that each dataset poses unique challenges. 
For instance, most models perform exceptionally well on GSM8K, while the AIME benchmarks reveal differentiators in advanced problem-solving capabilities.
\input{tables/math_benchmark}

\subsubsection{Discussion}

\textbf{Opportunities and Impact.} 
LLMs offer significant opportunities to enhance both mathematical research and education. 
In mathematical research, LLMs are capable of processing and synthesizing large amounts of information to facilitate researchers in literature reviews, novel conjecture generation, and even assist in theorem proving. 
Frameworks like FunSearch demonstrate the potential for LLMs to contribute to mathematical discovery by identifying hidden patterns. 
This can free up human mathematicians to focus on higher-level conceptual work and tackle problems previously deemed intractable due to computational complexity. 
The impact could be a faster pace of discovery and a broader exploration of mathematical landscapes.
As shown in Table~\ref{tab: math evidence}, LLMs demonstrate better performance than traditional proof assistants in the miniF2F dataset, showcasing the potential of LLMs as assistants in mathematical research. 

From the mathematical education perspective, LLMs can enhance personalized learning experiences by adapting to individual students' needs and providing tailored reasoning.
They can offer support for conceptual understanding by presenting information in multiple ways and aid in the development of problem-solving skills through engaging and relevant problems. 
As listed in Table~\ref{tab: math education}, students who received the LLM-generated explanations performed better than those who received only correct answers, demonstrating the potential of LLMs in personalized mathematical learning.
Automated assessment and feedback can also streamline the educational process, allowing teachers to focus on individual student support. 
Ultimately, LLMs have the potential to make mathematics more accessible and engaging for a wider range of learners.

\textbf{Challenges and Limitations.}
Despite the promising opportunities, significant challenges and limitations exist in the application of LLMs to mathematics.
A primary concern is their lack of mathematical reasoning and deep conceptual understanding. 
LLMs primarily operate through pattern matching and may struggle with multi-step reasoning, especially when faced with irrelevant information. 
Moreover, it may face hallucinations whose causes are still unknown, indicating the necessity for human verifications.
The inherent limitations in novelty and the potential for LLMs to merely reproduce existing knowledge raise questions about their ability to drive truly breakthrough discoveries without significant human guidance.

In education, these limitations translate to concerns about the reliability of LLM-generated explanations and solutions. LLMs struggle with visual and spatial reasoning, which poses a challenge for teaching, especially in geometric topics. Over-reliance on LLMs could also hinder the development of fundamental mathematical skills and critical thinking in students. 
Furthermore, ethical considerations regarding bias in training data and the potential impact on the integrity of mathematical knowledge need careful attention. 

\textbf{Future Research Directions.}
Future research should focus on addressing the limitations of LLMs in mathematics and exploring effective ways to integrate them into research and education. 
Here we summarize several key research directions:
\begin{itemize}[leftmargin=*]
    \item \textbf{Enhancing Mathematical Reasoning.} 
    Developing novel architectures and training recipes that enable LLMs to move beyond pattern matching toward genetic models. 
    This could involve incorporating symbolic computation capabilities or training on datasets designed to specifically test and improve reasoning skills.
    \item \textbf{Improving Reliability and Accuarcy.}
    Investigating methods to reduce hallucinations and errors in LLM outputs for mathematical tasks. 
    This could involve techniques like self-verification, the use of external validators like theorem provers, or reinforcement learning from human feedback focused on accuracy.
    \item \textbf{Effective Educational Tools.}
    There is a need for designing and evaluating LLM-empowered tools that effectively support mathematics learning without compromising conceptual understanding or fundamental skill development. 
    This includes exploring the utilization of LLMs in personalized tutoring, generating diverse explanations, and creating engaging problem-solving activities.
    \item \textbf{Integration Strategies Evaluation.}
    Exploring optimal strategies for incorporating LLMs into conventional educational practices to develop hybrid learning environments that capitalize on the advantages of both methodologies is crucial. 
    Additionally, it is vital to comprehend how to effectively instruct educators in the utilization of LLMs as pedagogical tools.
\end{itemize}

\textbf{Conclusion.}
The integration of LLMs into the landscape of mathematical research and education presents a transformative potential, but it is crucial to approach this integration with a balanced perspective. 
While LLMs offer exciting opportunities to augment human capabilities, accelerate discovery, and personalize learning, they are not without significant limitations, particularly in the core areas of mathematical reasoning and reliability. Traditional methods in both research and education, with their emphasis on rigor, conceptual understanding, and human intuition, remain indispensable.

We want to emphasize that \textbf{LLMs should serve as powerful tools to assist mathematicians and educators, rather than replacing them.}
By focusing on research that addresses the current limitations of LLMs and by thoughtfully integrating them into educational practices, we can harness their potential to advance mathematical knowledge and foster a deeper and more accessible understanding of mathematics for all. 
We believe the journey of integrating LLMs into mathematics is still in its early stages, and ongoing research, collaboration, and critical evaluation will be essential to navigate this evolving landscape effectively.

\subsection{Physics and Mechanical Engineering}
\subsubsection{Overview}

\noindent\textbf{5.2.1.1 Introduction to Physics}

Physics is a natural science that investigates the fundamental principles governing matter, energy, and their interactions through experimental observation and theoretical modeling~\cite{maxwell1878matter}. The focus of physics research varies from subatomic particles to cosmic structures, with the aim of establishing predictive, transparent and unified frameworks to explain everyday phenomena, such as why apples fall and why lights turn on~\cite{wigner1960unreasonable,maxwell1878matter}. As an important field of fundamental natural science, physics provides conceptual foundations and methodological tools that support other advanced scientific research and engineering projects~\cite{einstein1936physics}.
For example, the detection of gravitational waves marked one of the most significant breakthroughs in 21st-century physics~\cite{abbott2016observation}. The existence of gravitational waves was first predicted by Einstein’s theory of general relativity, but the direct detection failed for nearly a century due to their extremely weak nature~\cite{einstein1916naherungsweise}. In 2015, the LIGO interferometer in the United States successfully captured gravitational waves produced by the collision of two black holes~\cite{abbott2016observation}. This discovery not only confirmed theoretical predictions, but also established a new field, i.e. gravitational wave astronomy~\cite{sathyaprakash2009physics}, which has exerted far-reaching impacts on astrophysics, cosmology, and quantum gravity~\cite{sathyaprakash2012scientific,yunes2013gravitational}.

The discipline of physics is vast and typically organized into three core domains, each addressing a class of natural phenomena and associated methodologies~\cite{halliday2013fundamentals}:

\noindent\textbf{Fundamental Theoretical Physics.}
This domain reveals the basic laws of nature, establishing the theoretical foundation of the entire field of physics. It encompasses classical mechanics, electromagnetism, thermodynamics, statistical mechanics, quantum mechanics, and relativity~\cite{halliday2023principles}. The scope ranges from macroscopic motion (e.g., acceleration, vibration, and waves) to the behavior of subatomic particles (e.g., electron transitions, spin, self-consistent fields), including the structure of spacetime under extreme conditions such as near black holes~\cite{misner1973gravitation}.  
Researchers in this domain employ mathematical tools such as differential equations, Lagrangian and Hamiltonian mechanics, and group theory to construct theoretical models and derive predictions~\cite{goldstein2002classical}. These models not only guide experimental physics but also provide essential frameworks for engineering applications~\cite{arfken2011mathematical}.

\noindent\textbf{Physics of Matter and Interactions.}
This domain explores the microscopic structure and multi-scale interaction patterns of matter, and how these determine macroscopic material properties~\cite{kittel2018introduction}. Main subfields include condensed matter physics, atomic physics, molecular physics, nuclear physics, and particle physics~\cite{haken2006physics,griffiths2020introduction}. Key research topics cover crystal structure, electronic band theory, spin and magnetism, superconductivity, quantum phase transitions, and the classification and interaction of elementary particles~\cite{kittel2018introduction,griffiths2020introduction}.  
Common methodologies include first-principles calculations, quantum statistical modeling, and large-scale experiments involving synchrotron radiation, laser spectroscopy, or high-energy accelerators~\cite{martin2020electronic}. The findings from this domain have led to breakthroughs in semiconductor devices, materials development, quantum computing, and energy systems, driving significant technological innovation~\cite{marder2010condensed, zoller2005quantum}.

\noindent\textbf{Cosmic and Large-Scale Physics.}
This domain focuses phenomena at scales far beyond laboratory conditions and includes astrophysics, cosmology, and plasma physics~\cite{carroll2017introduction}. Astrophysics and cosmology investigate topics such as stellar evolution, galactic dynamics, gravitational wave propagation, early-universe inflation, and the nature of dark matter and dark energy~\cite{dodelson2024modern}.  
In parallel, plasma physics explores phenomena like solar wind, magnetospheric dynamics, and the behavior of ionized matter in space environments~\cite{kivelson1995introduction}. Research in this domain typically relies on astronomical observations (e.g., telescopes, gravitational wave detectors, space missions), theoretical models, and large-scale simulations, forming a “observation + simulation + theory” workflow~\cite{national2021pathways}.

Together, these three domains constitute the intellectual architecture of modern physics~\cite{heilbron2005oxford}. From particle collisions in accelerators to observations of distant galaxies, physics aims to understand the most fundamental laws of nature~\cite{greene2004fabric}. With the emergence of artificial intelligence, high-performance computing and advanced instrumentation, the physics research becomes increasingly intelligent, interdisciplinary and precise~\cite{carleo2019machine}.

\begin{figure}[!t]
    \centering
    \includegraphics[width=\linewidth]{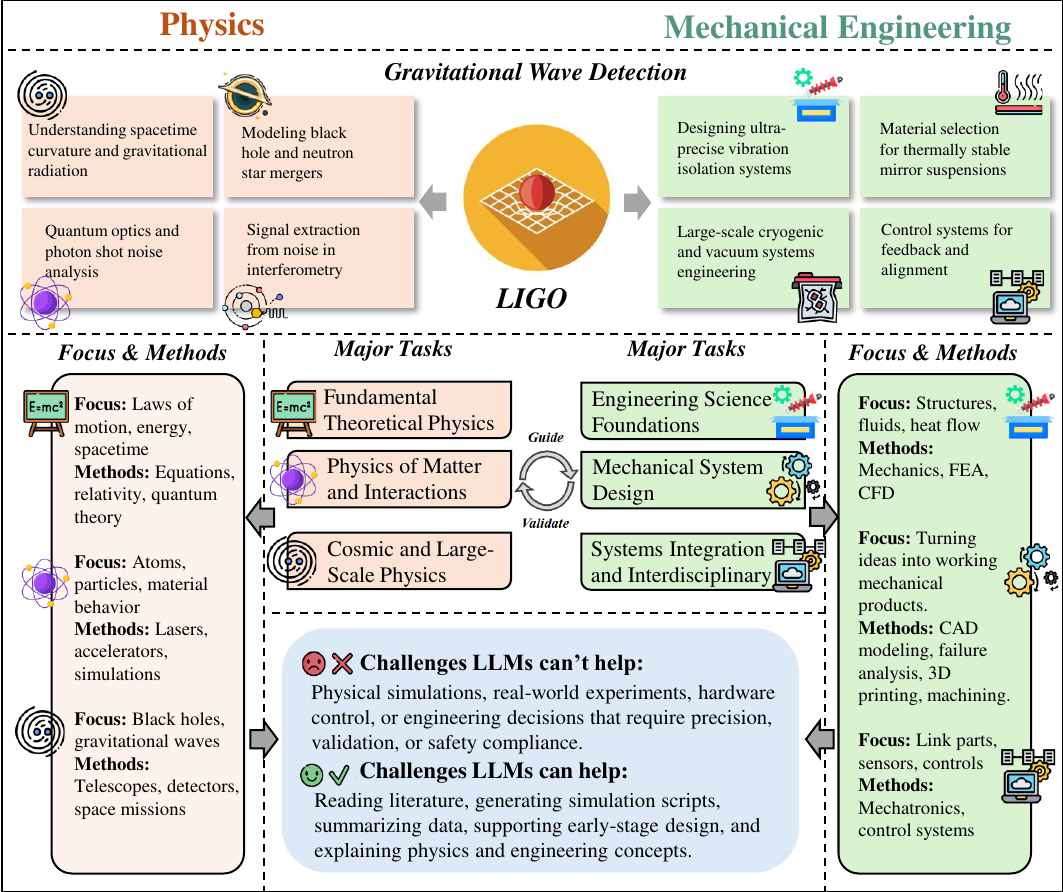}
    \caption{The relationships between major research tasks between physics and mechanical engineering.}
    \label{fig:physics_mechanical_framework}
\end{figure}

\noindent\textbf{5.2.1.2 Introduction to Mechanical Engineering}

Mechanical engineering is an applied science focusing on the design, analysis, manufacturing, and control of mechanical systems across a wide range of industries, which is driven by engineering mechanics, thermofluid sciences, materials engineering, control theory, and computational tools~\cite{avallone2006marks}. As a foundational engineering field, mechanical engineering has facilitated technological advancement in transportation, energy, robotics, aerospace, and biomedical systems~\cite{wickert2013introduction}.

From engines and turbines to robots and surgical devices, mechanical engineering continues to turn physical principles into functional products through design and manufacturing~\cite{wickert2013introduction}.
For example, the construction of the LIGO gravitational wave observatory represents not only a milestone in fundamental physics but also a triumph of mechanical engineering. LIGO's ultrahigh-vacuum interferometers required vibration isolation at nanometer precision, thermally stable mirror suspensions, and large-scale structural systems integrated with active control. These engineering achievements enabled the detection of extremely weak gravitational waves~\cite{abbott2016observation}.

The field of mechanical engineering is vast and is typically categorized into three core domains, each covering a range of methodologies and applications:

\noindent\textbf{Engineering Science Foundations.} This domain forms the analytical and physical core of mechanical engineering. It encompasses:
\textit{Mechanics}: including statics, dynamics, solid mechanics and continuum mechanics, used to model the deformation, motion, and failure of structures~\cite{craig2020mechanics}.
\textit{Thermal and fluid sciences}: covering heat conduction, convection, fluid dynamics, thermodynamics, and phase-change phenomena~\cite{moran2010fundamentals}.
\textit{Systems and control}: including system dynamics, feedback control theory, and mechatronic integration~\cite{ogata2009modern}.
Tools such as finite element analysis (FEA), computational fluid dynamics (CFD), and system modeling platforms like Simulink and Modelica are commonly used for corresponding implementations~\cite{zienkiewicz2005finite,lomax2001fundamentals}.

\noindent\textbf{Mechanical System Design and Manufacturing.} This domain aims to translate ideas into real-world products. It includes:
\textit{Mechanical design}: CAD modeling, mechanism design, stress analysis, failure prediction, and design optimization~\cite{budynas2011shigley}.
\textit{Manufacturing}: traditional subtractive methods (e.g., milling, turning), additive manufacturing (3D printing), surface finishing, and process planning~\cite{sharma2022additive}.
\textit{Smart manufacturing and Industry 4.0}: integration of sensors, data analytics, automation, and cyber-physical systems to create responsive and intelligent production environments~\cite{tyagi2024industry}.
These technologies bridge the gap between design and physical realization.

\noindent\textbf{Systems Integration and Interdisciplinary Applications.} Modern mechanical systems are often multi-functional and involve cross-disciplinary collaboration. This domain focuses on:
\textit{Robotics and mechatronics}: combining mechanics, electronics, and computing to build intelligent machines~\cite{soori2024intelligent}.
\textit{Energy and thermal systems}: engines, fuel cells, solar collectors, HVAC systems, and sustainable energy technologies~\cite{tian2013review}.
\textit{Biomedical and bioinspired systems}: development of prosthetics, surgical tools, and biomechanical simulations~\cite{gao2019biomechanical}.
\textit{Multiphysics modeling and digital twins}: simulation of interdisciplinary systems (thermal, fluidic, mechanical, electrical) and virtual prototyping~\cite{thelen2022comprehensive}.
This integration-driven domain reflects the evolution trend of mechanical engineering towards intelligent, efficient, and adaptive systems. Together, these three domains define the scope of modern mechanical engineering, 
shaping the physical world with ever-growing precision and complexity~\cite{avallone2006marks,wickert2013introduction}.

\noindent\textbf{5.2.1.3 Current Challenges}

Physics and mechanical engineering are closely interwoven disciplines that establish the foundation for understanding and shaping the material and technological world. Physics seeks to uncover the fundamental laws of nature, while mechanical engineering applies these principles to design, optimize, and control systems that power modern life. Together, they enable critical innovations across transportation, energy, manufacturing, healthcare, and space exploration. These disciplines are indispensable for solving complex challenges such as energy efficiency, automation, sustainable mobility, and precision instrumentation. Despite rapid progress in theoretical modeling, simulation, and intelligent design tools, both fields are still striving to address the complexity of nonlinear dynamics, multiphysics coupling, and real-world uncertainties in physical systems.

\noindent\textbf{Still Hard with LLMs: The Tough Problems.}
\begin{itemize}[leftmargin=10pt]

    \item \textbf{Complexity of Multiphysics Coupling and Governing Equations.}
    Physical and mechanical systems are often governed by a series of highly coupled partial differential equations (PDEs), involving nonlinear dynamics, continuum mechanics, thermodynamics, electromagnetism, and quantum interactions~\cite{zienkiewicz2005finite,anderson2002computational}. Solving such systems requires professional numerical solvers, high-fidelity discretization techniques, and physics-informed modeling assumptions. Although LLMs can retrieve relevant equations or suggest approximate forms, they are incapable of deriving physical laws, ensuring conservation principles, or performing accurate numerical simulations.
    
    \item \textbf{Simulation Accuracy and Model Calibration.}
    Accurate mechanical design and physical predictions typically rely on high-fidelity simulations such as finite element analysis (FEA), computational fluid dynamics (CFD), or multiphysics modeling~\cite{bathe2006finite,mosavi2014calibrating}. These simulations require precise input of geometry, boundary conditions, material models, and experimental validation. LLMs may assist in interpreting simulation reports or proposing modeling strategies, but they lack the resolution, numerical rigor, and feedback integration necessary to execute or validate such models.
    
    \item \textbf{Experimental Prototyping and Hardware Integration.}
    Validation through physical experiments is critical for engineering innovation, such as building prototypes, tuning actuators, installing sensors, and measuring performance under dynamic conditions~\cite{brooks1990elephants,zagal2004back}. These tasks depend on laboratory facilities, fabrication tools, and hands-on experimentation, all of which are beyond the operational scope of LLMs. Although LLMs can help generate test plans or documentation, they cannot replace real-world testing or iterative hardware development.
    
    \item \textbf{Materials and Manufacturing Constraints.}
    Real-world engineering designs must consider constraints such as thermal stress, fatigue life, manufacturability, and cost-efficiency~\cite{forster2015materials}. Addressing these challenges often relies on material testing, manufacturing standardization, and domain experience in processes such as welding, casting, and additive manufacturing. LLMs lack access to real-time physical data and material behavior and thus cannot support tradeoff decisions in design or production.
    
    \item \textbf{Ethical, Safety, and Regulatory Considerations.}
    Mechanical engineers must consider ethical impacts, user safety, and legal compliance when developing human-centered autonomous systems like biomedical devices~\cite{asme2021ethics}. Although LLMs can summarize policies or regulatory codes, it is hard for them to make decisions involving responsibility, risk evaluation, or normative judgment, which are essential for real-world systems.

\end{itemize}

\noindent\textbf{Easier with LLMs: The Parts That Move.}\\
Although current LLMs are limited in core tasks such as physical modeling and experimental validation, they have shown growing potential in assisting various supporting tasks in physics and mechanical engineering, particularly in knowledge integration, document drafting, design ideation, and educational support:

\begin{itemize}[leftmargin=10pt]

    \item \textbf{Literature Review and Standards Lookup.} 
    Both disciplines rely heavily on technical documentation such as material handbooks, design standards, experimental protocols, and scientific publications. LLMs can significantly accelerate the literature review process by extracting key information about theoretical models, experimental conditions, or engineering parameters. For instance, an engineer could use an LLM to compare different welding codes, retrieve thermal fatigue limits of materials, or summarize applications of a specific mechanical model~\cite{portenoy2020constructing,accuris2024engineering}.

    \item \textbf{Assisting with Simulation and Test Report Interpretation.} 
    In simulations such as finite element analysis (FEA), computational fluid dynamics (CFD), or structural testing, LLMs can help parse simulation logs, identify setup issues, or generate summaries of experimental findings. When integrated with domain-specific tools, LLMs may even assist in generating simulation input files, interpreting outliers in results, or recommending appropriate post-processing techniques~\cite{mudur2025feabench,ni2023mechagents}.

    \item \textbf{Supporting Conceptual Design and Parametric Exploration.} 
    During early-stage mechanical design or material selection, LLMs can suggest structural concepts, propose parameter combinations, or retrieve examples of similar engineering cases. For instance, given a prompt like “design a spring for high-temperature fatigue conditions,” the model might generate candidate materials, geometric options, and common failure modes~\cite{makatura2024large,wu2024cadvlm}.

    \item \textbf{Engineering Education and Learning Support.} 
    Education in physics and mechanical engineering involves both theoretical understanding and hands-on application. LLMs can generate step-by-step derivations, support simulation-based exercises, or simulate simple lab setups (e.g., free fall, heat conduction, beam deflection). They can also assist with terminology explanation or provide example problems to enhance interactive and self-guided learning~\cite{jiang2024beyond,abedi2023beyond}.

    \end{itemize}
    
In summary, although physical modeling, engineering intuition, and experimental testing remain essential in physics and mechanical engineering, LLMs are emerging as effective tools for information synthesis, design reasoning, documentation, and education. These disciplines may be reshaped by in-depth interactions among LLMs, simulation platforms, engineering software, and laboratory systems, paving the way from textual reasoning to intelligent system collaboration.

\begin{figure}[!t]
    \centering
    \includegraphics[width=\linewidth]{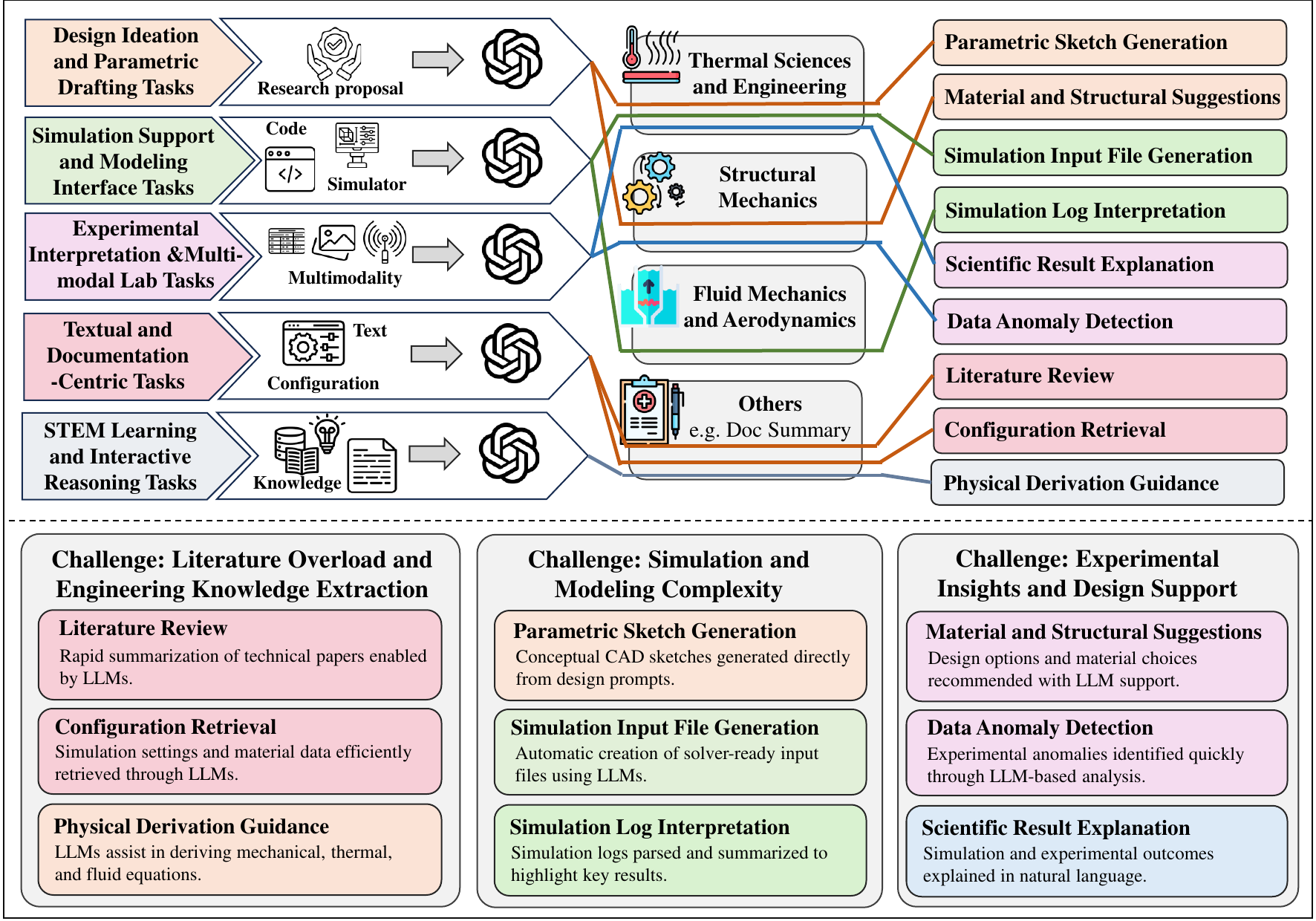}
    \caption{The pipelines of physics and mechanical engineering.}
    \label{fig:physics_mechanical_framework}
\end{figure}

\noindent\textbf{5.2.1.4 Taxonomy}

Research in physics and mechanical engineering spans from modeling fundamental laws of nature to designing and validating engineered systems. With the rapid development of LLMs, many of these tasks are being redefined through human-AI collaboration, automation, and intelligent assistance. Traditionally, physics and mechanical engineering are divided along disciplinary lines, e.g., thermodynamics, solid mechanics, control systems. However, from the perspective of LLMs, it is more productive to reorganize tasks based on their computational characteristics and data modalities.

This functional, task-driven taxonomy helps distinguish where LLMs can take on primary responsibilities, where they act in a supporting role, and where traditional numerical methods and expert reasoning remain indispensable. Based on this perspective, we propose five major categories that capture the current landscape of LLM-integrated research in physics and mechanical engineering:

\begin{itemize}[leftmargin=10pt]
\item \textbf{Textual and Documentation-Centric Tasks.}
LLMs are particularly effective in processing technical documents, engineering standards, lab reports, and scientific literature. For instance, Polverini and Gregorcic demonstrated how LLMs can support physics education by extracting and explaining key information from conceptual texts~\cite{polverini2023chatgpt}, while Harle et al. highlighted their use in organizing and generating instructional materials for engineering curricula~\cite{khan2024requirements}.

\item \textbf{Design Ideation and Parametric Drafting Tasks.}
In early-stage design and manufacturing workflows, LLMs can transform natural language prompts into CAD sketches, material recommendations, and parameter ranges. The MIT GenAI group systematically evaluated the capabilities of LLMs across the entire design-manufacture pipeline~\cite{makatura2024large}, and Wu et al. introduced CadVLM, a multimodal model that translates linguistic input into parametric CAD sketches~\cite{wu2024cadvlm}.

\item \textbf{Simulation-Support and Modeling Interface Tasks.}
Although LLMs cannot replace high-fidelity physical simulation, they can assist in generating model input files, translating specifications into solver-ready formats, and summarizing results. Ali-Dib and Menou explored the reasoning capacity of LLMs in physics modeling tasks~\cite{ali2024physics}, while Raissi et al.’s PINN framework demonstrated how language-driven architectures can help solve nonlinear partial differential equations by encoding physics into neural representations~\cite{raissi2019pinn}.

\item \textbf{Experimental Interpretation and Multimodal Lab Tasks.}
In experimental workflows, LLMs can support data summarization, anomaly detection, and textual explanation of multimodal results. Latif et al. proposed PhysicsAssistant, an LLM-powered robotic learning system capable of interpreting physics lab scenarios and offering real-time feedback to students and instructors~\cite{latif2024physicsassistant}.

\item \textbf{STEM Learning and Interactive Reasoning Tasks.}
LLMs are increasingly integrated into educational settings to guide derivations, answer conceptual questions, and simulate physical systems. Jiang et al. introduced a tutoring system that enhanced high school students’ understanding of complex physics concepts using LLMs~\cite{jiang2024beyond}, while Polverini’s work further confirmed the model’s utility in supporting structured and interactive learning~\cite{polverini2023chatgpt}.
\end{itemize}

\subsubsection{Textual and Documentation-Centric Tasks}
In physics and mechanical engineering, researchers and engineers routinely interact with large volumes of unstructured text: scientific papers, technical manuals, design specifications, test reports, and equipment logs. These documents are often dense, domain-specific, and heterogeneous in format. LLMs provide a promising tool for automating the extraction, summarization, and interpretation of this information.

One of the primary use cases is literature review and standard extraction. LLMs can parse multiple engineering reports and scientific articles to extract key findings, quantitative parameters, or references to specific standards, thereby reducing time-consuming manual review. For example, Khan et al. (2024) showed that LLMs can effectively assist in requirements engineering by identifying constraints and design goals from complex textual documents~\cite{khan2024requirements}.

Another growing application is in log interpretation and structured report analysis. In mechanical system testing and diagnostics, engineers often work with detailed experiment logs and operational narratives. Tian et al. (2024) demonstrated that LLMs can identify experimental conditions, setup parameters, and key outcomes from such semi-structured text logs, making them useful in experiment-driven engineering workflows~\cite{tian2024mechanicsllm}.

Furthermore, LLMs have been applied in sensor data documentation matching. Berenguer et al. (2024) proposed an LLM-based system that interprets natural language descriptions to retrieve relevant sensor configurations and data, effectively bridging the gap between textual requirements and structured data sources~\cite{berenguer2024sensors}.

These applications point to a broader role for LLMs as interfaces between human engineers and machine-readable engineering assets, enabling a smoother flow of information across documentation, modeling, and decision-making. Although challenges remain in domain-specific precision and context disambiguation, the utility of LLMs in handling technical documentation is becoming increasingly evident.

\subsubsection{Design Ideation and Parametric Drafting Tasks}

In physics and mechanical engineering, the early stage of design, or the idea formalization in the form of parameter-driven models, plays a critical role in shaping the final product. This process traditionally requires both deep understanding of domain knowledge and iterative exploration using CAD tools. With the emergence of LLMs, this early design workflow is being significantly transformed. LLMs can help engineers rapidly generate, interpret, and modify design concepts using natural language, thus improving both accessibility and productivity in the drafting process.

Recent studies have shown that LLMs are capable of generating design concepts from textual prompts that describe functional requirements or contextual constraints. For instance, Makatura et al. (2023) introduced a benchmark for evaluating LLMs on design-related tasks, showing that these models can generate reasonable design plans and material suggestions purely based on natural language input~\cite{makatura2024large}. This capability supports brainstorming and variant generation, especially in multidisciplinary systems where engineers must evaluate many trade-offs quickly.

Beyond concept generation, LLMs are increasingly used to support parametric drafting. This involves translating natural language into design specifications, such as dimensioned geometry, material choices, and assembly constraints. Wu et al. (2024) proposed CadVLM, a model capable of generating parametric CAD sketches from language-vision input, bridging LLMs with traditional CAD workflows~\cite{wu2024cadvlm}. Such models allow engineers to promptly iterate over different design options through language-driven instructions (e.g., “Make the slot wider by 2 mm” or “Add a fillet at the bottom edge”), greatly simplifying the communication between design intent and digital geometry.

Some systems have also incorporated LLMs directly into CAD environments, allowing interactive, prompt-based drafting and editing. Tools like SketchAssistant and AutoSketch use LLMs to assist with geometry creation and layout proposals. These interfaces reduce the learning curve for non-expert users and open up early-stage design to a broader range of collaborators. However, challenges remain in aligning generated outputs with engineering standards, ensuring the manufacturability of outputs, and maintaining traceability between design versions and decision logic.

Overall, LLMs are becoming valuable collaborators in the ideation-to-drafting pipeline of physics and mechanical engineering design. Although they can not replace domain expertise or formal simulation, they significantly accelerate exploration, reduce iteration costs, and expand accessibility to design tools.

\subsubsection{Simulation-support and Modeling Interface Tasks}

In physics and mechanical engineering, simulations play a critical role in modeling complex systems, validating designs, and predicting behavior. Traditionally, configuring and running simulations requires domain expertise, specialized tools, and manual scripting. The integration of LLMs has significantly improved efficiency and accessibility.

LLMs can translate natural language descriptions of physical setups into structured simulation code or configuration files. For example, \textit{FEABench} evaluates the ability of LLMs to solve finite element analysis (FEA) tasks from text-based prompts and generate executable multiphysics simulations, showing encouraging performance across benchmark problems~\cite{mudur2025feabench}. Similarly, \textit{MechAgents} demonstrates how LLMs work as collaborative agents to solve classical mechanics problems (e.g., elasticity, deformation) through iterative planning, coding, and error correction~\cite{ni2023mechagents}.

Beyond code generation, LLMs are being deployed as intelligent simulation interfaces. \textit{LangSim}, developed by the Max Planck Institute, connects LLMs to atomistic simulation software, enabling users to query and configure simulations via natural language~\cite{langsim2024}. Such systems lower the barrier for non-experts to engage in simulation workflows, automate routine tasks, and reduce friction in setting up complex models.

Moreover, LLMs can help interpret simulation results, summarize outcome trends, and generate human-readable reports that connect raw numerical output with engineering reasoning. This interpretability is especially valuable in multi-physics scenarios where simulation logs and visualizations are often overwhelming.

Overall, these advances are promising, but limitations still remain in LLMs' ability to ensure physical correctness, handle multiphysics coupling, and reason over temporal or boundary conditions. Nonetheless, their role as modeling assistants is becoming increasingly practical in early prototyping and parametric studies.

\subsubsection{Experimental Interpretation and Multimodal Lab Tasks}

In physics and mechanical engineering, laboratory experiments often generate complex datasets containing textual logs, numerical measurements, images, and sensor outputs. Interpreting these multi-modal datasets requires significant expertise and time. The advent of LLMs offers promising solutions to streamline this process by enabling automated analysis and interpretation of diverse data types.

LLMs can assist in translating experimental procedures and observations into structured formats, facilitating analysis and replication. For instance, integrating LLMs with graph neural networks has been shown to enhance the prediction accuracy of material properties by effectively combining textual and structural data \cite{turn0search1}. This multimodal approach allows for a more comprehensive understanding of experimental outcomes.

Moreover, LLMs have demonstrated capabilities in interpreting complex scientific data, such as decoding the meanings of eigenvalues, eigenstates, or wavefunctions in quantum mechanics, providing plain-language explanations that bridge the gap between complex mathematical concepts and intuitive understanding \cite{turn0search2}. Such applications highlight the potential of LLMs in making intricate experimental data more accessible.

Additionally, frameworks like GenSim2 utilize multi-modal and reasoning LLMs to generate extensive training data for robotic tasks by processing and producing text, images, and other media, thereby enhancing the training efficiency for robots in performing complex tasks \cite{turn0search3}.

Despite these progress, challenges remain in ensuring the accuracy and reliability of LLM-generated interpretations, especially when dealing with noisy or incomplete data. Recent research focuses on improving the robustness of LLMs in handling diverse and complex experimental datasets.

\subsubsection{STEM Learning and Interactive Reasoning Tasks}

LLMs are increasingly being integrated into STEM education to enhance learning experiences and support interactive reasoning tasks. Their ability to process and generate human-like text allows for more engaging and personalized educational tools.

LLMs can simulate teacher-student interactions, providing real-time feedback and explanations that adapt to individual learning needs. This capability has been utilized to improve teaching plans and foster deeper understanding in subjects like mathematics and physics \cite{turn0search3}. Additionally, LLMs have been employed to interpret and grade student responses, offering partial credit and constructive feedback, which aids in the learning process \cite{chen2025grading}.

Interactive learning environments powered by LLMs, such as AI-driven tutoring systems, have shown promise in facilitating inquiry-based learning and promoting critical thinking skills. These systems can guide students through problem-solving processes, encouraging them to explore concepts and develop reasoning abilities \cite{turn0search2}.

Despite these advancements, challenges remain in ensuring the accuracy and reliability of LLM-generated content. Ongoing research focuses on improving the alignment of LLM outputs with educational objectives and integrating multi-modal data to support diverse learning styles.

\begin{table}[!t]
    \centering
    \small
    \caption{Physics and Mechanical Engineering Tasks and Benchmarks}
    \label{tab:physics_bench}
    \resizebox{0.95\linewidth}{!}{
    \begin{tabular}{@{}p{5cm} p{4cm} p{9cm}@{}}
        \toprule
        \textbf{Type of Task} & \textbf{Benchmarks} & \textbf{Introduction} \\
        \midrule

        \multirow{4}{*}{\shortstack[l]{CAD and Geometric Modeling}} 
        & ABC Dataset~\cite{koch2019abc} \newline DeepCAD~\cite{qian2021deepcad} \newline Fusion 360 Gallery~\cite{willis2021fusion} \newline CADBench~\cite{du2024blenderllm}
        & The ABC Dataset, DeepCAD, and Fusion 360 Gallery together provide a comprehensive foundation for studying geometry-aware language and generative models. While ABC emphasizes clean, B-Rep-based CAD structures suitable for geometric deep learning, DeepCAD introduces parameterized sketches tailored for inverse modeling tasks. Fusion 360 Gallery complements these with real-world user-generated modeling histories, enabling research on sequential CAD reasoning and practical design workflows. CADBench further supports instruction-to-script evaluation by providing synthetic and real-world prompts paired with CAD programs. It serves as a high-resolution benchmark for measuring attribute accuracy, spatial correctness, and syntactic validity in code-based CAD generation. \\
        \midrule

        \multirow{1}{*}{Finite Element Analysis (FEA)} 
        & FEABench~\cite{mudur2025feabench}
        & FEABench is a purpose-built benchmark that targets the simulation domain, offering structured prompts and tasks for evaluating LLM performance in generating and understanding FEA input files. It serves as a critical testbed for bridging the gap between symbolic physical language and numerical simulation. \\
        \midrule

        \multirow{1}{*}{\shortstack[l]{CFD and Fluid Simulation}} 
        & OpenFOAM Cases~\cite{openfoam} 
        & The OpenFOAM example case library provides a curated set of fluid dynamics simulation setups, widely used for training models to understand solver configuration, mesh generation, and boundary condition specifications in CFD contexts. \\
        \midrule

        \multirow{1}{*}{\shortstack[l]{Material Property Retrieval}}
        & MatWeb~\cite{matweb} 
        & MatWeb is a widely-used material database containing thermomechanical and electrical properties of thousands of substances. It plays an essential role in supporting downstream simulation tasks such as material selection, constitutive modeling, and multi-physics simulation setup. \\
        \midrule

        \multirow{2}{*}{\shortstack[l]{Physics Modeling and PDE Learning}} 
        & PDEBench~\cite{hoellein2022pdebench} \newline PHYBench~\cite{qiu2025phybench}
        & PDEBench and PHYBench collectively advance the evaluation of LLMs in physical reasoning and numerical modeling. PDEBench focuses on classical PDEs like heat transfer, diffusion, and fluid flow in the context of scientific machine learning, while PHYBench introduces a broader spectrum of perception and reasoning tasks grounded in physical principles. Together, they support benchmarking across symbolic reasoning, equation prediction, and simulation-aware generation. \\
        \midrule

        \multirow{1}{*}{\shortstack[l]{Fault Diagnosis and Health Monitoring}}
        & NASA C-MAPSS~\cite{cmapss} 
        & NASA C-MAPSS provides real-world time-series degradation data from turbofan engines, serving as a benchmark for predictive maintenance, anomaly detection, and reliability modeling in aerospace and mechanical systems. \\
        \bottomrule
    \end{tabular}
    }
\end{table}

\subsubsection{Benchmarks}

In physics and mechanical engineering, tasks such as computer-aided design (CAD), finite element analysis (FEA), and computational fluid dynamics (CFD) are characterized by strong physical constraints, structured representations, and deep reliance on geometry or numerical solvers. The development of benchmarks to support LLMs in these domains is still in early stages. Although recent datasets have enabled initial exploration of LLMs in these fields, they present multiple challenges in scale, accessibility, and alignment with language-based modeling.

In the CAD domain, several large-scale datasets have been developed to support geometric learning and generative modeling. For example, the ABC Dataset~\cite{koch2019abc} provides over one million clean B-Rep (Boundary Representation) models, DeepCAD~\cite{qian2021deepcad} offers parameterized sketches for inverse modeling, and the Fusion 360 Gallery~\cite{willis2021fusion} includes real-world design sequences from professional and amateur CAD users. However, most of these datasets represent geometry using numeric or parametric formats that lack symbolic or linguistic structure. Specifically, B-Rep trees and STEP files are low-level and require domain-specific parsers, making them difficult for LLMs to interpret or generate in a meaningful way.

Although some recent efforts have attempted to represent CAD workflows through code-based formats such as FreeCAD Python scripts or Onshape feature code, these datasets are often small, sparse in supervision, and highly sensitive to syntactic or logical errors. Moreover, generating coherent and executable CAD programs remains a significant challenge due to the limited spatial reasoning capacity of current LLMs.

On the other hand, recent advances demonstrate that specialized instruction-to-code datasets and self-improving training pipelines can significantly improve LLMs' performance in CAD settings. For instance, BlenderLLM~\cite{du2024blenderllm} is trained on a curated dataset of instruction–Blender script pairs and further refined through self-augmentation. As shown in Table~\ref{tab:cadbench-blenderllm}, it achieves state-of-the-art results on the CADBench benchmark, outperforming models like GPT-4-Turbo and Claude-3.5-Sonnet across spatial, attribute, and instruction metrics, while maintaining a low syntax error rate. This indicates that domain-adapted LLMs, when paired with well-structured code-generation benchmarks, can overcome many of the geometric and syntactic limitations faced by general-purpose models.

\input{tables/physics_cadbench}

To address these issues, several strategies can be explored. One direction involves decomposing full modeling workflows into modular sub-tasks, such as sketch creation, constraint placement, extrusion operations, and feature sequencing. This allows the LLM to focus on smaller, interpretable segments of the modeling pipeline. Another direction is to reframe CAD problems into geometric reasoning tasks. For instance, by translating design problems into 2D or 3D visual logic similar to those found in geometry exams. Prior studies have shown that LLMs such as GPT-4 perform surprisingly well on geometric puzzles when the problem is represented symbolically or visually. Furthermore, retrieval-augmented generation (RAG) can be employed to provide contextual examples from past designs or sketches, thus improving generation quality through analogy-based learning. Overall, bridging the gap between high-dimensional geometric information and language representation remains a central challenge in CAD-focused LLM research.

Similarly, simulation-based tasks in FEA and CFD also require structured input generation, including mesh topology, material properties, solver settings, and boundary conditions. These tasks often involve producing complete simulation decks compatible with engines such as CalculiX or OpenFOAM, followed by interpreting complex field outputs such as stress distributions or velocity gradients.

Benchmarks such as FEABench~\cite{mudur2025feabench} and curated OpenFOAM case libraries~\cite{openfoam} provide valuable baselines for evaluating the simulation-awareness of LLMs. However, large-scale paired datasets with natural language descriptions, simulation input files and corresponding numerical results remains limited, posing a bottleneck for supervised fine-tuning and instruction-based evaluation.

To address this gap, FEABench introduces structured tasks that assess LLMs’ ability to extract simulation-relevant information. Table~\ref{tab:feabench-modelspecs} presents the performance of various LLMs across multiple physics-aware metrics, including interface factuality, feature recall, and dimensional consistency. Models like Claude 3.5 Sonnet and GPT-4o demonstrate strong results in retrieving factual and geometric descriptors, particularly in interface and feature extraction. However, all models show relatively low performance in recalling physical properties and structured feature attributes, reflecting ongoing challenges in capturing physical relationships from text. These results suggest that while LLMs can reliably recover high-level simulation inputs, deeper understanding of numerical structure and physical laws remains an open research problem.

\input{tables/physics_FEAbench}

A promising solution is to integrate LLMs with external numerical solvers in a simulator-in-the-loop framework. In this approach, an LLM is tasked with generating a complete simulation setup given a natural language prompt or design goal. The generated setup is then executed by a physics-based solver to produce ground-truth outputs. The input-output pairs, along with the original language prompt, can be stored as a triplet dataset and reused for supervised training. This method enables semi-automated dataset construction at scale, facilitates error correction via feedback from the simulator, and promotes the development of LLM agents that can reason across symbolic and physical domains. Additionally, by iterating through prompt refinement and result validation, such frameworks could enable reinforcement learning with human or physical feedback for high-fidelity simulation tasks.

Together, these benchmarks and emerging methodologies form the foundation of a growing research area at the intersection of language modeling, geometry, and physics. As more domain-specific tools and datasets are adapted for LLM-compatible formats, we expect substantial progress in generative reasoning, simulation co-pilots, and data-driven modeling for engineering systems.

\subsubsection{Discussion}

\noindent\textbf{Opportunities and Impact.}
LLMs are beginning to reshape workflows in physics and mechanical engineering, particularly in tasks such as CAD modeling, finite element analysis (FEA), material selection, simulation setup, and result interpretation. As demonstrated by models like CadVLM~\cite{wu2024cadvlm} which translates textual input into parametric sketches, FEABench~\cite{mudur2025feabench} which evaluates LLMs on FEA input generation, and LangSim~\cite{langsim2024} which enables natural language interaction with atomistic simulation tools, LLMs are emerging as intelligent intermediaries between domain experts and computational tools.

By converting natural language into structured engineering commands, LLMs greatly simplify early-stage design, parameter exploration, technical documentation, and preliminary simulation configuration. Through code generation, auto-completion, document retrieval, and example-based prompting, LLMs are becoming integral assistants in modern engineering workflows. As multimodal and multi-agent systems (e.g., MechAgents~\cite{ni2023mechagents}) become more common, LLMs are playing a key role in the next generation of “design–simulate–validate” engineering pipelines.

\vspace{0.5em}
\noindent\textbf{Challenges and Limitations.}
Despite these promising applications, multiple challenges persist. First, physical modeling tasks such as FEA and CFD involve highly coupled, nonlinear partial differential equations (PDEs) that require domain-specific inductive biases, numerical stability, and conservation principles, which current LLMs usually fail to solve.

Second, existing datasets in engineering domains present significant structural barriers. Most CAD datasets (e.g., B-Rep, STEP) are stored in numeric or parametric formats with minimal symbolic representation, making them difficult for LLMs to understand or generate. Code-based CAD datasets are more interpretable by LLMs, but they are often limited in size, brittle in syntax, and sensitive to logical correctness.

Moreover, LLMs struggle with tasks involving unit consistency, physical constraint enforcement, and boundary condition reasoning. In real-world engineering, even small errors in design parameters or simulation configurations can lead to system failure, safety risks, or structural inefficiencies. This makes it difficult to rely on LLMs for mission-critical design tasks without rigorous validation.

\vspace{0.5em}
\noindent\textbf{Research Directions.}
To further improve the effectiveness of LLMs in physics and mechanical engineering, several research directions are particularly promising:

\begin{itemize}[leftmargin=10pt]
    \item \textbf{Simulation-Augmented Dataset Generation.} Integrating LLMs with numerical solvers in a simulator-in-the-loop framework allows the generation of (language-input, simulation, output) triplets at scale. This enables supervised training, fine-tuning, and RLHF strategies grounded in physically valid feedback.
    
    \item \textbf{Task Decomposition and Geometric Reformulation.} Decomposing CAD workflows into modular sub-tasks (e.g., sketching, constraints, extrusion) and reformulating modeling problems as geometric reasoning tasks can align better with LLM capabilities and improve interpretability.
    
    \item \textbf{Multimodal and Multi-agent Integration.} Developing LLM systems that can call CAD tools, solvers, and databases autonomously will allow LLMs to reason, plan, and act across tools in complex design and simulation pipelines.
    
    \item \textbf{Standardized Benchmarks and Evaluation.} Creating large-scale, task-diverse, and format-unified benchmark datasets (e.g., combining natural language prompts, simulation files, and result summaries) will accelerate model evaluation and fair comparison in this field.
    
    \item \textbf{Physics Validation and Safety Assurance.} Embedding physical rule checkers and verification mechanisms into generation loops can help enforce unit consistency, structural validity, and simulation compatibility, ensuring that outputs are not just syntactically correct but physically plausible.
\end{itemize}

\vspace{0.5em}
\noindent\textbf{Conclusion.}
LLMs are becoming increasingly valuable assistants in physics and mechanical engineering, especially in peripheral tasks such as documentation, concept generation, parametric modeling, and simulation support. However, to integrate LLMs into real-world workflows, future systems must integrate LLMs with symbolic reasoning, geometric logic, physics-based solvers, and expert feedback. This synergy will enable the transition from language-based assistance to trustworthy, intelligent co-creation in complex engineering design and modeling workflows.

\subsection{Chemistry and Chemical Engineering}
\subsubsection{Overview}

\noindent Chemistry is the scientific discipline dedicated to understanding the properties, composition, structure, and behavior of matter. As a branch of the physical sciences, it focuses on the study of chemical elements and the compounds formed from atoms, molecules, and ions—their interactions, transformations, and the principles governing these processes \cite{wikipedia_chemistry, britannica_chemistry}. \textbf{Put simply, chemistry seeks to explain how matter behaves and how it changes} \cite{acs2023}.  We must acknowledge that the field of chemistry is vast and encompasses a variety of branches. Given the particularly rich application scenarios in areas such as organic chemistry, life sciences—especially in relation to LLM-related work—we will discuss these branches in the following chapter to provide a detailed introduction to works closely related to biology and life sciences.

In the field of chemistry, there are numerous sub-tasks, and many scientists have made significant contributions and achieved groundbreaking results over the past few hundred years. \textbf{Before diving into LLM-related topics, we would like to provide an overview of major tasks and traditional methods in chemistry research}. By integrating information from official websites and literature across various branches of the field \cite{carruthers2004modern, harvey2000modern, christian2013analytical, levine2009quantum, jensen2017introduction, patrick2023introduction, voet2010biochemistry, manahan2022environmental, ali2017environmental, corey1991logic}, we have summarized the research tasks in the domain of chemistry as follows:

\noindent\textbf{Analysis and Characterization.} This task involves identifying the substances present in a sample (qualitative analysis) and determining the quantity of each substance (quantitative analysis) \cite{vedantu_chemical_analysis}. In this section we emphasize experimental measurement and detection methods aimed at identifying which substances are present, as well as determining their composition, structure, and morphology; we do not here focus on how their properties change under varying conditions nor on prediction or modeling of those properties. It also includes elucidating the structure and properties of these substances at a molecular level \cite{rroij_chemical_analysis}. Traditional methods for analysis and characterization include techniques such as observing physical properties (color, odor, melting point), performing specific chemical tests to identify certain substances (like the iodine test for starch or flame tests for metals), and classical quantitative analysis using precipitation, extraction, and distillation \cite{britannica_chemical_analysis}. Modern research in this area heavily relies on sophisticated instruments. Spectroscopy, which studies how matter interacts with light, can provide significant insights into a molecule’s structure and composition \cite{rroij_chemical_analysis}. Chromatography is employed to separate complex mixtures into their individual components for analysis \cite{rroij_chemical_analysis}. Mass spectrometry (MS) is a powerful technique that can identify and quantify substances by measuring their mass-to-charge ratio with very high sensitivity and specificity \cite{rroij_chemical_analysis, duttacomprehensive}.

\noindent\textbf{Research on Properties.} Research on properties in chemistry refers to the systematic exploration and analysis of the physical and chemical characteristics of substances, with the main objective being to reveal the behavior and reaction characteristics of substances under different conditions \cite{atkins2013elements,brown2002chemistry}. We take “Research on Properties” to include both experimental determination and prediction or modeling of physical and chemical properties, with a primary interest in how those properties behave or change under different conditions. Traditionally, researchers have employed experimental methods to determine these properties. For thermodynamic properties, calorimetry is a key technique used to measure heat flow during physical and chemical processes \cite{robinson1997experimental,gill2010differential}. Equilibrium methods, such as measuring vapor pressure, can assist in determining energy changes during phase transitions \cite{robinson1997experimental}. For kinetic properties, traditional methods involve monitoring the changes in concentration of reactants or products over time \cite{zheng2015analytical}.

\textbf{Reaction Mechanisms. }The primary objective of studying reaction mechanisms in chemistry is to reveal the specific processes and steps involved in chemical reactions, including the microscopic mechanisms by which reactants are converted into products. This research field focuses on the formation of various intermediates during the reaction, the reaction pathways, rate-determining steps, and their corresponding energy changes \cite{atkins2013elements,silbey2022physical}. Traditional methods for investigating reaction mechanisms include kinetic studies, where the rate of a reaction is measured under different conditions to understand its progression \cite{smith2000reaction,britannica_reaction_mechanism,kraka2010computational}. Isotopic labeling involves using reactants with specific isotopes to trace the movement of atoms during the reaction \cite{britannica_reaction_mechanism}. Stereochemical analysis examines the spatial arrangement of atoms in reactants and products, providing insights into the reaction pathway \cite{britannica_reaction_mechanism}. Identifying the intermediate products formed during the reaction is also a crucial aspect of this research.

\noindent\textbf{Chemical Synthesis. } Chemical Synthesis refers to actually producing molecules in the laboratory or pilot‐scale. The synthesis of natural products is an important task in chemistry, aimed at using chemical methods to synthesize complex organic molecules found in nature \cite{butler2004role}. The realization of such synthesis in practice relies on several traditional experimental methods. Plant extraction separates compounds from plants using techniques like solvent extraction, cold pressing, or distillation, yielding various active ingredients. Fermentation technology utilizes microorganisms to produce natural products, commonly for antibiotics and bioactive substances \cite{zhou2014novel}. Organic synthesis constructs chemical structures through multi‐step synthesis and the introduction of functional groups \cite{gunstone2012fatty}. Lastly, semi‐synthetic methods modify simple precursors to create more complex natural compounds or their derivatives \cite{ikan1991natural}.

\textbf{Molecule Generation. } Molecule Generation involves computational chemistry and molecular modeling techniques to predict, optimize, or generate new molecular structures with desired functions or properties \cite{alkhzem2022design,o2011computational}. It includes computer‐aided design, virtual screening, property prediction, structure optimization, and theoretical modelling of molecules, etc. \cite{alkhzem2022design,o2011computational}. Molecular synthesis and design encompass both experimental synthesis \cite{o2011computational} and computer-aided design \cite{hassan2016computer}.

\textbf{Applied Chemistry. } Applied Chemistry refers to the branch of chemistry that focuses on practical applications in various fields such as industry, medicine, and environmental science. It involves using chemical principles to solve real-world problems and improve processes, including material chemistry and drug chemistry \cite{zhou2017drug, barbero2010introduction, baselt2014encyclopedia, AppliedChemistry}. Traditionally, several key methods are relied upon, including structure-activity relationship (SAR) studies, computer-aided drug design \cite{morris2009autodock4}, high-throughput screening \cite{berengut2006statistics}, and synthetic chemistry \cite{insuasty2020synthesis}.

\noindent\textbf{5.3.1.2 Introduction to Chemical Engineering}

Chemical engineering is an engineering field that deals with the study of the operation and design of chemical plants, as well as methods of improving production. Chemical engineers develop economical commercial processes to convert raw materials into useful products. Chemical engineering utilizes principles of chemistry, physics, mathematics, biology, and economics to efficiently use, produce, design, transport, and transform energy and materials \cite{wikipedia_chemical_engineering}. According to the Oxford Dictionary, \textbf{chemical engineering is a branch of engineering concerned with the application of chemistry to industrial processes}, particularly involving the design, operation, and maintenance of equipment used to carry out chemical processes on an industrial scale \cite{dictionary1989oxford}. In summary, it serves as the bridge that applies chemical achievements to industry.

Similar to chemistry, chemical engineering encompasses multiple fields, including not only chemistry, but also mathematics, physics, and economics. Through a comprehensive review of previous research \cite{towler2021chemical,westmoreland2014opportunities,ulrich1984guide,ottino2011chemical,sinnott2014chemical}, we have categorized the tasks in chemical engineering into the following types.

\textbf{Chemical Process Engineering. }Chemical process engineering includes chemical process design, improvement, control, and automation. Chemical process design focuses on the design of reactors, separation units, and heat exchange equipment to achieve efficient material conversion and energy utilization \cite{ulrich1984guide,smith2005chemical}, typically employing computer-aided design software and process simulation tools \cite{haydary2019chemical,west2008assessment}. Chemical process improvement involves the systematic analysis and optimization of existing chemical processes to enhance production efficiency, reduce resource consumption, and minimize environmental impact \cite{smith2005chemical}. It primarily relies on quality management tools \cite{yadav2016lean} and process simulation software \cite{al2022aspen}. Process control and automation aim to monitor and regulate chemical processes through control systems to ensure stable operation under set conditions, typically based on proportional–integral–derivative (PID) control systems \cite{mohindru2024review}, combined with advanced control technologies such as Model Predictive Control \cite{kumar2012model} to optimize processes. Distributed control systems and programmable logic controllers are also commonly used automation systems that can monitor and adjust process variables in real-time \cite{christofides2001control,liu2009distributed}.
    
\textbf{Equipment Design and Engineering. }Equipment design and engineering focus on the design, selection, and maintenance of chemical engineering equipment to ensure its efficient and safe operation within specific processes. The reliability and functionality of the equipment directly impact overall efficiency and safety \cite{couper2005chemical}. Equipment design is typically carried out in accordance with industry standards and regulations, such as American Society of Mechanical Engineers (ASME) and American Petroleum Institute (API). Engineers use computer‑aided design (CAD) software for detailed design and simulation \cite{parra20183d,fang2021discussion}. Additionally, strength analysis and fluid dynamics simulation are critical components, generally relying on computational fluid dynamics software to ensure the safety and efficiency of equipment under various operating conditions \cite{couper2005chemical}.
    
\textbf{Sustainability and Environmental Engineering. }Sustainability and environmental engineering focus on the impact of chemical processes on the environment and are dedicated to developing green chemical technologies to reduce pollution and resource consumption. This field emphasizes the importance of life cycle assessment and environmental impact assessment in achieving sustainability goals \cite{mihelcic2021environmental}.
    
\textbf{Scale-up and Technology Transfer.}The task of translating chemical achievements into practical applications in chemical engineering involves bridging the gap between laboratory discoveries and industrial-scale implementation, ensuring that innovative chemical processes and materials are effectively integrated into real-world production systems to meet societal and industrial demands \cite{siirola1996industrial}. Traditionally, the application of chemical achievements employs methods such as pilot scale testing to validate the feasibility and stability of the technology \cite{siirola1996industrial}, and process simulation and optimization (e.g., using tools like Aspen Plus and CHEMCAD) to model and optimize process flows, thereby reducing costs and improving efficiency. Simultaneously, factors such as economic viability, supply chain and market dynamics, and safety and environmental compliance are also evaluated and optimized \cite{towler2021chemical,ulrich1984guide}.

From the definition, we can see that there is a strong logical relationship between the fields of chemical engineering and chemistry at the macroscopic level. The main battleground of chemical science is in the laboratory, while the main battleground of chemical engineering is in the factory. Chemical engineering aims to translate processes developed in the lab into practical applications for the commercial production of products, and then work to maintain and improve those processes \cite{wikipedia_chemical_engineering,acsChemicalEngineering,wikipedia_chemistry,britannica_chemistry}.

\begin{figure}
\centering
\includegraphics[width=0.99\linewidth]{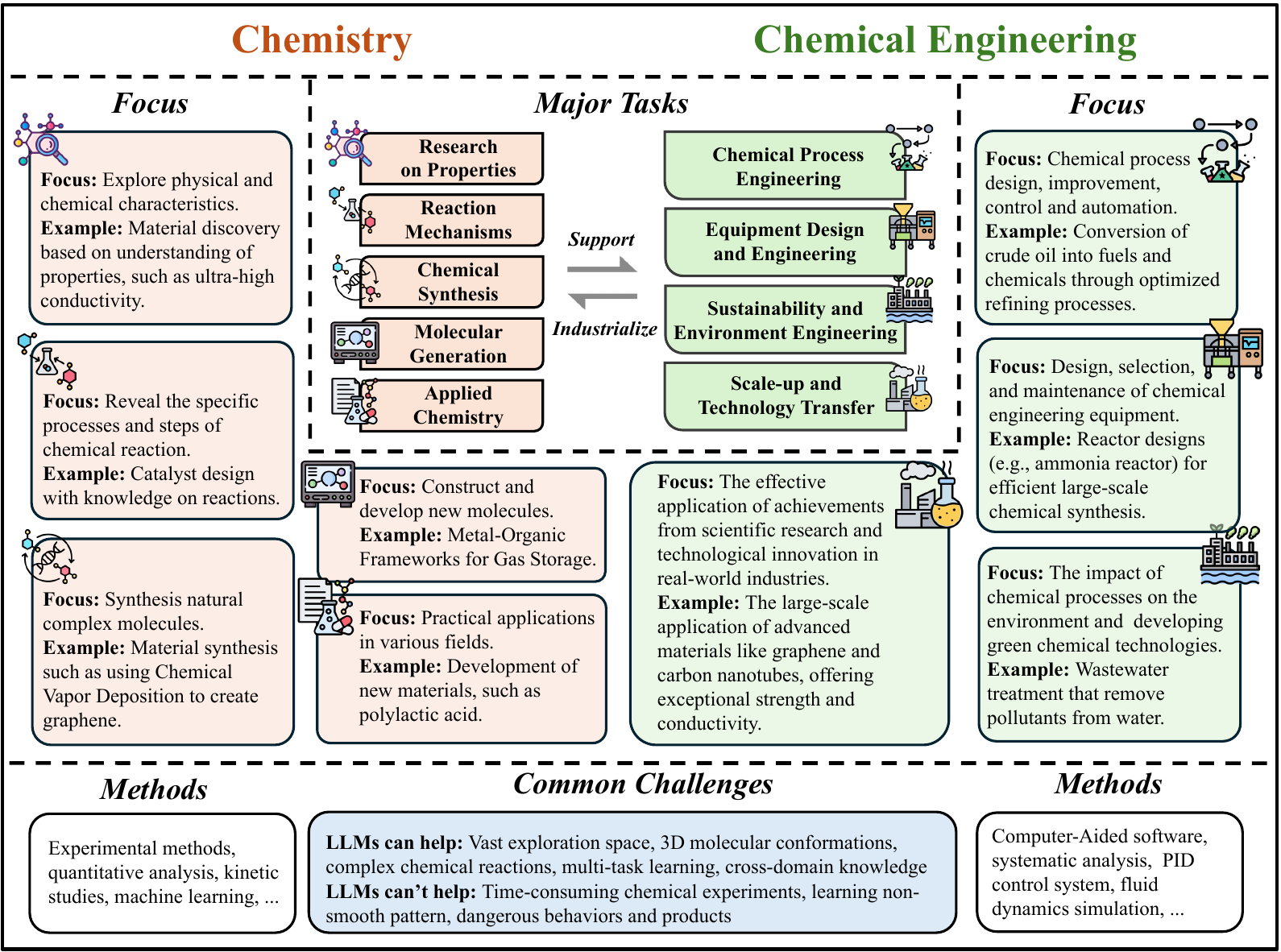}
\caption{The relationships between major research tasks in chemistry and chemical engineering.}
\label{fig:chem_diagram}
\vspace{-10pt}
\end{figure}

At the microscopic level, chemistry and chemical engineering share many common technologies, such as CAD and computational simulation. Moreover, there are varying degrees of connections between the different sub-tasks within these two fields. We have summarized the relationships among them in the form of a diagram in Figure-\ref{fig:chem_diagram}.

\noindent\textbf{5.3.1.3 Contribution of Chemistry and Chemical Engineering}

It is not difficult to imagine that chemistry, as a fundamental science, has profoundly impacted various aspects of human society, with its contributions evident in public health, materials innovation, environmental protection, and energy transition. Firstly, the contributions of chemistry to public health are significant. Through the synthesis and development of new pharmaceuticals, chemists have greatly improved human health \cite{gaynes2017discovery,gerber2008targeted}. For instance, the discovery of penicillin not only marked the beginning of the antibiotic era but also reduced the mortality risk associated with bacterial infections \cite{gaynes2017discovery,ligon2004penicillin}. In recent years, the development of targeted therapies \cite{gerber2008targeted,zhou2022targeted}, such as drugs aimed at specific cancers, relies on a chemical understanding of the internal mechanisms of tumor cells, thereby significantly enhancing patient survival rates. Secondly, chemistry has a revolutionary impact on materials innovation. Through the development of polymers \cite{seiler2006hyperbranched}, alloys \cite{starke1996application,niinomi2012development}, and nanomaterials \cite{barreto2011nanomaterials,kolahalam2019review}, chemists have not only enhanced material properties but also advanced technological progress. For example, the application of modern lightweight and high-strength composite materials has enabled greater energy efficiency and safety in the aerospace and automotive industries \cite{seiler2006hyperbranched}. Moreover, the emergence of graphene and other nanomaterials has opened new possibilities for the development of electronic products \cite{barreto2011nanomaterials,kolahalam2019review}.

In the realm of environmental protection, the contributions of chemistry cannot be overlooked. By developing efficient catalysts and clean technologies, chemists play a crucial role in reducing industrial emissions and tackling water pollution \cite{ong2010review}. For example, selective catalytic reduction reactions effectively convert harmful gases emitted by vehicles, significantly improving urban air quality \cite{lajili2022converting,makhanov2015new}. Furthermore, the role of chemistry in energy transition is becoming increasingly important \cite{centi2020smart,armaroli2016solar}. The development of renewable energy storage and conversion is fundamentally supported by chemical technologies \cite{dincer2000renewable}. For instance, the research and development of lithium-ion batteries \cite{kim2019lithium} and hydrogen fuel cells \cite{singla2021hydrogen} depend on the optimization of chemical reactions and material innovations, making the use of clean energy feasible.

\noindent\textbf{5.3.1.4 Challenges in the Era of LLMs}

Despite the significant achievements in the fields of chemical science and chemical engineering, there remain unresolved challenges in these areas. The emergence of LLMs presents an opportunity to address these issues. We must acknowledge that, unfortunately, LLMs are not omnipotent; they cannot solve all the challenges within this field. However, for certain tasks, LLMs hold promise in assisting chemists in overcoming these challenges. We have listed the following difficulties that LLMs cannot solve:

\textbf{The Irreplaceability of Time-consuming Chemical Experiments.} LLMs-generated outcomes in chemical research still require experimental validation. Assessing the true utility of these generated molecules, such as evaluating their novelty in real-world applications, can be a time-consuming undertaking \cite{guo2023can}. While LLMs have their advantages in data processing and information retrieval, solely relying on the results generated by the model may not accurately reflect the actual experimental conditions. Moreover, LLMs are trained on existing data and literature; if a specific field lacks sufficient data support, the outputs of the model may be inaccurate or unreliable \cite{himanen2019data}.

\textbf{Limitations in Learning Non-smooth Patterns.} Traditional deep learning struggles to learn non-smooth target functions that map molecular data to labels, as these target functions are frequently non-smooth in molecular property prediction. This implies that minor alterations in the chemical structure of a molecule can lead to substantial changes in its properties \cite{xia2023understanding}. Additionally, LLMs also find it difficult to solve this problem under the limited size of molecular datasets.

\textbf{Dangerous Behaviors and Products.} The field of chemistry carries certain inherent risks, as some products or reactions can be hazardous (e.g., flammable, explosive, toxic gases, etc.). LLMs may generate scientifically incorrect or unsafe responses, and in some cases, they may encourage users to engage in dangerous behavior \cite{zhao2024chemsafetybench}. Furthermore, LLMs can also be misused to create toxic or illegal substances \cite{guo2023can}. At the current stage of development, LLMs still cannot be fully trusted to ensure complete reliability.

On the other hand, despite the aforementioned limitations, the potential of LLMs in the fields of chemistry and chemical engineering is undeniable, as \textbf{they hold promise in addressing many challenges:}

\textbf{Decrease the Vast Chemical Exploration Space.} Inverse design enables the creation of synthesizable molecules that meet predefined criteria, accelerating the development of viable therapeutic agents and expanding opportunities beyond natural derivatives \cite{ramos2025review}. However, this quest faces a combinatorial explosion in the number of potential drug candidates—the chemical space made up of all small drug-like molecules—which is unimaginably large (estimated at \(10^{60}\)) \cite{ramos2025review}. Testing any significant fraction of these molecules, either computationally or experimentally, is simply impossible \cite{brenk2023escaping}. This field has been revolutionized by machine learning methods, particularly generative models, which narrow down the search space and enhance computational efficiency, making it possible to delve deeply into the seemingly infinite chemical space of drug-like compounds \cite{du2022molgensurvey}. LLMs, such as MolGPT \cite{bagal2021molgpt}, which employs an autoregressive pre-training approach, have proven to be instrumental in generating valid, unique, and novel molecular structures. The emergence of multi-modal molecular pre-training techniques has further expanded the possibilities of molecular generation by enabling the transformation of descriptive text into molecular structures \cite{xia2022systematic}.

\textbf{Generation of 3D Molecular Conformations.} Generating three-dimensional molecular conformations is another significant challenge in the field of molecular design, as the three-dimensional spatial structure of a molecule profoundly impacts its chemical properties and biological activity. Traditional computational methods are often resource-intensive and time-consuming, making it difficult for researchers to design and screen new drugs effectively. Unlike conventional approaches based on molecular dynamics or Markov chain Monte Carlo, which are often hindered by computational limitations (especially for larger molecules), LLMs based on 3D geometry exhibit remarkable superiority in conformation generation tasks, as they can capture inherent relationships between 2D molecules and 3D conformations during the pre-training process \cite{xia2022systematic}.

\textbf{Automate Chemical Agents.} Autonomous chemical agents combine LLM “brains” with planning, tool use, and execution modules to carry out experiments end‑to‑end. In the coscientist system, for example, GPT‑4 first decomposes a high‑level goal (“optimize a palladium‑catalyzed coupling”) into sub‑tasks (reagent selection, condition screening), retrieves relevant literature via a search tool, generates executable Python code for liquid‑handling robots, and then interprets sensor feedback to iteratively refine the protocol—closing the loop between design and execution \cite{boiko2023autonomous}. Similarly, Boiko et al. built an agent that plans ultraviolet–visible spectroscopy (UV–Vis) experiments by writing code to control plate readers and analyzers, automatically processing spectral data to identify optimal wavelengths, and even adapting to novel hardware modules introduced after the model’s training cutoff \cite{boiko2023emergent}. By leveraging LLMs for hierarchical task decomposition, self‑reflection, tool invocation (e.g., search APIs, code execution, robotics control), and memory management, these systems drastically accelerate repetitive experimentation and free researchers to focus on hypothesis generation rather than manual protocol execution \cite{boiko2023emergent}.

\textbf{Enhance Understanding of Complex Chemical Reactions.} The field of reaction prediction faces several key challenges that affect the accuracy of forecasting chemical reactions. A significant issue is reaction complexity, stemming from multi-step pathways and dynamic intermediates, which complicates product predictions, especially with varying conditions like different catalysts. Traditional models often struggle with these complexities, leading to biased outcomes. Utilizing advanced transformer architectures, LLMs can model complex relationships in chemical reactions and adjust predictions based on varying conditions. They excel in learning from unlabeled data through self-supervised pretraining, helping identify patterns in chemical reactions, particularly useful for rare reactions.

\textbf{Multi-task Learning and Cross-domain Knowledge.} The complexity of multi-task learning makes the simultaneous optimization of diverse prediction tasks difficult, while LLMs effectively handle this via shared representations and multi-task fine-tuning \cite{suvarna2024embracing}. Traditional methods also struggle to integrate cross-domain knowledge from chemistry, biology, and physics, yet LLMs address this seamlessly through pre-training and knowledge graph enhancement. 

\noindent\textbf{5.3.1.5 Taxonomy}

As summarized in Table~\ref{tab:chem-taxonomy}, in efforts to integrate chemistry research with artificial intelligence, particularly LLMs, many chemists primarily focus on tasks such as property prediction, property-directed inverse design, and synthesis prediction \cite{ramos2025review,suvarna2024embracing,mcdonald2023applied}. However, other chemists highlight additional significant tasks, including data mining and predicting synthesis conditions \cite{zheng2025large,pyzer2025foundation}. By synthesizing insights from these studies along with other
seminal works \cite{de2019synthetic,yu2023machine,butler2018machine}, we propose a more comprehensive classification method. This method accounts for both the rationality of chemical task classification and the characteristics of computer science.

From the chemistry perspective, our taxonomy echoes the field’s established research divisions—such as molecular property prediction, property‐directed inverse design, reaction type and yield prediction, synthesis condition optimization, and chemical text mining—ensuring that each category corresponds directly to a recognized experimental or theoretical task in chemical science. Concurrently, from the computer science perspective, by mapping every task onto a unified input–output modality framework, we add a computationally consistent structure that facilitates model development, benchmarking, and comparative analysis across diverse tasks within a single formal paradigm. Together, these dual alignments guarantee that our classification remains both chemically and algorithmically  meaningful.

\begin{figure}
\centering
\includegraphics[width=0.99\linewidth]{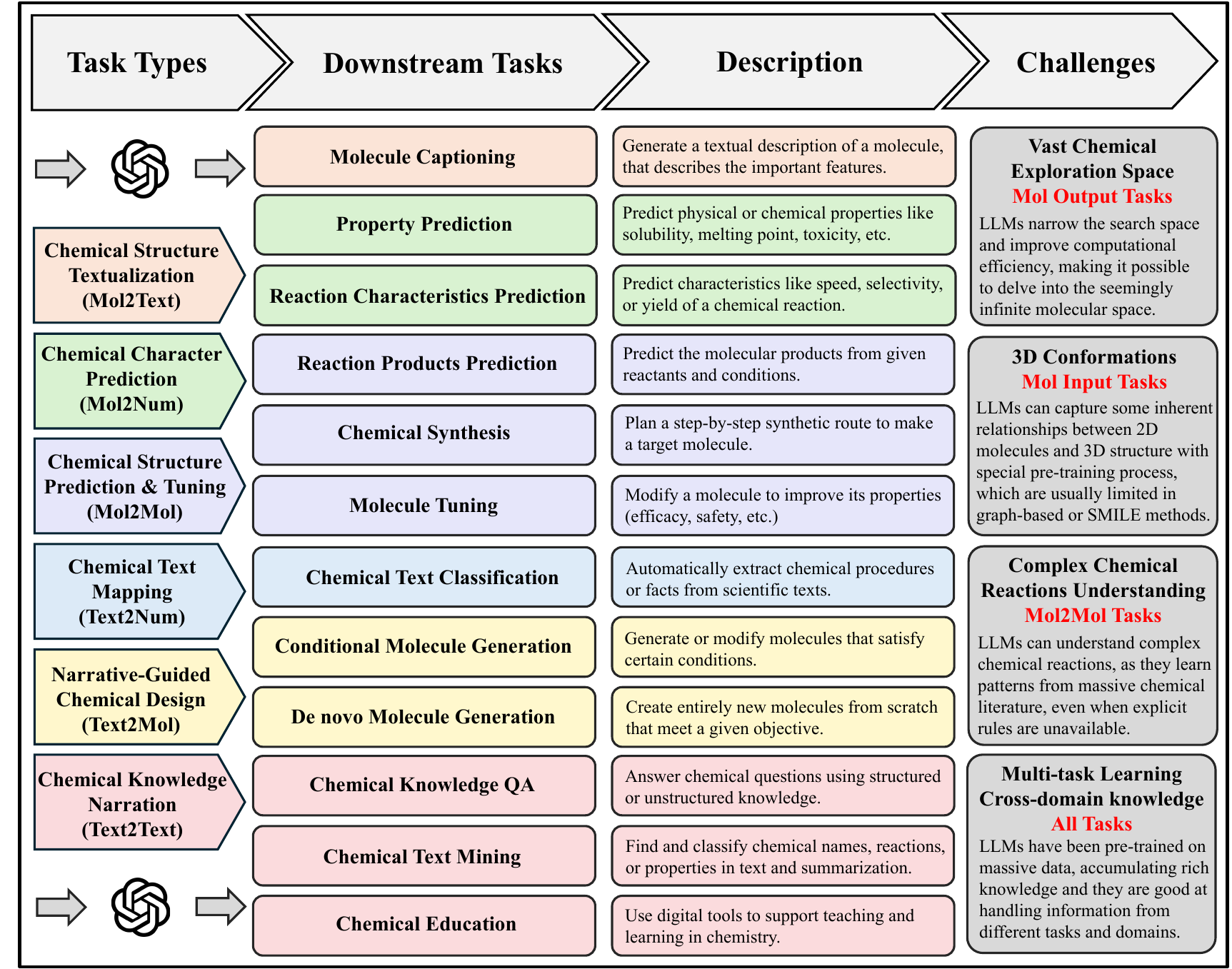}
\caption{A taxonomy of chemical tasks enabled by LLMs, categorized by input-output types and downstream objectives.}
\label{fig:chem_taxonomy}
\vspace{-10pt}
\end{figure}

\textbf{Chemical Structure Textualization.} Chemical structure textualization is the process of taking a molecule’s SMILES sequence as input and producing a detailed textual depiction that highlights its structural features, physicochemical properties, biological activities, and potential applications. Here, SMILES (Simplified Molecular Input Line Entry System) encodes a molecule’s atomic composition and connectivity as a concise, linear notation—for example, “CCO” denotes ethanol (each “C” represents a carbon atom and “O” an oxygen atom), while “C1=CC=CC=C1” represents benzene (the digits mark ring closure and “=” indicates double bonds)—enabling computational models to capture meaningful structural patterns and relationships for downstream text generation. Subtasks include molecule captioning, which exemplifies the goal of generating rich, human‑readable descriptions of molecules to give chemists and biologists rapid, accessible insights for experimental design and decision‑making.

\textbf{Chemical Characteristics Prediction.} Nowadays, SMILES provides a standardized method for encoding molecular structures into strings \cite{weininger1988smiles}. This string-based representation enables efficient parsing and manipulation by computational models and underpins a variety of tasks in cheminformatics, drug discovery, and reaction prediction. Notably, many machine learning models, including large-scale language models like GPT, are pre-trained or fine-tuned using corpora of SMILES sequences. Among the tasks leveraging SMILES input are property prediction and reaction characteristics prediction, where the model takes a SMILES sequence as input and outputs numerical values, categorical labels, or multi-dimensional vectors representing chemical properties, reactivity, bioactivity, and other experimentally relevant quantities.

\textbf{Chemical Structure Prediction \& Tuning.} Chemical structure prediction \& tuning tasks represent a classical form of sequence-to-sequence modeling \cite{wiki:seq2seq}, where the goal is to transform an input molecular sequence into an output sequence. In chemistry, this formulation is particularly intuitive because molecules are often represented as SMILES strings, which encode structural information in a linear textual format. Given an input SMILES sequence, the model learns to generate another SMILES string corresponding to a chemically meaningful transformation. This input–output structure underlies a variety of chemical modeling tasks, including reaction product prediction, chemical synthesis planning, and molecule tuning. For instance, the input may describe reactants or precursors, and the output may represent reaction products or structurally modified molecules, making these tasks central to computational reaction modeling and automated molecular design.

\textbf{Chemical Text Mapping.} Chemical text mapping tasks refer to the process of transforming unstructured textual input into numerical outputs such as labels, scores, or categories. At their core, these tasks involve analyzing chemical text—ranging from scientific articles to experimental protocols—and mapping the extracted information to structured numerical values for downstream applications like classification, relevance scoring, or trend prediction \cite{eltyeb2014chemical,ramos2025review}. A typical example is document classification, where the input is natural language text and the output is a discrete or continuous number representing, for example, a document’s category or relevance score. These tasks enable scalable analysis of chemical literature and facilitate integration of textual knowledge into data-driven modeling workflows.

\textbf{Narrative‑Guided Chemical Design.} Narrative‑Guided chemical design is a generative modeling paradigm extensively applied in chemistry and materials science, with the core objective of deriving molecular structures or material candidates that fulfill specific target properties or functional requirements \cite{zunger2018inverse,molesky2018inverse}. Unlike conventional forward design—which predicts properties from a given structure—inverse design begins with the desired outcome and works backward to propose compatible structures. In this context, the input is a description of the target properties, which may take the form of numerical constraints, categorical labels, or free-text descriptions, and the output is a molecular structure—typically represented as a SMILES string—that satisfies those specified criteria. This framework encompasses tasks such as de novo molecule generation and conditional molecule generation, enabling applications like targeted drug design, property-driven material discovery, and personalized molecular synthesis.

\textbf{Chemical Knowledge Narration.} Chemical knowledge narration tasks in chemistry refer to the transformation of one form of textual input into another, with both input and output grounded in chemical knowledge and language use \cite{ramos2025review}. These tasks leverage Natural Language Processing (NLP) techniques to process, convert, or generate chemistry-related textual data, thereby facilitating a range of downstream applications in research, education, and industry. For instance, given a textual input such as a paragraph from a research paper, the model may extract key information, translate it into another language, generate a summary, or answer domain-specific questions. Such tasks—encompassing chemical text mining, chemical knowledge question answering, and educational content generation—typically operate on natural language input and produce human-readable textual output, making them essential tools for improving access to and understanding of chemical information.

\input{tables/Chemistry_llm}

\subsubsection{Chemical Structure Textualization} 
Chemical tasks that map molecular structures to text serve as a bridge between the structural world of chemistry and human language. Chemical structure textualization is essentially “describing molecules in words,” akin to how one might recognize a complex object (like a new gadget) and then explain it in a common language \cite{guo2023can}. In everyday life, this is like looking at a detailed blueprint and verbally summarizing what it represents. In chemistry, such tasks are crucial: chemists often need to communicate structures verbally or in writing––for instance, by naming compounds or summarizing their features––so that others can understand them without seeing a drawing. Converting molecules to text makes chemistry more accessible and searchable (one can text‑search a compound name in literature, but not a structure diagram). For example, a medicinal chemist might draw a novel molecule and want the International Union of Pure and Applied Chemistry (IUPAC) name or a simple description to include in a report. An environmental chemist might identify an unknown substance and need an AI to generate a description like “a chlorinated hydrocarbon solvent.” These translations from structure to language are essential for reporting, education, and integrating chemical information into broader databases.

\textbf{Chemical Structure Textualization is fundamentally a Mol2Text task,} mapping chemical structure representations (e.g., SMILES) as input to corresponding natural language descriptions as output. Mol2Text encompasses any scenario of “structure to language.” Key subtasks include chemical nomenclature generation (e.g.\ converting a molecular structure into its IUPAC name or common name), molecule captioning (generating a sentence describing the molecule’s class or properties), and even property or hazard annotations (predicting textual labels like “flammable” from a structure). For instance, nomenclature generation might take a SMILES string of a molecule and output “2-(4‑Isobutylphenyl)propanoic acid” (the IUPAC name of ibuprofen). Molecule captioning could produce a phrase like “an aromatic benzene derivative with two nitro substituents” given a structure. They are LLM‑friendly because large databases of molecules with names or descriptions exist (providing plenty of textual training data), and the output is structured text that follows rules (ideal for sequence generation models). \textbf{We therefore highlight the area of molecule captioning in the following paragraph.} These illustrate how LLMs evolved from simply reading text to “reading” chemistry and writing text.

\textbf{Molecular Captioning.} Beyond formal nomenclature, a growing chemical structure textualization application is molecular captioning––generating a natural‑language description of a molecule’s structure or function. That is given a molecule’s representation (e.g. SMILES or graph), generating a coherent natural-language description that accurately captures its structural features, physicochemical properties, and potential biological activities. This is analogous to image captioning in computer vision (where an AI might look at a photo and say “a cat sitting on a sofa”). Here, the “image” is a chemical structure (which an LLM might ingest as a string like SMILES or another representation), and the output is a concise text description. For example, given the structure of caffeine, an ideal caption might be: “a bitter, white alkaloid of the purine class, commonly found in coffee and tea.” Or more straightforwardly: “Caffeine – a stimulant compound with a purine (xanthine) scaffold and multiple methyl groups.” This task is in its infancy, but its significance is clear: it would allow chemists to quickly get a textual summary of a molecule’s key features or known uses. In everyday life, this is akin to seeing a new plant and describing it as “a tall tree with waxy leaves and fragrant white flowers”––it communicates key identifying features in an intuitive way. 
MolT5 \cite{edwards2022translation} first achieved end‑to‑end text-SMILES translation by jointly training on vast amount of SMILES and textual descriptions, but relying solely on sequence information led to captions lacking intuitive structural understanding. To address this, MolFM \cite{luo2023molfm} introduced multimodal pretraining that combines SMILES, InChI, and text descriptions, significantly improving annotation accuracy and richness, yet it did not leverage molecular images for visual comprehension . Next, GIT‑Mol \cite{liu2024git} integrated molecular graphs, images, and text through cross‑modal fusion, achieving higher fidelity captions but at the cost of large model size and high deployment and inference overhead . To improve efficiency and deployment, MolCA \cite{liu2023molca} designed a cross‑modal projector and uni‑modal adapters, greatly reducing fine‑tuning and deployment costs while retaining multimodal capabilities, though its pretraining data coverage still needs expansion . Most recently, GraphT5 \cite{kim2025grapht5} employs multimodal cross‑token attention to tightly integrate molecular graph structures with a language model, balancing caption quality and model scale, and providing an efficient and scalable foundation for molecule captioning

\textbf{Problems Solved by LLMs.} LLMs have dramatically improved both chemical nomenclature generation and molecular captioning by learning from vast corpora of paired data. In the case of nomenclature, Transformer‑based models such as Struct2IUPAC have achieved accuracies of 98–99\% in converting SMILES strings to formal IUPAC names—performance that rivals rule‑based systems like Open Parser for Systematic IUPAC Nomenclature (OPSIN), which itself set the benchmark for open‑source name‑to‑structure parsing over a decade ago \cite{krasnov2021transformer}. Simultaneously, proof‑of‑concept captioning studies have shown that LLMs can associate common substructures with descriptive terms (e.g., “–NO$_2$” $\rightarrow$ “nitro compound”), enabling models like MolT5 to generate concise textual summaries of molecular features \cite{edwards2022translation}. Together, these successes illustrate that LLMs can both “read” and “write” chemistry, transforming structural representations into human‑readable language with high fidelity.

\textbf{Remaining Challenges.} Despite these advances, challenges remain in both domains. Chemical naming models, while highly accurate, operate as black boxes; when they err on novel or highly complex molecules, it is difficult to trace the decision pathway or understand which structural elements led to a misnaming. Moreover, evolving IUPAC standards—such as recent organometallic nomenclature updates—require continual model retraining or fine‑tuning to maintain correctness. In molecular captioning, the absence of a single “ground truth” means that errors are subtler and often manifest as partial truths or outright hallucinations (e.g., asserting a molecule is “used in perfumes” without evidence), and models struggle to calibrate the appropriate level of detail versus generalization. This fuzziness poses risks in scientific communication, as speculative or incorrect descriptors can mislead users.

\textbf{Future Work.} Looking ahead, integrating LLMs with external knowledge sources and multimodal inputs promises to address many of these limitations. Hybrid pipelines that combine neural proposals with rule‑based validation could ensure 100\% naming accuracy while preserving flexibility for new nomenclature conventions. Likewise, coupling captioning models with structured databases—so that, upon recognizing “glucose,” an LLM retrieves and incorporates the formula C$_6$H$_{12}$O$_6$—would enhance factual correctness. Finally, multimodal architectures capable of ingesting both SMILES and 2D structural images, or embedding property predictors to append numerical descriptors (e.g., molecular weight, logP), will yield richer, more reliable textual outputs and usher in a new era of AI‑driven chemical communication.

\subsubsection{Chemical Characteristics Prediction}
Chemical characteristics prediction task focuses on predictive modeling of molecular data, aiming to predict specific outcomes related to molecular structures using machine learning techniques. The primary input for this task is a detailed representation of molecules, typically in the form of SMILES strings, these representations allow models to learn meaningful patterns and relationships that inform predictions. Chemical \textbf{Characteristics Prediction is fundamentally a Mol2Number task}, taking molecular representations as input and producing quantitative property values as output.

In regression tasks, the goal is to predict continuous numerical values based on molecular features. For example, given the SMILES string "CCO" representing ethanol, a model might predict its boiling point as approximately 78.37°C. Similarly, for the SMILES "CC(=O)O" representing acetic acid, the model could predict a solubility of about 1000 g/L in water \cite{yu2023solvbert}. In another case, the SMILES "C1=CC=CC=C1" for benzene might be used to predict its logP value as around 2.13. Similarly, in classification tasks, the objective is to determine discrete outcomes. For instance, given the SMILES "CC(=O)OC1=CC=CC=C1C(=O)O" for aspirin, the model might classify it as a non-steroidal anti-inflammatory drug (NSAID). Similarly, an example of reaction classification involves the reaction with the SMILES notation Brc1ccccc1.B(O)O >> c1ccccc1B(O)O, where the task is to classify the reaction type. The model classifies this reaction as a "Suzuki coupling" with a confidence of 98.2\%.

The value of chemical characteristics prediction tasks in chemistry lies in their capacity to translate complex molecular and reaction information into quantitative predictions—enabling both experts and non‑specialists to anticipate how molecules will behave under given conditions, including chemical properties, reaction types, yields, and reaction rates. For example, in property prediction, the SMILES string “CCO” succinctly encodes the structure of ethanol (“C” for carbon, “O” for oxygen), allowing a model to infer its physical and chemical properties without recourse to time‑consuming quantum calculations. When we talk about reaction types, imagine mixing baking soda (NaHCO$_3$) and vinegar (CH$_3$COOH). A chemical characteristics prediction model would read the SMILES for each ingredient, see “NaHCOO” + “CC(=O)O,” and label it as an acid–base neutralization. That label helps chemists know that the reaction will produce CO$_2$ gas. Predicting yields is like estimating how much carbon dioxide you’ll collect in a balloon when you mix your soda‑and‑yeast “volcano” experiment. A yield‑prediction model might tell you, “Under these conditions, you’ll get about 80\% of the maximum CO$_2$ possible,” so you can plan ahead and avoid wasting ingredients. When discussing reaction‑rate prediction, think of Alka‑Seltzer fizzing in water. A model could predict how fast the tablet dissolves—does it take 30 seconds or three minutes?—based on temperature or how finely the tablet is crushed. In short, chemical characteristics prediction tasks let scientists—and even curious students—see, ahead of time, which “kitchen chemistry” will work best, how much product to expect, and how fast it will happen. This not only cuts down on costly trial‑and‑error in drug discovery or materials design, but also deepens our understanding of how molecules behave and reactions proceed.

Chemical characteristics prediction tasks encompass several categories: molecular property prediction (e.g., solubility — the maximum amount of a substance that can dissolve in a solvent, typically expressed in mol/L); bioactivity and binding‑affinity prediction (e.g., IC$_{50}$ — the concentration at which a compound inhibits 50\% of a target’s biological activity); reaction characters prediction (e.g., reaction‑type classification — categorizing a reaction by its mechanism, such as nucleophilic substitution or acid–base neutralization; and yield prediction — estimating the percentage of product obtained relative to the theoretical maximum under specified conditions); retrosynthesis‑route scoring (e.g., feasibility scoring — assessing whether a proposed synthetic pathway can be practically executed in the laboratory; and cost scoring — predicting the total economic expense of reagents and operations); force‑field and energy prediction (e.g., potential‑energy surfaces — energy landscapes that map molecular geometry to potential energy; and interatomic forces — forces between atoms calculated as gradients on that energy surface); and molecular descriptor or embedding generation (e.g., low‑dimensional embeddings — concise numerical vectors that capture key molecular characteristics in a reduced feature space).

Due to the intrinsic similarities among these tasks, we select two of the most representative ones—Property Prediction and Reaction Characters Prediction — for discussion in this paper. These two tasks benefit most from LLMs’ strengths: (1) they have abundant, well‐structured textual and experimental data (e.g., MoleculeNet benchmarks, USPTO reaction corpora) that LLMs can readily learn from; (2) LLMs can provide both numerical predictions and human‐readable rationales, enhancing interpretability over more opaque methods; and (3) improvements in these areas directly accelerate molecule prioritization and reaction planning in research and industry.

\textbf{Property Prediction.} The universal value of chemistry lies in accurately predicting compound properties to guide their practical use. Property Prediction is the task of predicting a molecule’s physicochemical or biological properties (e.g., solubility, binding affinity, toxicity) given its representation (such as a SMILES string, molecular graph, or descriptor vector). In pharmaceuticals, understanding how molecular structure influences bioactivity and toxicity enables the design of safer, more effective drugs; in materials science, predicting properties such as solubility, thermal stability, or mechanical strength from chemical structure accelerates the development of advanced polymers and functional materials. Traditional computational methods like quantum calculations and molecular dynamics offer high accuracy but demand extensive resources, whereas machine learning models can predict properties more efficiently. Recently, LLMs have demonstrated strong performance in molecular property and reaction‐outcome prediction by leveraging vast textual and experimental datasets, achieving competitive accuracy without the heavy computational overhead of physics‐based simulations. Combined with expert insight, AI‐driven property prediction promises to revolutionize compound prioritization and materials design by focusing experimental efforts on the most promising candidates.

In early studies, LLMs such as BERT were applied to chemical reaction classification tasks. A representative work by Schwaller et al. achieved an impressive classification accuracy of up to 98.2\%. The application focus then shifted from reaction classification to molecular property prediction, especially under scenarios with limited labeled data. Wang et al. proposed a semi-supervised model, SMILES-BERT \cite{wang2019smiles}, which was pretrained on large-scale unlabeled data via a "masked SMILES recovery" task. It achieved state-of-the-art performance across multiple datasets and marked the first successful application of BERT in drug discovery tasks. During the early exploration of molecular language models, Chithrananda et al. introduced ChemBERTa \cite{chithrananda2020chemberta}, systematically examining the impact of pretraining dataset size, tokenization strategies (BPE vs. SmilesTokenizer), and molecular representations (SMILES vs. SELFIES) on model performance. Results showed that increasing the pretraining data from 100K to 10M led to significant improvements in downstream tasks such as BBBP and Tox21. Although ChemBERTa did not outperform the GNN baseline Chemprop, the authors suggested that further scaling could close this gap. Tokenization comparisons showed a slight advantage for the custom tokenizer. While no significant difference was observed between SMILES and SELFIES, attention head visualization using BertViz revealed neuron selectivity to functional groups, highlighting the importance of proper benchmarking and awareness of model carbon footprint. Building on this, Ahmad et al. developed ChemBERTa-2 \cite{ahmad2022chemberta}, aiming to create a general-purpose foundation model. With a multi-task regression head and pretraining on 77 million molecules, ChemBERTa-2 achieved comparable performance to state-of-the-art models on MoleculeNet tasks. The study emphasized that different pretraining strategies had varying effects on downstream tasks, suggesting that model performance depends not only on pretraining itself, but also on the specific chemical context and fine-tuning dataset. Further extending this direction, Yuksel et al. proposed SELFormer \cite{yuksel2023selformer}, incorporating SELFIES to address concerns about the validity and robustness of SMILES. Pretrained on 2 million drug-like compounds and fine-tuned on a range of property prediction tasks (e.g., BBBP, SIDER, Tox21, HIV, BACE, FreeSolv, ESOL, PDBbind), SELFormer achieved leading performance in several cases. It demonstrated the ability to distinguish between molecules with varying structural properties, and suggested that future models should integrate structural data and textual annotations to build multimodal representations, enhancing generalizability and real-world utility. 

To further improve molecular structure representation, Maziarka et al. introduced the MAT (Molecule Attention Transformer) \cite{maziarka2020molecule}, incorporating atomic distances and molecular graph structure into the attention mechanism. This graph-structured self-attention led to performance gains in property prediction. Fabian focused on capturing molecular substructures and proposed Mol-BERT \cite{fabian2020molecular}, pretrained on 4 million molecules from ZINC and ChEMBL. Treating fingerprint fragments as "words" and using Masked Language Modeling (MLM) to learn sentence-level molecular semantics, Mol-BERT outperformed both GNNs and sequence models on tasks like Tox21 and SIDER. Ross et al. developed MoLFormer, trained on over 1.1 billion SMILES from ZINC and PubChem. By introducing Rotary Position Embeddings, it more effectively captured atomic sequence relationships. MoLFormer not only surpassed GNNs on various benchmarks but also achieved 60x energy efficiency, representing progress toward environmentally sustainable AI. 

On model generalization, Zhang et al. identified a bottleneck in the lack of correlation across different property datasets. They proposed MTL-BERT \cite{zhang2022pushing}, a multi-task learning model pretrained on large-scale unlabeled SMILES from ChEMBL. MTL-BERT improved prediction performance and enhanced interpretability of complex SMILES by extracting context and key patterns. 

On the task-specific level, Yu et al. proposed SolvBERT \cite{yu2023solvbert}, a multi-task regression model designed to predict both solvation free energy and solubility. Despite the traditional reliance on 3D structural modeling for such tasks, SolvBERT—using only SMILES—achieved performance competitive with, or even superior to, GNN-based approaches, showcasing the potential of text-based modeling in physical chemistry.

While model performance continues to improve, limited labeled data remains a major challenge. In 2024, Jiang et al. introduced INTransformer \cite{jiang2024intransformer}, which incorporated perturbation noise and contrastive learning to augment small datasets and improve global molecular representation, even under low-resource conditions. Similarly, MoleculeSTM \cite{liu2023multi} used contrastive learning to align SMILES strings and textual molecular descriptions extracted from PubChem using a LLM. Extending this idea to proteins, Xu et al. proposed ProtST \cite{xu2023protst}, which models protein sequences using a protein language model and aligns them with protein descriptions encoded by LLMs, exploring multimodal fusion for biomacromolecule modeling.

\textbf{Reaction Characters Prediction.} Typical tasks in chemical reaction property prediction include reaction type classification (determining which type or mechanistic category a given reaction belongs to), reaction yield prediction (estimating the yield of the target product under specific conditions), and reaction rate prediction (assessing the kinetics of the reaction, such as activation energy or rate constant). These studies are of great importance in the fields of pharmaceuticals, materials science, and chemical engineering. For example, as early as 2018, Ahneman et al. demonstrated that machine learning could predict the yields of untested combinations in coupling reactions based on a limited amount of experimental data, successfully identifying previously unknown high-yield catalytic systems \cite{ahneman2018predicting}. Moreover, in the search for more efficient organic photovoltaic materials, it is often necessary to synthesize a series of candidate molecules. By using models to predict the yield of each reaction step, researchers can eliminate candidates with expected low yields and poor scalability early in the process, instead prioritizing synthetic routes that are predicted to be high-yielding and require fewer steps. This approach accelerates the screening process and conserves reagents.

Reaction type prediction aims to determine which category a given reaction belongs to—such as Suzuki coupling or Diels–Alder—based on its reactants and products. Traditionally, chemical reaction classification has relied on manually crafted rules or template libraries, but these approaches are poorly robust to new data and require complex atom‐mapping preprocessing. To overcome this, Schwaller et al. introduced RXNFP \cite{schwaller2021mapping}, a Transformer-based encoder that learns fixed-length embeddings of entire reactions directly from unannotated SMILES in large datasets (e.g., USPTO) and then uses a simple k-NN or classifier to assign reaction classes. While RXNFP has been reported to achieve very high classification accuracy on reaction classification benchmarks (e.g.\ over 98\% on some USPTO subsets) \cite{schwaller2021mapping}, it remains primarily a static feature extractor and is not designed for generative tasks like product generation or sequence-to-sequence modeling for continuous outputs. T5Chem \cite{lu2022unified} addresses many of these gaps by casting reaction tasks—classification, product prediction, retrosynthesis, and yield regression—as text-to-text problems. A single T5 model, pretrained on large molecular datasets from PubChem and fine-tuned on public reaction sets, achieves strong performance across multiple tasks with one shared architecture, improving multitask efficiency and generalization.

The prediction of reaction yields has long been a central challenge in synthesis planning and industrial optimization, owing to the complex interplay among substrate structures, reagents, catalysts, solvents, temperature, and other factors. Initially, Schwaller et al. leveraged their RXNFP reaction fingerprints by feeding the learned fixed-length reaction embeddings into a regression head to provide preliminary yield predictions for Buchwald–Hartwig and Suzuki–Miyaura coupling reactions, demonstrating the feasibility of end-to-end transformer embeddings for yield regression. However, RXNFP was not specifically designed for yield prediction, and its static fingerprints lacked sensitivity to changes in reaction conditions. To address this, T5Chem \cite{lu2022unified}, a unified text-to-text multitask framework that, in addition to reaction classification and product prediction, incorporates a regression head for yield prediction. Pretrained on molecular data from PubChem and jointly fine-tuned on datasets such as USPTO, T5Chem matches or surpasses many baseline models across reaction prediction and yield tasks, showing that a single model can perform well on multiple reaction-related tasks. Building on this, the Schwaller team developed Yield-BERT \cite{schwaller2021prediction} in 2021 by fine-tuning ChemBERT on reaction SMILES to directly output yields; in high-throughput coupling reaction datasets Yield-BERT has been shown to achieve strong $R^2$ performance (e.g.\ values exceeding 0.90) compared to traditional methods using DFT-derived descriptors or handcrafted fingerprints. Yet Yield-BERT’s sensitivity to variations in catalysts, solvents, and other reaction conditions is limited, hindering its generalization across differing condition combinations. To enhance condition sensitivity, Yin et al. launched Egret \cite{yin2024enhancing} in 2023, combining masked language modeling with condition-based contrastive learning to teach the model to distinguish yield differences for the same substrates under varying conditions; Egret achieved improved $R^2$ scores in several public benchmarks. Subsequently, Sagawa and Kojima's ReactionT5 \cite{sagawa2023reactiont5} employed a two-stage pretraining strategy—first training CompoundT5 on a molecular library, then pretraining on a reaction-level database—enabling the model, with limited fine-tuning data, to achieve good performance ($R^2$ in challenging splits) in yield and product prediction tasks, highlighting the value of reaction-level pretraining. Most recently, the ReaMVP \cite{shi2024prediction} framework further incorporated 3D molecular conformations into pretraining alignment, aligning sequence and geometric views during a self-supervised stage, followed by fine-tuning on labeled yield data, resulting in modest boosts in $R^2$ on out-of-sample reactions and demonstrating the importance of multimodal information fusion for improving the generalizability of yield predictions.

\textbf{Problems Solved by LLMs.} LLMs have significantly advanced molecular property prediction by leveraging self‑supervised learning on large unlabeled datasets to learn robust molecular representations that improve generalization to limited labeled data. They also enable effective multi‑task learning through shared representations and task‑specific fine‑tuning strategies that mitigate interference between diverse prediction objectives. Furthermore, by integrating domain knowledge from chemistry, biology, and physics via pretraining on multimodal data and knowledge graph augmentation, these models can incorporate cross‑domain insights seamlessly. Finally, LLMs excel at processing contextual molecular representations such as SMILES and International Chemical Identifier (InChI) codes, automatically learning high‑dimensional features that capture complex structural interactions without the need for manual feature engineering.

\textbf{Remaining Challenges.} Despite these advances, several challenges persist. The vastness of chemical space requires models that can reliably generalize to structurally novel molecules and unseen scaffolds, a task that remains difficult for current architectures . Activity cliffs, where minor structural modifications lead to dramatic changes in molecular properties, continue to undermine prediction accuracy and demand models that are sensitive to such subtle variations. Moreover, the inherently graph‑structured nature of molecular data necessitates specialized neural architectures—such as graph neural networks and graph transformers—that can effectively capture both local and global structural patterns. Additionally, certain properties depend on three‑dimensional conformations or quantum mechanical effects, which two‑dimensional representations alone cannot fully capture, highlighting the need for methods that incorporate 3D structural information.

\textbf{Future Work.} Future work will focus on developing hybrid molecular representations that combine two‑dimensional graph features with three‑dimensional geometric descriptors—including conformer ensembles, steric effects, and electrostatic interactions—to more accurately model spatial relationships within molecules . Integrating molecular dynamics simulations and physics‑informed neural networks can further enrich these representations by providing dynamic and mechanistic insights into molecular behavior over time. These advances are expected to enhance the generalization of models across diverse reaction conditions, improve the reliability of reaction yield predictions, and accelerate the discovery of novel compounds with desired properties.

\subsubsection{Chemical Structure Prediction and Tuning}

In many chemistry problems, the desired output is another molecule. These tasks can be seen as “translating one molecule into another” – hence chemical structure prediction \& tuning. \textbf{Chemical Structure Prediction \& Tuning is inherently a Mol2Mol task.} This category includes chemical reaction predictions, retrosynthesis planning, and molecule optimization, among others. An analogy from daily life would be cooking: you start with ingredients (molecules) and through a recipe (reaction), end up with a dish (a new molecule). Alternatively, think of it as solving a jigsaw puzzle: you have pieces (fragments of molecules) and want to put them together into a final picture (target molecule) – the input pieces and the output picture are both made of the same stuff, just rearranged. Chemical structure prediction \& tuning tasks are central to chemistry because they essentially encompass chemical synthesis and design – predicting what will happen if molecules interact, or figuring out how to get from one molecule to another. For example, a forward reaction prediction might answer: “If I mix molecules A and B, what product will form?” A retrosynthesis task does the opposite: “I want molecule Z; what starting molecules could I use to make it?” These tasks directly assist chemists in the lab by suggesting likely outcomes or viable synthetic routes, thus speeding up research and discovery \cite{NamKim2016}.

Key subtasks under chemical structure prediction \& tuning include reaction outcome prediction (given reactant molecules and possibly conditions, predict the product molecules), chemical synthesis (particularly retrosynthesis, given a target product, propose one or more sets of reactants that could produce it), and molecule‐to‐molecule optimization (propose a structural modification to an input molecule to improve some property, e.g.\ “suggest a similar molecule with higher potency”). Another subtask is chemical pathway completion (extending a partial sequence of reactions by suggesting the next molecule). All these involve generating molecules from molecules. Reaction prediction (input: reactants, output: products) is a prime example: e.g.\ input SMILES for ethanol + acetic acid, output SMILES for ethyl acetate (the esterification product). Chemical synthesis is similarly crucial: input a drug molecule, output a plausible precursor like an aromatic halide plus a coupling partner. Molecule tuning is essential for fine‑adjusting a drug’s potency, selectivity, and pharmacokinetic properties while retaining its core active scaffold (e.g., introducing an amino group into the side chain of penicillin G produces ampicillin, thereby significantly improving oral bioavailability and antibacterial spectrum). We focus on these tasks because these have seen extensive development with LLM‐like models and there are large datasets (like USPTO patent reactions) to train on. 

\textbf{Reaction Product Prediction.} Reaction prediction is the task of predicting the products of a chemical reaction given the reactants (and sometimes reagents or conditions). It’s effectively a translation of a set of input molecules into a set of output molecules. In the language analogy, if reactants are words, the chemical reaction is a grammatical rule that rearranges those words into a new sentence (the product). A more concrete life analogy: consider mixing ingredients in cooking – if you combine flour, sugar, and butter and bake, you predict a cake as the outcome. Similarly, if you mix benzene with chlorine under certain conditions, you predict chlorobenzene (and hydrogen chloride as a byproduct) as the outcome. For decades, chemists approached this with rules (“if you see this functional group and add that reagent, you get this outcome”). The forward reaction prediction task asks an AI to learn those implicit rules from data. It’s significant because the space of possible products is enormous, and a human chemist’s intuition can be wrong or limited to known reactions. An accurate model can enumerate likely products, flagging surprises or confirming expectations. This can prevent wasted experiments and guide chemists toward successful reactions. It’s particularly useful in drug discovery or complex synthesis planning, where predicting side-products or main products can inform route selection.

Early machine learning models for reaction prediction often used template‐based systems: large libraries of reaction templates were extracted, and algorithms matched reactants to these templates to propose products. While effective, those methods required expert‐curated templates and struggled with reactions not in the template library. The turning point was realizing that chemical reactions could be encoded as strings (e.g.\ “CCO + O=C=O $\rightarrow$ CCC(=O)O” for esterification) and treated like a language translation problem. In 2016, Nam and Kim first applied a sequence‐to‐sequence RNN to reaction SMILES, showing that neural machine translation can predict organic reaction products directly from SMILES \cite{NamKim2016}. However, RNNs had limitations – they sometimes forgot parts of the input and struggled with very long SMILES or complex rearrangements.

The real leap came with Transformers. In 2019, Schwaller et al. introduced the Molecular Transformer, a Transformer‐based model for reaction outcomes that achieved 95\% top-1 accuracy on the USPTO benchmark, significantly outperforming both template-based and RNN approaches \cite{schwaller2019molecular}. By leveraging self-attention, the model considered all reactant tokens simultaneously, capturing reaction context more effectively. The Molecular Transformer also provided uncertainty estimates, enabling chemists to gauge prediction confidence. Despite its success, the first‐generation Transformer had limitations. It tended to memorize frequent reaction patterns, leading to overconfidence on well-represented reactions and underperformance on rare or novel chemistries. It also handled only single‐step reactions and did not predict yields or selectivities. To address memorization and improve generalization, SMILES augmentation was introduced—randomizing atom orders during training—which reduced overfitting and boosted top-accuracy metrics on large USPTO subset. More recently, general-purpose LLMs have been fine-tuned on reaction data. For example, GPT-3 was adapted to reaction prediction tasks and achieved performance that is competitive with specialized Transformer models on some USPTO-style datasets \cite{Guo2023PredictiveChemistry}. However, in zero-shot or few-shot settings, GPT-3.5 and GPT-4 still lag behind domain‐trained models in tasks requiring precise structural prediction, with top-1 accuracies substantially lower in unconstrained prediction tasks \cite{chen2024question}. These findings underscore the continued importance of task-specific training and data augmentation for reliable reaction outcome prediction.

\textbf{Chemical Synthesis.} Once a target molecule has been identified, the next major challenge is to predict its optimal synthetic route—including per-step and overall yield—under realistic chemical constraints \cite{shenvi2024natural}. While the demanding, elegant total syntheses of complex natural products have historically driven advances in organic chemistry, the past two decades have prioritized broadly applicable catalytic reactions \cite{shenvi2024natural}; only recently has complex synthesis become relevant again as a digitally encoded knowledge source that can be mined by LLMs \cite{ai2024extracting}. Unlike single-molecule property prediction, reaction planning must account for the multi-body nature of synthesis—modifying one reactant often requires re-optimizing all others under different mechanisms or conditions—and must balance multiple objectives, such as maximizing overall yield, minimizing the number of steps and cost of readily available starting materials, and ensuring chemical compatibility at each stage. Planning can proceed forward—from simple substrates to the target—or, more commonly, via retrosynthesis, introduced by E. J. Corey \cite{corey1988robert}, which deconstructs the target molecule into fragments that are reassembled most effectively from inexpensive, commercially available reagents. Retrosynthesis is the reverse of reaction prediction: given a target molecule (the product), the task is to predict one or more sets of reactant molecules that could form it in a single step. It’s essentially “un‑cooking” the dish to figure out the ingredients. In language terms, if a forward reaction is like forming a sentence from words, retrosynthesis is like taking a completed sentence and figuring out how to split it into two meaningful phrases that could combine to make that sentence. For example, the target “ethyl acetate” could be retrosynthesized to reactants “acetic acid + ethanol” (in the presence of an acid catalyst). Doing this well requires both creativity and extensive knowledge of known reactions, because the space of possible reactant combinations is enormous. For example, the first total synthesis of discodermolide required 36 individual steps (with a longest linear sequence of 24) to achieve only a 3.2 \% overall yield, vividly illustrating the vast combinatorial explosion of possible routes and the reliance on expert intuition. By coupling structure–activity relationship predictions with synthesis planning, LLM-based approaches now promise to select or even design molecules not only for optimal properties but also for tractable, high-yielding synthetic accessibility—enabling both rapid route discovery and the creation of novel non-natural compounds chosen for their ease of synthesis and predicted performance \cite{ramos2025review}.

Early computer-assisted synthesis planning began with recurrent architectures and handcrafted rules: Nam and Kim pioneered forward reaction prediction using a GRU-based translation model \cite{nam2016linking}, while Liu et al. applied an LSTM + attention seq2seq framework to retrosynthesis, achieving just 37.4 \% accuracy on USPTO-50K \cite{liu2017retrosynthetic}. Schneider et al. then enhanced retrosynthesis by algorithmically assigning reaction roles to reagents and reactants \cite{schneider2016s}, and rule-based, template-driven systems such as Chematica and Segler  \& Waller \cite{segler2017neural} captured explicit atomic and bond transformations in reverse planning—training on millions of reactions to deliver 95 \% top-10 retrosynthesis accuracy and 97 \% reaction-prediction accuracy—yet remained limited by their reliance on manually curated template libraries and inability to propose truly novel transformations. Semi-template methods struck a balance: Somnath et al.’s synthon-based graph model decomposed products into fragments and appended relevant leaving groups, boosting top-1 accuracy to 53.7 \% on USPTO-50K while retaining interpretability \cite{somnath2021learning}.

While early synthesis planning methods relied on RNNs and handcrafted templates, the advent of LLMs has transformed the field by treating chemical synthesis as a data-driven “translation” task. Schwaller et al. \cite{transformer2019model} first demonstrated this paradigm with a regex-tokenized LSTM-attention network that learned retrosynthetic rules directly from raw USPTO reactions—removing the need for explicit templates and uniquely tokenizing recurring reagents to distinguish solvents and catalysts. Building on that work, the Molecular Transformer  applied the full Transformer encoder–decoder architecture to both forward reaction and retrosynthetic prediction, inferring subtle correlations between reactants, reagents, and products without any handcrafted rules and achieving state-of-the-art accuracy on USPTO-MIT, USPTO-LEF, and USPTO-stereo benchmarks. To extend LLMs beyond single-step predictions, Schwaller et al. introduced a hypergraph exploration strategy in their 2020 Molecular Transformer model \cite{schwaller2020predicting}, dynamically expanding candidate routes using Bayesian-like scores and evaluating them with four new metrics—coverage (how much of chemical space is reachable), class diversity (variety of reaction types), round-trip accuracy (can predicted precursors regenerate the product), and Jensen–Shannon divergence (how closely the model’s predictions match real-world distributions). That same year, Zheng et al.’s SCROP \cite{zheng2019predicting} model combined a template-free transformer with a neural syntax corrector to self-correct invalid SMILES, boosting top-1 retrosynthesis accuracy on USPTO-50K to 59.0\%—over 6\% better than template-based baselines.

More recently, pretrained encoder–decoder LLMs have further elevated performance and flexibility. Irwin et al.’s Chemformer \cite{irwin2022chemformer} used a BART backbone pretrained on millions of SMILES strings, then fine-tuned for sequence-to-sequence synthesis tasks and discriminative property predictions (ESOL, Lipophilicity, FreeSolv), demonstrating that task-specific pretraining is essential for efficiency and accuracy. In 2023, Toniato et al. \cite{toniato2023enhancing} introduced prompt-engineering into retrosynthesis by appending classification tokens to target SMILES, guiding the model toward diverse disconnection strategies and producing multiple viable routes “out of the box.” Finally, Fang et al.’s MOLGEN \cite{fang2023domain} leveraged BART pretraining on 100 million SELFIES representations, domain-agnostic molecular prefix tuning, and an autonomous chemical feedback loop to ensure generated molecules are valid, non-hallucinatory, and retain their intended properties—foreshadowing autonomous LLM agents capable of end-to-end molecular design, synthesis planning, and iterative optimization.

\textbf{Molecule Tuning.} Molecule Tuning is the task of taking an initial molecule representation (e.g., SMILES string or graph) along with desired property modifications and generating structurally related analogs that optimize those specified properties while preserving the core scaffold. Molecule tuning leads compounds to simultaneously improve properties like potency, solubility, and safety—this is a cornerstone of drug design.

Early LLM-based approaches, such as DrugAssist \cite{ye2025drugassist}, introduced an interactive, instruction-tuned framework that lets chemists iteratively “chat” with the model to optimize one or more properties at a time. DrugAssist has achieved leading results in both single- and multi-property optimization tasks, showing strong potential in transferability and iterative improvement \cite{ye2025drugassist}. However, it requires fine-tuning on task-specific datasets and its generalization to entirely new property combinations remains a challenge in practice. To advance further, Chemlactica and Chemma \cite{guevorguian2024small} were developed by fine-tuning language models on a large corpus of 110 million molecules with computed property annotations. These models demonstrate strong performance in generating molecules with specified properties and predicting molecular characteristics from limited samples; relative improvements over prior methods (for example on the Practical Molecular Optimization benchmark) indicate they outperform earlier approaches in multi-property molecule optimization \cite{guevorguian2024small}. Despite these advances, fully zero-shot multi-objective optimization—where a model satisfies several new property constraints simultaneously without any additional training—remains difficult. Some approaches aim toward this goal (for example via prompt engineering, genetic methods, or sampling strategies), but no public model has yet been demonstrated to reliably achieve zero-shot control over wholly unseen property combinations. Finally, models that integrate richer structural context—such as using molecular graphs or fingerprint embeddings in addition to SMILES—are being explored. Early evidence suggests that these multimodal inputs can help propose chemically valid and synthetically accessible modifications under complex objectives, though again systematic evaluation under all multi-objective criteria is still emerging.

\textbf{Problems Solved by LLMs.} Thanks to LLM‐based models such as the Molecular Transformer, many routine reaction predictions are now essentially solved: medicinal chemists can predict likely metabolites and process chemists can foresee side products with high confidence \cite{schwaller2019molecular}. 

\textbf{Remaining Challenges.} However, challenges remain in handling multi‐step or “one‐pot” reactions where sequential transformations occur, since current one‐step models lack a mechanism to decompose complex cascades. Quantitative prediction of yields and selectivities is still out of reach, as these models only output the major product qualitatively. Additionally, out‐of‐domain reactions—those involving novel catalytic cycles or exotic reagents absent from training sets—often confound existing models \cite{Guo2023PredictiveChemistry}.

\textbf{Future Work.} Future LLMs for reaction prediction may incorporate mechanistic reasoning, internally decomposing reactions into elementary steps akin to chain‐of‐thought prompting \cite{Wei2022ChainOfThought}. There is growing interest in multi‐modal architectures that integrate text with molecular graphs or images, enabling a richer understanding of bond connectivity changes. Enhancing uncertainty quantification and explainability—such as highlighting which bonds form or break—will empower chemists to assess prediction confidence. Finally, embedding reaction prediction LLMs within autonomous laboratory systems could enable closed‐loop experimentation, where AI proposes, executes, and learns from chemical reactions in real time. Future retrosynthesis LLMs will likely integrate external knowledge bases indicating which building blocks are inexpensive or readily available, biasing suggestions toward practical routes. We also anticipate multi‑step planning architectures, where a higher‑level agent orchestrates sequential calls to a one‑step model, effectively planning entire synthetic routes. Finally, more interactive human–AI retrosynthesis tools may emerge, capable of asking clarifying questions or presenting alternative routes with pros and cons—transforming retrosynthesis from a static prediction task into a dynamic, collaborative design process.

\subsubsection{Chemical Text Mapping}

Chemical text mapping tasks take free‑form chemical text as input and output either discrete labels (classification) or continuous values (regression). \textbf{Chemical Text Mapping is fundamentally a Mol2Num task.} For example, in a document classification scenario, given the safety note “During the reaction, hydrogen gas evolved rapidly and ignited upon contact with air,” the model outputs the hazard label “flammable.” In a text‑based regression example, given the procedure description “1.0 g of reactant A yielded 0.8g of product B under standard conditions,” the model predicts a yield of 80\%. By automating the extraction of critical information—hazard classes, reaction types, success/failure flags, yields, rate constants, temperatures, solubilities, pK$_a$ values, and more—chemical text mapping dramatically reduces manual curation, accelerates the creation of structured databases for downstream modeling, and empowers chemists and students to query “How dangerous is this step?” or “How much product did they actually get?” at scale. 

Common tasks of this form include hazard classification, reaction‑type classification, procedure outcome classification, yield and rate constant regression, temperature and solubility prediction, etc. In this work, we concentrate on chemical text mining within the chemical text mapping framework—harnessing LLMs to transform narrative chemical descriptions into actionable categorical and numerical data.

\textbf{Chemical Text Classification.} Chemical Text Classification is the task of categorizing chemical documents or text segments into predefined labels—such as reaction type, property mentions, or entity categories—based on their unstructured textual content.

Chemical text classification has matured through successive generations of chemistry-tuned LLMs, each addressing the gaps of its predecessors. ChemBERTa-2 \cite{ahmad2022chemberta} was one of the first truly chemical-centric encoders—pretrained on 77 million SMILES strings (from PubChem) using masked-language modelling and multi-task regression—and it showed competitive performance on multiple downstream molecular property and classification benchmarks. However, as an encoder-only model, it requires separate fine-tuning heads for each task and lacks built-in generative or structured-output capabilities.

Recent studies, such as “Fine-tuning large language models for chemical text mining” \cite{zhang2024fine}, have explored unified frameworks that handle multiple chemical text mining tasks—compound entity recognition, reaction role labeling, MOF synthesis information extraction, NMR data extraction, and conversion of reaction paragraphs to action sequences. In these works, fine-tuned LLMs demonstrated exact match / classification accuracies in the range of approximately 69\% to 95\% across these tasks with only minimal annotated data \cite{zhang2024fine}. Nevertheless, challenges remain in cross-sentence relations, complex numeric extractions, and the consistency / validation of structured or JSON-style output formats.

\textbf{Problems Solved by LLMs.} Firstly, they can achieve high-precision tasks such as hazard classification, reaction type annotation, yield and rate regression, with minimal or even no labeled data, through prompt engineering or a small number of examples, greatly reducing the cost of manually building domain dictionaries and rules; secondly, LLMs, with their powerful ability to understand context, can handle multiple information extraction subtasks (such as named entity recognition, relation extraction, and numerical prediction) simultaneously, thereby integrating the originally scattered pipeline processes into a unified end-to-end model; thirdly, combined with retrieval enhancement and chaining thinking technologies, LLMs also show robustness in long documents and cross-sentence dependency scenarios, laying the foundation for the automated construction of large-scale structured databases. 

\textbf{Remaining Challenges.} However, there are still several lingering issues that need to be addressed: LLMs sometimes generate confident but incorrect predictions (hallucination phenomenon), and their adaptability to extremely professional or the latest literature is limited; long text processing is limited by the size of the context window, making it difficult to complete global information integration across chapters or documents; in addition, support for multi-level nested entities and complex chemical ontologies is still not perfect. 

\textbf{Future Work.} Future work can focus on introducing dynamic knowledge retrieval and knowledge graph fusion to build a continuously updated domain memory; exploring multimodal extraction (such as graph, spectrum, and text joint understanding); and combining uncertainty estimation and active learning strategies to improve the reliability and interpretability of the model, ultimately achieving a fully automated pipeline from laboratory notes to enterprise-level chemical knowledge bases.

\subsubsection{Property‐Directed Chemical Design} 

In property‐directed inverse design, one begins with a set of target property criteria (for example, minimum thresholds for cell permeability, binding affinity, or solubility), optional domain priors encoded by pretrained generative LLMs, and a synthesizability filter to ensure practical feasibility. A LLM then directly generates candidate molecular structures—expressed as SMILES strings or molecular graphs—that are chemically valid, synthetically accessible, and predicted to meet the specified property requirements. \textbf{Property‐Directed Chemical Design is fundamentally a Text2Mol task.} An everyday analogy might be describing a flavor or recipe you want (“something that tastes like chocolate but is spicier”) and having a chef create a new dish – here we describe a desired chemical property or scaffold, and the model devises a compound. The method’s objectives are to maximize compliance with target properties, promote novelty and diversity beyond natural‐product scaffolds, and guarantee synthesizability via rule‐based or learned retrosynthetic filters. Key constraints include chemical validity (proper valence and connectivity), synthesizability scores (e.g., predicted accessibility), and in-silico property feasibility. Analogous to random mutation screening but executed computationally at scale, only those molecules that satisfy both the validity/synthesizability filters and the predefined property thresholds are retained as viable candidates.

Key subtasks include conditional molecule generation, and text-conditioned molecule generation (where input is a description of desired properties or a prompt like “a molecule similar to morphine but non-addictive” and output is a new molecule suggestion). Another subtask is reaction-based text to molecule, such as “the product of acetone and benzaldehyde in aldol condensation” – where the input text implies a reaction and the output is the product structure. We will focus on (1) Conditional molecule generation and (2) De novo molecule generation, since these illustrate the spectrum from well-defined translation to open-ended generation.

\textbf{Conditional Molecule Generation.} Conditional molecule generation seeks to design novel compounds that satisfy user-specified criteria—whether a textual description, property targets, or structural constraints—directly in SMILES or 3D form. 

The earliest text-conditioned methods relied on prompt-based sampling: both MolReGPT \cite{li2024empowering}, which leverages GPT-3.5/4’s in-context few-shot learning to generate SMILES without any fine-tuning (albeit with variable chemical accuracy and prompt dependence), and Jablonka et al.’s GPT-3 adaptation \cite{jablonka2024leveraging}, which fine-tuned the base model via prompt prefixes to produce valid SMILES matching property labels, showed that off-the-shelf LLMs can be repurposed for prompt-based conditional generation. In contrast, MolT5 \cite{edwards2022translation} tackled text-to-SMILES translation via a T5 encoder–decoder fine-tuned on paired natural-language captions and SMILES, pioneering direct text-to-molecule mapping but without built-in multi-objective controls. To improve semantic alignment and diversity, TGM-DLM \cite{gong2024text} introduced a diffusion-based language model conditioned on text embeddings, yielding molecules that more faithfully match user descriptions at the cost of extra compute. Recognizing the need for scaffold specificity, SAFE-GPT \cite{noutahi2024gotta} adopted a fragment-based “SAFE” token representation, enforcing user-provided cores in each output while retaining peripheral variability. Extending conditioning into three dimensions, BindGPT \cite{zholus2025bindgpt} embeds protein pocket geometries alongside sequence tokens to perform 3D structure-conditioned generation, enabling de novo ligand design tailored to binding-site shapes but requiring specialized 3D inputs. Beyond text and structure, target- and context-specific generation has been advanced by cMolGPT \cite{wang2023cmolgpt}, which integrates protein–ligand embedding vectors into a MOSES-pretrained transformer to produce candidate libraries for EGFR, HTR1A, and S1PR1 with QSAR-predicted activities correlating at Pearson r > 0.75, and by PETrans \cite{wang2023petrans}, which couples a protein-sequence or 3D-pocket encoder with a SMILES decoder to generate ligands that respect detailed binding-site features. And ChemSpaceAL \cite{kyro2024chemspaceal} solves data-efficient, target-focused exploration by wrapping an uncertainty-driven active-learning acquisition loop around its transformer generator, iteratively sampling and scoring molecules against protein profiles to drastically reduce the need for large annotated inhibitor datasets while still uncovering high-affinity candidates.

\textbf{De novo Molecule Generation.} Here the task is: generate a molecule that fits a given textual description. This description could be very general (“a potent opioid painkiller that is less addictive”) or very specific (“a molecule with an ether linkage, and a molecular weight under 300 Da”). This is one of the holy grails of AI in drug discovery – allowing scientists to simply specify desired qualities and having the AI propose novel structures that meet them. It’s akin to an artist drawing a creature based on a myth description, except here it’s a chemist “drawing” a molecule based on a target profile. It could drastically speed up the brainstorming phase of drug design or materials design. Instead of manually tweaking structures, a researcher could ask the model for ideas: “Give me a drug-like molecule that binds to the serotonin receptor but doesn’t cross the blood-brain barrier” – a very high-level goal – and get some starting points. 

Early work in de novo molecular design leveraged Adilov’s  “Generative Pretraining from Molecules” \cite{adilov2021generative} , which adapted a GPT-2–style causal transformer to learn SMILES syntax via self-supervised pretraining and introduced adapter modules between attention blocks for minimal‐change fine-tuning. This approach provided a resource-efficient generative backbone for both molecule creation and downstream property prediction. Scaling up, MolGPT \cite{bagal2021molgpt} implemented a 6 million-parameter decoder-only model with masked self-attention to capture long-range SMILES dependencies, enforce valency and ring-closure rules for high-quality, chemically valid generation, and employ salience measures for token-level interpretability. MolGPT outperformed VAE-based baselines on MOSES and GuacaMol by metrics such as validity, uniqueness, Frechet ChemNet Distance, and KL divergence.

To better model global string context, Haroon et al. \cite{haroon2023generative} added relative attention heads to their GPT architecture, tackling the long-range dependency challenge and boosting validity, uniqueness, and novelty. ChemGPT \cite{frey2023neural} then systematically explored hyperparameter tuning and dataset scaling, revealing how pretraining corpus size and domain specificity drive generative performance. Subsequent work by Wang et al. further refined architectures and training strategies to surpass MolGPT benchmarks in de novo tasks. Departing from SMILES, Mao et al.’s iupacGPT \cite{mao2025iupac} trained on IUPAC name sequences using SELFIES masking and adapters, producing human-interpretable outputs that align with chemists’ naming conventions and streamline validation, classification, and regression workflows. GraphT5 \cite{kim2025grapht5} first introduced multi‑modal cross‑token attention between 2D molecular graphs and text, enabling text‑conditioned graph generation but lacking explicit control over scaffold or property constraints . MolCA \cite{liu2023molca} added uni‑modal adapters and a cross‑modal projector to improve robustness across representations, yet remained confined to 2D structures and did not follow complex textual instructions reliably. 

To capture spatial information, 3D‑MoLM \cite{li2024towards} incorporated 3D molecular coordinates alongside text, allowing generation of conformers matching optical or binding‑site descriptions, but it struggled with scaffold fidelity and multi‑objective trade‑offs. UTGDiff \cite{xiang2024instruction} addressed instruction fidelity by using a unified text‑graph diffusion transformer that follows detailed prompts for substructure and property constraints. Addressing chirality, Yoshikai et al. \cite{yoshikai2024novel} coupled a transformer with a VAE and used contrastive learning from NLP to generate multiple SMILES representations per molecule—enhancing molecular novelty and validity while capturing stereochemical information. AutoMolDesigner \cite{shen2024automoldesigner} wrapped de novo generation into an open-source pipeline for small-molecule antibiotic design, emphasizing domain-specific automation with heuristic filters and reaction-feasibility checks. Taiga \cite{mazuz2023molecule} introduced a two-stage approach—unsupervised SMILES $\rightarrow$ latent mapping followed by REINFORCE-based fine-tuning on metrics like QED, IC$_50$, and anticancer activity—to achieve property-optimized design via reinforcement learning. Finally, cMolGPT \cite{wang2023cmolgpt} demonstrated flexible mode switching, operating unconditionally to explore chemical space de novo and switching seamlessly to conditional, target-focused generation under the same architecture, thus unifying both paradigms in a single LLM framework.

\textbf{Problems Solved by LLMs.} LLMs have demonstrated that there exists a learnable mapping from natural language to chemical structures, allowing chemists to “draw” molecules with words instead of manually constructing SMILES strings. For example, MolT5—jointly trained on text and 100 million SMILES—can generate precisely “COc1ccccc1” in response to “Give me a molecule containing a phenyl ring, an ether linkage, and a molecular weight under 300Da.” MolReGPT goes further: using ChatGPT’s few‑shot prompting, it can output valid candidate structures matching “phenyl ring + ether + MW<300” with no fine‑tuning required. This capability drastically lowers the design barrier—researchers need only describe desired features to obtain testable structures. And more importantly, it dramatically narrows the chemical search space, focusing billions of possible molecules down to hundreds or thousands of the most relevant candidates and thereby greatly accelerating discovery. Moreover, LLMs support real‑time, multi‑objective, and multi‑constraint generation—such as “increase hydrophilicity while retaining a hydrophobic ring” or “balance potency and synthetic accessibility”—and can explore chemical space rapidly even in low‑ or zero‑data scenarios. 

\textbf{Remaining Challenges.} The imagination of LLMs is limited by their training. If certain property correlations were never seen, the model might not know how to fulfill a prompt. Also, validity of generated molecules is a concern – models like ChemGPT when generating freely can sometimes produce invalid SMILES or chemically impossible structures (though less often as training improves). When guided by text, the risk of hallucinating a molecule that meets text but is chemically nonsensical is real. For example, an LLM might attempt to satisfy “nonflammable gas” and produce something like “XeH” – which is not a stable molecule, but fits the prompt superficially (xenon is nonflammable, but xenon hydride is not a thing). Ensuring chemical validity often requires adding a post‑check (like using cheminformatics software to validate and correct if needed). Another issue is the evaluation of success: if an AI generates a molecule from a prompt “potent opioid with less addiction,” how do we know it succeeded? We would have to test the molecule in silico or in lab. So often these models are used to generate candidates which are then fed into predictive models or experiments. It’s more of an ideation tool currently.

\textbf{Future Work.} We expect to see tighter integration of narrative‑guided chemical design generation with other models and databases. One likely scenario is an interactive system: the user gives a prompt, the AI generates a molecule, then another AI (or the same with a different prompt) evaluates the molecule’s properties and explains why it might or might not meet the criteria, then the user or an automated agent refines the prompt or adds constraints, and the cycle continues – essentially an AI‑driven design loop. Another direction is combining this with reinforcement learning or Bayesian optimization: use the text prompt to generate an initial population of molecules, then optimize them using property predictors (some recent work uses LLMs with in‑context learning to do Bayesian optimization for catalysts) \cite{Guo2023PredictiveChemistry}, hinting at possibilities of optimization within the model’s latent space. Also, as these generative models improve, one can imagine integrating hard constraints (like no toxic substructures or obey Lipinski’s rules for drug‑likeness) directly via prompt or via a filtering step in the generation process (some have tried using fragment‑based control tokens, e.g., telling the model “include a benzene ring” or “avoid nitro groups”). Another interesting future aspect is diversity vs. focus: LLMs might have a bias to generate molecules that are similar to what they know (so‑called mode collapse around familiar structures). Future models might include techniques to encourage more novelty (perhaps via lower sampling temperature or specialized training objectives) when novelty is desired. Conversely, if a very specific structure is needed, one might combine text prompt with a partial structure hint (like providing a scaffold SMILES and asking the model to complete it with substituents that confer certain properties).

\subsubsection{Chemical Knowledge Narration}
Chemical knowledge narration tasks take unstructured chemical text as input and produce another, more structured or user‐friendly text output. \textbf{It is fundamentally a Text2Text task.} For example, given the free‐form procedure “Add 5g of sodium hydroxide to 50mL of water, stir for 10 min, then slowly add 10g of benzaldehyde at $0^\circ C$,” a model can generate a standardized protocol: “1. Dissolve 5g NaOH in 50mL H\textsubscript{2}O. 2. Cool to $0^\circ C$. 3. Add 10g benzaldehyde dropwise over 5 min. 4. Stir at $0^\circ C$ for 10 min.” Likewise, from “The oxidation of cyclohexanol to cyclohexanone was performed using PCC in dichloromethane,” it can produce the concise summary “Cyclohexanol was oxidized to cyclohexanone with PCC in DCM,” and given the SMILES “CC(=O)OC1=CC=CC=C1C(=O)O” it can output the IUPAC name “2‐acetoxybenzoic acid (aspirin).” These transformations standardize and clarify experimental descriptions, automate nomenclature and summarization, and enable seamless integration into electronic lab notebooks, saving researchers hours of manual editing. 

Common chemical knowledge narration applications include protocol standardization, reaction summarization, SMILES-IUPAC conversion, literature summarization, question‐answer generation, and explanatory paraphrasing. In our work, we harness these capabilities for chemical knowledge QA, chemical text mining, and chemical education.

\textbf{Chemical Knowledge QA.} Chemical Knowledge QA is the task of answering natural-language queries about chemical concepts, reactions, and properties by retrieving and reasoning over relevant information from unstructured or structured chemical sources. 

Chemical knowledge QA first saw major gains with LlaSMol \cite{yu2024llasmol}, an instruction-tuned model using the SMolInstruct dataset (over three million samples, covering 14 chemistry tasks), which outperformed GPT-4 on several chemistry benchmarks such as SMILES-to-formula conversion and other canonical tasks \cite{yu2024llasmol}. Nevertheless, it remains bounded by its training data cutoff, and it does not explicitly handle visual structure figure input in its evaluated tasks. To fill in more visual reasoning ability, ChemVLM \cite{li2025chemvlm} introduces a multimodal model integrating chemical images and text, enabling it to perform tasks like chemical OCR, multimodal chemical reasoning, and molecule understanding from visual plus textual cues. It achieves competitive performance across these tasks \cite{li2025chemvlm}. However, like many static QA models, its update of chemical knowledge is limited by its training corpus, and occasional incorrect or incomplete answers remain a concern.

\textbf{Chemical Text Mining.} Chemical Text Mining is the task of extracting and structuring relevant chemical information—such as reactions, molecular properties, or entity relationships—from unstructured textual sources (e.g., scientific articles, patents, and lab reports).

Chemical text mining has seen clear advances with models such as LlaSMol \cite{yu2024llasmol}, which uses the SMolInstruct dataset (over three million samples across 14 chemistry tasks) to fine-tune open-source LLMs, achieving substantial improvements over general-purpose models \cite{yu2024llasmol}. More recently, the work “Fine-tuning Large Language Models for Chemical Text Mining” \cite{zhang2024fine} demonstrated that fine-tuned models (including ChatGPT, GPT-3.5-turbo, GPT-4, and open-source LLMs such as Llama2, Mistral, BART) can handle five extraction tasks—compound entity recognition, reaction role labeling, MOF synthesis information extraction, NMR data extraction, and conversion of reaction paragraphs to action sequences—with exact accuracy in the range of approximately 69\% to 95\% using minimal annotated data \cite{zhang2024fine}. There are also works which fine-tune pretrained LLMs (GPT-3, Llama-2) to jointly perform named entity recognition and relation extraction across sentences and paragraphs in materials chemistry texts, outputting structured or JSON-like formats with good performance in linking entities like dopants-host, MOF information, and general composition/phase/morphology/application extraction \cite{dagdelen2024structured}.

\textbf{Chemistry Education LLMs.} Chemistry education LLMs have evolved from generic chatbots to increasingly specialized, curriculum-aligned tutors. For instance, ChemLLM \cite{zhang2024chemllm} is among the first LLMs dedicated to chemistry: trained on chemical literature, benchmarks, and dialogue interactions, it can provide explanations, perform interpretation of core concepts, and respond to student queries in a dialogue style with reasonable domain accuracy. Another relevant study\cite{guo2023can}, establishes a benchmark comprising eight chemistry tasks (e.g.\ explanation, reasoning, formula derivation), showing that models like GPT-4, GPT-3.5, Davinci-003 etc. outperform generic LLMs in zero-shot and few-shot settings on many such tasks. A further perspective by Du et al. \cite{du2024large} discusses how LLMs can assist in lecture preparation, guiding students in wet-lab and computational activities, and re-thinking assessment styles, though it does not report a system that simultaneously generates new problems, offers misconception warnings, and supports dialogic tutoring. Each generation—from ChemLLM’s dialogue capability, to benchmark studies demonstrating explanation + reasoning, to educational perspectives exploring scalable assessment—shows how chemistry-tuned LLMs are gradually moving toward more capable teaching assistants, though fully interactive, curriculum-aligned AI tutors remain an open challenge.

\textbf{Problems Solved by LLMs.} LLMs have revolutionized chemical knowledge narration tasks by enabling end‑to‑end transformations—standardizing free‑form procedures, summarizing complex reaction descriptions, and converting between SMILES and IUPAC names—all in a single, unified model without the need for multiple specialized tools or extensive manual intervention. Their strong contextual understanding and few‑shot/in‑context learning capabilities allow them to adapt quickly to new tasks with minimal examples, dramatically cutting the time researchers spend editing protocols, writing summaries, or updating electronic lab notebooks. 

\textbf{Remaining Challenges.} LLMs still occasionally “hallucinate” confidently incorrect details, struggle with long documents that exceed their context windows, and lack robust mechanisms for handling deeply nested or multi‑step reaction descriptions. 

\textbf{Future Work.} Future work must therefore focus on integrating retrieval‑augmented generation to ground outputs in real literature, fusing text with experimental figures or spectra for more accurate multimodal summaries, incorporating chain‑of‑thought prompting to produce auditable reasoning paths, maintaining a dynamically updated chemical knowledge base to stay current with new findings, and developing specialized evaluation benchmarks for protocol standardization, reaction summarization, and nomenclature conversion. These advances will make LLMs even more reliable, explainable, and indispensable for chemical knowledge QA, text mining, and education.

\subsubsection{Benchmarks}
As summarized in Table~\ref{tab:chem_bench}, we have compiled a comprehensive list of datasets that have been employed across a broad range of chemistry tasks using LLMs. This table includes benchmarks for tasks such as molecular property prediction, reaction yield prediction, reaction type classification, reaction kinetics, molecule captioning, reaction product prediction, chemical synthesis planning, molecule tuning, conditional and de novo molecule generation, chemical knowledge question answering, chemical text mining, and chemical education.

Systematically cataloging current research benchmarks is essential to bridge the gap between generic language modeling advances and chemistry‐specific tasks. By unifying evaluation protocols—standardizing data splits, SMILES preprocessing, and task formulations—we ensure that performance comparisons are both fair and reproducible. Moreover, a holistic survey reveals not only the breadth of existing benchmarks (from molecular property regression and reaction‐type classification to molecule captioning and generative design) but also their inherent biases: most datasets focus on drug‐like organic molecules or patent reactions, while domains such as inorganic chemistry, environmental pollutants, and negative‐result reporting remain underrepresented. This comprehensive overview therefore lays the foundation for more rigorous model development and benchmarking, guiding researchers toward data curation and experimental designs that fully exploit LLM capabilities in chemical contexts.

In the sections that follow, we will select several of the most influential datasets from Table~\ref{tab:chem_bench} for in‐depth discussion. For each chosen dataset, we will describe its scope, annotation scheme, and typical use cases, and then survey the performance of representative LLMs on these benchmarks to elucidate current capabilities and remaining challenges.

\textbf{MoleculeNet.(Mol2Num)} MoleculeNet is a consolidated benchmark suite that currently bundles sixteen public datasets spanning quantum chemistry, physical chemistry, biophysics and physiology.  All tasks share a uniform, text--first layout: each row of the .csv file starts with a canonical SMILES string followed by one or more property labels---binary for classification (e.g. BBBP, BACE, HIV, Tox21, SIDER) or floating--point for regression (ESOL, FreeSolv, Lipophilicity, the QM series).  Where 3‑D information is required, companion .sdf or NumPy archives store Cartesian coordinates and energies.  Official JSON split files define random, scaffold and temporal partitions so that every study can reproduce the same train/validation/test folds.  A typical classification row from BACE reads \verb|CC1=C(C2=C(N1)C(=O)N(C(=O)N2)Cc3ccccc3)O,1|, the trailing "1" indicating an active \(\beta\)-secretase inhibitor.  In the regression task ESOL a sample line might be \verb|OC1=CC=CC=C1,-2.16|, pairing a phenol SMILES with its experimental log‑solubility in mol\,L\(^{-1}\).
\input{tables/property_benchmark}

\input{tables/chem_bench}



\textbf{USPTO.(Mol2Num, Mol2Mol)} Starting from Daniel Lowe’s 1.8‑million raw patent dump—which stores un‑mapped SMILES triples like \texttt{O=C=O.OCCN>>O=C(O)NCCO}—researchers have carved out several task‑specific subsets.  The \emph{USPTO‑MIT forward‑prediction split} keeps 479035 atom‑mapped reactions and is the de‑facto benchmark for single‑step product prediction, where the input string \texttt{CC(=O)Cl.O=C(O)c1ccccc1>>?} asks the model to regenerate the ester product. \emph{USPTO‑50K} retains 50014 lines and appends a ten‑class label, enabling reaction‑type classification exemplified by \texttt{CCBr.CC(=O)O>[base]>CCOC(=O)C,7}, where “7” represents an acylation class.  Building on the same patent source, \emph{USPTO‑Yield} merges textual yield phrases so that a row such as \texttt{C1=CC=CC=C1Br.CC(=O)O>[Cu]>C1=CC=CC=C1CO,72} allows numeric yield regression, while \emph{USPTO‑Stereo} preserves wedge‑bond chirality, demanding stereochemically exact output for inputs like \texttt{C[C@H](Cl)Br.O>>?}.  Beyond single steps, \emph{PaRoutes} links ~450000 Lowe reactions into multistep route graphs so a model must recreate the full path ending in \texttt{C1=CC=C(C=C1)C(=O)O} rather than just its terminal disconnection.  Finally, \emph{ORDerly} re‑formats selected USPTO lines into Open‑Reaction‑Database JSON with timestamped splits—each entry like \texttt{"inputs":[{"smiles":"CCOC(=O)Cl"}],"outputs":[{"smiles":"CCOC(=O)O"}]}, "temperature": 298—so that forward prediction, condition inference and genuine time‑splitting can be assessed simultaneously.  Together these sub‑corpora let large chemical language models be probed across product generation, type classification, yield regression, stereochemical accuracy and multistep planning without ever leaving the USPTO domain.

\textbf{ChEBI‑20.(Mol2Text)} ChEBI‑20 is a medium‑sized molecular–caption corpus that links 33,010 small‑molecule SMILES strings to concise English sentences distilled from the ChEBI ontology.  The data are released as a UTF‑8 CSV whose first column stores the canonical SMILES and whose second column holds the free‑text caption; a third column specifies the standard 8:1:1 train/validation/test split.  Because each record couples structure and language, the collection naturally supports molecule‑to‑text caption generation, text‑to‑molecule retrieval and cross‑modal representation learning.  In the captioning task a model receives the input \verb|CC(C)OC(=O)C1=CC=CC=C1C(=O)O| and is expected to output a sentence such as “Ibuprofen is a propionic acid derivative with an isobutyl side chain and an aromatic core.”  In the inverse retrieval task the same caption is fed to the system, which must rank the correct SMILES ahead of thousands of distractors; the ground‑truth pair above therefore serves both roles without modification.
\input{tables/ChEBI_bench}

\textbf{USPTO-MIT.(Mol2Mol, Mol2Num)} The USPTO-MIT dataset, curated by MIT from Lowe’s extraction of original USPTO patent reactions and cleaned through atom mapping validation and SMILES normalization, comprises approximately 470,000 single-step forward reaction records split into training, validation, and test sets of 409,035, 30,000, and 40,000 examples respectively; each record encodes atom-mapped reactants, reagents, and products in SMILES (for example, CC(=O)O.CCCO>>CCCOC(=O)C denotes acetic acid and propanol yielding propyl acetate), and these high-quality, atom-mapped reactions support a variety of AI-driven chemistry tasks such as forward reaction prediction, single-step retrosynthesis, reaction classification, template extraction with atom mapping, and reagent prediction.
\input{tables/USPTO-MIT-forward_bench}

\textbf{GuacaMol.(Mol2Mol)} GuacaMol is an open-source de novo molecular design benchmarking suite built from approximately 1.8 million deduplicated SMILES strings standardized from the ChEMBL database. The construction pipeline includes salt removal, charge normalization, element filtering (retaining only H, B, C, N, O, F, Si, P, S, Cl, Se, Br, I), truncation to under 100 characters, and removal of any compounds overly similar to the hold-out set(holdout\_set\_gcm\_v1.smiles). GuacaMol defines 20 goal-directed tasks—ranging from simple property optimization (e.g., log P, TPSA) and rediscovery of known drugs to similarity-guided generation and scaffold-hopping—and, most centrally, molecule tuning multi-objective optimization tasks. These tuning tasks challenge models to perform fine-grained adjustments against scoring functions like QED, log P, and synthetic accessibility rather than merely reproducing the training distribution. For example, in the Cobimetinib multi-objective tuning task, generative models apply Pareto optimization strategies (such as NSGA-II or NSGA-III) to iteratively modify Cobimetinib’s SMILES substituents, maximizing a weighted combination of drug-likeness (QED) and solubility (log S) scores to produce novel candidates balanced across multiple property dimensions. This emphasis on molecule tuning not only tests a model’s ability to replicate known chemical spaces but also measures its practical value in accelerating lead optimization during early drug discovery by finely balancing multiple molecular properties.

GuacaMol not only supports goal-directed multi‐property tuning tasks, but also provides two key generative scenarios: conditional molecule generation and de novo generation. In conditional generation, models must produce compounds that satisfy user‐specified property or scaffold constraints. For example, MolGPT achieves strong control over QED and log P in GuacaMol’s conditional benchmarks, attaining validity $\approx$0.98, high uniqueness, and novelty close to 1.000, while cMolGPT extends these approaches by prepending target property values to the input, enabling precise conditional generation. More recently, LigGPT introduces flexible multi-constraint conditioning, allowing a single model to balance multiple property targets while retaining synthesizability and validity.

In the de novo setting, GuacaMol evaluates models on validity, uniqueness, novelty, Fréchet ChemNet Distance (FCD), and KL divergence. Here, MolGPT achieves validity 0.981, uniqueness 0.998, and novelty 1.000, LigGPT improves further with validity 0.986 and novelty 1.000, and SELF-BART excels on distributional similarity metrics such as FCD and KL divergence. Graph-based masked generation approaches also show competitive performance on these benchmarks, highlighting the impact of molecular representation on generation quality. Early generative frameworks such as ChemGAN and Entangled Conditional AAE serve as important references in the distribution-learning tasks, helping the community understand the strengths and limits of deep learning methods in exploring chemical space. Together, GuacaMol’s conditional and de novo tasks offer a comprehensive, rigorous testbed for chemical LLMs, driving continued innovation in model architectures and training strategies.
\begin{table}[!h]
    \centering
    \small
    \caption{De novo molecule generation performance on GuacaMol for chemical-domain LLMs. Best per metric in \colorbox{Blue4}{blue}.}
    \label{tab:guacamol-denovo-llms}
    \resizebox{0.5\linewidth}{!}{
    \begin{tabular}{lccc}
        \toprule
        \textbf{Model} & \textbf{Validity↑} & \textbf{Uniqueness↑} & \textbf{Novelty↑} \\
        \midrule
        MolGPT & 0.981 & 0.998 & \cellcolor{Blue4}1.000 \\
        LigGPT & 0.986 & 0.998 & \cellcolor{Blue4}1.000 \\
        SmileyLlama & 0.958 & \cellcolor{Blue4}1.000 &  0.987 \\
        \bottomrule
    \end{tabular}
    }
\end{table}

\textbf{MOSES.(Text2Mol)} MOSES is a molecular generation benchmarking platform introduced by Polykovskiy et al. in 2020 in Frontiers in Pharmacology, derived from 4,591,276 SMILES in the ZINC Clean Leads collection and filtered by molecular weight (250–350 Da), rotatable bonds ($\leq$ 7), XlogP ($\leq$ 3.5), removal of charged/ non-C/N/S/O/F/Cl/Br/H atoms, PAINS, and medicinal chemistry filters to yield 1,936,962 drug-like molecules. These molecules are split into training ($\approx$ 1,584,664), test ($\approx$ 176,075), and scaffold-test ($\approx$ 176,089 with unique Bemis–Murcko scaffolds) sets to assess model performance on both seen and unseen scaffolds. MOSES supports de novo generation, where the CharRNN baseline achieves validity 0.975, uniqueness 0.999, novelty 0.842, IntDiv1 0.856 and IntDiv2 0.850 on the test set; and conditional generation, for example SELF-BART applies property-conditioned decoding to generate molecules with desired constraints, attaining validity 0.998, uniqueness 0.999, novelty 1.000, and strong internal diversity scores. Consequently, MOSES serves as a unified and rigorous benchmark for core tasks in molecular generation, spanning distribution-learning to property-driven generation.

\textbf{Summary.} Current benchmark datasets for LLMs in chemistry emphasize three main task categories. First, \textbf{molecular property prediction} dominates, with collections such as MoleculeNet and Therapeutics Data Commons providing numerous binary and regression targets (e.g.\ solubility, toxicity, binding affinity). Second, \textbf{reaction outcome and classification} tasks, primarily drawn from USPTO and Reaxys repositories, assess models on product prediction, reaction‐type labeling, and yield regression. Third, \textbf{molecule generation} benchmarks (MOSES, GuacaMol) evaluate de novo and condition-driven design by measuring validity, novelty, and property optimization. By contrast, complex tasks like multi-step synthesis planning, reaction condition optimization, and 3D conformer reasoning remain underrepresented.

Compared to traditional cheminformatics and rule-based methods, domain-trained transformer models have demonstrated quantitative gains. Here, we illustrate such improvements using three representative tasks that are widely applied: Property Prediction, Reaction Prediction and Classification, and Molecule Generation. \textbf{In property prediction}, specialized SMILES‐BERT variants outperform random forests on multiple ADMET assays by several percentage points. \textbf{In single-step reaction tasks}, sequence-to-sequence transformers surpass template-based systems, improving top-1 accuracy by 5–10\%. \textbf{In generative settings}, LLM‐based generators achieve near-perfect chemical validity (> 98 \%) and higher diversity metrics than earlier recurrent or graph-based approaches, while also enabling multi-objective optimization of drug-like properties with average improvements of 10–20 \% over heuristic baselines.

Despite encouraging results, current LLMs face important limitations when applied to chemistry domains. 

A core challenge is that \textbf{chemical data are highly structured (e.g. graphs), yet LLMs operate on linear token sequences.} This mismatch means that transformers struggle to natively represent molecular topology and 3D geometry. For instance, an LLM given a SMILES string has no direct encoding of the molecule’s shape or stereochemistry, which can be crucial for many properties. This leads to errors when tasks fundamentally depend on spatial or structural reasoning – a known example is that models predicting quantum chemistry properties from SMILES (instead of actual 3D coordinates) perform poorly and misrepresent the true task. Another limitation is the knowledge cutoff of LLMs and their tendency to hallucinate. Without explicit chemical rules, an LLM may propose an impossible reaction or a nonsensical molecule, especially if it hasn’t seen similar examples in training. Ensuring validity and consistency in outputs remains non-trivial; even with grammar-constrained decoding, models might violate subtle chemical constraints or overlook rare elements. 

\textbf{Data scarcity and bias are additional concerns}: many benchmark datasets are relatively small and biased toward drug-like molecules, so LLMs may generalize poorly to larger chemical space or unusual chemotypes. Researchers also report that LLM performance can be brittle – small changes in input format (SMILES vs another notation) or prompt phrasing can yield different results, reflecting an unstable understanding. From a practical standpoint, the resource requirements of large models pose a challenge: using cutting-edge LLMs (like GPT-4) can be orders of magnitude more costly and slower than using task-specific models. This makes it difficult for researchers to fine-tune or deploy the largest models on private data. 

\textbf{Finally, the interpretability of LLM decisions is limited} – unlike human chemists or simpler models that can point to a mechanistic rationale, a transformer’s prediction is hard to dissect, which can erode trust in sensitive applications (e.g. drug discovery). In summary, today’s chemical LLMs are constrained by input representations, data quality, model transparency, and computational cost, highlighting the need for new strategies to realize their full potential.

From these observations we derive three core insights. \textbf{First, dataset limitations in both scale and scope} constrain LLM performance: public repositories often contain errors, inconsistent annotations, and limited chemical diversity—popular benchmarks such as QM9 can be misused and fail to represent realistic molecular spaces. To overcome this, a community effort should curate larger, cleaner datasets by aggregating high‑quality experimental results (e.g. ADME assays, comprehensive reaction outcomes) and expanding initiatives like the Therapeutic Data Commons to include negative results and broader chemistries. Benchmarks must also incorporate crucial chemical information—3D conformations, stereochemistry, reaction conditions (catalysts, solvents, yields)—to foster deeper chemical reasoning beyond simple SMILES pattern matching. \textbf{Second, model and methodology strategies require refinement}. Treating chemical structures as linear text has merits but introduces tokenization and validity challenges, motivating exploration of alternative representations such as SELFIES or fragment‑based vocabularies. Generic LLMs lack embedded chemical rules (valence, aromaticity) and benefit from domain‑specific pretraining on extensive chemical corpora (molecules, patents, protocols). Hybrid architectures—integrating LLMs with graph neural networks, physics‑based modules, or explicit inter‑atomic distance matrices—can bridge the gap between sequence models and spatial structure. \textbf{Finally, improving reliability and usability demands thoughtful task formulation and validation}. Reformulating regression tasks as classification or ranking problems, developing chemistry‑specific prompting (few‑shot, chain‑of‑thought, multi‑step retrosynthesis prompts), and embedding chemical validation loops (reinforcement learning with validity rewards or critic models) can reduce hallucinations and ensure chemical soundness. Coupling LLMs with external tools (“LLM + tools” paradigms) and advancing interpretability—via attention analysis, attribution methods, and standardized evaluation metrics—will build trust and utility. In conclusion, converging richer data, smarter representations, hybrid modeling, and robust benchmark design will propel LLMs toward becoming reliable, powerful instruments for chemical research., the convergence of richer data, smarter representations, hybrid modeling strategies, and thoughtful benchmark design will help overcome current limitations and guide LLMs to become more reliable and powerful tools for chemical research.

\subsubsection{Discussion}

\noindent\textbf{Opportunities and Impact.}  
LLMs are becoming transformative tools in chemistry and chemical engineering, bridging traditional chemical methods with cutting‑edge computational advancements. 

In molecular textualization (Mol2Text), the traditional rule-based naming system relies on manually written chemical naming rules, which often find it difficult to cover all cases of novel or complex molecules, while LLMs can learn naming patterns from large-scale chemical corpora, achieving better generalization and robustness. Transformer models such as Struct2IUPAC achieve 98.9 \% correct SMILES-IUPAC conversions on a 100 k test set, halving the residual error (∼1 \% vs 3 \%) observed for the long-standing rule-based parser OPSIN on comparable benchmarks \cite{krasnov2021struct2iupac}.  For example, when researchers discover a new antibiotic with a highly unusual structure, these models can immediately generate its official chemical name, saving chemists hours or days of manual naming efforts.

In property prediction (Mol2Number), traditional methods require the training of independent models for each property, making it difficult to share underlying chemical knowledge; while LLMs have absorbed the correlations between properties during the pre-training phase, achieving one-time multi-task predictions. LLM-driven approaches like ChemBERTa unify this process—after training on vast molecular datasets, a single model can simultaneously predict properties like solubility (e.g., "will the compound dissolve easily in water?"), toxicity (e.g., "is the compound safe?"), and expected chemical yields. Scaling ChemBERTa-2 pre-training from 0.1 M to 10 M molecules lifts average ROC-AUC across the MoleculeNet suite by +0.11 (e.g., 0.67 to 0.78) without training separate models per property, whereas classic GCN/GAT baselines plateau near 0.70 \cite{ahmad2022chemberta}. For instance, pharmaceutical chemists designing new drugs can quickly assess multiple crucial drug properties simultaneously, significantly streamlining the early drug discovery phase. 

For complex reaction planning (Mol2Mol), traditional reaction prediction software largely relies on manual reaction templates, which are difficult to capture complex mechanisms; while LLM based on Transformer can accurately plan synthetic routes by learning long-range dependencies between steps through self-attention mechanisms. LLM-based models, such as those inspired by Reaction Transformer, can break complex reactions into simpler, understandable stages. For example, when developing a complex cancer treatment molecule, these models can accurately suggest each intermediate step in synthesis, increasing the success rate of predicted synthetic routes by 10-20\% compared to older rule-based methods. The Molecular Transformer attains 90.4 \% top-1 accuracy on USPTO product prediction, while the best contemporary template-driven system (RetroComposer) reaches only 65.9 \%, a 24-point jump that translates into far fewer manual template overrides during route design \cite{schwaller2019molecular}.  

In chemical text classification (Text2Num), traditional text mining often relies on manual rules or shallow features, making it difficult to handle context-dependent chemical descriptions; while LLMs can accurately extract reaction conditions and results through deep semantic understanding. For example, researchers fine-tuned LLaMA-2-7B on 100,000 USPTO reaction processes, enabling it to directly generate structured records that comply with the Open Reaction Database (ORD) architecture—the model achieved an overall message-level accuracy of 91.25\%, and a field-level accuracy of 92.25\%, capable of stably identifying key numerical information such as temperature, time, and yield \cite{ai2024extracting}. In comparison, the best feature-engineered CRF/SSVM patent-NER pipeline achieved an F-measure of only 88.9\% on the CHEMDNER-patents CEMP task, highlighting the significant performance improvement of LLM in chemical text information extraction \cite{zhang2016chemical}.

Inverse design (Text2Mol) benefits from LLMs’ ability to generate chemically valid candidates under user‑specified property constraints, reducing trial‑and‑error cycles from weeks to minutes and expanding exploration of novel chemical space \cite{bagal2021molgpt}. Traditional reverse engineering requires a large amount of trial and error and manual filtering, resulting in low efficiency; while LLM acquires distributed knowledge in the chemical space through pre-training and combines it with conditional generation, which can quickly output high-quality candidate molecules. With LLM-based generative models such as MolGPT \cite{bagal2021molgpt} and ChemGPT \cite{frey2023neural}, chemists can now simply describe the needed properties (e.g., "a molecule that lowers blood pressure but doesn't cause dizziness") and instantly receive hundreds of suitable, chemically viable molecule suggestions. This dramatically shortens the molecule discovery process from potentially weeks of trial and error down to just minutes. Conditional generators such as Adapt-cMolGPT now yield 100 \% syntactically valid molecules under SELFIES-based sampling, compared with 88–94 \% validity for earlier SMILES-based GPT decoders—eliminating one in ten invalid proposals and narrowing medicinal-chemist triage \cite{yoo2024adapt}.

Finally, in chemical text mining (Text2Text), traditional text mining often relies on manual rules or shallow features, making it difficult to handle context-dependent chemical descriptions; while LLMs can accurately extract reaction conditions and results through deep semantic understanding. For example, researchers fine-tuned LLaMA-2-7B on 100,000 USPTO reaction processes, enabling it to directly generate structured records that comply with the Open Reaction Database (ORD) architecture—the model achieved an overall message-level accuracy of 91.25\%, and a field-level accuracy of 92.25\%, capable of stably identifying key numerical information such as temperature, time, and yield \cite{ai2024extracting}. In comparison, the best feature-engineered CRF/SSVM patent-NER pipeline achieved an F-measure of only 88.9\% on the CHEMDNER-patents CEMP task, highlighting the significant performance improvement of LLM in chemical text information extraction \cite{zhang2016chemical}.  

\noindent\textbf{Challenges and Limitations.}  
Despite these advances, LLM‑driven chemical models still face critical hurdles. 

First, the “experimental validation gap” persists: Despite impressive predictive power, LLM-driven chemical models still require extensive laboratory verification before their results can be trusted for critical decisions. For example, an LLM acts like an experienced chef who can predict delicious recipes based on prior knowledge, but ultimately you still have to cook the dish yourself and taste it to confirm whether it's actually good \cite{tom2024self}. Without deeper integration with automated robotic experimental setups or closed-loop experimental cycles, this validation bottleneck remains a significant "last mile" barrier \cite{seifrid2022autonomous}.  

Second, LLMs lack explicit mechanistic reasoning. Current LLM models predominantly learn associative patterns from large-scale datasets, often lacking explicit chemical reasoning or mechanistic insights. Imagine a student who memorizes a vast number of math solutions without understanding the underlying principles; they will struggle when encountering slightly altered problems. Similarly, predicting complex, multi-step chemical reactions (e.g., radical-based cascades) demands a mechanistic understanding that pure memorization from data cannot reliably provide, leading to frequent mistakes in subtle yet critical chemical details and limiting industrial adoption \cite{dutta2024think,bran2025chemical}.

Third, LLMs struggle to generalize to novel and sparsely represented chemical spaces. LLMs heavily depend on the breadth and quality of their training data, causing them to perform inadequately in chemically novel scenarios or sparsely represented reaction classes. For example, an LLM trained mostly on common organic reactions is like a chef proficient in everyday home cooking who suddenly faces preparing sophisticated French cuisine; the unfamiliar ingredients and methods may lead to frequent mistakes. This limitation restricts their predictive reliability in cutting-edge research or niche industrial applications, where innovation frequently occurs \cite{han2024generalist}.

Fourth, chemical “hallucinations” remain problematic: Generative chemical models powered by LLMs often produce chemically invalid or practically unsynthesizable molecules, a phenomenon known as chemical "hallucination." For example, an LLM could resemble an imaginative but inexperienced architect designing visually appealing buildings that are impossible to construct due to practical limitations of materials and construction methods. Although integrating rule-based filters partially mitigates this issue, systematic validation approaches remain inadequate, undermining trust in their use for real-world synthesis planning \cite{le2024utilizing}. 

Lastly, domain‑specific fine‑tuning demands high‑quality annotated datasets, which are expensive or impractical to produce at scale for every niche subfield. Without robust few‑shot or low‑data learning methods, many specialized applications remain out of reach \cite{sutanto2024llm}.

\noindent\textbf{Research Directions.}  
\begin{itemize}[leftmargin=10pt]
  \item \textbf{Hybrid LLM–Mechanistic Frameworks.}  
    Combine LLMs with rule‑based and physics‑informed modules to integrate statistical language understanding with chemical theory.

  \item \textbf{Multimodal Chemical Representations.}  
    Develop architectures that jointly process SMILES, molecular graphs, and spectroscopic or crystallographic data to capture 3D and electronic structure.

  \item \textbf{Closed‑Loop Experimental Pipelines.}  
    Integrate LLM outputs into automated synthesis and analysis platforms, enabling rapid hypothesis testing and feedback‑driven model refinement.

  \item \textbf{Data‑Efficient Fine‑Tuning.}  
    Leverage transfer learning, few‑shot prompting, and synthetic augmentation of underrepresented reaction data to improve performance in sparse domains.

  \item \textbf{Explainability and Uncertainty Quantification.}  
    Incorporate attribution methods and probabilistic modeling to provide confidence metrics and mechanistic rationales alongside predictions.

  \item \textbf{Governance‑First Deployment.}  
    Establish standards for model validation, transparent reporting (model cards), and ethical guidelines to ensure responsible use in chemical research and industry.
\end{itemize}

\noindent\textbf{Conclusion.}  
LLMs have significantly reshaped workflows in chemical discovery and engineering, but realizing their full potential requires innovations that marry linguistic intelligence with chemical reasoning, robust validation workflows, and ethical governance. By pursuing hybrid, multimodal, and closed‑loop approaches, the community can overcome current limitations and drive the next wave of breakthroughs in chemical science and industrial application.

\subsection{Life Sciences and Bioengineering}

\subsubsection{Overview} 
\noindent\textbf{5.4.1.1 Introduction to Life Sciences} 

Life sciences refer to all branches of sciences that involve the scientific study of living organisms and life processes~\cite{blumenthal1997withholding,blumenthal1996relationships,louis1989entrepreneurs}. In other words, they encompasses fields like biology, medicine, and ecology that explore how organisms (from micro-organisms to plants and animals) live, grow, and interact. Life scientists seek to understand the structure and function of living things, from molecules inside cells up to entire ecosystems, and to discover the principles that govern life~\cite{5d397ee1275ded87f97c4f9f,6330320990e50fcafdbb4639}. In simple terms, life sciences are about studying living things (such as humans, animals, plants, and microbes) to learn how they work and affect each other and the environment~\cite{blumenthal1996relationships,lewis2013economic,fisher1976environment}. This knowledge not only satisfies human curiosity about nature but also underpins applications in health, agriculture, and environmental conservation.

Life sciences are vast domains, so their research tasks range from decoding genetic information to observing animal behavior. Traditionally, each type of task has relied on specific methods and tools developed over decades (or even centuries) of biological research~\cite{blumenthal1997withholding,blumenthal1996relationships}. Through its detailed subdivision into specific subfields, we have categorized research in life sciences based on investigations ranging from the microscopic to the macroscopic level ~\cite{wikipedia_life_sci} and summaries of the review topics written by life sciences scientists~\cite{international2001initial,lodish2008molecular,carpenter2023principles,bayliss1915principles,darwin2023origin}. The main research tasks and their classic approaches include:

\noindent\textbf{Deciphering Genetic Codes. }Understanding heredity and gene function has been a core task. Early geneticists used breeding experiments to infer how traits are inherited~\cite{581a4c800cf245bd0c5d1adb,correns1924gregor}. In the 20th century, methods like DNA extraction~\cite{hearn2010dna,di2007comparison,rohland2007comparison} and Sanger sequencing~\cite{crossley2020guidelines,sikkema2013targeted,beck2016systematic} enabled reading the genetic code, while PCR (polymerase chain reaction)~\cite{erlich1989pcr,zhu2020pcr} revolutionized gene analysis by allowing DNA amplification. Today, high-throughput genome sequencing~\cite{kircher2010high,lou2013high} and bioinformatics~\cite{baxevanis2020bioinformatics} are standard for genetic research.

\noindent\textbf{Studying Cells and Molecules. }A fundamental task is to uncover how cells and their components (proteins, nucleic acids, etc.) function. Biochemists use centrifugation~\cite{majekodunmi2015review,brakke1951density} and chromatography~\cite{smith2013chromatography} to separate molecules, and X-ray crystallography~\cite{smyth2000x,woolfson1997introduction} or NMR to determine molecular structures~\cite{schwieters2006using}. For instance, gel electrophoresis~\cite{hames1998gel} became a routine method to separate DNA or proteins by size, and later innovations like Western blots~\cite{kurien2006western} provided ways to detect specific molecules. These methods, combined with controlled experiments in test tubes, have traditionally powered discoveries in molecular and cell biology.

\noindent\textbf{Physiology and Medicine. }Life sciences also deals with whole organisms, how organ systems work and how to treat their maladies~\cite{lewin1996physiology,verkhratsky2018physiology,saba1970physiology,boron2016medical,hahnemann2005organon,castiglioni2019history,sigerist1987history}. Physiological experiments on model organisms (from fruit flies and mice to primates) have been crucial~\cite{verkhratsky2018physiology,saba1970physiology}. For example, testing organ function~\cite{van2019personalised} or disease processes~\cite{krishnamoorthy2006inflammation} often involves animal models where interventions can be done. In medicine, clinical observations and clinical trials (systematic testing of treatments in human volunteers) are standard for linking biological insights to health outcomes. Additionally, fields like immunology and neuroscience have developed specialized methods (e.g. antibody assays~\cite{yunginger2000quantitative,kricka1999human}, brain imaging~\cite{raichle2006brain}) to probe complex systems. Life scientists in these areas often use a combination of laboratory experiments, medical imaging (like MRI, X-rays), and longitudinal studies to unravel how the human body (and other organisms) maintain life and what goes wrong in diseases~\cite{kekelidze2013colorectal,schwartzman1993apoptosis}.

\noindent\textbf{Ecology and Evolution. }A broader task is understanding life at the population, species, or ecosystem level, how organisms interact with each other and their environment, and how life evolves over time~\cite{amaral2020ecology,polis1990ecology,odum1971fundamentals}. Field observations and experiments are the cornerstone of ecology – researchers might count and tag animals, survey plant growth, or manipulate environmental conditions in the wild~\cite{sopinka2015manipulating}. Long-term ecological research (e.g. observing climate effects on forests~\cite{burt2015seeing}) and paleontological methods~\cite{hammer2024paleontological} have illuminated evolutionary history. In evolution, apart from the fossil record, comparing DNA/protein sequences across species (made possible by sequencing methods)~\cite{pearson1997comparison,huang1996fast} is a modern approach, but historically, comparative anatomy and biogeography were used by Darwin and others to infer evolutionary relationships. Today, computational models and DNA analysis complement classical fieldwork to address ecological and evolutionary questions~\cite{peltz2011next}.

\begin{figure}
    \centering
    \includegraphics[width=0.99\linewidth]{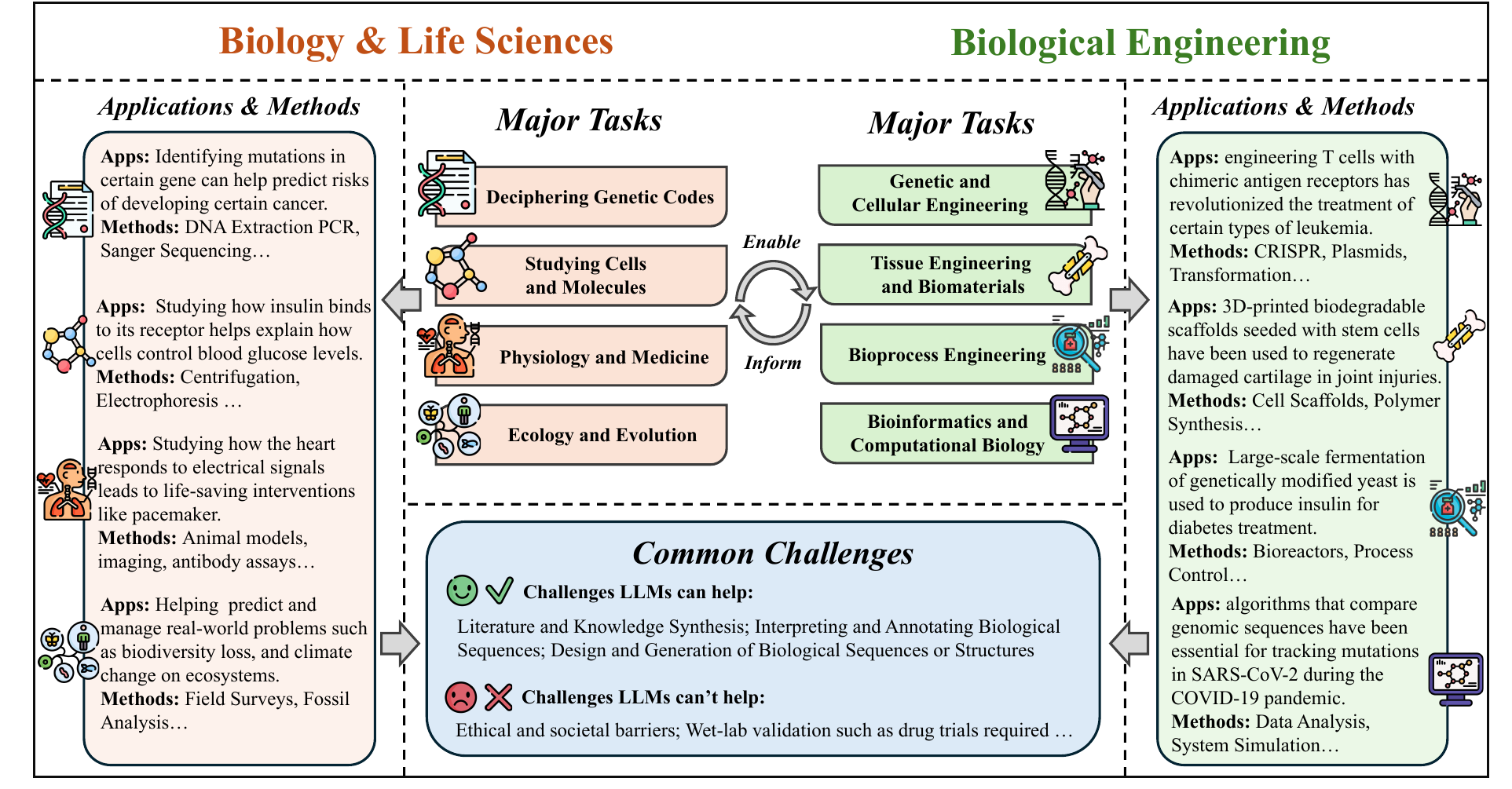}
    \caption{The relationships between major research tasks between biology and bio-engineering.}
    \label{fig:bio_diagram}
    \vspace{-10pt}
\end{figure}

\noindent\textbf{5.4.1.2 Introduction to Bioengineering} 

Bioengineering (also called biological engineering) is the application of biological principles and engineering tools to create usable, tangible, and economically viable products~\cite{5c1ed0553a55acc9dcffff7b,magin2004fractional}. Essentially, bioengineering leverages discoveries from life sciences by applying engineering design to develop technologies addressing challenges in biology, medicine, or other fields involving living systems~\cite{kumar2020bioengineering,jaklenec2012progress}. Put simply, bioengineering combines biological understanding with engineering expertise to design and build solutions, such as medical devices~\cite{camara2015security}, novel therapies~\cite{smolen2007new}, or biomaterials~\cite{ratner2004biomaterials,tathe2010brief}. It is inherently interdisciplinary: a bioengineer may employ mechanical engineering to construct artificial limbs, electrical engineering for biomedical sensors, chemical engineering for bioprocessing, and biological sciences across all applications~\cite{619bb9ed1c45e57ce96df7ab}. Thus, bioengineering bridges pure science with practical engineering, translating biological knowledge into innovations that enhance lives.

To better organize and understand bioengineering's scope, traditional research tasks are categorized into well-established domains. This classification is based on historical developments and practical engineering workflows~\cite{wikipedia_bio_e,Weidner2018Facilitation,Helms2009Biologically}, dividing the discipline according to how biological knowledge translates into engineering solutions and tangible products~\cite{Helms2009Biologically,Nagel2010Function-based}. Each category corresponds to a major application domain within bioengineering, representing distinct pathways for integrating biology with engineering.

\noindent\textbf{Genetic and Cellular Engineering.} Many bioengineers modify biological cells or molecules for new functions—such as engineering bacteria to produce pharmaceuticals or editing genes to treat diseases~\cite{nowakowski2013genetic,zhou2022engineered,lee2009metabolic,riglar2018engineering}. Genetic engineering techniques from molecular biology are foundational. Since the 1970s, recombinant DNA technology (using restriction enzymes to manipulate genes)~\cite{khan2016role,wright1986recombinant,johnson1983human,micklos1990dna} and cell transformation~\cite{rubio2005spontaneous} have enabled scientists to insert genes into organisms. Practically, bioengineers often use plasmids to introduce genes into bacteria and employ fermentation bioreactors (borrowed from chemical engineering) for cultivating genetically modified microbes at scale. More recently, CRISPR-Cas9 gene editing (developed in the 2010s) has allowed precise genome modifications~\cite{jiang2017crispr,redman2016crispr,doudna2014new,hsu2014development,ran2013genome}. Typical workflows include designing genetic constructs, altering cells, and scaling selected cell lines. This domain overlaps significantly with biotechnology and biomedical sciences~\cite{wikipedia_biology,kumar2020bioengineering}.

\noindent \textbf{Tissue Engineering and Biomaterials.} A significant bioengineering area involves engineering or replacing biological tissues using methods such as cultivating cartilage, skin, or organs in laboratories~\cite{ikada2006challenges,chapekar2000tissue,lanza2020principles,park2007biomaterials}. Core techniques include cell culture and scaffold fabrication—bioengineers cultivate cells on biodegradable scaffolds (often polymers) to form tissues~\cite{chapekar2000tissue}. Late 20th-century innovations demonstrated that seeding cells onto 3D scaffolds could produce artificial tissues (e.g., synthetic skin)~\cite{dar2002optimization,holy2000engineering}. Biomaterials science contributes materials (e.g., polymers, ceramics) designed to safely interact with the body, such as titanium or hydroxyapatite implants~\cite{ripamonti2012osteoinductive,de1989interface,cook1988hydroxyapatite}. This domain has produced synthetic skin grafts and advances toward lab-grown organs like bladders~\cite{ferber1999lab} and blood vessels~\cite{niklason2020bioengineered}.

\noindent \textbf{Bioprocess Engineering.} Bioengineers design processes to scale up biological products (e.g., mass-producing vaccines, biofuels, or fermented foods)~\cite{doran1995bioprocess,harun2010bioprocess,liu2020bioprocess}, drawing on chemical engineering principles adapted to biological contexts. Engineers design bioreactors, optimize conditions (temperature, pH, nutrients), and ensure sterile, efficient processes~\cite{Rio-Chanona2018Deep}. Traditional methods include continuous culture~\cite{herbert1956continuous,andrews1968mathematical} and process control systems. The large-scale production of penicillin in the 1940s exemplifies bioprocess engineering, involving optimizing Penicillium mold growth in industrial tanks~\cite{abraham1941further,gaynes2017discovery,ligon2004penicillin}. Today's production of monoclonal antibodies or industrial enzymes similarly employs refined classical fermentation and purification techniques~\cite{Shukla2010Recent,Yokoyama1999Production}.

\noindent \textbf{Bioinformatics and Computational Biology.} Although occasionally considered separate fields, bioengineers frequently engage in computational modeling of biological systems or analyze biological data (genomic or protein structures) to guide engineering designs~\cite{gentleman2005bioinformatics,tang2019recent,ranganathan2018encyclopedia,lio2003wavelets}. This domain involves algorithms and simulations—for example, modeling physiological systems using differential equations and computational methods from control theory. Computational approaches have long contributed to bioengineering, supporting prosthetic design optimization (via CAD and finite element analysis) and genomic analyses (software for DNA sequence analysis)~\cite{bhavikatti2005finite,szabo2021finite,taylor2014feap,marck1988dna,mckenna2010genome,lu20033dna}. This domain underscores bioengineering's combination of wet-lab experimentation and computational methods.

\noindent\textbf{5.4.1.3 Current Challenges} 

Life sciences and bioengineering are foundational disciplines that have transformed our understanding of life and significantly improved human health and well-being. Life sciences uncover the fundamental principles of biology, from Darwin’s theory of evolution and Mendel’s laws of inheritance to the germ theory of disease. These discoveries led to major advances such as vaccines, antibiotics like penicillin, and the molecular revolution sparked by the discovery of DNA’s double-helix structure. Tools like PCR and the Human Genome Project further deepened our ability to decode and manipulate genetic information, ushering in the era of personalized medicine.

Bioengineering complements these insights by applying them to solve real-world problems. The development of X-ray imaging allowed non-invasive diagnosis, while innovations like the implantable pacemaker and artificial organs expanded the scope of life-saving care. The production of human insulin through recombinant DNA technology marked a milestone in biopharmaceuticals. Later, tissue engineering demonstrated that lab-grown organs could be transplanted into humans, and gene-editing tools like CRISPR have opened new frontiers in treating genetic diseases.

Together, life sciences and bioengineering form a powerful synergy: the former provides deep biological insight, while the latter transforms that knowledge into tangible solutions. Their joint progress continues to revolutionize medicine, agriculture, and environmental science—improving quality of life and shaping a more advanced future.

Despite remarkable advancements in life sciences and bioengineering, numerous challenges persist due to the complexity and inter-connectivity inherent to biological systems. Intriguingly, many of the most formidable obstacles overlap between these fields, as they frequently address complementary facets of the same intricate biological phenomena. \textbf{In this section, we systematically examine several critical common challenges,} distinguishing between those that currently remain beyond the capability of artificial intelligence tools such as LLMs and those that can already benefit from LLM integration.

\noindent\textbf{Still Hard with LLMs: The Tough Problems.}

Here we analyze some key common challenges that are currently still beyond the reach of LLMs, we acknowledge LLMs' limitations related to experimental design, interpretative complexity, and practical hands-on tasks, underscoring domains where human expertise remains indispensable.

\begin{itemize}[leftmargin=10pt]
    \item \textbf{Ethical and Safety Challenges. }Life sciences and bioengineering are inherently intertwined with ethical and societal considerations that transcend purely technical challenges~\cite{golding1967ethical}. These fields routinely grapple with questions surrounding the responsible use of gene editing technologies like CRISPR~\cite{dzau2015responsible,jasanoff2015crispr}, the long-term ecological effects of genetically modified organisms~\cite{tiedje1989planned,snow2005genetically,wolfenbarger2000ecological}, and the protection of sensitive patient data in genomics and biomedical research~\cite{karagyaur2019ethical}. While LLMs can assist in synthesizing scientific literature and outlining stakeholder perspectives, they lack the capacity for moral reasoning or normative judgment. Their outputs are constrained by the biases present in their training data, which poses risks when applied to ethically sensitive domains~\cite{piergentili2021crispr,national2017human}. Ethical decision-making in bioengineering—such as determining the acceptability of human germline editing, setting standards for clinical trials, or regulating synthetic biology applications—remains the responsibility of human experts, policymakers, and the broader public. These decisions require inclusive debate, value alignment, and legal oversight that go beyond algorithmic capabilities~\cite{floridi2018ai4people,cath2018artificial}. As technologies advance, both the Life sciences and AI communities must collaboratively develop ethical frameworks that reflect societal values while fostering innovation.

    \item \textbf{Needs for Empirical Validation. }Both life sciences and bioengineering ultimately rely on physical experimentation~\cite{enderle2012introduction}. Scientific hypotheses must be empirically verified, and bioengineered solutions require testing under real-world conditions~\cite{rabinow2012designing}. However, such experiments often pose significant bottlenecks due to their inherent slowness, high costs, and ethical limitations—particularly in human studies, where experimentation is strictly constrained, and animal models often fail to fully replicate human biology.~\cite{williams2004use,mcgonigle2014animal,akhtar2015flaws} 
    While computational models can alleviate some of these burdens, they cannot fully substitute for wet-lab or clinical experiments. Similarly, LLMs are incapable of conducting physical experiments or collecting new empirical data. Although they can assist in designing experimental protocols, they cannot implement or validate them~\cite{li2014direct}. Consequently, research challenges that fundamentally require novel data acquisition—such as identifying new drug targets or evaluating biomaterials—remain beyond the scope of LLMs alone. The crucial \textbf{last mile} of validation in biology and engineering - demonstrating something works in actual living systems - remains dependent on laboratories, clinical trials, and real-world testing~\cite{berger2004last}.
    
    \item\textbf{Complexity of Biological Systems. }The overarching challenge is that living systems are astonishingly complex and multi-scale. Scientists struggle with this complexity as small changes can have cascading, unpredictable effects. For life scientists, this means incomplete understanding of many diseases and biological processes~\cite{menche2015uncovering,uversky2013decade,cho2012chapter}. For bioengineers, it means difficulty designing interventions without unintended consequences~\cite{manojlovich2016systematic,ni2021mitigating}. LLMs cannot reliably solve this because much of biological complexity stems from unknown factors requiring empirical observation and quantitative modeling beyond text-pattern recognition~\cite{riesselman2017deep}. While LLMs process information well, the emergent behavior of complex biological networks often requires specialized modeling that correlation-based systems can't provide without explicit mathematical frameworks. Major challenges like understanding neural circuits or curing cancer remain unsolved because they require new scientific discoveries and experimental validation, not just knowledge retrieval~\cite{mjolsness2019prospects}.
    
    \item \textbf{Data Quality and Integration. }Modern life sciences and bioengineering generate enormous volumes of data from genomic sequences, proteomics, patient records, and sensors. Making sense of this data reliably presents significant challenges because it's often noisy, comes from disparate sources, and lacks integration~\cite{boughorbel2018alternating,zhu2021hard}. While LLMs excel at processing text, they struggle with heterogeneous scientific data that includes numbers, images, and experimental measurements~\cite{zhao2023survey}. LLMs don't have native capabilities to process raw experimental data like gene expression matrices or medical imaging unless specifically augmented with specialized tools~\cite{pal2020big}. Challenges in biological big data - ensuring reproducibility, establishing causal relationships from observational data, or analyzing complex multi-modal datasets - still require specialized algorithms and human expertise in statistics and domain knowledge. LLMs might help report findings or suggest hypotheses, but they cannot replace the sophisticated analytical pipelines needed for rigorous scientific data analysis in these fields.
    
\end{itemize}

In summary, many of the grand challenges – decoding all the details of human biology, curing major diseases, sustainably engineering biology for the environment which are still open. LLMs, in their current state, are tools that can assist researchers but not solve these on their own, because the challenges often require new empirical discovery or involve complex systems and judgments beyond pattern recognition. An LLM might speed up literature review or suggest plausible theories, but it won’t automatically unravel the secrets of life that scientists themselves are still grasping at.

\noindent\textbf{Easier with LLMs: The Parts That Move.}

On a more optimistic note, there are challenges within life sciences and bioengineering where LLMs are already proving useful or have clear potential to contribute. These tend to be problems involving knowledge synthesis, pattern recognition in sequences/text, or generating hypotheses – tasks where handling language or symbolic representations is key. A few examples of such challenges that LLMs can tackle (and why they are suitable) include:

\begin{itemize}[leftmargin=10pt]
    \item \textbf{Literature Overload and Knowledge Synthesis.} One critical challenge uniquely pronounced in biology and bioengineering is managing the vast, rapidly growing, and fragmented body of research literature. \textit{Unlike fields such as mathematics, law, or finance, biological disciplines inherently encompass a multitude of interconnected subspecialties, each producing large volumes of highly specialized research.} For example, understanding a complex disease like cancer may require integrating findings from genetics, immunology, cell biology, pharmacology, and bioinformatics—each field publishing detailed, specialized studies that must be synthesized for comprehensive insights. The complexity arises not only from the sheer volume but also from interdisciplinary connections, intricate experimental details, and extensive supplementary materials required for reproducibility. Consequently, researchers face significant difficulty in identifying relevant literature, extracting key insights, and synthesizing knowledge efficiently. This is precisely where LLMs show strong potential as intelligent literature reviewers~\cite{agarwal2024litllm,an2024vitality,glickman2024ai}. Advanced LLMs, such as GPT-4, can proficiently read, summarize, and contextualize complex biomedical texts, rapidly extracting relevant findings from extensive corpora~\cite{mcginness2025highlighting,glickman2024ai}. For example, an LLM could swiftly provide researchers with an overview of current biomarkers for Alzheimer’s disease or consolidate recent advancements in biodegradable stent materials~\cite{glickman2024ai}. By effectively navigating dense technical language and complex sentence structures inherent to biological literature, LLMs mitigate literature overload, facilitate interdisciplinary integration, and enable literature-based discovery—highlighting connections between seemingly disparate research findings~\cite{baek2024researchagent,wu2024automated,matarazzo2025survey}.
    
    \item \textbf{Interpreting and Annotating Biological Sequences (Genomics/Proteomics). }In both life sciences research and bioengineering applications like synthetic biology, understanding DNA, RNA, and protein sequences is crucial~\cite{wang2025large,wu2025generator,zheng2023structure}. These sequences can be thought of as strings of letters (A, T, C, G for DNA; amino acids for proteins) – in other words, a language of life. Recent work has shown that language models can be applied to these biological sequences, treating them like natural language, where “words” are motifs or codons and “sentences” are genes or protein domains. This is a challenge where LLM-like models shine~\cite{nguyen2023hyenadna,ross2024chaining,xiao2024rna}. This means LLMs can help annotate genomes (predicting genes and their functions in a newly sequenced organism) or predict the effect of mutations (important for understanding genetic diseases)~\cite{xiao2024rna}. In proteomics, models can suggest which parts of a protein are important for its structure or activity~\cite{fang2022helixfold,truong2023poet}. The advantage of LLMs here is their ability to handle long-range dependencies in sequences – biology often has context-dependent effects, and language models are designed to handle such context. Moreover, LLMs can generate sequences too, which leads to the next point.
    
    \item \textbf{Design and Generation of Biological Sequences or Structures. }In bioengineering, a cutting-edge challenge is designing new biological components – for instance, designing a protein that catalyzes a desired chemical reaction, or an RNA molecule that can serve as a therapeutic. Traditionally, this is very hard (the search space of possible sequences is astronomically large). However, LLMs have a generative capability that can be harnessed here. Already, models like ProGen~\cite{madani2020progen,nijkamp2023progen2} have shown they can generate novel protein sequences that have a predictable function across protein families. In simpler terms, an LLM trained on a vast number of protein sequences can be prompted to create a new sequence that looks like, say, an enzyme, and those sequences have been experimentally verified in some cases to fold and function~\cite{madani2020progen,madani2023large,nijkamp2023progen2,elnaggar2021prottrans,brandes2022proteinbert}. This is a remarkable development because it means LLMs can assist in protein engineering and drug discovery by proposing candidate designs that humans or simpler algorithms might not think of. Similarly, for DNA/RNA, an LLM could suggest a DNA sequence that regulates gene expression in a certain way (useful for gene therapy designs)~\cite{wu2025generator,xiao2024rna} or propose improvements to a biosynthetic pathway by modifying enzyme sequences~\cite{fang2022helixfold,truong2023poet}. LLMs are suitable for these creative tasks because, much like with natural language, they can interpolate and extrapolate learned patterns to create new, coherent outputs (here, “coherent” means biologically plausible sequences). While any generated design still needs to be tested in the lab (to confirm it works as intended), LLMs can dramatically accelerate the ideation phase of bioengineering design.
    
\end{itemize}

\begin{figure}
    \centering
    \includegraphics[width=0.99\linewidth]{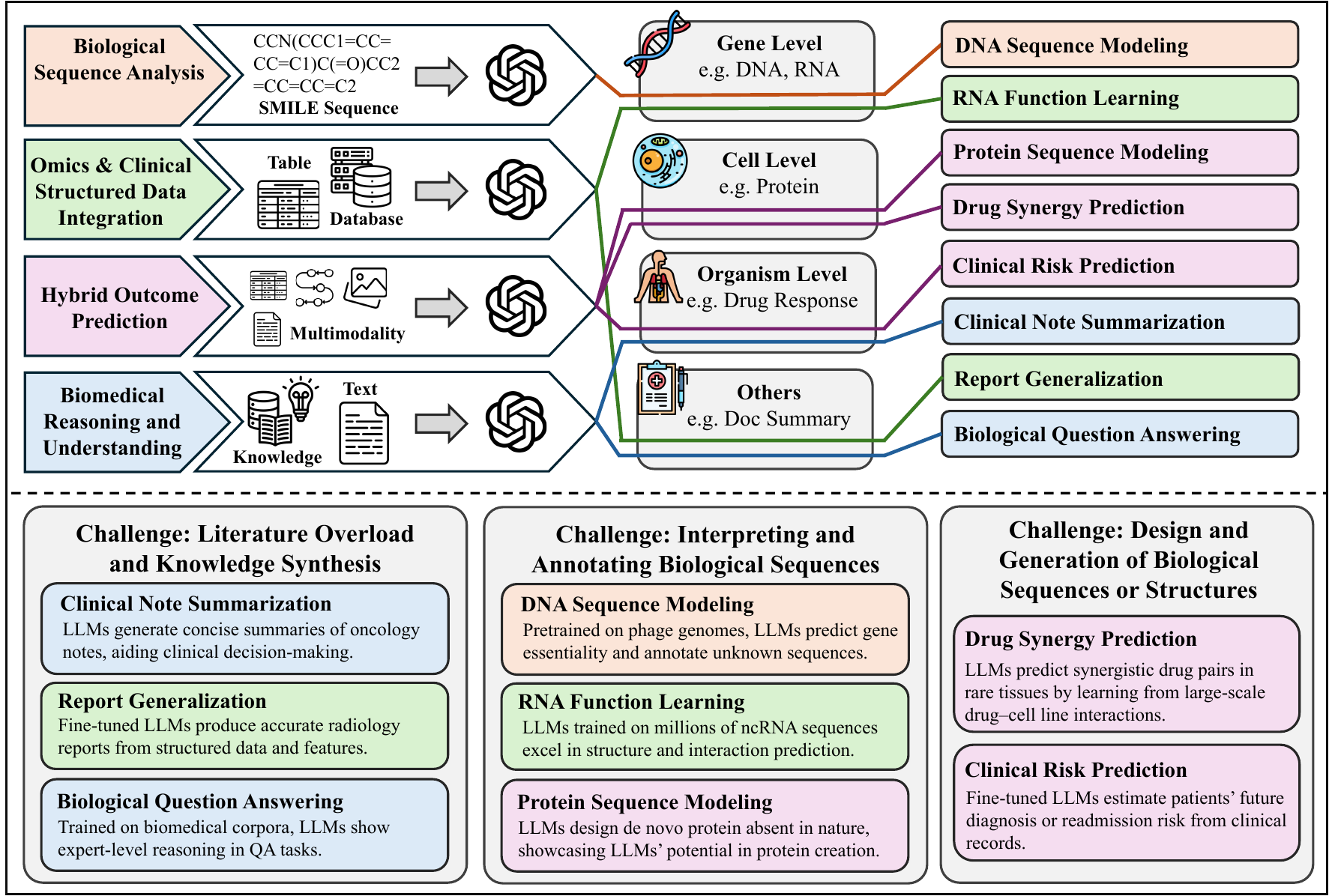}
    \caption{Our taxonomy for life sciences and bio-engineering.}
    \label{fig:bio_diagram}
    \vspace{-10pt}
\end{figure}

\noindent\textbf{5.4.1.4 Taxonomy}

Throughout the development of life sciences and bioengineering, research has gradually branched into increasingly specialized subfields. Nevertheless, many fundamental commonalities persist across domains. For instance, identifying potential folding targets within long protein sequences is methodologically similar to locating functional nucleotide fragments within DNA, both involve extracting biologically meaningful patterns from symbolic sequences~\cite{liu2024large,koonin2003principles,li2014direct}. In addition, biological tasks often span multiple levels and modalities, encompassing data from molecular to system scales and ranging from unstructured text to structured graphs~\cite{rajendran2024learning,zhang2025scientific}. This inherent diversity renders traditional task-type-based classifications insufficient, as they obscure the subset of tasks where LLMs are particularly well-suited.

To address this, we propose a taxonomy that emphasizes the computational characteristics of tasks, enabling a more precise alignment with the capabilities and limitations of LLMs. This data-centric perspective offers three key advantages:

\begin{itemize}[leftmargin=10pt]

\item{\textbf{Model compatibility}}: LLMs exhibit significantly different performance depending on input modality (e.g., excelling at natural language but struggling with numerical matrices)~\cite{de2024show,wang2024accurate,mirzadeh2024gsm}. A modality-aware taxonomy makes these alignments explicit.

\item{\textbf{Cross-domain transferability}}: Many life sciences tasks share structural similarities in their input data, making it easier to adapt and transfer methods across domains.

\item{\textbf{Support for multimodal integration}}: As biological data increasingly spans multiple modalities~\cite{rajendran2024learning,zhang2025scientific}, this taxonomy facilitates the design of composite pipelines, in which LLMs can serve diverse roles such as generation, orchestration, or interpretation.

\end{itemize}

\noindent\textbf{Sequence-Based Tasks.} These tasks involve analyzing sequential biological data, such as DNA, RNA, or protein sequences, which are essentially strings of nucleotides or amino acids. Typical examples include genome annotation, mutation impact prediction, protein structure prediction from sequences, and the design of genetic circuits. The input for these tasks generally comprises one or multiple biological sequences, with outputs including sequence annotations or newly designed sequences. From an artificial intelligence standpoint, these tasks are analogous to language processing problems because biological sequences possess a form of syntax (e.g., motifs and domains) and semantics (functional implications). LLMs and other sequence-based AI models are thus particularly effective for these applications, interpreting biological sequences similarly to languages. For instance, predicting the pathogenicity of a DNA mutation is akin to detecting grammatical errors in a language, where biological viability parallels grammatical correctness~\cite{yang2023dna}. Recent advancements, such as protein language models, exemplify successful AI applications leveraging this analogy~\cite{yang2023dna,zhang2023dnagpt}.

\noindent\textbf{Structured and Numeric Data Tasks.} These tasks involve handling structured datasets, numerical measurements, and graphs commonly encountered in physiology, biochemistry, and bioengineering. Examples include analyzing patient heart rate time-series data, interpreting metabolomics datasets, optimizing metabolic network models, and designing control systems for prosthetics. Inputs typically consist of numerical or tabular data (sometimes time-series), while outputs could involve predictions (e.g., forecasting patient adverse events) or control decisions. Such tasks traditionally rely on statistical methods or control theory, and they are generally less naturally suited for LLMs unless translated into textual or coded representations. A creative utilization of LLMs in this context is their ability to generate computational code from natural language descriptions, bridging descriptive problem statements to numeric analyses. For example, researchers can prompt an LLM to produce Python code to analyze specific datasets, utilizing the model as an intermediary tool between natural language and computational implementation~\cite{nascimento2024llm4ds,shen2025autoiot,maddigan2023chat2vis,nejjar2025llms}. However, purely numerical tasks involving complex calculations or optimizations typically remain better served by specialized algorithms.

\noindent\textbf{Textual Knowledge Tasks.} These tasks involve managing, interpreting, and generating text-based information prevalent in biology and engineering. Examples include literature searches and question-answering, proposal writing, extracting critical information from research articles, summarizing electronic health records, and analyzing biotechnology patents. Inputs here predominantly include unstructured textual documents (such as research articles, clinical notes, or patent filings), with outputs comprising synthesized textual summaries, detailed answers, or structured reports. This area represents the inherent strength of LLMs, encompassing various subtasks such as knowledge retrieval and question-answering, summarization and literature review, and protocol or technical report generation. Given their core competency in processing and synthesizing textual data, LLMs are exceptionally well-suited to these tasks.

\noindent\textbf{Predictive Modeling Tasks (Hybrid).} Many research questions in biology and bioengineering can be formulated as predictive problems, such as determining drug toxicity, predicting crop yields from genetic modifications, or forecasting protein folding success. These tasks frequently involve integrating multiple modalities—including sequences, structural data, textual descriptions—and require robust extrapolative capabilities. Inputs often combine diverse data formats, with outputs focusing on biological or engineering outcomes. Although many predictive tasks overlap with sequence-based analyses, this category explicitly emphasizes multidimensional predictions that leverage various feature sets. LLMs can contribute to these tasks through interpretive roles, qualitative reasoning, or as orchestrators within computational pipelines. For instance, an LLM could manage interactions between specialized bioinformatics tools, interpret computational outputs, and provide coherent explanations or qualitative predictions, highlighting their integrative and explanatory potential within complex predictive frameworks.

\input{tables/bio_llms}

\subsubsection{Genomic Sequence Analysis}
Sequence-based tasks focus on the learning and modeling of sequential data in life science and bioengineering, aiming to assist researchers in extracting meaningful biological insights from sequence-encoded information. The primary inputs to these tasks are biological sequences, such as nucleotide sequences in DNA/RNA or amino acid sequences in peptides. For instance, DNA is composed of four nucleotides, adenine (A), guanine (G), thymine (T), and cytosine (C), which are inherently stored in organisms in a sequential format, requiring no additional transformation. These biologically grounded representations allow models to learn structural patterns and latent dependencies, thereby enabling effective downstream prediction and classification tasks.

In sequence-based tasks, the objective is to leverage learned representations to accomplish specific downstream goals. For example, given genomic DNA adjacent to genes that may contain enhancers, a binary label is predicted for each 128-base-pair segment to determine whether it belongs to an enhancer region. Enhancers are short, non-coding DNA elements that regulate gene expression and can exert influence across distances of thousands to over a million base pairs by physically interacting with gene promoters. Another representative task is predicting the gene-editing efficiency of a given single guide RNA (sgRNA) sequence when guided by Cas proteins in CRISPR-based applications.

The value of sequence-based tasks in life sciences and bioengineering lies in their capacity to capture long-range dependencies and hierarchical patterns in extremely long and sparse sequences—often in a semi-supervised or unsupervised manner. For instance, in the human genome, protein-coding regions account for only about 1\% of the total DNA sequence. Likewise, functionally important non-coding regions such as promoters also constitute a very small fraction of the genome~\cite{mattick2022human}. By training on massive sequence datasets, models can learn to identify such regions with high accuracy. Researchers can then input unknown sequences into the model to annotate potential coding regions or predict functionally significant non-coding elements. In addition to classification, sequence-based models are also applied in predictive tasks. For example, they can estimate the likelihood and frequency of off-target mutations induced by CRISPR systems at unintended genomic loci, thereby improving the safety and efficacy of gene editing. In summary, sequence-based tasks significantly reduce the time and cost of sequence analysis, while deepening our understanding of the functional and regulatory roles encoded in biological sequences.

Although both DNA and RNA exist in the form of sequences, they differ significantly in biological function and modeling objectives~\cite{cheatham1997molecular}. DNA sequences primarily encode genetic information, and related modeling tasks focus on identifying regulatory elements such as promoters, enhancers, and transcription factor binding sites, as well as capturing long-range dependencies across the genome. In contrast, RNA is more involved in functional execution, including splicing, modification, translational regulation, and interactions with proteins. Typical tasks involve RNA secondary structure prediction, modification site identification, and functional classification of non-coding RNAs~\cite{ozsolak2011rna,stark2019rna}. Therefore, we categorize Sequence-Based Tasks into DNA Sequence Modeling and RNA Function Learning, and the subsequent discussion will be centered around these two directions. These two tasks share several common characteristics that make LLMs particularly well-suited for this domain. First, the abundance of unlabeled or sparsely labeled sequence data provides a rich resource for training. Second, LLMs not only deliver reliable predictions but also offer highly interpretable textual reasoning to support their outputs.

\textbf{DNA Sequence Modeling. }DNA Sequence Modeling refers to the task of computationally analyzing and learning patterns from nucleotide sequences—typically represented as strings composed of A, T, C, and G—to understand the underlying biological functions, regulatory mechanisms, and genetic variations encoded in the genome. In recent years, LLMs have rapidly emerged in the field of DNA, advancing our understanding of the information encoded within the genome. Transformer-like LLMs are now capable of reading, reasoning over, and even designing the 3.2Gb human genome, evolving from early convolutional neural networks (CNNs) that handled short windows to billion-parameter foundation models that process megabase-scale contexts. Early convolutional frameworks such as DeepSEA~\cite{zhou2015predicting} demonstrated that raw sequence alone is sufficient to predict chromatin features, inspiring a decade-long progression from hybrid CNN-RNN models and Transformer encoders to long-context state-space models. Today’s genomic LLMs often outperform traditional physics-based or motif-based methods across various tasks, including enhancer detection, cell type-specific expression, and non-coding variant effect prediction, while also providing saliency maps that highlight learned regulatory syntax.

DNABERT~\cite{ji2021dnabert} adapted the BERT architecture to human reference DNA by tokenizing sequences into 6-mers and using masked language modeling (MLM) to learn bidirectional representations, which proved transferable to promoter, enhancer, and splice site prediction tasks. Building on this, DNABERT-2~\cite{zhou2023dnabert2} introduced byte-pair encoding (BPE)~\cite{gage1994new}, breaking the fixed k-mer limitation, reducing memory usage by 30\%, and improving average MCC by 2 percentage points across 28 datasets. Techniques like self-distillation and adaptive masking further refined embeddings under limited data conditions.

Unlike typical LLMs, DNA inputs are significantly longer than those in standard NLP tasks. Even when models support large context windows, performance can degrade. To address these issues, Enformer~\cite{Avsec2021.04.07.438649} combines ConvNet~\cite{liu2022convnet} downsampling with 1D self-attention over 200kb regions, doubling the correlation with gene expression compared to prior models and significantly improving eQTL effect sign prediction. HyenaDNA~\cite{nguyen2023hyenadna} scales context length to 1 million nucleotides using sub-quadratic implicit convolutions, enabling 160× faster training than FlashAttention Transformers~\cite{dao2022flashattention}, while maintaining single-base resolution and outperforming Enformer on proximal promoter tasks.

GROVER~\cite{sanabria2024dna} uses frequency-balanced byte-pair encoding vocabularies to learn “sequence contextuality” directly from the human genome, outperforming prior k-mer baselines in next-token prediction and fine-tuning tasks like promoter detection and CTCF binding prediction. Its unsupervised embeddings can recover GC content, repeat categories, replication timing, and other functional signals purely from sequence, underscoring how tokenization design can unlock richer biological grammar.

With increased computational power and model scaling, larger models and training corpora have emerged. The Nucleotide Transformer~\cite{dalla2024nucleotide} with 2.5 billion parameters was pretrained on 3,202 human and 850 non-human genomes, generating general-purpose embeddings that improved performance across 18 downstream tasks—including pathogenicity scoring and enhancer–promoter linking—without requiring task-specific architectural changes. GenomeOcean~\cite{zhou2025genomeocean}, a 4-billion-parameter model, extended this paradigm to 220TB of metagenomic data, capturing rare microbial taxa for better ecologically driven generalization.

LLMs have substantially advanced DNA sequence modeling by enabling scalable interpretation of genomic sequences, capturing long-range dependencies, and learning regulatory syntax directly from raw nucleotide strings. They have addressed core challenges such as integrating context across megabase-scale windows, improving prediction of non-coding variant effects, and providing transferable embeddings for diverse downstream tasks without extensive feature engineering. However, key challenges remain: performance often degrades with increasing sequence length despite architectural innovations, and current models still struggle with integrating multi-omic signals, rare variant generalization, and interpretability in clinical settings. Future directions include developing more efficient architectures for ultra-long sequences, incorporating cross-modal biological data (e.g., epigenomic or transcriptomic layers), and aligning model predictions with mechanistic biological knowledge to support hypothesis generation and therapeutic discovery.

\textbf{RNA Function Learning. }RNA Function Learning refers to the task of modeling RNA sequences to uncover their structural attributes and functional roles, often leveraging sequence-structure relationships to predict biological behaviors and interactions. Understanding RNA sequences and their structural-functional relationships is essential for numerous molecular biology applications, including splicing regulation, RNA-protein interactions, and non-coding RNA functional annotation. Traditional bioinformatics methods such as sequence alignment and thermodynamic folding models (e.g., RNAfold) provide accurate predictions but suffer from limitations like heavy computational demands and dependency on handcrafted features.

Recently, leveraging advances in natural language processing (NLP), large-scale pretrained language models adapted to RNA sequences have emerged, significantly improving our capacity to interpret biological information embedded in nucleotide sequences. Early efforts, such as RNABERT~\cite{akiyama2022informative}, marked a shift toward learning biological grammar directly from data. RNABERT combined masked language modeling (MLM) with a structural alignment objective (SAL), enabling the model to internalize pairwise structural relationships by training on alignment scores derived from the Needleman-Wunsch algorithm~\cite{rnacentral2021rnacentral}.

Subsequent models expanded on this foundation by enhancing both scale and methodological complexity. RNA-FM~\cite{chen2022interpretable} significantly scaled the training dataset, employing 23.7 million deduplicated RNA sequences from RNAcentral~\cite{rnacentral2021rnacentral}, thus improving generalization capabilities for functional prediction tasks. RNA-MSM~\cite{zhang2024multiple} further advanced this approach by incorporating evolutionary context through homologous sequence modeling with multiple sequence alignments (MSAs), inspired by MSATransformer~\cite{rao2021msa}. Notably, RNA-MSM strategically excluded families with known structures during training, effectively reducing overfitting and enhancing performance on structure-aware tasks.

Parallel developments have addressed specific functional RNA elements, refining the modeling approach based on targeted biological contexts. For instance, SpliceBERT~\cite{chen2023self} specifically targeted splicing regulation by training on 2 million vertebrate pre-mRNA sequences from UCSC~\cite{haeussler2019ucsc}. By focusing explicitly on pre-mRNA rather than mature transcripts, SpliceBERT captured sequence features critical for identifying splicing junctions, splice sites, and regulatory motifs (e.g., exonic splicing enhancers or silencers), aspects typically overlooked in traditional modeling frameworks~\cite{marin2023bend}. Consequently, this model supports tasks such as splice site prediction, detection of alternative splicing events, and the identification of novel, tissue-specific regulatory elements.

Complementing this targeted functional perspective, 3UTRBERT~\cite{yang2024deciphering} specialized in modeling 3' untranslated regions (3'UTRs) to facilitate studies of post-transcriptional regulation. Building further upon the integration of structure into sequence modeling, UTR-LM~\cite{chu20245} explicitly incorporated structural supervision alongside MLM pretraining. It employed two biologically informed auxiliary tasks: secondary structure prediction and minimum free energy (MFE) regression. Secondary structures, predicted using the ViennaRNA toolkit~\cite{lorenz2011viennarna,gruber2008vienna}, were utilized as local structural constraints during masking, while global thermodynamic stability (MFE) values were predicted from global contextual embeddings ([CLS] token). The training dataset included carefully curated natural and synthetic 5'UTR sequences from databases like Ensembl and high-throughput assays, ensuring robust learning of biologically relevant patterns~\cite{cunningham2022ensembl,sample2019human,cao2021high}. These methods strengthened the link between structural and functional RNA predictions, demonstrating applicability in translation efficiency prediction and synthetic RNA design.

Lastly, BEACON-B and BEACON-B512 expanded RNA language modeling into extensive datasets comprising over 500,000 human non-coding RNAs from RNAcentral~\cite{rnacentral2021rnacentral}, exploring broader functional landscapes beyond coding transcripts~\cite{marin2023bend}. These models highlighted the importance of tailored training objectives, domain-specific masking strategies, and carefully curated datasets, all contributing to enhanced interpretability and biological accuracy.

LLMs have revolutionized RNA function learning by shifting the paradigm from alignment- or thermodynamics-based methods to data-driven, end-to-end models that learn both structural and functional features directly from sequences. These models have improved prediction accuracy for splicing patterns, RNA-protein interactions, and post-transcriptional regulatory elements, while also enabling interpretability through structural supervision and auxiliary tasks. Nonetheless, key challenges remain: many models still struggle with generalizing to novel RNA classes, integrating evolutionary and tertiary structural information, and explaining model decisions in biologically meaningful ways. Future research directions include scaling models to more diverse and comprehensive RNA datasets, incorporating multi-resolution structural priors, and aligning language model outputs with experimentally validated functional annotations to bridge the gap between sequence modeling and functional genomics.

\subsubsection{Clinical Structured Data Integration}
Clinical structured data integration focuses on the intelligent utilization of structured clinical information—such as Electronic Health Records (EHRs), laboratory test results, and coded diagnoses—to support and automate critical healthcare decision-making processes. The goal is to leverage artificial intelligence, particularly LLMs, to understand structured clinical inputs, build predictive or generative models, and produce meaningful outputs that improve clinical workflows, enhance patient care, and enable personalized medicine. These tasks primarily rely on structured or semi-structured datasets, including tabular EHR entries, time-series vital signs, coded diagnoses and treatments (e.g., ICD, CPT, LOINC), and structured questionnaire responses. Unlike free-form text, such data is inherently aligned with medical ontologies and clinical protocols, enabling models to reason with high factual precision and temporal awareness.

The primary objective of clinical structured data integration is to perform downstream tasks that generate clinically useful outputs based on structured patient data. For instance, given longitudinal EHRs containing timestamped diagnoses, prescriptions, and lab values, a model can forecast disease onset, stratify patient risk, or suggest treatment plans. In other scenarios, structured data is transformed into human-readable summaries—such as clinical progress notes or discharge instructions—to reduce the documentation burden on clinicians. A persistent challenge lies in bridging the gap between machine-readable formats and clinical narratives: generated content must be not only factually accurate but also contextually appropriate and linguistically coherent. Furthermore, since EHR data often contains missing values, noise, or institutional heterogeneity, models must be robust to irregular sampling and generalize across diverse healthcare settings.

In the broader context of life sciences and bioengineering, clinical structured data integration serves as a cornerstone of evidence-based medicine, offering scalable solutions for personalized care, automated documentation, and proactive health monitoring. By seamlessly connecting the structured backbone of clinical practice with the expressive power of language models, this work marks a critical step toward an intelligent, interoperable, and human-centered healthcare system.

To reflect the dual nature of this field, clinical structured data integration can be categorized into two main types: \textbf{Clinical Language Generation} and \textbf{EHR-Based Prediction}. The former focuses on converting structured clinical data into fluent, accurate natural language reports, enabling applications such as automated drafting of medical notes, radiology impression generation, and ICU event summarization. This task emphasizes controllable generation, temporal summarization, and medical factuality, requiring models to balance conciseness with informativeness. In contrast, EHR-based prediction aims to extract actionable insights from patient records to support tasks such as sepsis alerts, readmission prediction, and personalized risk scoring. These tasks demand strong temporal modeling, integration of clinical knowledge, and high interpretability, especially when informing critical medical decisions.

Across both task types, incorporating domain-specific inductive biases—such as hierarchical coding systems, medical knowledge graphs, or treatment ontologies—has been shown to enhance model performance. LLMs have demonstrated great potential in unifying diverse input modalities and producing clinically meaningful outputs, particularly when structured prompting or graph-aware architectures are employed. Moreover, the growing availability of publicly accessible, de-identified datasets such as MIMIC-III/IV~\cite{johnson2016mimic,johnson2023mimic} and eICU has fostered the development of standardized evaluation benchmarks. These advances not only enable rigorous comparison across methods but also promote the creation of generalizable and trustworthy AI systems for real-world clinical applications.

\textbf{Clinical Language Generation. }Clinical Language Generation refers to the use of natural language processing techniques to automatically produce coherent and clinically meaningful text from structured inputs such as electronic health records, diagnostic codes, or medical templates. Clinical Language Generation (CLG) is rapidly emerging as a foundational infrastructure in smart healthcare. By leveraging structured data or text recorded using templates, CLG models can automatically draft outpatient/inpatient notes, generate radiology report impressions, rewrite patient-friendly versions, and even transcribe real-time doctor-patient conversations. These capabilities significantly reduce the documentation burden on healthcare professionals, while improving the quality and readability of medical records, thereby supporting evidence-based decision-making and interdisciplinary collaboration.

With the maturation of large-scale pretraining corpora and instruction tuning techniques, CLG has evolved from early small-parameter models to multi-modal systems with tens of billions of parameters, offering unprecedented text generation capabilities in clinical settings.

One of the earliest representative works, ClinicalT5~\cite{lu2022clinicalt5}, adapted the T5~\cite{raffel2023exploringlimitstransferlearning} framework to hospital notes from datasets like MIMIC-III/IV~\cite{johnson2016mimic,johnson2023mimic}, pretraining on text-to-text tasks for long clinical narratives. It achieved a 3.1 ROUGE-L improvement on discharge summary generation and outperformed long-text baselines such as BART~\cite{lewis2019bart}, demonstrating that generative models can effectively capture key information in complex, structured medical records. However, ClinicalT5's training data was primarily composed of single-center English inpatient notes, which limits its generalization across languages and institutions.

In contrast, general-purpose LLMs are often trained on multilingual, multi-source datasets and inherently possess cross-domain generalization capabilities~\cite{radford2019language,brown2020language,achiam2023gpt}. With the advancement of such models, GPT-4~\cite{achiam2023gpt} and Med-PaLM 2~\cite{singhal2025toward}, through instruction tuning, can generate high-quality clinical drafts. GPT-4 achieved near-human accuracy in outpatient record analysis across three languages and can draft standardized clinical progress notes in zero-shot settings~\cite{schwieger2024large}. Med-PaLM 2 excelled in the MultiMedQA evaluation framework, particularly in reasoning and safety dimensions, showcasing the strength of large decoder-based models in long-form clinical text generation.

For patient communication, Jonah Zaretsky et al. demonstrated that LLMs can rewrite structured discharge summaries to a 6–8th grade reading level, with readability scores 18\% higher than physician-authored versions, greatly enhancing patient understanding of medication and follow-up instructions~\cite{zaretsky2024generative}. Meanwhile, model scale and multimodal capabilities are also advancing. Me-LLaMA~\cite{xie2024me}, built on LLaMA 2~\cite{touvron2023llama}, integrates PubMed, clinical guidelines, and knowledge graphs, and supports 13–70B parameter ranges. With medical instruction tuning, it enables multimodal prompt-based generation for case summaries and diagnostic explanations.

In fact, many models designed for clinical or medical tasks possess some degree of CLG capabilities. However, as many focus more on medical QA or comprehension tasks, we will introduce those models in detail in the following sections.

LLMs have transformed Clinical Language Generation (CLG) by enabling automatic, fluent synthesis of complex clinical narratives from structured inputs, thereby alleviating documentation burdens and enhancing the accessibility of medical records for both professionals and patients. They have addressed key challenges such as adapting outputs to various clinical tasks, and generating patient-friendly text. Nonetheless, several challenges persist: generalization across institutions and languages remains limited due to training data biases; factual consistency and clinical safety must be rigorously validated; and integrating multimodal signals (e.g., images, vitals) into text generation is still nascent. Future work should prioritize domain adaptation techniques, fine-grained clinical factuality evaluation, multimodal integration, and collaborative frameworks involving clinicians to ensure that generated content is both medically reliable and practically useful in diverse healthcare environments.

\textbf{EHR Based Prediction. }An electronic health record (EHR) is the systematized collection of electronically stored patient and population health information in a digital format. EHRs play a critical role in modern healthcare systems. Beyond the advantages of digitization—such as easier storage and review—EHRs significantly improve the quality and efficiency of medical care. They provide comprehensive, accurate, and real-time patient information, enabling clinicians to make more informed and precise clinical decisions. Moreover, EHRs facilitate the sharing of patient health information across departments and institutions, enhancing collaborative efficiency and ensuring continuity of care across different healthcare settings. The vast amount of data accumulated in EHR systems also provides a solid foundation for training LLMs, as these records often contain structured annotations—such as specific diseases and severity levels—that typically require little to no transformation, making them highly suitable for LLM-based learning.

A foundational model in this domain is BEHRT~\cite{li2020behrt}, a transformer-based model developed for healthcare representation learning. BEHRT adapts BERT to longitudinal EHR data by encoding structured sequences of medical codes (e.g., diagnoses, medications) along with age embeddings. By learning temporal dependencies, BEHRT achieved strong performance in tasks such as disease onset prediction (e.g., predicting diabetes based on early comorbidities), and it demonstrated robust performance in downstream stratification tasks with minimal fine-tuning.

However, BEHRT was trained on relatively small datasets, limiting its potential. In contrast, Med-BERT~\cite{rasmy2021med}, a context-based embedding model, was pre-trained on a large-scale structured EHR dataset comprising 28,490,650 patients and clinical coding systems like ICD-10 and CPT. Fine-tuning experiments showed that Med-BERT significantly improved prediction accuracy. Notably, Med-BERT performed exceptionally well with small fine-tuning datasets, achieving AUC scores that surpassed baseline deep learning models by over 20\%, and even matched the performance of models trained on datasets ten times larger~\cite{rasmy2021med}.

Building on this trend, GatorTron~\cite{yang2022large} scaled up the model size to 8.9 billion parameters and was trained on over 90 billion clinical narratives and structured labels. It demonstrated remarkable generalization capabilities in tasks such as phenotype prediction and cohort selection. Its scalability enables modeling of complex inpatient trajectories and supports patient-level reasoning even in low-resource scenarios.

In the multimodal space, models like MultiMedQA~\cite{singhal2023large} and Clinical Camel~\cite{toma2023clinical} integrate structured EHR entries (e.g., vital signs, lab results) with textual prompts to generate clinical answers from tabular data. For example, given a prompt such as “Is this patient at risk for acute kidney injury?” and a series of lab values and medication records, the model outputs a response like “Yes, due to elevated creatinine levels and concurrent use of nephrotoxic drugs.”

LLMs have significantly advanced EHR-based prediction by leveraging structured medical codes, temporal information, and multimodal clinical signals to support personalized forecasting of disease onset, treatment outcomes, and patient risk stratification. These models, ranging from BEHRT and Med-BERT to the billion-parameter GatorTron, have demonstrated strong generalization across clinical tasks with minimal fine-tuning and are particularly effective even in low-resource settings. However, challenges remain in modeling long, sparse, and irregular patient timelines, ensuring clinical interpretability, and addressing domain shifts across institutions and EHR systems. Future work should focus on integrating heterogeneous modalities (e.g., imaging, genomics), improving temporal reasoning across fragmented records, and developing explainable frameworks that align model decisions with clinician expectations to foster trust and deployment in real-world healthcare settings.

\subsubsection{Biomedical Reasoning and Understanding}

Biomedical Reasoning and Understanding focus on the understanding and modeling of textual information in the fields of life sciences and bioengineering. The goal is to leverage artificial intelligence technologies to enhance the semantic parsing of natural language content such as scientific literature, clinical case records, and diagnostic reports. The primary inputs for these tasks are natural language texts—for example, research abstracts from PubMed or clinical notes from patient records. Similar to DNA or RNA sequences, natural language inherently contains rich semantic information and can be directly processed by language models without additional transformation. This representation allows models to learn semantic patterns, reasoning cues, and contextual dependencies embedded in the language, thereby providing strong semantic support and reasoning capabilities for downstream tasks such as disease diagnosis, clinical report generation, and biomedical literature question answering.

The main objective of Biomedical Reasoning and Understanding is to perform a variety of practically meaningful downstream tasks based on effective modeling of natural language texts. For example, given a research abstract from PubMed, the model needs to accurately identify and extract biomedical entities such as drug names, disease types, and gene symbols, and further uncover functional relationships among them, such as “Drug X treats Disease Y” or “Gene A is significantly associated with Disease B.” Additionally, the model can assist researchers in quickly understanding complex oncology reports and automatically answering questions such as “What treatment methods were used in this study?” or “Which patient subgroups benefited the most according to the results?” More generally, scenarios also include using medical examination questions (e.g., USMLE) as input to evaluate the model's question-answering and reasoning capabilities across broad medical knowledge domains. These tasks rely on extensive biomedical knowledge found in literature, clinical notes, and databases, posing high demands on LLMs in terms of factual recall, domain-specific reasoning, and complex language interpretation.

In the life sciences and bioengineering domains, the value of Biomedical Reasoning and Understanding lies in their ability to extract critical information from massive volumes of text that is vital for research and clinical decision-making. Similar to the “information sparsity” seen in sequence-based tasks, biomedical texts also exhibit low information density but high-value key content—for instance, a clinical case report may be lengthy, yet the truly decisive content for diagnosis or treatment planning is often minimal. Therefore, models must possess robust capabilities in long-text modeling, information retrieval, and semantic compression to effectively accomplish the task objectives. Furthermore, in certain application scenarios, the model can also automatically generate clinical decision-making suggestions based on existing research findings, or explain treatment plans to patients in more accessible language—thus promoting research transparency and improving doctor-patient communication.

Although all these tasks involve text processing, they differ fundamentally in task structure, reasoning focus, and model design. Some subtasks revolve around retrieving or generating answers to biomedical questions—such as determining a diagnosis, choosing a treatment plan, or interpreting research outcomes—and typically require models to possess strong knowledge recall and evidence-based reasoning capabilities. In contrast, others focus on identifying semantic relationships, logical entailment, or classification problems within or across texts—for example, determining whether two sentences entail each other, or classifying text based on medical intent. These two categories reflect two long-standing paradigms in natural language processing: retrieval/generation and reasoning/classification, which also align with widely adopted benchmarking methods today. Therefore, we further subdivide Biomedical Reasoning and Understanding into two categories: \textbf{Question Answering} and \textbf{Language Understanding}.

Both categories benefit from the abundance of unlabeled or partially labeled biomedical text resources, including research papers, clinical notes, and medical examination datasets, which provide rich materials for self-supervised or weakly supervised pretraining. Furthermore, LLMs are not only capable of generating accurate answers or performing effective classification, but also excel at providing clear and interpretable reasoning, thereby significantly enhancing the transparency and trustworthiness of predictions. These characteristics enable LLMs to transcend individual task boundaries and provide robust technical support for knowledge-intensive biomedical reasoning.

\textbf{Question Answering. }Biomedical Question Answering focuses on enabling models to accurately extract or generate answers from scientific literature, clinical notes, or medical guidelines in response to domain-specific natural language queries. A series of LLMs and domain-specific models have been applied to biomedical question answering (QA). Early approaches employed transformer models such as BioBERT~\cite{lee2020biobert} and PubMedBERT~\cite{gu2021domain}, BERT~\cite{devlin2019bert} based models pre-trained on biomedical corpora—and fine-tuned them for QA tasks. Compared to general-purpose language models, these domain-specific models demonstrated higher accuracy on biomedical QA benchmarks. For example, BioBERT~\cite{lee2020biobert} achieved higher F1 scores than baseline BERT in the BioASQ~\cite{krithara2023bioasq} challenge tasks, owing to its domain-specific pretraining. Generative transformer models tailored to biomedicine have also been developed, such as BioGPT~\cite{luo2022biogpt} (a GPT-2-style~\cite{radford2018improving} model trained on biomedical texts) and BioMedLM~\cite{bolton2024biomedlm} (also known as PubMedGPT 2.7B, a GPT-based model trained on PubMed abstracts). These models have achieved strong results in QA tasks.

Subsequently, instruction tuning and conversational LLMs entered the biomedical QA domain. Med-PaLM~\cite{singhal2023large} (and its successor Med-PaLM 2~\cite{singhal2025toward}) fine-tuned Google’s PaLM~\cite{chowdhery2022palm} model on medical QA tasks and achieved near-expert-level performance on the United States Medical Licensing Examination (USMLE), with accuracy around 86.5\%, approaching that of expert physicians (87\%).

To move toward truly doctor-like LLMs—beyond simply answering questions—researchers have fine-tuned pretrained models on more novel datasets. For example, ChatDoctor~\cite{li2023chatdoctor} was created by fine-tuning LLaMA~\cite{touvron2023llama} on medical dialogue data, enabling interactive QA in a patient-doctor chat format. HuatuoGPT~\cite{huatuogpt-2023} posits that an intelligent medical advisor should proactively ask questions rather than respond passively. Huatuo-2~\cite{chen2023huatuogptii} uses an innovative domain-adaptive approach to significantly improve its medical knowledge and conversational skills. It demonstrated optimal performance on several medical benchmark tests, notably surpassing the GPT-4 on the Specialist Assessment and the new version of the Physician Licensing Exam. Similarly, models such as Clinical Camel~\cite{toma2023clinical} and DoctorGLM~\cite{xiong2023doctorglm} are LLM-based medical chatbots designed specifically to answer medical questions in a conversational style.

At the same time, thanks to LLMs' inherent zero-shot capabilities, large general-purpose models like GPT-4~\cite{achiam2023gpt} remain competitive, which has demonstrated strong performance in medical QA and often outperforms smaller domain-specific models in zero-shot settings~\cite{nori2023capabilities}.

Recently, reasoning has played an increasing role in this subtask. For example, Huatuo-o1~\cite{chen2024huatuogpto1medicalcomplexreasoning} enhances complex reasoning ability by (1) guiding the search for complex reasoning trajectories using a verifier and fine-tuning the LLM accordingly, and (2) applying reinforcement learning (RL) with verifier-based rewards. FineMedLM-o1~\cite{yu2025finemedlm} further introduced Test-Time Training~\cite{sun2020test} into the medical domain for the first time, promoting domain adaptation and ensuring reliable and accurate reasoning.

LLMs have substantially advanced biomedical question answering by enabling precise information extraction and fluent generation from diverse medical texts, ranging from clinical guidelines to patient dialogues. They have outperformed traditional domain-specific baselines by incorporating instruction tuning, conversational capabilities, and advanced reasoning techniques. However, several challenges remain: factual consistency and hallucination still pose risks in high-stakes clinical applications; models often struggle with ambiguous queries, underrepresented diseases, or multimodal reasoning; and real-world deployment requires careful alignment with clinical workflows and regulations. Future efforts should focus on integrating more life science and bio-engineering knowledge, enhancing traceable multi-step reasoning, and developing evaluation protocols that reflect real-world clinical utility, ensuring that QA systems can support clinicians safely and effectively.

\textbf{Language Understanding. }Language Understanding in the life science and bio-engineering involves modeling a system’s ability to comprehend, interpret, and reason over domain-specific texts, enabling accurate semantic inference and contextual judgment across diverse biomedical and scientific narratives. Beyond direct question answering, LLMs are increasingly applied to various language understanding tasks in the biomedical domain. These tasks require interpretation and reasoning over biomedical texts such as clinical narratives, scientific abstracts, or exam questions to support judgments or classifications. A typical example is natural language inference (NLI) in medicine: given a textual premise (e.g., a statement from a patient report) and a hypothesis, the model must determine whether the premise entails, contradicts, or is neutral with respect to the hypothesis. For instance, consider the premise “The patient denies any history of diabetes,” and the hypothesis “The patient has a history of diabetes.” A model with true understanding should correctly classify this as a contradiction, since the hypothesis directly conflicts with the premise. Language understanding is crucial for clinical decision support and information extraction, as it determines whether conclusions are genuinely grounded in clinical observations or life science facts. It also underpins tasks such as document classification, textual entailment, and reading comprehension. In essence, these tasks assess whether LLMs demonstrate deep comprehension of biomedical language, rather than mere memorization.

Many LLMs originally developed for QA have also been used for understanding tasks, often via fine-tuning for classification. Early models like BioBERT~\cite{lee2020biobert} and PubMedBERT~\cite{gu2021domain} pioneered performance improvements in biomedical text classification and NLI, achieving strong results on tasks such as MedNLI~\cite{romanov2018lessons}. Fine-tuned BioBERT on the MedNLI dataset significantly outperformed earlier RNN-based models in reasoning accuracy, because of its better grasp of clinical terminology and context. ClinicalBERT~\cite{clinicalbert}, initialized from BioBERT~\cite{lee2020biobert} and further trained on electronic health records, proved particularly effective in clinical NLI and related tasks, as it captured domain-specific syntax and abbreviations from structured data. More recent domain-specific models, such as BioLinkBERT~\cite{yasunaga2022linkbert} and BlueBERT~\cite{peng2019transfer}, report MedNLI accuracy in the mid-80\% range—approaching human expert performance.

Meanwhile, large general-purpose LLMs have demonstrated capability in language understanding via prompting. For example, GPT-4~\cite{achiam2023gpt} can perform NLI without explicit fine-tuning, when prompted with queries like, “Does the following statement logically follow from the previous one?”~\cite{nori2023capabilities} Trained on a broad corpus—including some medical content—these models often achieve decent accuracy in zero-shot or few-shot settings.

However, instruction-finetuned biomedical models are pushing the boundaries further. A recent method, BioInstruct~\cite{tran2023instruction}, compiled around 25,000 biomedical instruction-response pairs, covering tasks such as NLI and QA, and used them to fine-tune a LLaMA model. This resulted in significant improvements across multiple benchmarks, indicating that targeted instruction tuning can effectively teach LLMs the reasoning patterns required for biomedical language understanding. Similarly, models like ChatDoctor~\cite{li2023chatdoctor} and Clinical Camel~\cite{toma2023clinical} (based on LLaMA~\cite{touvron2023llama}), which were introduced for QA, can also perform classification or inference in a dialogue format when guided appropriately through prompts or lightweight fine-tuning. In summary, a wide range of models—from domain-specific BERTs to large GPT-style models—have been leveraged for understanding tasks. The trend is moving away from training small task-specific models from scratch and toward adapting large foundation language models (e.g., LLaMA-7B or 13B) via fine-tuning or prompting, to better transfer their general knowledge and linguistic capability to the complex biomedical domain.

LLMs have significantly advanced language understanding in the biomedical and life science domains by enabling contextual reasoning, semantic inference, and classification across complex and specialized texts such as patient reports, scientific literature, and medical examinations. These models have proven effective in tasks like natural language inference (NLI), document classification, and reading comprehension—particularly through domain-adaptive pretraining and instruction tuning. However, key challenges remain: understanding nuanced clinical negations, reasoning over long and fragmented documents, and ensuring interpretability in high-stakes decision-making. Future work should focus on improving zero-shot generalization across clinical subdomains, integrating structured biomedical ontologies for more grounded reasoning, and developing explainable evaluation frameworks to assess whether models truly comprehend rather than memorize biomedical language.

\subsubsection{Hybrid Outcome Prediction}

Hybrid Outcome Prediction refers to a class of tasks where LLMs are employed to predict complex biological or therapeutic outcomes by integrating diverse types of biological, chemical, and contextual information. Unlike traditional sequence-only or structure-only modeling, hybrid prediction tasks often require models to simultaneously reason over multiple heterogeneous inputs—such as chemical structures, genetic profiles, and cellular environments—to forecast functional outcomes or treatment effects. These tasks are of paramount importance in life sciences and bioengineering, as many real-world biological phenomena—such as drug response, synergistic effects, or protein function—arise from the interplay of diverse molecular and cellular factors rather than from single-modality information.

Typical inputs to hybrid prediction tasks may include combinations of small molecules, amino acid sequences, gene expression profiles, mutation data, or even broader multi-omics signatures. The outputs range from continuous measurements (e.g., synergy scores, binding affinities) to categorical labels (e.g., synergistic vs. antagonistic drug pairs, functional vs. non-functional protein variants). Hybrid outcome prediction challenges models not only to capture complex intra- and inter-modality relationships but also to generalize across biological contexts that may differ substantially between training and deployment scenarios.

The importance of hybrid outcome prediction is amplified in translational research and therapeutic development, where accurate computational forecasts can dramatically reduce experimental costs, prioritize candidate interventions, and uncover novel biological mechanisms. However, this class of tasks poses unique challenges: input modalities are often high-dimensional and noisy; the relationships between features and outcomes can be nonlinear and context-dependent; and biological interpretability remains a significant hurdle. LLMs, with their ability to integrate multimodal data, model contextual dependencies, and adapt to new tasks through fine-tuning or prompting, are particularly well-suited to address these complexities.

In this section, we focus on two major sub-directions within Hybrid Outcome Prediction: Drug Synergy Prediction and Protein Modeling. Both represent critical applications where LLMs have demonstrated transformative potential, yet where significant challenges and opportunities for future development remain.

\textbf{Drug Synergy Prediction. }Drug synergy prediction involves forecasting the therapeutic efficacy of drug combinations. In many diseases—particularly cancer—combination therapies can improve treatment outcomes and prevent resistance. Drug synergy refers to a phenomenon where the combined effect of two (or more) drugs exceeds the effect of each drug administered individually. Identifying synergistic drug pairs is critical for accelerating the design of combination therapies while reducing the need for extensive laboratory testing. However, this task is highly challenging due to the combinatorial explosion of possible drug pairs and the complex biological mechanisms underlying their interactions. The synergy of a given drug pair can vary depending on the context—such as cell type or disease environment—making generalization difficult. Despite these challenges, accurate synergy prediction can dramatically narrow the search space for effective multi-drug treatment regimens.

Models designed for this task typically take two drugs as input—often represented by their chemical structures, such as SMILES strings or molecular fingerprints—along with contextual features like genomic profiles of the target cell line. The output is a synergy score or class label indicating whether the combination exhibits synergistic behavior. Specifically, the input may consist of a pair of SMILES strings \textit{\{Drug A, Drug B\}} and a cell line ID, while the output could be a continuous synergy metric, such as a Bliss or Loewe additivity score, or a binary label (synergistic vs. non-synergistic). Some models use drug-pair dose-response matrices, though many modern approaches simplify the task to predicting a single synergy score per drug pair. Incorporating contextual information (e.g., gene expression or mutation data of the cell line) makes this a multimodal prediction task, as synergy is often conditional on biological context.

Early synergy prediction methods used feature-engineered machine learning models. DeepSynergy~\cite{preuer2018deepsynergy}, for example, employed deep neural networks to combine molecular descriptors with gene expression profiles. More recently, transformer-based and LLM-inspired models have emerged. One notable example is DFFNDDS~\cite{xu2023dffndds}, which integrates a BERT-like language model~\cite{devlin2019bert} for encoding drug SMILES and introduces a dual-feature fusion attention mechanism to capture drug-cell interactions. The BERT module in DFFNDDS~\cite{xu2023dffndds} jointly attends to drug-drug and drug-cell features to learn non-linear synergy effects. This architecture helps discover subtle interaction patterns—such as complementary mechanisms of action—that may be missed by simpler models or naive feature concatenation.

CancerGPT~\cite{li2024cancergpt} introduced a few-shot approach using a GPT-style model, transforming tabular synergy data into natural language format and fine-tuning GPT-3 to predict drug synergy in rare cancers. This method leverages the prior knowledge embedded in the language model's weights, enabling accurate predictions even with zero or few training samples in new tissue types. Another cutting-edge method, SynerGPT~\cite{edwards2023synergpt}, pretrains a GPT model to perform contextual learning of a “synergy function.” It is trained to take a personalized dataset of known synergistic pairs as a prompt and then predict new pairs under the same context. This context-based approach avoids reliance on fixed molecular descriptors or domain-specific biological knowledge, instead extrapolating from patterns embedded in the prompt—achieving competitive results.

Furthermore, LLMs can serve as foundation models to address a diverse set of tasks in this domain. One such approach, BAITSAO~\cite{Liu2024.04.08.588634}, is a foundation model strategy that integrates multiple datasets and tasks. It uses context-rich embeddings from LLMs as initial representations of drugs and cell lines, and performs pretraining on large drug combination databases within a multitask learning framework. BAITSAO~\cite{Liu2024.04.08.588634} outperformed both classical models (like DeepSynergy~\cite{preuer2018deepsynergy}) and more recent tabular or transformer-based models on benchmark datasets, thanks to its multitask training and transfer learning across drug combination contexts. Overall, these LLM-based strategies—from fine-tuned GPT models to transformer fusion networks—highlight the growing role of language model architectures in capturing the complex relationships underlying drug synergy.

LLMs have brought significant advances to drug synergy prediction by enabling the modeling of complex drug–cell line interactions through contextual embeddings, attention mechanisms, and prompt-based reasoning. These models reduce reliance on handcrafted features, generalize better across biological contexts, and support few-shot or even zero-shot inference, which is especially valuable for rare diseases or under-studied drug pairs. However, key challenges remain: biological interpretability is limited, especially in identifying mechanistic pathways; synergy predictions often lack consistency across datasets or experimental conditions; and integrating multi-omics data with chemical and pharmacological knowledge in a unified framework is still an open problem. Future work should focus on enhancing cross-dataset generalization, embedding biological priors into LLM architectures, and developing transparent, mechanistically grounded models that can support experimental design and clinical translation in combination therapy development.

\input{tables/bio_bench}

\input{tables/bio_bend}

\input{tables/rna_bench}

\input{tables/bio_qa}

\input{tables/bio_lu}

\input{tables/bio_ehrqa}

\input{tables/bio_drug}

\input{tables/bio_protrein}

\textbf{Protein Modeling. }Protein Modeling refers to the task of learning structural, functional, or evolutionary patterns from amino acid sequences, enabling predictions of protein properties such as folding, function, or interaction based. The development of protein LLMs has been driven by the deepening integration of computational biology and natural language processing techniques. Early efforts focused on leveraging traditional deep learning architectures such as LSTMs for representation learning of protein sequences~\cite{alley2019unified,bepler2019learning,hie2021learning}. Models like UniRep~\cite{alley2019unified} and Bepler \& Berger~\cite{bepler2019learning} made initial progress in constructing protein embedding vectors. 

AlphaFold~\cite{jumper2021highly} is a protein structure prediction model released by DeepMind, which revolutionized the long-standing protein folding problem through deep learning. The model integrates evolutionary homologous sequences, residue-pair geometric maps, and physical constraints using an attention-based network. In CASP14, it achieved an average GDT\_TS of 92.4, marking the first time atomic-level accuracy was reached. The subsequent release of a database containing over 2 million predicted structures has significantly accelerated drug discovery, enzyme engineering, and pathogenic mutation annotation.

With the rise of Transformers~\cite{vaswani2017attention} in natural language processing, this paradigm was rapidly transferred to protein modeling. Transformer-based models such as ProtTrans~\cite{elnaggar2021prottrans} and ESM-1b~\cite{rives2021biological} emerged, offering enhanced capabilities in capturing long-range dependencies within sequences, significantly improving the accuracy of protein structure and function prediction.

The ESM series has since expanded in both model size and task scope—from ESM-1v~\cite{meier2021language} to ESM-2~\cite{lin2022language} and the latest ESM-3~\cite{hayes2024simulating}—achieving end-to-end sequence-to-structure prediction (e.g., ESMFold~\cite{lin2022language}), incorporating multimodal information, and enabling complex reasoning and even generative design for protein function. These advancements signify a shift toward universal modeling and reasoning capabilities in protein LLMs.

Beyond foundational modeling capabilities, researchers have begun to explicitly inject structural information into the training process to enhance the models’ ability to capture 3D protein conformations. Models such as SaProt~\cite{su2023saprot} and ESM-GearNet~\cite{zhang2023enhancing} integrate local or global structural features to enrich sequence representations, while approaches like OntoProtein~\cite{zhang2022ontoprotein} and ProteinCLIP~\cite{wu2024proteinclip} leverage knowledge graphs and contrastive learning with text to improve semantic understanding and generalization. These structure-informed and knowledge-enhanced strategies have not only improved model expressiveness on tasks such as mutation effect prediction, functional domain annotation, and binding site identification, but also extended the applicability of protein LLMs to drug target identification and molecular interaction prediction.

Building on foundational understanding and reasoning, protein LLMs have further evolved toward generative modeling. ProGen~\cite{madani2023large} and ProtGPT2~\cite{ferruz2022protgpt2} were among the first to apply the autoregressive language modeling paradigm to protein sequence generation, capable of producing diverse, biologically active sequences conditioned on functional labels or species. ProGen2~\cite{nijkamp2023progen2} scaled up both model size and training data, significantly enhancing its ability to model protein adaptability and diversity. Meanwhile, ProLLaMA~\cite{lv2024prollama} incorporated protein sequence learning into the LLaMA architecture, achieving joint understanding and generation within a single framework and demonstrating the potential of multi-task and cross-modal pretraining. In contrast, models like Pinal~\cite{dai2024toward} and Ankh~\cite{elnaggar2023ankh} explore structure-guided, efficient encoder-decoder architectures to balance generation quality with parameter efficiency.

At the same time, several integrated frameworks have emerged to support protein design and engineering. For example, ProteinDT~\cite{liu2025text} enables zero-shot generation of protein sequences from textual functional descriptions, while PLMeAE~\cite{zhang2025integrating} integrates with automated biological experimentation platforms to construct a “design-build-test-learn” loop for automated protein engineering. Innovative interactive tools such as ProteinGPT~\cite{xiao2024proteingpt} and ProteinChat~\cite{guo2023proteinchat} have also appeared, supporting structure input, language interaction, and functional Q\&A, further advancing protein language models toward intelligent agents with cognitive and interactive capabilities.

Overall, the evolution of protein LLMs has clearly progressed from small-scale LSTM-based semantic embeddings, to large-scale Transformer-based structural predictions, and toward multimodal-enhanced generative design. This trajectory has not only significantly expanded the frontiers of protein science but also laid a robust foundation for the next generation of biomolecular design, functional prediction, and clinical applications.

\subsubsection{Benchmarks}

The rapid adoption of LLMs in life sciences and bio-engineering has spurred the development of specialized benchmarks designed to systematically assess their performance across diverse biological and clinical tasks. Benchmarks such as BEND for DNA language models and BEACON for RNA language models rigorously evaluate the ability of LLMs to interpret complex genomic and transcriptomic information, encompassing tasks that range from functional element annotation in genomic sequences to predicting RNA secondary structures. Complementing these biological benchmarks, medical QA datasets like MedQA, MedMCQA, and PubMedQA focus on evaluating clinical knowledge, reasoning capabilities, and contextual understanding of biomedical literature. Together, these benchmarks offer a comprehensive framework to evaluate and drive progress in applying LLMs to real-world biomedical challenges.

\textbf{BEND. }BEND is a unified evaluation framework designed to systematically assess the performance of DNA language models (DNA LMs) on realistic biological tasks. The benchmark suite comprises seven tasks based on the human genome, covering functional elements at varying length scales, such as promoters, enhancers, splice sites, and transcription units. Each task provides input sequences and labels in a standardized format, supporting a range of downstream tasks including both classification and regression.

The task design of BEND reflects the core challenges of genome annotation: wide variation in sequence length, sparsity of functional regions, and low signal density. To evaluate the performance of DNA LMs on these tasks, BEND offers a scalable framework for generating embedding representations and training lightweight supervised models. Experimental results demonstrate that while certain DNA LMs can approach the performance of expert methods on specific tasks, they still face difficulties in capturing long-range dependencies (such as enhancer recognition). Moreover, different models display varying preferences for modeling gene structure and non-coding region features.

For example, in the enhancer annotation task, BEND formulates the problem as binary classification: for each 128-base-pair segment of gene-adjacent DNA, the model predicts whether it contains an enhancer. Data are sourced from CRISPR interference experiments and integrated with major transcription start site (TSS) information, with a 100,096-bp sequence extracted for each gene and annotated in 128-bp segments. The main challenge of this task lies in identifying distal regulatory relationships, which tests the model's ability to capture long-range dependencies.

\textbf{BEACON. }BEACON is a comprehensive evaluation benchmark specifically designed for RNA language models, encompassing 13 tasks related to RNA structural analysis, functional studies, and engineering applications. All tasks adopt a unified data format and support both classification and regression evaluations, applicable to both sequence-level and nucleotide-level predictions.

For example, in the RNA secondary structure prediction task, the model is required to determine whether each pair of nucleotides forms a base pair, with the F1 score used as the evaluation metric. The data for this task is sourced from the bpRNA-1m database.

BEACON also includes a systematic evaluation of various models and finds that single-nucleotide tokenization and ALiBi positional encoding demonstrate superior performance across multiple tasks. Based on these findings, a lightweight baseline model named BEACON-B is proposed.

\textbf{QA Benchmarks. }

The landscape of biomedical and clinical QA benchmarks spans a diverse range of tasks, from licensing examination questions to domain-specific reasoning over scientific literature. These datasets challenge models not only on factual recall but also on higher-order reasoning, reading comprehension, and the ability to synthesize information from complex biomedical contexts. Together, they provide a comprehensive evaluation suite for assessing the medical knowledge, reasoning ability, and contextual understanding of AI systems in healthcare and biomedical research.

Current benchmarks are primarily concentrated within the English language domain, often based on medical licensure examinations from different English-speaking countries. There are also benchmarks in other languages, such as Chinese. These benchmarks fully simulate real-world exams, providing only the question stem and answer choices. In addition, there are benchmarks that supply LLMs with a reference document, requiring the model to combine its own knowledge with the provided context to generate a more informed answer.

\begin{itemize}[leftmargin=10pt]

\item \textbf{MedQA. }The MedQA dataset consists of multiple-choice questions from the United States Medical Licensing Examination (USMLE). It covers general medical knowledge and includes 11,450 questions in the development set and 1,273 questions in the test set. Each question has 4 or 5 answer choices, and the dataset is designed to assess the medical knowledge and reasoning skills required for medical licensure in the United States.

\item \textbf{MedMCQA. }
MedMCQA is a large-scale multiple-choice QA dataset derived from Indian medical entrance examinations (AIIMS/NEET). It covers 2.4k healthcare topics and 21 medical subjects, with over 187,000 questions in the development set and 6,100 questions in the test set. Each question has 4 answer choices and is accompanied by an explanation. MedMCQA evaluates a model's general medical knowledge and reasoning capabilities.

\item  \textbf{PubmedQA. }Different from MedQA and MedMCQA, PubMedQA is a closed-domain QA dataset, In which each question can be answered by looking at an associated context (PubMed abstract). It is consists of 1,000 expert-labeled question-answer pairs. Each question is accompanied by a PubMed abstract as context, and the task is to provide a yes/no/maybe answer based on the information in the abstract. The dataset is split into 500 questions for development and 500 for testing. PubMedQA assesses a model's ability to comprehend and reason over scientific biomedical literature.

\end{itemize}

\noindent\textbf{Drug Synergy Prediction}

\textbf{CancerGPT.} CancerGPT~\cite{li2024cancergpt} assesses the capability of LLMs to predict drug pair synergy in rare cancer tissues with limited structured data. The evaluation framework involves testing LLMs' performance in few-shot and zero-shot learning scenarios across seven rare tissue types, comparing the results to those of larger models like GPT-3~\cite{brown2020language}.

The evaluation process includes:
\begin{itemize}[leftmargin=10pt]

\item \textbf{Few-shot and Zero-shot Learning:} Assessing the model's ability to predict drug synergy with minimal or no training examples, highlighting the LLM's capacity to generalize from limited data.

\item \textbf{Benchmarking Across Multiple Tissues:} Testing the model's predictive performance across seven different rare cancer tissue types to ensure robustness and generalizability.
    
\end{itemize}

This evaluation framework demonstrates that LLMs, even with fewer parameters, can effectively predict drug pair synergies in contexts with scarce data, offering a promising approach for biological inference tasks where traditional structured data is lacking.

\textbf{BAITSAO}. The benchmark framework integrates both regression and classification tasks, based on synergy scores (e.g., Loewe, Bliss, HSA, ZIP) and binary synergy labels derived from large-scale drug combination datasets such as DrugComb. Each sample consists of a drug pair and a cell line, with input features constructed from Large Language Model (LLM) embeddings of descriptive prompts about drugs and cell lines, standardized into numerical vectors.

The design of the BAITSAO evaluation suite reflects key challenges in drug synergy prediction: sparse synergy signals, heterogeneous data formats, and limited generalization to novel drug combinations. To evaluate model performance, BAITSAO pre-trains on large-scale synergy data under a multi-task learning (MTL) framework, capturing both single-drug inhibition and pairwise synergy. The model is then assessed on held-out datasets using metrics such as Pearson correlation, mean squared error, ROC-AUC, and accuracy. Ablation and sensitivity analyses are further conducted to study embedding strategies, training data scales, and model scaling laws.

For example, in the synergy classification task, BAITSAO formulates the problem as a binary prediction: given a pair of drugs and a cell line, predict whether the combination yields a synergistic effect. Inputs are constructed by averaging the embeddings of both drugs and concatenating with the cell line embedding, with synergy labels binarized using a threshold on the Loewe score. This setup evaluates the model’s ability to generalize to unseen drug-cell line combinations, including out-of-distribution (OOD) samples, and serves as a robust benchmark for multi-drug reasoning and zero-shot prediction.

\noindent\textbf{Protein Modeling}

\textbf{TAPE.} TAPE (Tasks Assessing Protein Embeddings) is a large-scale benchmark designed to evaluate transfer learning methods on protein sequences. It comprises five biologically relevant tasks, including protein structure prediction, remote homology detection, and protein engineering. Each task features carefully curated splits to assess models’ ability to generalize in biologically meaningful ways. TAPE evaluates various self-supervised learning approaches for protein representation and shows that pretraining significantly improves performance across nearly all tasks, although traditional non-neural methods still outperform in some cases. The benchmark promotes standardized evaluation and method comparison in protein modeling.

\textbf{PEER.} PEER (Protein sEquence undERstanding) is a comprehensive and multi-task benchmark designed to evaluate deep learning methods on protein sequences. It encompasses 17 biologically relevant tasks across five categories: protein function prediction, localization prediction, structure prediction, protein-protein interaction prediction, and protein-ligand interaction prediction. Each task includes carefully curated training, validation, and test splits to assess models' generalization capabilities in real-world scenarios. PEER evaluates various sequence-based approaches, including traditional feature engineering methods, different sequence encoding techniques, and large-scale pre-trained protein language models.


\textbf{Summary.} Benchmarking efforts in life‑science and bio‑engineering LLMs now coalesce around four broad task families. First, \textbf{sequence‑based evaluation} dominates DNA and RNA modeling. Suites such as BEND (DNA) and BEACON (RNA) probe classification and regression across functional‑element annotation, secondary‑structure inference, and variant‑effect prediction. Second, \textbf{clinical structured‑data tasks} assess models on Electronic Health Records, splitting into clinical language generation (e.g., ClinicalT5, GPT‑4 hospital‑note drafting) and EHR‑based prediction (e.g., BEHRT, Med‑BERT risk scoring). Third, \textbf{textual knowledge tasks} test biomedical reasoning and understanding via QA (MedQA, MedMCQA, PubMedQA) and natural‑language inference benchmarks such as MedNLI, measuring factual recall, chain‑of‑thought reasoning, and long‑context comprehension. Finally, \textbf{hybrid outcome‑prediction benchmarks}—drug‑synergy suites (e.g., DrugCombDB subsets) and protein‑modeling challenges (ESMFold, ProGen)––demand multimodal integration across chemistry, omics and cellular context.

Across these categories, domain‑trained transformers consistently outperform classical baselines. \textbf{In sequence modeling}, long‑context LLMs (Enformer, HyenaDNA) improve enhancer or eQTL effect prediction correlations by 20–40\% over CNN/RNN hybrids, while bidirectional masked models (DNABERT‑2) raise MCC scores on promoter/enhancer detection by 2–5 pp versus 6‑mer CNNs. \textbf{In clinical language generation}, instruction‑tuned GPT‑4 drafts discharge summaries that clinicians rate as equal or superior in accuracy and readability to human‑written notes, and models like Med‑PaLM 2 reach 86\% accuracy on USMLE‑style exams, narrowing the gap to licensed physicians. \textbf{EHR‑based predictors} such as GatorTron boost AUROC for onset prediction tasks by 3–6 pp relative to GRU or logistic‑regression baselines, even under low‑data fine‑tuning. \textbf{In drug‑synergy prediction}, transformer fusion networks (DFFNDDS) and prompt‑based few‑shot GPT variants (CancerGPT) lift balanced accuracy by 5–12 pp over DeepSynergy, while LLM‑generated protein sequences (ProGen2) exhibit in‑vitro activities on par with natural enzymes in ≥50\% of tested families.

Yet substantial limitations persist. \textbf{Ultra‑long genomic context} still degrades accuracy despite linear‑time attention variants; distal enhancer–promoter linkage and rare‑variant generalization remain open. \textbf{Multimodal fusion} is ad‑hoc: most benchmarks isolate a single modality, leaving cross‑omics reasoning and image‑augmented clinical tasks underexplored. \textbf{Data quality and bias} are acute—human‑centric genomes, single‑institution EHRs, and English‑only QA corpora skew performance and hamper species‑, population‑, or language‑level transfer. \textbf{Safety and interpretability} issues mirror those seen in chemistry: hallucinated diagnoses or biologically implausible sequence designs can slip through, and attention maps alone rarely satisfy domain experts’ need for mechanistic insight.

From these observations we derive three actionable insights. \textbf{(1) Benchmark breadth and depth must expand}. Community curation of larger, more diverse genomes (e.g., non‑model organisms), multilingual clinical notes, and truly multimodal datasets (sequence+structure+phenotype+imaging) is essential. \textbf{(2) Representation and architecture choices require re‑thinking}. Treating kilo‑ to megabase sequences as flat text overlooks 3‑D chromatin contacts; integrating graph, spatial or physics‑aware modules with transformers, and exploring alternative encodings (e.g., byte‑pair k‑mers, SELFIES‑like bio‑tokens) can bridge this gap. \textbf{(3) Reliability hinges on task design and validation}. Embedding biological priors, tool‑augmented prompting (e.g., retrieval of wet‑lab evidence), and post‑hoc critic models can curb hallucination and enforce mechanistic plausibility; standardized factuality and safety metrics—analogous to clinical adjudication—should accompany benchmark scoreboards.

In sum, life‑science–oriented LLM benchmarks have revealed impressive gains over traditional pipelines, but progress is gated by richer data, modality‑aware architectures, and rigorous, biology‑centric evaluation. Aligning these elements will accelerate LLMs from promising assistants to dependable engines for discovery and precision medicine.

\subsubsection{Discussion}

\textbf{Opportunities and Impact. }
LLMs are now deeply integrated across the life sciences pipeline, supporting a broad spectrum of tasks ranging from genomic sequence interpretation to drafting clinical documentation. Their greatest impact has emerged in areas that align with their linguistic strengths—such as literature summarization, clinical note generation, and biomedical question-answering—where abundant data, low-cost supervision, and linguistic evaluation metrics have enabled rapid progress.

A major advantage lies in the \textbf{tokenizable structure} of biological data. Representations like k-mers for genomic sequences, SMILES for chemical compounds, and ICD codes in medical records are inherently suited to masked language modeling or autoregressive learning. As a result, LLMs like DNABERT-2, Nucleotide Transformer, BEHRT, and Med-PaLM 2~\cite{zhou2023dnabert2,dalla2024nucleotide,singhal2025toward} offer unified frameworks to model complex biological substrates. For example, RNA oligomers of various sizes require different experimental strategies—from NMR~\cite{johnson1994nmr} and FRET~\cite{jares2003fret} for small structures to cryo-EM and CLIP-seq for large complexes~\cite{bai2015cryo,kikhney2015practical}. Traditional computational methods often rely on size-specific architectures with manual feature engineering, while LLMs can tokenize all scales using shard tokenizers and learn a size-agnostic representation space.

Furthermore, LLMs exhibit \textbf{context extrapolation} capabilities that allow modeling of long-range dependencies in genomic data. For example, predicting MYC expression traditionally required multiple CNN-based sliding windows, Hi-C loop assemblies, and handcrafted features~\cite{dhanasekaran2022myc,roayaei2020mustache}, which struggled to capture distal interactions. In contrast, models like HyenaDNA~\cite{nguyen2023hyenadna} process 1 Mb genomic windows in a single forward pass, leveraging sub-quadratic convolutions to directly learn enhancer–promoter logic, thereby eliminating the need for fragmented, manually-curated pipelines.

The \textbf{instruction-following and multitask capabilities} of LLMs further enable unified handling of diverse biomedical applications. For instance, Med-PaLM 2 can simultaneously draft clinical notes, generate ICD-10/LOINC codes, and rewrite instructions at a 6th-grade level—all in a single prompt. This integration replaces three siloed hospital systems and reduces development timelines from months to hours.

\textbf{Challenges and Limitations. }
Despite these advancements, significant limitations persist. LLMs excel in symbolic and text-rich domains but underperform in tasks requiring deep experimental grounding or multi-scale biological reasoning.

First, the gap in \textbf{empirical grounding} remains a major bottleneck. Models such as ProGen2 can propose novel peptides, but their real-world efficacy is limited. For instance, the initial validation of ProGen2-generated incretin peptides showed an activity success rate of merely 7\%, emphasizing the indispensable role of iterative wet-lab testing and retraining.

Second, life science problems often involve \textbf{system-level complexity}, where token-based reasoning is insufficient. Predicting off-target effects of CRISPR editing or long-term drug toxicity demands multiscale modeling across molecular, cellular, and organismal levels. Although models like CRISPR-GPT~\cite{huang2024crispr} show promise, they still miss over 30\% of off-target sites in whole-genome data with complex chromatin interactions~\cite{jiang2017crispr,redman2016crispr,piergentili2021crispr}.

Third, the rise of powerful generative models introduces new \textbf{ethical, safety, and provenance concerns}. LLMs capable of generating accurate protocols may inadvertently facilitate dual-use research or propagate hallucinations. Open-source toxicity predictors like ToxinPred~\cite{rathore2024toxinpred,sharma2022toxinpred2} can potentially be misused to design harmful biological sequences. Without clear traceability or accountability mechanisms, the risks of misuse escalate.

\textbf{Research Directions.}
To address these challenges, we propose a forward-looking research agenda focused on hybrid architectures and responsible integration.
\begin{itemize} [leftmargin=10pt]

    \item First, future LLM systems should aim to \textbf{unify diverse biological modalities}—including genomic sequences, protein structures, cell images, clinical time-series, and textual notes—within a cohesive multimodal framework. Such models can enable integrated diagnosis and prediction by capturing complex biological correlations across data types.

    \item Second, LLMs should evolve from passive tools into \textbf{active hypothesis-generating agents}. This requires coupling with laboratory automation systems, real-time EHR streams, and high-throughput simulation platforms. For instance, an LLM-guided robotic lab could autonomously design, test, and refine molecular hypotheses in closed experimental loops, dramatically accelerating the discovery cycle.

    \item Third, the training of LLMs should incorporate \textbf{biologically-informed learning techniques}. Self-distillation improves interpretability through reasoning chains, contrastive alignment ensures consistency with biomedical knowledge bases, and physics-informed regularization grounds models in biophysical laws (e.g., thermodynamics in MD simulations), reducing hallucinations and enhancing trustworthiness.

    \item Finally, \textbf{proactive governance} must be embedded from the outset. Techniques such as differential privacy for sensitive patient data, watermarking synthetic DNA sequences to differentiate them from natural ones, and rigorous human oversight mechanisms are crucial for ensuring ethical deployment. Building responsible AI systems is not an afterthought—it must be integral to model development.

\end{itemize}

\textbf{Conclusion.}
LLMs have transformed information processing in the life sciences, accelerating literature review, genomic annotation, and clinical documentation. Their most pronounced successes lie in tasks with high symbolic complexity but relatively low experimental demands. However, realizing their full potential in experimental biology and bioengineering will require overcoming structural limitations.

This transformation demands \textbf{more than model scaling}; it necessitates innovations in architecture, training paradigms, and system integration. Bridging the gap between computational prediction and empirical validation calls for hybrid systems that fuse LLMs with biological priors, experimental platforms, and domain-specific constraints.

When thoughtfully designed and ethically deployed, LLMs can serve not merely as intelligent assistants but as generative partners in hypothesis formation, experimental design, and therapeutic innovation—ultimately accelerating the transition from scientific discovery to clinical application.

\subsection{Earth Sciences and Civil Engineering}

\subsubsection{Overview}

\noindent\textbf{5.5.1.1 Earth Sciences Introduction}

Earth sciences encompass the study of Earth's physical structure, processes, and history, as well as its surrounding systems such as the atmosphere, oceans, and outer space~\cite{fu2001earth,national2001basic,national2007earth}. These sciences aim to understand how the Earth and related environments function and change over time \cite{von1988theory}. Researchers in this field investigate diverse phenomena ranging from earthquakes and volcanoes to ocean circulation, weather systems, and planetary formation. Earth science not only satisfies human curiosity about our planet but also supports practical applications such as natural disaster prediction \cite{keum2020real,ge2022disaster}, climate modeling \cite{neelin2010climate}, resource exploration \cite{liu2022review}, and environmental protection~\cite{esty2004environmental}.

For instance, the early detection of seismic waves through global monitoring networks has enabled the development of earthquake early warning systems, helping to save lives in vulnerable regions like Japan and California~\cite{li2018machine}. Similarly, reconstructions of past climate from ice cores in Antarctica have revealed how small changes in atmospheric carbon dioxide were linked to major glacial cycles—insights that now guide our understanding of human-driven climate change~\cite{waters2016anthropocene}.

Given the interconnected nature of Earth's systems, research in earth sciences is typically categorized by major environmental domains: the solid Earth, water systems, the atmosphere, and space. Each category relies on specialized observational and analytical techniques developed across centuries of geoscientific exploration~\cite{fowler1990solid,hoff2010greening,bohme2021atmosphere,toth2005space}. Below, we outline the primary research tasks and traditional methods used in these four domains:

\noindent\textbf{Solid Earth.} This category involves studying Earth's interior, crust, and surface features to understand geological processes such as mountain building, earthquakes, and volcanic activity. Geologists use tools like field mapping~\cite{hornak1988magnetic,flick2014mapping,wandell2007visual}, rock sampling~\cite{baecher1977statistical}, and seismic imaging~\cite{scales1995theory,biondi20063d} to probe the planet's subsurface. Techniques such as radiometric dating~\cite{olsson1986radiometric} reveal the age of rocks and events, while plate tectonic theory~\cite{falvey1974development} provides a framework for explaining continental drift, subduction zones, and the formation of Earth's landforms.

\noindent\textbf{Water Systems.} This area includes both freshwater and marine environments. Hydrologists and oceanographers study the movement, distribution, and quality of water using instruments like stream gauges, sonar systems, and deep-sea submersibles~\cite{sun2021review,purser2018ocean,busby1976manned}. Ocean circulation, sea level rise, and the global water cycle are key focus areas. Research methods include remote sensing, water sampling, and the deployment of autonomous floats to measure temperature, salinity, and current flow. These investigations are crucial for understanding climate change, water resource management, and marine ecosystems~\cite{taylor2013ground}.

\noindent\textbf{Atmosphere.} The atmospheric sciences examine weather patterns, climate dynamics, and the composition of the air enveloping Earth. Meteorologists use weather stations, balloons, radar, and satellites to observe atmospheric conditions~\cite{larom1997influence}. Climate scientists build numerical models based on physical laws~\cite{mcguffie2001forty} to predict weather and analyze long-term climate trends. Historical data from sources like ice cores~\cite{robin1977ice} and tree rings~\cite{tei2017tree} help reconstruct past climates. Understanding the atmosphere is vital for forecasting extreme events, assessing air quality, and responding to global warming~\cite{hou2016long,he2024air}.

\noindent\textbf{Outer Space.} Also called planetary or space science, this domain explores celestial bodies and their relevance to Earth. Techniques include astronomical observation, space probe missions, spectroscopy, and sample return analysis~\cite{rast2019earth}. Comparative studies of planets, moons, and asteroids help scientists infer Earth's formation and evolutionary history. Data from missions like those to Mars or the Moon, along with meteorite analysis, deepen our understanding of planetary geology, habitability, and solar system dynamics~\cite{rossi2018planetary,horton2021assessing,murray1999solar}.

\noindent\textbf{5.5.1.2 Introduction to Civil Engineering}

Civil engineering is an engineering discipline dealing with the design, construction, and maintenance of physically or naturally built environment, including roads, bridges, residence and pipelines ~\cite{zavala2014survey, buildings11020066}. It is essentially the application of physical and scientific principles for addressing real-world challenges like climate crisis, encompassing fields like structures, material science, geology, soils, hydrology, environmental science, mechanics and other fields ~\cite{wikipedia_civil}. In simpler terms, civil engineering is about the management of buildings and infrastructures.

Civil engineering is a broad field composed of various branches, ranging from geotechnical engineering to urban planning. To better organize and understand the scope of civil engineering, by integrating information from relevant literature ~\cite{harle2024advancements,VADYALA2022100316}, we have summarized the common research tasks and corresponding classic solutions in civil engineering as following categories:

\textbf{Structural Design Optimization.} Structural optimization aims to find the best arrangement of structural components to meet requirements in terms of size, shape, topology and other aspects~\cite{tsiptsis2019structural}. Early research on structural optimization proposed mathematical programming and numerical search techniques~\cite{kuhn2013nonlinear,aldwaik2014advances}, which are still one of the most commonly applied approaches. Recently, metaheuristic methods have become popular due to their suitability in combinational optimization problems~\cite{saka2016metaheuristics}. However, they also suffer from high complexity and inadequacy for high-dimensional problems~\cite{sorensen2015metaheuristics,mahdavi2015metaheuristics}. A substantial number of studies have been proposed to address these problems~\cite{mortazavi2020new,zheng2020new}.

\textbf{Structural Health Monitor (SHM).} SHM refers to the process of structural data collection, assessment and damage detection in order to test the safety risks of new structures and the remaining time of existing structures~\cite{tokognon2017structural,gharehbaghi2022critical}. With the rapid development of IoT and wireless communication techniques, civil engineers usually use smart sensors to access a variety of structural parameters, such as mechanical and optical parameters~\cite{hodge2014wireless,cai2012structural}. Anomaly detection methods are commonly adopted for structural assessment and detection, including response-based~\cite{huseynov2020bridge,gomes2019review}, reliability-based~\cite{jamali2019reliability}, feature-based~\cite{babajanian2019damage} and computer vision methods~\cite{sari2019road,sarrafi2018vibration}.

\textbf{Material Behavior Modeling.} Based on historical data, the behavior of construction materials in different environments can be predicted, so that civil engineers can make informed decisions about material selection and construction arrangement~\cite{harle2024advancements}. Different materials usually adopts different behavior modeling strategies. For example, non-destructive experimental assessment methods are commonly used for concrete strength estimation~\cite{BREYSSE2012139,helal2015non}, numerical analysis methods and physical models are developed for soil behavior modeling~\cite{mitchell2005fundamentals,lou2011structure}, while finite element method is widely adopted to fill the heterogeneity gap among various composite materials~\cite{polym12040818,matthews2000finite}. 

\textbf{Product Behavior Modeling.} To ensure rational resource allocation and decision-making, it is also necessary to study the behavior of built systems, such as canal systems and transportation systems~\cite{harle2024advancements}. Machine learning algorithms have been widely applied to product behavior modeling for a long time, where commonly used traditional methods include SVM, fuzzy logic, Markov chains and evolutionary computation~\cite{YASEEN2015829,8344781,LIANG2023127067}. Recently, cutting-edge spatial-temporal network analysis techniques like Graph Neural Networks have received growing attention~\cite{JIANG2022117921,belbute2020combining}.

\begin{figure}[!t]
    \centering
    \includegraphics[width=\linewidth]{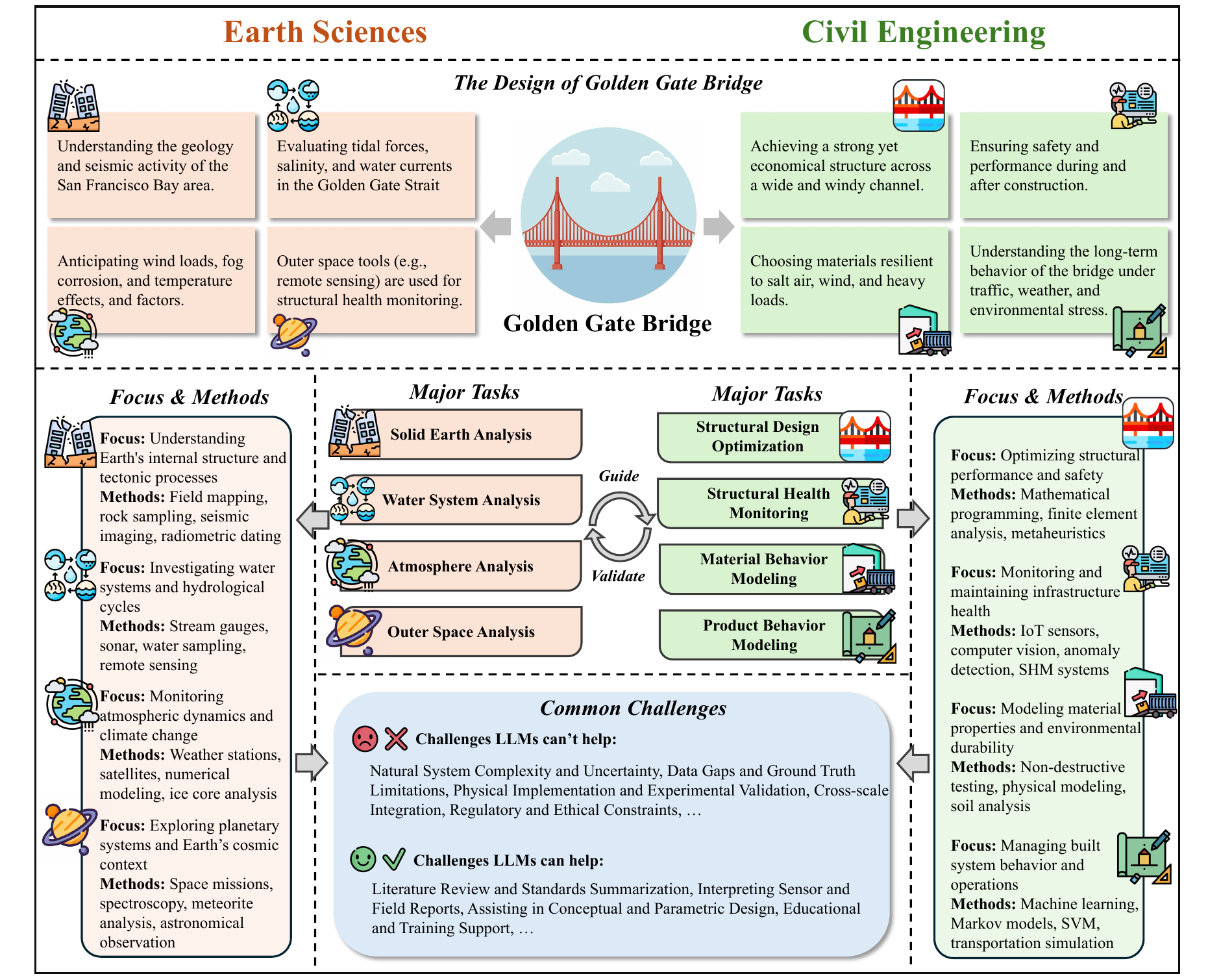}
    \caption{The relationships between major research tasks between earth sciences and civil engineering.}
    \label{fig:earch_civil_framework}
\end{figure}

\noindent\textbf{5.5.1.3 Current Challenges}

Earth sciences and civil engineering are deeply interconnected disciplines essential for understanding and shaping the physical world we live in. Earth sciences seek to decipher the natural processes that govern our planet, while civil engineering applies this knowledge to construct resilient, functional infrastructure. These fields play a vital role in addressing global issues such as natural disasters, climate adaptation, urbanization, and sustainability. Despite major advances in observational tools, modeling, and design techniques, both domains continue to face substantial challenges stemming from the complexity, unpredictability, and dynamic nature of Earth systems and human-built environments.

\noindent\textbf{Still Hard with LLMs: The Tough Problems.}

\begin{itemize}[leftmargin=10pt]
    \item \textbf{Natural System Complexity and Uncertainty.} Earth systems are governed by highly nonlinear, interdependent physical processes—from tectonic shifts and groundwater flow to atmospheric circulation and sea-level dynamics~\cite{neelin2010climate,fowler1990solid}. Predicting these phenomena accurately requires not only long-term observational data but also high-resolution numerical simulations. Civil engineers must incorporate these uncertainties into infrastructure design—such as accounting for variable soil conditions or future climate extremes. However, LLMs, while capable of interpreting related literature or summarizing best practices, lack the ability to model dynamic systems governed by partial differential equations or simulate stochastic environmental behavior. Capturing such processes requires specialized physics-based modeling and empirical calibration, far beyond the capacity of current text-based AI systems.
    \item \textbf{Data Gaps and Ground Truth Limitations.} Earth and civil systems often involve inaccessible or hazardous environments (e.g., deep subsurface, remote ocean basins, or aging underground infrastructure), where direct measurement is difficult or incomplete~\cite{sun2021review,baecher1977statistical}. As a result, both domains suffer from sparse and noisy datasets that hinder model calibration and decision-making. For example, limited geological borehole data may be insufficient for accurate subsurface mapping, and historical infrastructure records may be missing or outdated. While LLMs can help process available documents, they cannot compensate for missing sensor data, field measurements, or satellite coverage, which are essential for building reliable models or simulations.

    \item \textbf{Physical Implementation and Experimental Validation.} In civil engineering especially, the ultimate test of a solution lies in its physical realization—constructing structures, monitoring performance, and conducting stress tests under real-world conditions~\cite{hodge2014wireless,helal2015non}. Earth scientists similarly depend on empirical fieldwork, such as drilling ice cores or deploying seismographs. These tasks involve tools, logistics, and materials that LLMs cannot access or control. While LLMs may assist in designing monitoring protocols or reviewing standards, they cannot carry out experiments or validate hypotheses in the field.

    \item \textbf{Cross-scale Integration.} A recurring challenge in both fields is integrating phenomena across vastly different spatial and temporal scales—for instance, linking microscale soil composition to large-scale slope stability, or connecting millennial-scale climate changes to today’s hydrological models~\cite{mcguffie2001forty,liu2022review}. This requires sophisticated multi-scale modeling, often coupling discrete and continuous frameworks, something well beyond the representational capacity of LLMs trained primarily on textual corpora. Such integration typically involves customized simulations and domain-specific algorithms informed by physics and engineering principles.

    \item \textbf{Regulatory and Ethical Constraints.} Civil engineers must design within strict regulatory, economic, and ethical frameworks—ensuring safety, sustainability, and community impact are considered~\cite{tokognon2017structural}. Earth scientists face ethical concerns in interventions like geoengineering or resource extraction. While LLMs can provide policy summaries or ethical perspectives from the literature, they cannot weigh trade-offs, assess context-specific risks, or make normative judgments. These decisions require human oversight, societal debate, and legal frameworks beyond what AI can resolve.
\end{itemize}

\noindent\textbf{Easier with LLMs: The Parts That Move.}

Despite these limitations, there are many areas in which LLMs can meaningfully assist researchers and practitioners in earth sciences and civil engineering—particularly in tasks related to knowledge synthesis, documentation, and early-stage design:

\begin{itemize}[leftmargin=10pt]
    \item \textbf{Literature Review and Standards Summarization.} Both fields rely on vast and highly technical bodies of knowledge, including regulatory documents, geological surveys, engineering design codes, and academic papers. LLMs can significantly streamline the literature review process by summarizing scientific reports, extracting key metrics, and comparing standards across regions~\cite{agarwal2024litllm,glickman2024ai}. For instance, an LLM could help an engineer quickly retrieve the seismic design requirements for bridges in a specific country or summarize recent advances in landslide risk prediction models.

    \item \textbf{Interpreting Sensor and Field Reports.} As structural health monitoring (SHM) and geoscience increasingly adopt IoT and sensor networks, LLMs can help translate raw sensor metadata or technician logs into structured, actionable insights~\cite{tokognon2017structural,sari2019road}. They can assist in automatically annotating inspection reports, flagging anomalies in sensor readings, or identifying trends across multiple sources, especially when integrated with domain-specific tools.

    \item \textbf{Assisting in Conceptual and Parametric Design.} In civil engineering tasks such as structural layout or materials selection, LLMs can be useful in generating design suggestions, proposing parametric options, or reviewing existing case studies. For example, they can draft potential designs based on textual constraints or help identify suitable sustainable materials from engineering databases~\cite{tsiptsis2019structural,zheng2020new}.

\end{itemize}

In conclusion, earth sciences and civil engineering face enduring challenges related to physical constraints, empirical validation, and systemic complexity. While LLMs are unlikely to replace domain experts or physical experimentation, they are emerging as valuable assistants in literature synthesis, document interpretation, early-stage design, and communication—accelerating workflows and enhancing accessibility. The future of these disciplines may lie in effective collaboration between human ingenuity, physical modeling, and intelligent AI support systems.

\begin{figure}[!t]
    \centering
    \includegraphics[width=\linewidth]{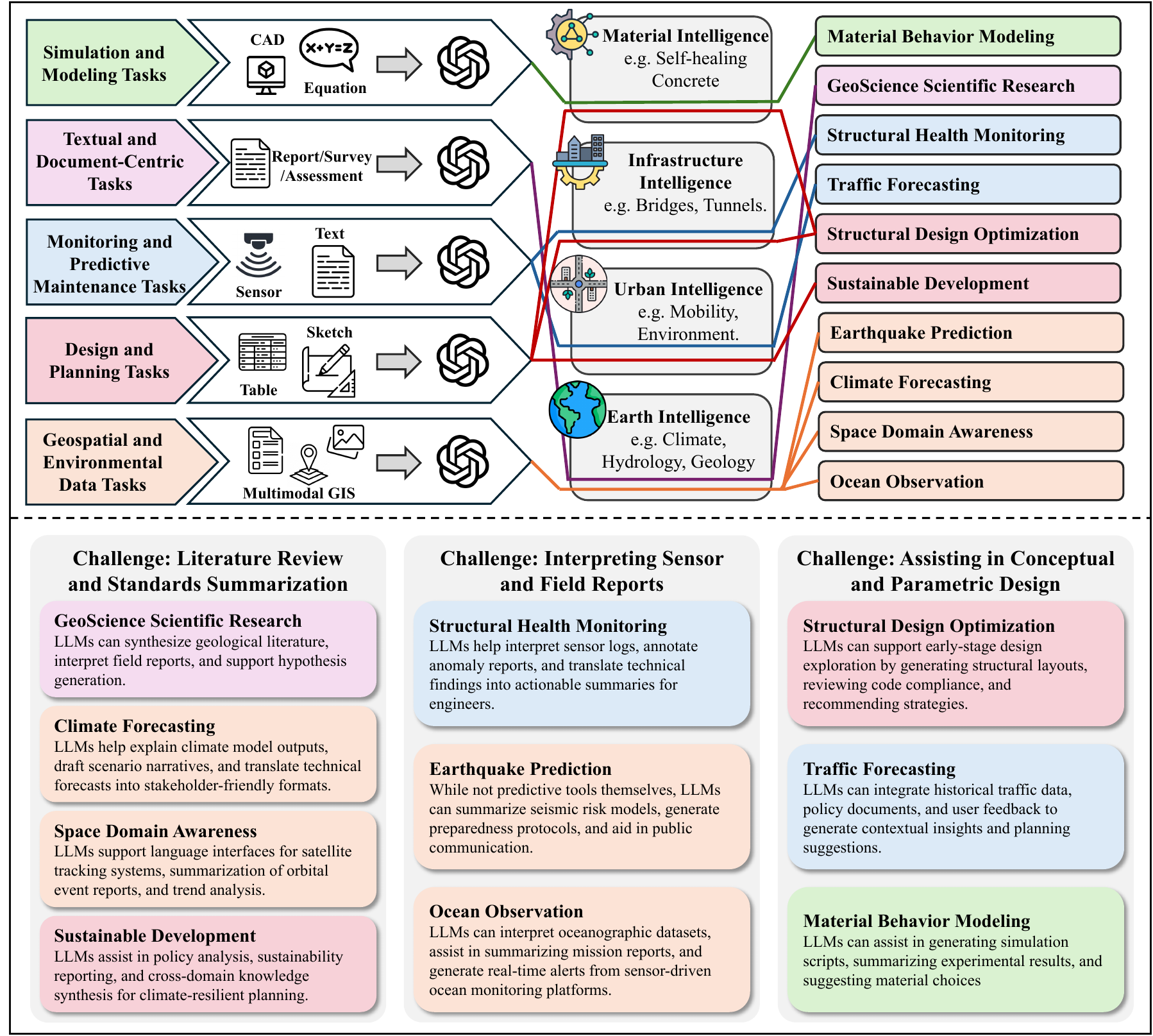}
    \caption{The pipelines of Earth sciences and civil engineering.}
    \label{fig:earch_civil_pipeline}
\end{figure}

\noindent\textbf{5.5.1.4 Taxonomy}

Research in Earth sciences and civil engineering encompasses a broad array of problems, spanning from natural system modeling to infrastructure design and monitoring. While traditionally organized by disciplinary boundaries (e.g., geology, hydrology, structural engineering), these fields also share deep methodological similarities, particularly in their use of spatial data, physical modeling, and complex environmental measurements~\cite{hoff2010greening,harle2024advancements}. Importantly, many tasks require interpreting unstructured field reports, processing geospatial sensor data, or modeling physical systems under uncertainty—domains where the potential role of LLMs is still being explored. To support more effective use of LLMs in these fields, we propose a taxonomy grounded in computational characteristics rather than disciplinary labels. This perspective helps reveal where LLMs can directly contribute, where they serve supporting roles, and where traditional numerical modeling remains essential. It offers three practical advantages:

\begin{itemize}[leftmargin=10pt]
    \item \textbf{Model compatibility}: Clearly distinguishes tasks suitable for LLMs (e.g., text synthesis) versus those requiring numerical simulation or high-dimensional optimization~\cite{wang2024accurate}.
    \item \textbf{Cross-domain transferability}: Highlights common computational structures across environmental and infrastructure problems, facilitating the reuse of AI workflows.
    \item \textbf{Pipeline orchestration}: Supports hybrid modeling pipelines where LLMs coordinate, explain, or augment scientific and engineering workflows using diverse data types.
\end{itemize}

\noindent\textbf{Geospatial and Environmental Data Tasks.}
These tasks involve interpreting and reasoning over georeferenced data from remote sensing, field instruments, or environmental simulations. Examples include analyzing satellite images to detect deforestation, processing LiDAR or GIS data to monitor urban growth, and interpreting topographic or hydrological maps for flood modeling~\cite{taylor2013ground,sun2021review}. Inputs typically consist of raster, vector, or tabular geospatial datasets; outputs may be spatial predictions (e.g., landslide zones), environmental classifications, or time-series trends. Although LLMs are not inherently designed for numerical geospatial data, they can assist in generating geospatial analysis scripts (e.g., in Python/ArcPy or QGIS), interpreting metadata or documentation, and explaining geostatistical outputs. Additionally, LLMs may serve as interfaces between users and GIS tools, enabling natural language-based queries about geospatial datasets~\cite{nascimento2024llm4ds}.

\noindent\textbf{Engineering Simulation and Physical Modeling Tasks.}
This category includes tasks that rely on numerical simulations to model physical phenomena, such as seismic wave propagation, structural stress analysis, fluid flow in soil, or traffic modeling. Inputs often include CAD models, finite element meshes, or system equations; outputs range from displacement fields to safety margins or failure probabilities~\cite{mcguffie2001forty,helal2015non,polym12040818}. These tasks are typically computationally intensive and governed by physics-based principles (e.g., Navier–Stokes or elasticity theory), which are poorly suited to LLMs. However, LLMs can still assist by auto-generating simulation code, translating natural-language specifications into solver-ready scripts (e.g., ANSYS or Abaqus macros), and summarizing simulation results for decision-makers. In this sense, they act as connective tissue between engineering intent and computational execution.

\noindent\textbf{Textual and Document-Centric Tasks.}
A substantial portion of Earth science and civil engineering knowledge is encoded in unstructured documents: geological surveys, environmental impact assessments, building codes, inspection reports, and academic literature. Tasks in this domain include extracting key findings from regulatory reports, summarizing design guidelines, or generating technical proposals~\cite{tokognon2017structural,harle2024advancements}. Inputs are typically plain text, PDFs, or semi-structured tables, with outputs including summaries, structured annotations, or natural language responses. These tasks represent the natural strength of LLMs. For instance, an LLM can assist an engineer in drafting a permit application by synthesizing local zoning codes and environmental constraints or help a geologist extract lithology descriptions from archival drilling logs.

\noindent\textbf{Monitoring and Predictive Maintenance Tasks (Hybrid).}
This class includes tasks where sensor data and models are combined to anticipate failures or evaluate performance. Examples include structural health monitoring (e.g., identifying bridge fatigue), monitoring dam stability, predicting groundwater level fluctuations, or evaluating pavement degradation from satellite imagery and sensor inputs~\cite{cai2012structural,sari2019road,busby1976manned}. Inputs span across time-series sensor data, weather models, images, and inspection text logs. Outputs typically include alerts, risk scores, or suggested interventions. While predictive modeling often depends on statistical learning or physics-based simulations, LLMs can play important interpretive roles: translating sensor anomalies into natural-language diagnostics, integrating multi-source logs into incident summaries, or suggesting maintenance schedules by referencing historical case reports.

\noindent\textbf{Design and Planning Tasks.}
These tasks involve defining, reasoning about, and communicating high-level engineering plans—such as choosing structural layouts, evaluating site suitability, or balancing cost and safety trade-offs in urban planning. Inputs include sketches, specifications, tabular datasets, and design goals, while outputs may be schematic proposals, material selections, or scenario comparisons. Though the core design logic is domain-driven and often quantitative, LLMs can support rapid ideation, generate initial alternatives, or provide justifications based on case studies and standards~\cite{tsiptsis2019structural,zheng2020new}. They can also assist in multi-objective trade-off analysis, especially when paired with simulation tools and optimization libraries.

\input{tables/earth_science_civil_engineer}

\subsubsection{Geospatial and Environmental Data Tasks}

Geospatial and environmental data analysis focuses on data processing and interpretation, in order to assist researchers in data understanding. The primary inputs to these tasks are geospatial datasets, such as GIS data, satellite images and hydrological maps. These datasets can not be easily understood by humans. For instance, GIS data is usually in form of vector and raster, which is not as intuitive as natural language. Besides, despite the existence of APIs for data processing, practitioners are still plagued by challenges such as the accessibility of software and technological complexity. Interacting with these data through textual information can make them more accessible and understandable for researchers, which is essential for efficient and automated data processing.

Researchers have explored the incorporation of LLMs into the processing and analysis of geospatial data. The initial efforts mainly use LLMs to process text-based inputs to assist GIS tasks~\cite{li2023autonomous,deng2024k2,lin2023geogalactica,zhang2024mapgpt,roberts2023gpt4geo}. For instance, ~\cite{li2023autonomous} introduced Autonomous GIS, an LLM-powered GIS to solve geospatial tasks including toponym recognition, location description and time series forecasting. ~\cite{lin2023geogalactica} introduced GEOGALACTICA to compensate geoscience-specific knowledge to general LLMs. These studies demonstrated the potential of LLM in geospatial data tasks. Considering that visual information is critical in the analysis of certain geospatial data like remote sensing data, the application of large vision-language models (VLMs) has emerged as a promising research direction. However, due to the unique characteristics of remote sensing images, such as high resolution, diverse scales, and complex acquisition angles, existing general VLMs perform poorly in remote sensing tasks. To address these issues, several studies have been conducted. Pioneering work RSGPT~\cite{hu2025rsgpt} constructed a high-quality human-annotated Remote Sensing Image Captioning dataset (RSICap) and introduced RSGPT, a GPT-based model fine-tuned on RSICap with advanced performance in image captioning and visual question-answering tasks. RS-LLaVA~\cite{bazi2024rs} improved LLaVA for remote sensing imagery and created the RS-instructions dataset by integrating four single-task datasets related to captioning and VQA. GeoChat~\cite{kuckreja2024geochat} further built a novel RS multimodal instruction-following dataset and introduced the first versatile remote sensing VLM with multitask conversational capabilities, which can answer image-level and region-specific queries and generate visually grounded responses. EarthGPT~\cite{zhang2024earthgpt} integrated a visual-enhanced perception mechanism, a cross-modal mutual comprehension approach, and a unified instruction tuning method to come up with a universal multimodal LLM for multisensor remote sensing image comprehension.

Despite the remarkable achievements of these models in various geospatial tasks, the challenges within geospatial and environmental data analysis lie not only in data comprehension, but dadata selection and the utilization of professional tools like APIs like APIs. Recent studies attempt to convert LLMs from an operator to a decision-maker~\cite{zhang2023geogpt,singh2024geollm}. In other words, they leverage the reasoning capability of LLMs to assess appropriate tools and generate executable programs and generate executable programs based on given requirements before addringessing downstream tasks. As one of the earliest studies, GeoGPT~\cite{zhang2023geogpt} is designed to automate geospatial tasks. It uses an LLM to understand user demands from natural language descriptions and then selects and executes appropriate GIS tools from a pre-defined pool to conduct procedures like data collection, processing, and analysis. Change-Agent~\cite{liu2024change} can follow user instructions to comprehensively interpret and analyze remote sensing change, which integrates a multilevel change interpretation (MCI) model as the toolbox and an LLM as the "brain". GeoLLM-Engine~\cite{singh2024geollm} further equips more geospatial API tools and external knowledge bases, enabling agents to handle more complex tasks. ~\cite{ning2025autonomous} designed a GIS agent framework for autonomous geospatial data retrieval, consisting of a data source index and a handbook inventory. It uses an LLM as the decision-maker to select appropriate data sources from a pre-defined list and generate programs to fetch data. To further enhance agents' ability to process complex sequential tasks and avoid hallucination caused by the lack of domain knowledge, ~\cite{chen2024llm} proposed a novel interactive framework GeoAgent, integrating a code interpreter, static analysis, and Retrieval-Augmented Generation (RAG) within a Monte Carlo Tree Search (MCTS) algorithm; ~\cite{hou2024chain} proposed a Chain-of-Programming (CoP) framework, decomposing API selection and code generation process into five separate steps and incorporating external knowledge base and user feedback; ~\cite{hou2025geocode} proposed GeoCode-GPT specifically for geospatial data analysis code generation, which is pre-trained and fine-tuned on their built GeoCode geospatial code corpus. 

In summary, the evolution of geospatial LLMs has clearly progressed from "workers" to "experts". They are required to formulate a complete solution specific to user demands, and meanwhile generate executable programs to obtain the expected outcomes, which significantly helps improve data processing efficiency. However, they still suffer from limited domain knowledge, inferior data quality and inability to handle complex sequential reasoning.

\subsubsection{Engineering Simulation and Physical Modeling Tasks}

Engineering simulation and physical modeling also focus on data understanding and interpretation. Compared with geoscience tasks, the primary inputs are in more complex forms, like CAD models and system equations. It is highly challenging to achieve automatic simulation with LLMs, and the reasons are presented as follows: (1) Numerical modeling of structures is complex, involving numerous inputs, simulation configurations and closely coupled sets of equations in calculation~\cite{JIANG2024123431,ZHANG2024114788}. Even the slightest error can not be tolerated in this process. (2) General LLMs only have a vague understanding of the physical and dynamic rules of the real world~\cite{cherian2024llmphy}. However, the translation between natural language and simulation code is essential for reducing modeling efforts and facilitating automated data and modeling workflow. 

Although significant progress has been made in LLM-based code generation, LLM-assisted numerical simulation for earth science and civil engineering is still under-explored. Relevant research can be divided into two branches, simulation code generation and user interface for automatic simulation. Existing LLM-based simulation code generation are usually aimed at specific applications~\cite{ZHANG2024114788,pursnani2024hydrosuite}. For instance, ~\cite{JIANG2024123431} proposes an automated building modeling platform Eplus-LLM that uses a fine-tuned T5 to translate natural language descriptions of buildings into EnergyPlus models. HydroSuite-AI~\cite{pursnani2024hydrosuite} is designed for hydrology and environmental science research, which integrates HydroLang, HydroCompute, and HydroRTC open-source libraries to help researchers generate code snippets. GeoSim.AI~\cite{bekele2025geosim} designs a suite of AI assistants for numerical simulations in geomechanics, translating natural language and image inputs into simulation parameters and scripts. On the other hand, there are other researchers using natural language descriptions to guide LLM and conduct automatic simulation. ~\cite{SONG2024114983} proposes a GPT-based building performance simulation system that integrates simulation engines and data analytics in the GPT environment. ~\cite{10770822} designs an LLM-based agent ChatSUMO to simplify SUMO simulations for traffic modeling, which can convert user text inputs into relevant keywords for running Python scripts and generating traffic simulation scenarios. 

While these advances are promising, only relatively simple modeling cases have been studied, and further enhancement is needed for more complex practical modeling. There are also concerns about model transparency and inability to handle interdependent user requirements. Nonetheless, the role of LLMs as simulation assistants is becoming increasingly practical in early prototyping studies.

\subsubsection{Textual and Document-Centric Tasks}
\label{sec:ec_text}

Textual and document-centric tasks mainly focus on knowledge extraction and translation for unstructured textual data. Knowledge extraction aims to extract required information from textual data like government documents and inspection reports, while knowledge translation aims to translate raw text data into other data forms to make them more understandable by humans or computers. 

Although some studies have suggested that LLMs encoded preliminary geoscience knowledge from their training corpora~\cite{bhandari2023large,haas2023learning,mooney2023towards}, the lack of domain knowledge still hinders the application of LLMs in earch science and civil engineering due to the unique properties of relevant textual data. Therefore, various strategies have been proposed to adapt general LLMs to achieve domain-specific question answering and knowledge extraction~\cite{deng2024k2,lin2023geogalactica}. For instance, GeoBERT~\cite{denli2021geoscience} retrained BERT with geoscientific record dataset, which was further fine-tuned for geoscience question answering and query-based summarization task. ~\cite{bi2023oceangpt} proposed the first-ever LLM specialized for ocean science tasks. ~\cite{zhang2024bb} constructed a new pretraining dataset BB-GeoPT with GIS-specific knowledge, retrained an open-source LLM and obtained BB-GeoGPT, an LLM for GIS domain.~\cite{dong2024geo} customized an RAG-enhanced LLM to efficiently process geological document information.

Compliance check is one of the knowledge translation tasks that receives growing attention~\cite{buildings14071983,wan2024generative,kumar2024architectural}, which aims to ensure that built structures or environment are free of safety risks and consistent with design regulations and industrial standards. The application of LLMs in automatic compliance check is still limited. Pioneering work~\cite{zheng2023llm} proposes LLM-FuncMapper to identify predefined atomic functions for regulatory clauses interpretation with the help of LLMs. ~\cite{liu2023gpt} adopts GPT-based models and achieves automatic compliance check through prompt engineering. ~\cite{PU2024124601} presents the AutoRepo framework, which combines unmanned vehicles and multimodal LLMs for automated construction inspection report generation. To further automate regulatory text processing and improve efficiency, ~\cite{buildings14071983} integrates deep learning models, LLMs and ontology knowledge models, which are responsible for preliminary text classification, structured information extraction and compliance check respectively.
Textual data only accounts for a small portion of all geoscience and urban science data, and there are several cross-modality knowledge translation tasks have been discussed, including urban profiling~\cite{yan2023urban} and traffic report summarization~\cite{da2023llm}.

In summary, the introduction of LLMs has made significant progress in textual and document-centric tasks, which has laid foundation for further development of domain-specific foundation models. However, existing methods still suffer from hallucination and the lack of high-quality training data. 

\subsubsection{Monitoring and Predictive Maintenance Tasks}

Monitoring and maintenance tasks are responsible for the health management of environment and built structures, which is a critical stage of the lifecycle of bridges, residence and other research objects in civil engineering and earth science. Most of these tasks are within the scope of Operation and Maintenance (O\&M)~\cite{chen2024revolutionizing,LI2025115515}. O\&M is a complex systematic process, including data collection and analysis, performance prediction, condition assessment, strategy development and emergency response. There are also some other tasks within the construction stage, targeting for newly built structures. These procedures are used to be labor-intensive, which is inefficient and prone to errors. The introduction of LLMs into data processing and real-time inspection can significantly reduce the cost of time and data compared to traditional data mining methods. 

For newly built structures, automatic compliance check is a critical step of construction inspection, in order to make sure the structures are free of safety risks and consistent with design regulations and industrial standards. Detailed introduction can be referred to Section \ref{sec:ec_text}.
The application of LLMs is more extensive in O\&M, which focuses more on system optimization. Some existing works tend to enhance traditional O\&M tools with LLMs~\cite{FORTH2024110312,zhang2024large}. For example, \cite{ZHANG2024113877} proposes an automated data mining framework that combines maximal frequent itemset mining and GPT to detect energy waste patterns and optimize energy use. \cite{XIAO2024114691} proposes an RAG-enhanced LLM-based multi-agent framework to deal with unstructured building data and achieve automatic energy optimization. \cite{FORTH2024110312} using Semantic Textual Similarity and fine-tuned LLMs to automatically enrich Building Information Models (BIM) and assist building energy performance simulation. On the other hand, digital twin technologies have been widely applied to building maintenance and environmental monitoring~\cite{hosamo2022review,hamzah2024drone} , where LLMs can potentially make significant contributions to the construction of digital twin~\cite{10570372}. For instance, ~\cite{10590311} proposes an LLM-based digital twin solution to collect real-time human preference data and enhance the optimization of human-in-the-loop systems within the context of Cyber-Physical Systems and the Internet of Things (CPS-IoT). ~\cite{jia2024natural} explores LLM-driven sensor data acquisition protocol to achieve automatic capturing of real-time sensor data based on the requirements of user inputs. LLMs can also contribute to digital twin construction for railway health monitoring~\cite{ferdousi2024defecttwin}, smart transportation~\cite{hong2024llm}, water distribution network management~\cite{syed2024smart} and many other applications. 

Overall, researchers have made initial attempts to apply LLM techniques to monitoring and maintenance tasks, but current research still faces challenges like reliance on high-quality training data and limited transferability. The few-shot learning power of LLMs should be further exploited, and more application scenarios are worth exploring. 

\subsubsection{Design and Planning Tasks}

Design and planning are the beginning of lifecycles of both individual buildings and complex systems. The core of design and planning is trade-off. For building structural design, engineers aim to achieve the trade-off among safety, cost, aesthetics and other constraints. As for more complex urban planning, more requirements need to be considered, such as functional section arrangement and commute. LLMs can function as both data analyzers and user interface in this stage~\cite{urbanfoundation,qin2024intelligent,JIANG2024123431}. On the one hand, LLMs can analyze vast amount of collected data and come up with design verifications and recommendations. On the other hand, given user feedback and requirements, LLMs can generate inclusive solutions and conduct optimization accordingly.

The applications of LLMs in building structural design and optimization is still under-explored. LLMs have been applied to automatic design in this stage. In automatic design, attempts have been made to utilize LLMs as user interface, where LLMs accept user inputs and generate architectural details and simulation code. For instance, ~\cite{qin2024intelligent} uses an LLM as the core controller, which can interpret engineers' natural language descriptions and translate them into executable simulation code for shear wall structural design and optimization. ~\cite{jang2024automated} proposes the NADIA framework, which integrates LLMs and BIM authoring tools to enable architectural design consulting and detailing via natural language. However, the application of LLMs in other design tasks like building layout planning and design rendering has not been explored~\cite{JANG2025106174}. 

As for urban planning, researchers have shown growing interests in the development of urban foundation models to achieve urban general intelligence~\cite{zhang2024urban}. Although general LLMs have difficulty in dealing with urban science related textual information due to their unique properties in text style and knowledge requirements~\cite{zhu2024plangpt}, recent works have proved that LLMs can provide urban planners with valuable insights through proper fine-tuning. To name a few, PlanGPT~\cite{zhu2024plangpt} is the first specialized LLM for urban and spatial planning, which proposes a customized embedding model Plan-Emb and hierarchical search strategy Plan-HS for accurate information extraction and an LLM-based PlanAgent for tool invocation and information integration. ~\cite{zhou2024large} explores the application of LLMs in participatory urban planning, proposing an LLM-based multi-agent collaboration framework and a fishbowl discussion mechanism to simulate the communication between planners and residents. ~\cite{zhang2024trafficgpt,da2024open,lai2023llmlight} explores the applications of LLMs in transportation system, utilizing natural language prompts and the reasoning ability of LLMs to analyze traffic data and provide decision support for traffic control. TrafficGPT~\cite{zhang2024trafficgpt} integrates LLMs and traffic foundation models, while Open-ti~\cite{da2024open} bridges the gap between academic research and industry in traffic simulation and control via ChatZero control agent. 

Overall, the introduction of LLMs into design and planning tasks have been proved to significantly reduce the cost of time and human labor. However, researchers have only conducted preliminary explorations on this topic. More application scenarios could be explored, and the agents' inability to handle complex sequential actions need to be addressed in future research~\cite{da2024open}.

\input{tables/earth_civil_benchmark}

\input{tables/earth_civil_geobench}

\subsubsection{Benchmarks}

The application of LLMs in earth science and civil engineering is still quite limited, and available public benchmark datasets can not meet the requirements of various tasks in this field. Case studies and human evaluation are widely adopted for the evaluation of some domain-specific tasks, such as numerical simulation, compliance check, maintenance tasks and structural design. No effort has been made to construct a comprehensive benchmark for them. Therefore, in this section, we only summarize benchmarks for geoscientific understanding and urban planning. We will also select several representative benchmarks from Table~\ref{tab:earth_civil_bench} for in‐depth discussion. For each chosen benchmark, we will describe its scope and data form, and then survey the performance of representative LLMs on it. 

\input{tables/earth_civil_rsi}

\textbf{GeoBench.} GeoBench is the first benchmark for evaluating the ability of LLMs to understand and utilize geoscience knowledge to solve geoscientific problems. The authors collected pure text questions related to geology, geography and environmental science from NPEE and APTest examinations, including 1,578 objective questions and 939 subjective questions, which are represented by multi-choice questions and essay questions respectively. Objective questions are evaluated based on accuracy, while subjective questions are evaluated based on perplexity and GPTScore~\cite{fu2023gptscore}. Benchmark results are listed in Table \ref{tab:geobench}.

\input{tables/earth_civil_geocode}

\textbf{RSIEval.} RSIEval is constructed to assess VLMs in multi-modal remote sensing tasks, especially remote sensing specific image captioning and VQA. Images from the validation set of DOTA-v1.5 are selected, divided into 512×512 patches, and 100 patches are picked for manual annotation. It contains 100 high-quality image-caption pairs  and 936 diverse image-question-answer triplets. The questions used for VQA task are further divided into four categories. Object-related questions include presence, quantity, color, absolute position, relative position, area comparison, and road direction. Image-related questions involve high/low resolution and panchromatic/color images. Scene-related questions are about the main theme/scene and urban/rural scenes. Reasoning-related questions are like inferring the season of image capture, the dryness or wetness of the area, the moving speed of a ship, and the state of the water surface. Compared with previous RSVQA datasets, RSIEval provides a more diverse set of questions and answers, enabling a more comprehensive assessment. Benchmark results are listed in Table \ref{tab:rsieval_image} and Table \ref{tab:rsieval_vqa}.

\input{tables/earth_civil_plan}

\textbf{GeoCode.} GeoCode benchmark is used to evaluate LLM-based geospatial data analysis method, measuring whether LLMs can follow complex task instructions and generate proper code for data analysis. It contains 18,148 single-turn tasks and 1,356 multi-turn tasks, involving 2,313 function calls from 28 widely used Python libraries across 8 key domains. These tasks and function calls reflect complex instructions and implementation logic, requiring models to have strong compositional reasoning abilities. The evaluation of single-turn tasks focuses on the accuracy of function calls and the success rate of individual tasks, while multi-turn task evaluation emphasizes the task completion rate. Benchmark results are listed in Table \ref{tab:geocode}. Note that the authors did not conduct comprehensive evaluation for multi-turn tasks, and we only display single-turn task evaluation results here.

\textbf{PlanGPT.} PlanGPT benchmark is constructed to examine LLMs' capabilities in urban planning from different dimensions. Evaluation data is collected from multiple sources, including spatial planning documents from different administrative levels and authoritative textbooks. Four downstream tasks are considered for evaluation, including text generation, text style transfer, information extraction, and text evaluation. Text generation aims to evaluate the quality of the urban planning documents generated by the model; Text style transfer aims to evaluate the model's ability to convert text into the urban planning style; Information extraction aims to evaluate the model's ability to extract key information from text; Text evaluation is designed to evaluate the model's ability to evaluate urban planning proposals. Benchmark results are listed in Table \ref{tab:plangpt}.

\textbf{Summary.} In tasks related to textual data, ChatGPT and LLaMA consistently achieve satisfactory performance. In image-based tasks, LLaVA consistently outperforms other general LLMs. Besides, CodeGemma is capable of handling code generation tasks in the field of geoscience. However, the aforementioned benchmarks can not sufficiently evaluate more complex situations where LLMs are treated as user interface for knowledge retrieval, data analysis and numerical simulation. GeoLLM-Engine makes initial and valuable attempts by categorizing various downstream tasks based on user intents, while it is highly limited by the pre-defined system behavior functions. Further effort should be invested to achieve more extensive evaluation.

\subsubsection{Discussion}

\noindent\textbf{Opportunities and Impact.}
LLMs are beginning to reshape workflows in earth sciences and civil engineering, offering new capabilities for knowledge extraction, document synthesis, conceptual design, and early-stage simulation support. As shown across geospatial data processing~\cite{li2023autonomous, zhang2023geogpt}, engineering simulation~\cite{JIANG2024123431, bekele2025geosim}, document-centric tasks~\cite{dong2024geo, zhu2024plangpt}, monitoring and predictive maintenance~\cite{hosamo2022review, chen2024revolutionizing}, and design and planning~\cite{qin2024intelligent, zhang2024urban}, LLMs significantly lower the barriers to access technical knowledge, accelerate tedious documentation tasks, and enhance preliminary design ideation.

By serving as interfaces between domain experts and computational tools, LLMs can democratize access to specialized workflows, enabling faster prototyping, enhanced interdisciplinary collaboration, and more accessible education and training pipelines. Their ability to interpret natural language specifications, summarize complex regulatory documents, and support tool-assisted reasoning holds particular promise for infrastructure resilience, environmental monitoring, and sustainable urban development. Moreover, the emergence of domain-specific foundation models (e.g., PlanGPT~\cite{zhu2024plangpt}, EarthGPT~\cite{zhang2024earthgpt}) suggests that specialized LLMs could become integral components of future scientific and engineering ecosystems.

\noindent\textbf{Challenges and Limitations.}
Despite these opportunities, fundamental limitations persist when applying LLMs to earth sciences and civil engineering. Physical system modeling—ranging from seismic simulation to hydrodynamic forecasting—requires accurate, high-dimensional numerical computations governed by partial differential equations~\cite{mcguffie2001forty}. Current LLMs lack the inductive biases, precision, and physical fidelity needed for such tasks, relegating them to supporting roles in workflow orchestration rather than core simulation.

Similarly, data sparsity and uncertainty, particularly in hazardous or inaccessible environments~\cite{sun2021review, baecher1977statistical}, cannot be resolved through language-based reasoning alone. Real-world challenges like infrastructure resilience under extreme events or predictive maintenance of aging systems demand empirical validation and physical fieldwork beyond the reach of AI models.

Other notable concerns include the lack of transparency and verifiability in LLM outputs—particularly problematic for compliance verification and safety-critical applications~\cite{wan2024generative, kumar2024architectural}. Furthermore, issues such as regulatory complexity, ethical trade-offs in interventions (e.g., resource extraction, geoengineering), and societal value judgments require human oversight, collective deliberation, and domain-specific expertise that LLMs cannot replace.

\noindent\textbf{Research Directions.}
To maximize the impact of LLMs while respecting domain constraints, several promising research directions emerge:

\begin{itemize}[leftmargin=10pt]
    \item \textbf{Domain-Specific Fine-Tuning and Hybridization.} Building specialized LLMs for Earth sciences and civil engineering—such as those incorporating geospatial, structural, and regulatory corpora~\cite{zhang2024urban, dong2024geo}—can enhance factual grounding and reasoning capabilities, particularly when paired with physics-based simulators.
    
    \item \textbf{Integration with Physical Modeling Pipelines.} Rather than replacing numerical solvers, LLMs should serve as front-end interfaces and code generators for simulation frameworks, automating tedious configuration tasks and improving accessibility~\cite{JIANG2024123431, pursnani2024hydrosuite}.
    
    \item \textbf{Autonomous Decision-Making with Domain Constraints.} The development of geospatial agents~\cite{zhang2023geogpt, hou2024chain} that reason over available tools, APIs, and datasets presents a blueprint for autonomous yet domain-constrained action planning in complex scientific workflows.
    
    \item \textbf{Digital Twin and Smart Infrastructure Integration.} Leveraging LLMs for real-time monitoring, data fusion, and user interaction within digital twin environments offers a promising avenue for dynamic asset management, environmental monitoring, and resilience planning~\cite{hosamo2022review, hong2024llm}.
    
    \item \textbf{Ethical AI Frameworks for Infrastructure and Environmental Decisions.} Future research must address bias mitigation, explainability, regulatory compliance, and human-centered design when deploying LLM-augmented systems in critical infrastructure and environmental governance~\cite{wan2024generative}.
\end{itemize}

\noindent\textbf{Conclusion.}
LLMs offer powerful new tools for supporting and accelerating workflows in earth sciences and civil engineering, particularly in domains related to knowledge synthesis, document understanding, and conceptual design. However, realizing their full potential requires carefully integrating them into hybrid pipelines alongside empirical observation, physical modeling, and human expertise. Moving forward, the synergy between specialized LLMs, domain-specific simulations, and expert-driven validation will define the next generation of scientific discovery and infrastructure innovation, ensuring that technological advances align with societal needs, environmental stewardship, and engineering excellence.

\subsection{Computer Science and Electrical Engineering}

\subsubsection{Overview}
\textbf{Computer Science (CS).} 
CS is the study of computers and the algorithmic processes that support their operation~\cite{knuth1974computer, shagrir1999computer, cortina2007introduction}. 
In other words, CS studies all topics relevant to computers, which encompasses a broad range of focus, from fundamental computer principles to the design of hardware and software systems, along with diverse real-world applications. 
The interdisciplinary nature of CS~\cite{lehman2010mathematics}, explicitly drawing from mathematics, engineering, and logic, incorporating techniques from fields such as electronic circuit design, probability, and statistics, positions it as a central driving force behind innovation in numerous sectors and further enables CS to contribute to a wide array of applications, from scientific modeling to the development of intelligent systems.

\textbf{Electrical Engineering (EE).} 
EE focuses on the study and application of electricity, electronics, and electromagnetism~\cite{dorf1997electrical, rizzoni2009fundamentals}, covers a multitude of areas, including power engineering, telecommunications, control systems, electronics, signal processing, and computer engineering.
Fundamentally, EE are the architects and builders of the complex electrical and electronic infrastructure that sustains modern life~\cite{JSTSP2024toward}.
In particular, EE is about understanding and manipulating electricity to create practical solutions that benefit society.
This involves a wide range of activities, from the generation and distribution of electrical power that lights our homes and powers industries to the design of intricate electronic circuits found in everyday devices such as smartphones, computers, and vehicles~\cite{abdollahi2024hardware}.

As discussed above, CS and EE represent two distinct yet deeply interconnected disciplines. 
While CS primarily focuses on algorithms, software development, and computational theory, EE deals with the physical systems, hardware, and electronic components that enable computing. 
Their relationship is characterized by a complex interplay of shared foundations, complementary approaches, and evolving boundaries that continue to shape modern technology. 
Moreover, addressing every viewpoint in CS and EE is non-trivial, as these fields are pertinent to a wide range of areas.
Therefore, in this section, we focus on two fundamental topics in CS and EE, i.e., \textbf{software development and circuit design}.

\begin{figure}
    \centering
    \includegraphics[width=0.99\linewidth]{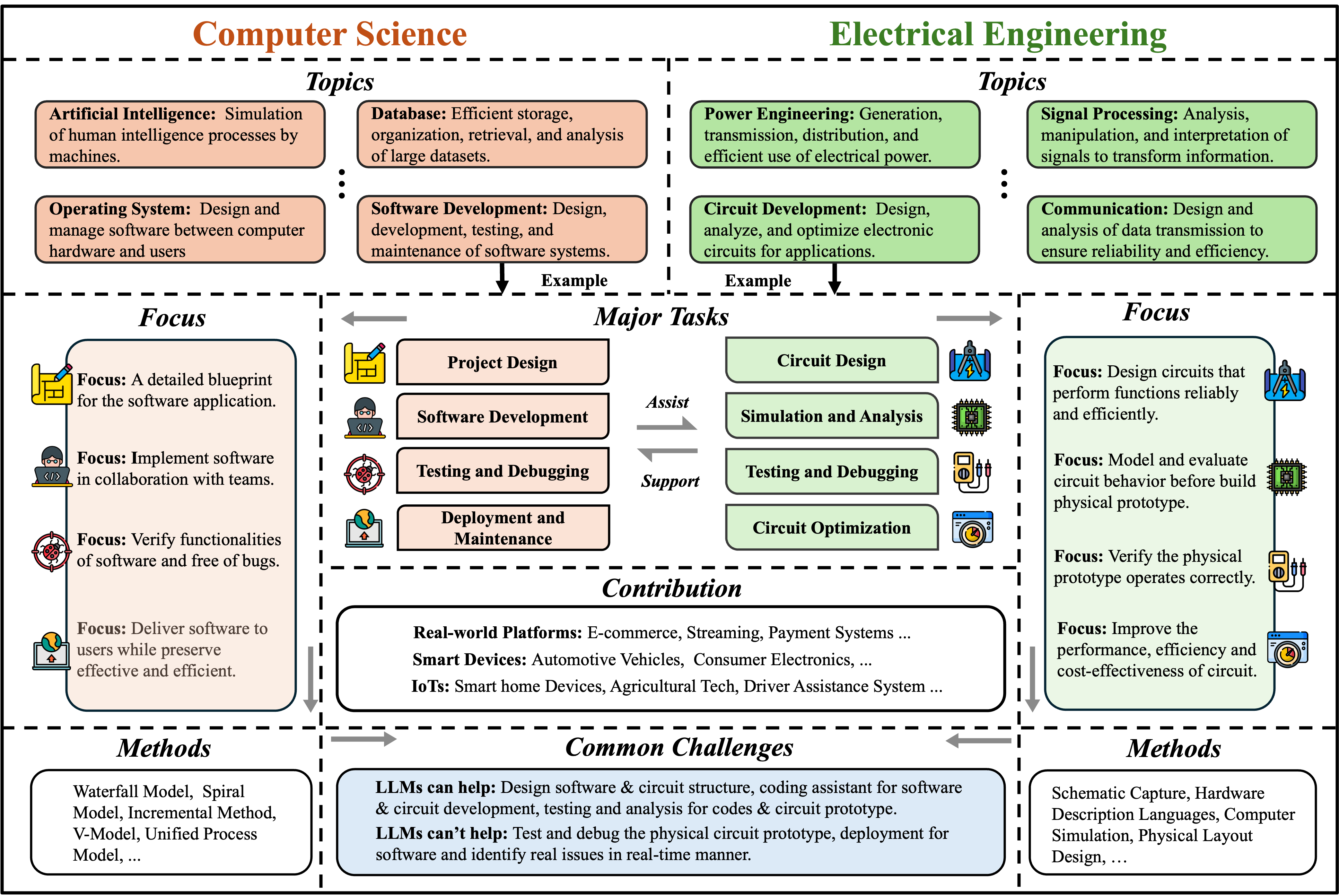}
    \caption{The relationships between software development in computer science and circuit design in electrical engineering.}
    \label{fig:csee_diagram}
    \vspace{-10pt}
\end{figure}

\textbf{Conventional Software Development.} 
Traditional programming approaches typically involve manual coding via programming languages, following the software development life cycle (SDLC). 
These methods have evolved from low-level machine code to high-level languages, but they generally rely on human developers to write, test, and debug code line by line~\cite{liu2024empirical, thorne2013reducing}.
Among traditional methodologies, several models have been widely adopted, including the Waterfall model, the Spiral model, the V-model, the Incremental Approach, and the Unified Process model~\cite{stoica2013software}. 
Next, we will highlight each model in detail:
\begin{itemize}[leftmargin=*]
    \item \textbf{Waterfall Model}: 
    It represents a linear, sequential approach where each phase of the program development lifecycle, from requirements gathering to maintenance, must be completed before the next phase can begin~\cite{awad2005comparison, risener2022study, leong2023hybrid}.
    This model assumes that all requirements can be clearly defined and understood at the outset of the project. 
    While straightforward and easy to manage for small projects with stable requirements, the Waterfall model struggles with complex projects or those where requirements are likely to change~\cite{stoica2013software}. 
    Its rigidity makes this model difficult to accommodate new requirements or to revisit earlier phases once completed, leading to potential issues and increased costs if errors or omissions are discovered late in the development cycle~\cite{pawar2015comparative}. 
    The lack of early prototypes also hinders client feedback, potentially resulting in a final product that does not fully meet user needs.
    \item \textbf{Spiral Model}: 
    It offers a risk-driven approach that combines elements of the Waterfall model with iterative development~\cite{boehm1986spiral, leong2023hybrid}. 
    Projects in the Spiral model progress through several iterations, with each iteration involving planning, risk analysis, engineering, and evaluation. 
    By explicitly addressing risks at each stage, this model is better suited for complex projects with high potential risks compared to the linear Waterfall model~\cite{boehm2002spiral, risener2022study}.
    The cyclical nature allows for revisiting and refining requirements and designs based on ongoing risk assessment, leading to a more adaptable outcome for projects where risks are a significant concern~\cite{leong2023hybrid}.
    The success of the Spiral Model heavily relies on accurate and thorough risk assessments at each iteration. 
    If risk analysis is inadequate or incorrect, the project may suffer from poor planning and execution.
    Moreover, specialized knowledge in risk management is essential, but this expertise is not always available within every team
    \item \textbf{V-Model}:
    This is an extension of the Waterfall model that emphasizes the relationship between each development phase and a corresponding testing phase, focusing on verification and validation~\cite{balaji2012waterfall, gracias2024transitioning}. 
    Executed in a V-shape sequence, this model ensures that testing activities are planned and executed in parallel with the development stages, highlighting the importance of quality assurance throughout the lifecycle.
    By linking verification and validation to each stage, the V-Model aims to improve software quality and ensure that the final product meets both the specified requirements and the user's needs.
    \item \textbf{Incremental Approach}:
    This approach involves developing a program in a series of functional increments~\cite{shaikh2019comparison}. 
    Each increment adds more functionality to the previous ones, allowing for early delivery of working software to the client and the incorporation of feedback in subsequent increments. 
    This approach offers greater flexibility compared to the Waterfall model, enabling the development team to adapt to changing requirements and reduce the risk associated with large, monolithic development efforts by delivering working software in smaller, manageable parts~\cite{stoica2013software}.
    \item \textbf{Unified Process Model}: 
    This model is an iterative and incremental program development framework driven by use cases, centered around architecture, and supporting component-based design~\cite{shaikh2019comparison}. 
    Utilizing the Unified Modelling Language to represent various phases and deliverables, the unified process model emphasizes iterative development, where the software evolves over multiple cycles, and is built piece by piece. 
    This framework provides a flexible and adaptable approach suitable for a wide range of software projects, focusing on continuous improvement and active stakeholder involvement through use-case-driven development and an architectural focus that promotes a robust and scalable system design.
\end{itemize}

Beyond aforementioned conventional methodologies, the program development process commonly involves \textbf{manually coding}, where developers write source code in programming languages such as C/C++, Java, and Python, via tools like compilers, interpreters, and debuggers, while adhering to organizational coding standards~\cite{stoica2013software}.
This manual translation of design specifications into executable code is inherently susceptible to human errors in logic, syntax, and implementation, particularly as the complexity of the program increases.
\textbf{Debugging} relies on developers manually stepping through the code via debuggers, setting breakpoints to pause execution at specific points, and inspecting the variables to understand the program states and identify the causes of errors~\cite{roychoudhury2010debugging, fitzgerald2008debugging}. 
Techniques like inserting print statements at strategic locations in the code and conducting manual code reviews are also commonly employed to aid in the debugging process.
However, these methods can prove particularly inefficient in large and complex systems, often requiring significant time and effort to trace the execution flow and pinpoint the source of bugs~\cite{fitzgerald2008debugging}.
Afterward, \textbf{testing strategies} in traditional SDLC models involve a distinct phase that occurs after or in parallel with the implementation of the program. 
This typically includes creating detailed test plans and test cases based on the software requirements, followed by the execution of these tests to identify defects and ensure that the program meets the specified standards. 
Testing in traditional methodologies is characterized as "difficult and late", implying that critical bugs are frequently discovered towards the end of the development cycle, making them more costly and time-consuming to rectify. 
Furthermore, clients are not typically involved in the coding and testing stages of traditional software development~\cite{stoica2013software}.

\textbf{Traditional Circuit Design.}
Next, we discuss the circuit design in EE.
The design of digital circuits has traditionally relied on a set of well-established methodologies to transform high-level specifications into physical hardware.
These methods include schematic capture, the utilization of Hardware Description Languages (HDLs)~\cite{weste2015cmos, palnitkar2003verilog, ashenden2007digital} such as VHDL~\cite{weste2015cmos} and Verilog~\cite{palnitkar2003verilog, thomas2008verilog}, simulation techniques, and the process of physical layout design:
\begin{itemize}[leftmargin=*]
    \item \textbf{Schematic Capture:}
    It involves the graphical creation of a circuit diagram using Electronic Design Automation (EDA) tools~\cite{haas2007ensuring, lin2019beyond}.  
        In this approach, designers place and connect symbols that represent electronic components to construct the circuit architecture and interconnections visually~\cite{nelson2012using}. 
    While schematic capture provides an intuitive way to represent smaller circuits, it becomes increasingly cumbersome and less scalable for complex designs containing a large number of components~\cite{jones2015less}. 
    Managing the intricate web of interconnections and ensuring accuracy through a purely graphical representation can be challenging for large integrated circuits.
    \item \textbf{Hardware Description Languages (HDLs):}
    HDLs, such as VHSIC HDL (VHDL)~\cite{dewey1983vhsic} and Verilog~\cite{thomas2008verilog}, are textual languages used to describe the behavior and structure of digital circuits at abstraction levels, from the high-level Register Transfer Level down to the gate level. 
    Specifically, VHDL was originally designed for military applications and is popular for its strong typing and structured syntax, making it a favored solution for large and critical projects. 
    Verilog, on the other hand, gained widespread adoption in the semiconductor industry for the design and verification of Application-Specific Integrated Circuits (ASICs) and Field-Programmable Gate Arrays (FPGAs) due to its relative simplicity and ease of use. 
    HDLs allow circuit designers to describe the flow of digital signals and the logical operations performed among them~\cite{mano_ciletti_2012, vemeko_2024}.
    As the textual representation facilitates the use of automated EDA tools for simulation and synthesis, HDLs offer a more scalable and manageable approach for designing complex digital systems compared to schematic capture. 
    \item \textbf{Simulation:}
    Simulation techniques are crucial in traditional digital circuit design for verifying the functionality and timing of a circuit described in an HDL before it is physically implemented.
    This process involves test benches creation (sets of input stimuli applied to the circuit model) and result monitoring (ensuring the design behaves as intended). 
    Although indispensable for early detection of design errors, the simulation of large, complex circuits is often computationally demanding and time-consuming.
    In particular, thorough verification requires the careful planning and execution of extensive test cases to cover all possible operating conditions and scenarios, which can be a significant challenge for intricate designs.
    \item \textbf{Physical Layout Design:}
    At this stage, the abstract circuit design is transformed into a physical implementation on a substrate, such as a silicon die for an integrated circuit or a printed circuit board (PCB). 
    This process involves determining the precise placement of components and the routing of the conductive interconnections between them, taking into account various factors such as signal integrity, power distribution, thermal management, and manufacturability. 
    The physical layout significantly impacts the final performance, reliability, and cost of the circuit. Traditionally, achieving an efficient and effective physical layout, especially for high-performance designs, requires significant manual effort, expertise, and often involves iterative refinement to optimize for various conflicting requirements like area, speed, and power consumption.
\end{itemize}

\textbf{Limitations and Challenges.}
Traditional methodologies in programming and circuit design, despite their long-standing use, face several limitations and challenges in the context of modern software development and increasing demands for complexity, performance, and faster time-to-market.
From a programming perspective, these limitations relate to scalability, the complexity of large codebases, the time-consuming nature of debugging, and the inherent potential for human error.
For circuit design, these challenges include managing the increasing complexity of integrated circuits, the difficulty in verifying and testing these complex designs, the often long design cycles, and the high level of specialized experience required.

\textbf{LLMs as solutions.}
LLMs offer potential solutions to several limitations inherent in traditional software programming methodologies. 
Their ability to generate and understand code, along with their natural language processing skills, can be leveraged to address issues related to routine coding tasks, debugging, code complexity, and human error.
Moreover, the application of LLMs in digital circuit design is an emerging and rapidly evolving field. 
Researchers are exploring the potential of LLMs to tackle some of the limitations of traditional hardware design methodologies in areas such as HDL code generation, functional verification, and high-level synthesis.
To summarize the main applications of LLMs in CSEE, we introduce the following taxonomy.
(i) Code Generation Tasks; (ii) Code Assistant in Debugging; (ii) Code Analysis in Codebases; (iii) HDL Code Generation; (iv) Functional Verification; (v) High-Level Synthesis (HLS).

\subsubsection{Code Generation Tasks}
Several studies have demonstrated that LLMs excel at code generation tasks~\cite{hou2024large, belzner2023large, fan2023large}. 
By generating boilerplate code, standard data structures, common algorithms, and even entire functions based on natural language descriptions or specifications, LLMs can significantly reduce the amount of manual coding required from developers. 
Tools such as GitHub Copilot have already demonstrated the potential for substantial time savings by automating the creation of routine code elements~\cite{ma2023ai}.
This automation allows developers to focus their efforts on more complex and creative problem-solving, rather than spending time on repetitive and often error-prone coding.

Several recent studies~\cite{peng2023impact, song2024impact} have systematically conducted empirical experiments to demonstrate the effectiveness of LLMs in assisting developers with code generation tasks.
Peng et al.~\cite{peng2023impact} conducted an experiment involving 95 developers from diverse backgrounds, dividing them into treatment and control groups. The results showed that, among those who completed the task, the average completion time was 71.17 minutes for the treatment group and 160.89 minutes for the control group—a 55.8\% reduction in completion time. The t-test yielded a p-value of 0.0017, with a 95\% confidence interval for the improvement ranging from [21\%, 89\%]. Notably, there were four outliers with completion times exceeding 300 minutes, all of whom were in the control group; the findings remain robust even when these outliers are excluded.
These results indicate that Copilot significantly enhances average productivity in the studied population.

\subsubsection{Code Assistant in Debugging}
LLMs also play a crucial role in assisting with debugging~\cite{zhong2024debug, levin2024chatdbg, lee2024unified, majdoub2024debugging}. 
For example, LDB~\cite{zhong2024debug} introduced a step-by-step LLM debugger that mimics human reasoning by verifying runtime execution, showing that LLMs can systematically analyse program states to guide developers through complex bugs.
By analysing code and error messages, LLMs can help developers understand the nature of the problem, identify potential causes, and even suggest fixes~\cite{yan2024better, jiang2024ledex, yang2024enhancing}. 
ChatDBG~\cite{levin2024chatdbg} demonstrated that an AI-powered assistant can interpret error logs, recommend debugging strategies, and interactively answer developer queries, significantly reducing the cognitive load during troubleshooting.
Moreover, LLMs' ability to understand the context of the code and the meaning of error messages can be particularly valuable in pinpointing the root cause of issues more quickly than traditional manual debugging techniques.
Recent work by Yan et al.~\cite{yan2024better} showed that combining static analysis with LLM reasoning leads to more explainable and accurate localisation of crashing faults, while LeDex~\cite{jiang2024ledex} trained LLMs to self-debug and explain their reasoning, further improving transparency and reliability in automated debugging.
Furthermore, some LLMs are being developed with the capability to automatically identify and even repair certain types of bugs.
By automating code generation and completion, LLMs can contribute to mitigating human error in software programming. 
For instance, UniDebugger~\cite{lee2024unified} proposed a multi-agent LLM-based system that leverages the synergy of several models to collaboratively diagnose and fix bugs, outperforming single-agent approaches. 
Another study~\cite{majdoub2024debugging} provided a comprehensive evaluation of open-source LLMs, highlighting their growing competence in real-world debugging tasks.
The real-time code suggestions and automated generation capabilities can reduce the likelihood of common mistakes such as typos, syntax errors, and simple logical errors. 
COAST~\cite{yang2024enhancing} further showed that refining LLM training data using communicative agent strategies can substantially boost the model’s ability to catch and correct subtle logical errors, expanding the practical utility of LLMs in everyday development.
By predicting and generating code based on learned patterns and best practices, LLMs can help enforce consistency and reduce the introduction of errors during the coding process.

\subsubsection{Code Analysis in Codebases}
Addressing the complexity of large codebases is another area where LLMs can offer assistance \cite{bhattacharya2023exploring, arakelyan2023exploring, li2022generation}. 
By generating natural language explanations of code snippets and providing summaries of code functionality, LLMs can make it easier for developers to understand and navigate complex or unfamiliar code. 
This can be particularly helpful when working with legacy systems or when onboarding new team members. 
Additionally, LLMs can potentially aid in refactoring code by suggesting improvements to its structure, identifying areas of high complexity, and even generating refactored code, thereby making large codebases more manageable and maintainable. 

\subsubsection{Hardware Description Language Code Generation}
LLMs can potentially help in generating HDL code to manage the increasing complexity of digital circuits. 
By translating natural language descriptions of hardware functionality into Verilog or VHDL, LLMs automate the code generation for complex modules and systems, allowing designers to focus on higher-level architectural decisions. 
While challenges remain in preserving the functional correctness and code generation efficiency, this direction holds the promise to reduce manual coding effort and accelerate the design process.
For example, AutoChip~\cite{thakur2023autochip} introduces an automated, feedback-driven method that utilizes LLMs to generate HDL. 
It combines LLMs with output from Verilog compilers to iteratively complete the design. 
An initial module is generated using design prompts. 
Afterwards, this module is corrected based on compilation errors and simulation messages.
Besides circuit designing, many studies~\cite{liu2024craftrtl, collini2024c2hlsc, xiao2024prefixllm, tsai2024automatically, bai2023towards, sohrabizadeh2022autodse,abi2024hlsfactory, xu2024automated, fang2024assertllm,xu2024meic}
have attempted to leverage LLMs for assistant code generation in hardware development. 
For example, MEIC~\cite{xu2024meic} employs a dual LLMs mechanism for automatic Verilog code development, while RTLFixer~\cite{tsai2024automatically} leverages Retrieval-Augmented Generation to address syntax errors in HDL.
LLMs can be used for HDL debugging and automatic testing~\cite{ma2024verilogreader, zhang2023llm4dv, qiu2024autobench, orenes2023using}.
AutoSVA~\cite{orenes2023using} introduces an iterative framework that leverages formal verification to generate assertions from given hardware modules. 
VerilogReader~\cite{ma2024verilogreader} leverages LLMs to read Verilog code and coverage, aiming at generating code coverage convergence tests.

\subsubsection{Functional Verification}
LLMs are capable of assisting with functional verification by generating testbenches, assertions, and potentially aiding in formal verification. 
For example, AutoBench~\cite{qiu2024autobench} represents a significant advancement as the first LLM-based testbench generator for digital circuit design that requires only the description of the design under test (DUT) to automatically generate comprehensive testbenches.
Their ability to understand design specifications could enable the automatic creation of test scenarios to ensure the circuit behaves as intended under various conditions, potentially addressing the difficulty in verifying complex designs and reducing the time required for this critical stage.
Besides, assertion-based verification is critical for ensuring that design circuits comply with the architectural engineers.
AssertLLM~\cite{fang2024assertllm} addresses this challenge by automatically processing complete specification files and generating functional verification assertions. 
It breaks down the complex task into three phases with customised LLMs-extracting structural specifications, mapping signal definitions, and generating assertions. 
Moreover, test stimuli generation is another crucial component of hardware verification, and has been transformed through frameworks like LLM4DV~\cite{zhang2023llm4dv} that harness the power of LLMs to create efficient test conditions. 
Specifically, LLM4DV introduces a prompt template for interactively eliciting test stimuli from LLMs along with innovative prompting improvements to enhance performance. 
When compared to traditional constrained-random testing (CRT), LLM4DV excels in efficiently handling straightforward DUT scenarios by leveraging basic mathematical reasoning and pre-trained knowledge. 
Although effectiveness decreases with more complex task settings, it still outperforms CRT in relative terms.

\subsubsection{High-Level Synthesis}
Furthermore, LLMs are being explored for their role in guiding High-Level Synthesis (HLS). 
For instance, HLSPilot~\cite{xiong2024hlspilot}, the first LLM-enabled high-level synthesis framework, can fully automate high-level application acceleration on hybrid CPU-FPGA architectures.
The framework applies a set of C-to-HLS optimisation strategies catering to various code patterns through in-context learning, which provides the LLMs with exemplary C/C++ to HLS prompts. 
Table~\ref{tab: csee evidence} lists the performance of the LLMs-based and HLS-based methods in real-world application benchmarks.
The results demonstrate that HLSPilot achieves comparable performance to manually crafted counterparts and can even outperform them in some cases.
By understanding high-level language descriptions or natural language specifications of hardware algorithms, LLMs could help in translating these into HLS-compatible code or even directly guide the HLS tools to produce efficient hardware implementations. This could potentially make hardware design more accessible to engineers with software backgrounds and contribute to shorter design cycles.
This capability addresses the challenge of the substantial semantic gap between natural language expressions and hardware design intent~\cite{liao2024llms}. 
Specifically, LLMs provide a more intuitive design process where engineers can specify hardware functionality in natural language, making hardware design more accessible to those without extensive HDL expertise.

\input{tables/csee_evidence}

\input{tables/csee.tex}

\subsubsection{Benchmarks}
\input{tables/csee_bench.tex}
To evaluate the capabilities of LLMs in tasks related to coding and circuit design, a variety of benchmark datasets have been introduced. 
Table~\ref{tab: csee benchmark dataset} enumerates the key benchmark datasets within the CSEE domain. 
Subsequently, we will delve into a more comprehensive discussion of the commonly utilized benchmark datasets, beginning with these pertaining to coding tasks.

\textbf{HumanEval.} 
It is a benchmark dataset developed by OpenAI specifically designed to evaluate an LLM's code generation capabilities. 
HumanEval consists of 164 hand-crafted programming problems comparable to software interview questions, focusing on functional correctness rather than textual similarity to the reference solution.
Each problem includes a function signature, docstring, code body, and several unit tests, and the performance is measured via PASS@K~\cite{chen2021evaluating}.
HumanEval has been recognized as a standard evaluation for many code-generation models.

\textbf{LiveCodeBench.} 
LiveCodeBench encompasses a comprehensive array of code-related functionalities, extending beyond mere code generation to include self-repair, code execution, and test output prediction. 
Notably, it places a significant emphasis on self-repair, wherein models, upon generating initial code, receive error feedback and are required to amend their solutions accordingly. 
This aspect aligns with the code debugging category within our taxonomy. 
Furthermore, LiveCodeBench necessitates that models predict the outcomes of code execution for specified inputs, which pertains to the code analysis category in our taxonomy. 
A pivotal innovation of LiveCodeBench lies in its methodology for mitigating data contamination, a critical concern in the evaluation of LLMs.

\textbf{APPS.} Automated Programming Progress Standard~\cite{hendrycksapps2021} attempts to mirror how human programmers are evaluated by posting coding problems in natural language and testing solution correctness.
The APPS dataset comprises 100,000 coding problems sourced from platforms such as Codeforces and Kattis, spanning a range from beginner to collegiate competition levels. 
Within this collection, there are 232,444 solutions crafted by human programmers. The average length of each problem is 203.2 words, reflecting the inherent complexity of the tasks.

\textbf{CodeXGLUE.} 
It stands for General Language Understanding Evaluation benchmark for CODE. 
CodeXGLUE offers an extensive array of code intelligence tasks and serves as a platform for the evaluation and comparison of models. 
It encompasses 14 datasets across 10 diverse programming-language tasks, including code-code (e.g., clone detection, defect detection), text-code (e.g., code search, text-to-code generation), code-text (e.g., summarization), and text-text (e.g., documentation translation).

\textbf{CodeContests.} 
It is a competitive programming dataset used for training AlphaCode~\cite{doi:10.1126/science.abq1158}, DeepMind's code generation system.
It contains programming problems from sources like Aizu, AtCoder, CodeChef, Codeforces, and HackerEarth. 
Unlike other benchmarks that merely provide the gold solutions, it also contains incorrect human solutions in various languages.

\textbf{Additional Benchmarks for Software Engineering.}
MBPP consists of around 1K crowd-sourced Python programming problems designed for entry-level programmers.
It focuses on programming fundamentals and standard library functionality.
Each problem includes a task description, code solution, and three automated test cases.
LeetCodeDataset is a high-quality benchmark for evaluating and training code-generation models, particularly focused on reasoning abilities. 
LeetCodeDataset is sourced from LeetCode Python problems with rich metadata and contains over a hundred test cases per problem.
Moreover, it enables contamination-free evaluation and efficient supervised fine-tuning.
WikiSQL is a corpus of hand-annotated SQL queries and natural language question pairs for database-related tasks. 
WikiSQL comprises 87,726 SQL queries and natural-language question pairs, split into training (61,297), development (9,145), and test (17,284) sets.

\textbf{RTLLM.} A benchmark for design RTL generation with LLMs. 
The latest version, i.e., RTLLM 2.0, expands on the original RTLLM benchmark, augmenting it to 50 hand-crafted designs across various applications: (i) Arithmetic Modules; (ii) Memory Modules; (iii) Control Modules and (iv) Miscellaneous Modules.

\textbf{VerilogEval}, A benchmark designed for evaluating LLMs in Verilog code generation for hardware design and verification. 
VerilogEval consists of 156 diverse problems from the Verilog instructional website HDLBits, ranging from simple combinational circuits to complex finite-state machines.
Its evaluation framework uses simulation to verify the functional correctness of generated code by comparing outputs with golden solutions.
The problem statements are provided in two types: machine-generated and human-written.

We further report several available performance reports in Table~\ref{tab: csee performance}.
According to the table, we find that 
(i) Claude 3.7 Sonnet consistently delivers the strongest results wherever it is reported, topping three of the six benchmarks—CodeMMLU (67 \%), Aider-Polyglot (64.9 \%), and SWE-Bench (62.3 \%).
(ii) Claude 3.5 Sonnet remains competitive: although it trails its newer sibling on the knowledge-heavy benchmarks, it records the overall best HumanEval score (93.7 \%) and the second-best EvalPlus (81.7 \%).
(iii) GPT-o1 places second on HumanEval (92.4 \%) and first on EvalPlus (89 \%), but its lead narrows or disappears on the more engineering-oriented datasets (CodeMMLU and SWE-Bench)
(iv) Llama 3.3 matches proprietary giants on HumanEval (88.4 \%) but lags on CodeMMLU (43.9 \%), highlighting the gap that still exists for open models on reasoning-heavy evaluation.

In summary, in general, coding tasks, Claude 3.5 Sonnet, and GPT-o1 give the best pass rates, with GPT-4o, Qwen 2.5, and Llama-3 70 B as cost-efficient runners-up.
On analysis or multilingual engineering tasks, Claude 3.7 Sonnet is the clear leader, while Gemini 2.0 Flash and DeepSeek R1 provide solid mid-tier options.
For large-scale debugging- and circuit design-related tasks, Claude 3.7 Sonnet again excels, but GPT-4o delivers the top score on PICBench and remains the most balanced all-rounder.

\input{tables/csee_benchmark.tex}

\subsubsection{Discussion}
\textbf{Opportunities and Impact.}
The integration of LLMs into both software programming and digital circuit design presents significant opportunities to improve production, streamlining workflow, and fostering innovation~\cite{jin2024llms, kochar2024ledro}.
In software programming, LLMs can automate repetitive coding tasks, leading to substantial time saving and allowing developers to concentrate on more complex jobs~\cite{ma2023ai}.
Moreover, LLMs are capable of assisting developers in debugging in static and dynamic scenarios~\cite{zhong2024debug, levin2024chatdbg, lee2024unified, majdoub2024debugging}.
Furthermore, LLMs can improve code comprehension by generating human-readable explanations of complex codebases, facilitating maintenance and collaboration~\cite{bhattacharya2023exploring, arakelyan2023exploring, li2022generation}.
The impact of these capabilities is a potentially accelerated software development lifecycle, improved code quality, and increased developer satisfaction.

Similarly, in digital circuit design, LLMs offer the potential to automate the generation of HDL code from natural language specifications, which can help manage the increasing complexity of modern integrated circuits~\cite{liu2024craftrtl, collini2024c2hlsc, xiao2024prefixllm, tsai2024automatically, bai2023towards, sohrabizadeh2022autodse,abi2024hlsfactory, xu2024automated, fang2024assertllm,xu2024meic}. 
They can also assist in functional verification by generating testbenches and assertions, potentially reducing the verification bottleneck~\cite{zhang2023llm4dv,fang2024assertllm, qiu2024autobench}. Moreover, LLMs are being explored for their role in high-level synthesis, potentially making hardware design more accessible to software engineers and shortening design cycles. The impact of LLMs in this domain could be a faster time-to-market for hardware innovations and a more efficient design process.

\textbf{Challenges and Limitations.}
Despite the promising opportunities, the application of LLMs in software programming and digital circuit design also entails challenges and limitations~\cite{jiang2024survey}. 
In software, while LLMs can generate code, ensuring the correctness and reliability of complex logic remains a challenge, often requiring significant human oversight.
LLMs may also struggle with maintaining state and reasoning about long-term dependencies in software systems. The issue of "hallucinations," where LLMs generate incorrect or nonsensical code, is a significant concern that needs to be addressed to ensure the trustworthiness of LLM-generated software.

In digital circuit design, the limitations of current LLMs are even more pronounced. Fundamental hardware architecture design, which requires a deep understanding of physics and microarchitectural innovations, is beyond the current capabilities of LLMs. 
Optimizing the physical layout for complex ICs demands specialized knowledge that LLMs do not yet possess. Verifying highly novel or safety-critical hardware designs often requires rigorous formal verification techniques that may exceed the reasoning abilities of current LLMs. Furthermore, the application of LLMs in analog circuit design is still in its infancy and presents unique challenges. 
Limited data sources in the hardware domain also pose a significant challenge for training effective LLMs for circuit design tasks.

\textbf{Conclusion.}
Traditional methodologies have long served as the foundation for software programming and digital circuit design.
The emergence of LLMs has opened a new pathway to address several limitations in these topics.
In software development, LLMs can automate repetitive coding tasks, assist developers with debugging, help manage the complexity of large codebases through explanations and refactoring suggestions, and mitigate human error through intelligent code generation and completion. In digital circuit design, LLMs offer promising solutions for generating HDL code, assisting with functional verification, and guiding high-level synthesis, potentially streamlining the design process and making it more accessible.
Further research and development are crucial to improve the accuracy and reliability of LLM-generated code and verification components, particularly in critical applications. Seamlessly integrating LLMs into existing design workflows and EDA tools in hardware engineering, exploring their potential in more advanced design and optimization tasks, and addressing challenges like data scarcity in hardware design are essential areas for future investigation. As LLMs continue to evolve, they are poised to become valuable partners for engineers in both software and hardware, augmenting human capabilities and shaping the future of technological innovation.

\newpage
\section{Conclusion: Navigating the Present, Shaping the Future}
\label{sec:conclusion}

In this chapter, we explore where LLMs stand today and where they are heading. We begin by outlining new frontiers in LLMs. We then synthesize lessons across three domains. For arts, letters, and law, we distill shared opportunities and common limitations and formalize use paradigms that span historical analysis to legal reasoning. For economics and business, we translate signals from finance, accounting, economics, and marketing into strategy, again pairing opportunities and constraints with concrete paradigms of use. For science and engineering, we treat models as instruments: after brief probes into individual disciplines, we surface cross-cutting opportunities and limitations and propose paradigms suited to experimental and computational workflows. Finally, we pilot the present and plot the future by assessing the current state and articulating a future path that integrates schema-aligned multimodality; grounded, verifiable attribution; tool-augmented computation with formal constraints; rule-governed, reproducible agent-based simulation; temporal and causal adaptation; decision support with calibrated uncertainty and domain controls; human-in-the-loop oversight and transparent governance; and education-led capacity building with embedded safety—collectively yielding a practical, auditable, and scalable blueprint for cross-disciplinary adoption.

\subsection{New Frontiers in LLMs}
\label{LLMs+}

LLMs have evolved from research curiosities into ubiquitous tools driving a new wave of productivity across many sectors. Their impact on the economy is projected to be enormous – one analysis estimates generative AI could add roughly \$2.6–4.4 trillion USD in value annually to the global economy in the long run \cite{chui2023economic_potential}. As of 2025, LLM research is advancing at breakneck speed, as new models and techniques are rapidly extending LLM capabilities, making them more efficient, specialized, and powerful. Equally important, researchers and policymakers are focusing on how to deploy these models responsibly at scale. Drawing on the lifecycle of LLM development and aligning with both academic research trends and industry practices, we identify five major frontiers in LLM research—advances in model architecture, innovations in training paradigm, knowledge and tool integration, domain specialization and customization, and responsibility and ethical considerations—capturing advances from core design through deployment and governance.

\noindent\textbf{Advances in Model Architecture.}
A central frontier in LLM research lies in architectural innovation. Earlier progress often depended on scaling Transformers with ever larger parameter counts, but this strategy faces diminishing returns in both cost and energy. The new wave of models is therefore characterized by designs that embed efficiency, multimodality, and reasoning into the core structure. These advances mark a structural evolution in how LLMs are built, focusing less on brute-force scale and more on intelligent specialization.

One breakthrough is the Mixture-of-Experts (MoE) paradigm~\cite{baldacchino2016variational}, which activates only a subset of parameters for each input rather than the entire network. This approach drastically reduces computation while preserving accuracy, enabling models to expand total parameter counts without proportional increases in inference cost. DeepSeek-R1 \cite{guo2025deepseek} and DeepSeek-V2 illustrate this approach by routing queries across specialized expert subnetworks, achieving state-of-the-art reasoning performance at training costs orders of magnitude lower than dense predecessors \cite{mann2025deepseek,hanbury2025deepseek}. Google’s Gemini 2.0 applies similar principles, leveraging expert routing for efficiency at scale~\cite{google_gemini}. MoE design represents more than a technical optimization: it reflects a new paradigm that scale and efficiency can coexist, expanding access to capable models beyond many technology giants.

In parallel, multimodal architectures are extending LLMs beyond text into vision, speech, and audio, reflecting how humans integrate multiple channels of perception. GPT-4o, Claude 3.5 Sonnet, and Gemini 2.0 exemplify this trend by enabling unified input-output pipelines across modalities. This unlocks new application domains: a medical assistant that analyzes both patient notes and radiology scans, or a workplace AI that interprets screenshots and executes contextual actions \cite{masri2025traffic}. Cross-lingual capabilities deepen the promise of multimodality. Meta’s xLLaMA-100, for instance, scales instruction tuning to 100 languages, significantly improving coverage for low-resource contexts \cite{lai2024xllama100}. By integrating modalities and languages, these architectures begin to function as "world models", systems that approximate the richness of human perception and communication rather than remaining confined to text.

Reasoning-oriented architectures further expand capability by embedding structured cognition into the model workflow. Whereas early prompting tricks like Chain-of-Thought (CoT) showed that LLMs could benefit from step-by-step reasoning, newer designs such as Tree-of-Thought (ToT) and graph-based reasoning allow models to branch into alternatives and refine their logic. OpenAI’s o1 is a flagship example: deployed with reasoning as a default mode, it consistently outperforms earlier systems on mathematics, logic, and scientific benchmarks by producing and verifying intermediate steps \cite{theverge_o1_free_2025}. DeepSeek-R1 also demonstrates how reinforcement learning with verifiable rewards (RLVR) can amplify reasoning without brute scale \cite{mann2025deepseek}. Together, these approaches shift the paradigm from fluency to cognition: models are no longer judged solely by their ability to generate coherent text, but by their capacity to reason in structured, inspectable ways. Further, neuro-symbolic AI \cite{garcez2023neurosymbolic}—by combining the intuition of neural models with the rigor of symbolic reasoning—has been recognized as the emerging paradigm that could lead the next significant shift in AI.

Collectively, MoE efficiency, multimodal extensions, and reasoning-centric workflows define the architectural frontier. The next generation of models will not be remembered for parameter counts alone, but for embedding efficiency, perception, and reasoning as foundational design principles. This reorientation promises not only stronger performance but also broader adoption, as models become more sustainable, context-aware, and capable of structured thought.

\noindent\textbf{Innovations in Training Paradigms. } As we approach the limits of simply "scaling up" Transformers with more data and GPUs, beside model architectures, researchers are exploring new training paradigms to achieve the next leap in capabilities. This frontier is critical: improved training methods can make LLMs more capable, more reliable, more efficient, and more adaptable without just making them bigger.

One promising direction is integrations of alignment approaches for model training. The alignment approaches could encompass Reinforcement Learning from Human Feedback (RLHF) \cite{ouyang2022training}, Reinforcement Learning from AI Feedback (RLAIF) \cite{lee2023rlaif}, and multi-stage alignment strategies \cite{li2022mvptr}, designed to adapt model behavior to human values, societal norms, and safety requirements, while further instilling enhanced reasoning and tool-use capabilities. DeepSeek-R1, for example, was largely trained via reinforcement learning with verifiable rewards (RLVR), enabling it to solve math and logic tasks by checking answers against automated validators \cite{deepseek_r1_2025}. Similarly, recent work extends RLVR and related strategies to broader reasoning domains \cite{su_crossing_2025}. These approaches illustrate how training signals derived from verifiable outcomes can replace the need for massive labeled datasets, making training more scalable and reliable.

In parallel, researchers are investigating radically new architectures inspired by neuroscience. The SpikingBrain project proposes a brain-inspired model that combines spiking neurons with efficient attention mechanisms, yielding strong performance on ultra-long contexts while reducing inference memory costs \cite{spikingbrain_2025}. Such designs challenge the assumption that scaling Transformers is the only path forward, suggesting that new algorithmic principles—closer to how the human brain processes information—could define the next era of LLMs. Complementary work reveals that LLMs contain "functional networks", modular subsets of neurons disproportionately responsible for specific functions. Masking these subnetworks drastically degrades performance, while preserving only about 10\% of them can still retain acceptable accuracy \cite{liu2025pruning}. This hints at a modularity and redundancy reminiscent of biological systems, opening opportunities for leaner and more targeted training. In summary, researchers are increasingly drawing inspiration from the human brain, examining principles such as modularity \cite{alkhamissi2025llmnetwork}, continual learning \cite{wu2024continual}, and memory augmentation \cite{memorag2025} that support knowledge updates and long-context processing (e.g., long-context GPT-4o and Claude with extended document support), as well as latent dynamics \cite{pan2025spikingbrain,gu2024mamba}. These lines of inquiry aim to inform the development of architectures that go beyond Transformers, pointing toward new directions for future LLMs.

Additionally, Parameter-Efficient Fine-Tuning (PEFT)—a family of methods that enable large models to be adapted to downstream tasks without retraining all parameters—has also drawn significant attention. Approaches such as Low-Rank Adaptation (LoRA) \cite{hu2022lora} significantly reduce computational and storage requirements, making fine-tuning feasible even on resource-constrained hardware. By introducing only a small number of additional trainable parameters or compressing weight representations, PEFT allows models to retain the general capabilities learned during pretraining while rapidly adapting to domain-specific tasks. This strategy not only lowers the barrier to customization, as exemplified by lightweight models such as Qwen-1.5B and Phi-3, but also facilitates flexible deployment across diverse applications.

\noindent\textbf{Knowledge and Tool Integration.} 
Equally transformative is the frontier of knowledge and tool integration. Unlike purely parametric models, which are limited by the static data they were trained on, integrated systems dynamically connect to external knowledge repositories, computational engines, and software environments. This integration allows models not only to generate language but to ground their answers in facts, reason through external computation, and act through APIs. It addresses some of the most pressing limitations of early LLMs: hallucination, staleness, and lack of action.

Retrieval-Augmented Generation (RAG) is the clearest example of this shift. By combining model outputs with results from external search engines or databases, RAG reduces hallucination and ensures outputs are both current and verifiable. Studies show that RAG-based assistants outperform base models in knowledge-intensive domains such as law, medicine, and research, where citing evidence is critical \cite{gao2023retrieval}. This architecture allows enterprises to connect proprietary corpora—legal filings, medical guidelines, technical manuals—directly to LLMs, turning them into domain-aware assistants. By anchoring generation in retrieval, RAG ensures that models evolve from static predictors to knowledge-grounded systems supporting high-stakes decisions.

Tool use and AI agents represent a second strand. By linking LLMs with APIs, execution environments, and user interfaces, researchers have created systems that do not just converse but act. Early prototypes like AutoGPT illustrated this principle by chaining prompts and tool calls to pursue long-term goals autonomously~\cite{yang2023auto}. More specialized deployments now show its practical utility: Salesforce’s Large Action Models (xLAM) integrate LLMs with enterprise software, enabling tasks such as updating customer records, orchestrating workflows, and triggering procurement actions \cite{salesforce_xlam_v2_2025}. Devin, billed as an “AI software engineer,” exemplifies the frontier by autonomously handling software development lifecycles—reading tickets, editing code, running tests, and submitting patches against live repositories \cite{cognition2024devin}. These systems highlight how LLMs are evolving from assistants to operators, compressing cycle times across industries by completing tasks end-to-end.

A third and complementary development is program-aided reasoning (PAR). Rather than relying solely on natural language reasoning, PAR pipelines let models produce symbolic representations—such as code or equations—which are then executed by external engines. This hybrid approach, exemplified by PAL (Program-Aided Language Models), substantially improves accuracy on mathematical and scientific benchmarks~\cite{gao2023pal}. By delegating computation to deterministic systems, PAR reduces hallucinations and opens new horizons in domains requiring precision, from quantitative finance to physics simulations. In doing so, it demonstrates how symbolic and neural paradigms can coexist, combining the generative flexibility of LLMs with the rigor of formal computation.

In summary, RAG, tool use, and program-aided reasoning redefine what an LLM is: no longer a static predictor but a dynamic collaborator. By grounding outputs in retrieval, extending reach through APIs, and leveraging symbolic reasoning for precision, integrated models promise to be more trustworthy, more capable, and more aligned with real-world needs. This frontier turns language models into knowledge workers and problem solvers, capable not just of generating text but of acting intelligently within complex environments.

\noindent\textbf{Domain Specialization and Customization. } As LLMs mature, it has become increasingly clear that general-purpose models cannot optimally serve the full breadth of real-world applications. A major frontier is therefore the specialization of LLMs for specific domains, in which models are fine-tuned or adapted to industry datasets, proprietary corpora, or disciplinary knowledge. As discussed extensively in earlier chapters, the logic is straightforward: a finance assistant must interpret regulatory language and market jargon, while a biomedical model must parse clinical notes and research articles. Tailoring models to these contexts improves accuracy and ensures closer alignment with the terminology, constraints, and reasoning styles of each field. The benefits extend beyond raw performance: customization also enhances trust and usability, as stakeholders can engage with an AI that is contextually fluent and institutionally compliant rather than one that only approximates domain knowledge.

The impact of this specialization is already visible across diverse disciplines. As we have discussed previously, while considerable progress has been made in domains such as biomedicine and chemistry, there remains ample room for expansion. At the same time, the next frontier lies in less frequently highlighted fields such as art, history, and philosophy, where customized LLMs are beginning to function as cultural interpreters, educational guides, and tools for archival research. Crucially, the emergence of open-source foundation models has lowered barriers to entry, enabling organizations to fine-tune systems like LLaMA or Mistral without prohibitive computational cost. This democratization ensures that domain-specific LLMs are no longer confined to technology giants but can proliferate across laboratories, enterprises, and cultural institutions alike.

In sum, the rise of domain-specialized and customized LLMs represents more than a natural extension of scale—it marks a structural evolution in the development of language technologies. By embedding expertise directly into the models, researchers and practitioners are producing systems that are not only more accurate and trustworthy but also more impactful in solving domain-specific problems.


\noindent\textbf{Trustworthiness and Governance.} As LLMs enter high-stakes applications, their misuse can directly harm individuals and society. Without safeguards, models may reinforce social biases, leak private data, or be exploited to generate disinformation at scale. In employment or finance, unfair outputs could entrench discrimination; in healthcare, errors could compromise patient safety. Regulators and researchers emphasize that fairness, privacy, and security must be built into model design rather than treated as afterthoughts \cite{meding_2025_fairness,edpb_privacy_2025}. Emerging frameworks such as the EU AI Act, NIST’s risk management updates, and the Open Worldwide Application Security Project (OWASP) Top 10 for LLMs call for proactive governance, bias auditing, privacy-by-design, and stronger protections against adversarial misuse \cite{eu_ai_act_compliance_2025,owasp_llm_top10_2025}.  

Looking forward, the legitimacy of LLMs will hinge on embedding responsible practices at the core of their development. Models must be transparent in how they reason, accountable for errors, and resilient to manipulation. Equally, international coordination will be necessary to avoid fragmented protections across jurisdictions. The future frontier of responsible AI is thus less about technical benchmarks and more about trust: ensuring that increasingly capable LLMs operate in ways that uphold human values, protect users, and strengthen rather than destabilize social institutions.  

\subsection{Beyond the Brief: LLMs for Arts, Letters, and Law}
\label{sec:cluster-1}

In this section, we examine the expanding role of LLMs across the humanistic and social disciplines, including history, philosophy, political science, the arts and architecture, and law. We first outline shared opportunities, showing how LLMs can support interpretation, analysis, and creative production across these fields, from assisting historical research and philosophical inquiry to augmenting artistic design and legal drafting. It then turns to common limitations, addressing persistent concerns such as bias, casual reasoning, contextual grounding, interpretability, and originality. Synthesizing insights across these disciplines, we finally identify five paradigms—role play, understanding, generation, education, and human-in-the-loop collaboration—that structure how LLMs mediate knowledge work in these domains, highlighting both their transformative potential and epistemic limitations, and pointing toward directions for future research.

\subsubsection{Shared Opportunities}

Arts, letters and law encompass a wide range of human expression, thought and governance. These disciplines highly rely on the processing of massive documents, including narratives, disclosures, artworks and regulations to conduct interdisciplinary information retrieval, document analysis, human behavior simulation and content generation. The strong document processing ability of LLMs is potentially changing research paradigms and extending the boundaries of these disciplines. 

\textbf{Automation of Labor-Intensive Document Analysis.} LLMs are increasingly capable of analyzing long, interdisciplinary and multi-modal documents, which are very common in the social science sector. Such ability of LLMs enables the automation of traditionally labor-intensive interpretive tasks in social science, like literature transcription~\cite{blevins2024}, multi-modal artwork interpretation~\cite{bin2024gallerygpt}, 
entity recognition~\cite{hauser2024large} and text classification~\cite{ornstein2025train, le2025positioning, o2023measurement}. This progress democratizes the access to social science research via information retrieval systems with natural language interfaces~\cite{zheng2024disciplink,nay2023llm,savelka2023explaining}, which could have a broader impact in the future. 

\textbf{Reasoning-based Document Understanding.} LLMs possess the ability to perform complex logical reasoning across diverse contexts, while reasoning is unavoidable in social science. Researchers not only need to understand massive documents, but also need to make comparisons among available documents based on various logic to deduce convincing conclusions. In history, LLMs are utilized to perform analogical reasoning~\cite{li2024past} to augment the productivity of historians. In philpsophy, normative reasoning~\cite{dillion2025moralexpertise, kempt2024conversationalethics} and analytical logic~\cite{myung2025epistemology, heersmink2024phenomenology} could be performed by LLMs to assist the interpretation of dense arguments. Legal reasoning has also been assisted by LLMs in the discipline of law~\cite{cao2024pilot, wu2023precedent}. Based on different logical conditions, the potential of LLMs in logical analysis could be further exploited. 

\textbf{Social Behavior Simulation.} LLMs can learn from complex human interactions and model social behaviors of humans, which potentially enables novel methodologies for social science research. Recent studies demonstrate their capacity to replicate human behavior from different eras and with different stances~\cite{varnum2024large,qu2024performance,karanjai2025synthetic}, which facilitates social behavior prediction from quantitative perspective. For example, LLMs can be used to design campaign messages based on the behavior prediction of voters with different backgrounds~\cite{hackenburg2024evidence,matz2024potential}. This lays foundation for large-scale behavior modeling and agent-based modeling. 

\textbf{Decision-Making under Complex Social Scenarios.} The ability of LLMs in decision-making in situations with complex contextual factors has been demonstrated. The decision-making process in real-world society is never easy due to the involvement of various stakeholders, which LLMs could potentially assist. For example, in judgment prediction, LLMs are designed to capture the nuances of precedent, legal reasoning,
and jurisdictional variation to make convincing decisions~\cite{cao2024pilot,wu2023precedent}. The application of LLMs in more complex and realistic scenarios could be further explored and improved.

\textbf{Conceptual Ideation.} One of the most transformative features of LLMs is their ability to generate meaningful content based on user requirements. Initial attempts have been made in art creation~\cite{mirowski2023co} and legal document drafting. All the progress reveals more possibilities for LLMs to further assist social science research via data synthesis, AI-aided experimental design and hypothesis testing.

\subsubsection{Common Limitations}

\noindent Despite their transformative promise, the application of LLMs across history, philosophy, political science, the arts, and law reveals a series of recurring limitations that cut across these domains. These challenges are not isolated to one field but emerge wherever models are asked to move beyond surface-level fluency into tasks that demand rigorous reasoning, contextual sensitivity, cultural awareness, or normative accountability.

\textbf{Lacking Logical Depth and Causal Reasoning.} Although LLMs excel at producing grammatically correct and stylistically fluent text, their argumentative structures often collapse under closer scrutiny. In philosophy, models are able to mimic argumentative form but frequently fail to maintain internal consistency or sustain normative frameworks, resulting in outputs that are rhetorically plausible yet conceptually shallow~\cite{shanahan2023evaluating}. In political science, similar issues arise when models are tasked with simulating voter behavior or forecasting ideological dynamics: they can generate distributions of preferences or hypothetical debates, but these outputs often lack causal grounding or institutional nuance, limiting their value for theory-building or hypothesis testing~\cite{Guo2024EconNLIEL}. This structural weakness points to a broader problem: current architectures rely on pattern recognition rather than genuine logical inference, making it difficult to ensure reasoning fidelity in domains where causal rigor is indispensable.

\textbf{Struggling with Contextual Grounding.} LLMs may consistently struggle with contextual grounding, especially in disciplines where meaning is tightly bound to historical, cultural, or institutional settings. In history, for example, models can produce compelling narratives but often miss critical temporal markers or distort the socio-political conditions that shape historical interpretation, yielding decontextualized or anachronistic accounts~\cite{denzin2003performance}. In the arts, LLMs may describe works of visual or literary art while neglecting embedded symbolism or cultural resonance, effectively flattening creative production into dehistoricized text~\cite{bin2024gallerygpt}. In architecture and law, the problem becomes particularly acute: design traditions or legal precedents are deeply tied to institutional and jurisdictional contexts, which LLMs may fail to respect, leading to outputs that are formally accurate but substantively misaligned~\cite{Zhang2024ArchGPTHL, shu2024lawllm}.

\textbf{Boundaries of Creativity and Authorship.} In artistic disciplines, LLMs demonstrate impressive capabilities in stylistic imitation—replicating genres, voices, or compositional strategies—but they rarely achieve authentic novelty or intentional innovation~\cite{yuan2023artgpt}. This limitation raises difficult questions about authorship and authenticity: if a model generates a painting description, a script, or a poem that mirrors human creativity, who owns the output, and does it carry cultural or aesthetic value independent of human intent? In domains such as theater and performance, where LLMs have been used as collaborative partners, scholars emphasize that while these systems can assist in expanding creative options, they lack the embodied intentionality, emotional grounding, and lived experience that characterize human expression~\cite{mirowski2023co, Branch2024DesigningAE}. Without frameworks for attributing originality and recognizing the limits of machine-generated creativity, these outputs risk being derivative, undermining rather than expanding cultural production.

\textbf{Opacity and Trust in High-Stakes Domains.} Another limitation is the opacity and lack of interpretability of LLM outputs, which undermines trust in high-stakes fields such as law, political governance, and even historical analysis. In contrast to traditional scholarly or legal reasoning—where arguments are accompanied by citations, precedents, and explicit rationales—LLM-generated responses often present conclusions without traceable justification~\cite{shu2024lawllm}. This black-box quality is particularly problematic in regulated environments, where accountability requires the ability to audit reasoning processes and verify sources. For instance, in legal judgment prediction, models may propose plausible outcomes but cannot provide the structured doctrinal reasoning that judges or lawyers require to ensure fairness and legitimacy~\cite{nigam2024rethinking}. Similar issues arise in political applications, where the absence of transparent sourcing makes it difficult to separate genuine insights from spurious correlations.

\textbf{Bias Propagation and Ethical Misuse.} Finally, across all of these disciplines, LLMs face persistent risks of bias propagation and ethical misuse. Because they inherit patterns from training corpora, they tend to replicate and amplify demographic, cultural, or ideological biases embedded in their data. In political science, this manifests as the reinforcement of dominant narratives and marginalization of dissenting voices, raising concerns about fairness and representativeness~\cite{bang2024measuring}. In law, models trained on judicial records may reproduce historical inequities or discriminatory practices, thereby embedding bias into automated decision support~\cite{williams2025disinfo}. In the arts, generative systems may privilege dominant cultural styles while neglecting minority traditions, leading to homogenization of creative expression. Beyond these structural biases, LLMs can also be deliberately misused—for example, to generate disinformation campaigns, manipulate public opinion, or fabricate legal or historical documents—posing risks that extend beyond academia into broader civic and cultural life.

Taken together, these shared limitations underscore the enduring gap between the surface-level fluency of LLMs and the deeper epistemic, cultural, and ethical requirements of humanistic and social inquiry. They highlight the need for systems that can move beyond pattern replication toward genuine reasoning, contextual sensitivity, and transparency, while also demanding robust ethical frameworks to safeguard against misuse. Addressing these cross-cutting challenges will be crucial if LLMs are to serve as responsible collaborators in the advancement of knowledge across the humanities, social sciences, and law.

\subsubsection{LLM Paradigms from History to Law}

Humanistic and social science research follows a distinctive lifecycle: interpreting cultural and historical materials \citep{maccormick2016interpreting}, reasoning across disciplinary boundaries \citep{nicolini2012understanding}, integrating multi-source and multi-modal evidence \citep{yuan2023artgpt}, training practitioners through education \citep{brock2017findings}, and embedding ethical constraints to ensure legitimacy and responsibility \citep{liu2023trustworthy}. Each stage presents opportunities for LLMs to augment scholarly practice by extending interpretive reach, supporting structured reasoning, and enabling interactive collaboration. By aligning our discussion with this lifecycle, we highlight how LLMs are beginning to transform research workflows, owing to the centrality of meaning, context, and representation in humanistic inquiry. We therefore identify five paradigms of adoption: (1) LLMs as role-players to simulate historical, political, and artistic figures, (2) LLMs as domain experts via knowledge-augmented reasoning, (3) multi-modal learning to unify and reason over heterogeneous sources, (4) education and training assistants that scaffold interpretive skills, and (5) ethically constrained generators that foreground bias, legitimacy, and responsibility. Overall, these paradigms illustrate both the promise and responsibility of integrating LLMs into the humanities and social sciences.  

\noindent\textbf{LLM-driven Role-players to Simulate Historical, Political, and Artistic Figures.}  
The ability to inhabit historical, political, or artistic figures is central to research and pedagogy in the humanities and social sciences. Such role-playing enables scholars to explore alternative perspectives, reconstruct debates, and test hypotheses \citep{zeng2024histolens,dillion2025moralexpertise,argyle2023synthetic} about decision-making, interpretation, and cultural expression. Traditional methods of reenactment or simulation \citep{lane2007use} are constrained by limited sources, participant availability, and the difficulty of sustaining diverse voices. LLMs introduce a novel affordance: they can adopt and maintain specific personae, simulating the perspectives, voices, and behaviors of both real and imagined figures. In history and philosophy, LLMs have been used to reenact Confucian debates~\cite{zeng2024histolens} and conduct moral dialogues~\cite{dillion2025moralexpertise}; in political science, they act as synthetic respondents for opinion polling and preference elicitation~\cite{argyle2023synthetic}; and in literature, education, and performance studies, they enrich narrative design by inhabiting fictional or theatrical roles~\cite{mirowski2023co}. These applications open new avenues of inquiry, providing interactive, adaptable, and scalable methods for examining complex intellectual and cultural terrains. Despite this potential, LLM-based role-play presents persistent challenges \citep{tseng2024two}, such as maintaining persona consistency, addressing knowledge gaps, balancing creativity with coherence, etc. Emerging research highlights three key directions: {simulation fidelity}, ensuring responses align with the epistemic and stylistic attributes of the assumed figure \citep{louie2024roleplay}; {interpretive flexibility}, enabling adaptation across diverse research and pedagogical contexts \citep{shao2023character}; and {responsibility and ethics}, foregrounding transparency, accountability, and safeguards against misuse \citep{liu2023trustworthy}.

\noindent\textbf{LLM-assisted, Knowledge-augmented Expert Reasoning.}  
Expert reasoning in the humanities and social sciences often requires navigating interdisciplinary sources and applying complex modes of logical or normative deduction \citep{sunstein1993analogical}. When acting as domain experts, large language models can support this process by parsing heterogeneous and multi-modal materials \citep{turn0search1}, structuring information into coherent frameworks \citep{dagdelen2024structured}, and drawing connections across disciplinary boundaries \citep{xiang2025survey}. LLMs contribute to this paradigm in several ways. In history, they have been used to identify past events analogous to contemporary situations across multiple dimensions~\cite{li2024past}; in philosophy, they enable normative reasoning~\cite{dillion2025moralexpertise, kempt2024conversationalethics} and integrate diverse sources for comparative analysis~\cite{colombatto2024folk, overgaard2024conscious}; and in law, systems such as PILOT \citep{cao2024pilot} and PLJP \citep{wu2023precedent} demonstrate how LLMs can assist in judgment prediction and structured legal reasoning. These applications extend LLMs beyond summarization, positioning them as exploratory thinkers that generate hypotheses, surface analogies, and provide structured insights to augment human expertise.  

However, challenges remain in this paradigm. Knowledge-augmented reasoning requires consistency in applying domain-specific logic and reliable grounding in authoritative sources. Risks include overgeneralization, shallow analogies, or spurious reasoning when models rely on incomplete or biased data. To address these issues, prior work has experimented with curated knowledge bases \citep{celli2024knowledge}, retrieval-augmented generation \citep{gao2023retrieval}, hybrid symbolic--neural frameworks \citep{wang2024python}, and evaluation protocols tailored to expert reasoning tasks \citep{wu2023precedent}. Future directions include improving transparency in reasoning chains \citep{liu2023trustworthy}, developing benchmarks that capture expert-level interpretive nuance \citep{guha2023legalbench}, and designing interactive workflows where LLMs collaborate with scholars and practitioners as partners in discovery \citep{louie2024roleplay}.

\noindent \textbf{Multi-source Understanding via Multi-modal Learning.} Humanistic and social science research is inherently multi-modal, drawing on sources that extend far beyond text. Historical inquiry relies on archives, visual artifacts, and material culture \citep{what_is_history}; philosophy engages both natural language and symbolic representations such as logic and formal models \citep{philpapers}; political science integrates speeches, survey data, and network graphs \citep{easton1953political}; the arts and architecture combine textual descriptions with visual images and spatial models \citep{yuan2023artgpt,groat2013architectural}; and law requires the synthesis of statutes, precedents, and structured legal formats \citep{corbin1919legal}. These examples illustrate why knowledge in these domains cannot be reduced to text alone, and why the transition from text-only LLMs to systems capable of aligning heterogeneous sources of evidence marks a critical frontier. Looking ahead, multi-modal learning extends beyond representation toward reasoning and co-creation. Such systems could generate richer historical narratives by combining textual records with visual or material evidence \citep{zeng2024histolens,ghaboura2025time}; connect natural language with symbolic inference in philosophy \citep{segler2017neural}; integrate discourse with empirical data in political science \citep{aoki2024large}; foster cross-modal creativity and understanding in the arts and architecture \citep{bin2024gallerygpt,Zhang2024ArchGPTHL,yuan2023artgpt}; and provide transparent reasoning chains in law by linking free-form text with structured legal sources \citep{shu2024lawllm}. However, key challenges remain, including difficulties in aligning meaning across modalities, uneven availability of high-quality data, limited interpretability, and risks of bias or cultural insensitivity. Mitigations explored in prior work include grounding outputs in verifiable sources \citep{wang2023survey}, curating domain-specific corpora \citep{shen2024tag}, developing interpretability tools \citep{singh2024rethinking}, and engaging domain experts in dataset design and evaluation \citep{guha2023legalbench}.

\noindent\textbf{Instructional Assistants Scaffolding Interpretive Skills.} 
Education and training are central to cultivating critical reasoning and interpretive sensitivity across the humanities and social sciences. Traditional methods face challenges of scale, limited access to individualized feedback, and constraints in simulating diverse perspectives or real-world problem contexts. LLMs offer a new paradigm by serving as interactive tutors or collaborators. They have been applied to support moral growth in philosophy~\cite{wang2024moralgrowth}, provide simulated exercises in legal reasoning~\cite{choi2023llmforempirical}, generate creative prompts in literature~\cite{shanahan2023evaluating}, and aid historical literacy by translating or annotating primary sources~\cite{blevins2024}. Yet, the integration of LLMs into education introduces new challenges. These include the risks of oversimplification, epistemic overconfidence, and the potential reinforcement of biases when model outputs are mistaken for authoritative content. Without critical framing, learners may uncritically adopt generated perspectives as normative or overlook their limitations. Existing works have begun to mitigate these risks through strategies such as embedding critical reflection prompts \citep{cai2025impact}, integrating retrieval from curated corpora \citep{chung2024legalrag}, and designing assessment frameworks that foreground transparency and interpretive rigor \citep{fei2023lawbench}.

\noindent\textbf{Ethical Constraints in Text Generation.}  
Ethical constraints are essential in the humanities and social sciences, where interpretation, representation, and authority carry deep cultural and political weight. LLMs risk reproducing and amplifying biases already embedded in training data, thereby marginalizing underrepresented perspectives and reinforcing existing power structures~\cite{xu2025better, colombatto2024folk}. In law and politics, such distortions can misinform, erode legitimacy, or skew debates toward dominant ideologies~\cite{potter2024bias, nigam2024rethinking}. In art and education, they can distort cultural meaning, misattribute authorship, or commodify creative labor~\cite{shanahan2023evaluating, mirowski2023co}. Moreover, LLMs often fail to provide transparent reasoning: hallucinated citations, misaligned analogies, and opaque decision-making complicate trust, accountability, and epistemic responsibility~\cite{savelka2023explaining, li2025system}. These failures underscore why ethical constraint is not peripheral but central to deploying LLMs in socially consequential domains.  

The challenges are multifaceted. Beyond technical accuracy, responsible deployment requires value alignment, cultural sensitivity, and democratic oversight. Existing approaches include bias audits and dataset diversification, interpretability tools for exposing reasoning paths, and governance frameworks that embed stakeholder participation. Some works further explore constrained generation through retrieval-augmented methods \citep{gao2023retrieval} or rule-based overlays, aiming to reduce hallucinations and anchor outputs in verifiable sources. Looking ahead, research must move toward participatory evaluation frameworks \citep{li2024llms}, context-sensitive alignment strategies \citep{dai2023safe}, and interdisciplinary standards \citep{what_is_history,polis1990ecology,groat2013architectural} that embed accountability at every stage of LLM use. Such extensions would ensure that ethical constraints operate not as external checks but as integral design principles, enabling text generation that is both epistemically responsible and socially legitimate.

\subsection{Signals to Strategy: LLMs for Economics and Business}

LLMs are reshaping the landscape of economics and business by introducing language-driven intelligence into fields traditionally anchored in numerical models, symbolic reasoning, and empirical inference. Their ability to parse unstructured data, generate strategic insights, simulate human behavior, and provide natural language interfaces is transforming how decisions are modeled, forecasted, and executed across finance, economics, accounting, and marketing. This section synthesizes these developments by examining the shared opportunities, common limitations, and emerging paradigms that define the role of LLMs in decision-making and knowledge work within these disciplines.

\subsubsection{Shared Opportunities}

Economics and business represent domains where human behavior is modeled, forecasted, and influenced through quantitative abstraction. Whether optimizing capital allocation, designing market mechanisms, or interpreting consumer sentiment, these disciplines rely on structured models and empirical inference to guide decision-making. The emergence of LLMs marks a significant epistemological shift: they introduce language-based reasoning into domains traditionally dominated by numerical and symbolic methods. More fundamentally, LLMs suggest that language itself can become a computationally actionable asset. That is, textual inputs such as statements, disclosures, or consumer narratives can now be parsed, reasoned over, and transformed into structured decisions or economic interventions through language-driven interfaces.

The financial sector exemplifies this shift with a highly optimistic outlook toward LLM adoption. According to the 2024 IIF-EY survey~\cite{iif2024aiml}, \textit{89\%} of financial institutions are using or piloting GenAI/LLMs, and \textit{100\%} increased their AI/ML investments in 2024. McKinsey estimates that technology-driven productivity improvements in banking could raise operating profits by \textit{9--15\%}, equivalent to \textit{\$200--340 billion} annually~\cite{kamalnath2023capturing}. Yet, despite this enthusiasm, widespread deployment remains limited. A 2024 Alan Turing Institute survey found that while most BFSI leaders plan to integrate GenAI within two years, \textit{over 70\%} remain in the proof-of-concept phase~\cite{maple2024impact}. Additionally, \textit{80\%} of financial institutions prioritize internal process optimization over customer-facing applications~\cite{iif2024aiml}, reflecting a cautious and deliberate adoption strategy.

This momentum has also spread to adjacent fields, notably accounting~\cite{kpmg2024aiadoption}. KPMG’s 2024 survey reports that \textit{76\%} of organizations have adopted AI in accounting functions~\cite{kpmg2024aiadoption}. A separate survey of 273 CPA decision-makers shows that \textit{30\%} of business executives are experimenting with GenAI, up from \textit{23\%} the previous year~\cite{strickland2024genai}. Deloitte and the IMA further report that \textit{16\%} of finance and accounting professionals already use or adopt GenAI, with another \textit{44\%} planning adoption within five years~\cite{deloitte_ima_2024_ai_controllership}. Despite growing interest, \textit{34\%} of CPA decision-makers express significant concerns about data privacy, ethics, and AI output accuracy~\cite{strickland2024genai}, highlighting the need for robust governance and risk management.

In marketing, GenAI and LLM adoption has accelerated rapidly. A September 2024 AMA survey found that \textit{nearly 90\%} of marketing professionals have integrated GenAI tools~\cite{cashion2024generative}, primarily for content creation, ideation, and campaign development. Marketers also leverage GenAI to enhance writing quality, customer communications, and social media management. However, full-scale implementation remains rare. A McKinsey survey of 52 Fortune 500 retail executives revealed that while \textit{90\%} have launched GenAI pilot projects, only \textit{two} reported successful, organization-wide deployment~\cite{sukharevsky2024llm}, underscoring ongoing operational and governance challenges.

Taken together, these adoption patterns illustrate both the promise and the cautious trajectory of GenAI integration across business domains. Beyond survey statistics, it is useful to highlight several cross-cutting opportunities that characterize how LLMs are beginning to transform economic and business practices.

\textbf{Bridging Text and Numbers in Decision-Making.}
At the foundation, LLMs create new synergies between unstructured and structured data sources. In finance, this allows for the integration of narrative sources such as earnings calls, regulatory filings, and news sentiment into trading or risk models~\cite{kim2024financial, li2025extracting}. In economics, LLMs enable the simulation of policy discourse or agent behavior using naturalistic language~\cite{Taghikhah2023ACF, Mei2024ATT}. In marketing and accounting, they support extraction of insights from customer feedback and disclosure narratives~\cite{Jain2023ThePA, street2023let}. These capabilities enable a more holistic understanding of economic and business phenomena, where qualitative nuance complements the quantitative structure.

\textbf{Cognitive Automation of Expert Tasks.}
Building on this integration capacity, LLMs are increasingly applied to automate tasks that previously required expert human judgment. Examples include financial document summarization~\cite{han2024xbrl}, fraud risk annotation~\cite{alshurafat2023usefulness}, tax law interpretation~\cite{dong2024scoping}, or marketing campaign ideation~\cite{Aghaei2025HarnessingTP}. Their flexibility across domains allows them to serve as language-enabled research assistants, scaling expert workflows without requiring rigid template-based automation.

\textbf{Simulating Human Economic Behavior.}
Beyond automation, LLMs also enable the simulation of human-like decision processes at scale. Recent studies demonstrate their capacity to replicate behavior in classic experimental economics tasks~\cite{Filippas2023LargeLM}, simulate agents in macroeconomic environments~\cite{Taghikhah2023ACF}, and engage in recursive reasoning akin to game-theoretic interactions~\cite{Shapira2024GLEEAU, Guo2024EconNLIEL}. This opens new avenues for large-scale behavioral simulation, agent-based modeling, and pre-testing of economic experiments with synthetic populations.

\textbf{Natural Language Interfaces for Complex Systems.}
Finally, LLMs enable more intuitive access to economic and financial systems through natural language interfaces. Users can query databases, generate dashboards, or explore regulatory documents conversationally, lowering the barrier to complex analytics and democratizing access to domain-specific insights~\cite{yang2024financial, wu2024susgen}.

\subsubsection{Common Limitations}

LLMs have demonstrated remarkable progress in natural language understanding and generation, yet their integration into domains such as economics and finance reveals several enduring limitations. While they excel at producing fluent, contextually relevant text and synthesizing vast amounts of information, their underlying architecture imposes structural constraints that limit reliability in high-stakes or technically rigorous settings. These challenges are not merely technical inconveniences but touch on issues of precision, consistency, adaptability, and fairness—each of which has direct implications for decision-making, compliance, and trust. The following paragraphs outline some of the most salient limitations that should be considered when deploying LLMs in these domains.

\textbf{Lack of Numerical Grounding and Symbolic Precision.}
Despite their linguistic fluency, LLMs often underperform in tasks that require rigorous arithmetic reasoning or symbolic manipulation. This limitation is particularly acute in finance and accounting, where decimal-level precision is non-negotiable, and in game-theoretic modeling or economic equilibrium analysis, where consistency across assumptions is essential~\cite{drinkall2025forecasting}.

\textbf{Limitations in Logical Consistency and Structured Reasoning.}
Despite their fluency and versatility, LLMs often lack the ability to maintain internal logical consistency or adhere to domain-specific reasoning protocols across extended contexts. Tasks such as financial forecasting, policy analysis, or strategic decision modeling typically require consistent assumptions, conditional reasoning, and multi-step logic capabilities that LLMs frequently approximate only heuristically. This can result in outputs that are coherent at the surface level but structurally or factually flawed when examined rigorously. Without mechanisms for enforcing logical constraints or incorporating symbolic reasoning, LLMs risk generating plausible-sounding yet misleading content~\cite{Guo2024EconNLIEL}.

\textbf{Temporal Fragility in Dynamic Environments.}
Economic and business domains are characterized by constant change, whether in policy regimes, market structures, or consumer preferences. LLMs, especially those based on static corpora, often lack the ability to adapt in real-time. This poses challenges in domains like trading or policy forecasting, where temporal sensitivity is critical.

\textbf{Interpretability and Compliance in High-Stakes Settings.}
LLMs are often black-box systems, making it difficult to audit or justify their outputs. In regulated domains such as lending, taxation, or corporate disclosure, this lack of transparency raises legal and ethical concerns~\cite{huang2025survey}. Explainability remains an open challenge, particularly for decisions with material consequences.

\textbf{Bias Propagation and Ethical Risks.}
LLMs inherit biases present in their training data. This presents significant risks in domains where decisions affect individuals or communities, such as credit scoring, hiring, or marketing segmentation~\cite{feng2023empowering, fotoh2023exploring, albrecht2022despite}. Ensuring fairness, representativeness, and accountability requires careful system design and ongoing evaluation.

\subsubsection{LLM Paradigms from Finance to Marketing}

\noindent\textbf{Grounding LLMs in Economics and Financial Disclosures.}
LLMs are increasingly used to integrate unstructured text with tables, ledgers, time series, and sustainability reports, enabling more comprehensive interpretation of financial information. Retrieval-augmented methods enhance extraction from lengthy documents such as annual filings and regulatory texts~\cite{han2024xbrl, zhao2024revolutionizing}. Advances in factor modeling establish links between textual disclosures and asset behavior~\cite{wang2024llmfactor}, while audit tasks benefit from structured evidence analysis~\cite{gu2024artificial}. Nonetheless, model capacity remains strained by very long documents, and attribution errors pose risks to reliability~\cite{gupta2024systematic}. Emerging work focuses on schema-aware encoders aligned with financial ontologies, hierarchical retrieval strategies for complex reports~\cite{tao2025treerag}, and mechanisms that ensure outputs are explicitly grounded in cited passages or data tables~\cite{berger2023towards}.

\noindent\textbf{Controlled Text Generation with Compliance Safeguards.}
LLMs are increasingly applied to the automated drafting of financial summaries, audit reports, tax memos, policy briefs, and client communications. Controlled text generation has been shown to enhance efficiency in accounting and tax reporting, with case studies documenting substantial cost savings in finance operations~\cite{kim2023bloated, li2024applying, hechtner2025design, Deloitte2025ZoraAI}. However, risks remain: fabricated references and inconsistencies can undermine compliance, while authorship and authenticity concerns persist in external-facing communication~\cite{huang2025survey, GolabAndrzejak2023}. To mitigate these issues, emerging research includes developing grounding mechanisms with verifiable citations, enforcing structured output formats, and incorporating human review for materials carrying regulatory or legal implications~\cite{berger2023towards, fohr2023deep}.

\noindent\textbf{Toward Reliable Agent-Based Simulation with LLMs.}
LLM-driven agent-based environments are increasingly used to evaluate policies, trading strategies, and consumer responses prior to real-world deployment. Economic models now employ language-based agents to approximate strategic interactions and macroeconomic scenarios~\cite{Li2023EconAgentLL, woo2024llm}, while marketing studies deploy synthetic consumers to test campaigns and anticipate likely outcomes~\cite{Sarstedt2024UsingLL, Wang2024LargeLM}. In finance, agent societies are applied to investigate microstructure dynamics~\cite{xiao2024tradingagents}. Despite these advances, the limited depth of reasoning, prompt sensitivity, and reproducibility concerns reduce overall reliability~\cite{Guo2024EconomicsAF, Ross2024LLMEM, Patten2023EvaluatingDS}. Emerging research highlights the integration of explicit constraints from economics and accounting principles, the application of structured reasoning techniques to improve logical consistency~\cite{huang2022towards}, and the validation of simulations against experimental or real-world data~\cite{Wei2022ChainOT, Quan2024EconLogicQAAQ}.

\noindent\textbf{Adapting LLMs to Dynamic Environments with Causal Awareness.}
This line of work focuses on aligning model outputs with evolving conditions and distinguishing correlation from underlying causal mechanisms. Recent studies demonstrate potential in linking narratives to market factors and adapting to changing financial environments~\cite{wang2024llmfactor, wang2024quantagent}. Yet static fine-tuning remains ill-suited to non-stationary settings~\cite{ni2024harnessing}, and outputs frequently capture surface-level correlations without valid causal inference, which constrains their utility in pricing, policy, and marketing applications~\cite{albrecht2022despite}. Promising research explores online learning approaches capable of detecting shifts, integrates econometric methods for causal identification~\cite{kiciman2023causal}, and develops benchmarks to evaluate counterfactual reasoning in business and economic contexts~\cite{Quan2024EconLogicQAAQ}.

\noindent\textbf{LLMs in High-Stakes Decision-Making.}
Interactive LLM-driven systems that synthesize diverse information and recommend actions under uncertainty are increasingly applied in domains such as trading, corporate finance, credit scoring, and pricing. Specialized designs incorporating LLMs have enhanced responsiveness to market narratives and integration of structured data for investment and corporate decision-making~\cite{Li2023TradingGPTMS, wang2024quantagent, gomes2024automating}, while LLM-powered multi-agent approaches now coordinate distinct roles including analysis, risk evaluation, and execution ~\cite{xiao2024tradingagents, yu2024fincon}. Despite these advances, precision in numerical reasoning remains limited \cite{drinkall2025forecasting}, performance often deteriorates under shifting conditions \cite{ni2024harnessing}, and outputs must be sufficiently transparent to withstand regulatory scrutiny~\cite{huang2025survey}. To move forward, promising research directions integrating LLMs with external tools for exact calculation~\cite{chen2022program}, designing methods that attach confidence levels to recommendations~\cite{fu2025deep}, and embedding domain-specific controls that reflect supervisory requirements in finance and insurance~\cite{koa2024learning, yang2025fraud}.

\noindent\textbf{Building Trust in Regulated and Consumer-facing Applications.}
Recent work has introduced evaluation benchmarks in auditing and taxation to reveal both strengths and weaknesses of LLM-supported workflows~\cite{wangauditbench, choi2025taxation}, while growing evidence documents risks of bias in credit scoring, hiring, and market segmentation~\cite{feng2023empowering, fotoh2023exploring, albrecht2022despite}. At the same time, opaque reasoning processes hinder review, and accountability often remains unclear in LLM-assisted decision-making~\cite{huang2025survey, de2024will}. To address these concerns, emerging approaches emphasize standardized logging of prompts, retrievals, and outputs, domain-specific fairness audits, and the integration of Human-in-the-Loop pipelines for high-stakes contexts~\cite{cashion2024generative}.

\subsection{Models as Instruments: LLMs for Science and Engineering}

LLMs are best understood here not as stand-alone "oracles," but as \emph{instruments} that scientists and engineers wield alongside established apparatus such as microscopes, sequencers, finite-element solvers, and CAD/EDA toolchains. As language-native interfaces, they read and write across papers, protocols, code, units, and standards; compose with simulators and databases; and help translate intent ("what to study or build") into operational artifacts ("how to compute, test, and document"). In this sense, LLMs extend the laboratory and the design studio: they lower access costs to sophisticated workflows, accelerate iteration on hypotheses and prototypes, and surface relevant prior art—while remaining dependent on physically grounded models, measurements, and validation.

This section examines LLMs as "instruments" along four axes. First, we probe individual disciplines to characterize where LLMs already help (literature synthesis, code/configuration, early-stage design) and where they fall short (mechanistic fidelity, geometric reasoning, safety-critical accuracy). Second, we identify shared opportunities that recur across fields, especially simulator-in-the-loop workflows and domain-specific grounding. Third, we analyze common limitations—including validation bottlenecks and hallucination risks—that constrain deployment in high-stakes settings. Finally, we articulate cross-domain paradigms that organize effective use (model interfaces, knowledge translation, generative design, education, and human-in-the-loop verification), emphasizing that rigorous science and engineering require LLMs to be embedded within auditable, constraint-aware pipelines rather than replacing them.

\subsubsection{Probe into Individual Disciplines}

\noindent \textbf{Mathematics.} In mathematics, LLMs are beginning to reshape both research and education by enhancing proof development, conjecture generation, and individualized learning. While they show notable promise in symbolic reasoning and problem solving, challenges remain in scaling formal verification and ensuring rigor across diverse mathematical domains.

\begin{itemize}[leftmargin=10pt]
\item \textbf{Mathematical Research.} LLMs can assist mathematicians by streamlining the proof development process, automatically verifying logical arguments, and converting heuristic reasoning into formally structured proofs. They can analyze large collections of mathematical literature and datasets, thus identifying novel patterns and generating new conjectures that might otherwise remain undetected. By reducing the time and effort required for routine tasks and offering alternatives to traditional proof assistants, LLMs help lower the barriers to entry in specialized mathematical domains. Furthermore, through benchmarking on datasets such as MATH, AIME, and GSM8K, LLMs have demonstrated potential in reliably addressing complex problem-solving tasks, making them a compelling tool for both theoretical exploration and practical applications.

\item \textbf{Education in Mathematics.}
LLMs offer promising support in math education by providing personalized tutoring that adapts to each student’s learning pace and style. They can generate tailored, step-by-step explanations and alternative problem-solving approaches, which enable students to gain a deeper understanding of mathematical concepts. Empirical studies have shown that students benefit significantly when they first try to solve problems on their own and then consult LLM-generated guidance, as this approach reinforces learning and improves performance on subsequent assessments.
Additionally, LLMs can assist teachers by automating routine tasks such as creating practice problems and grading assignments, allowing educators to devote more time to direct student interaction and conceptual teaching.
\end{itemize}

\noindent \textbf{Physics and Mechanical Engineering.} In physics and mechanical engineering, LLMs are beginning to show promise across a variety of highly structured and physically grounded tasks, yet face fundamental limitations due to the complexity of geometric reasoning and numerical simulation.

\begin{itemize}[leftmargin=10pt]
\item \textbf{CAD modeling and generative design} represent one of the most active areas of exploration. Systems like CadVLM~\cite{wu2024cadvlm} demonstrate the potential of LLMs to translate textual descriptions into parametric sketches, while datasets such as the Fusion 360 Gallery~\cite{willis2021fusion} and FreeCAD-based code corpora offer initial scaffolds for learning CAD workflows. However, geometric data remains difficult to represent in language, and most current CAD code datasets are brittle, small in scale, and prone to syntactic or logical errors. LLMs struggle with spatial reasoning and lack a robust understanding of geometric constraints, making end-to-end design generation an open challenge.

\item \textbf{Simulation input generation for FEA and CFD} is another emerging application. Benchmarks such as FEABench~\cite{mudur2025feabench} evaluate how well LLMs can produce valid simulation input decks from natural language. LangSim~\cite{langsim2024} further showcases the integration of LLMs with simulation tools for materials and atomistic modeling. These examples demonstrate early success in bridging language and numerical domains. Nonetheless, large-scale paired datasets linking natural language to simulation files and results are still rare, and LLMs often lack the inductive biases and unit reasoning required for physically meaningful output.

\item \textbf{Multimodal and multi-agent systems} like MechAgents~\cite{ni2023mechagents} and LangSim~\cite{langsim2024} highlight a promising direction in which LLMs orchestrate tool use across CAD, solvers, and databases. Such systems show that closed-loop pipelines---design, simulate, validate---can be coordinated via LLM-based agents. Still, these pipelines require careful prompting, are sensitive to function hallucination, and rely heavily on handcrafted tool interfaces.

\item \textbf{Text generation and report summarization} is a relatively mature use case. LLMs can assist in drafting simulation logs, experiment summaries, and design documentation. While valuable, these outputs often lack the physical accuracy or constraint enforcement necessary for high-stakes engineering decisions.
\end{itemize}

\noindent \textbf{Chemistry and Chemical Engineering.} In chemistry and chemical engineering, LLMs are beginning to reshape both fundamental scientific inquiry and industrial applications by accelerating molecular discovery, reaction prediction, and process optimization. While they show strong potential in unifying symbolic and numerical reasoning for chemical systems, current limitations in generalization, 3D structural understanding, and multi-step synthesis planning highlight the need for domain-specific adaptations and robust benchmarks.

\begin{itemize}[leftmargin=10pt]

    \item \textbf{Fundamentals of Chemistry.}  
    Chemistry is fundamentally concerned with understanding the composition, structure, properties, and transformations of matter, spanning diverse tasks including qualitative and quantitative analysis, reaction mechanism elucidation, molecular synthesis, and computational modeling \cite{wikipedia_chemistry}. Traditional methods leverage spectroscopic \cite{claridge2016high}, chromatographic \cite{smith2013chromatography}, and mass-spectrometric analyses \cite{aksenov2017global} to characterize molecular structures and quantify components, while experimental studies of thermodynamics and kinetics reveal deeper insights into chemical behavior \cite{mortimer2002chemical}.

    \item \textbf{Principles of Chemical Engineering.}  
    Chemical engineering complements these scientific principles by translating laboratory discoveries into scalable industrial operations \cite{wikipedia_chemical_engineering}. It focuses on process engineering—including reactor design, separation technologies, and process control—equipment optimization, fluid dynamics analyses adhering to industrial standards, and sustainability considerations through environmental and lifecycle assessments \cite{fogler2010essentials}. Together, chemistry and chemical engineering bridge the gap between molecular-level understanding and large-scale implementation, addressing applications in pharmaceuticals, materials, and environmental sciences.

    \item \textbf{Transformative Role of LLMs.}  
    Recently, LLMs have demonstrated transformative potential across chemistry and chemical engineering, providing unified computational frameworks for molecular description, property prediction, reaction outcome forecasting, inverse design, and chemical knowledge extraction \cite{coley2019graph,xie2024chemical}. Domain-specialized LLMs have considerably accelerated progress by replacing manual, feature-intensive approaches with unified latent representations \cite{m2024augmenting}, streamlining reaction planning via integrated forward and retrosynthesis modeling \cite{jablonka2024leveraging}, automating molecular optimization through intuitive prompt-based generation, and converting unstructured chemical narratives into actionable structured data \cite{ramos2025review}.

    \item \textbf{Current Progress and Limitations.}  
    Despite these advances, substantial challenges remain. Models trained on biased datasets—dominated by well-studied, drug-like molecules—often struggle to generalize to novel chemical spaces, leading to inaccuracies or chemically implausible outputs \cite{liu2022chemical,mccloskey2019using}. Molecular textualization methods still face limitations in handling complex, novel structures or evolving naming conventions, and prediction methods struggle with capturing nuanced 3D conformations, quantum mechanical effects, and activity cliffs \cite{ramos2025review}. Synthetic planning algorithms frequently underperform on multi-step or cascade reactions, especially under underrepresented conditions \cite{berreziga2025combining}. Moreover, the limited context window of current LLM architectures hampers lengthy procedural reasoning, multi-step synthesis planning, and accurate provenance tracking—critical for laboratory adoption and regulatory compliance \cite{liu2024large}.

    \item \textbf{Benchmark Landscape and Gaps.}  
    The current benchmarks for LLMs in chemistry and chemical engineering span a diverse array of tasks, from molecular property prediction (e.g., MoleculeNet) \cite{wu2018moleculenet} to molecular generation (e.g., GuacaMol) \cite{brown2019guacamol} and chemical text mining. These datasets employ unified SMILES preprocessing, standardized split strategies (random, scaffold, temporal), and consistent task definitions to ensure reproducible and fair evaluation. However, most benchmarks remain focused on drug-like organic molecules and patent reactions, offering limited support for domains such as inorganic chemistry, environmental pollutants, failed experiments, or multi-step synthesis planning. The mainstream benchmarks can be broadly categorized into three paradigms—molecular property prediction (Mol2Num), reaction outcome and classification (Mol2Mol/Mol2Num), and molecule generation (Text2Mol/Text2Text)—yet their reliance on linear token sequences limits direct encoding of 3D structure and chemical rules, resulting in suboptimal performance on complex tasks such as stereoselectivity prediction and multi-step synthesis planning.

\end{itemize}

\noindent \textbf{Life Sciences and Bio-engineering.} In the life sciences, LLMs are increasingly integrated into biological discovery and clinical practice, while in bio-engineering they support the translation of fundamental insights into practical technologies. These models show promise in sequence analysis, medical text understanding, and generative design, yet still face critical challenges in handling multimodal data, ensuring safety, and grounding outputs in biological mechanisms.

\begin{itemize}[leftmargin=10pt]

    \item \textbf{Fundamentals of Life Sciences.}
Life sciences investigate the origin, structure, function, and evolution of living systems, from biomolecules to whole ecosystems \cite{blumenthal1997withholding,lodish2008molecular}.  Classical research tasks include decoding genetic information through sequencing and PCR \cite{hearn2010dna,erlich1989pcr}, resolving macromolecular structures via X-ray crystallography or cryo-EM \cite{smyth2000x,schwieters2006using}, and elucidating physiological mechanisms with animal models and clinical trials \cite{verkhratsky2018physiology,van2019personalised}.  These endeavours supply the empirical foundations on which modern biomedical science is built.

    \item \textbf{Principles of Bio-engineering.}
Bio-engineering translates biological insight into practical technologies, integrating chemical, mechanical, and electrical engineering with molecular and cellular biology \cite{kumar2020bioengineering,jaklenec2012progress}.  Core domains include genetic and cellular engineering (e.g., CRISPR-based editing) \cite{jiang2017crispr}, tissue engineering and biomaterials \cite{ikada2006challenges,ratner2004biomaterials}, bioprocess scale-up in fermenters \cite{doran1995bioprocess}, and computational modeling of complex biological systems \cite{gentleman2005bioinformatics}.  These principles bridge discovery and deployment—turning laboratory breakthroughs into vaccines, biologics, and medical devices.

    \item \textbf{Transformative Role of LLMs.}
LLMs have recently emerged as unifying tools that read, reason about, and even design biological sequences and clinical narratives \cite{ji2021dnabert,singhal2023large}.  In genomics, transformer-based encoders such as Enformer and HyenaDNA capture megabase-scale regulatory syntax, improving enhancer and eQTL prediction by 20–40\% over CNN/RNN baselines \cite{Avsec2021.04.07.438649,nguyen2023hyenadna}.  Clinical variants like BEHRT and GatorTron convert longitudinal EHRs into patient-level risk scores and decision support, outperforming traditional models even with limited fine-tuning data \cite{li2020behrt,yang2022large}.  Generative models (e.g., ProGen2 for proteins, CancerGPT for drug synergy) accelerate hypothesis generation by proposing functional sequences or synergistic drug pairs that are later validated in the lab \cite{nijkamp2023progen2,li2024cancergpt}.

    \item \textbf{Current Progress and Limitations.}
Despite rapid gains, LLMs still face key obstacles.  Ultra-long genomic contexts (>1Mb) degrade accuracy, and integrating 3-D chromatin contacts or multi-omic layers remains an open problem \cite{zhou2023dnabert2}.  In clinical settings, models can hallucinate diagnoses or misinterpret rare conditions, raising safety concerns \cite{nori2023capabilities}.  Data bias is pervasive—human-centric genomes, single-institution EHRs, and English-only corpora limit cross-population and cross-language generalization \cite{johnson2016mimic}.  Finally, interpretability lags behind domain expectations; attention heat-maps seldom satisfy clinicians who need mechanistic explanations for high-stakes decisions \cite{tran2023instruction}.

    \item \textbf{Benchmark Landscape and Gaps.}
Benchmark suites now span four task families.  (i) \emph{Sequence-based} benchmarks—BEND for DNA and BEACON for RNA—evaluate functional-element annotation, structure prediction, and variant effect scoring \cite{zhou2025genomeocean,marin2023bend}.  (ii) \emph{Clinical structured-data} tasks quantify performance in language generation (e.g., ClinicalT5) and EHR prediction (e.g., Med-BERT) \cite{lu2022clinicalt5,rasmy2021med}.  (iii) \emph{Textual knowledge} benchmarks such as MedQA, MedMCQA, PubMedQA, and MedNLI probe factual recall and reasoning over biomedical literature and clinical notes \cite{krithara2023bioasq,romanov2018lessons}.  (iv) \emph{Hybrid outcome-prediction} datasets cover drug synergy (DrugCombDB derivatives) and protein modeling (ESM-Fold, ProGen) \cite{xu2023dffndds,lin2022language}.  However, most benchmarks isolate single modalities, underrepresent non-model organisms, rare diseases, and multilingual corpora, and seldom test cross-modal reasoning or end-to-end “design-build-test” loops.  Expanding dataset diversity, embedding biological priors into architectures, and coupling LLMs with wet-lab feedback loops are critical next steps toward trustworthy, generalizable bio-AI systems.
    
\end{itemize}

\noindent \textbf{Earth Science and Civil Engineering.} LLMs in Earth sciences and civil engineering are emerging as valuable tools for analyzing geospatial data, supporting compliance and safety tasks, and enabling natural language interaction with engineering workflows. Despite progress in domain-specific fine-tuning and tool integration, key challenges include the lack of grounded physical reasoning, data scarcity, and the need for reliable validation mechanisms.

\begin{itemize}[leftmargin=10pt]

    \item \textbf{Earth Sciences.}
Recent works such as RSGPT~\cite{hu2025rsgpt}, RS-LLaVA~\cite{bazi2024rs}, and GeoChat~\cite{kuckreja2024geochat} demonstrate the utility of vision-language models (VLMs) fine-tuned for remote sensing imagery. They enable captioning, visual question answering, and spatial reasoning on satellite data. GeoGPT~\cite{zhang2023geogpt} further integrates LLMs with GIS toolchains, automating geospatial tasks through tool selection and code generation. Remote sensing imagery presents unique difficulties, such as high resolution, varied scale, and complex viewing angles. Generic VLMs perform poorly in this domain without specialized fine-tuning. Additionally, current LLM agents struggle with tool orchestration and often hallucinate function calls, limiting the reliability of autonomous geospatial workflows.

    \item \textbf{Scientific Literature and Report Summarization.}
LLMs can effectively summarize geoscience documents, such as environmental impact assessments and geological surveys. Tools like LitLLM~\cite{agarwal2024litllm} and GeoBERT~\cite{denli2021geoscience} have shown success in domain-specific QA and summarization tasks. GeoBench~\cite{fu2023gptscore} provides a benchmark demonstrating LLMs’ competence in factual recall and reasoning in geoscientific contexts. While LLMs perform well on surface-level summaries, they may miss domain nuances or generate hallucinated facts, especially when faced with unstructured or outdated data. There is also a lack of high-quality, standardized corpora for fine-tuning LLMs in Earth sciences.

    \item \textbf{Natural Language Interfaces to Geospatial Tools.}
Systems like GeoLLM-Engine~\cite{singh2024geollm} and Change-Agent~\cite{liu2024change} demonstrate how LLMs can be used to translate user queries into API calls, enabling natural-language access to complex Earth observation pipelines. LLMs lack grounded physical reasoning and struggle with multistep logical planning required for sequential tool invocation. Most systems still rely on hand-curated tool schemas and cannot generalize across tasks without extensive prompting or supervision.

    \item \textbf{Civil Engineering.}
LLMs have been applied to translate regulatory texts and check inspection documents for compliance. LLM-FuncMapper~\cite{zheng2023llm}, AutoRepo~\cite{PU2024124601}, and GPT-based compliance check systems~\cite{liu2023gpt} automate interpretation of building codes and structural standards. Despite promising results, current systems lack precision guarantees. Hallucinations in regulatory interpretation can lead to unsafe outcomes. Moreover, design codes vary by region and often contain exceptions and edge cases that are hard to encode in prompts or models. Other applications include structural health monitoring, design support, simulation-aware code generation, and urban planning, but these remain dependent on human oversight and face fundamental challenges in constraint reasoning and validation.

\end{itemize}

\noindent\textbf{Computer Science and Electrical Engineering.} 
LLMs are emerging as valuable partners alongside traditional methods in computer science and electrical engineering by automating repetitive coding, assisting debugging, explaining and refactoring large codebases, generating HDL, aiding functional verification, and guiding high-level synthesis.
Although the benefits of LLMs in these domains, further research to improve accuracy and reliability, seamless integration with existing tools and EDA flows, exploration of advanced design and optimization roles, and solutions to hardware data scarcity is necessitate. Overall, the integration of LLMs in both software development and circuit design holds the potential to accelerate design processes, reduce human error, and enhance overall productivity by automating routine tasks, improving debugging efficiency, and facilitating better code and design comprehension. Continued research and development in this area promise to further harmonize traditional engineering methodologies with state-of-the-art AI assistance, ultimately leading to faster time-to-market and improved reliability in both software and hardware projects.

\begin{itemize}[leftmargin=10pt]
    \item \textbf{Software Engineering.}
LLMs can automate repetitive coding tasks by generating boilerplate code, standard data structures, and common algorithms from natural language descriptions. This capability allows developers to focus on higher-level reasoning and complex problem solving rather than manual implementation. LLMs also serve as effective code assistants by explaining complex code snippets, offering real-time suggestions, and supporting debugging—analyzing error messages and identifying the causes of bugs. In addition, LLMs can help manage large codebases by summarizing functionality and suggesting improvements, thus facilitating refactoring and maintenance, which ultimately leads to improved code quality and reduced development time.

\item \textbf{Circuit Design.}
In the domain of digital circuit design, LLMs show promise by automating the generation of HDL code (e.g., Verilog and VHDL) from natural language specifications. This process enables designers to quickly prototype complex digital circuits and focus on architectural decisions rather than routine coding. LLMs can assist with functional verification by generating testbenches and assertions, thereby reducing the time spent on manual testing and debugging of circuit designs. Furthermore, LLMs are being explored for high-level synthesis, where they help translate high-level descriptions into efficient hardware implementations, making hardware design more accessible and shortening design cycles.

\end{itemize}

\subsubsection{Shared Opportunities}

Science and engineering are disciplines fundamentally concerned with modeling, measurement, and controlled transformation of the physical world. Whether formulating governing equations, designing experiments, or scaling technologies into applications, these fields rely on structured representations, such as mathematical formalisms, simulation codes, and empirical datasets, to generate reliable knowledge. The advent of LLMs introduces a new epistemic layer: language itself becomes a computational interface to scientific reasoning. Instead of relying solely on symbolic or numerical methods, scientists and engineers can now interact with models, data, and tools through natural language, enabling qualitative insight and quantitative rigor to coexist within unified workflows.

Evidence of this shift is already visible across domains. In mathematics, LLMs demonstrate capabilities in conjecture generation and automated proof verification~\cite{romera2024mathematical}. In chemistry and materials science, pretrained molecular language models achieve strong performance on benchmarks such as MoleculeNet and GuacaMol~\cite{wu2018moleculenet, brown2019guacamol}. In life sciences, genomics models such as DNABERT and HyenaDNA~\cite{ji2021dnabert,nguyen2023hyenadna} capture long-range sequence dependencies, while clinical models like GatorTron~\cite{yang2022large} improve patient risk forecasting. Physics and engineering prototypes such as FEABench, PDEBench, and GeoGPT~\cite{mudur2025feabench,hoellein2022pdebench,zhang2023geogpt} show that LLMs can already interface with solvers, digital twins, and geospatial pipelines. Despite uneven adoption, the trajectory across disciplines are similar: enthusiasm is strong, prototypes are proliferating, but broad deployment remains constrained by issues of validation, interpretability, and domain safety.

Across the sciences and engineering disciplines, LLMs are opening up new possibilities that cut across traditional boundaries. While the specific applications vary, from theorem proving in mathematics to drug discovery in chemistry or digital twins in civil engineering, common opportunities emerge in how these models unify symbolic reasoning, empirical data, and computational workflows. The following themes highlight shared directions where LLMs are reshaping research and practice.

\textbf{Domain-Specific Grounding and Hybridization.}  
LLMs benefit significantly from corpora tailored to scientific and engineering domains. In Earth sciences and civil engineering, this means incorporating geospatial, structural, and regulatory texts; in physics and materials, fine-tuning on equations, databases, and solver input files; and in chemistry, textualizing molecular representations such as SMILES or IUPAC names. When coupled with physics-based simulators or domain ontologies, such specialization strengthens factual grounding and reduces hallucination, creating hybrid models that connect symbolic language with mechanistic understanding.

\textbf{Integration with Modeling and Simulation Pipelines.}  
Rather than replacing established solvers, LLMs serve as natural front-ends to modeling environments. They can generate finite element decks, PDE solver configurations, and computational chemistry workflows, automating tedious setup steps while lowering the barrier to entry for multiphysics tools. This role extends across fields: configuring hydrological models in Earth sciences, assisting with CalculiX or OpenFOAM \cite{openfoam} inputs in physics, or supporting retrosynthetic route planning in chemistry. In each case, LLMs augment rather than supplant numerical rigor.

\textbf{Exploration, Design, and Autonomous Decision-Making.}  
By reasoning over constraints and large search spaces, LLMs enable exploratory and design-oriented tasks. AlphaFold~\cite{jumper2021highly} represents the clearest demonstration of how learned models can autonomously explore combinatorial search spaces. In this case, protein conformations, yielding solutions that reshape both life sciences and downstream engineering. Beside that, geospatial agents demonstrate constrained action planning in environmental workflows; and materials informatics systems assist in property retrieval and selection. These examples illustrate a broader opportunity: using language-driven agents to navigate high-dimensional design landscapes while remaining anchored to domain constraints.

\textbf{Interfaces, Education, and Human-Centered Integration.}  
A final opportunity lies in making complex systems more accessible. LLMs enable natural language interaction with digital twins, building information models, and geospatial APIs; they also help contextualize physics or math concepts for students, or draft clinical and engineering reports. This educational and communicative role complements technical automation, fostering human-centered design, transparency, and ethical alignment. Especially in critical infrastructure or healthcare contexts, pairing accessibility with explainability and governance is central to responsible deployment.

\subsubsection{Common Limitations}

Despite promising early progress, fundamental barriers constrain the broader adoption of LLMs across science and engineering. While each discipline highlights different technical bottlenecks, recurring challenges emerge that reflect the current architectural limits of language models as well as the structural realities of scientific practice. These limitations must be addressed before LLMs can transition from proof-of-concept tools to reliable components of critical workflows.

Evidence of these limitations can be seen across domains. In Earth sciences and engineering, models cannot capture the fidelity of numerical solvers for seismic or hydrodynamic systems~\cite{mcguffie2001forty}. In chemistry and life sciences, generative models often produce chemically invalid or biologically implausible candidates~\cite{berreziga2025combining,ramos2025review}, with even promising predictions requiring costly validation in wet labs or clinical trials~\cite{nijkamp2023progen2}. In mathematics, LLMs reproduce surface patterns rather than true abstractions, while in CSEE, generated code for complex scenarios remains brittle and hallucination-prone~\cite{zhang2023llm4dv}. These examples illustrate that while enthusiasm is justified, practical deployment requires caution and new safeguards.

\textbf{Insufficient Physical and Mechanistic Modeling.}  
LLMs tend to capture shallow patterns and struggle to encode conservation laws, governing equations, or mechanistic reasoning. For example, multiphysics PDEs and nonlinear continuum models require stability and boundary guarantees that language models cannot enforce~\cite{zienkiewicz2005finite,bathe2006finite}. In chemistry, models memorize statistical patterns but cannot reason through mechanistic steps such as radical cascades. As a result, LLMs can generate code for solvers or reactions but cannot yet provide mechanistic fidelity comparable to validated tools.

\textbf{Experimental and Validation Gaps.}  
Across life sciences, chemistry, and materials, in-silico outputs must be confirmed experimentally. Molecules generated by Adapt-cMolGPT or ProGen2 require synthesis and activity assays~\cite{nijkamp2023progen2}, proteins must be expressed, and materials tested before real-world use~\cite{forster2015materials}. This “last-mile” validation bottleneck means that LLMs accelerate hypothesis generation but adoption remains limited without automated design–build–test–learn loops in many science and engineering fields.

\textbf{Hallucination and Lack of Verifiability.}  
A common limitation is the generation of plausible but false outputs. Chemistry benchmarks such as QCBench document "valid-looking" molecules that violate valence or synthesis rules~\cite{xie2025qcbench}. Clinical NLP systems have fabricated diagnoses or medications~\cite{nori2023capabilities}, while engineering contexts show unit inconsistencies or fabricated parameters~\cite{wan2024generative}. Without reliable verification pipelines, hallucinations undermine trust in safety-critical workflows.

\textbf{Data Sparsity and Generalization Limits.}  
Many scientific fields are constrained by data scarcity. Rare subfields in chemistry (e.g., inorganic catalysis, polymer chemistry) lack sufficient curated datasets, leading to degraded performance compared to common drug-like molecules. Genomics models such as Enformer and HyenaDNA degrade in accuracy over ultra-long contexts~\cite{nguyen2023hyenadna}, while geoscience applications suffer from incomplete records in hazardous or inaccessible regions~\cite{sun2021review}. Bias toward well-studied domains hampers transfer to frontier problems.

\textbf{Limits of Reasoning and Innovation.}  
Finally, LLMs often mimic surface patterns rather than demonstrating genuine abstraction or novelty. In mathematics, models fail on distractor-laden or multi-step reasoning tasks~\cite{romera2024mathematical}, echoing known results without principled innovation. In software and hardware design, generated code or architectures lack robustness for complex logic or micro-architectural innovation~\cite{fang2024assertllm}. While valuable as assistants, current LLMs contribute little to advancing conceptual frontiers without significant human guidance.

\subsubsection{LLM Paradigms for Science and Engineering}
Science and engineering research follows a distinctive lifecycle: structuring knowledge into usable forms, generating and testing new hypotheses, validating designs through simulation and experimentation, and training practitioners while safeguarding against dual-use risks. Each stage presents opportunities for LLMs to act as mediators, accelerators, validators, trainers and safeguards. This section highlights the challenges and future directions for researchers to explore in each stage. By aligning our discussion with this lifecycle, we capture how LLMs are beginning to transform research workflows in ways that differ from other domains, owing to the uniquely formal, verifiable, and safety-critical nature of scientific and engineering practice. We therefore identify the following five paradigms of adoption: (1) LLMs as mediators to unify diverse knowledge representations, (2) constraint-grounded hypothesis generation for early-stage filtering, (3) navigating vast designs through verifiable simulation, (4) education and training as adaptive assistants, and (5) safeguards against dual-use risks. Taken together, these paradigms illustrate both the promise and responsibility of integrating LLMs into the fields of science and engineering.

\noindent\textbf{LLM as Mediator to Unify Diverse Knowledge Representations.}  
A central promise of LLMs is their ability to unify heterogeneous sources of knowledge—papers, technical manuals, databases, and design protocols—within a single natural language interface. By acting as mediators, they can lower barriers to information access, making specialized tools and workflows more widely usable across disciplines.  What makes this frontier unique to science and engineering is that much of the critical knowledge in science and engineering is not written in prose or tables but encoded in formats such as equations, symbolic grammars, CAD sketches, PDE solver decks, or molecular graphs, where validity depends on strict adherence to physical laws, mathematical consistency, and domain-specific syntax, making translation a non-trivial challenge. Researchers address this by combining LLMs with structured representations and external validators. In chemistry, Struct2IUPAC converts between molecular string formats with near-99\% accuracy~\cite{krasnov2021struct2iupac}, while SELFIES enforces syntactic validity in molecule generation~\cite{krenn2022selfies}. In civil engineering, PlanGPT parses building codes into enforceable design requirements~\cite{zhu2024plangpt}, and in earth sciences, systems like GeoGPT and HydroSuite generate GIS queries and hydrological models from natural language prompts~\cite{zhang2023geogpt,pursnani2024hydrosuite}. Physics and engineering models such as CadVLM and FEABench~\cite{wu2024cadvlm,mudur2025feabench} translate prompts into valid CAD sketches or PDE solver configurations, directly bridging language and symbolic computation. These advances suggest that, as a future direction for researchers, LLMs can become robust semantic front-ends to the formal languages that underpin scientific and engineering practice, which are both linguistically accessible and technically rigorous.

\noindent\textbf{Constraint-Grounded Hypothesis Generation for Early-Stage Filtering.}
Foundational sciences and engineering are governed by hard constraints (e.g., conservation laws, symmetries, chemical valence grammars), which enables LLMs to generate and cheaply pre-filter hypotheses before costly experimentation. Thus, it benefits researchers because these constraint-grounded pipelines narrow vast design spaces quickly, and shift effort from expensive wet-lab/field trials to inexpensive automated filtering, accelerating iterations of ideas while maintaining plausibility. 

For example, chemistry-focused models such as ChatMol and Hyformer propose molecules that satisfy multi-property objectives~\cite{zeng2024chatmol,yan2023hyformer}, while life science models like ProGen2 and ESM-3 generate foldable proteins to seed design–build–test–learn cycles~\cite{nijkamp2023progen2}. These advances parallel the transformative impact of AlphaFold, which reframed protein folding as a representation learning problem and unlocked new possibilities in molecular biology~\cite{jumper2021highly}. In physics and materials, equivariant architectures embed conservation laws and symmetry priors, allowing LLM-orchestrated proposals to be triaged against surrogate models that respect physical constraints~\cite{batzner2022nequip, xu2025envirodetanet}. In computer systems, LLM agents increasingly delegate logical consistency checks to SMT/SAT solvers during program or specification synthesis, automatically filtering out infeasible hypotheses without human input~\cite{ssv2025}. Together, these constraint-grounded strategies demonstrate how LLMs can accelerate hypothesis generation while preserving mechanistic fidelity, enabling researchers to locate promising ideas earlier, reduce wasted experimental effort, and iterate at scale in ways that are both efficient and scientifically credible.

\noindent\textbf{Navigating Vast Designs Through Verifiable Simulation. }  
A central challenge in science and engineering is the need to search enormous combinatorial spaces, such as proteins, PDE configurations, or CAD geometries, which is fundamentally different from domains like law, art, or finance, where outputs are subjective, context-dependent, or influenced by non-deterministic social factors. This combination of expansive search and verifiable feedback makes LLMs uniquely suited to technical discovery, where LLMs can generate candidate solutions that can be rapidly filtered through verification. Current systems already illustrate this promise: GeoGPT and HydroSuite accelerate environmental modeling~\cite{zhang2023geogpt, pursnani2024hydrosuite}, CadVLM and FEABench enable rapid prototyping in engineering~\cite{wu2024cadvlm, mudur2025feabench}, HDL copilots support verifiable hardware design~\cite{zhang2023llm4dv}, and models like ChatMol or ProGen2 generate testable molecules and proteins~\cite{zeng2024chatmol, nijkamp2023progen2}. Yet challenges remain, as many outputs are superficially plausible but chemically invalid or physically unstable. Researchers are responding by embedding verification directly into generation pipelines, from chemistry’s forward–retro loops~\cite{berreziga2025combining} and QCBench screening~\cite{xie2025qcbench} to solver-integrated orchestration agents like LangSim~\cite{langsim2024}. The path forward is clear: pair LLM-driven exploration with rigorous validation so that vast design spaces can be navigated responsibly and translated into quick and reliable scientific advances.

\noindent\textbf{Education and Training in Sciences and Engineering as Adaptive Assistants. }  
The diffusion of new knowledge requires not only discovery but also education and training. However, foundational subjects such as chemistry, mathematics, and physics are often perceived as challenging, in part because the knowledge is abstract and traditional teaching emphasizes memorization and procedural fluency over conceptual understanding~\cite{thurston2005mathematical, kajander2014mathematical, peter2012critical}. LLMs offer new opportunities to lower these barriers by acting as adaptive assistants. For example, in chemistry, ChemLLM provides dialogue-based tutoring aligned with curricula~\cite{zhang2024chemllm}, and benchmark studies show that large models like GPT-4 outperform older systems on explanation and reasoning tasks~\cite{guo2023can}. In mathematics, LLMs can generate stepwise solutions tailored to learners~\cite{romera2024mathematical}, while in physics and engineering, prototypes contextualize advanced concepts and guide novices through toolchains~\cite{polverini2023chatgpt}. These examples illustrate the potential of LLMs to scaffold learning and democratize access to technical knowledge in a personalization fashion. Key challenges now lie in ensuring conceptual accuracy, handling misconceptions, and connecting tutoring with experiential learning, requiring researchers to focus on these directions by developing robust evaluation protocols, integrating LLMs into blended learning environments, and designing oversight mechanisms that allow these models to complement, rather than replace, traditional educational practice. Current efforts such as curriculum-aligned benchmarks~\cite{guo2023can}, misconception-aware tutoring frameworks~\cite{du2024large}, and domain-specific systems like ChemLLM~\cite{zhang2024chemllm} point to practical strategies, suggesting that with sustained refinement LLMs could become indispensable partners in making science and engineering knowledge more accessible, engaging, and effective.

\noindent\textbf{Mitigating Dual-use Risks in Sciences and Engineering with Embedded Verification.}  
Beside common ethical concerns such as bias and reliability, a unique safety concern in the sciences and engineering arises from the possibility of dual-use problem. LLM-generated designs in sciences and engineering domains such as chemistry, biology, or computer sciences can directly enable and accelerate the creation of hazardous materials or malicious software. For example, in chemistry and life sciences, models have been shown capable of suggesting toxic or biologically harmful molecules by reversing drug-discovery pipelines~\cite{tran2023instruction,xie2025qcbench}. In computer and electrical systems, rapid generation of malware or unverified HDL code introduces the risk of systemic vulnerabilities at both software and hardware levels~\cite{fang2024assertllm, kumar2024architectural}. These risks underscore that in technical domains, unsafe outputs are not just misleading, they may be directly weaponizable.

Researchers are beginning to address these challenges with safeguards that are unique to the scientific and engineering context. In chemistry, domain-specific benchmarks such as QCBench screen for chemically invalid or unsafe molecules before they are considered usable~\cite{xie2025qcbench}, while forward–retro synthesis loops embed automated verification into discovery pipelines~\cite{berreziga2025combining}. In engineering, hybrid pipelines integrate LLMs with solvers and simulation sandboxes, ensuring that generated parameters and designs satisfy physical constraints before release~\cite{ni2023mechagents,langsim2024}. In computing, prototype systems combine LLM copilots with formal verification and automated test benches to catch unsafe or malicious code before deployment~\cite{fang2024assertllm}. Parallel to these technical solutions, scholars emphasize the importance of regulatory embedding and human-in-the-loop governance, drawing on traditions of safety codes in engineering and biosafety levels in laboratory science~\cite{tokognon2017structural, wan2024generative}. Together, these strategies illustrate a path forward: by pairing LLM innovation with domain-specific safety checks, simulation-based validation, and oversight frameworks, researchers can mitigate dual-use risks and ensure that powerful models serve as enablers of scientific progress rather than accelerants of harm.

\subsection{Pilot the Present, Plot the Future}

The integration of LLMs across diverse fields has begun to move beyond proof-of-concept applications, offering tangible progress while also highlighting shared opportunities and common challenges for further advancement. In this section, we synthesize insights across three broad domains—(a) arts, letters, and law, (b) economics and business, and (c) science and engineering—to take stock of current developments and chart how LLMs are transforming the landscapes of knowledge creation, disciplinary methods, and cross-field integration.

\subsubsection{Current State}

Across the domains of arts, letters, and law, economics and business, and science and engineering, LLMs have demonstrated several shared advances and achieved breakthroughs in practice:

\begin{itemize}[leftmargin=10pt]
\item  \textbf{Scaling Information Processing and Multimodal Integration.} In history and law, LLMs can now process millions of archival documents or court cases, for example by enabling cross-search within the U.S. Supreme Court case database \cite{shu2024lawllm,law2015dictionary,guha2023legalbench}. In finance, they are used to parse lengthy 10-K filings and extract risk factors (e.g., through the LLM-Factor framework) \cite{harris2024managers, wang2024llmfactor}. In the sciences, LLMs combined with chemical databases assist in molecular screening {\cite{frey2023neural,taylor2022galactica}} and property prediction {\cite{yuksel2023selformer,ahmad2022chemberta}}, significantly shortening drug discovery cycles {\cite{ramos2025review,m2024augmenting,ruan2024automatic}}.

\item  \textbf{Facilitating Knowledge-augmented Decision Support.} In legal contexts, LLMs support normative argumentation and judicial prediction—for instance, Harvard CaseLaw GPT generates reasoning chains grounded in legal precedent \cite{shu2024lawllm}. In economics, agent-based systems replicate game-theoretic scenarios to pre-test policy interventions \cite{Li2023EconAgentLL, Guo2024EconomicsAF, Hao2025AMF}. In engineering, LLMs assist in FEA and PDE modeling \cite{hoellein2022pdebench}, helping structural engineers verify design options more efficiently.

\item \textbf{Enabling Pre-experimental Simulations.} In political science, LLM-driven agents are used to simulate voter reactions under different policy scenarios \cite{louie2024roleplay}, offering early insight into social impact. In financial markets, multi-agent systems such as TradingGPT distribute roles for analysis \cite{Li2023TradingGPTMS}, risk evaluation, and execution, providing a ``pre-experimental'' platform for portfolio management. In life sciences, LLM-driven virtual cell models support preliminary validation of drug responses, reducing costly trial-and-error in the lab \cite{tang2025cellforge,bunne2024build}.

\item \textbf{Lowering Barriers to Complex Technical Workflows.} In accounting and taxation, LLMs are embedded in firm-level reporting pipelines (e.g., Deloitte’s ZoraAI) to generate client reports and assist in audit workflows \cite{Deloitte2025ZoraAI}. In engineering education, LLMs provide natural language interfaces where students can generate MATLAB or COMSOL simulations with conversational commands \cite{jiang2024beyond,abedi2023beyond}. In geosciences, integration with GIS platforms allows natural language–driven geospatial analysis, enabling non-experts to perform complex modeling~\cite{nascimento2024llm4ds,taylor2013ground}.

\item \textbf{Fostering Human–AI Co-creation and Co-production.} In arts and architecture, LLMs have been used to co-generate design iterations with architects, rapidly producing stylistic variations \cite{kumar2024architectural,Galanos2023ArchitextLG}. In marketing, firms employ LLMs to co-create ad copy and user personas, accelerating campaign design \cite{Sarstedt2024UsingLL, Wang2024LargeLM}. In research, models like Galactica can co-author draft surveys with scientists, speeding up knowledge synthesis \cite{taylor2022galactica}.
\end{itemize}

Collectively, these examples illustrate how LLMs are evolving from “language generators” into cross-disciplinary intelligent infrastructures: systems capable of scaling information processing, supporting advanced reasoning, enabling pre-experimental simulations, fostering human–AI co-creation, and lowering the barriers to complex technical workflows—ultimately democratizing access to knowledge and practice.

While the applications of LLMs in arts, letters, and law, economics and business, and science and engineering demonstrate remarkable promise, several common challenges limit their reliability and broader adoption: limited reasoning depth and causal grounding (fluent yet hollow arguments， weak multi-step/game-theoretic reasoning， difficulty honoring mechanisms and conservation constraints); insufficient contextual and institutional anchoring (loss of temporal–cultural–jurisdictional and institutional nuance); deficits in numerical, symbolic, and physical precision; gaps in verification and attribution (hallucinations, mis-citations); opacity and unclear accountability; poor adaptation to non-stationarity and causal structure; weak reproducibility and robustness (prompt sensitivity, unstable simulations/designs); data scarcity and bias amplification; concerns over originality, authorship, and legitimacy; and a “last-mile” burden of validation and compliance. These challenges are not confined to single domains but recur across fields, exposing structural limitations of current approaches. \textbf{These limitations crystallize into three bottlenecks:} (1) moving from surface pattern-matching to verifiable reasoning; (2) adapting static knowledge to dynamic, causal settings; and (3) shifting from merely usable outputs to auditable, accountable systems—prerequisites for deploying LLMs as reliable cross-disciplinary infrastructure in high-stakes contexts.

\subsubsection{Future Path}

Given current state discussed above, LLMs reveal a common pattern: clear opportunities—unprecedented access to knowledge, multimodal and structured integration, agentic ``pre-experiments'', tool-augmented decision support, and education—while simultaneously exposing common challenges in factuality and attribution, depth and causal grounding of reasoning, numerical and physical precision, robustness under non-stationarity, reproducibility, and governance and accountability. Structured from foundational evidence infrastructure to modeling and foresight, then to decision and deployment, and ultimately to governance, capability building, and safety, we outline the path ahead below which translates those opportunities into design principles that directly address the obstacles, providing a roadmap for reliable, auditable, and scalable deployment across disciplines.

\begin{itemize}[leftmargin=10pt]

\item \textbf{Reducing Data-Format Silos via Schema-Aligned Multimodality.} LLMs are expected to natively operate across multimodal data—including text, tables and ledgers, time series, figures, CAD/FEA models, equations, and geospatial layers—by leveraging schema alignment and hierarchical retrieval to preserve structure and semantics, thereby reducing data-format silos and long-document errors.

\item \textbf{Tool-Orchestration and Physics-Consistency .} LLMs operate in concert with calculators, optimizers, numerical solvers, and theorem provers, while validators enforce units, boundary conditions, conservation laws, and other invariants. This coupling improves numerical accuracy and physical fidelity in finance, science, and engineering tasks.

\item \textbf{Grounded and Verifiable LLMs.} In this frontier, the generated claims are coupled with recoverable sources through retrieval, structured citations, and domain ontologies, with programmatic attribution checks that expose the evidence behind each step of reasoning to to reduce hallucination and ensure traceability in domains such as law, auditing, and engineering.

\item \textbf{Temporal and Causal Adaptation.} Another promising direction is advancing LLMs beyond static fine-tuning through online learning, drift detection, and integration with econometric and causal inference methods to address non-stationary environments and distinguish correlation from causation.

\item \textbf{Rule-Governed, Reproducible Agent-Based Simulation.} Role-specialized agents, governed by explicit domain rules, simulate policy debates, market microstructure, consumer response, or experimental protocols. Seeds and prompts are logged, and outcomes are validated against historical or experimental data to achieve reproducibility.

\item \textbf{Decision Support with Uncertainty and Domain Controls.} Recommendations are accompanied by calibrated confidence measures and risk thresholds, and deployments encode compliance or physical limits together with auditable trails. These mechanisms enable reliable use of LLMs in high-stakes contexts such as finance, regulation, and engineering safety.

\item \textbf{Cost-effective Domain-specialized Models.} 
Currently, pre-trained foundation models can be extremely cost-expensive. For example, GPT-3 was trained by using a transformer-based architecture with up to 175 billion parameters based on an enormous text corpus with the size of over 45 terabytes, costing an estimated \$12 million for a single training run \cite{brown2020language}. This is usually infeasible and unaffordable for research community and most of the practitioners. The next wave of domain-specialized models should embed domain customization and efficiency in their core architectures, prioritizing intelligent specialization over brute-force scaling.

\item \textbf{Human-in-the-loop Oversight and Governance.}
Pipelines standardize logging of prompts, retrievals, tool calls, and outputs, and adopt role-based approvals, fairness audits, and transparent model/data statements. These practices delineate responsibility, facilitate review, and support external accountability.

\item \textbf{Education, Capacity Building, and Embedded Safety.}
Discipline-aligned tutoring and workflow scaffolds carry users from language to tools, accelerating upskilling across studios, clinics, and labs. In parallel, rigorous red-teaming (i.e., stress-testing for jailbreaks, data leakage, model misuse (including inversion), bias, toxicity, misinformation, and tool-use abuse), dual-use screening, sandboxed execution, and incident-response playbooks will help embed safety-by-design, while preserving legitimate research.

\end{itemize}

As these frontiers (and more) mature, LLMs advance from capable generators to dependable infrastructures—systems that connect evidence to reasoning, adapt to changing conditions, interoperate with tools and standards, and operate within auditable governance. This progression enables end-to-end pipelines—analysis, simulation, and decision support—at disciplinary scale, with reliability sufficient for scholarly, commercial, and safety-critical use.

\textbf{Conclusion.}  In this paper, we survey cutting-edge LLMs across Arts, Letters and Law, Economics and Business, and Science and Engineering. Rather than aiming for exhaustive coverage, we take selected domains as a first step to bridge the humanities and technology, examining how LLMs shape research and practice while outlining key limitations, open challenges, and promising directions. Some claims may be contested, and—as technology, especially AI, evolves rapidly—the landscapes we review will continue to shift. Even so, we hope the insights from this cross-disciplinary review help researchers and practitioners exploit LLMs to advance their works in real-world practice.

\end{CJK*}
\newpage 
\bibliographystyle{unsrt}  
\bibliography{reference_all}
\newpage
\appendix

\end{document}